%% file: meshlets_cvpr20.tex
\documentclass[10pt,twocolumn,letterpaper]{article}

\usepackage{cvpr}

\makeatletter
\@namedef{ver@everyshi.sty}{}
\makeatother
\usepackage{tikz}

\usepackage{times}
\usepackage{epsfig}
\usepackage{graphicx}
\usepackage{amsmath}
\usepackage{amssymb}
\usepackage{dsfont}
\usepackage{color}
\usepackage{comment}
\usepackage{subfig}
\usepackage{wrapfig}
\usepackage{nicefrac}
\usepackage{float}
\usepackage[normalem]{ulem}
\usepackage{animate}
\usepackage{calc}
\usepackage[export]{adjustbox}
\usepackage{enumitem}
\usepackage[font=small]{caption}
\usepackage{xspace}
\usepackage{multirow}
\usepackage{authblk}
\usepackage{algorithm2e}
\usepackage[]{algorithmic}
\usepackage[normalem]{ulem}

\newcommand*\cP{\hspace{-2.5pt}{\color{darkred}{ \tikz[baseline=(char.base)]{\node[shape=circle,draw,inner sep=1pt](char){P};}}}\xspace}
\newcommand*\cN{\hspace{-2.5pt}{\color{darkred}{ \tikz[baseline=(char.base)]{\node[shape=circle,draw,inner sep=1pt](char){N};}}}\xspace}
\newcommand*\cT{\hspace{-2.5pt}{\color{darkred}{ \tikz[baseline=(char.base)]{\node[shape=circle,draw,inner sep=1pt](char){T};}}}\xspace}

\usepackage[pagebackref=true,breaklinks=true,letterpaper=true,colorlinks,bookmarks=false]{hyperref}

\definecolor{darkgreen}{RGB}{30,150,30}
\definecolor{darkblue}{RGB}{0,0,127}
\definecolor{darkred}{RGB}{180,20,20}
\definecolor{darkmagenta}{RGB}{127,0,127}
\definecolor{darkcyan}{RGB}{0,127,127}
\cvprfinalcopy

\begin{document}

\title{{M}eshlet Priors for 3{D} Mesh Reconstruction}
\makeatletter
\renewcommand\Authfont{\fontsize{11.5}{14.4}\selectfont}
\renewcommand\AB@affilsepx{\qquad \protect\Affilfont}
\makeatother
\author[1,2]{Abhishek Badki}
\author[1]{Orazio Gallo}
\author[1]{Jan Kautz}
\author[2]{Pradeep Sen}
\affil[1]{NVIDIA}
\affil[2]{University of California, Santa Barbara}
\renewcommand*{\Authsep} { \ \ \ \ \  }%
\renewcommand*{\Authands}{ \ \ \ \ \  }%

\maketitle

\begin{abstract}
\input{abstract}
\end{abstract} 

\section{Introduction}\label{sec:intro}
\input{intro}

\section{Related Work}\label{sec:related}
\input{related}

\section{Method}\label{sec:method}
\input{overview}
\input{method}

\section{Implementation Details}\label{sec:implementation}
\input{implementation}

\section{Experiments}\label{sec:results}
\input{results}

\section{Discussion and Limitations}\label{sec:discussion}
\input{discussion}

\section{Conclusions}\label{sec:conclusions}
\input{conclusions}

\section*{Acknowledgments}
We would like to thank Kihwan Kim, Aleandro Troccoli, and Ben Eckart for the discussions on comparisons and evaluation.
Arash Vahdat for the discussions and feedback on latent space regularization.
UCSB acknowledges partial support from NSF grant IIS 16-19376 and an NVIDIA fellowship for A.\ Badki. 

{\small
\bibliographystyle{ieee_fullname}
\bibliography{meshlets_cvpr20}
}

\clearpage

\twocolumn[
  \begin{@twocolumnfalse}
{
   \newpage
   \null
   \begin{center}
      {\Large \bf {M}eshlet {P}riors for 3{D} {M}esh {R}econstruction \par ({S}upplementary)}
      {
      \large
      \lineskip .5em
      \begin{tabular}[t]{c}
          
      \end{tabular}
      \par 
      }
      \vskip .5em
      \vspace*{0pt}
   \end{center}
}
  \end{@twocolumnfalse}
]

\setcounter{section}{0}
\setcounter{figure}{0}
\setcounter{table}{0}
\setcounter{footnote}{0}

\section{Additional Experimentation Details}\label{sec:expdetails}
\input{meshlets_cvpr20_supplementary_additional_exp_details}
\

\section{Additional Results}\label{sec:discussion}
\input{meshlets_cvpr20_supplementary_additional_results}

\

\section{Additional Algorithm Details}\label{sec:algodetails}
\input{meshlets_cvpr20_supplementary_additional_algo_details}

\end{document}

%% file: abstract.tex

Estimating a mesh from an unordered set of sparse, noisy 3D points is a challenging problem that requires to carefully select priors.
Existing hand-crafted priors, such as smoothness regularizers, impose an undesirable trade-off between attenuating noise and preserving local detail.
Recent deep-learning approaches produce impressive results by learning priors directly from the data.
However, the priors are learned at the \emph{object level}, which makes these algorithms class-specific and even sensitive to the pose of the object.
We introduce meshlets, small patches of mesh that we use to learn \emph{local} shape priors.
Meshlets act as a dictionary of local features and thus allow to use learned priors to reconstruct object meshes in any pose and from unseen classes, even when the noise is large and the samples sparse.
{\let\thefootnote\relax\footnote{This work was done while A.\ Badki was interning at NVIDIA.}}
{\let\thefootnote\relax\footnote{Code available at \url{https://github.com/NVlabs/meshlets}.}}

%% file: intro.tex

The ability to capture, represent, and digitally manipulate objects is crucial for a wide range of important applications, from content creation to animation, robotics, and virtual reality.
Among the different representations for 3D objects (which also include depth maps, occupancy grids, and point clouds), meshes are particularly appealing.

\begin{figure}
\centering
\includegraphics[width=\columnwidth]{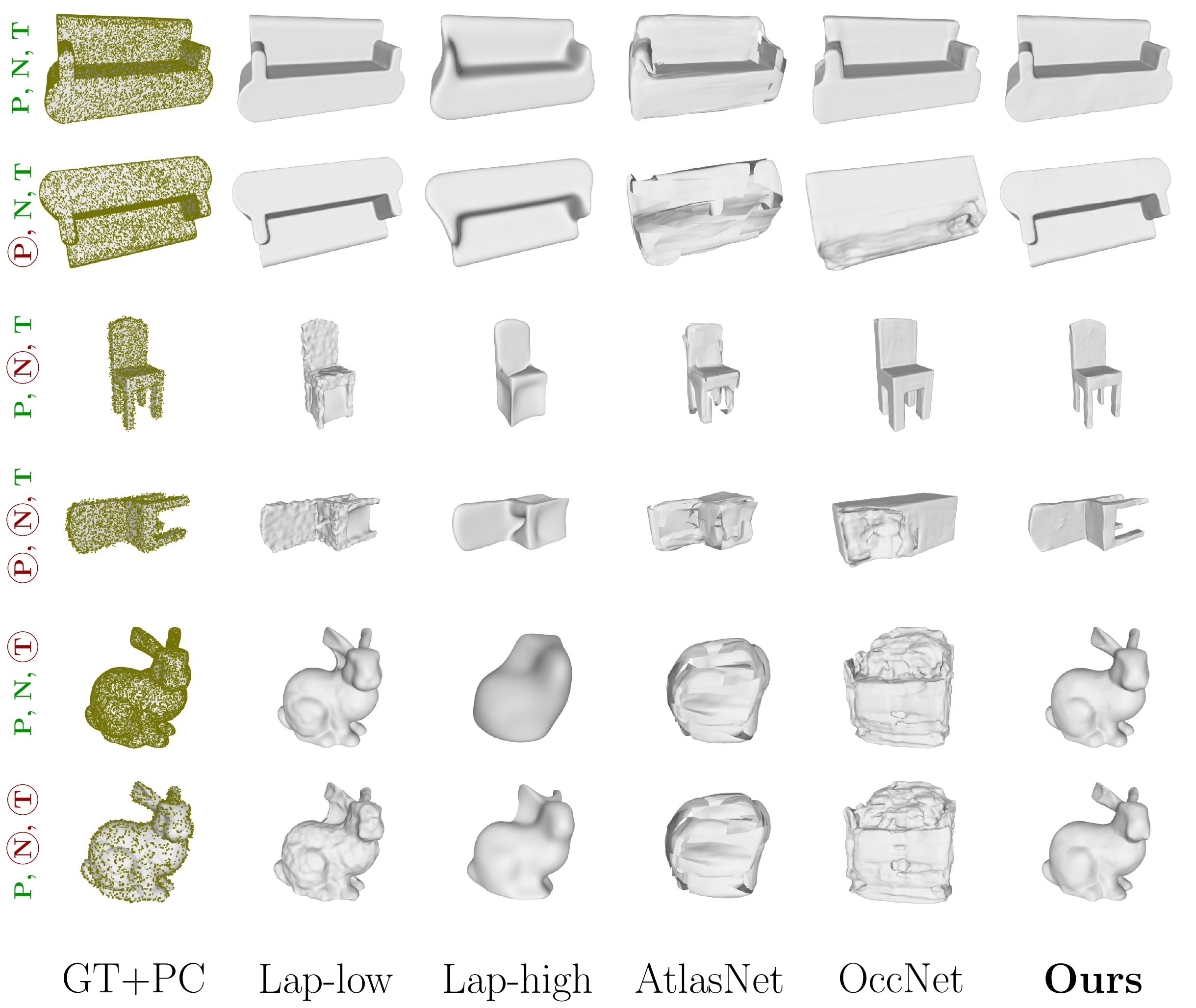}
\caption[]{Mesh reconstructions for objects in pose ({\color{darkgreen}P}) and out of pose (\cP), under low ({\color{darkgreen}N}) and moderate (\cN) noise, and for objects in ({\color{darkgreen}T}) and out (\cT) of training. GT+PC ground truth mesh and the point cloud used by the various methods to estimate the mesh. Traditional methods introduce a noise vs.\ smoothness trade-off (Laplacian low/high~\cite{laplacian}). State-of-the-art, deep-learning methods (AtlasNet~\cite{Groueix_2018_CVPR} and OccNet~\cite{occNet}) learn object-level priors, which causes them to fail on objects not seen in training (\cT), or even on objects that are \emph{just rotated} w.r.t. the training set (\cP). Our method learns local priors and forces global consistency with the point cloud.}\label{fig:teaser}
\end{figure}

Estimating meshes of real-world objects, however, is not straightforward since
common capture strategies, such as structured light~\cite{ScharsteinStructuredLight} or multi-view stereo~\cite{huang2018deepmvs,yao2018mvsnet}, produce point clouds or depth maps instead.
These intermediate representations are noisy, sparse, and, when used to estimate a continuous surface, they introduce a trade-off between over-fitting to the noise and over-smoothing.
Traditional methods require hand-crafted priors (\eg, local smoothness) to balance noise and details, as is the case for Laplacian reconstruction~\cite{laplacian}.
Figure~\ref{fig:teaser} shows that this balance is difficult to strike when the point cloud is noisy (rows marked as \cN).
Recent learning-based methods learn priors directly from a large number of examples~\cite{kato2018renderer,wang2018pixel2mesh,cmrKanazawa18,occNet,Park_2019_CVPR}.
Because of their ability to learn priors directly from the data, these approaches can produce impressive results from both point clouds and single images.
However, they learn priors \emph{at the object level}, which limits their ability to reconstruct objects from classes not seen during training (rows marked as \cT).
They also struggle to disentangle the shape priors with the object pose:
state-of-the-art learning methods can fail completely on a rotated point cloud (rows marked as \cP) even though they can reconstruct the  same point cloud when its pose resembles that of the training set (rows marked as {\color{darkgreen}P}).
Tatarchenko~\etal suggest that many of these methods may actually learn a form of classification and nearest-neighbor retrieval from the dataset, rather than a proper 3D reconstruction~\cite{tatarchenko2019single}.
In fact, even for rows marked with {\color{darkgreen}P} and {\color{darkgreen}T} in Figure~\ref{fig:teaser}, a closer look seems to indicate that AtlasNet~\cite{Groueix_2018_CVPR} and OccNet~\cite{occNet} are reconstructing a different couch and chair from the training set.

We present a learning-based method to extract a 3D mesh from a set of sparse, noisy, unordered points that bridges traditional and learning-based approaches.
Our key intuition is to learn geometric priors \emph{locally} while enforcing their consistency \emph{globally}. 
To represent shape priors, we introduce \emph{meshlets}, small patches of mesh that, loosely speaking, serve as a learned dictionary of local features.
Specifically, we use a variational auto-encoder (VAE)~\cite{kingma2013auto} to learn the latent space of meshlets that can be observed in natural shapes.
We call these ``natural meshlets''.
Learning these local features offers two key advantages.
First, it allows to reconstruct objects from classes never seen in training: at some scale, a couch exhibits similar local features to those of a bunny.
Second, it disentangles the global pose of the object and the parametrization learned by the network, which allows our algorithm to be robust to dramatic changes of the object's pose, as shown in Figure~\ref{fig:teaser}.

To fit the meshlets to a point cloud, we minimize their distance to the points, while enforcing that they belong to the latent space of natural meshlets.
Therefore, the resulting surface will locally satisfy the priors we learned.
However, because the meshlets are optimized independently of each other, the mesh extracted from their union will not be watertight.
Therefore, we define an auxiliary, watertight mesh and propose to use it in an alternating optimization that ensures that the meshlets are consistent with each other and with the observed point cloud.

We show with extensive comparisons that this iterative method produces results that outperform the state-of-the-art on challenging scenarios such as noisy points clouds in arbitrary poses (Figure~\ref{fig:teaser}). 
In summary, our contributions are as follows:
\begin{itemize}
\item We present meshlets, a new way of representing local shape priors in the latent space of a variational autoencoder that is trained on local patches from a dataset of real-world objects.
\item We propose an alternating optimization which fits meshlets to the measured point samples (enforcing local constraints) while maintaining global consistency for the mesh.
\item We demonstrate for the first time, to our knowledge, successful reconstructions of 3D meshes from very sparse, noisy point measurements with a category-agnostic, learning-based method.
\end{itemize}

%% file: related.tex

Extracting a mesh from a point cloud is an important problem that has been the focus of much research since the early days of graphics.
Traditional methods such as marching cubes~\cite{marchingcubes} or Ball-Pivoting~\cite{bernardini1999ball} work well for cases where the noise is small as compared to the density of the point cloud.

In general, however, noise does raise issues.
One traditional solution, then, is to use the points and their normals to compute a signed-distance function whose zero crossing is the desired surface~\cite{tsdf,hoppe1992surface,bajaj1995automatic,poisson,screenedpoisson}.
An alternative is to use hand-crafted priors, such as smoothness of the vertices and normals of the estimated mesh~\cite{laplacian}.
However, these priors introduce a trade-off between suppressing the noise and preserving sharp features that becomes increasingly brittle for sparser and noisier point clouds (Figure~\ref{fig:teaser}).

Priors can be more effectively learned from data with neural networks.
Deep learning methods, for instance, have shown great success in estimating depth maps from images, whether from multiple views~\cite{huang2018deepmvs,yao2018mvsnet}, stereo~\cite{kendall2017end}, or even a single-image~\cite{Eigen:2014:DMP:2969033.2969091,monodepth17,zhou2017unsupervised,lasinger2019towards}.
Even meshes can be directly extracted from a single image, provided that the class of the object is known~\cite{kato2018renderer,wang2018pixel2mesh,cmrKanazawa18}.

Rather than requiring to manually tinker with the traditional noise/sharpness trade-off, methods that learn priors to extract meshes from point clouds introduce a new one:
generally speaking, the lower the quality of the observations (\eg, strong noise or sparsity of the point cloud), the stronger the priors need to be, thus affecting the algorithm's ability to generalize to different and unseen classes.
For instance, methods that learn local priors are class-agnostic but tend to need dense point clouds with low levels of noise~\cite{pcpnet,Yu_2018_CVPR,Yu_2018_ECCV}.
Similar to our approach, Williams~\etal~\cite{Williams2019} propose to use a local 3D patch-based representation.
However, their approach uses a deep neural network as a geometric prior and does not use any data for training. 
This makes it difficult to leverage good priors in presence of high noise and sparsity.
The recent works or Park~\etal~\cite{Park_2019_CVPR}, Groueix~\etal~\cite{Groueix_2018_CVPR}, and Mesheder~\etal~\cite{occNet} produce impressive results even with sparser and potentially noisier data, but fail to generalize to completely new classes of objects.
They even struggle when the point cloud is in a pose that differs significantly from the training pose, as show in Figure~\ref{fig:teaser}, rows marked with \cP.
This issue is due, in part, to the fact that these methods lack a mechanism to enforce geometric constraints at inference time.
Our method is class-agnostic thanks to its ability to learn and enforce local priors while minimizing the error with respect to the point cloud at inference time.

The idea of learning priors from data and enforcing geometric constraints at inference time was recently explored for depth map~\cite{Bloesch_2018_CVPR}, point cloud~\cite{zhu2017object} and surface estimation~\cite{Litany2018, lin2019photometric, Park_2019_CVPR}. 
These approaches use low dimensional representations that allow inference time optimization. 
However, approaches that learn priors at the object level tend to be category specific. 
Our meshlet priors directly encode the (local) shape of the surface instead of a viewer centric depth~\cite{Bloesch_2018_CVPR}. 
Meshlets are class agnostic and can be used to learn and enforce priors at different scales. 

Key to solving the mesh estimation problem is how to represent it. 
Different representations for meshes exist that are amenable to use with neural networks, but they tend to also be class specific~\cite{Sinha_2017_CVPR,Ben-Hamu}.
One key ingredient of our method is the use of small mesh patches, called meshlets, which simplify the processing of the mesh, among other things.
A related approach is the work of Groueix~\etal who also represent the mesh as a collection of large parts, which they call charts~\cite{Groueix_2018_CVPR}.
However, their method does not offer a mechanism to enforce global consistency and does not leverage local shape priors.

%% file: overview.tex

Our goal is to estimate a mesh from a set of unordered, non-oriented points.
The task is easy when the point cloud is dense and the noise is low.
However, when the quality of the observations degrades, \eg, sparser or noisier points, the choice of priors and heuristics becomes central.
Hand-crafted priors, such as smoothness, introduce a trade-off between overly smooth and noisy reconstructions, as shown in Figure~\ref{fig:teaser} (Lap-low/high).
On the other hand, neural networks can learn priors directly from data, but they introduce other challenges.
First, capturing the distribution of generic objects requires training on a large number of examples, possibly larger than what existing datasets can supply.
Moreover, generalization can be an issue:
the performance of existing learning-based methods quickly degrades when the test objects differ from the ones used in training, as shown in the rows marked as \cT in Figure~\ref{fig:teaser}.
Finally, it is not straightforward to disentangle object-level priors and the pose of the object.
Figure~\ref{fig:teaser} shows that OccNet~\cite{occNet} and AtlasNet~\cite{Groueix_2018_CVPR}, both recent state-of-the-art works, fail for classes never seen in training \cT, or even when the pose of the object is significantly different from the poses seen in training \cP.

To overcome these issues, we propose to learn priors \emph{locally}: even if the Stanford Bunny in Figure~\ref{fig:teaser} was never seen in training, its local features are similar to those found in more common objects from the training set.
We introduce \emph{meshlets}, which can be regarded as small patches of a mesh, as shown in Figure~\ref{fig:meshlet_interp}.
Loosely speaking, meshlets act as a dictionary of basic shape features.
Meshlets are local and of limited size, and thus offer a simple mechanism to disentangle the (local) priors from the object's pose.
If meshlets are adapted to the point cloud independently of each other, however, they may not result in a watertight surface.
Therefore, we explicitly enforce their consistency \emph{globally}.
In the following we describe these two stages, and the overall process to extract a mesh.

%% file: method.tex

\begin{figure}
\input{figures/meshlet_interp/meshlet_interp_v0}
\caption{Smoothness of the latent space. We can progressively deform one mesh onto another by interpolating between the corresponding points in latent space (see Section~\ref{sec:meshlets}).}\label{fig:meshlet_interp}
\end{figure}
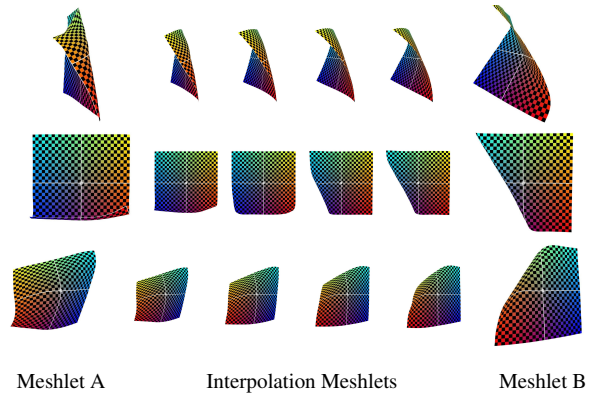

\subsection{Local Shape Priors with Meshlets}\label{sec:meshlets}

In this section we introduce meshlets, and describe how we leverage them to enforce local shape priors.
Intuitively, a meshlet $m$ is a small patch of mesh deformed to adhere to a region of another, larger mesh, as shown in Figure~\ref{fig:meshlet_interp}~and~\ref{fig:union}(a).
To extract meshlet $m$ at vertex $v$ of mesh $\mathcal{M}$, we first compute a local geodesic parametrization~\cite{melvaer2012geodesic} that maps the 3D coordinates of the vertices in a neighborhood of $v$ to coordinates on $\pi_T$, the plane tangent to $\mathcal{M}$ at $v$.
We then  re-sample the geodesic distance function $\pi_T$ at integer coordinates $(\mu,\nu)$.
This gives us the correspondence between a vertex on the meshlet at $(\mu,\nu)$, and a vertex on the mesh in the neighborhood of $v$.

Because they only require a local parametrization computed with respect to the center vertex $v$, meshlets work well even for objects with large, varying curvature.
Because they are local, they can learn shape priors that are independent of the pose and class of the object.

\paragraph{Learning local shape priors with meshlets.}
We want to learn the distribution of ``natural meshlets,'' \ie, those meshlets that capture the local features of real-world objects.
Inspired by recent methods~\cite{zhu2017object,Bloesch_2018_CVPR}, we use a variational auto-encoder (VAE).
By training the VAE to reconstruct a large number of meshlets, we force its bottleneck to learn the latent space of natural meshlets.
Differently put, vectors sampled on this manifold and fed into the decoder result in natural meshlets.
We extract meshlets from objects from the ShapeNet dataset~\cite{shapenet2015} and we feed their 3D coordinates for training.
However, we first translate and rotate the meshlets to bring them into a canonical pose.
This transformation is necessary to make sure that similar meshlets sampled from different 3D locations and orientations map to similar regions in the VAE's latent space.
More specifically, given a meshlet $m_i$, we first translate and rotate it so that its center $c_i$ is at the origin, and the normal at $c_i$ is aligned with the z-axis, then we rotate it around the $z$ axis so that the local $(\mu,\nu)$ coordinates of the meshlet are aligned with the $x$ and $y$ axes.
We call this the canonical pose.
A meshlet, then, is completely defined by $P_i$, the transformation from global to canonical pose, and $l_i$, the latent vector corresponding to the meshlet in canonical pose:
\begin{equation}
    m_i = \{l_i, P_i\}.
\end{equation}
Section~\ref{sec:train_details} details the network's architecture.
Since we disentangle pose and shape, smoothly traversing the latent space will smoothly vary the shape of the reconstructed meshlet as shown in Figure~\ref{fig:meshlet_interp}, where we take the latent vectors $l_A$ and $l_B$ corresponding to meshlets A and B, and we progressively interpolate between them to get vectors $l_I$'s.
The meshlets reconstructed from the $l_I$'s smoothly interpolate between the shape of meshlets A and B.

\begin{figure}[t]
\centering
\includegraphics[width=\linewidth]{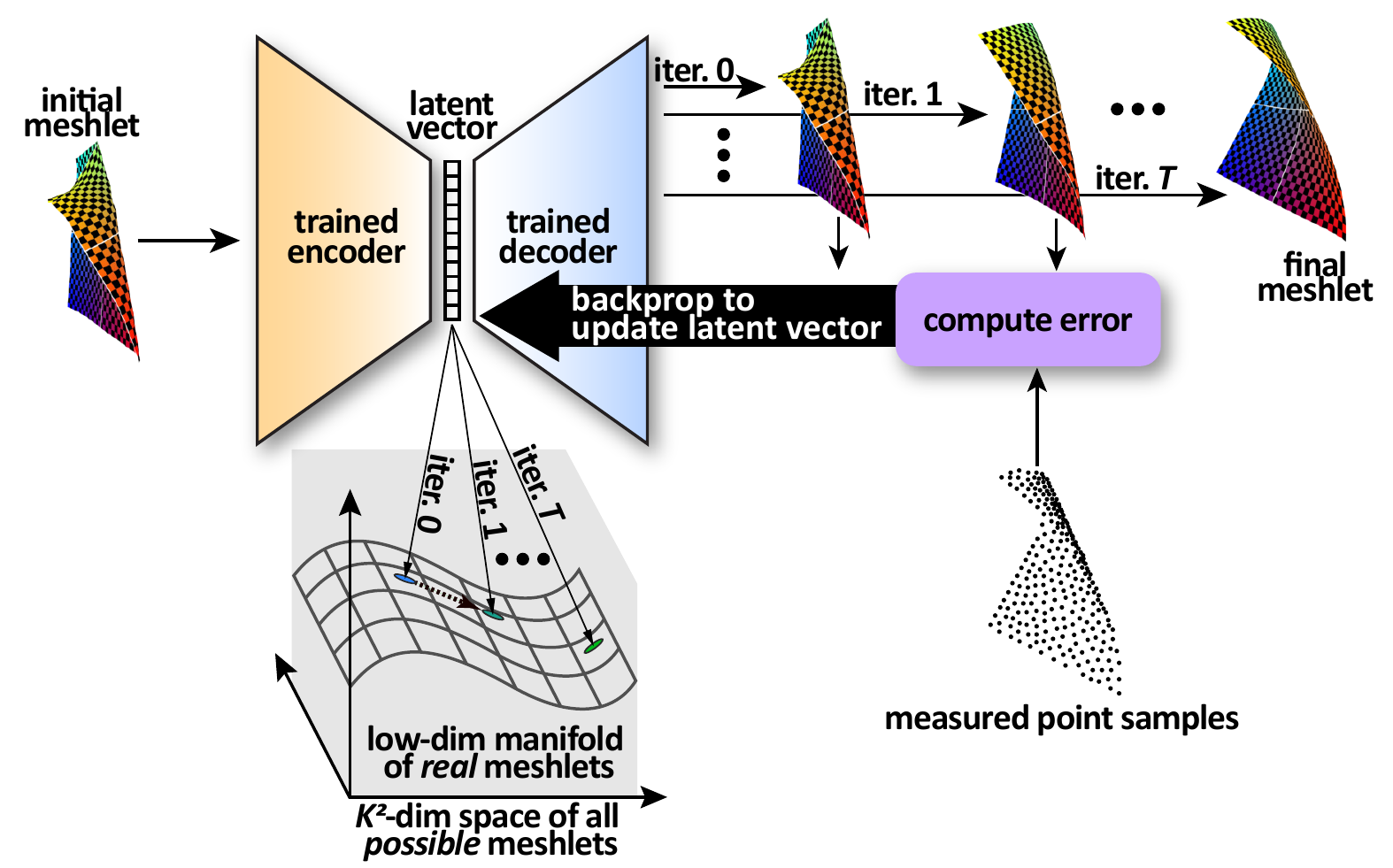}
\caption{Optimization of our meshlets using the learned latent-space as a prior. By backpropagating the error with respect to the measured points and using it to update the meshlet's latent vector, we are effectively moving along the low-dimensional manifold of real meshlets while fitting the points.}\label{fig:meshlet_optimization}
\vspace{1mm}
\end{figure}

\paragraph{Fitting a meshlet to 3D points.}
Assume now that we are given a set of 3D points roughly corresponding to the size of a meshlet (we will generalize this to a complete point cloud in Section~\ref{sec:overall}).
Deforming a natural meshlet to fit it is now straightforward: we simply traverse the latent space learned by the VAE to minimize the distance between the meshlet and the points.
Specifically, we take $m_i(t_0)$, an initialization of the meshlet, and run it through the encoder to find the corresponding latent vector $l_i(t_0)$.
This is the starting point of our optimization.
With the weights of the VAE frozen, we compute the error between the meshlet and the points, and take a gradient descent step through the decoder.
This brings us to a new point in latent space, $l_i(t_1)$, and the corresponding meshlet $m_i(t_1)$.
Meshlet $m_i(t_1)$ is a natural meshlet that is closer to the given 3D points. 
We iterate until convergence (Figure~\ref{fig:meshlet_optimization}).
We note that, although other approaches have also proposed to optimize the latent vector of a VAE to match some measured samples (\eg,~\cite{Park_2019_CVPR,Bloesch_2018_CVPR,Bagautdinov2018,Litany2018}), they do so at the object (or scene) level.
Because our method learns local surface patches, and therefore reuse surface priors across different object categories, it can better generalize.

\subsection{Overall Optimization}\label{sec:overall}
\begin{figure}[t]
\centering
\includegraphics[width=\linewidth]{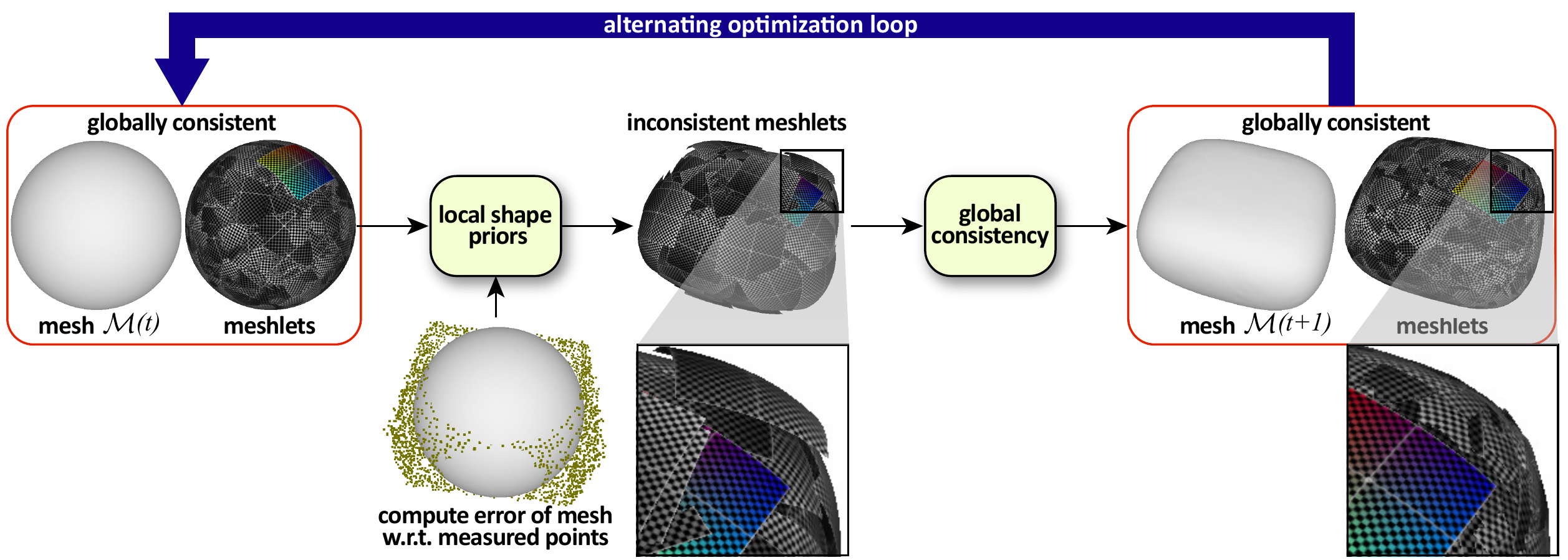}
\vspace{-0.44in}
\begin{small}
\begin{tabbing}
   ddddddddddddddddddddddddddddd \=  dddddddddddd \= \kill
   \> \hspace{-0.04in}(a) \> \hspace{0.04in}(b)   \\
\end{tabbing}
\end{small}
\vspace{-0.3in}
\caption{Alternating optimization used in our algorithm. The first stage updates the meshlets based on the errors between the underlying mesh and the measured point cloud. However, because meshlets are localized representations, optimizing them individually causes inconsistencies across the object. Hence, in the second stage we enforce global consistency across all meshlets to reconstruct an updated version of the mesh which is used in the next iteration of the algorithm.}\label{fig:optimization_loop}
\end{figure}

\begin{figure}[t]
\centering
\includegraphics[width=\linewidth]{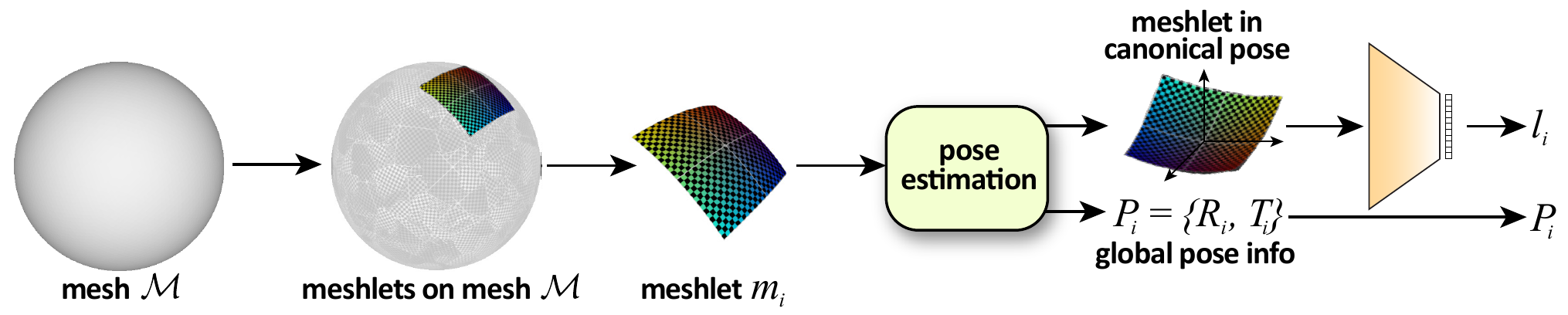}
\caption{Encoding of meshlets on a given mesh, $\mathcal{M}$. Each meshlet is represented by a latent vector $l_i$ in the low-dimensional manifold as well as its global pose $P_i$ (composed of rotation $R_i$ and translation $T_i$ between the canonical space and global coordinates).}\label{fig:meshlet_encoding}
\end{figure}

Having explained how our meshlets can be used to learn local priors, and can be fit to a set of 3D points, we can describe the overall algorithm, which is fairly straightforward at its core.
Throughout our optimization, we use an auxiliary mesh $\mathcal{M}$, which we initialize to a rough approximation of the complete mesh, $\mathcal{M}(t_0)$.
This could be a sphere, or any other surface that satisfies our meshlet priors, \ie, meshlets extracted from $\mathcal{M}(t_0)$ lie on the manifold learned by the VAE.
From $\mathcal{M}(t_0)$ we extract $N$ overlapping meshlets $m_i(t_0)$'s and find the corresponding $l_i(t_0)$'s and $P_i(t_0)$'s (Figure~\ref{fig:meshlet_encoding}).
We select $N$ so that each vertex on $\mathcal{M}(t_0)$ is covered by at least $3$ meshlets.
Generally, this results in $500$ to $1{,}500$ meshlets.   
We also find the distance between $\mathcal{M}(t_0)$ and the point cloud, whose gradient we can propagate to the meshlets since we have the correspondences between mesh and meshlets by construction.
This allows us to update the meshlets to adapt to the points points (Section~\ref{sec:meshlets}).
However, this optimization is performed on each meshlet independently, so it results in small gaps between the meshlets (Figure~\ref{fig:optimization_loop}(a)).
Therefore, we enforce and maintain global consistency by adding a step in which we deform $\mathcal{M}$ to match the meshlets and update the meshlets to match $\mathcal{M}$.
Deforming $\mathcal{M}$ brings it closer to the point cloud, deforming the meshlets forces them to be globally consistent.
Finally we iterate through the following two steps:
\begin{enumerate}[leftmargin=*] 
    \item Optimize meshlets to fit the point cloud (\ref{sec:local_priors}).
    \item Optimize meshlets and mesh to match each other (\ref{sec:global_consistency}).
\end{enumerate}
At convergence, the auxiliary variable $\mathcal{M}$, watertight by construction, is our estimation of the mesh.
We now explain the two steps in detail.

\subsubsection{Enforcing Local Shape Priors}\label{sec:local_priors}
To optimize the $N$ meshlets $\{m_i(t_0)\}_{i=1:N}$ with respect to the point cloud we need to define an error.
Unfortunately, the correspondences between the point cloud and the vertices of the meshlets are not readily available.
A Chamfer distance, then, is not straightforward to use because without correspondences all the points in the point cloud would contribute to the error of all the meshlets---even if they are on opposite sides of the object.
However, we do have the correspondences between the vertices of $\mathcal{M}(t_0)$ and the meshlets.
Therefore, we compute the Chamfer distance between the point cloud and the mesh instead:
\begin{equation}\label{eq:PC}
C^{\text{PC}} = \sum_{v_j \in \mathcal{M}(t_0)} \min_{p \in \text{PC}} ||v_j-p||_2^2
+ \sum_{p \in \text{PC}} \min_{v_j\in \mathcal{M}(t_0)} ||v_j-p||_2^2,
\end{equation}
where $p$ is a 3D point in the input point cloud, PC.
Equation~\ref{eq:PC} gives us per-vertex error on the mesh, which we can propagate to the corresponding meshlets.
We then update the meshlets to minimize $C^{\text{PC}}$ as explained in Section~\ref{sec:meshlets}, and get a new set of natural meshlets $\{m_i(t_1)\}_{i=1:N}$.

\subsubsection{Enforcing Global Consistency}\label{sec:global_consistency}
To enforce that meshlets $\{m_i(t_1)\}_{i=1:N}$ are globally consistent, \ie, that their union is a watertight mesh, we use, once again, $\mathcal{M}$.
Specifically, we compute the Chamfer distance between the vertices of $\mathcal{M}$ and the vertices of \emph{all} the meshlets as
\begin{eqnarray}\label{eq:Cm}
C^m &=& \sum_{v_j \in \mathcal{M}} \min_{v_k \in \{m_i\}_{i=1:N}} ||v_j-v_k||_2^2\nonumber\\
&+& \sum_{v_k \in \{m_i\}_{i=1:N}} \min_{v_j\in \mathcal{M}} ||v_j-v_k||_2^2.
\end{eqnarray}
First we keep the meshlets fixed and deform $\mathcal{M}(t_0)$ to minimize $C^m$.
Then we fix the resulting mesh $\mathcal{M}(t_1)$ and adjust the meshlets with the algorithm described in Section~\ref{sec:meshlets}, but this time to minimize $C^m$.
We iterate until Equation~\ref{eq:Cm} is minimized.
At this point the meshlets will be consistent with the mesh and, in turn, \emph{globally}.
This process corresponds to the block ``global consistency'' in Figure~\ref{fig:optimization_loop}.

%% file: figures/meshlet_interp/meshlet_interp_v0.tex

\newcommand{\widthMul}{0.19}

\newcommand{\interpScaling}{0.65}

\newlength{\meshletWidth}
\newlength{\meshletHeight}
\settowidth{\meshletWidth}{\includegraphics[width=\widthMul\columnwidth]{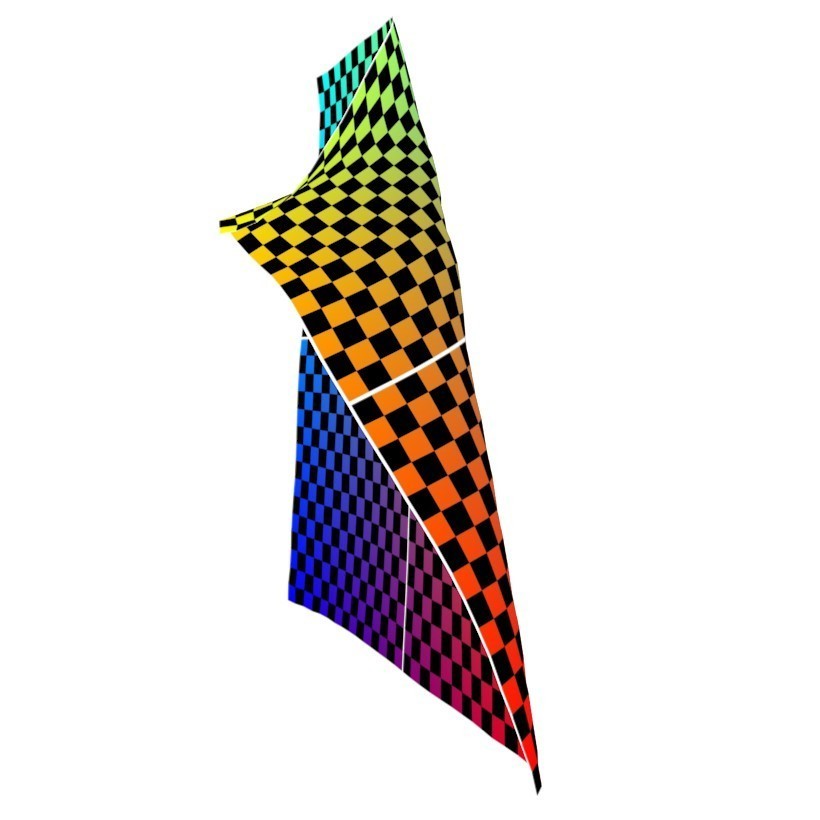}}
\settoheight{\meshletHeight}{\includegraphics[width=\widthMul\columnwidth]{figures/meshlet_interp/000000/m_000.jpg}}

\newlength{\raiseupby}
\setlength{\raiseupby}{(\meshletHeight-\interpScaling\meshletHeight)*\real{0.5}}

\captionsetup[subfigure]{labelformat=empty}
\centering
\subfloat{\includegraphics[width=\meshletWidth]{figures/meshlet_interp/000000/m_000}}
\raisebox{\raiseupby}{
\subfloat{\includegraphics[width=\interpScaling\meshletWidth]{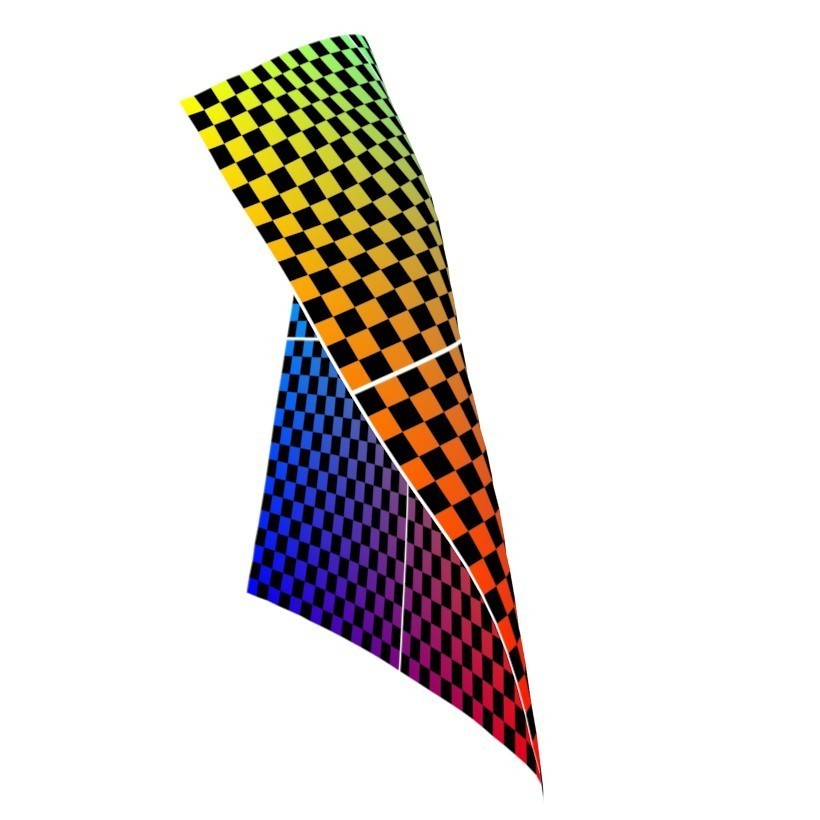}}
\subfloat{\includegraphics[width=\interpScaling\meshletWidth]{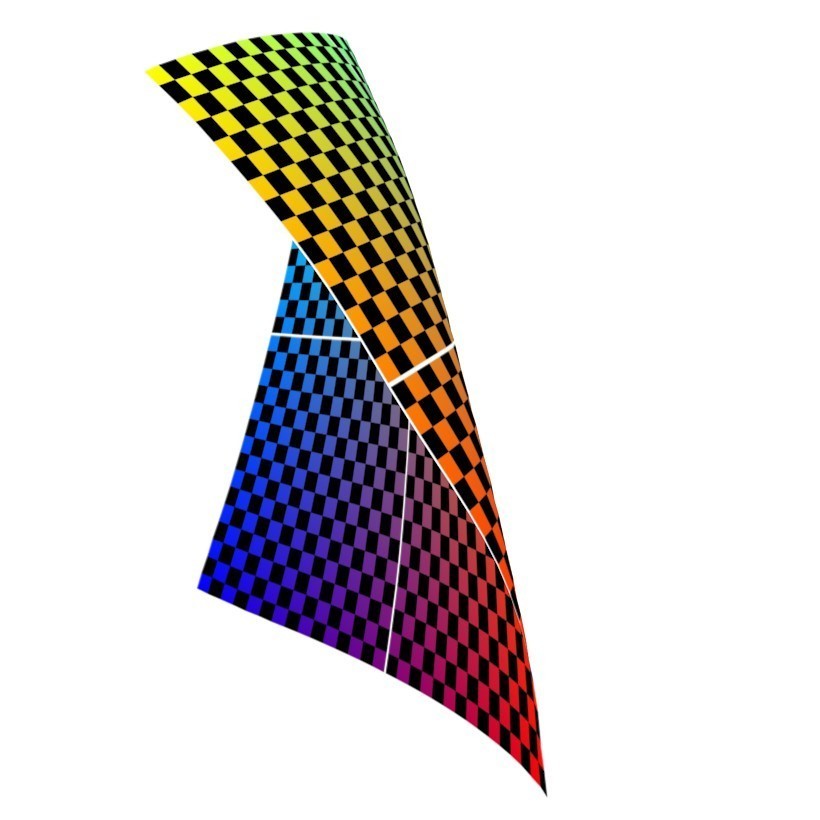}}
\subfloat{\includegraphics[width=\interpScaling\meshletWidth]{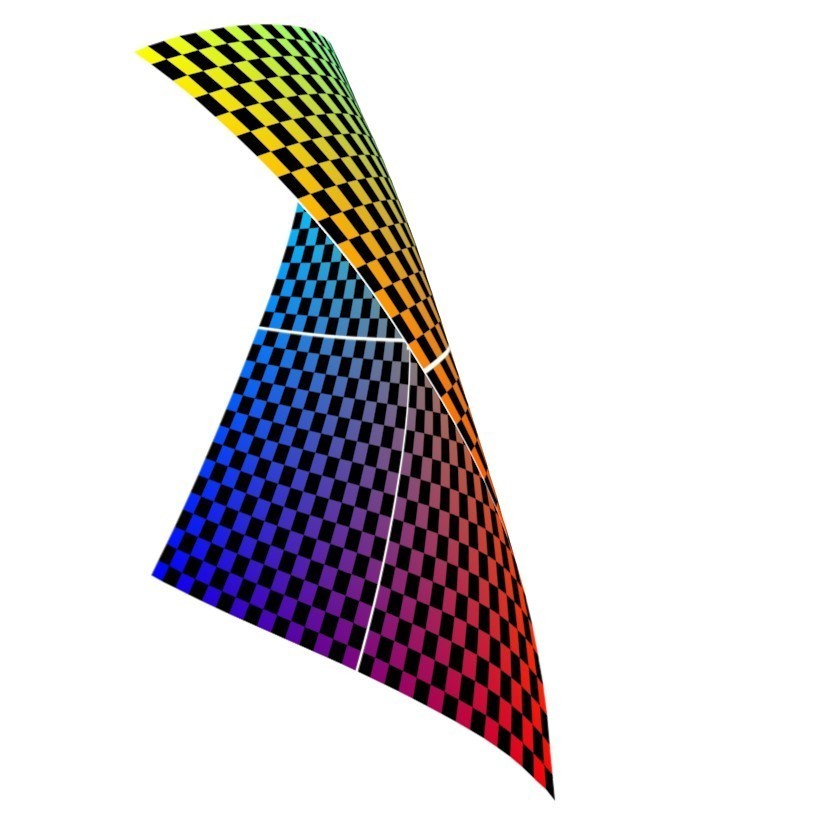}}
\subfloat{\includegraphics[width=\interpScaling\meshletWidth]{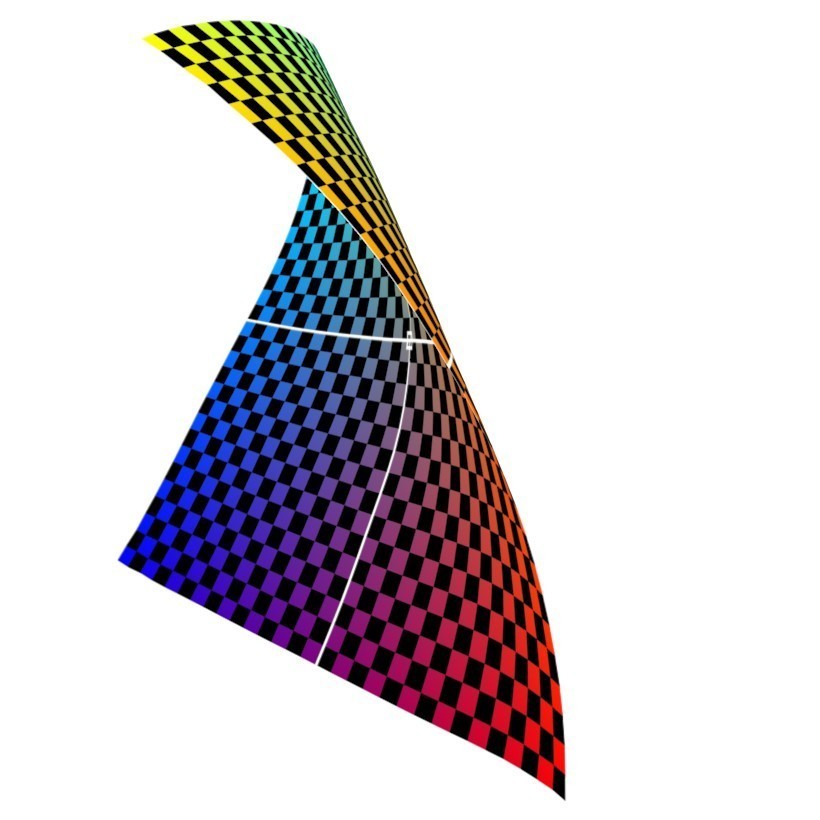}}
}
\subfloat{\includegraphics[width=\meshletWidth]{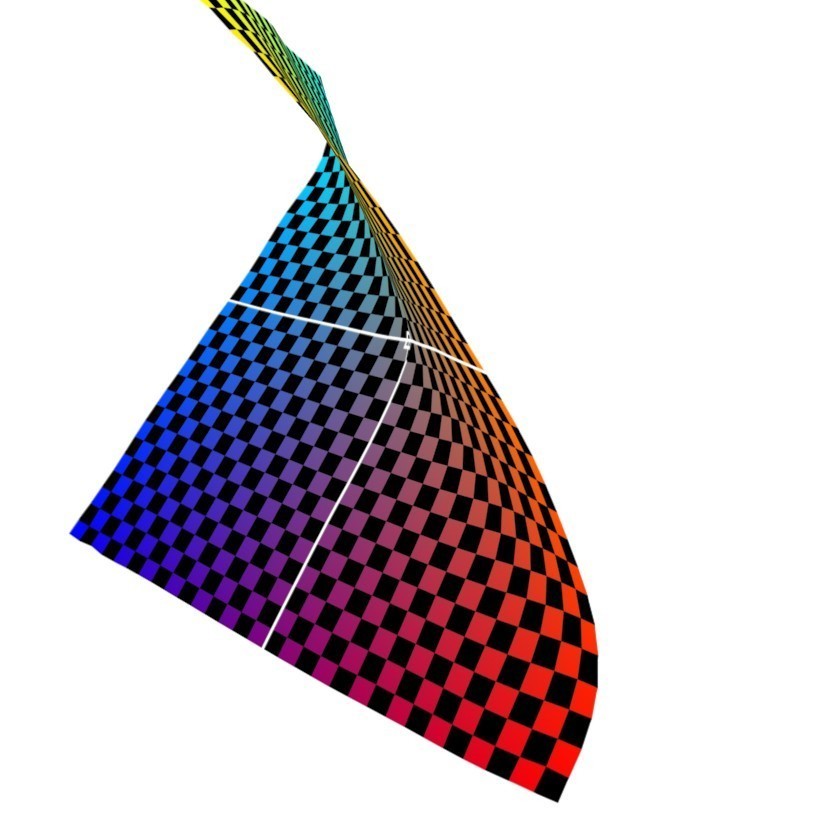}}\\
\vspace{-4mm}
\subfloat{\includegraphics[width=\meshletWidth]{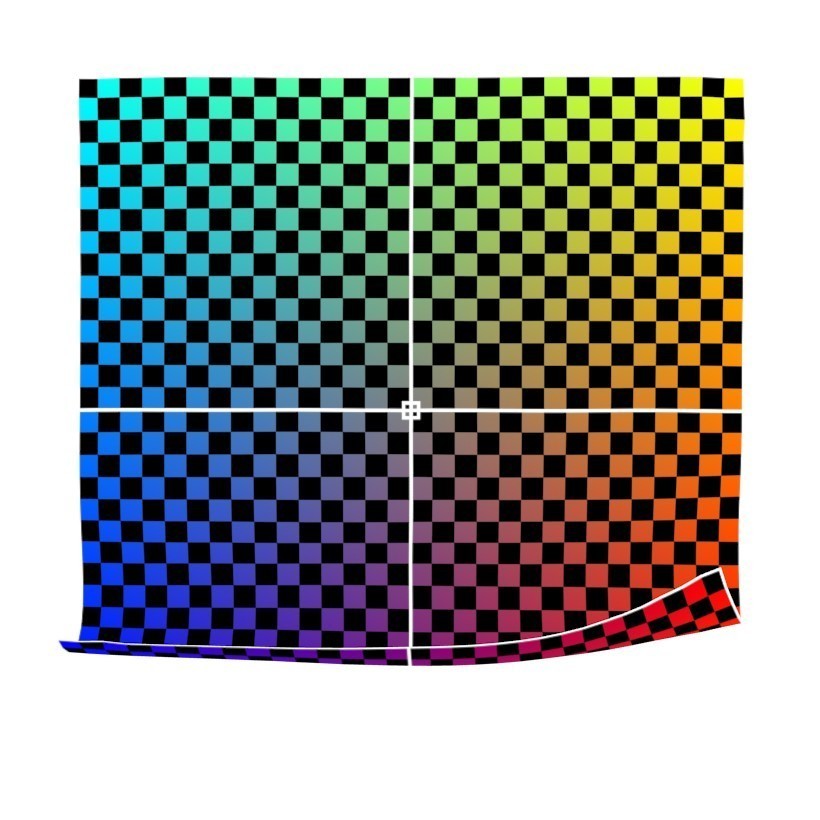}}
\raisebox{\raiseupby}{
\subfloat{\includegraphics[width=\interpScaling\meshletWidth]{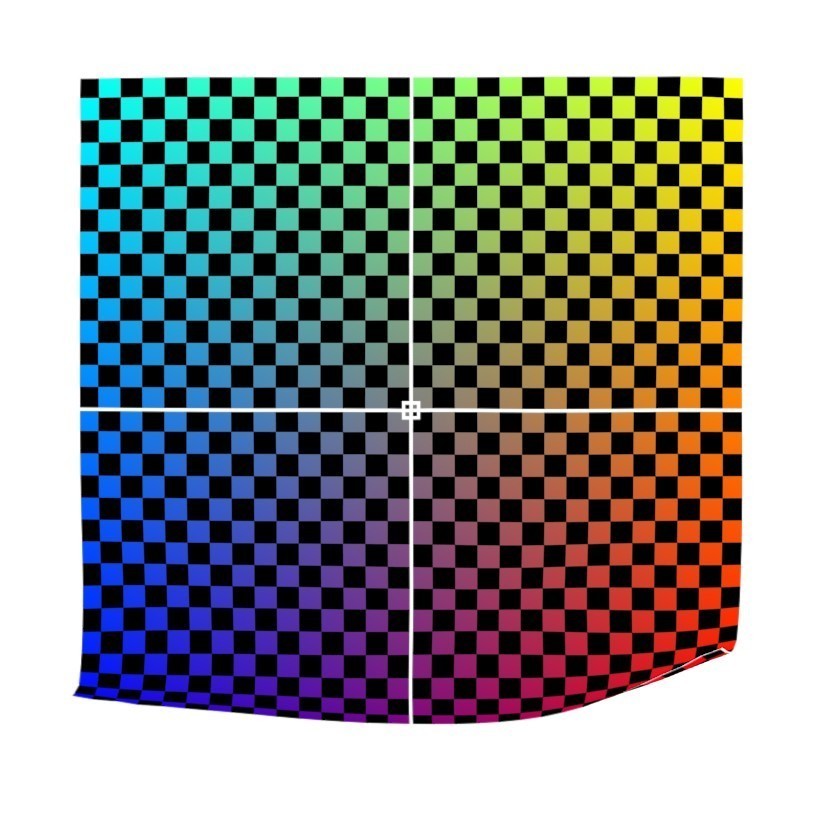}}
\subfloat{\includegraphics[width=\interpScaling\meshletWidth]{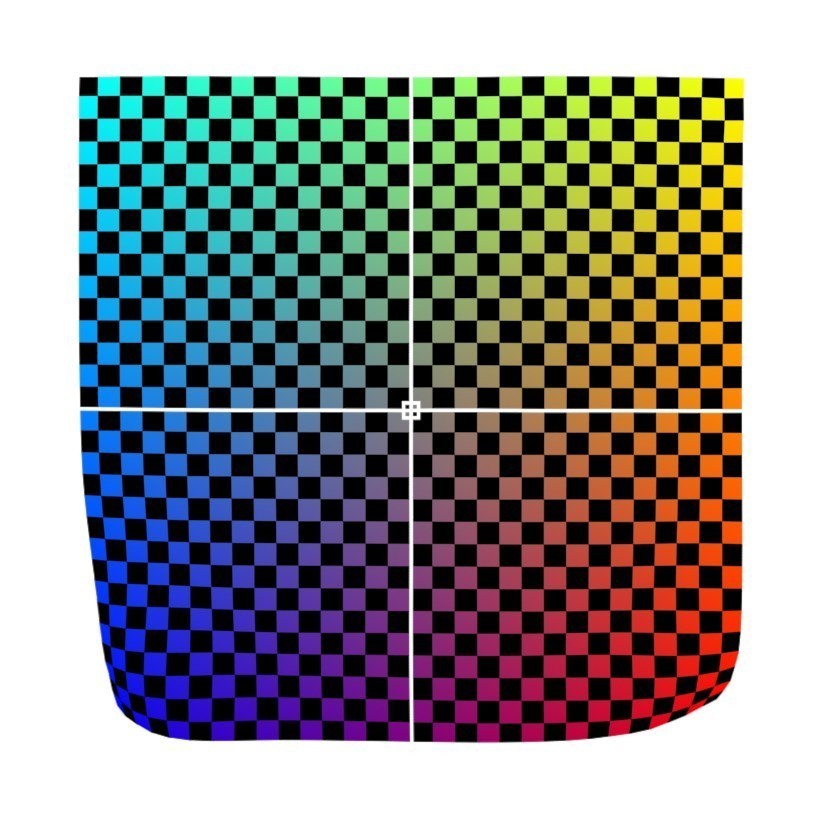}}
\subfloat{\includegraphics[width=\interpScaling\meshletWidth]{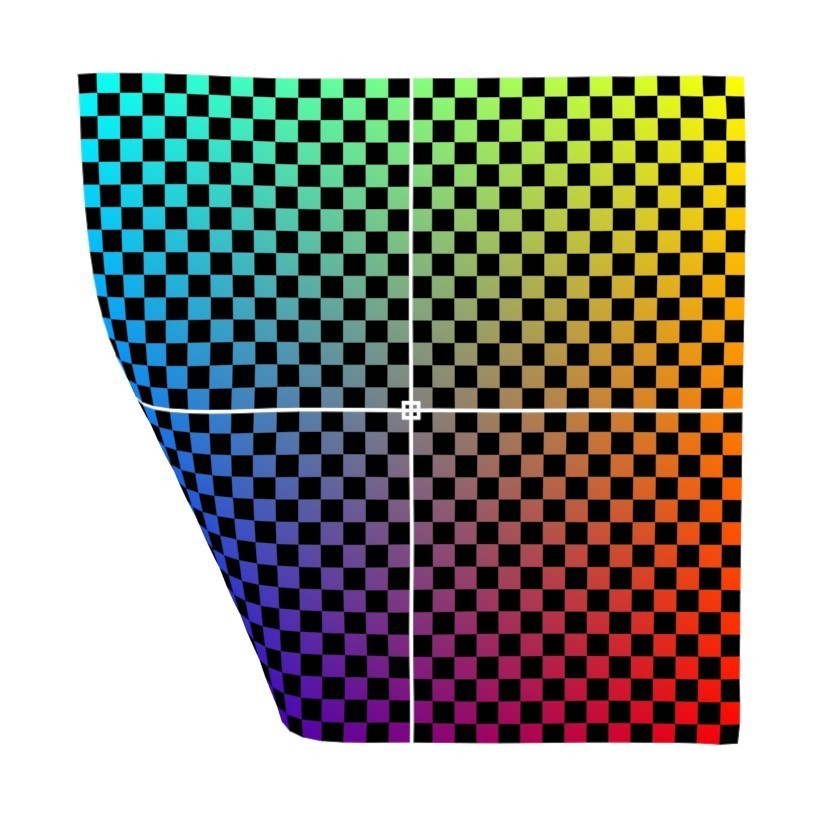}}
\subfloat{\includegraphics[width=\interpScaling\meshletWidth]{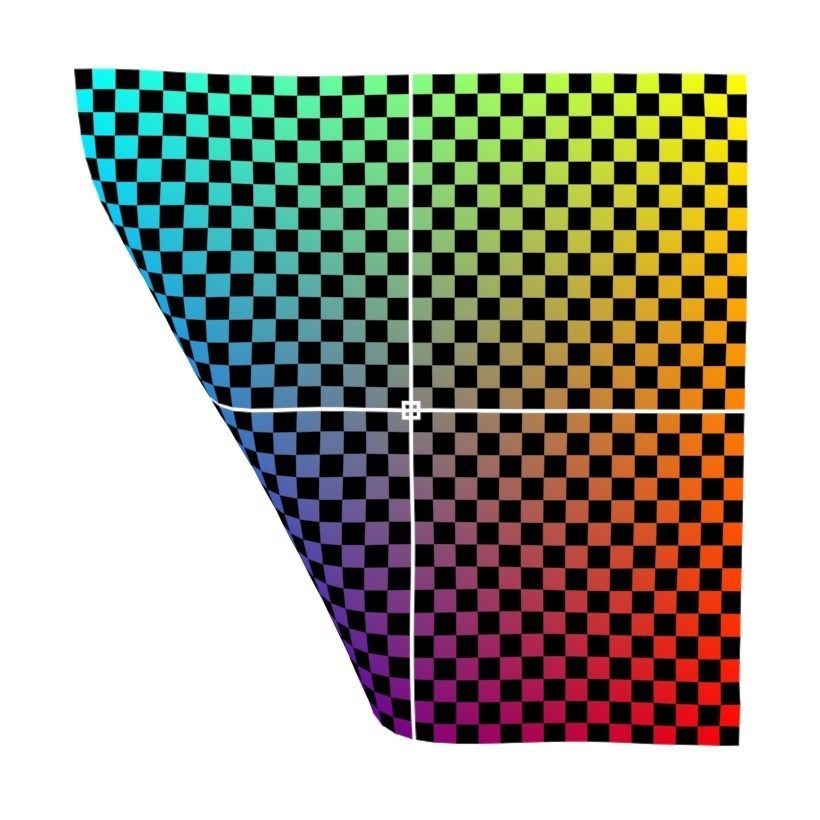}}
}
\subfloat{\includegraphics[width=\meshletWidth]{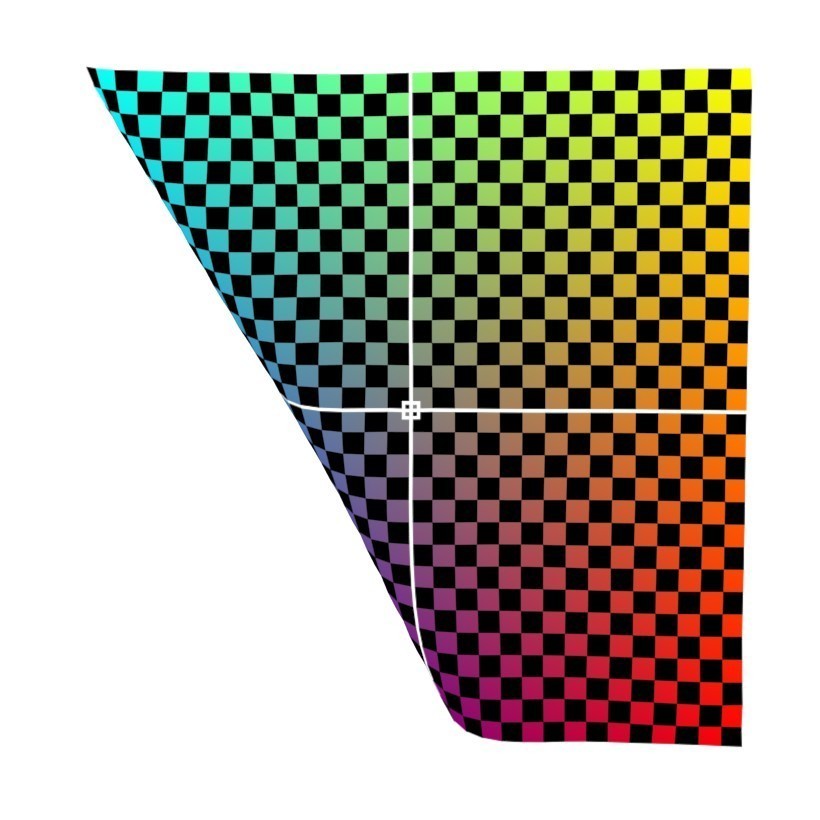}}\\
\vspace{-4mm}
\subfloat[Meshlet A]{\includegraphics[width=\meshletWidth]{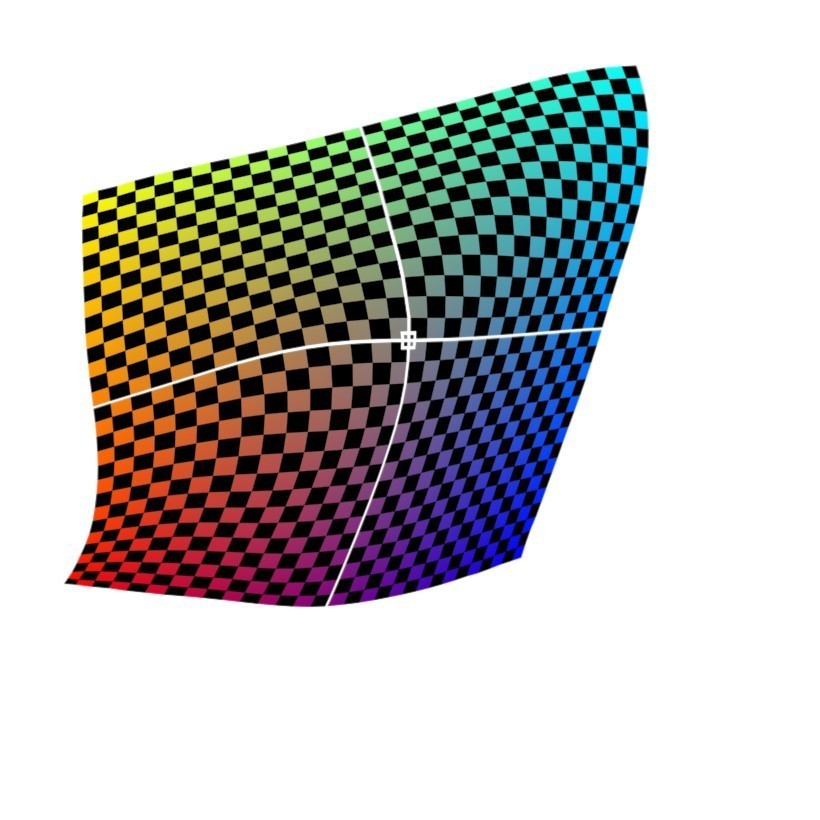}}
\subfloat[Interpolation Meshlets]{\raisebox{\raiseupby}{
\includegraphics[width=\interpScaling\meshletWidth]{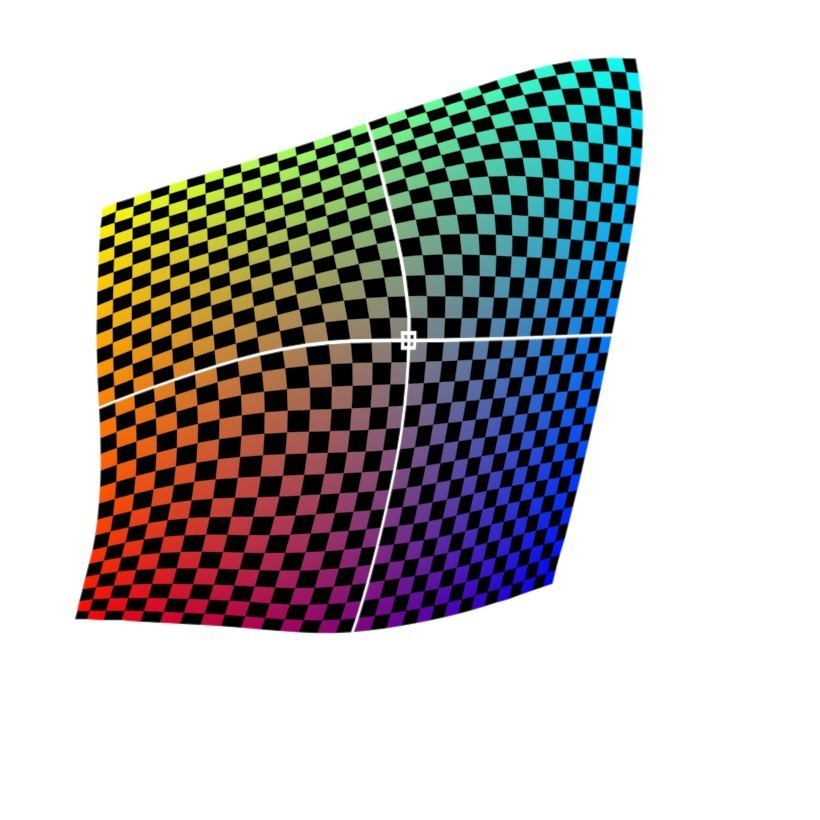}~
\includegraphics[width=\interpScaling\meshletWidth]{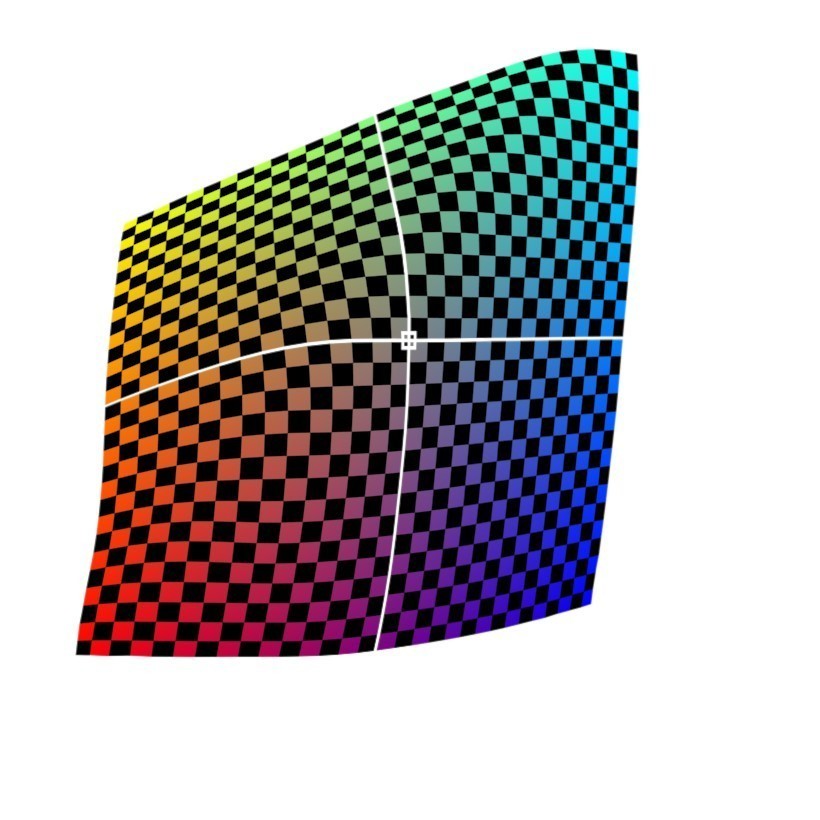}~
\includegraphics[width=\interpScaling\meshletWidth]{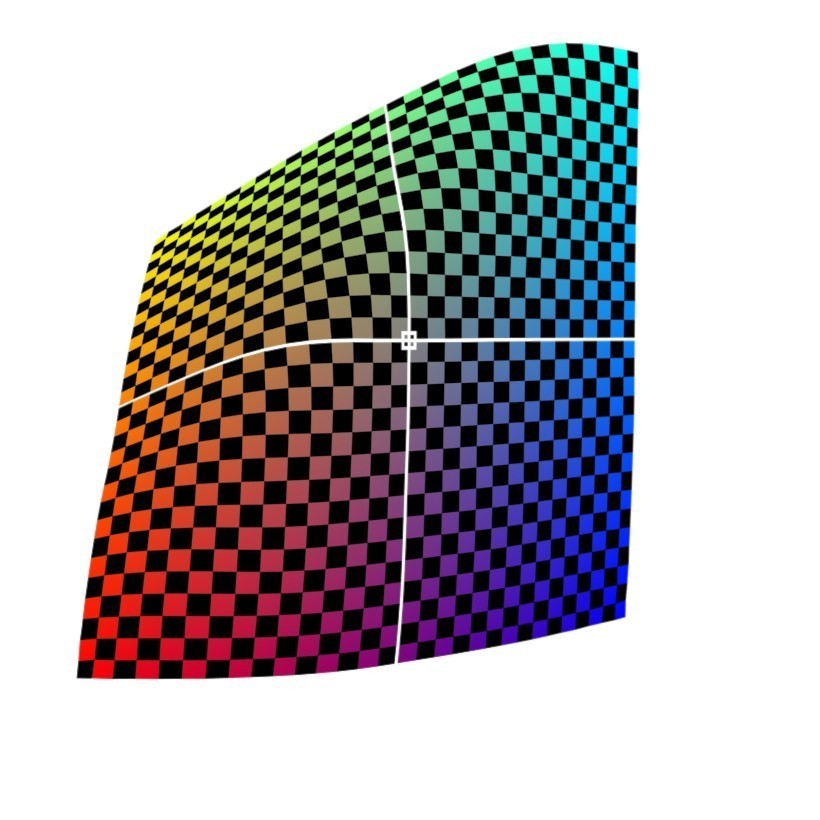}~
\includegraphics[width=\interpScaling\meshletWidth]{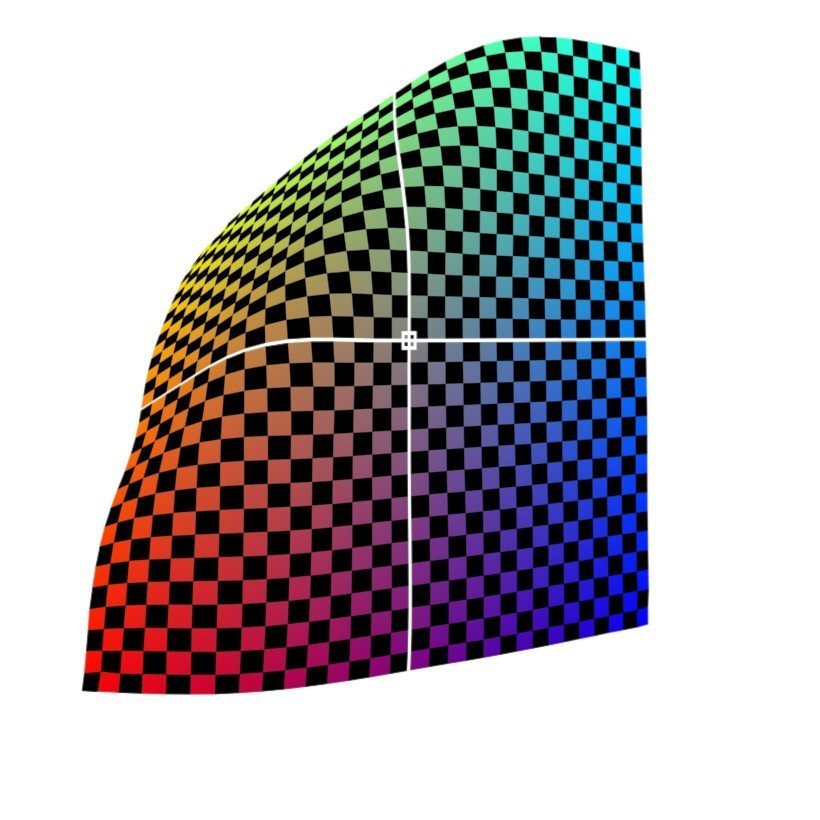}
}}
\subfloat[Meshlet B]{\includegraphics[width=\meshletWidth]{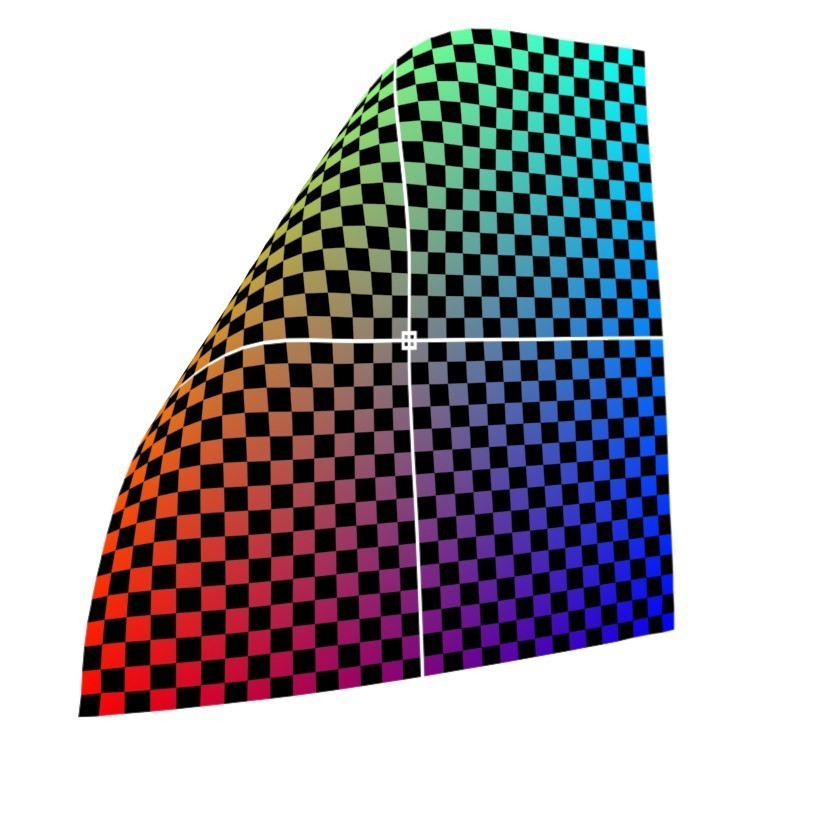}}

%% file: implementation.tex

\subsection{Optimization}\label{sec:optim_details}
We start by describing a few details that improve the efficiency of the optimization procedure or the quality of the resulting meshes.

\paragraph{Mesh initialization.} The auxiliary mesh $\mathcal{M}(t_0)$ can be any genus-zero mesh that satisfies the meshlets' priors (see Section~\ref{sec:overall}).
The actual choice, however, does affect the number of iterations required to converge.
We initialize our approach with an overly-smoothed Laplacian reconstruction. 
Empirically, we have observed that the results of our algorithm initialized in this way are effectively indistinguishable from the results obtained by using a sphere as an initialization; convergence, however, does take a fraction of the time.
For reference, we show a few examples of $\mathcal{M}(t_0)$ in the Supplementary.

\paragraph{Meshlets re-sampling.} As the optimization progresses, the shape and the size of the auxiliary mesh $\mathcal{M}$ may change significantly.
On the one hand, this is a desirable behavior: if the mesh can scale to arbitrary sizes, it can properly match the size of the underlying mesh, even when the initialization is far from it.
On the other, it results in a sparser meshlet coverage and, potentially, no coverage in some areas.
Moreover, it could cause meshlets to be overly-stretched.
Therefore, every 20 iterations of enforcing local shape priors and global consistency (blue arrow in Figure~\ref{fig:optimization_loop}), we re-sample the meshlets on the current mesh.

\paragraph{Re-meshing.} Large changes from the initialization may also cause issues to the mesh itself, which may stretch in some regions or become otherwise irregular.
One way to prevent this is to use strong smoothness priors when enforcing global consistency, but that would hinder our ability to reconstruct sharp features.
At the end of every iteration, we re-mesh $\mathcal{M}$ using Screened Poisson Reconstruction~\cite{screenedpoisson} to encourage smoothness while respecting the priors enforced by our approach, \ie, preserving the sharpness of local features.
We provide more details in the Supplementary.

\subsection{Meshlet training}\label{sec:train_details}
To train the meshlets network, we sample $2.2\times10^6$ meshlets from the ShapeNet dataset~\cite{shapenet2015}.
We extract meshlets by randomly selecting objects across several classes.
We then apply three different scales to each object and extract 256 meshlets for each scale, so that our meshlet dataset captures both fine and coarse details.
Note that we disregard meshlets that are problematic. 
Specifically, we use the geodesic distance algorithm by Melv{\ae}r~\etal~\cite{melvaer2012geodesic} and reject those meshlets for which the geodesic distance calculation results in a large anisotropic stretch, or fails altogether.
The network, then, is trained to reconstruct these meshlets using $\ell_2$ as a loss.
In all of our experiments we use meshlets of size $31\times 31 \times 3$.
To exploit the latent space of natural meshlets, we use a fully-connected encoder decoder network that takes as input a $(31\cdot 31)\times 3$ vector (\ie, a vectorized version of the meshlet).
The encoder and the decoder are symmetric with 6 layers each, and the latent code vector is one third of the input dimension.

%% file: results.tex

\paragraph{Comparisons.}

\begin{figure*}
\input{figures/results/result_page_cvpr}
\caption{Qualitative comparison of several reconstruction methods and our approach. On the left is the input, the ground-truth (GT) mesh overlaid with the sparse, noisy point cloud (PC). We show results with normals estimated with Meshlab~\cite{meshlab} and PCPNet~\cite{pcpnet}. We reconstruct the resulting point clouds with both RILMS~\cite{oztireli2009feature} and Screened Poisson~\cite{screenedpoisson}. Laplacian regularizer~\cite{laplacian} is shown for two levels of smoothing. We also show three recent deep learning approaches: Deep Geometric Prior (DGP)~\cite{Williams2019}, AtlasNet~\cite{Groueix_2018_CVPR} and OccNet~\cite{occNet}. All of these methods struggle to cope with noise, classes not seen in training, or both.}\label{fig:big_table}
\end{figure*}
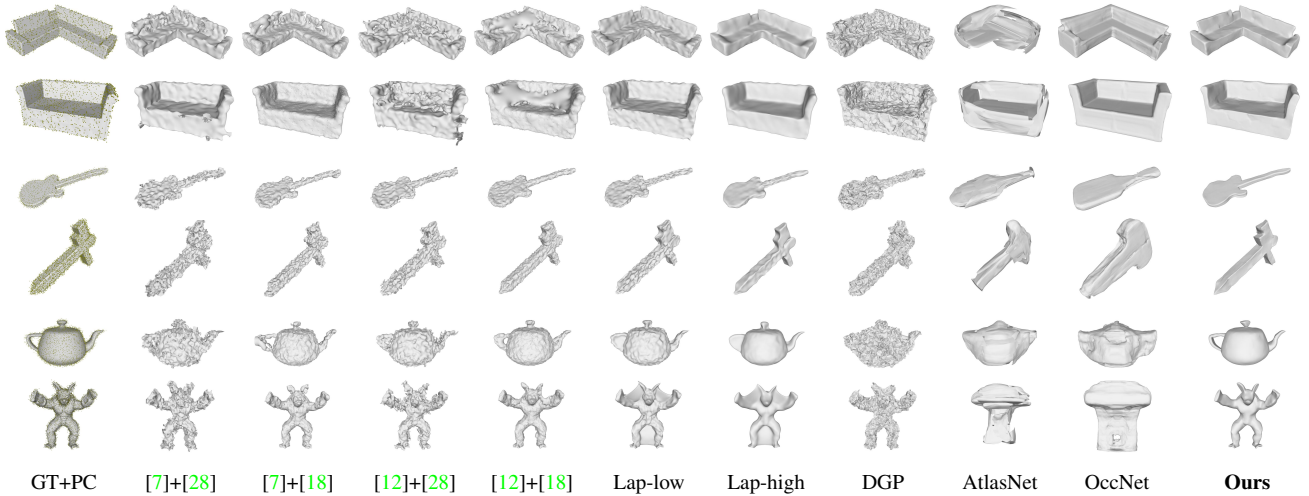

In this section we evaluate our method against state-of-the-art approaches.
Then, we compare our meshlets to other local shape priors to validate their importance.

We compare our method with several state-of-the-art mesh reconstruction approaches.
The first is Screened Poisson~\cite{screenedpoisson}, a widely used, traditional technique that creates watertight surfaces from oriented point clouds.
Because our input is a raw point cloud, we need to estimate normals.
We use two methods to estimate normals. 
One is MeshLab's normal estimation, which fits local planes and uses them to estimate normals~\cite{meshlab}.
The other is a recently published, learning-based method called PCPNet~\cite{pcpnet}.
The second mesh reconstruction approach is RILMS, a method by {\"O}ztireli~\etal~\cite{oztireli2009feature}, which also requires oriented points.
They propose a recent variant of marching cubes that preserves sharp features using non-linear regression.
In addition, we compare against Laplacian mesh optimization~\cite{laplacian}.
Leveraging the fact that the norm of the mesh's Laplacian captures the local mean curvature, this mesh optimization algorithm optimizes the Laplacian at the vertices in a weighted least-square sense.
The algorithm has a free parameter that regulates the smoothness of the resulting surface.
After a parameter sweep we found that no single parameter would yield the best results over the whole dataset.
Therefore we settled for two values, each offering a different compromise between denoising and over-smoothing.
We also compare with Deep Geometric Prior (DGP)~\cite{Williams2019}, OccNet~\cite{occNet}, and AtlasNet~\cite{Groueix_2018_CVPR}, all of which are deep learning methods.
The last two approaches learn priors at the object level.

\paragraph{Data.}
We test all the methods on 20 objects.
To validate that our method generalizes well, we also include four objects that are commonly used by the graphics community (Suzanne, the Stanford Bunny, Armadillo, and the Utah Teapot).
We select the rest of the meshes from the test set of ShapeNet dataset~\cite{shapenet2015}.
We show all the objects in the Supplementary.
However, because the ShapeNet meshes are not always watertight, we pre-process them with the algorithm proposed by Stutz and Geiger~\cite{Stutz2018ARXIV}.
To generate the input point clouds for evaluation, we randomly decimate the number of vertices of the watertight meshes by different factors, obtaining three different sparsity levels.
For each sparsity level we also add an increasingly large amount of Gaussian noise.
We describe the parameters we use, and offer visualization of the different levels of noise in the Supplementary.

\paragraph{Numerical evaluation.}
For our numerical evaluation we use the symmetric Hausdorff distance, which reports the largest vertex reconstruction error for each mesh, and the Chamfer-$\ell_1$ distance, which computes the distance between two meshes after assigning correspondences based on closest vertices.
Table~\ref{tab:numbers} shows that our method performs consistently better than all the competitors.
The gap is most apparent when comparing with deep-learning methods that learn priors at the object level, further suggesting that our strategy to learn local priors is a promising direction.
We list the numbers for each object across the different noise settings in the Supplementary.

\begin{table}
\footnotesize
\centering
\begin{tabular}{cccc}
&&Chamfer-$\ell_1$&Hausdorff\\
\hline
\multirow{2}{*}{Meshlab~\cite{meshlab}~~+}&Scr.Pois.~\cite{screenedpoisson} & 0.0285 / 0.0112 & 0.339 / 0.102\\
\cline{2-4}
& RILMS~\cite{oztireli2009feature} & 0.0177 / 0.0166 & 0.149 / 0.148\\
\hline
\multirow{2}{*}{PCPNet~\cite{pcpnet}~~+}&Scr.Pois.~\cite{screenedpoisson} & 0.0122 / 0.0109 & 0.147 / 0.140\\
\cline{2-4}
& RILMS~\cite{oztireli2009feature} & 0.0181 / 0.0176 & 0.151 / 0.153\\
\hline
\multirow{2}{*}{Laplacian~\cite{laplacian}} & Low & 0.0104 / 0.0103 & {\color{darkred}{0.100}} / {\color{darkred}{0.065}}\\
\cline{2-4}
& High & {\color{darkred}{0.0096}} / {\color{darkred}{0.0094}} & 0.103 / 0.069\\
\hline
\multicolumn{2}{l}{Deep Geometric Prior~\cite{Williams2019}} & 0.0128 / 0.0130 & 0.147 / 0.148\\
\hline
\multicolumn{2}{l}{AtlasNet~\cite{Groueix_2018_CVPR}} & 0.0415 / 0.0377 & 0.293 / 0.263\\
\hline
\multicolumn{2}{l}{OccNet~\cite{occNet}} & 0.0630 / 0.0627 & 0.304 / 0.285\\
\hline
\multicolumn{2}{l}{Ours} & {\color{darkgreen}{0.0090}} / {\color{darkgreen}{0.0092}} & {\color{darkgreen}{0.054}} / {\color{darkgreen}{0.047}}
\end{tabular}
\caption{We compare our method with state-of-the-art approaches, both traditional and learning-based, using two metrics. For each metric we report mean/median values over all of the objects reconstructed, and across multiple levels of noise. {\color{darkgreen}{Green}} and {\color{darkred}{Red}} indicate the best and second best method respectively.}\label{tab:numbers}   
\end{table}

\paragraph{Qualitative evaluation.}
\begin{figure}
\centering
\subfloat[]{
\includegraphics[width=.245\columnwidth, trim={80 0 80 0}, clip]{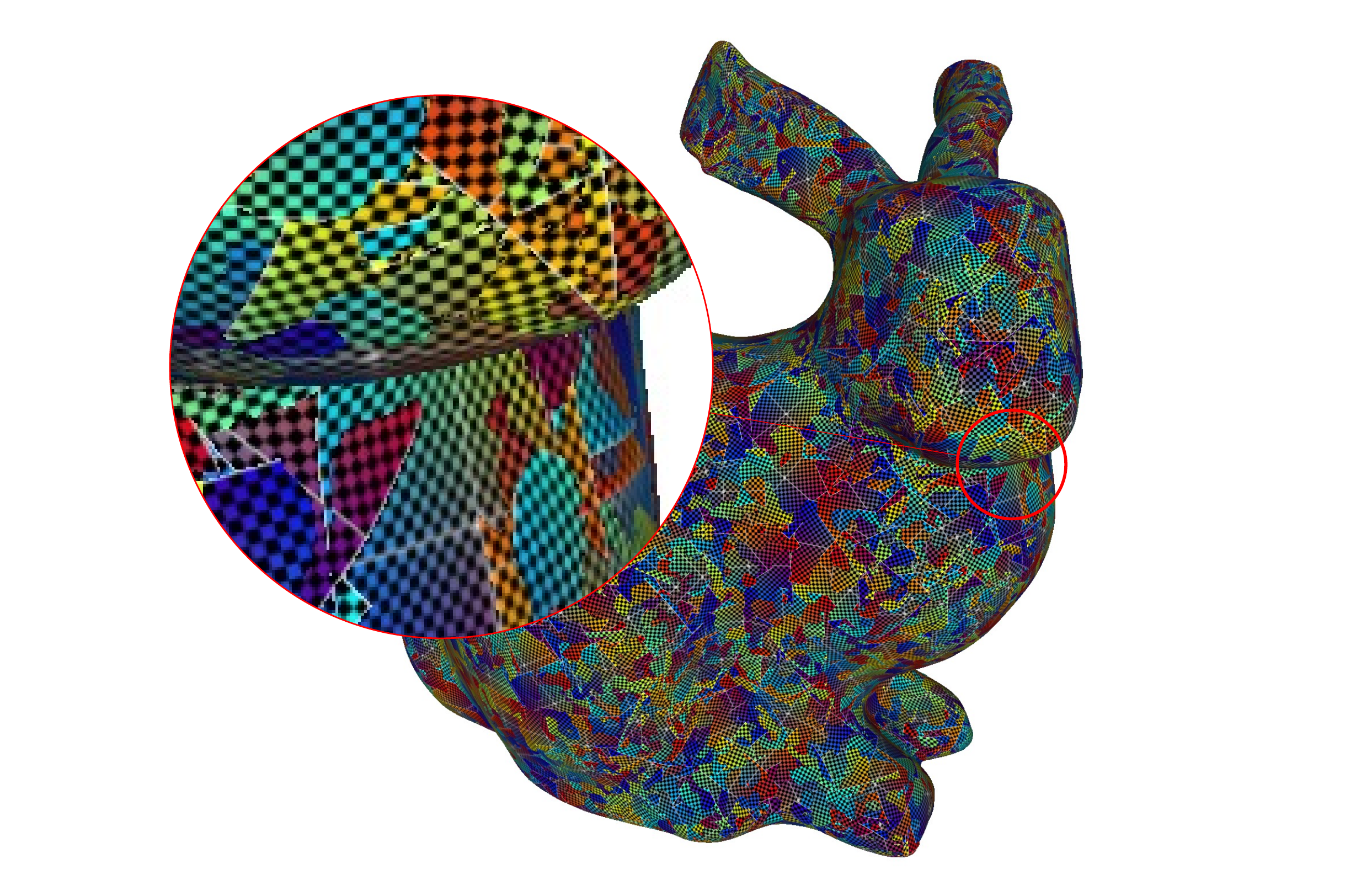}
\includegraphics[width=.245\columnwidth, trim={200 30 0 0}, clip]{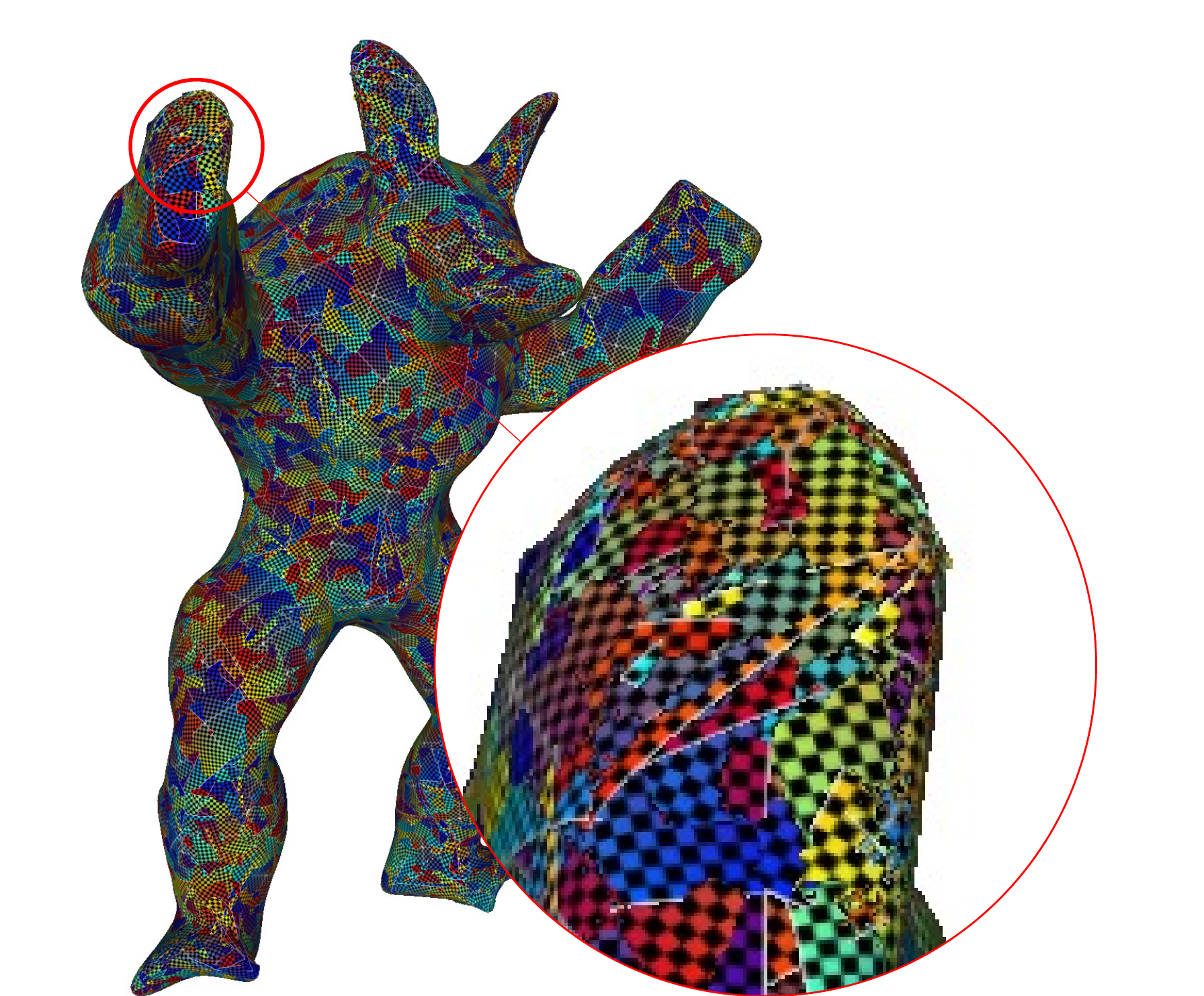}
}
\subfloat[]{
\includegraphics[width=.245\columnwidth, trim={80 0 80 0}, clip]{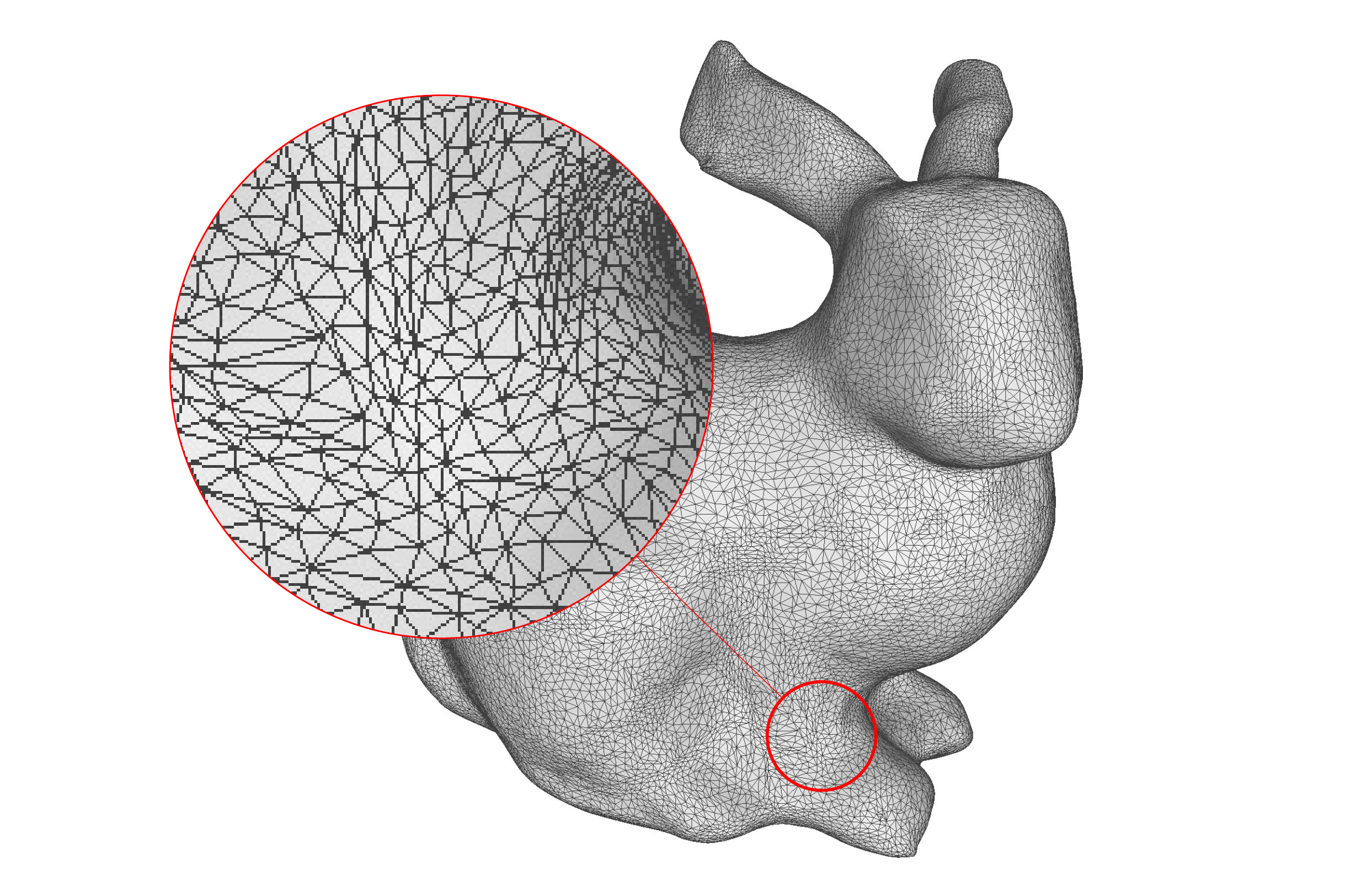}
\includegraphics[width=.245\columnwidth, trim={200 30 0 0}, clip]{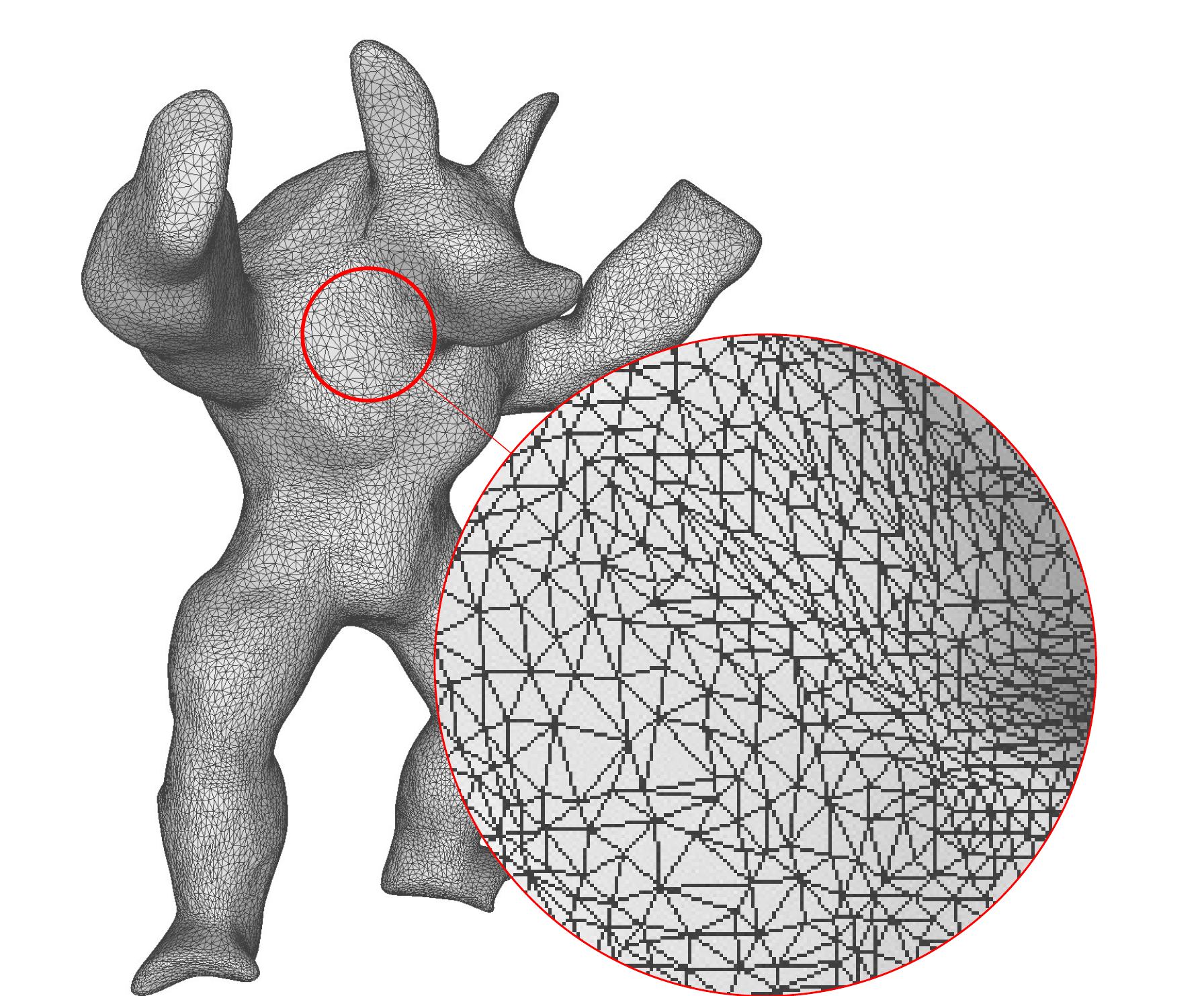}
}
\caption{Our final meshlets are globally consistent and capture the local shape of the mesh (a). The resulting mesh is regular over the whole reconstructed object (b).}\label{fig:union}
\end{figure}
In Figure~\ref{fig:union} we show both the meshlets at the end of our optimization (a) and the quality of the final mesh reconstructed by our algorithm (b).
Note that, thanks to the re-meshing steps during our optimization procedure (Section~\ref{sec:optim_details}), our output is a high-quality, regular mesh.   
We also show a subset of the objects used in the numerical evaluation in Figures~\ref{fig:teaser}~and~\ref{fig:big_table} for different levels of sparsity and noise.
Additional results are in the Supplementary.
Competing methods are significantly impacted by noise and produce overly smooth results to attenuate its effect.
For example, the Laplacian reconstructions obtained with low regularization (Lap-low) are still noisy, while those for which we used high regularization (Lap-high) are over-smoothed.
Even Screened Poisson reconstruction~\cite{screenedpoisson}, the de facto standard among traditional methods, used in conjunction with PCPNet~\cite{pcpnet} to estimate the normals, produces visibly noisy results.
Finally, as also shown in Figure~\ref{fig:teaser}, state-of-the-art deep learning methods only work on objects seen in training and for low levels of noise.
Our results, on the other hand, offer the best trade-off between detail and noise by recovering locally sharp features and small details, despite the sparsity and noise of the point clouds.

\paragraph{On the importance of natural meshlets.}
\begin{figure}
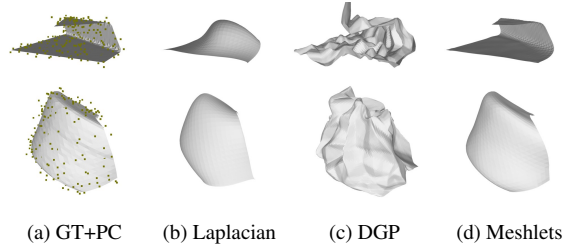

\include{figures/meshlets_are_awesome/priors}
\caption{Meshlets reconstruct local features more accurately than other priors. Input point clouds shown on the GT mesh in (a).}\label{fig:meshlets_are_the_best}
\end{figure}
Our meshlet priors are the core of our method.
Here we compare our natural meshlets with other shape priors to isolate their contribution to the overall quality of the result.
The first is a Laplacian regularizer, which is a standard smoothness prior~\cite{laplacian}.
The second, is Deep Geometric Priors, the recent work by Williams~\etal, which suggests that a neural network is also, in itself, a prior for local geometry~\cite{Williams2019}.
We use these priors and our natural meshlet prior to optimize small patches of mesh to small point clouds extracted from real objects.
Figure~\ref{fig:meshlets_are_the_best} shows two representative examples.
Despite the complexity of the local shape, and the level of noise, the optimization that uses our strategy (Section~\ref{sec:meshlets}) is able to correctly estimate the underlying meshlet.
On the contrary, Deep Geometric Prior over-fits to the noise, and the Laplacian regularizer over-smooths the surface.
On $10^5$ meshlets, the average symmetric Hausdorff distance is $0.027$ for DGP, $0.029$ for Laplacian, and $0.024$ for our method.

\begin{figure}
    \vspace{-5mm}
    \input{figures/ablation/ablation.tex}
    \caption{Replacing our meshlet priors with Laplacian priors leads to overly-smoothed results. However, using our meshlet priors without using global consistency (GC) produces irregular meshes.}\label{fig:meshlets_are_necessary}
\end{figure}
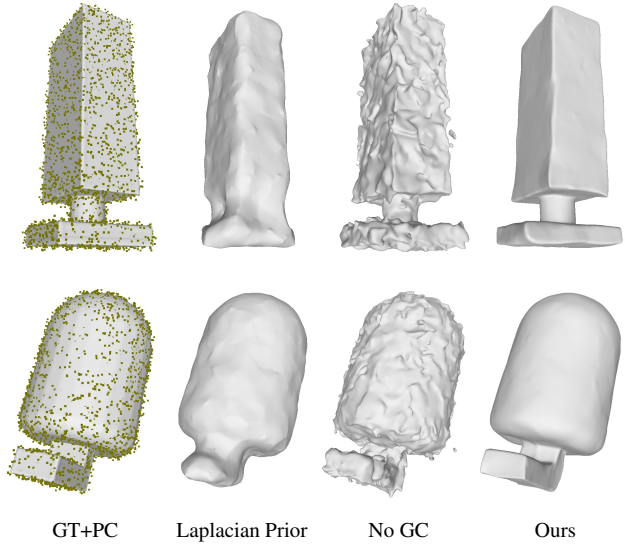

To further validate the importance of our natural meshlet priors towards our overall optimization procedure, we swap our latent space search with a Laplacian prior, and leave the rest of the algorithm untouched.
As shown in Figure~\ref{fig:meshlets_are_necessary}, using meshlet priors allows to better smooth noise while still reconstructing sharp features, such as edges and corners.
However, we also note that enforcing global consistency (GC) is an important step of our method, without which the quality of the final reconstruction degrades significantly, as shown in Figure~\ref{fig:meshlets_are_necessary}. 

%% file: figures/results/result_page_cvpr.tex

\newlength{\numcrops}
\setlength{\numcrops}{11pt}
\newcommand{\fitscale}{.98}

\newlength{\cropwidth}
\settowidth{\cropwidth}{\includegraphics{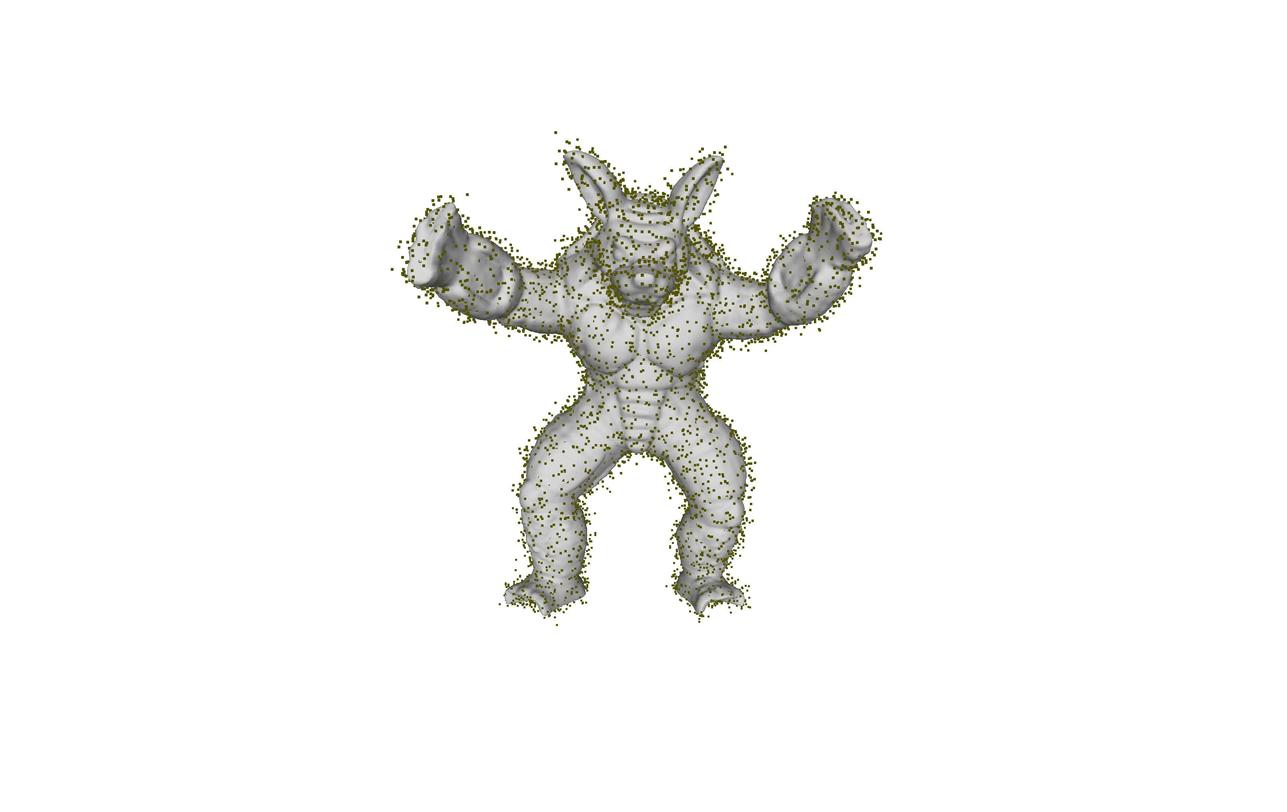}}
\newlength{\one}
\setlength{\one}{1pt}
\newlength{\tgtwidth}
\setlength\tgtwidth{\textwidth*\ratio{\one}{\numcrops}}

\captionsetup[subfigure]{labelformat=empty}
\centering
\vspace{-10mm}
\subfloat{\includegraphics[width=\fitscale\tgtwidth, trim={240 80 240 80}, clip]{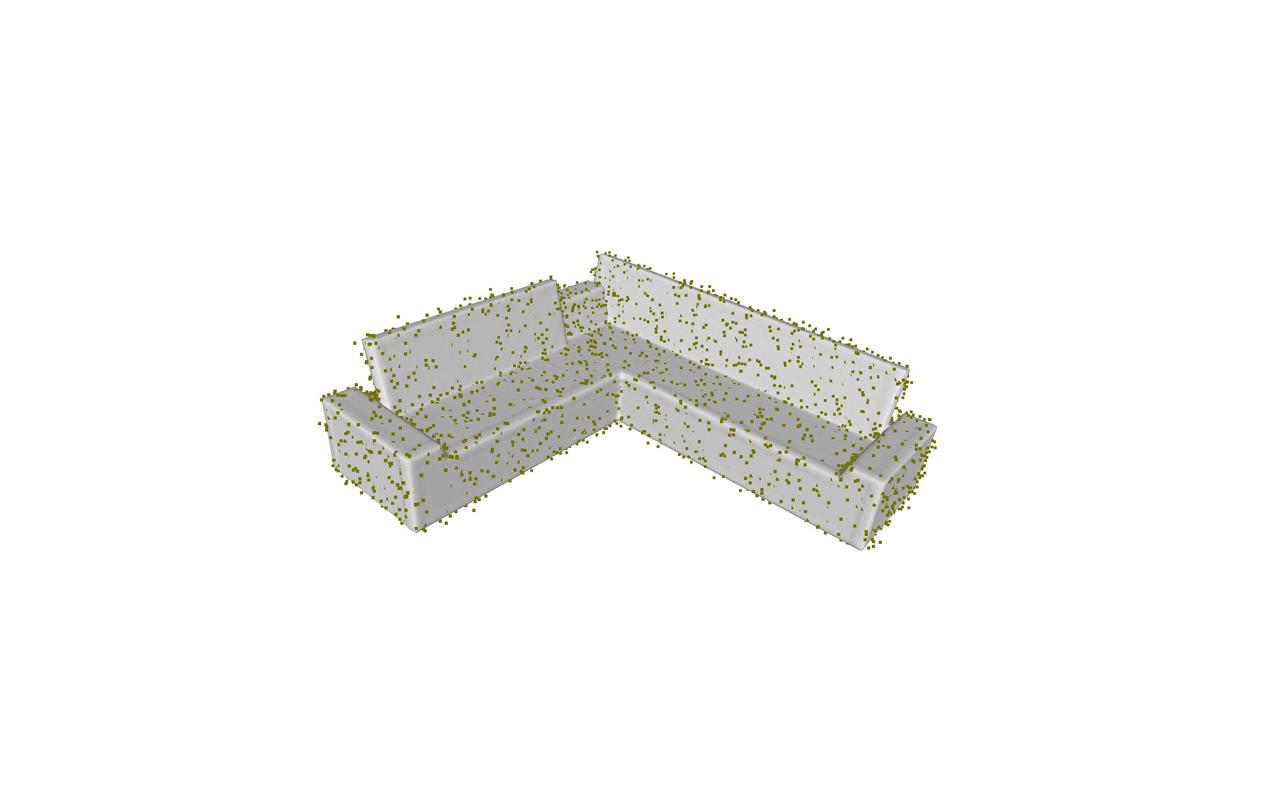}}
\subfloat{\includegraphics[width=\fitscale\tgtwidth, trim={240 80 240 80}, clip]{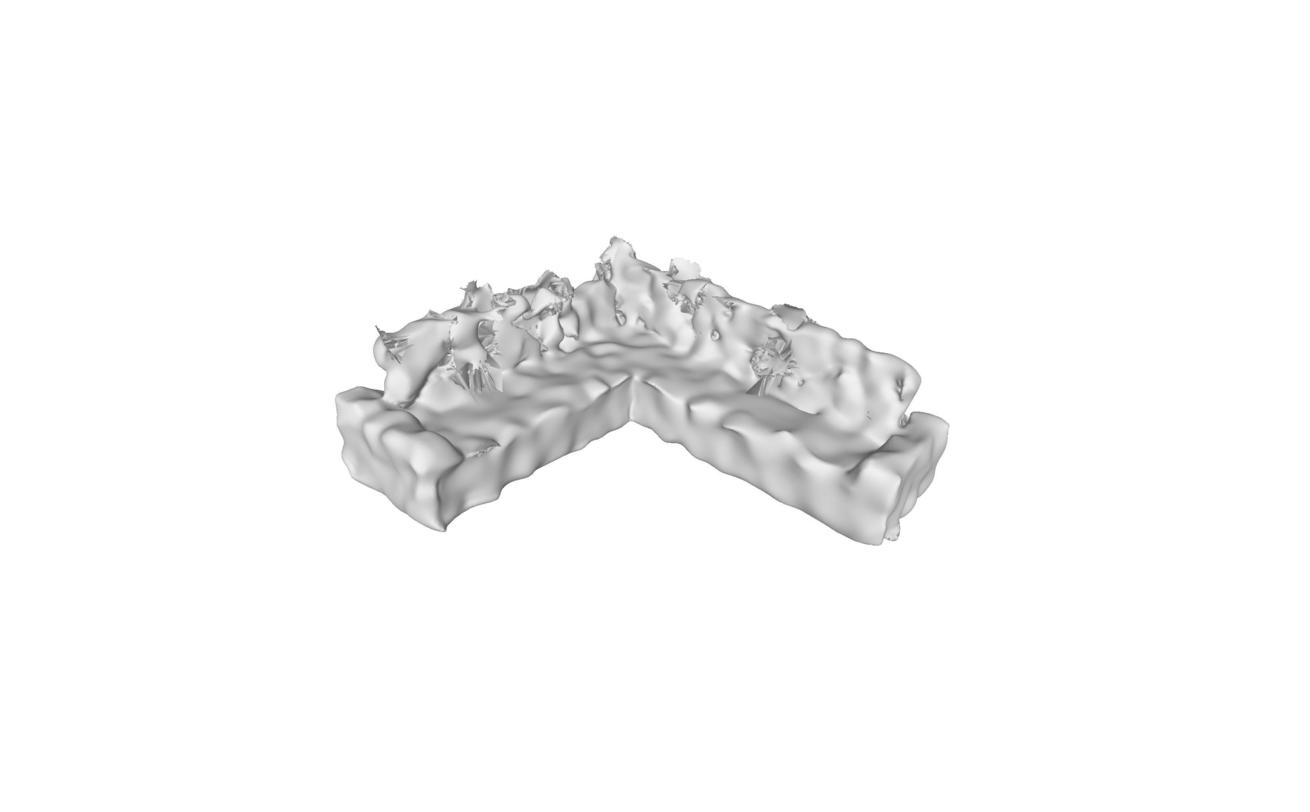}}
\subfloat{\includegraphics[width=\fitscale\tgtwidth, trim={240 80 240 80}, clip]{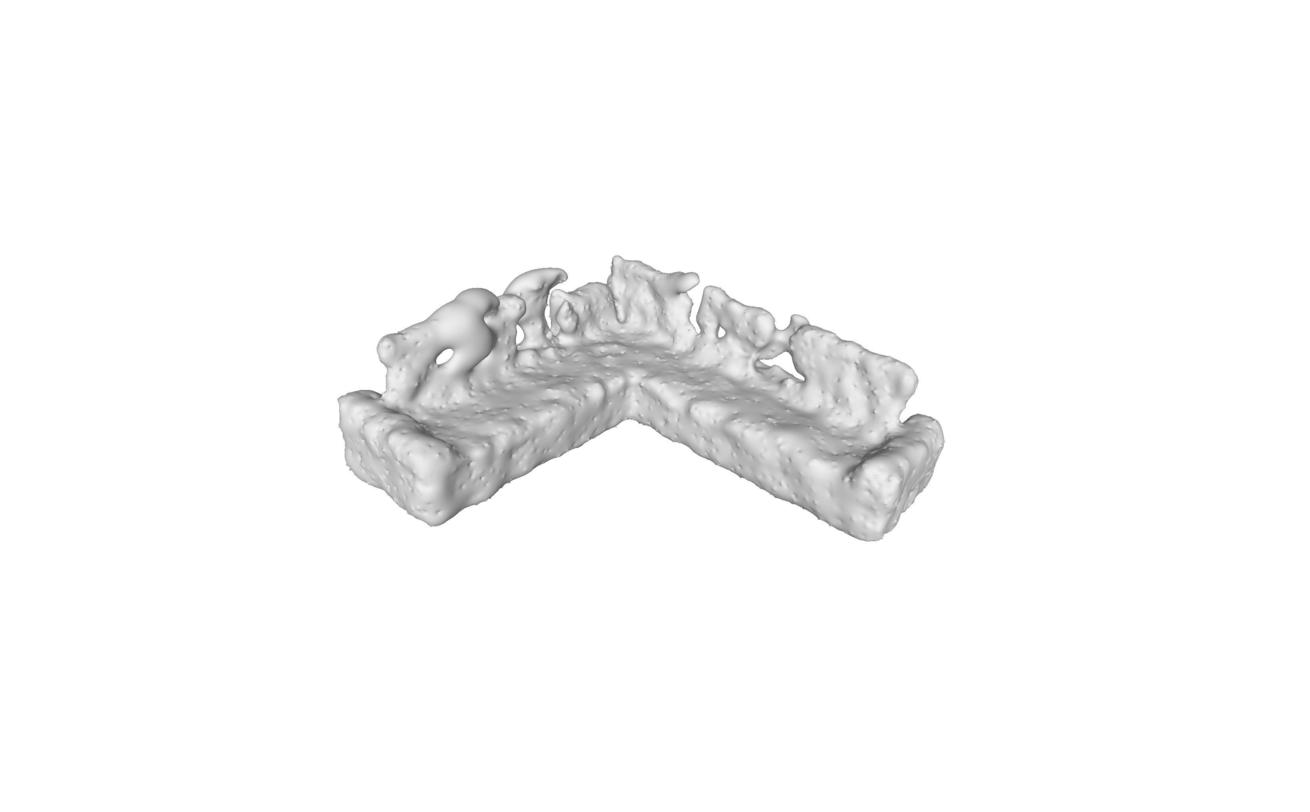}}
\subfloat{\includegraphics[width=\fitscale\tgtwidth, trim={240 80 240 80}, clip]{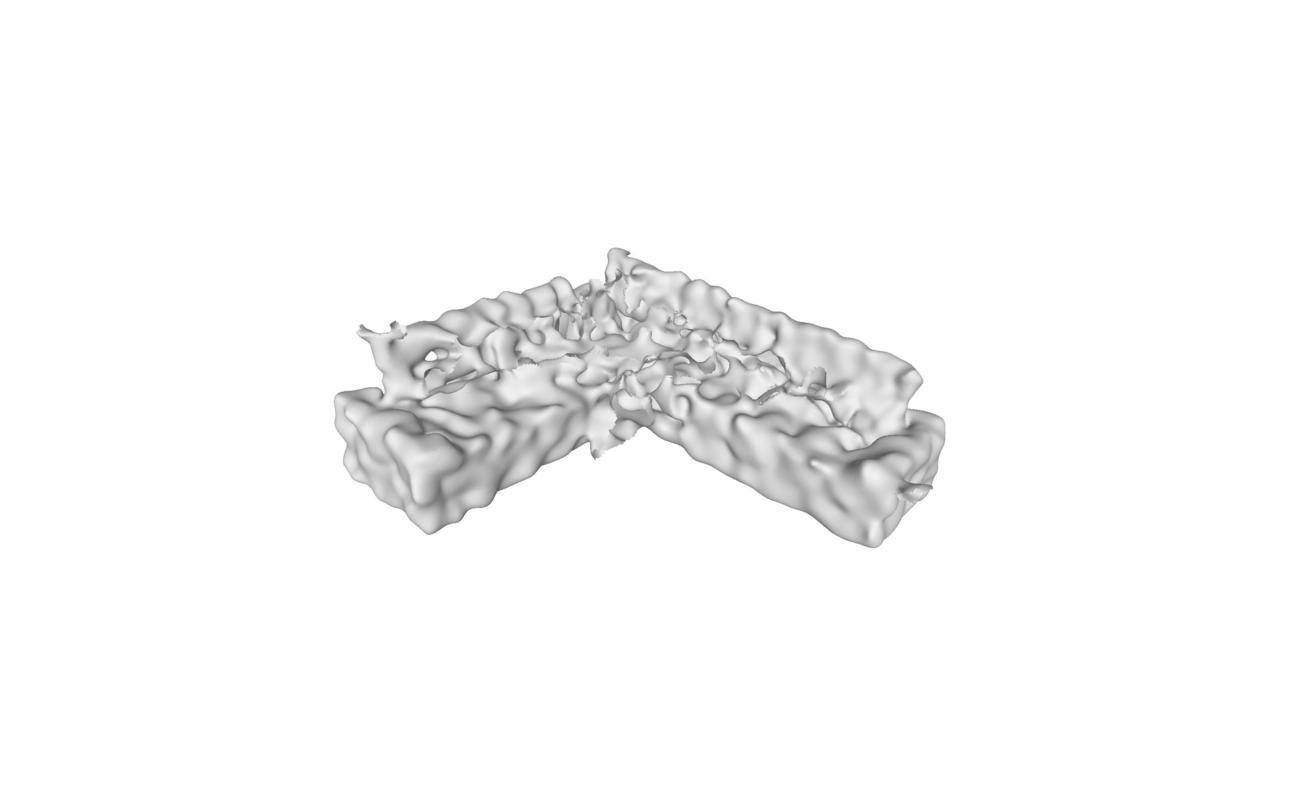}}
\subfloat{\includegraphics[width=\fitscale\tgtwidth, trim={240 80 240 80}, clip]{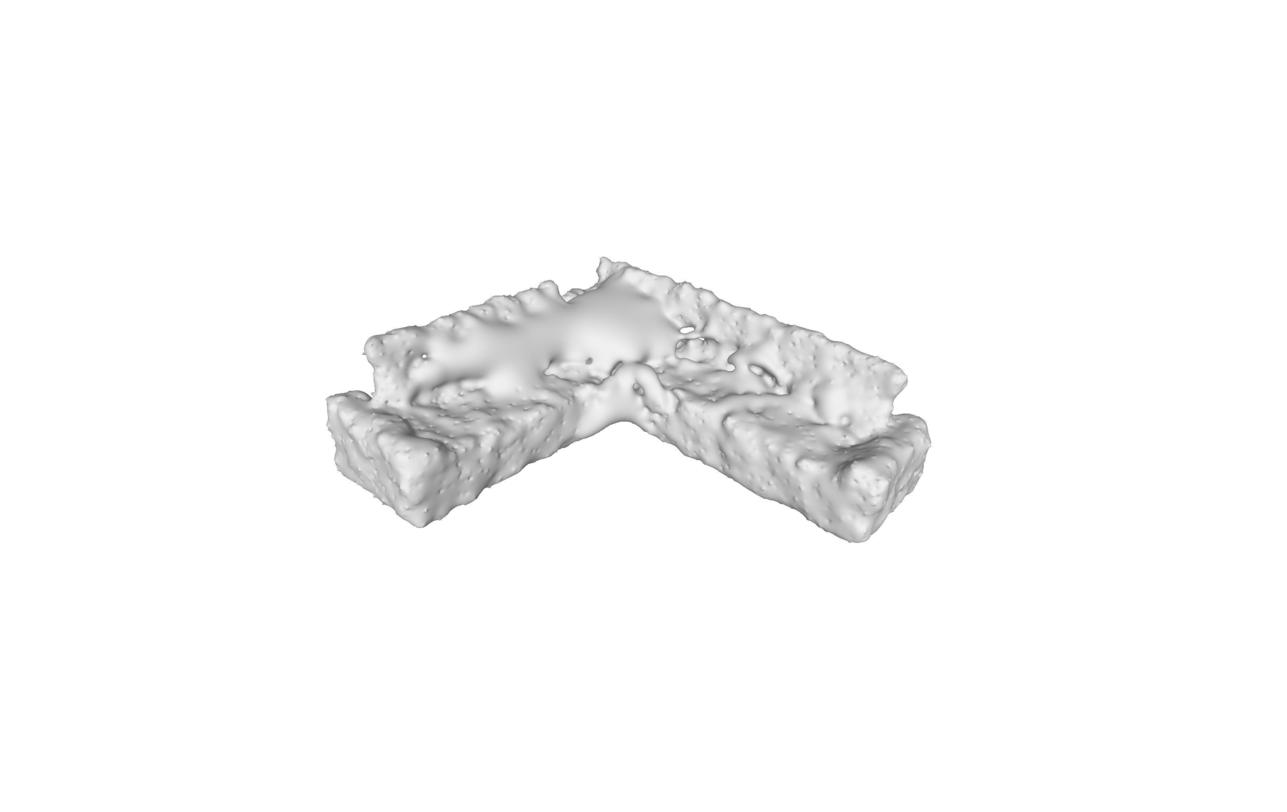}}
\subfloat{\includegraphics[width=\fitscale\tgtwidth, trim={240 80 240 80}, clip]{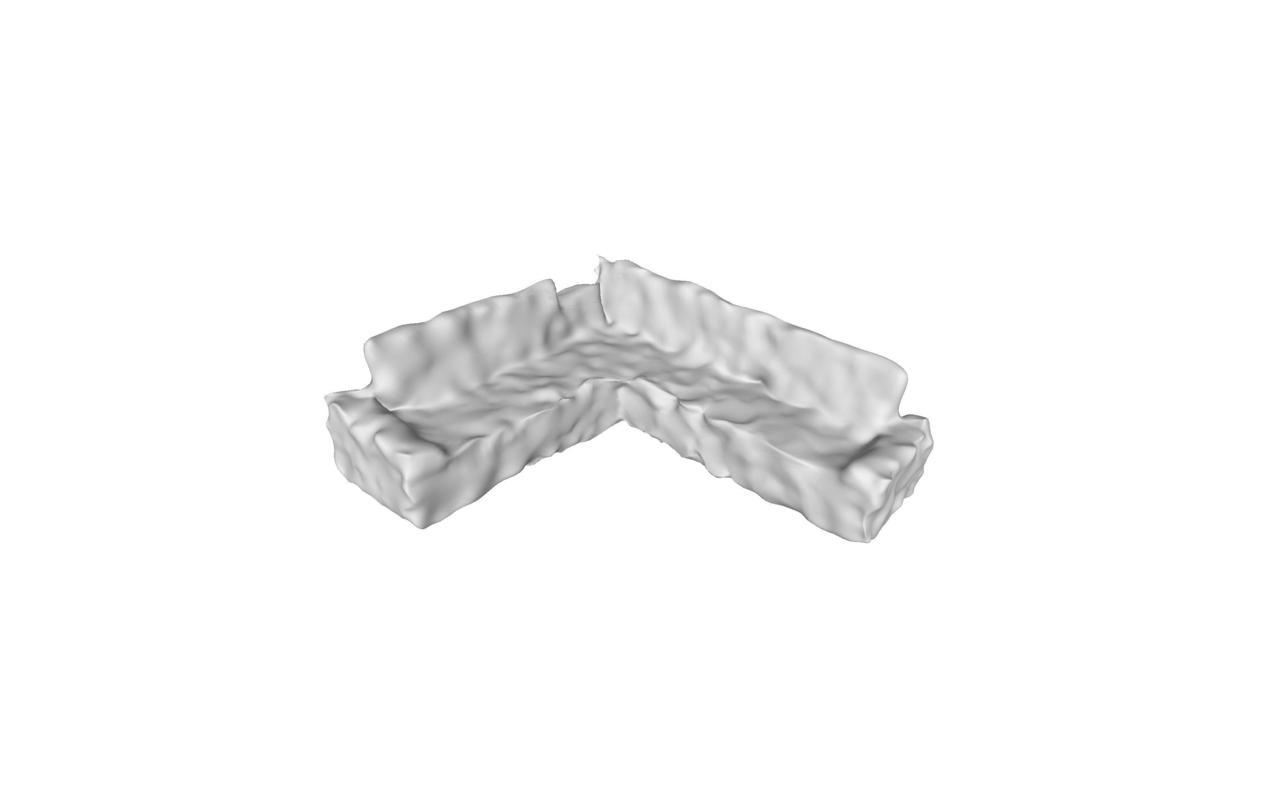}}
\subfloat{\includegraphics[width=\fitscale\tgtwidth, trim={240 80 240 80}, clip]{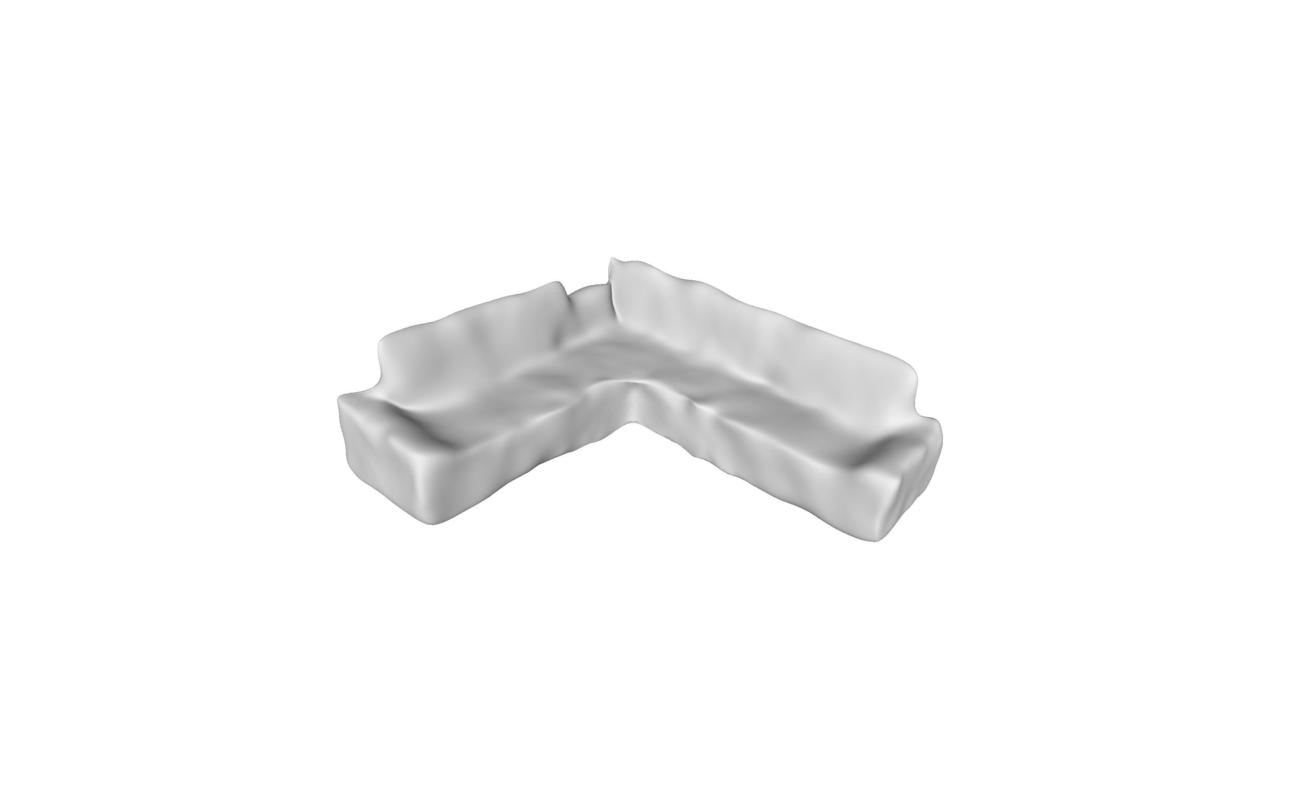}}
\subfloat{\includegraphics[width=\fitscale\tgtwidth, trim={240 80 240 80}, clip]{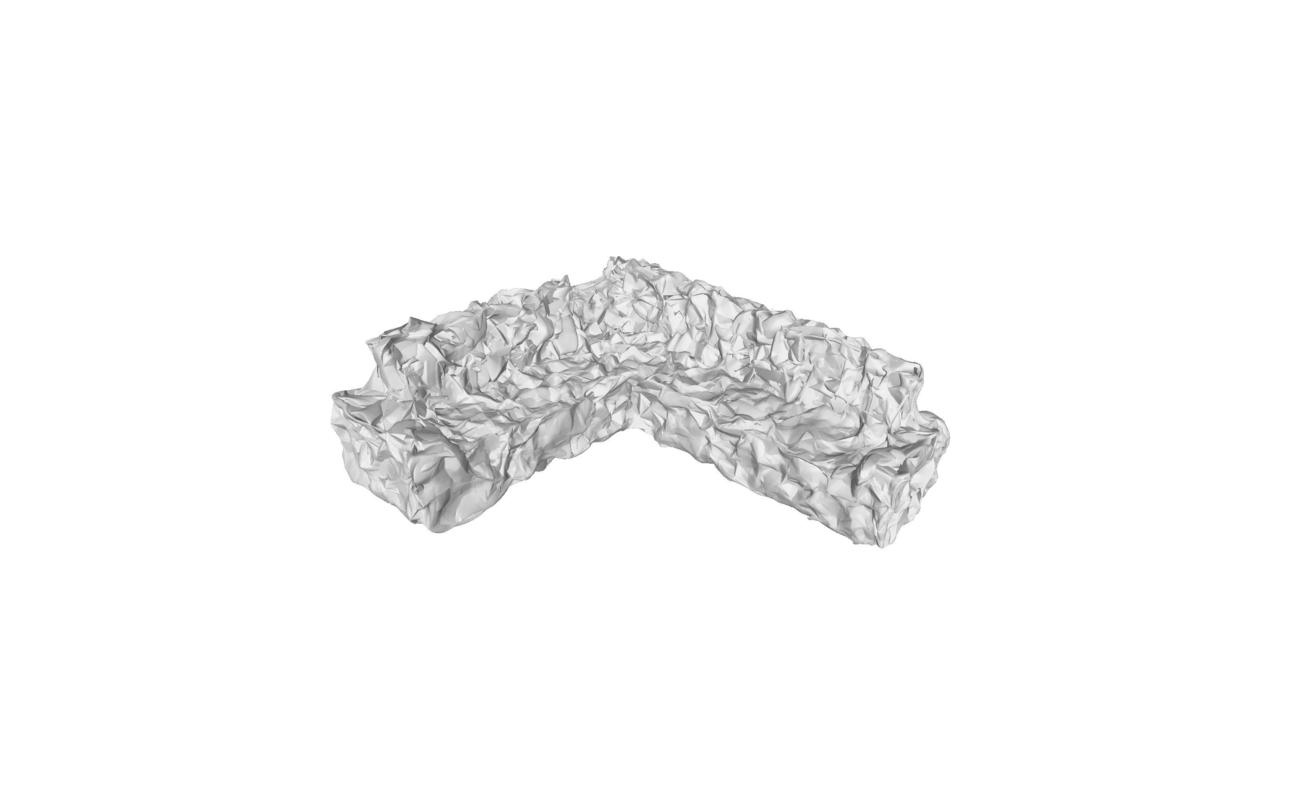}}
\subfloat{\includegraphics[width=\fitscale\tgtwidth, trim={240 80 240 80}, clip]{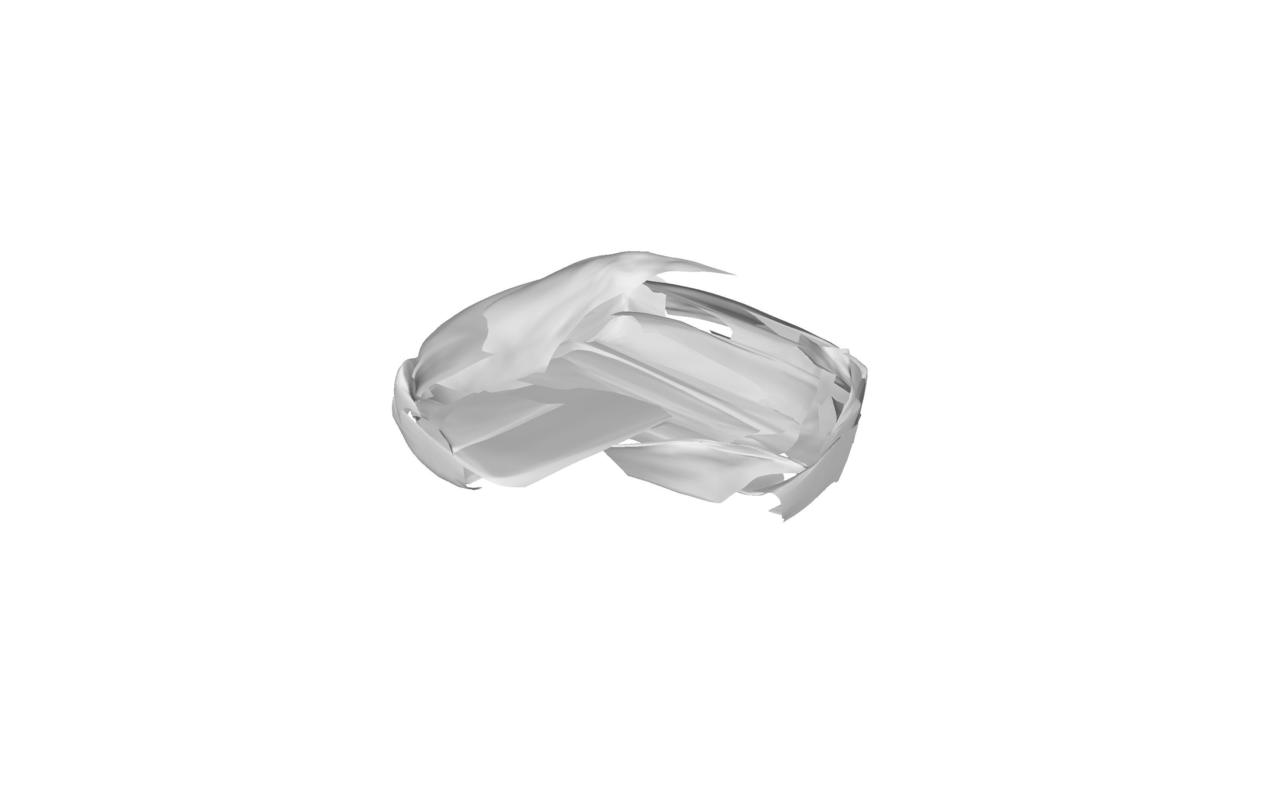}}
\subfloat{\includegraphics[width=\fitscale\tgtwidth, trim={240 80 240 80}, clip]{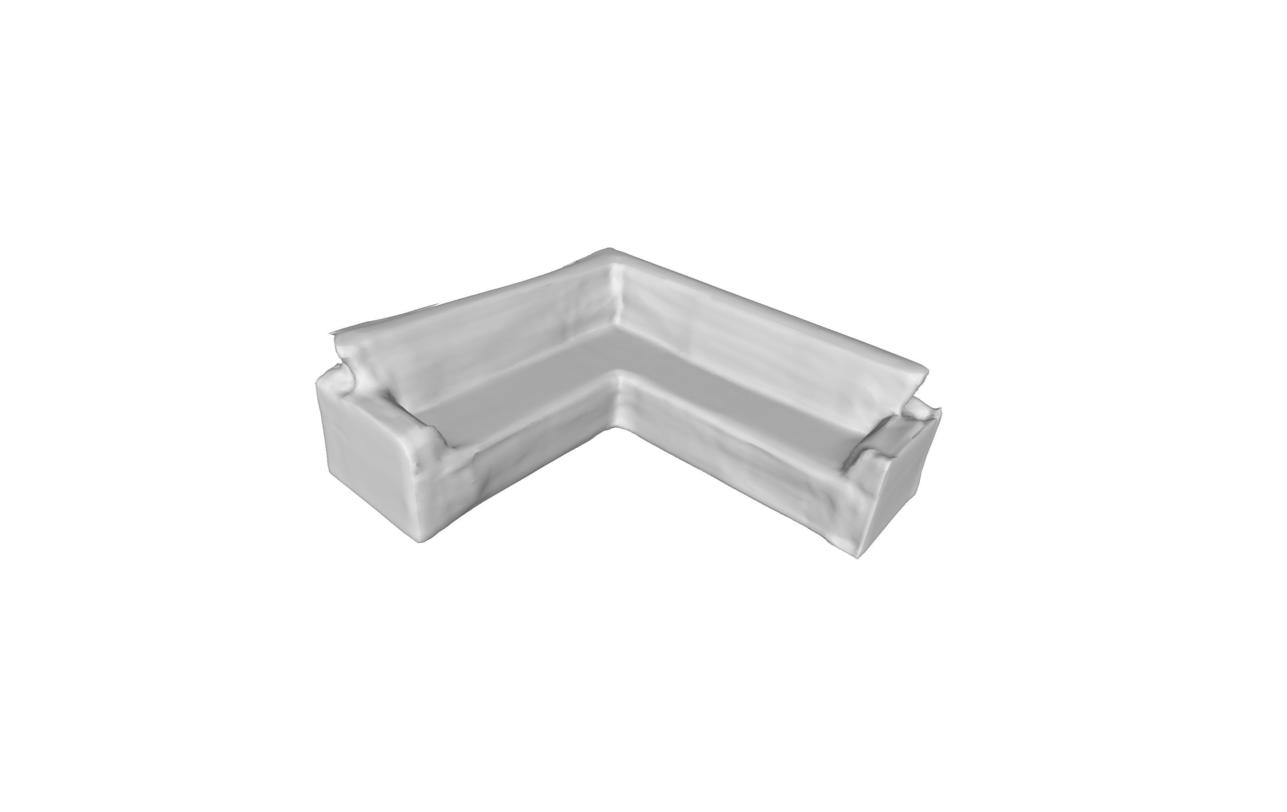}}
~
\subfloat{\includegraphics[width=\fitscale\tgtwidth, trim={240 80 240 80}, clip]{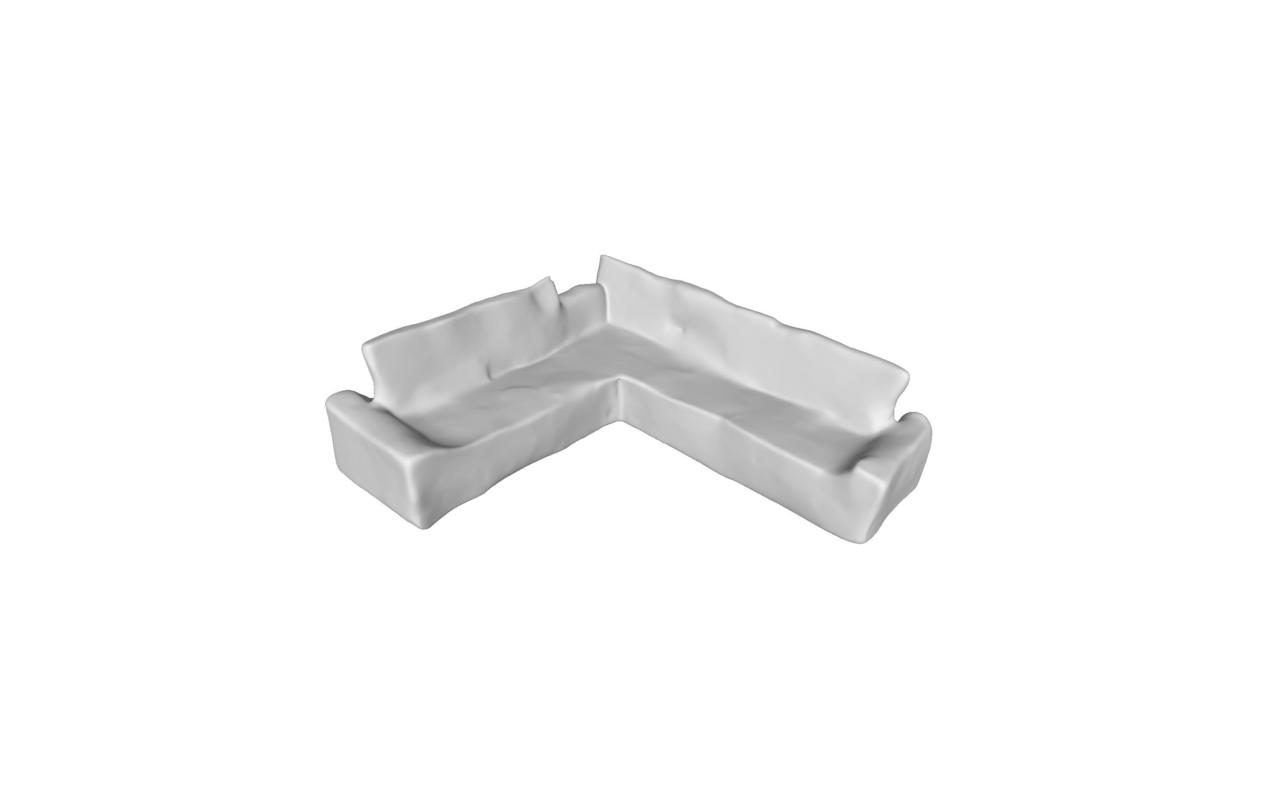}}
\\
\vspace{-5mm}
\subfloat{\includegraphics[width=\fitscale\tgtwidth, trim={180 40 210 180}, clip]{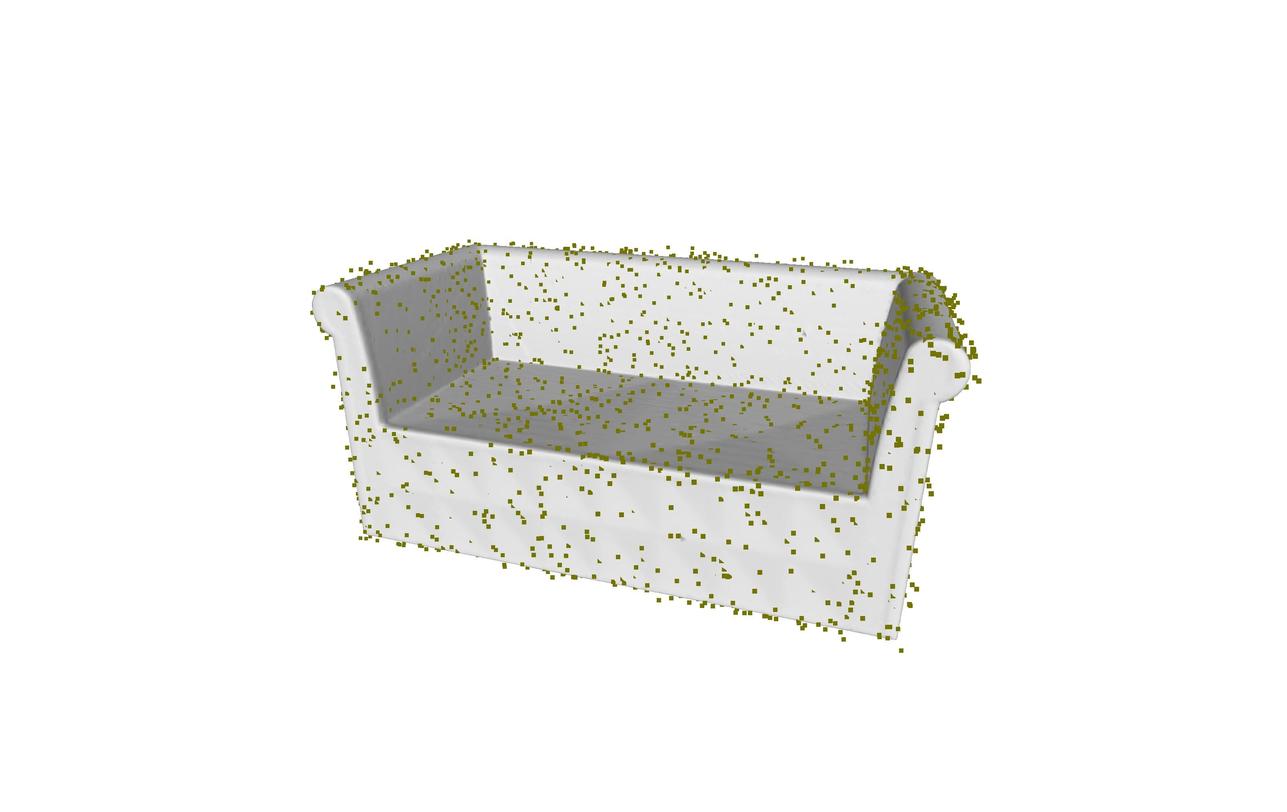}}
\subfloat{\includegraphics[width=\fitscale\tgtwidth, trim={180 40 210 180}, clip]{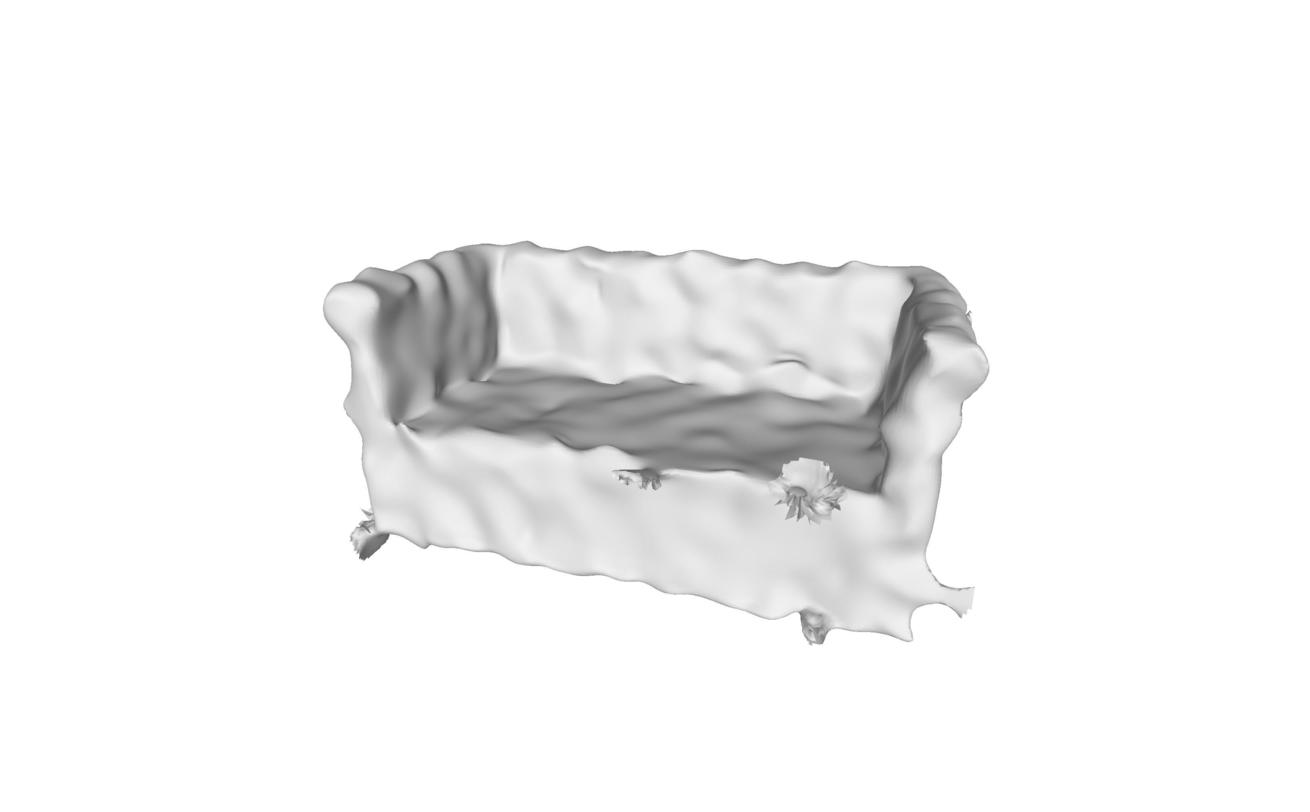}}
\subfloat{\includegraphics[width=\fitscale\tgtwidth, trim={180 40 210 180}, clip]{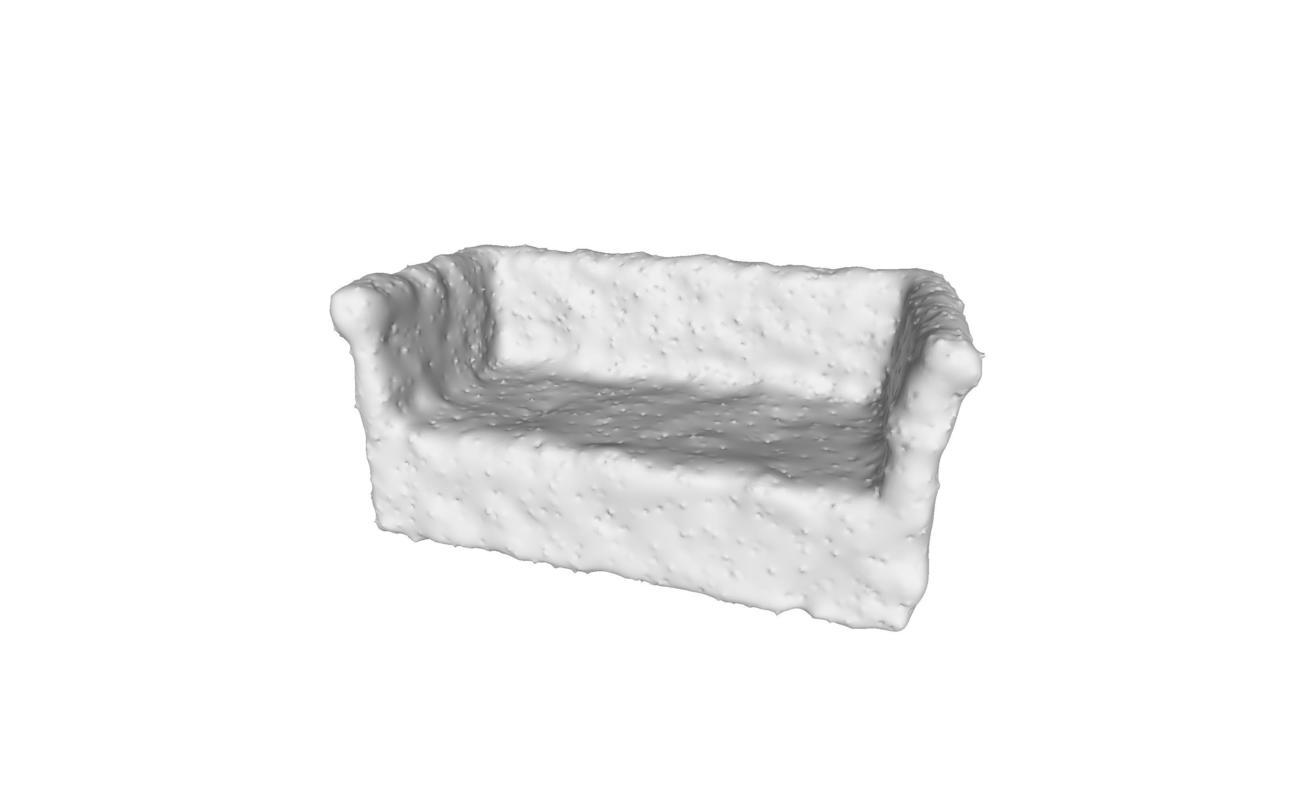}}
\subfloat{\includegraphics[width=\fitscale\tgtwidth, trim={180 40 210 180}, clip]{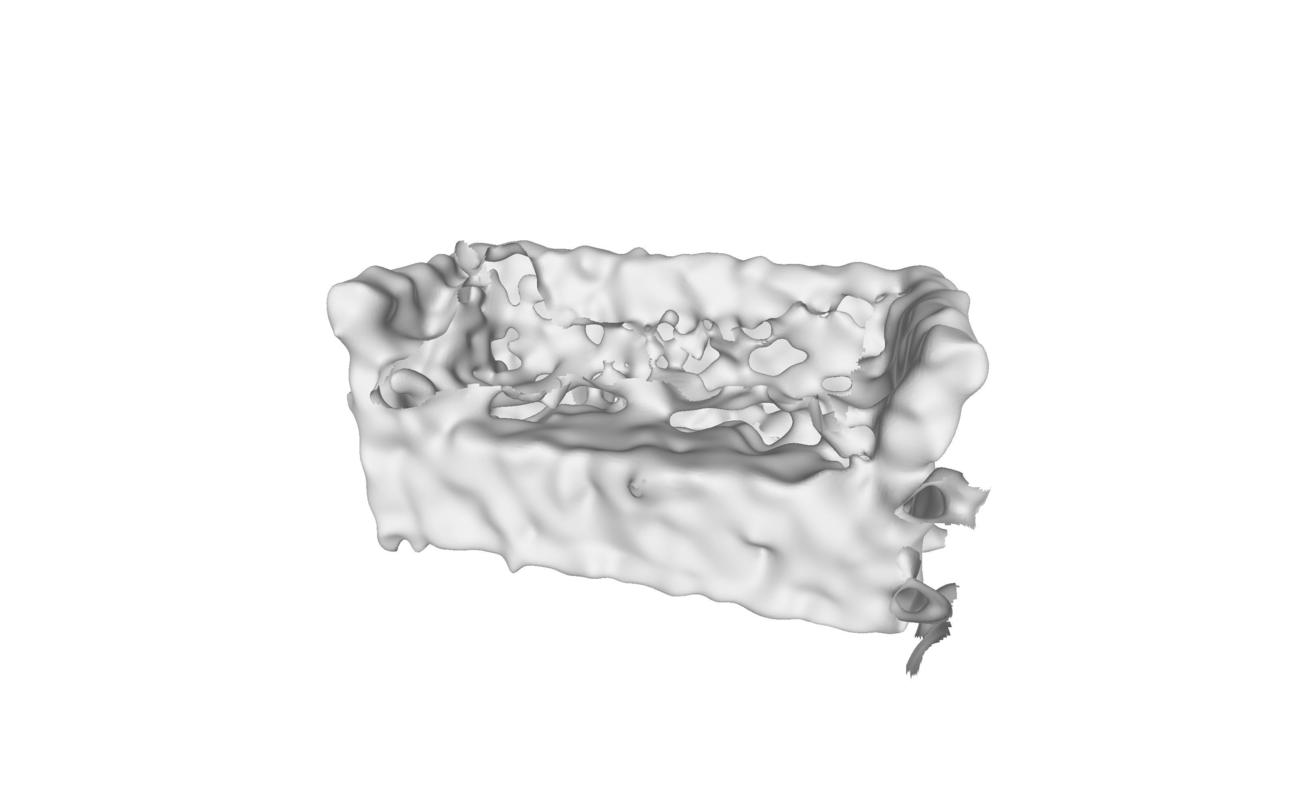}}
\subfloat{\includegraphics[width=\fitscale\tgtwidth, trim={180 40 210 180}, clip]{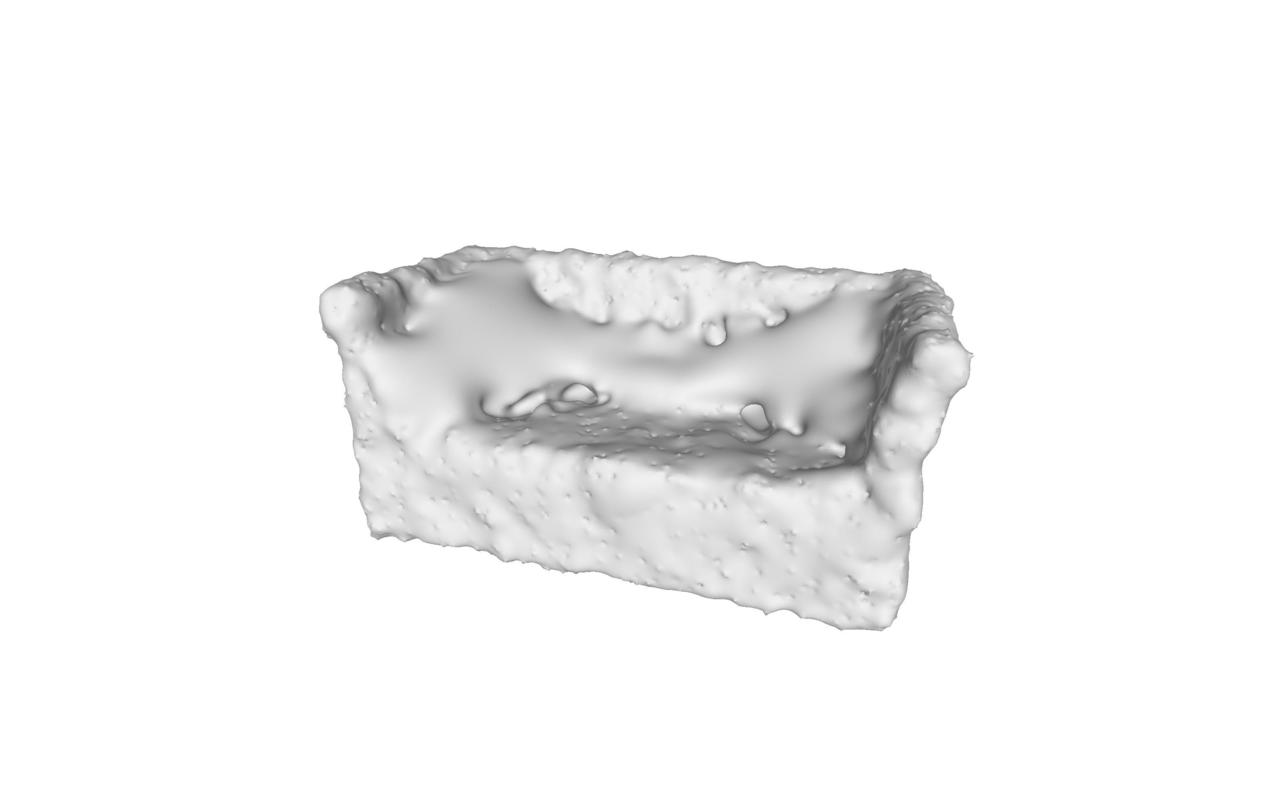}}
\subfloat{\includegraphics[width=\fitscale\tgtwidth, trim={180 40 210 180}, clip]{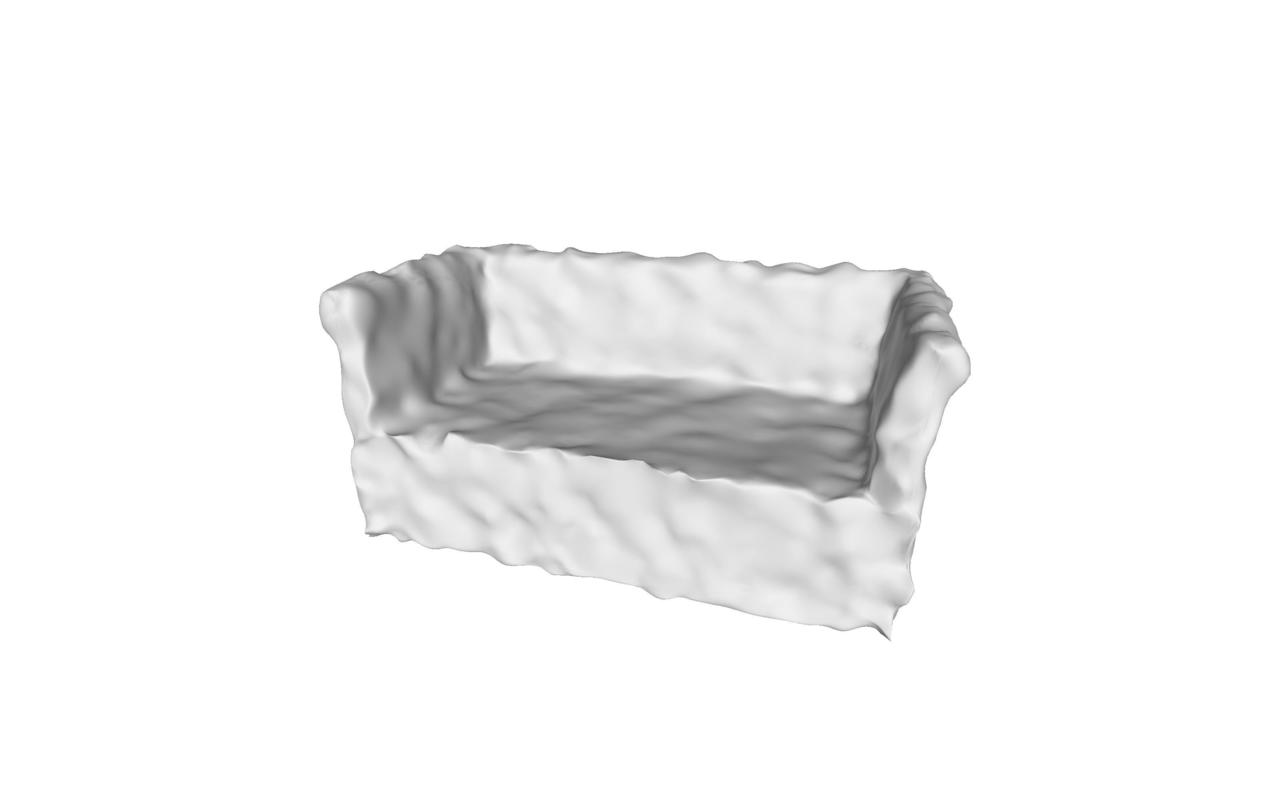}}
\subfloat{\includegraphics[width=\fitscale\tgtwidth, trim={180 40 210 180}, clip]{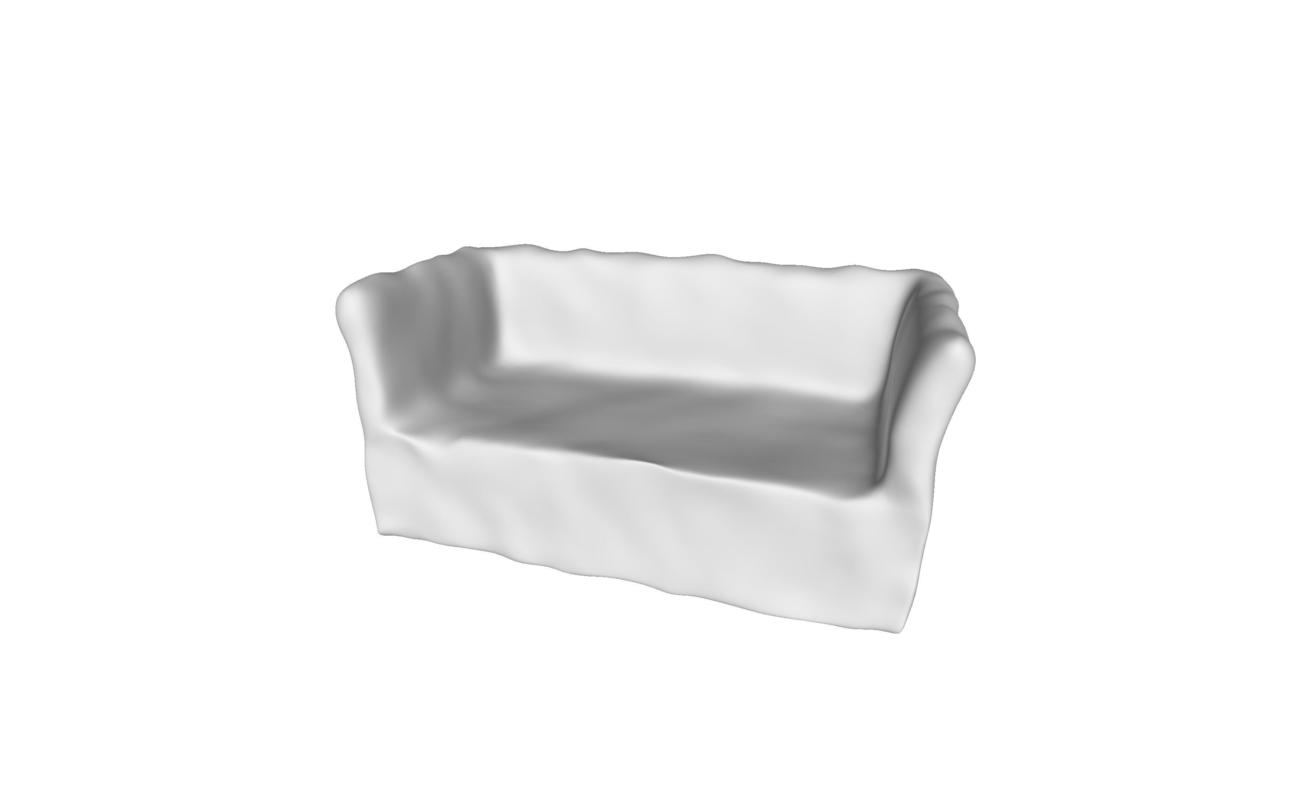}}
\subfloat{\includegraphics[width=\fitscale\tgtwidth, trim={180 40 210 180}, clip]{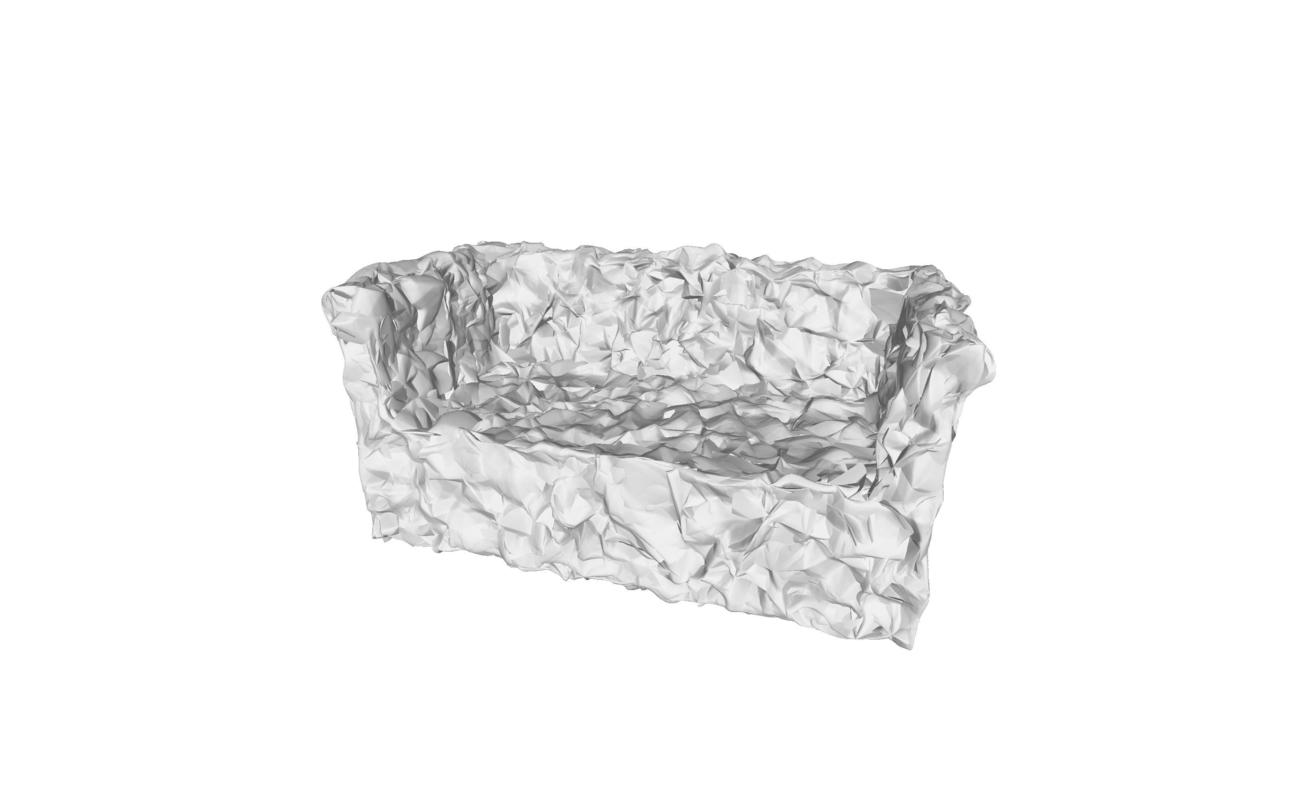}}
\subfloat{\includegraphics[width=\fitscale\tgtwidth, trim={180 40 210 180}, clip]{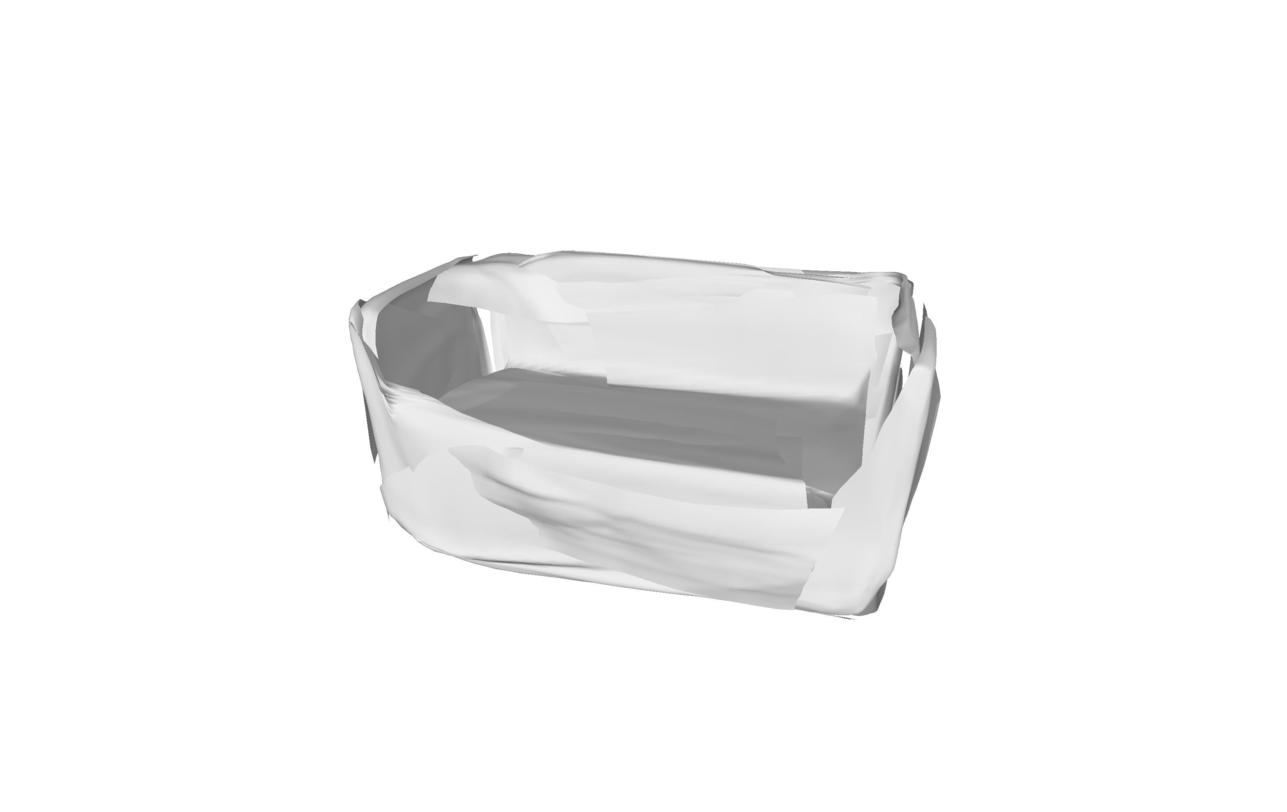}}
\subfloat{\includegraphics[width=\fitscale\tgtwidth, trim={180 40 210 180}, clip]{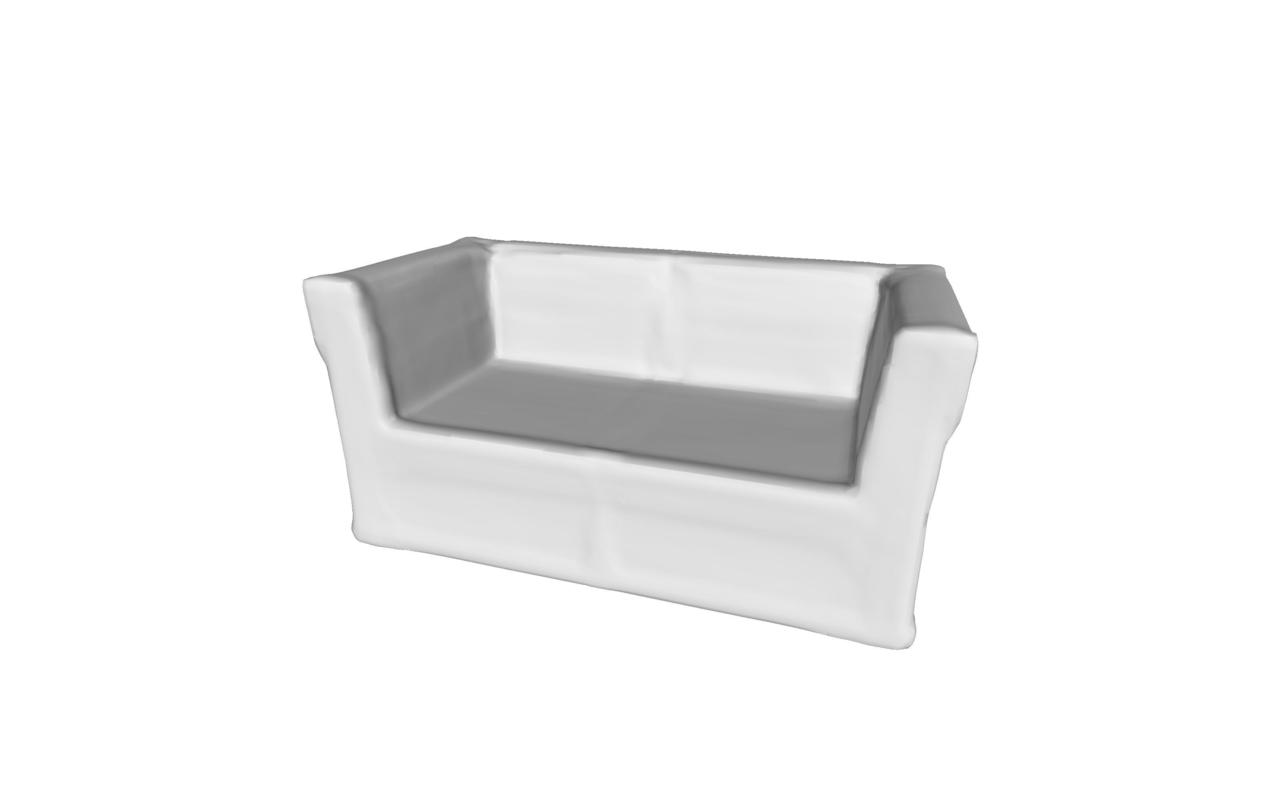}}
~
\subfloat{\includegraphics[width=\fitscale\tgtwidth, trim={180 40 210 180}, clip]{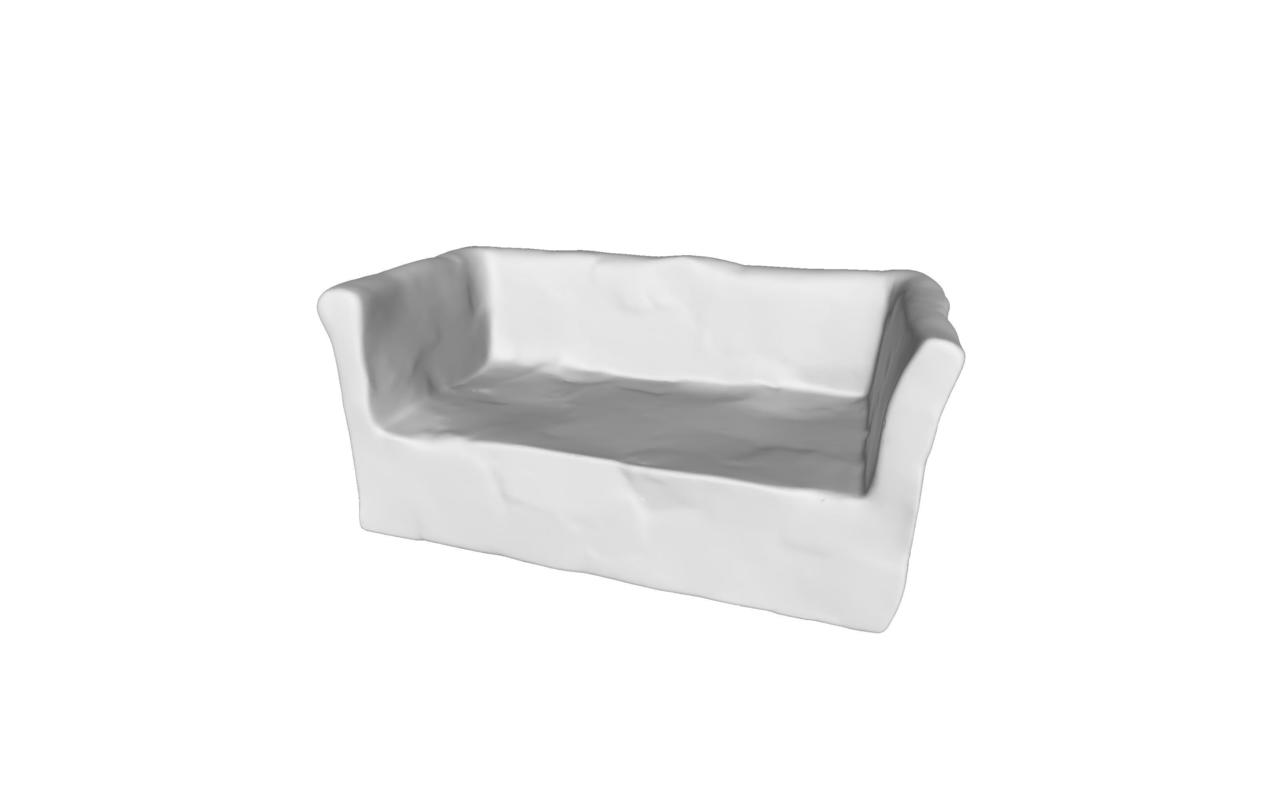}}
\\
\vspace{-5mm}
\subfloat{\includegraphics[width=\fitscale\tgtwidth, trim={188 80 190 140}, clip]{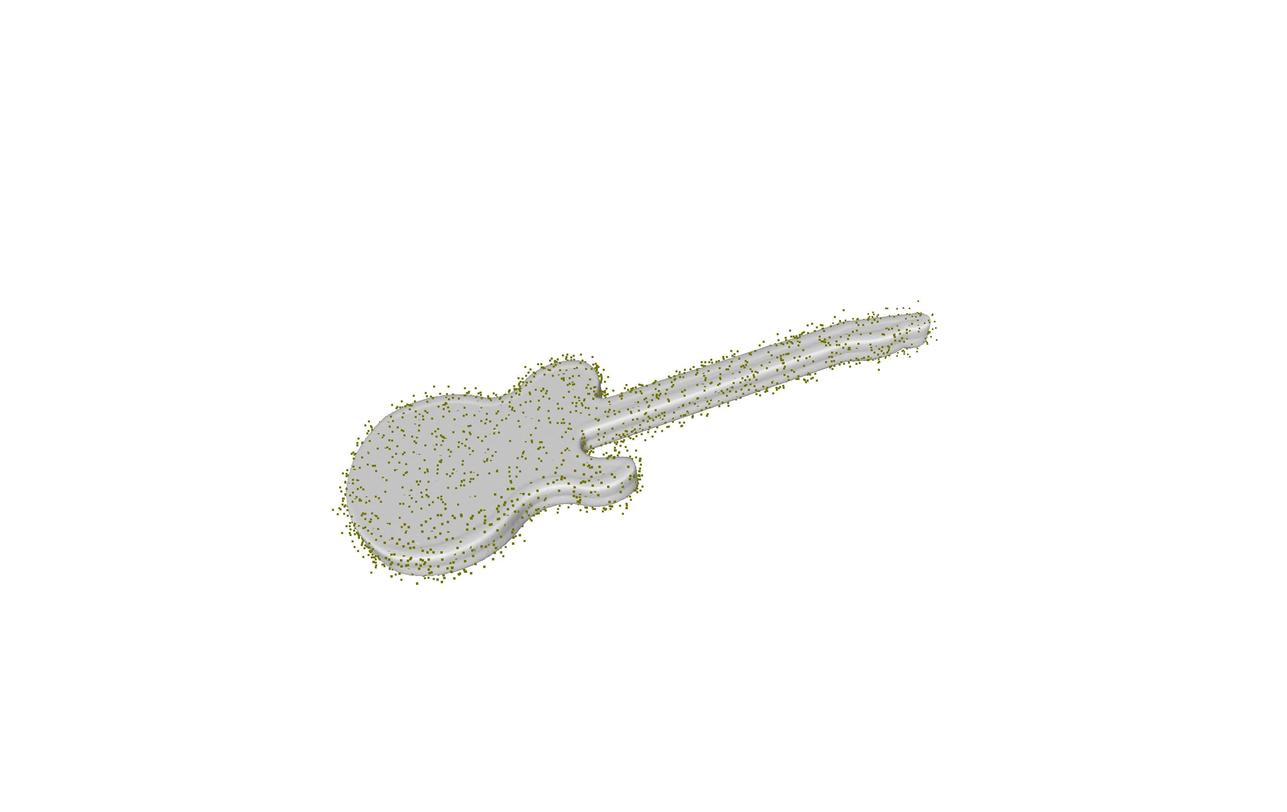}}
\subfloat{\includegraphics[width=\fitscale\tgtwidth, trim={188 80 190 140}, clip]{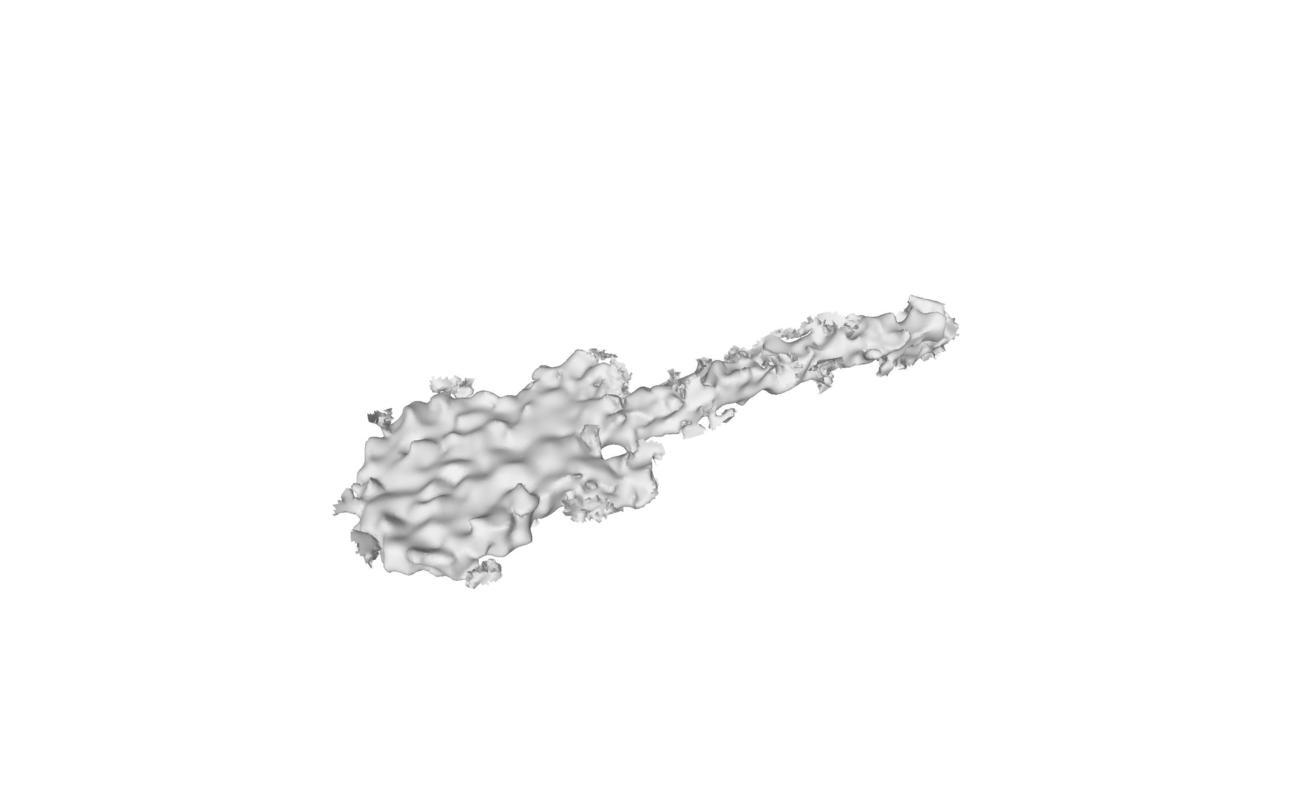}}
\subfloat{\includegraphics[width=\fitscale\tgtwidth, trim={188 80 190 140}, clip]{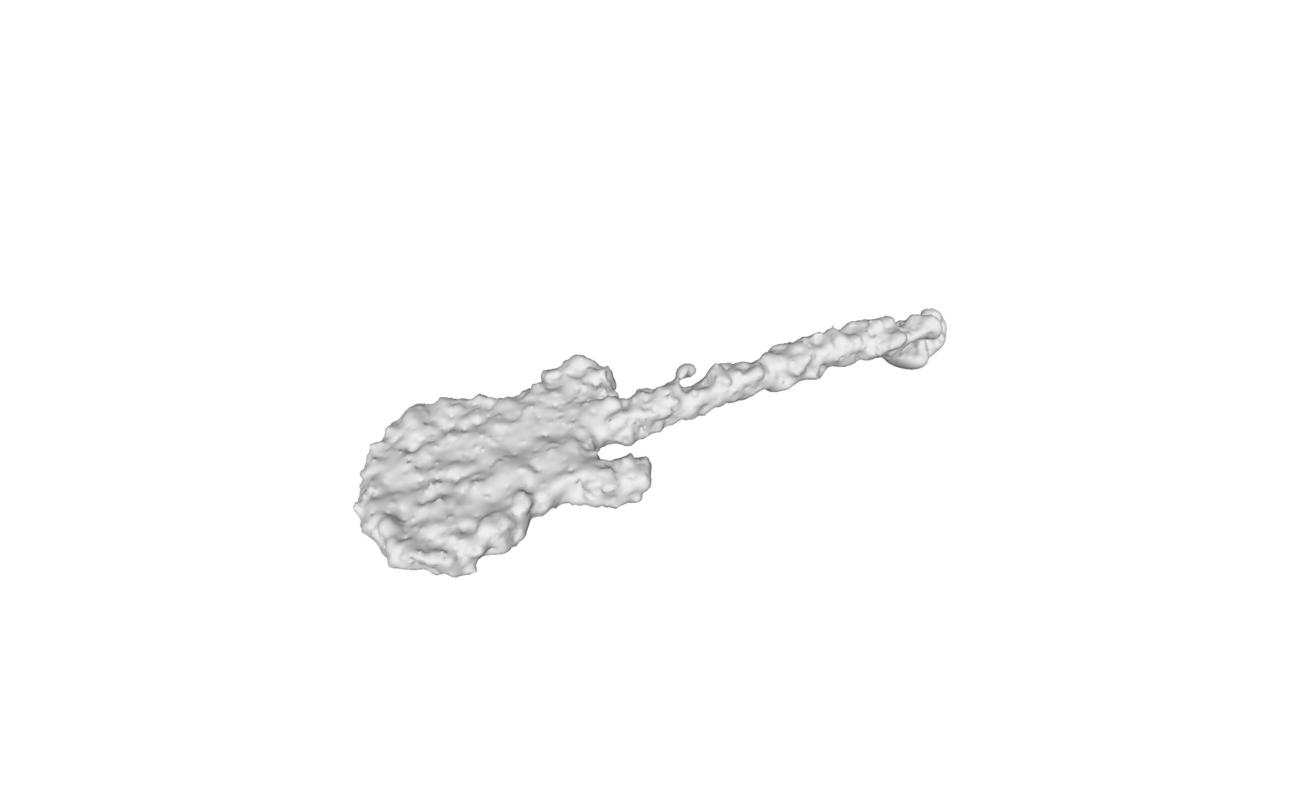}}
\subfloat{\includegraphics[width=\fitscale\tgtwidth, trim={188 80 190 140}, clip]{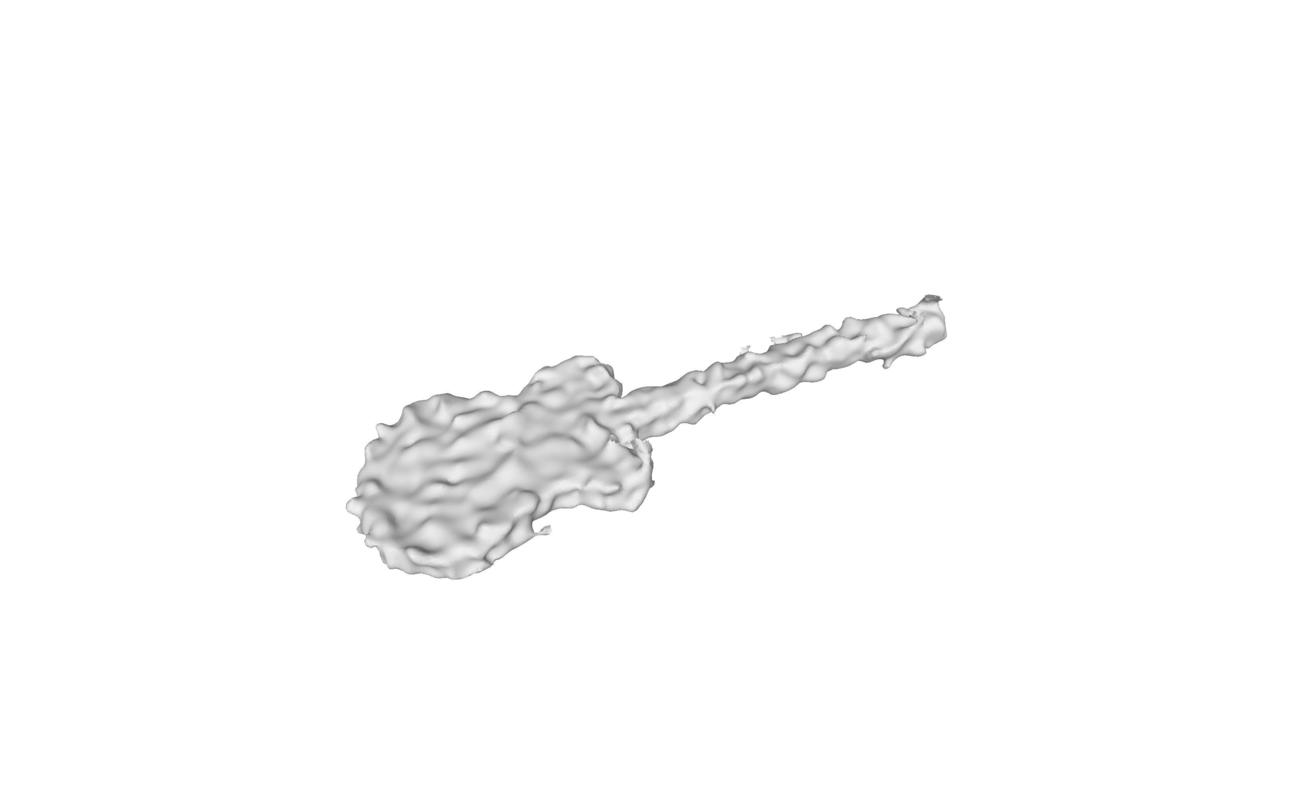}}
\subfloat{\includegraphics[width=\fitscale\tgtwidth, trim={188 80 190 140}, clip]{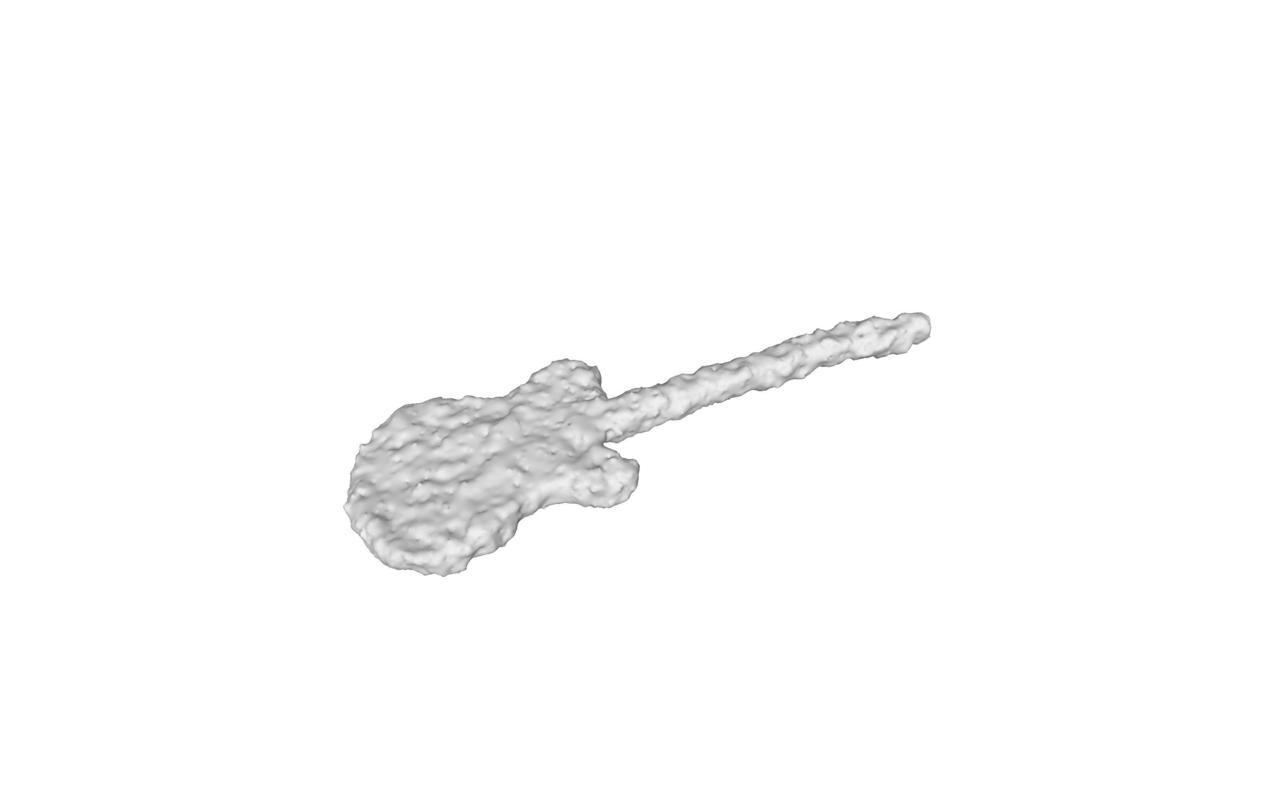}}
\subfloat{\includegraphics[width=\fitscale\tgtwidth, trim={188 80 190 140}, clip]{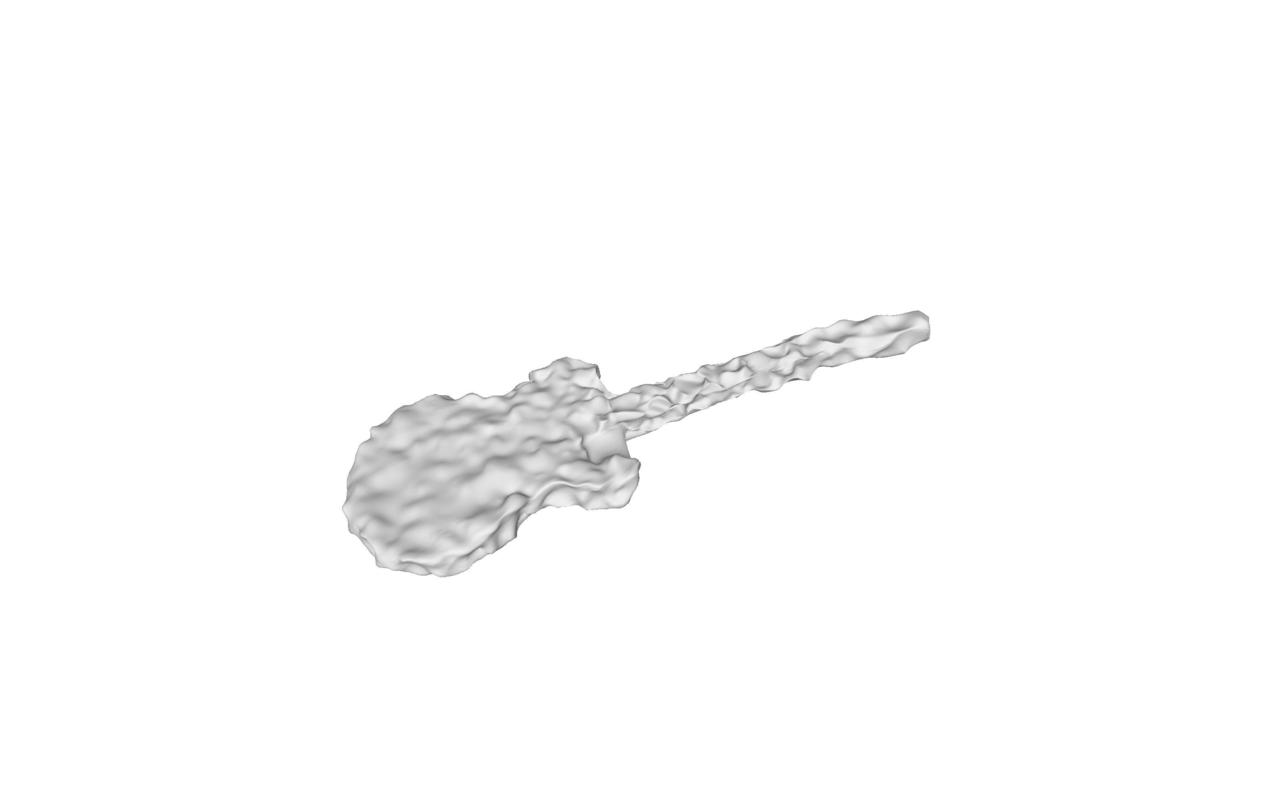}}
\subfloat{\includegraphics[width=\fitscale\tgtwidth, trim={188 80 190 140}, clip]{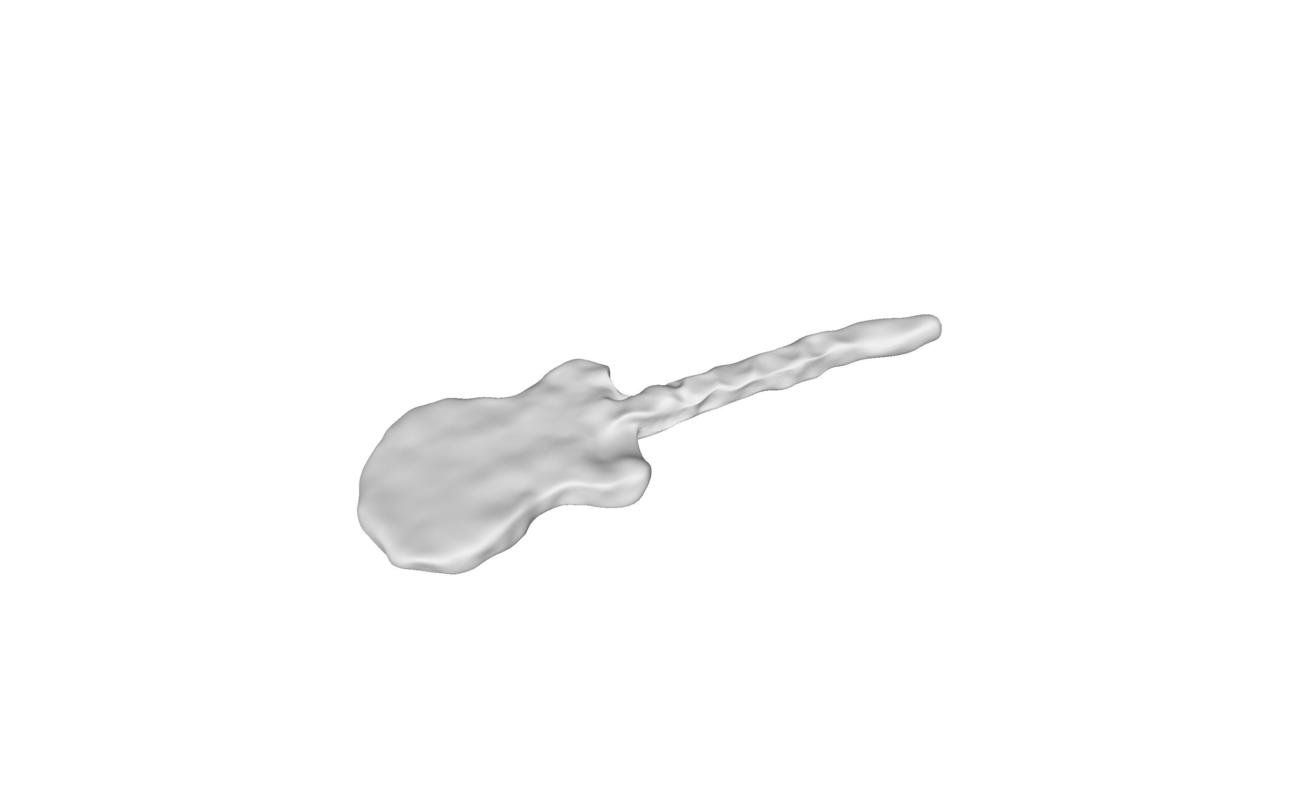}}
\subfloat{\includegraphics[width=\fitscale\tgtwidth, trim={188 80 190 140}, clip]{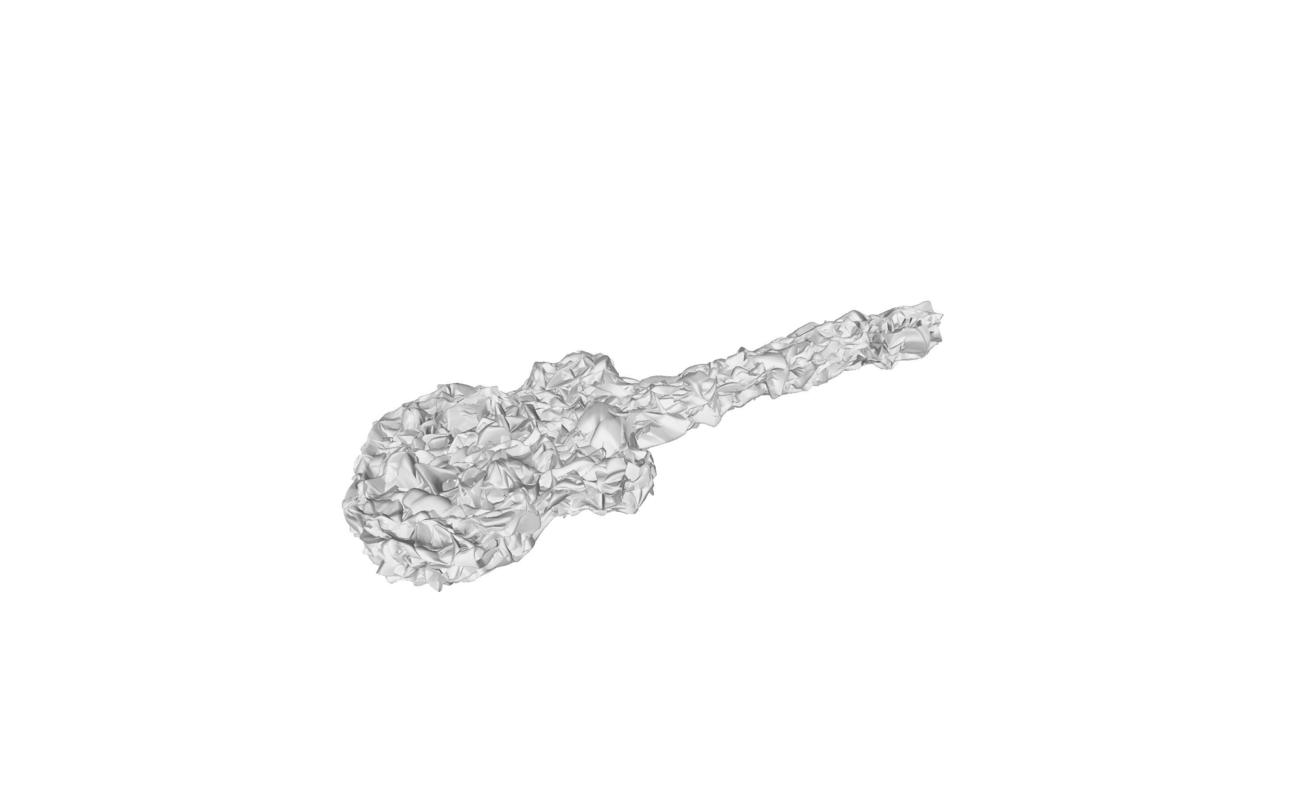}}
\subfloat{\includegraphics[width=\fitscale\tgtwidth, trim={188 80 190 140}, clip]{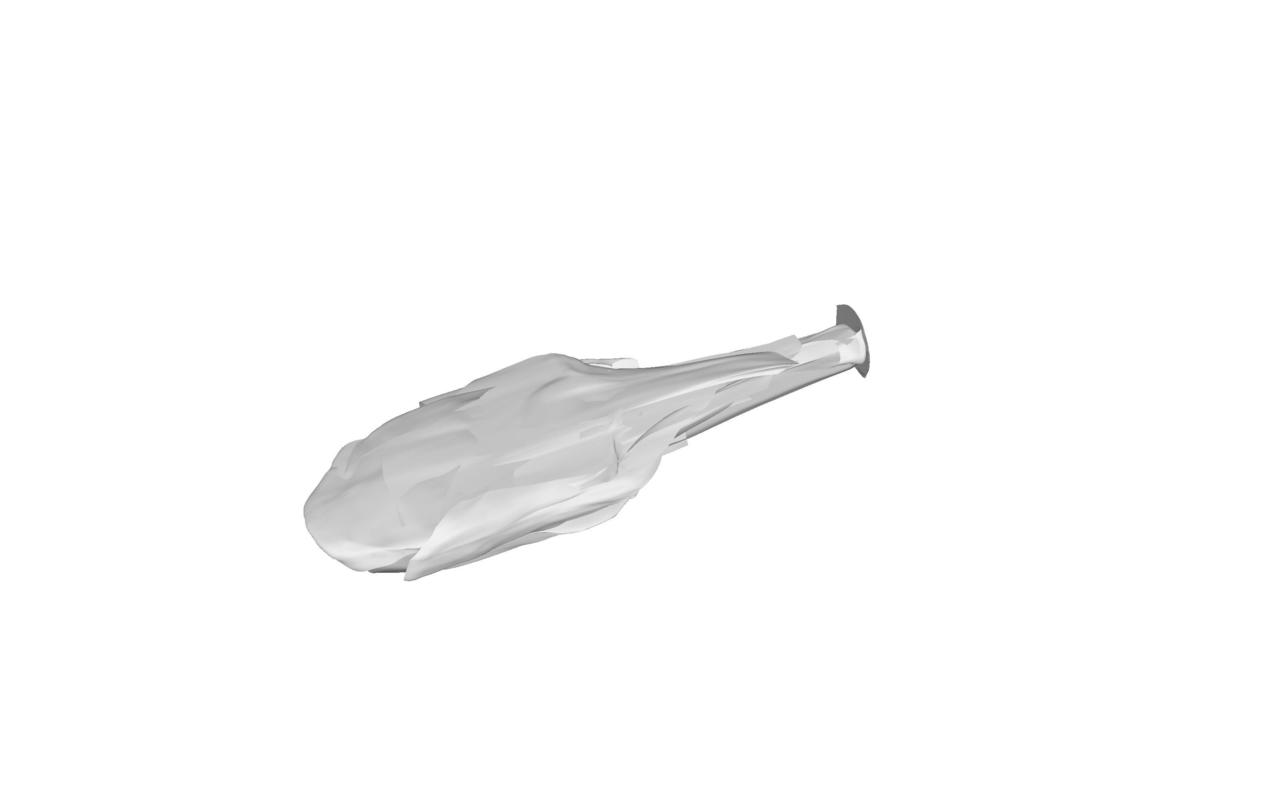}}
\subfloat{\includegraphics[width=\fitscale\tgtwidth, trim={188 80 190 140}, clip]{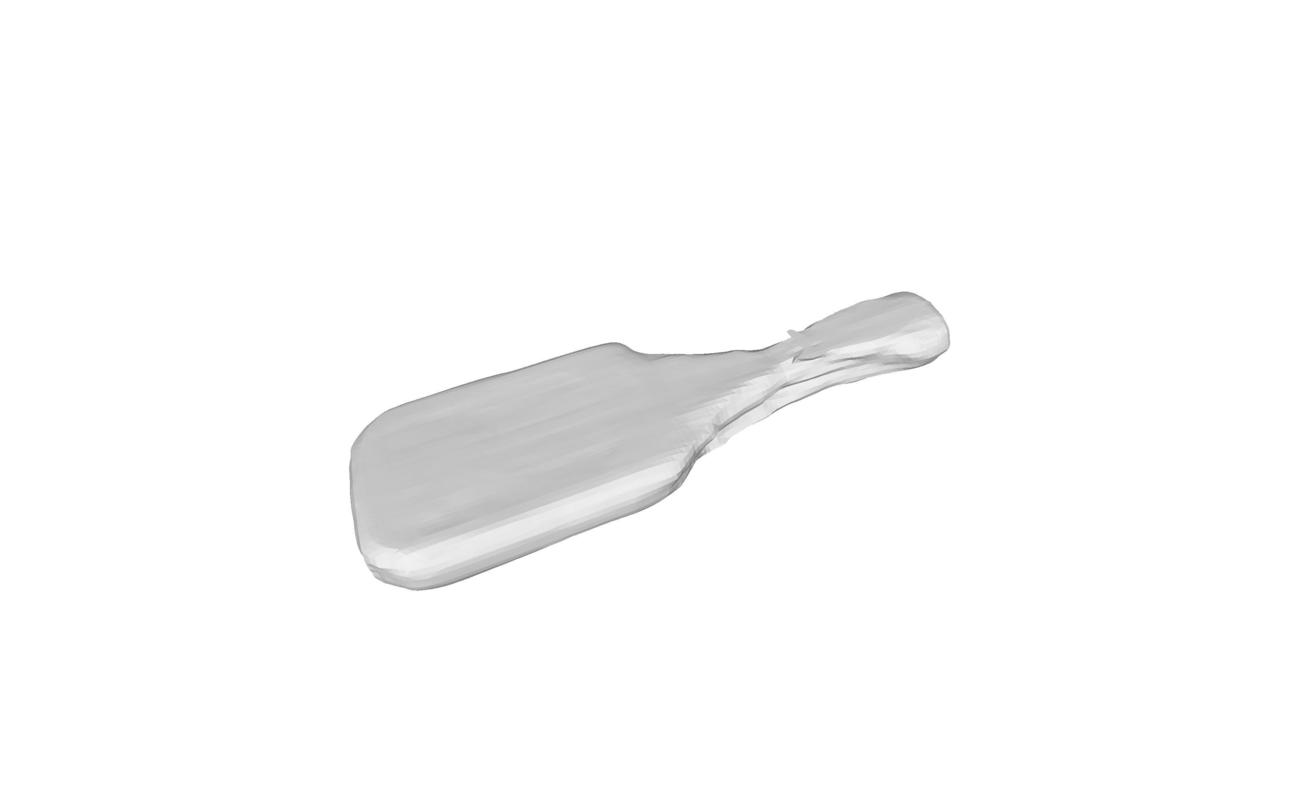}}
~
\subfloat{\includegraphics[width=\fitscale\tgtwidth, trim={188 80 190 140}, clip]{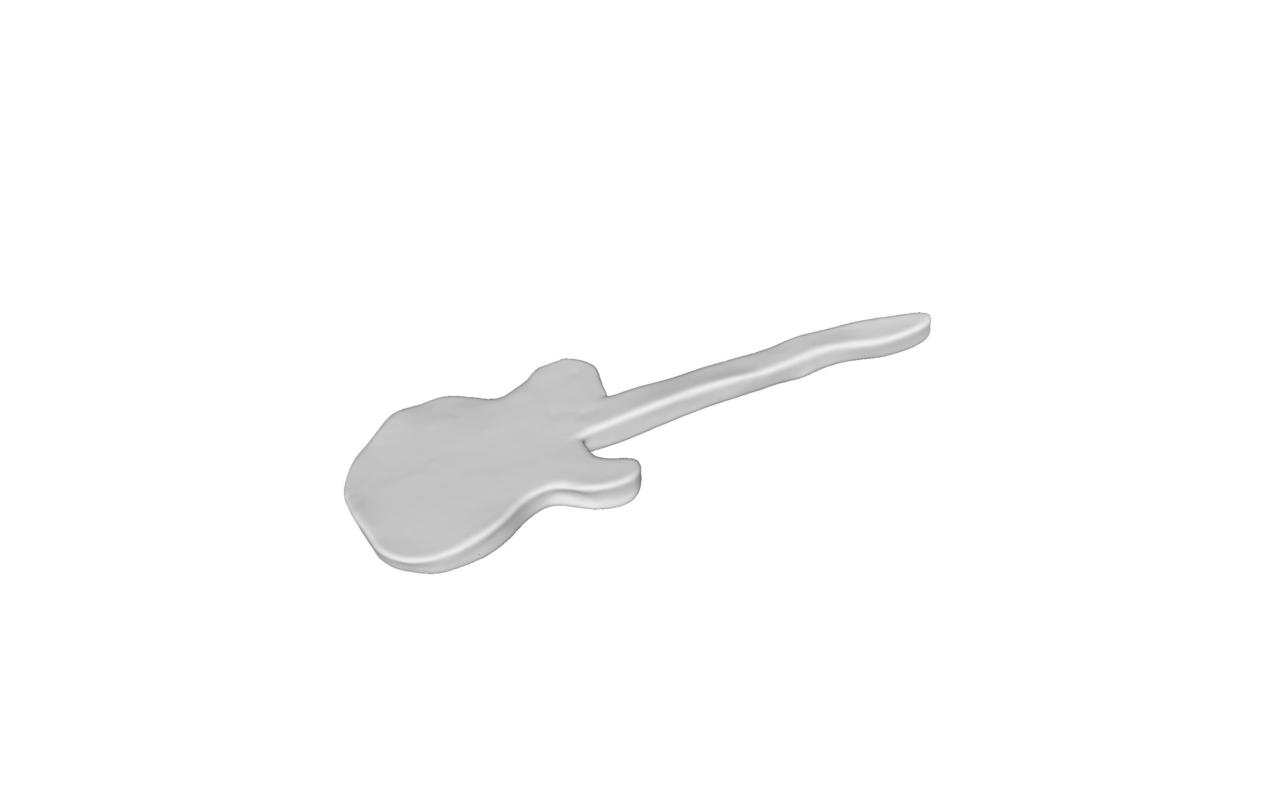}}
\\
\vspace{-5mm}
\subfloat{\includegraphics[width=\fitscale\tgtwidth, trim={188 80 190 80}, clip]{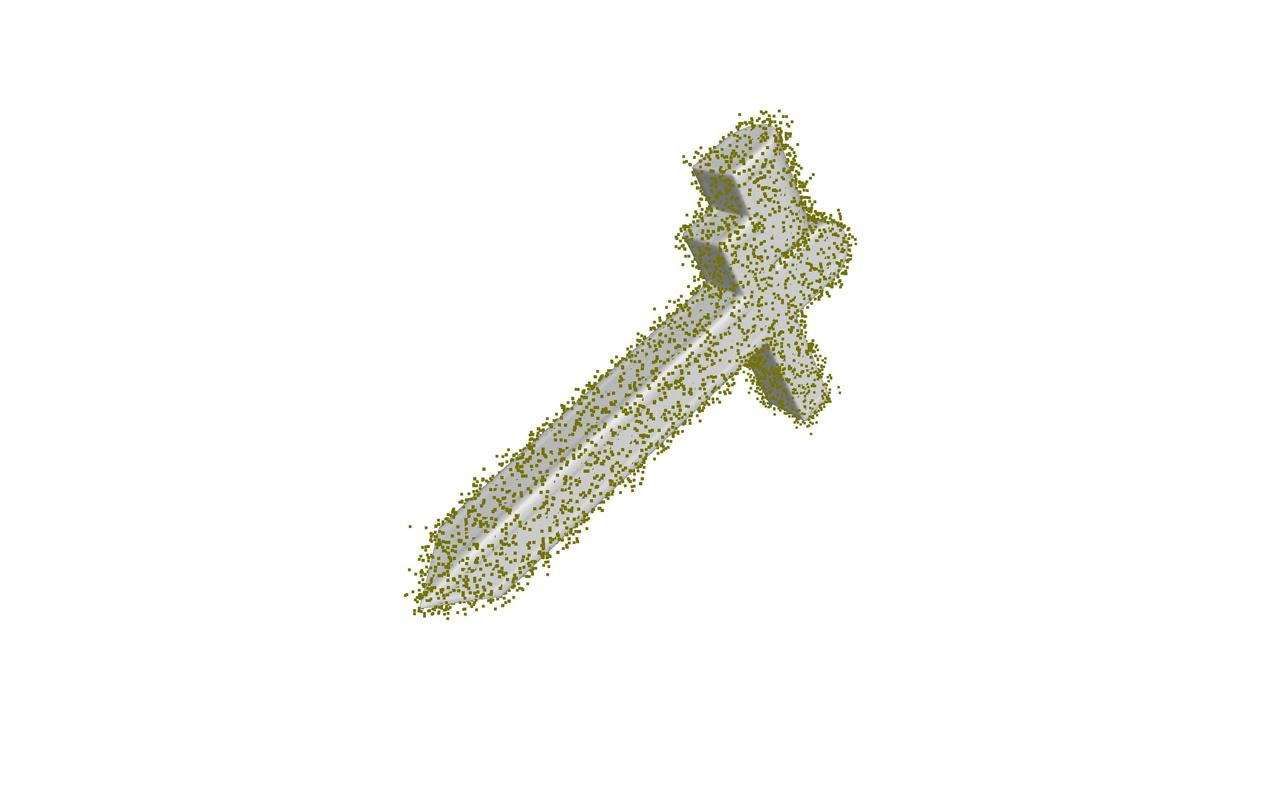}}
\subfloat{\includegraphics[width=\fitscale\tgtwidth, trim={188 80 190 80}, clip]{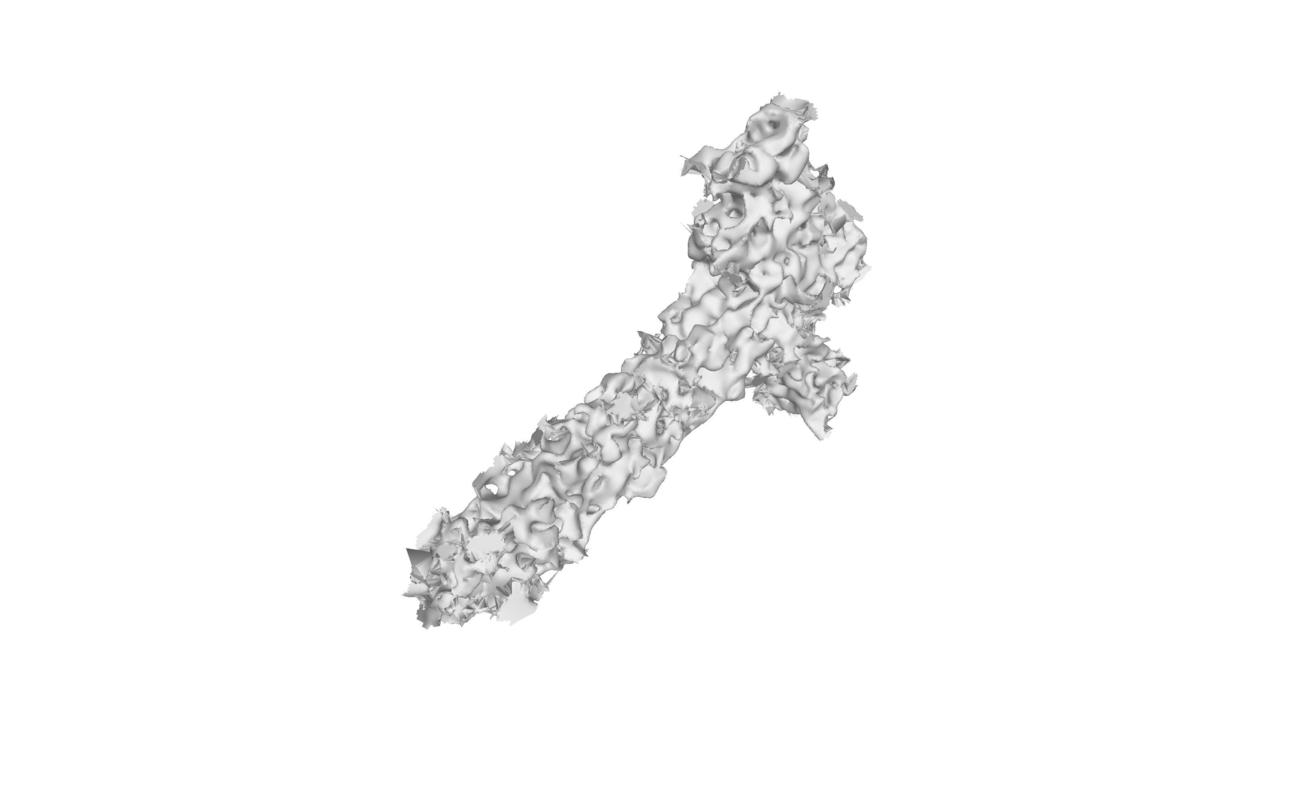}}
\subfloat{\includegraphics[width=\fitscale\tgtwidth, trim={188 80 190 80}, clip]{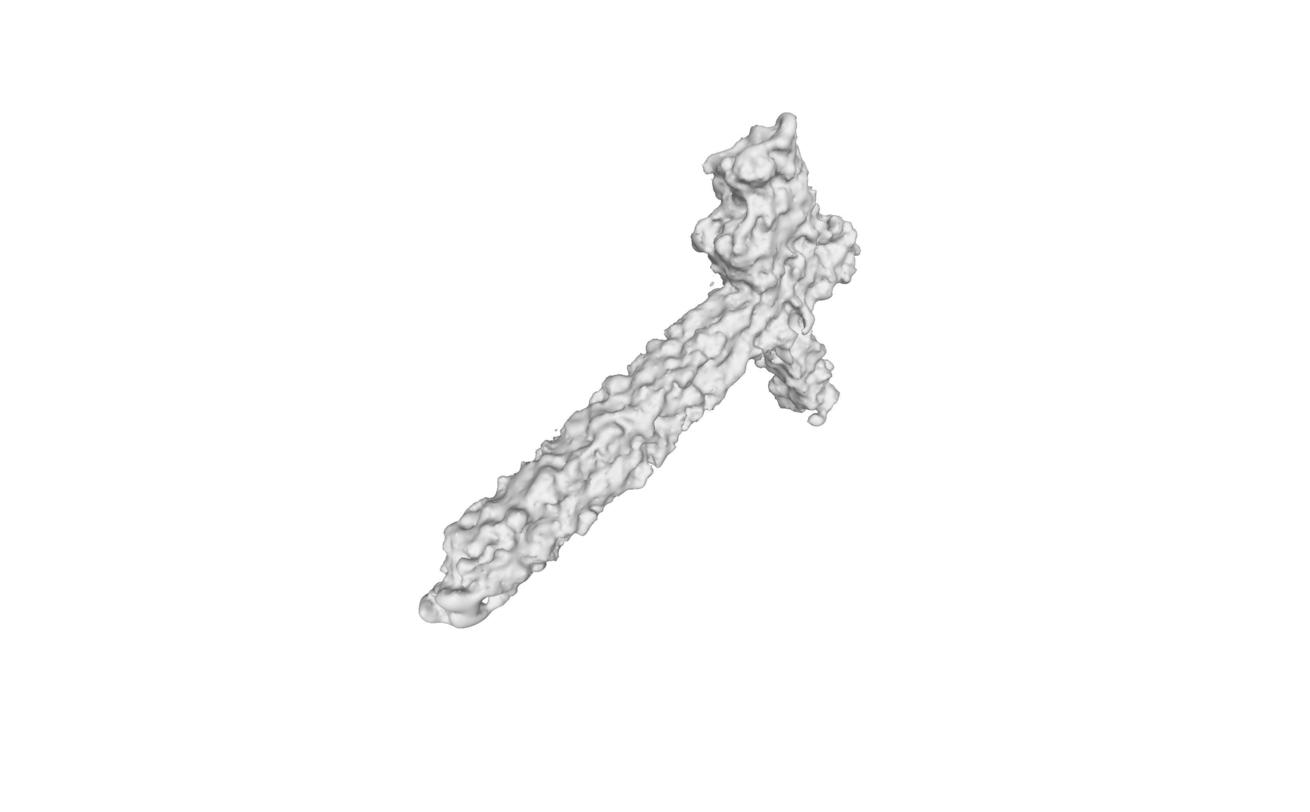}}
\subfloat{\includegraphics[width=\fitscale\tgtwidth, trim={188 80 190 80}, clip]{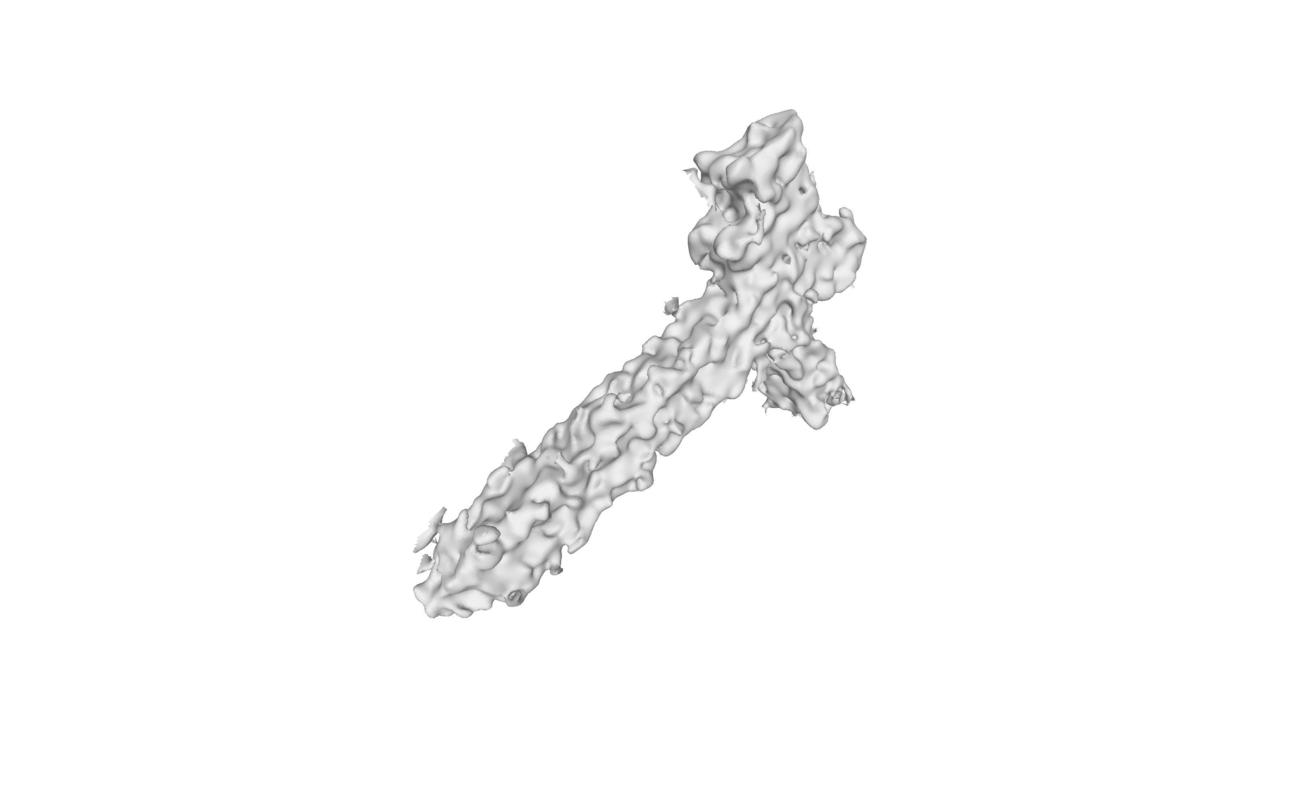}}
\subfloat{\includegraphics[width=\fitscale\tgtwidth, trim={188 80 190 80}, clip]{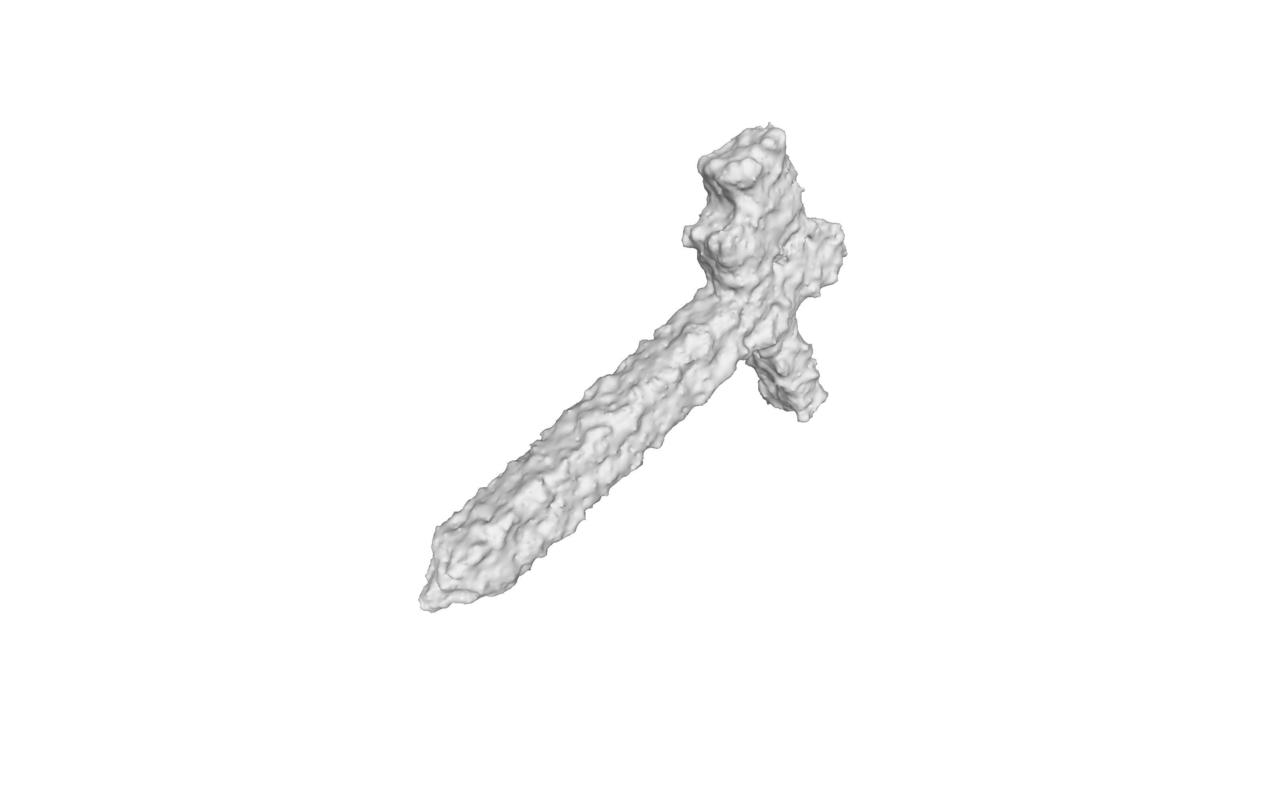}}
\subfloat{\includegraphics[width=\fitscale\tgtwidth, trim={188 80 190 80}, clip]{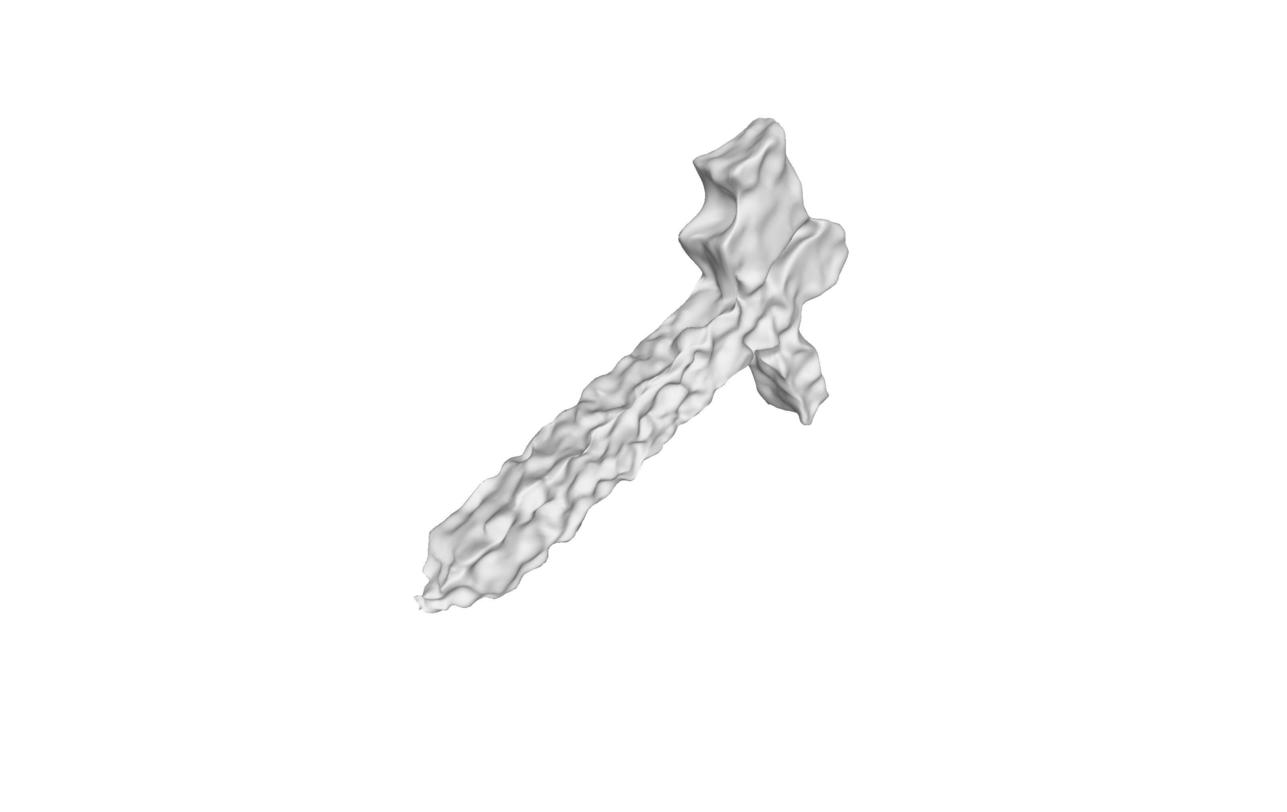}}
\subfloat{\includegraphics[width=\fitscale\tgtwidth, trim={188 80 190 80}, clip]{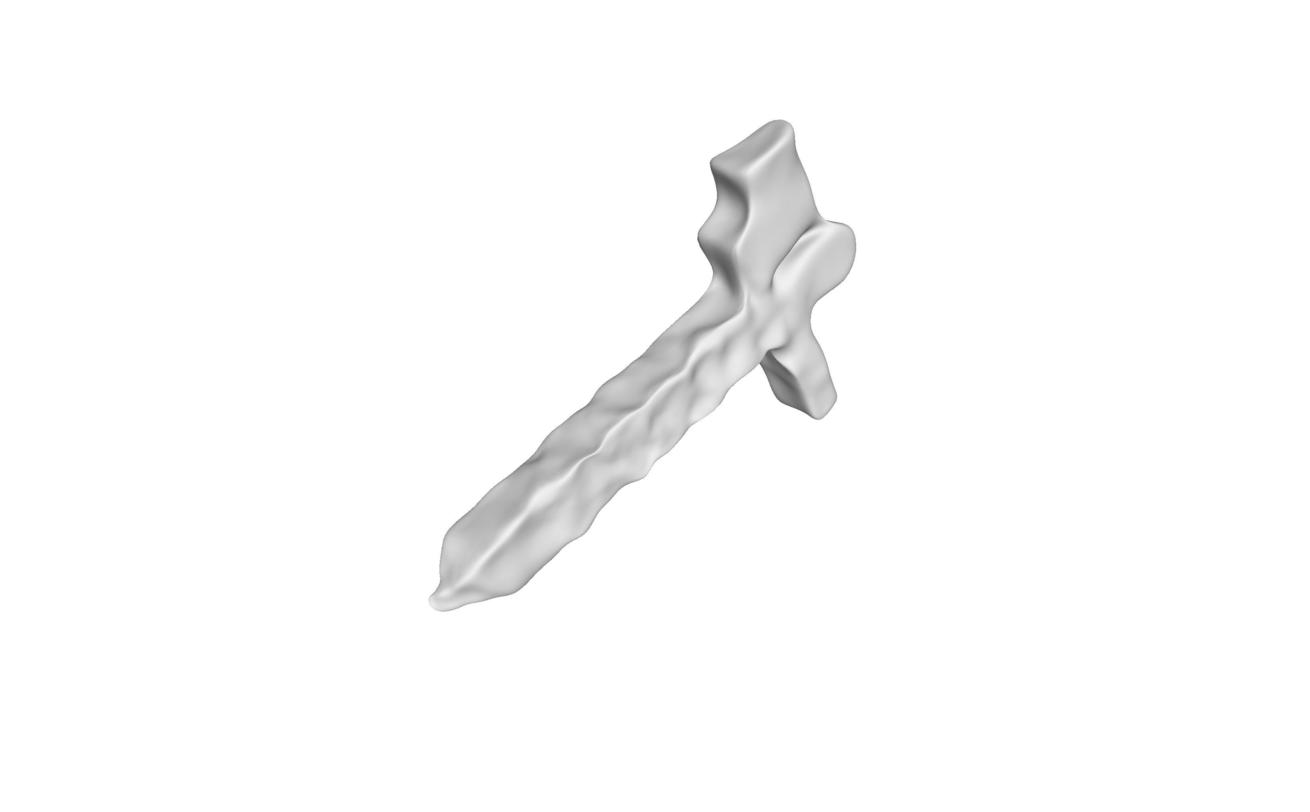}}
\subfloat{\includegraphics[width=\fitscale\tgtwidth, trim={188 80 190 80}, clip]{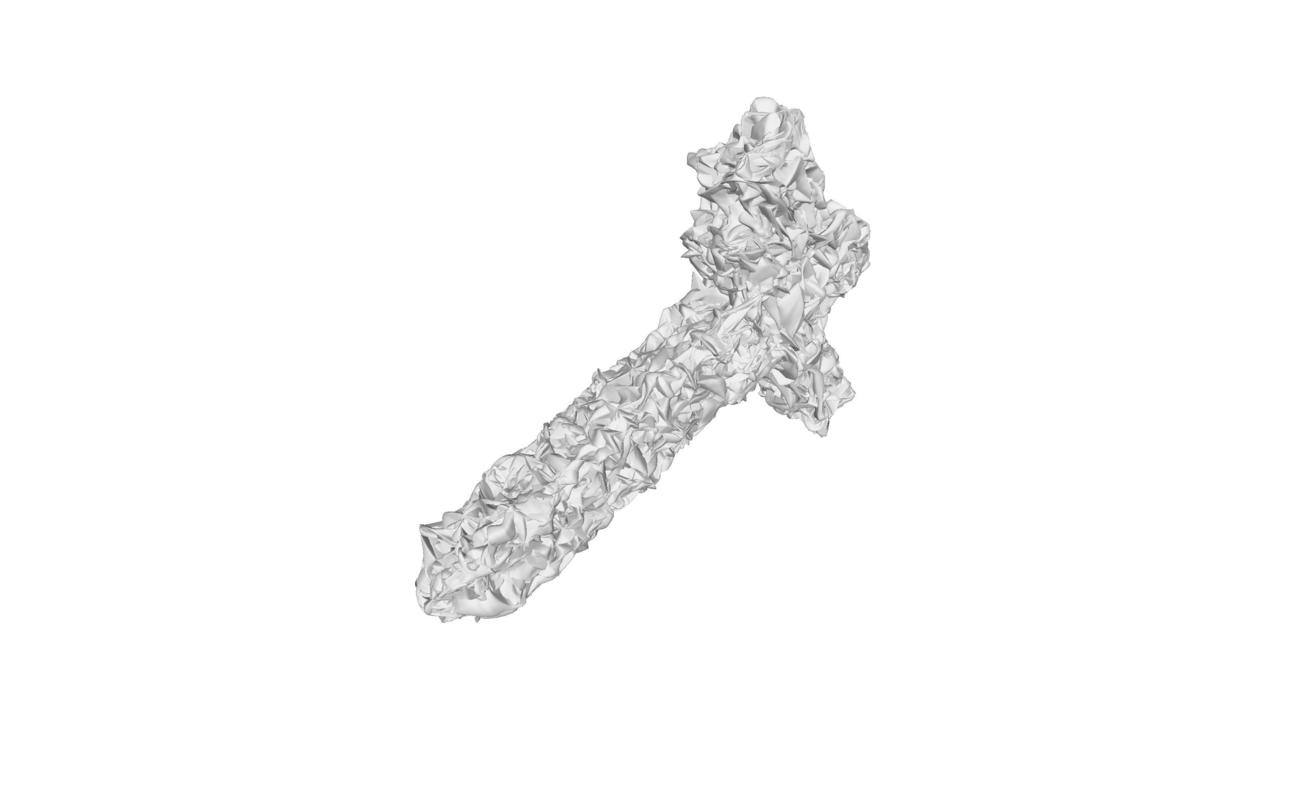}}
\subfloat{\includegraphics[width=\fitscale\tgtwidth, trim={188 80 190 80}, clip]{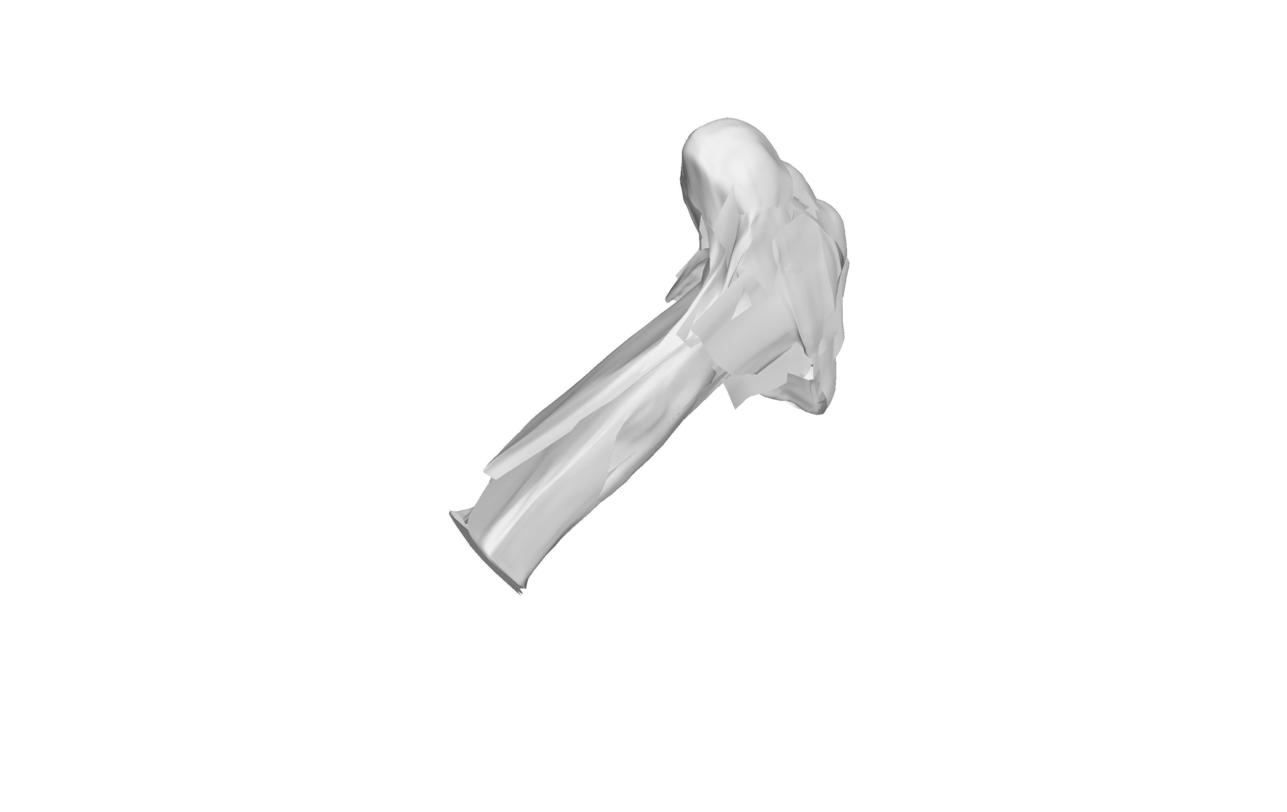}}
\subfloat{\includegraphics[width=\fitscale\tgtwidth, trim={188 80 190 80}, clip]{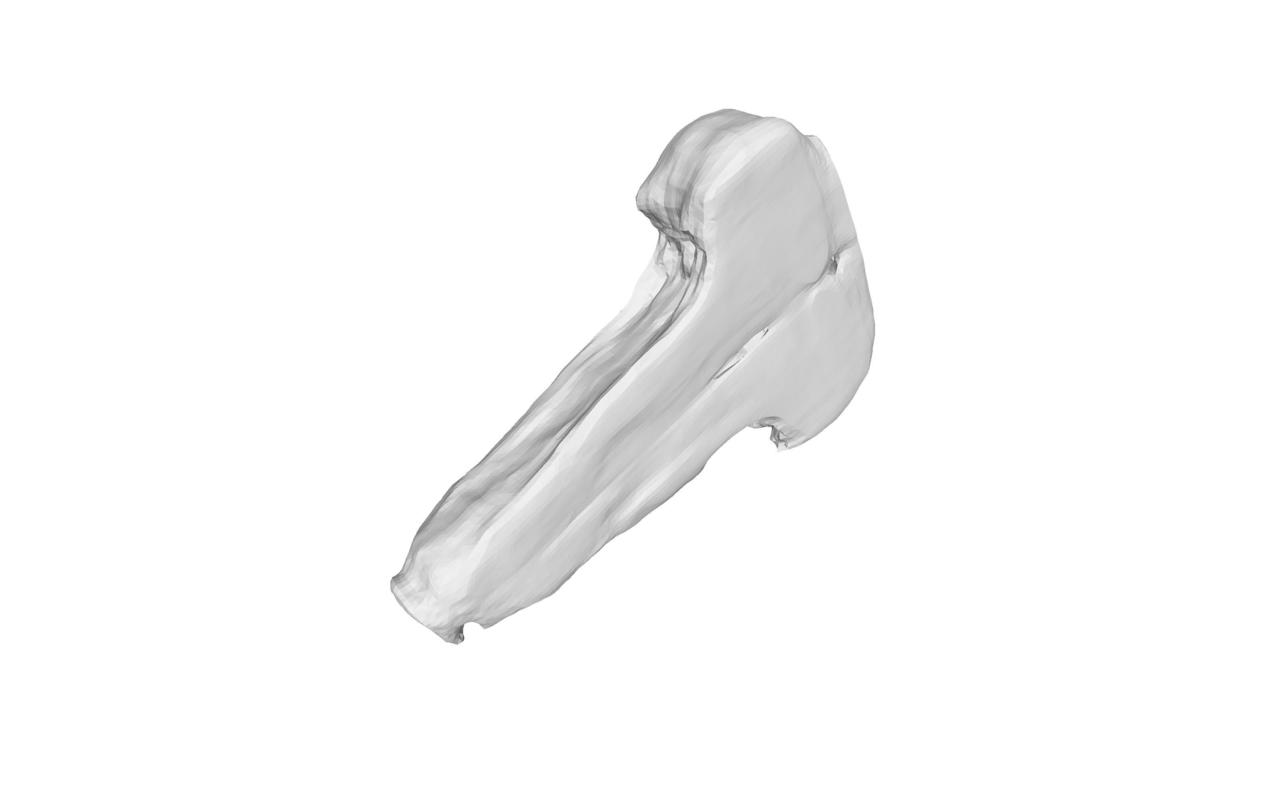}}
~
\subfloat{\includegraphics[width=\fitscale\tgtwidth, trim={188 80 190 80}, clip]{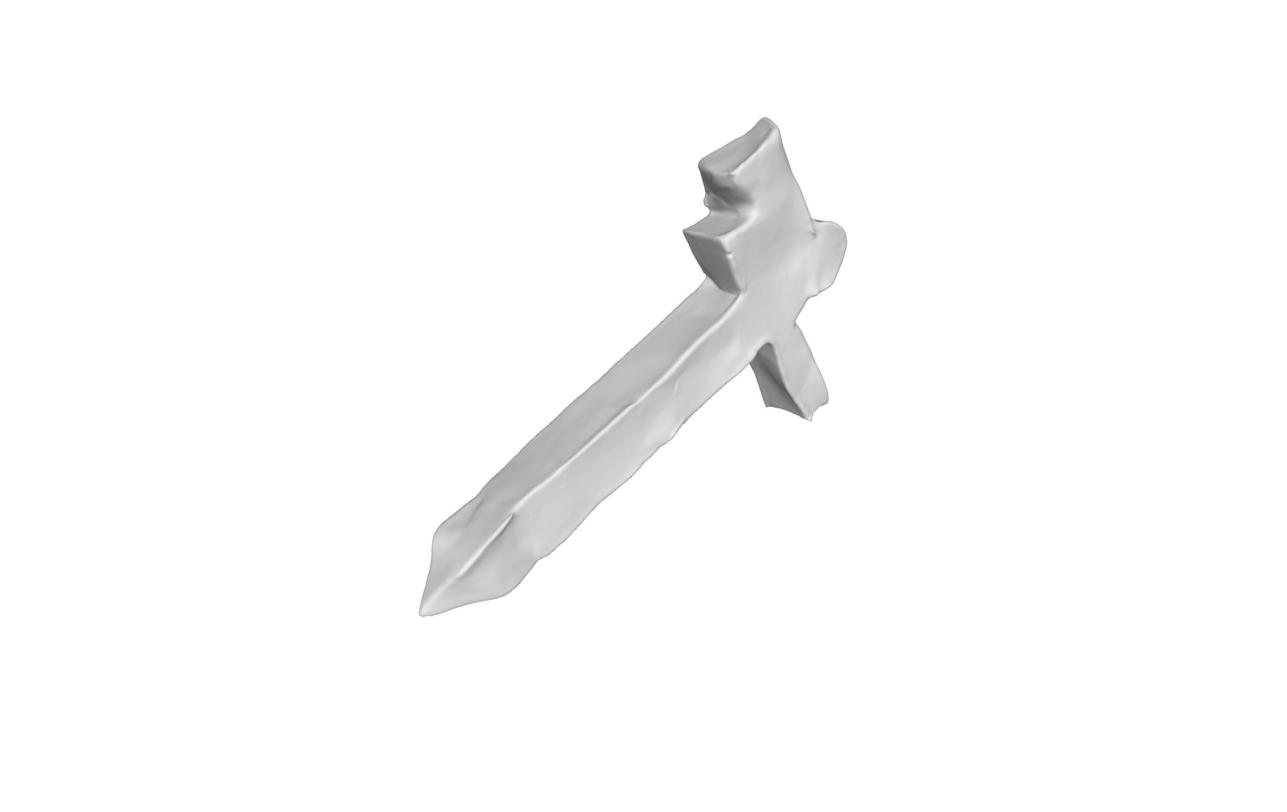}}
\\
\vspace{-5mm}
\subfloat{\includegraphics[width=\fitscale\tgtwidth, trim={188 80 190 96}, clip]{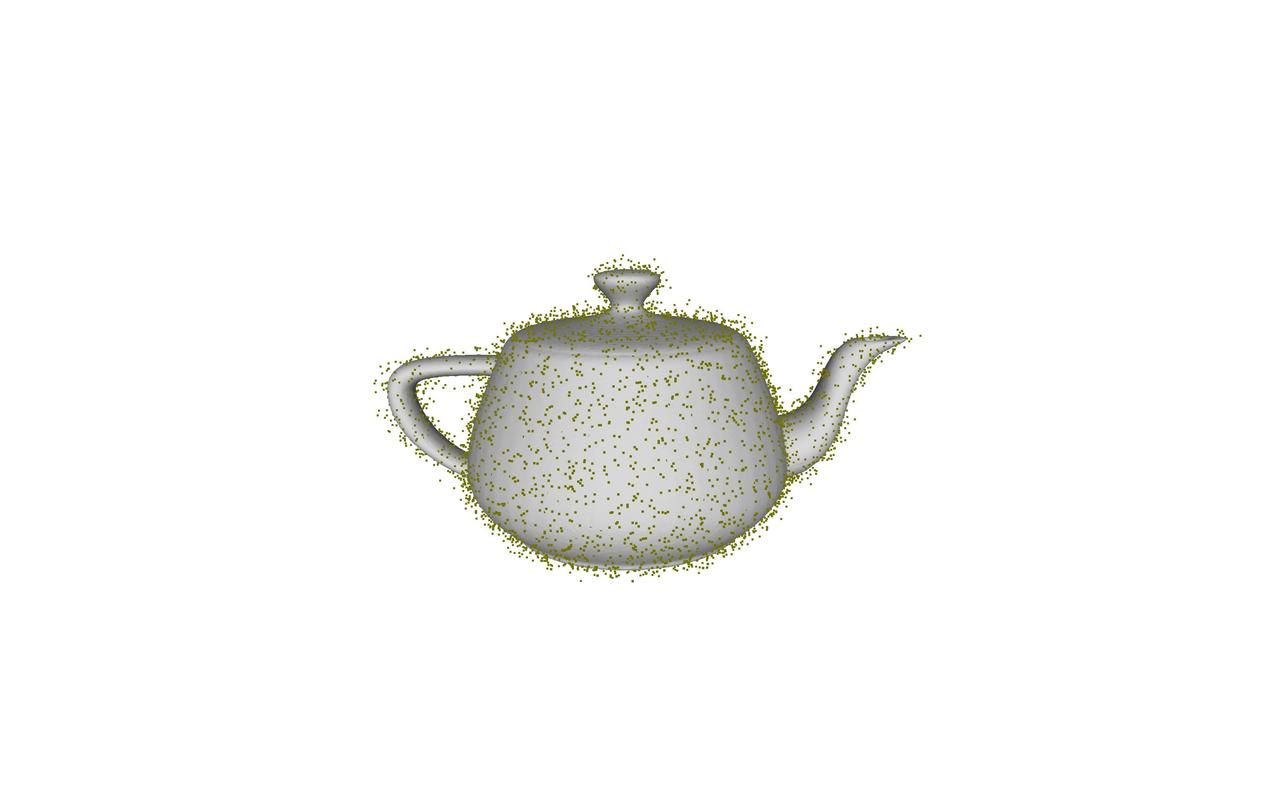}}
\subfloat{\includegraphics[width=\fitscale\tgtwidth, trim={188 80 190 96}, clip]{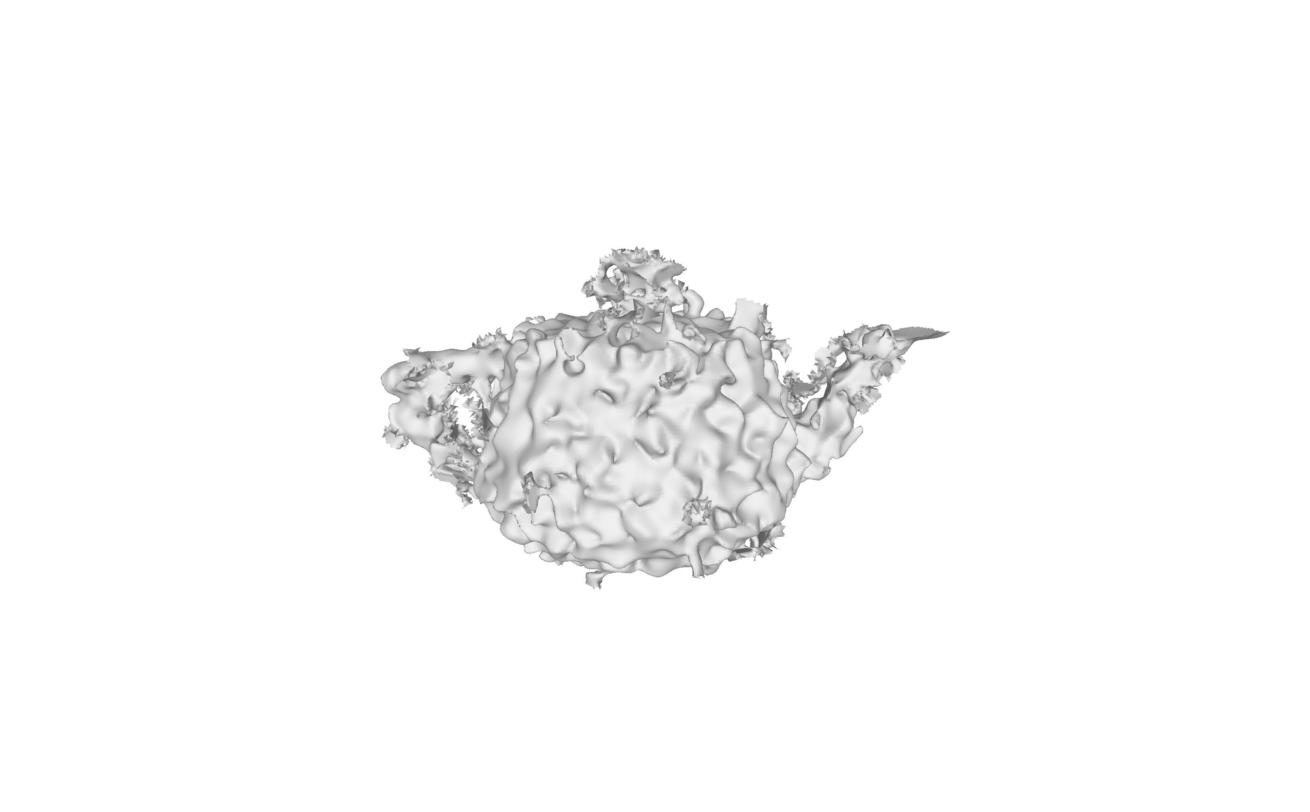}}
\subfloat{\includegraphics[width=\fitscale\tgtwidth, trim={188 80 190 96}, clip]{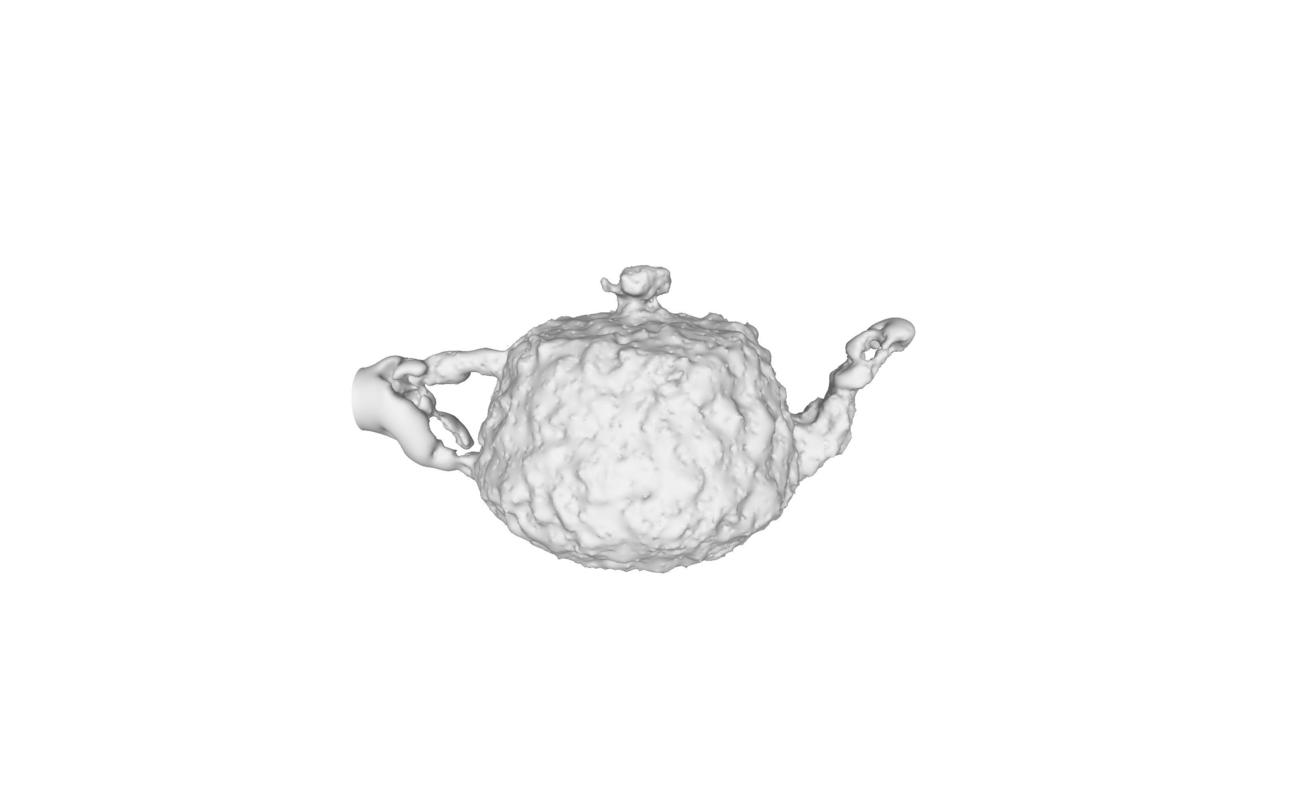}}
\subfloat{\includegraphics[width=\fitscale\tgtwidth, trim={188 80 190 96}, clip]{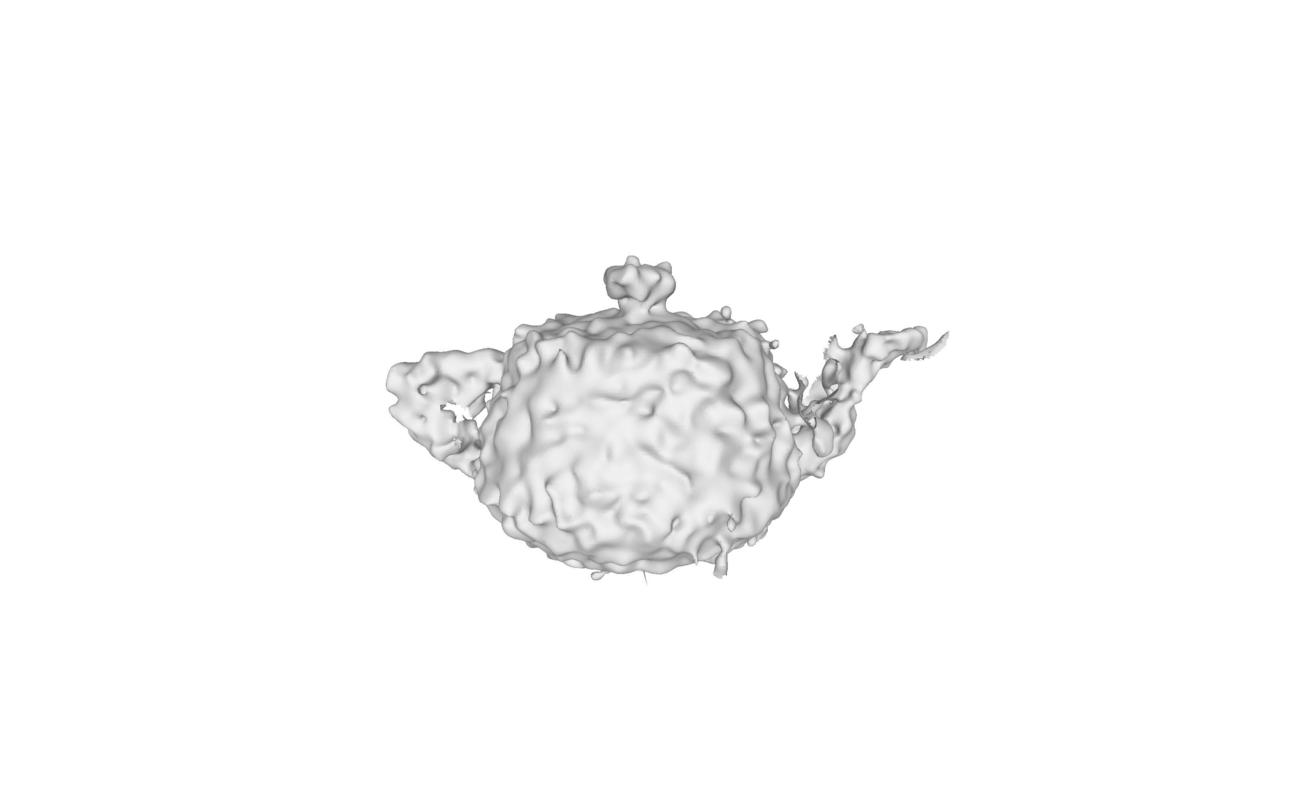}}
\subfloat{\includegraphics[width=\fitscale\tgtwidth, trim={188 80 190 96}, clip]{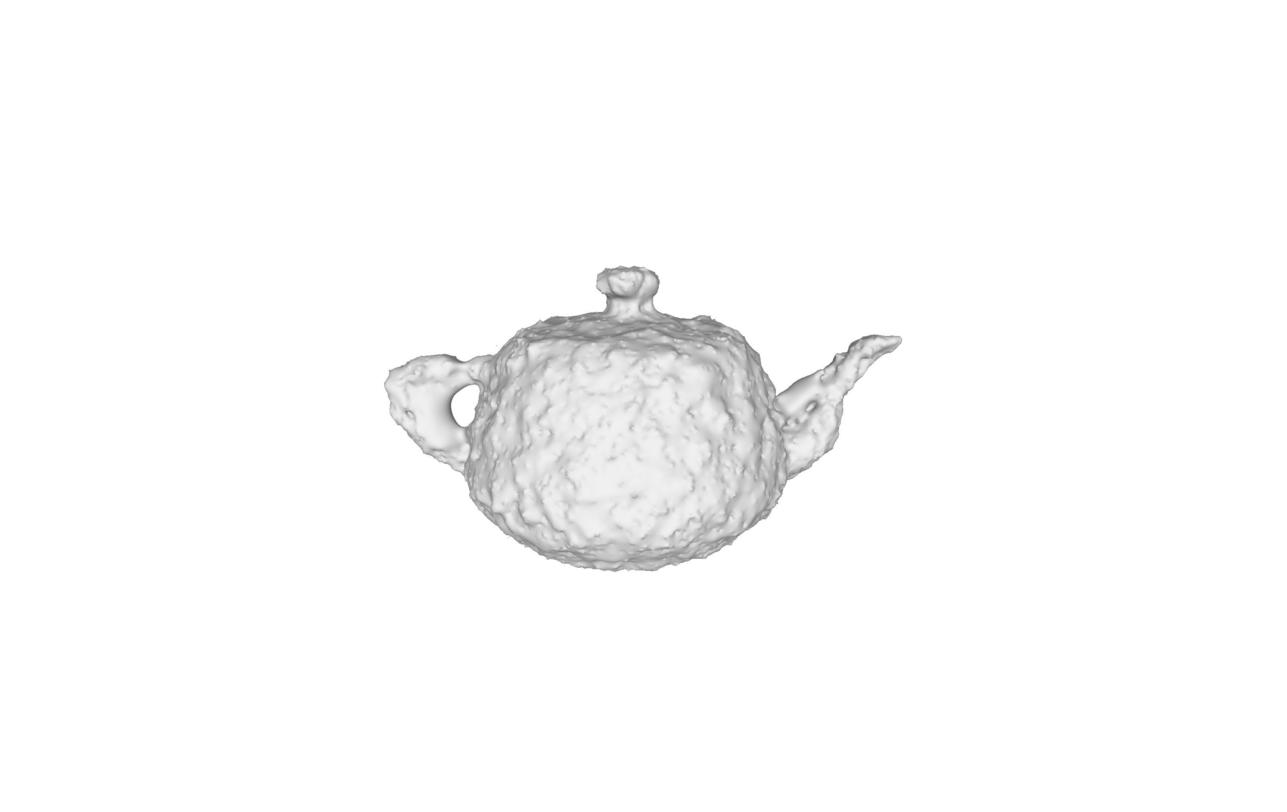}}
\subfloat{\includegraphics[width=\fitscale\tgtwidth, trim={188 80 190 96}, clip]{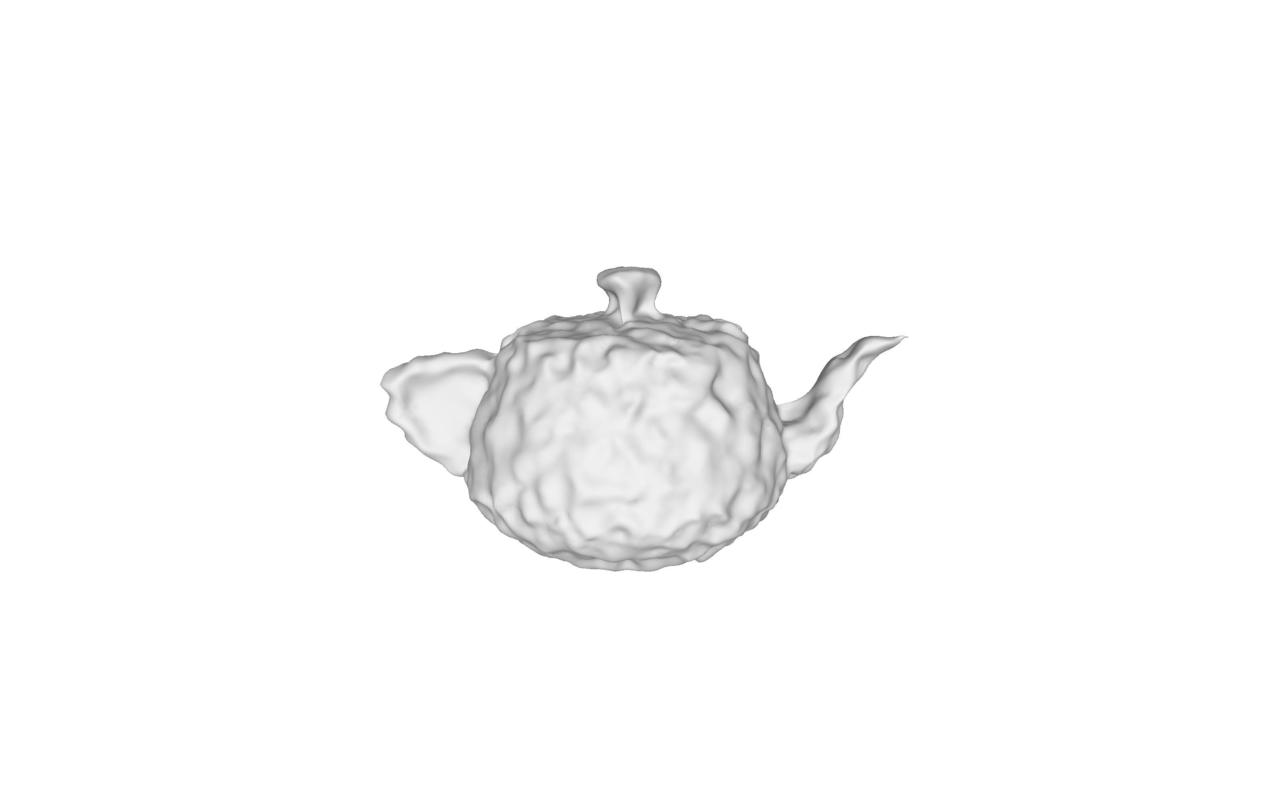}}
\subfloat{\includegraphics[width=\fitscale\tgtwidth, trim={188 80 190 96}, clip]{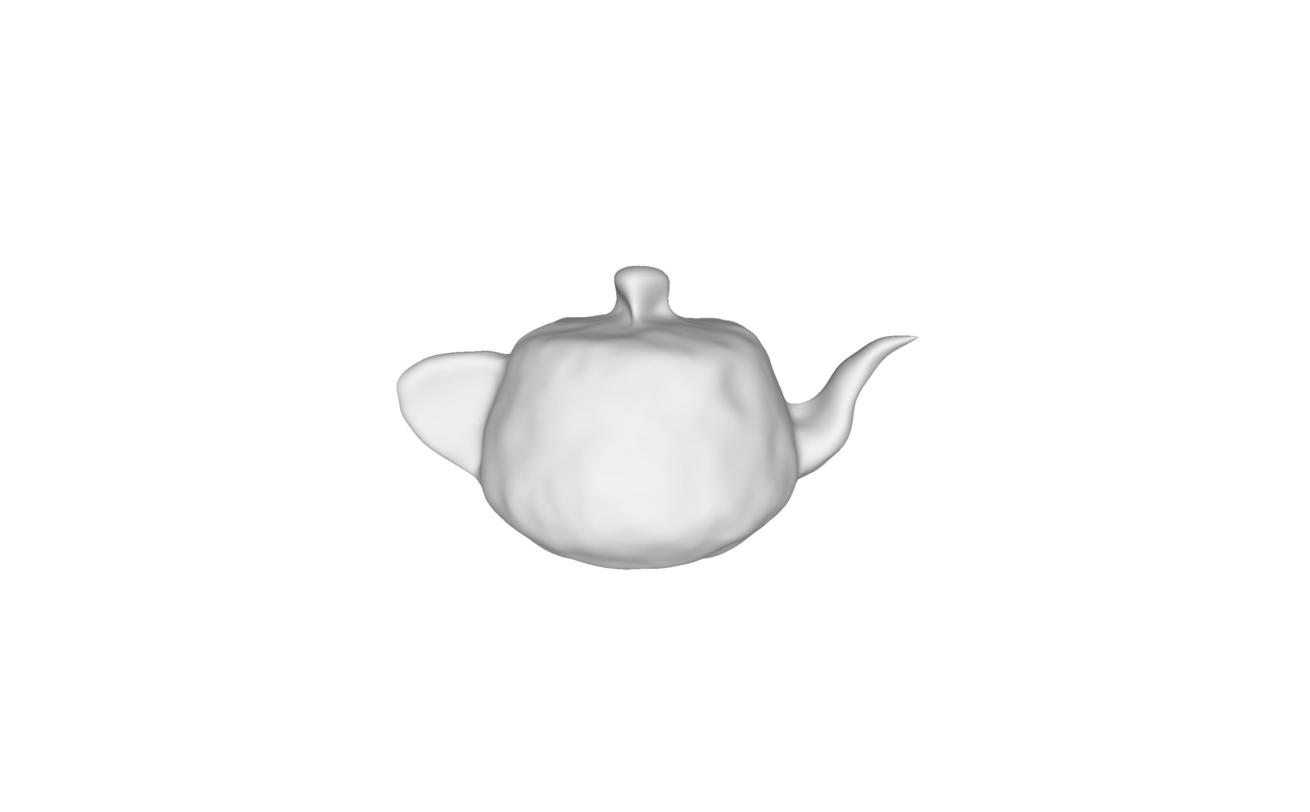}}
\subfloat{\includegraphics[width=\fitscale\tgtwidth, trim={188 80 190 96}, clip]{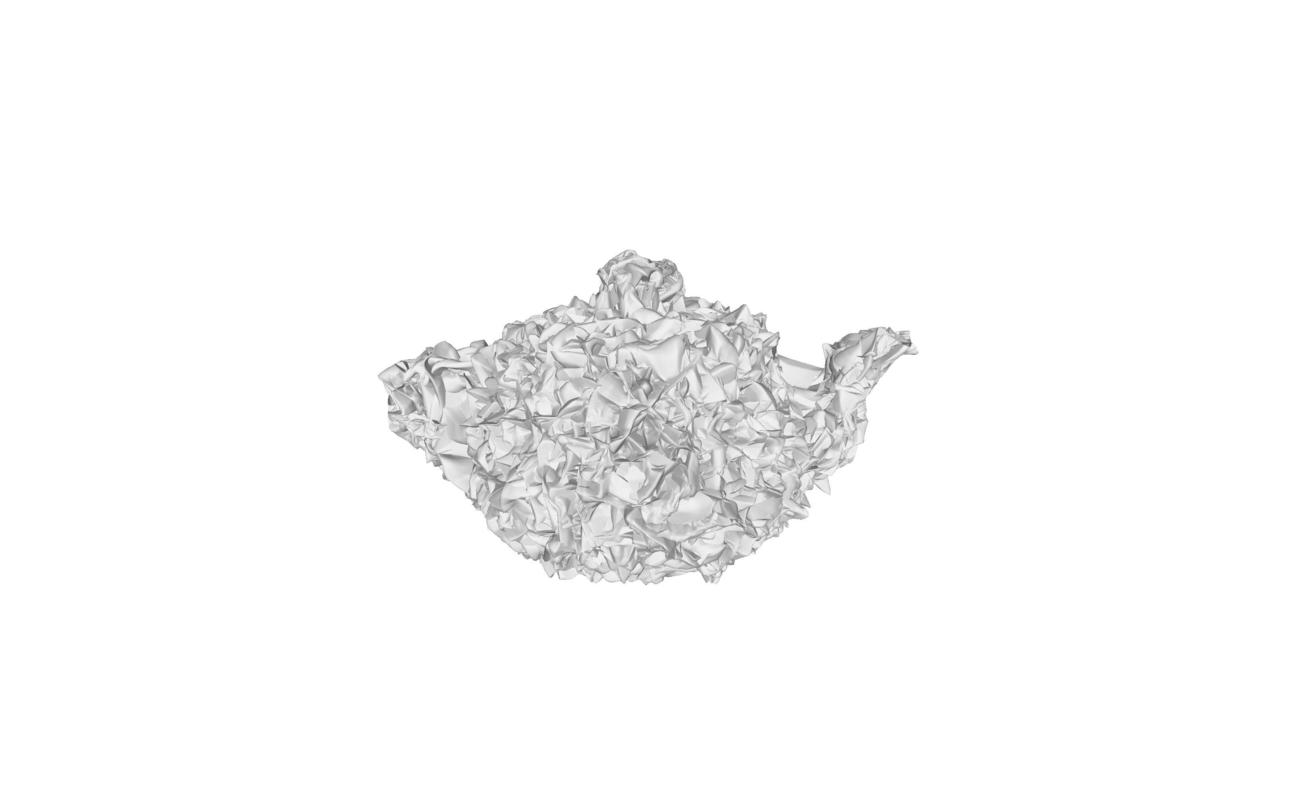}}
\subfloat{\includegraphics[width=\fitscale\tgtwidth, trim={188 80 190 96}, clip]{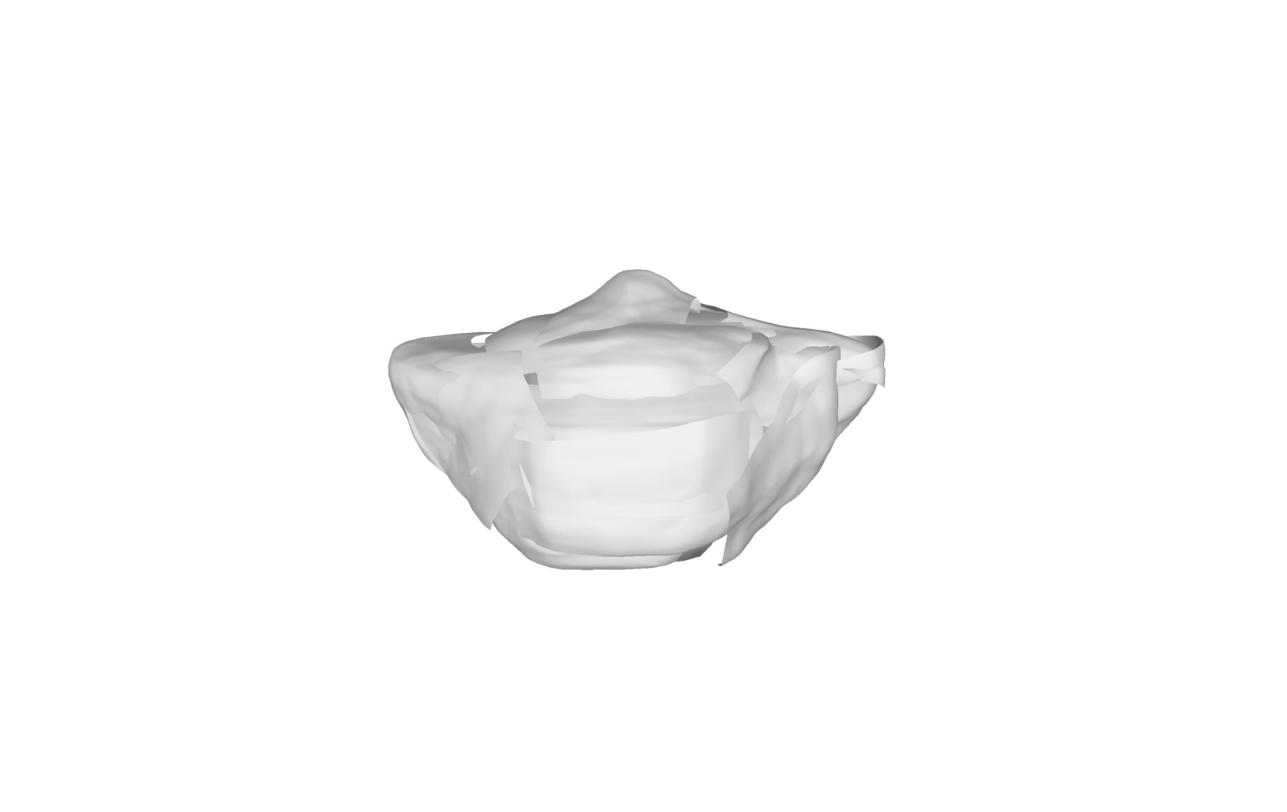}}
\subfloat{\includegraphics[width=\fitscale\tgtwidth, trim={188 80 190 96}, clip]{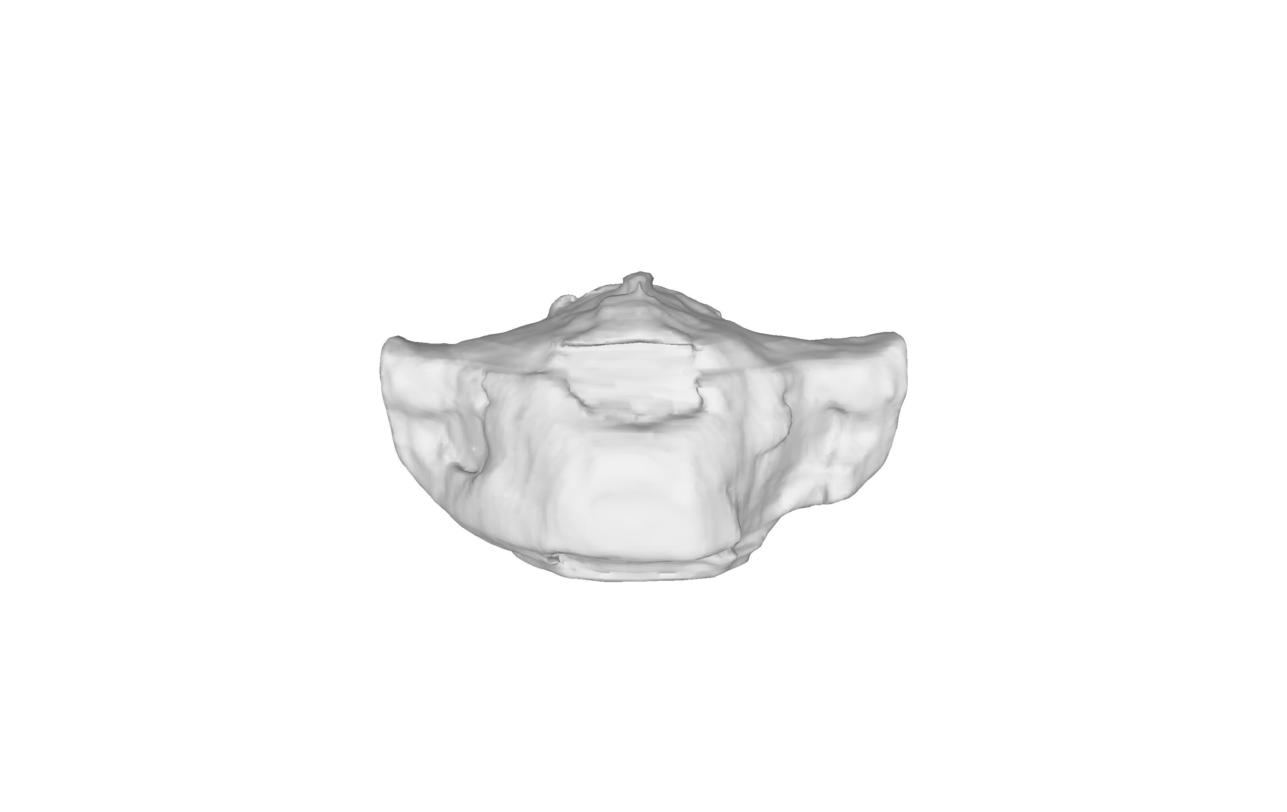}}
~
\subfloat{\includegraphics[width=\fitscale\tgtwidth, trim={188 80 190 96}, clip]{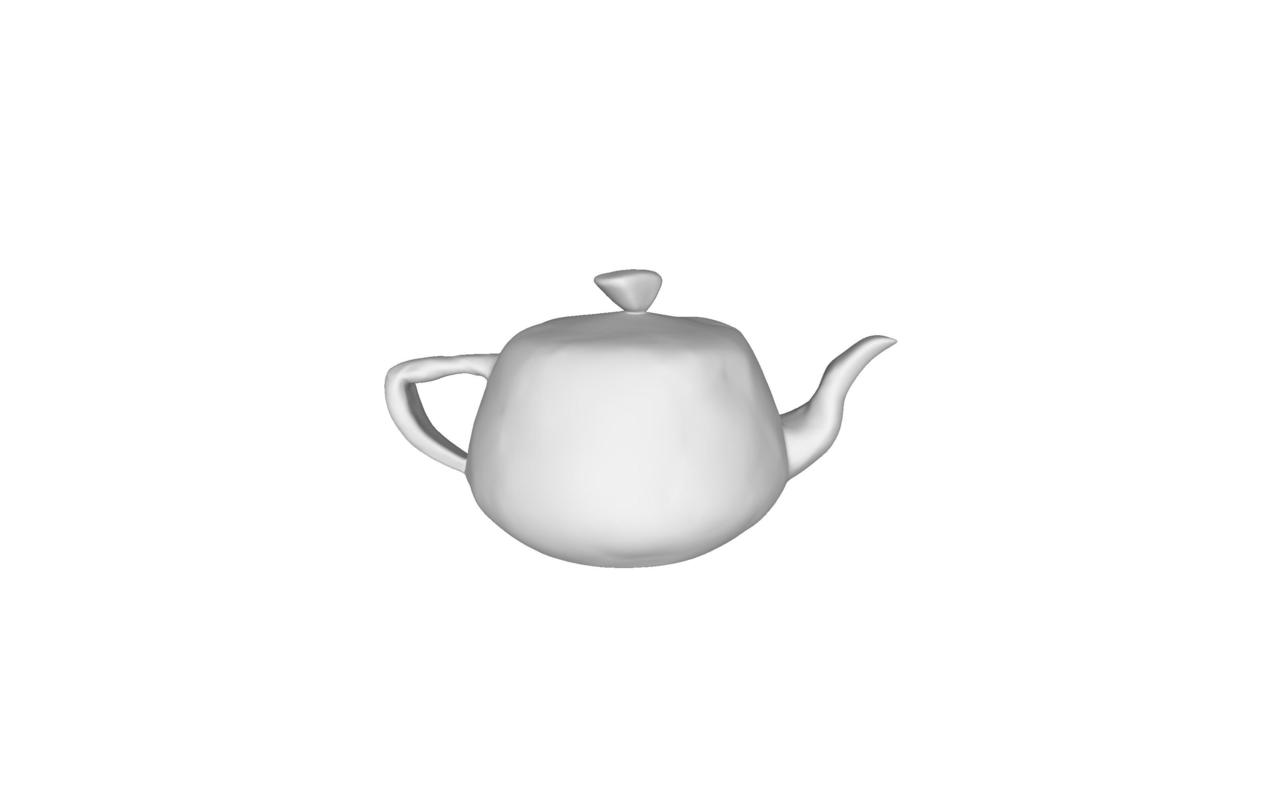}}
\\
\vspace{-5mm}
\subfloat[GT+PC]{\includegraphics[width=\fitscale\tgtwidth, trim={170 80 172 80}, clip]{figures/results/armadillo_new/snapshot00.jpg}}
\subfloat[\cite{meshlab}+\cite{oztireli2009feature}]{\includegraphics[width=\fitscale\tgtwidth, trim={170 80 172 80}, clip]{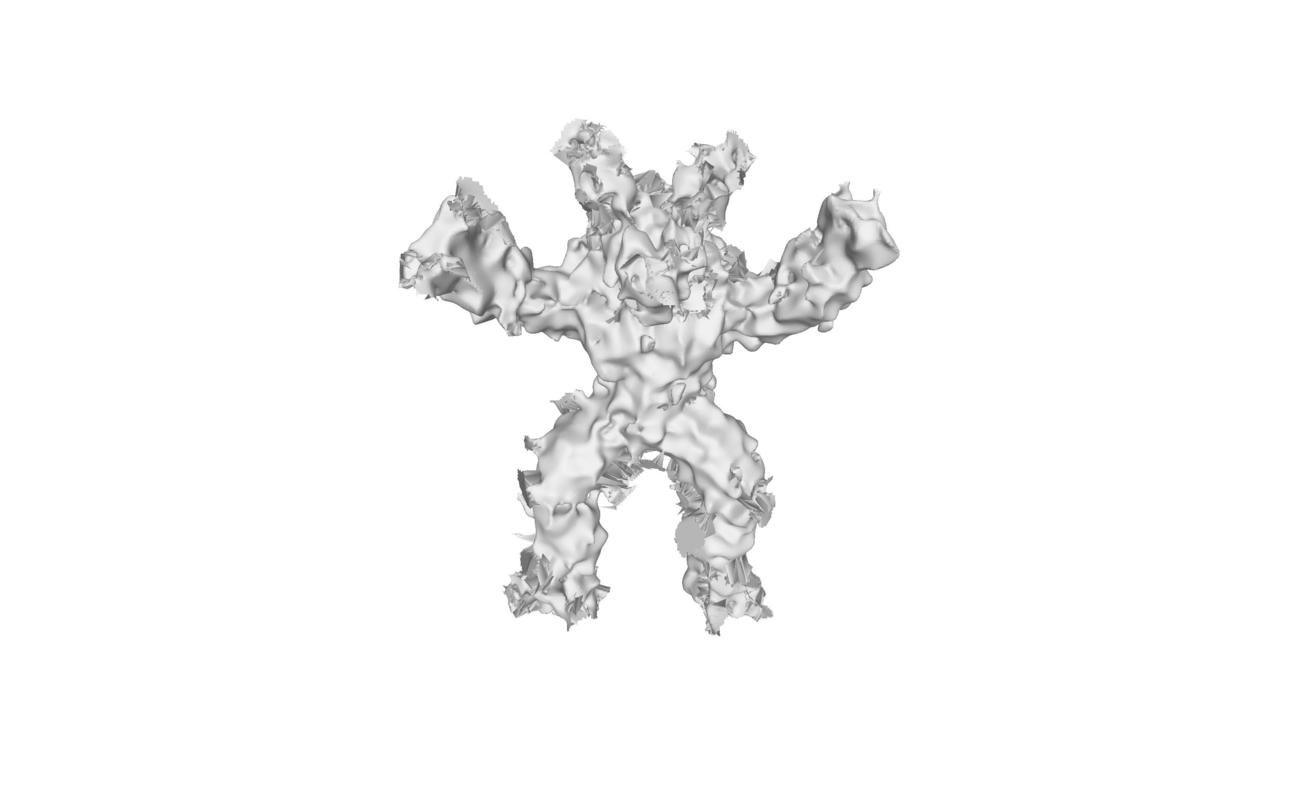}}
\subfloat[\cite{meshlab}+\cite{screenedpoisson}]{\includegraphics[width=\fitscale\tgtwidth, trim={170 80 172 80}, clip]{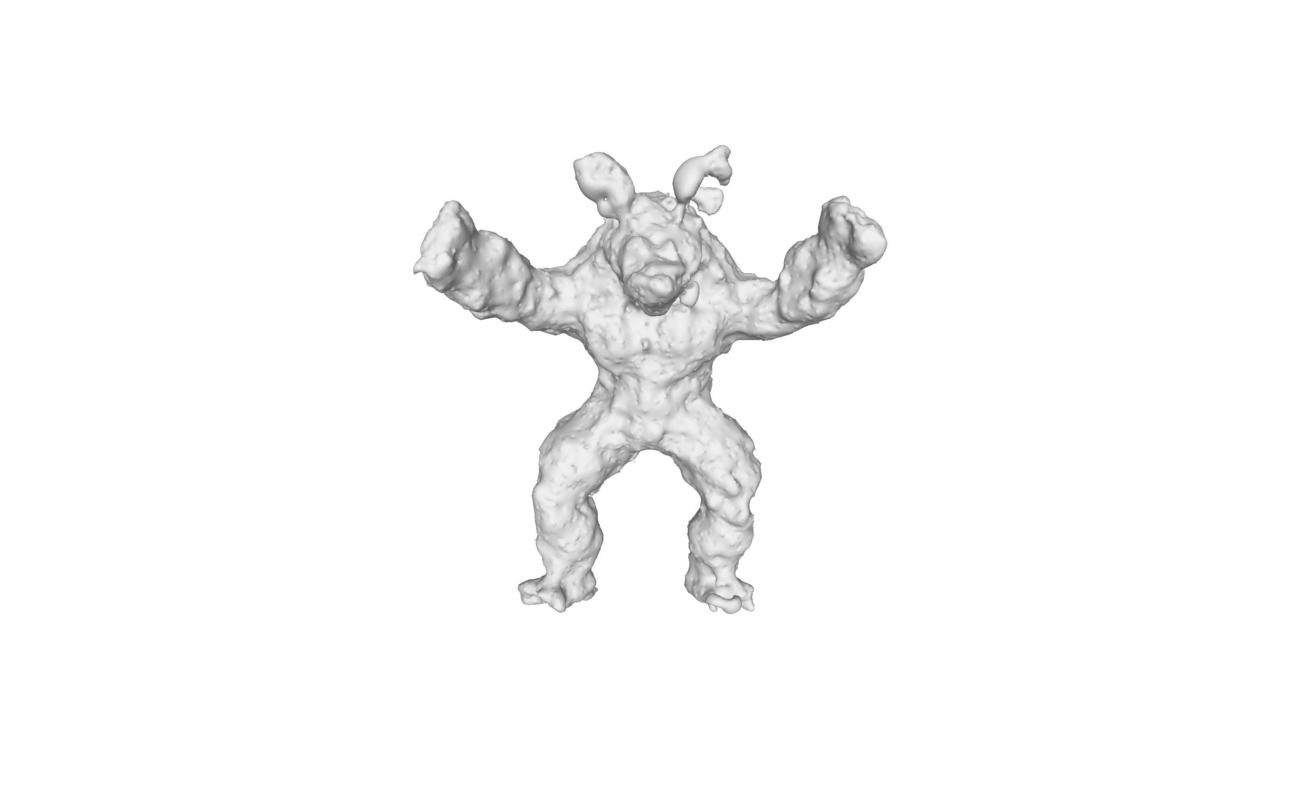}}
\subfloat[\cite{pcpnet}+\cite{oztireli2009feature}]{\includegraphics[width=\fitscale\tgtwidth, trim={170 80 172 80}, clip]{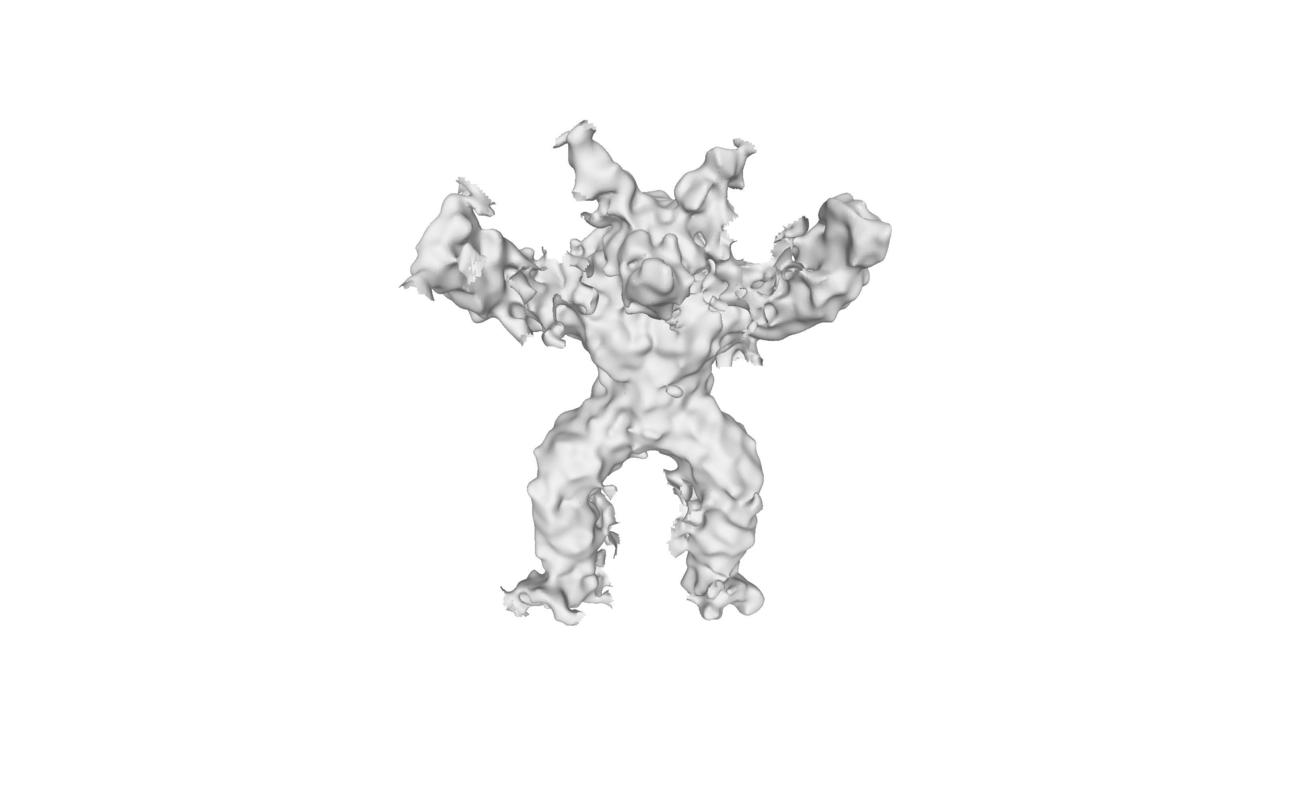}}
\subfloat[\cite{pcpnet}+\cite{screenedpoisson}]{\includegraphics[width=\fitscale\tgtwidth, trim={170 80 172 80}, clip]{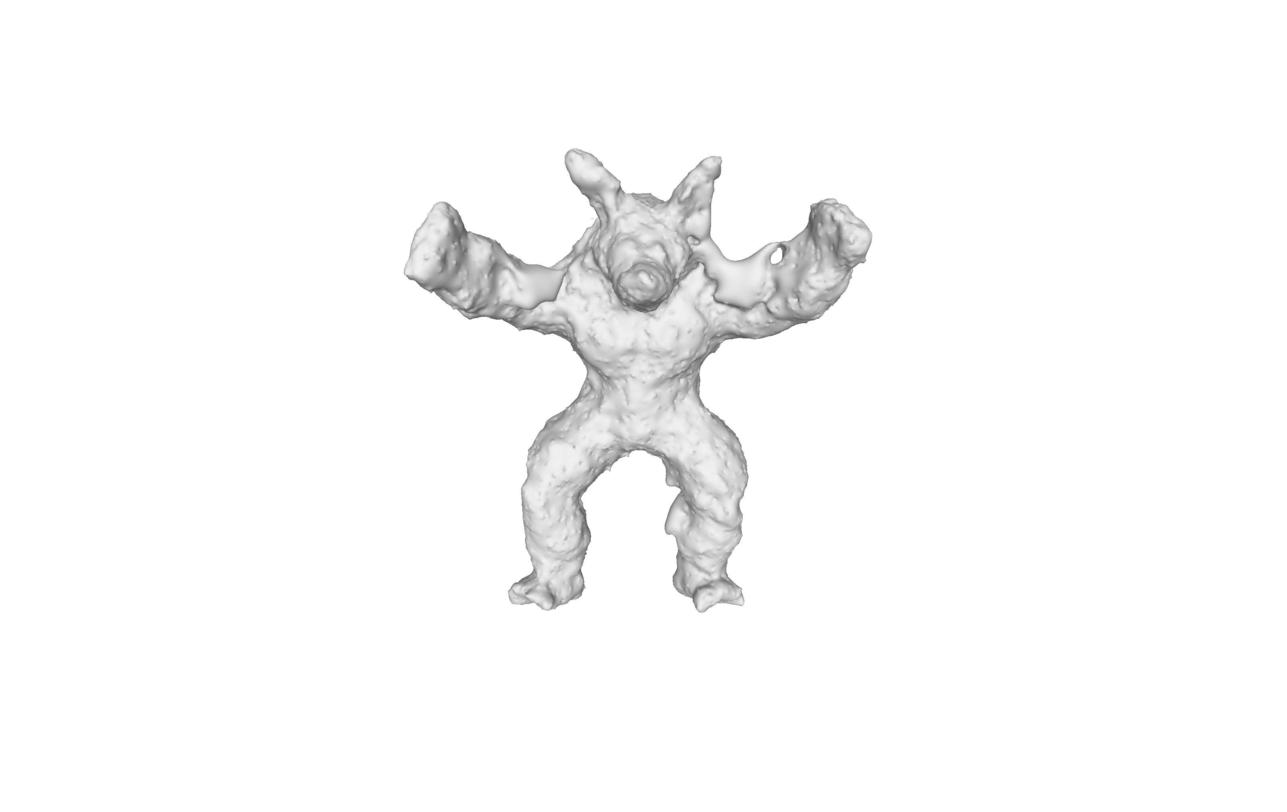}}
\subfloat[Lap-low]{\includegraphics[width=\fitscale\tgtwidth, trim={170 80 172 80}, clip]{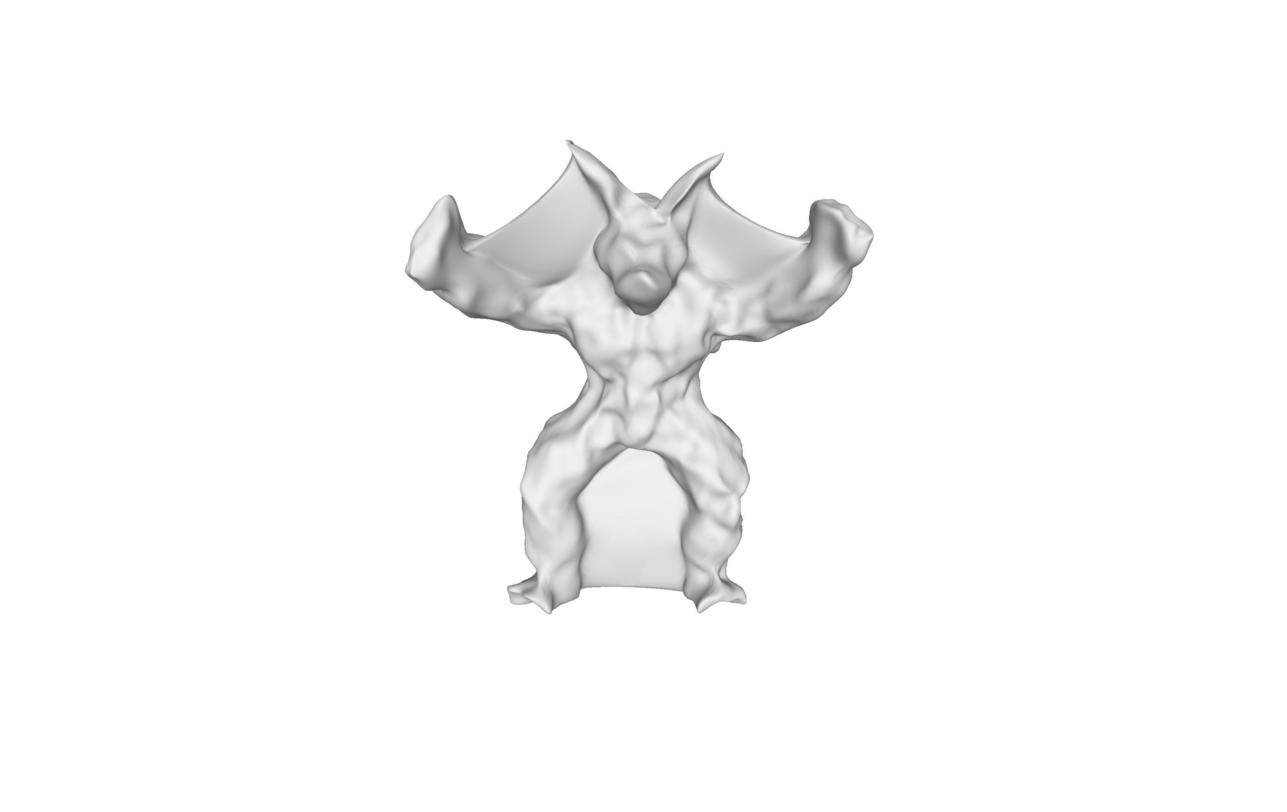}}
\subfloat[Lap-high]{\includegraphics[width=\fitscale\tgtwidth, trim={170 80 172 80}, clip]{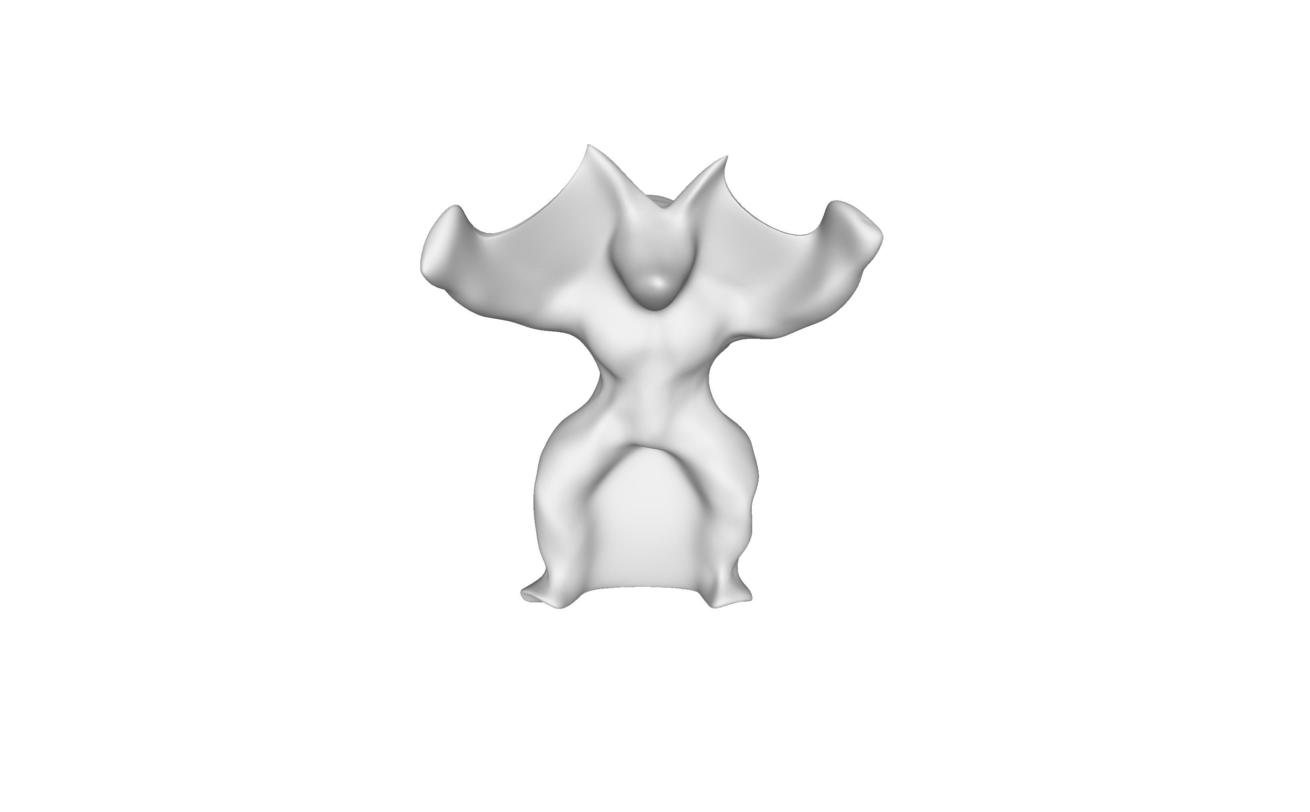}}
\subfloat[DGP]{\includegraphics[width=\fitscale\tgtwidth, trim={170 80 172 80}, clip]{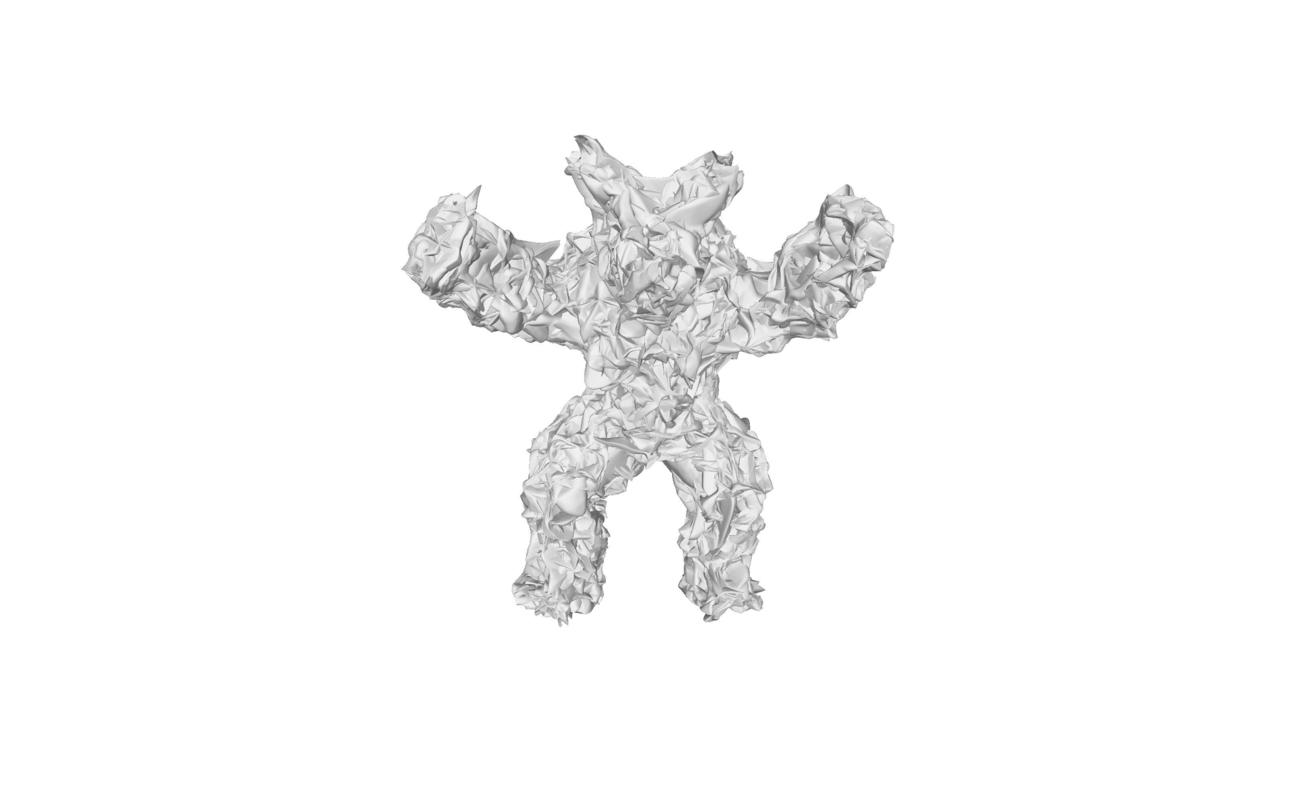}}
\subfloat[AtlasNet]{\includegraphics[width=\fitscale\tgtwidth, trim={170 80 172 80}, clip]{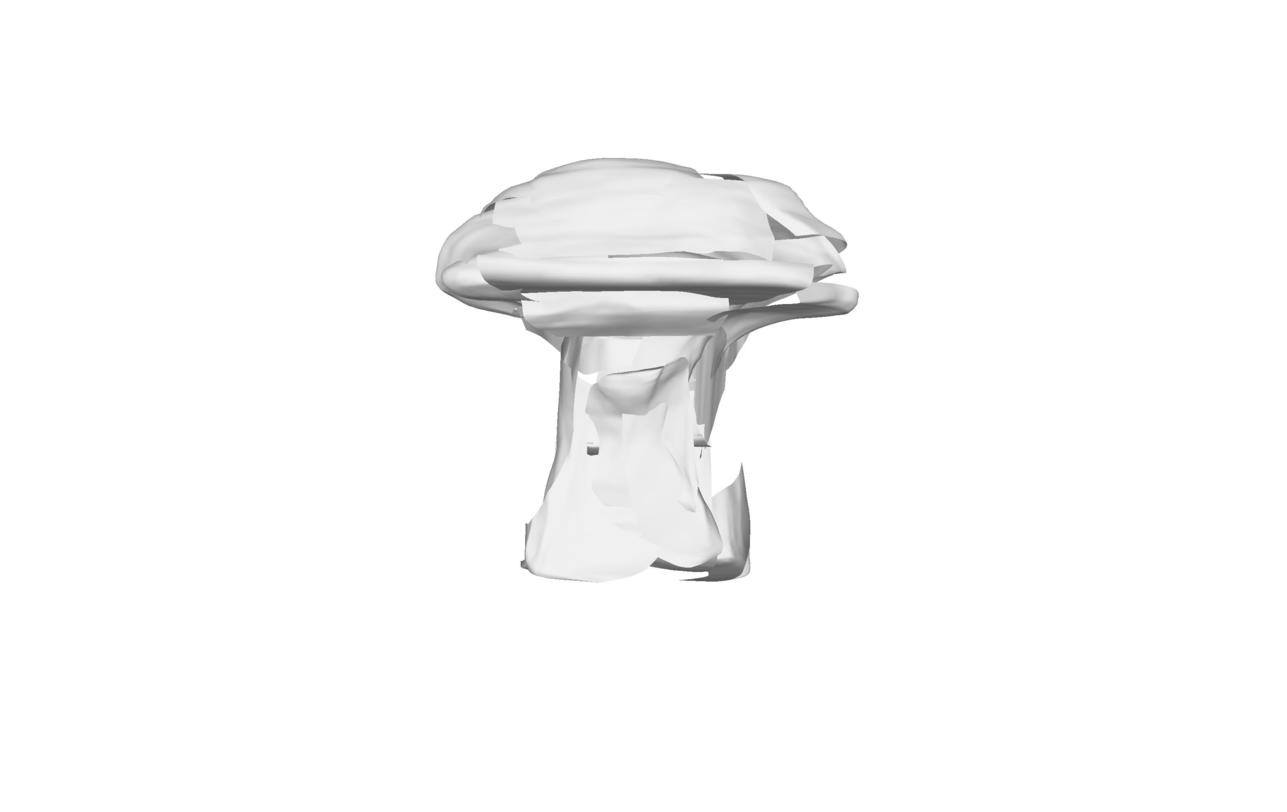}}
\subfloat[OccNet]{\includegraphics[width=\fitscale\tgtwidth, trim={170 80 172 80}, clip]{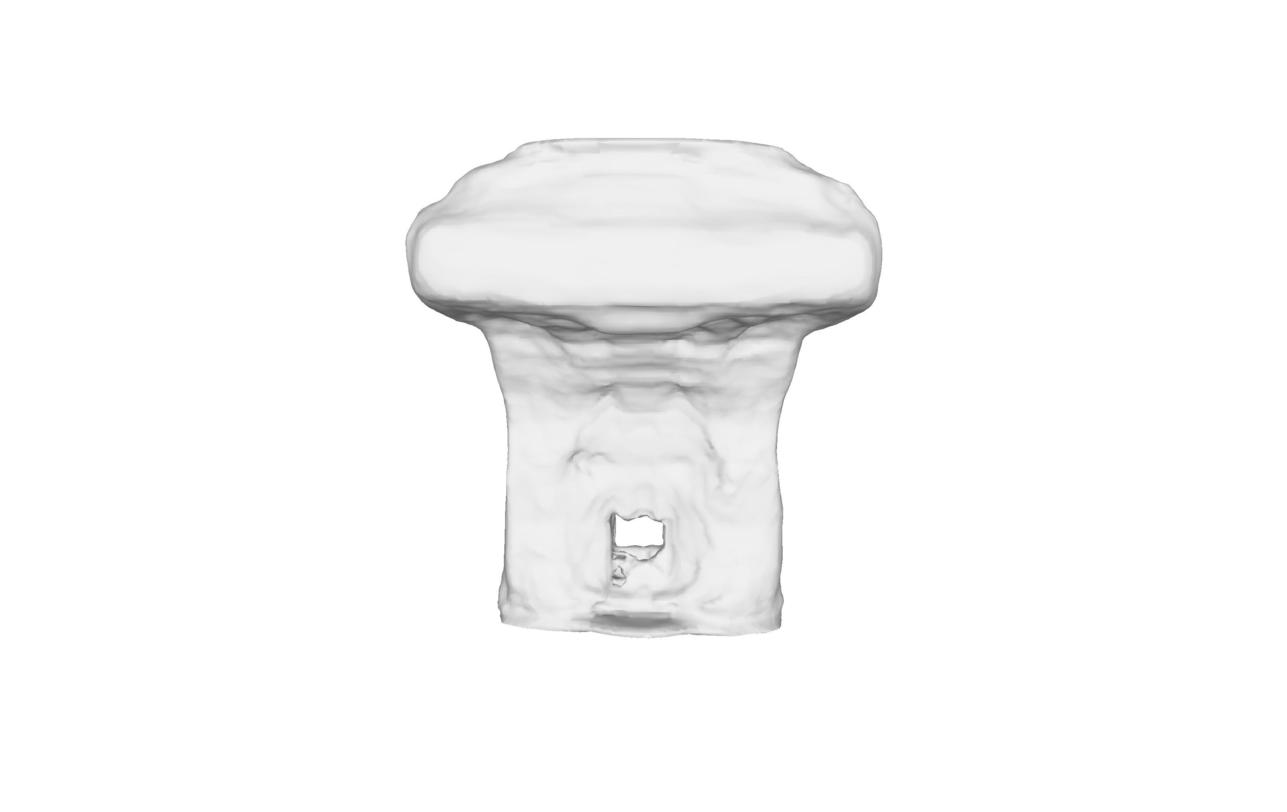}}
~
\subfloat[{\bf Ours}]{\includegraphics[width=\fitscale\tgtwidth, trim={170 80 172 80}, clip]{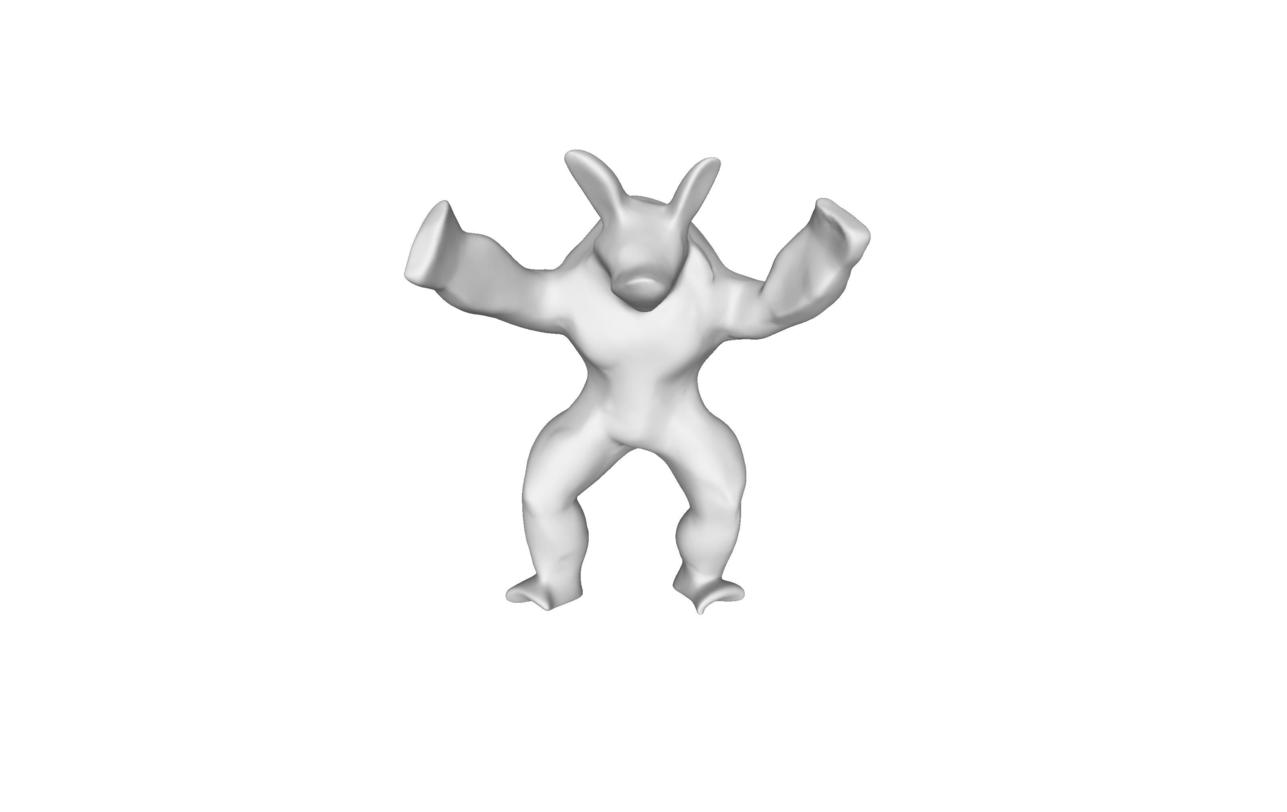}}

%% file: figures/meshlets_are_awesome/priors.tex

\newlength{\numcropsteaser}
\setlength{\numcropsteaser}{4pt}
\newcommand{\fitscale}{0.925}

\newlength{\cropwidthteaser}
\settowidth{\cropwidthteaser}{\includegraphics{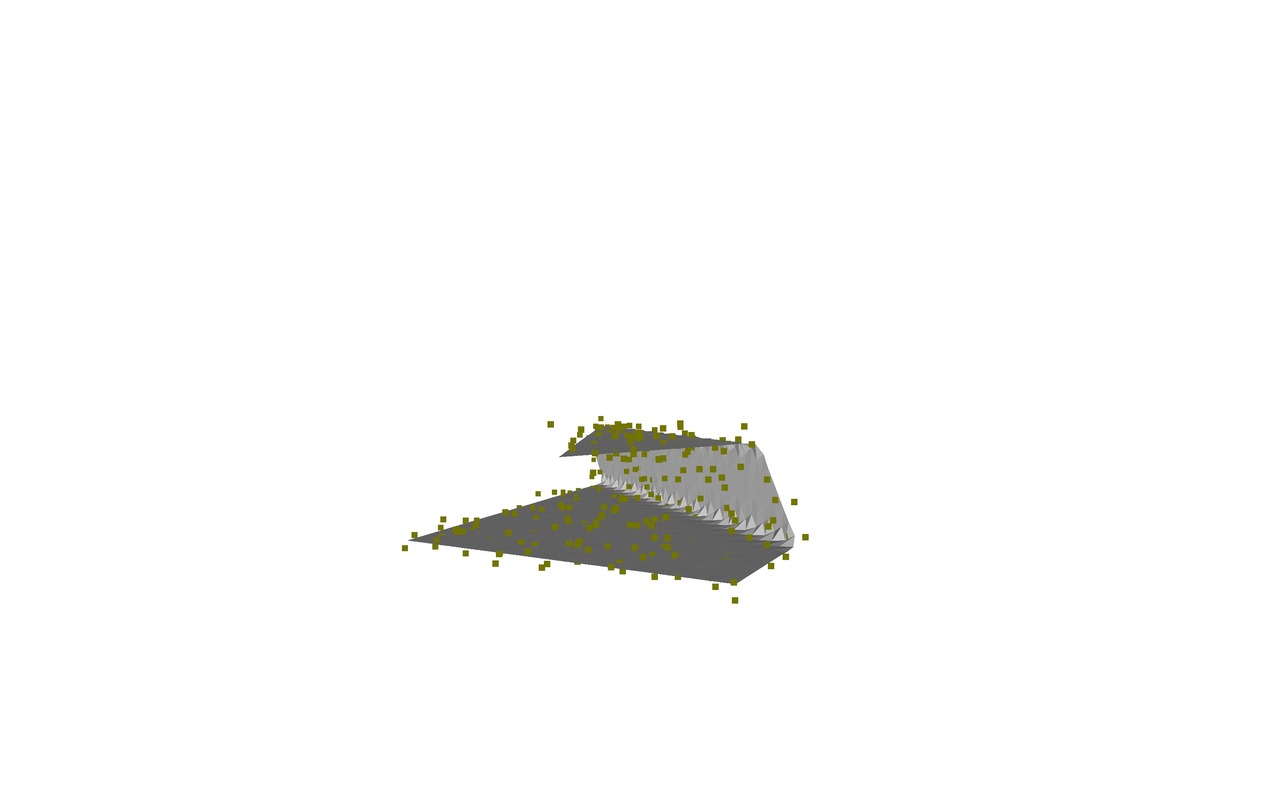}}
\newlength{\oneteaser}
\setlength{\oneteaser}{1pt}
\newlength{\tgtwidthteaser}
\setlength\tgtwidthteaser{\columnwidth*\ratio{\oneteaser}{\numcropsteaser}}

\begin{small}
    
\centering
\subfloat{\includegraphics[width=\fitscale\tgtwidthteaser, trim={280 110 300 280}, clip]{figures/meshlets_are_awesome/00/gt00.jpg}}
\subfloat{\includegraphics[width=\fitscale\tgtwidthteaser, trim={280 110 300 280}, clip]{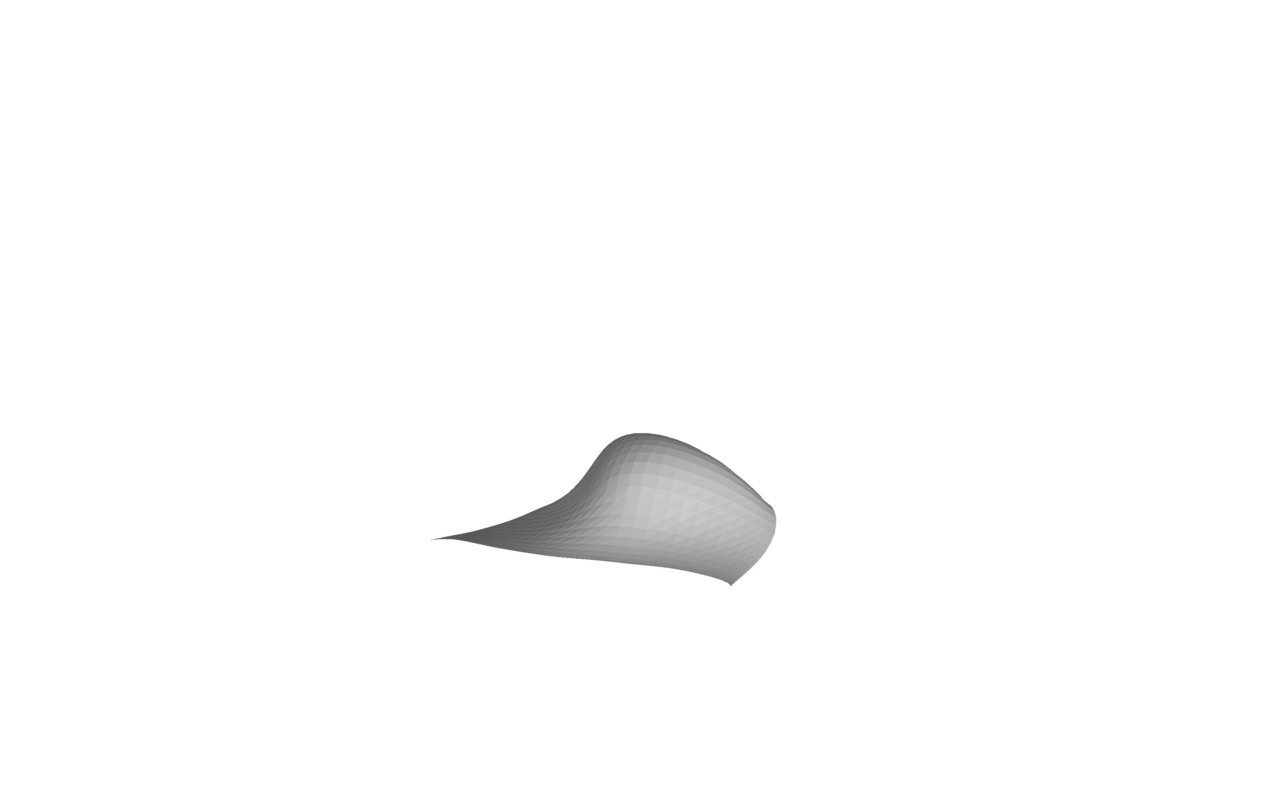}}
\subfloat{\includegraphics[width=\fitscale\tgtwidthteaser, trim={280 110 300 280}, clip]{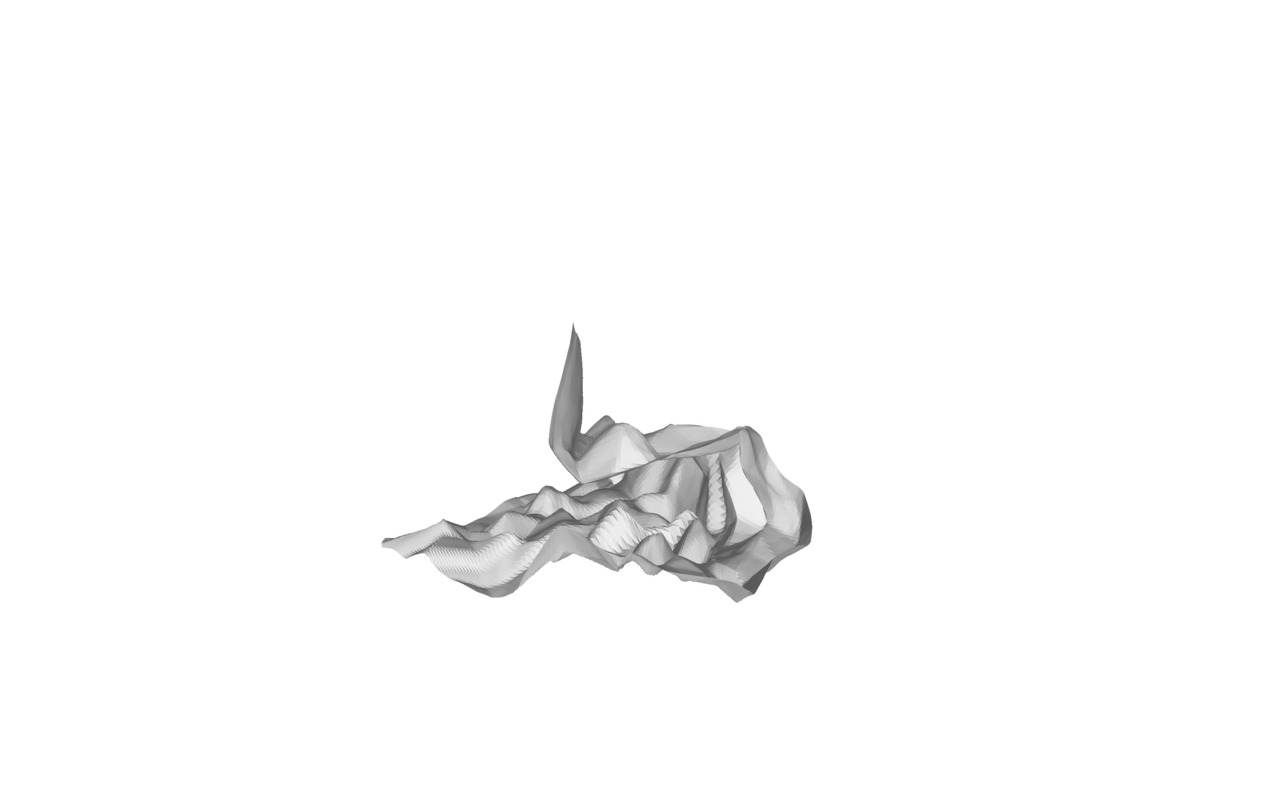}}
\subfloat{\includegraphics[width=\fitscale\tgtwidthteaser, trim={280 110 300 280}, clip]{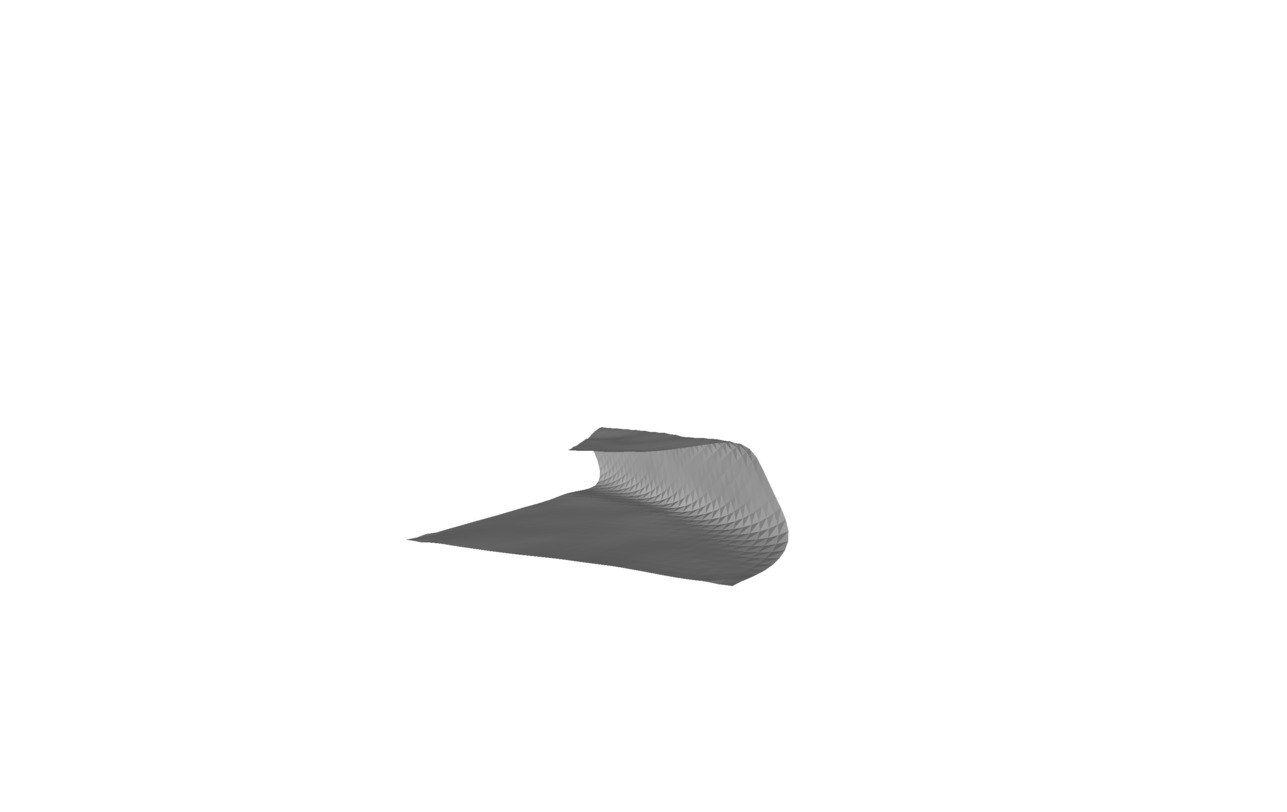}}
\\
\vspace{-5mm}
\setcounter{subfigure}{0}
\subfloat[\footnotesize GT+PC]{\includegraphics[width=\fitscale\tgtwidthteaser,      trim={280 110 300 140}, clip]{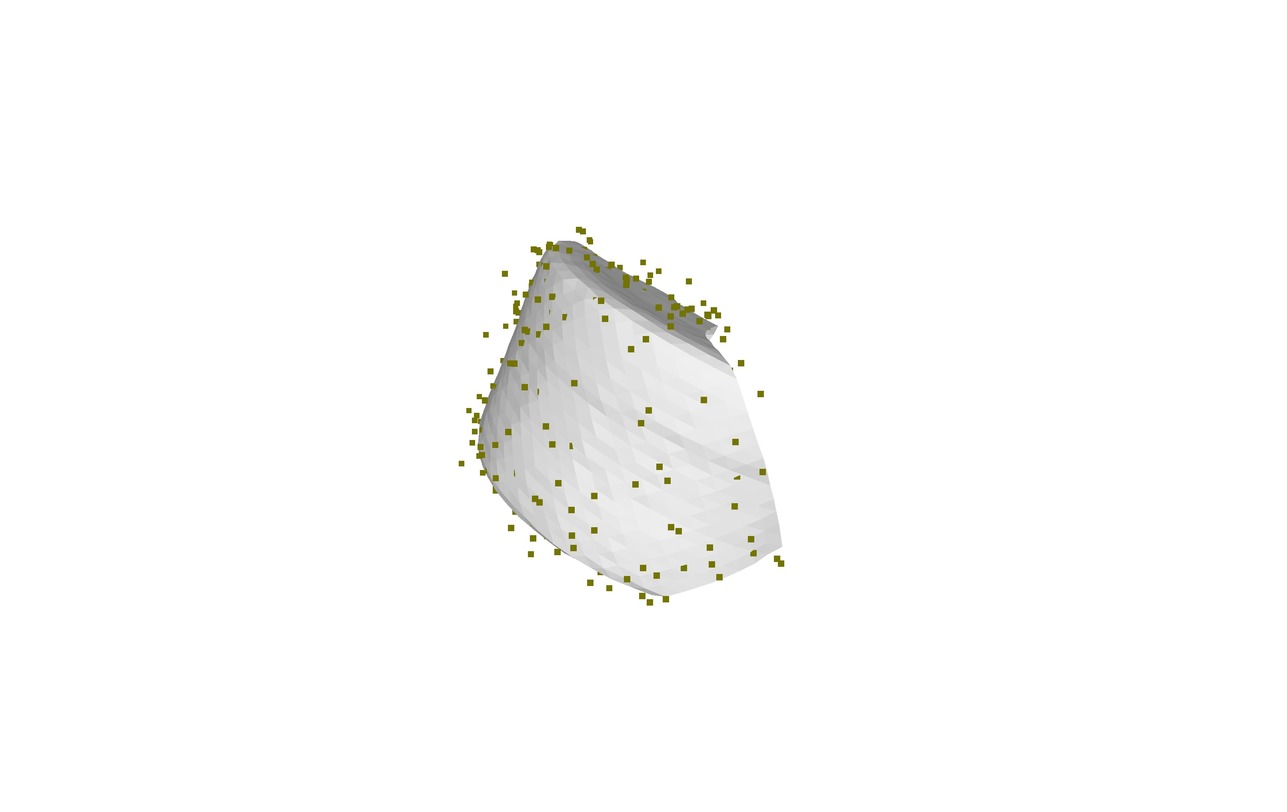}}
\subfloat[\footnotesize Laplacian]{\includegraphics[width=\fitscale\tgtwidthteaser,  trim={280 110 300 140}, clip]{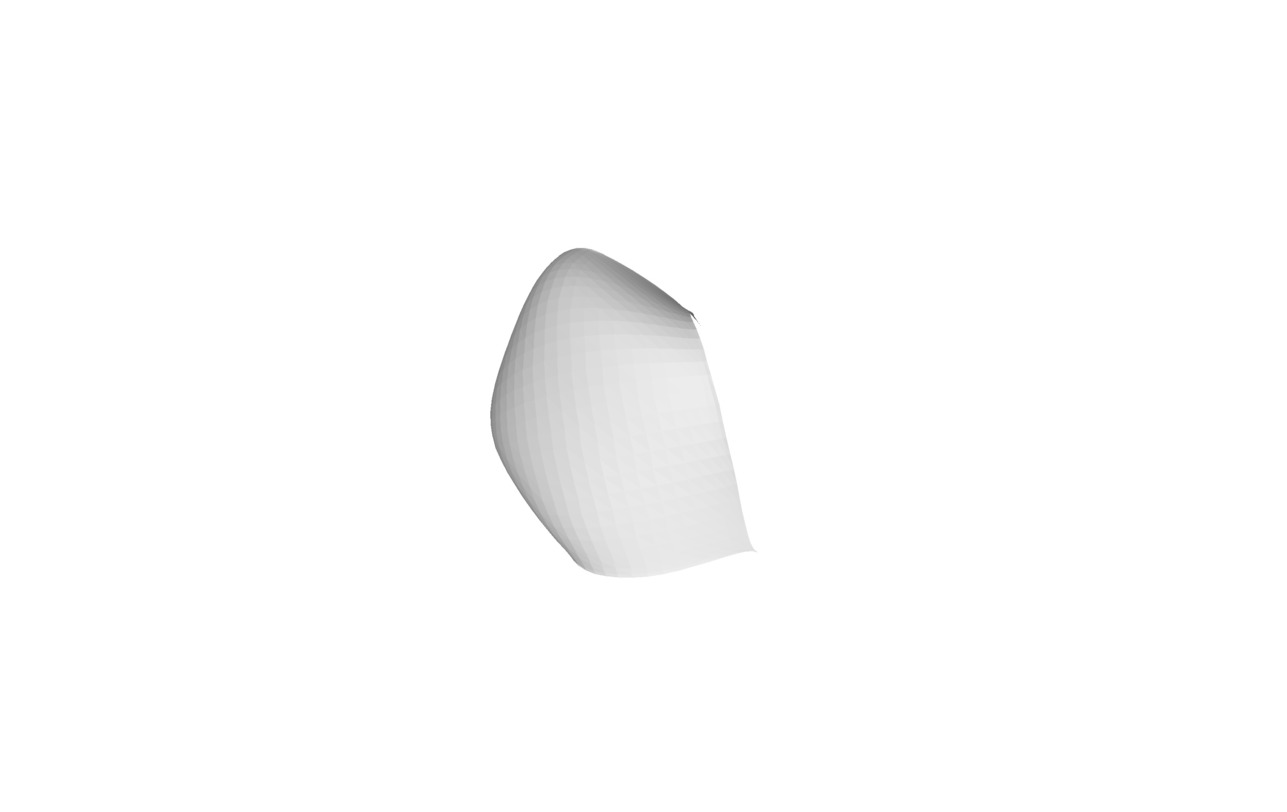}}
\subfloat[\footnotesize DGP]{\includegraphics[width=\fitscale\tgtwidthteaser,        trim={280 110 300 140}, clip]{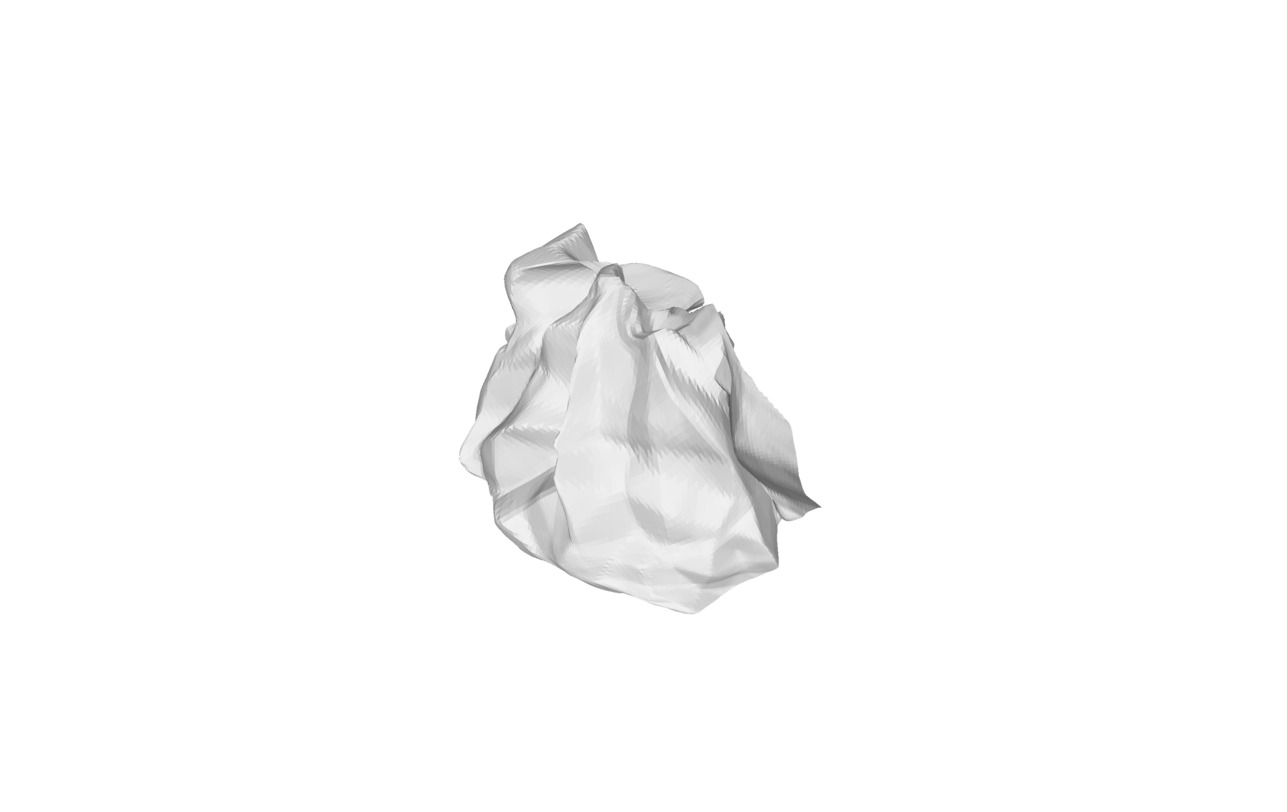}}
\subfloat[{\footnotesize Meshlets}]{\includegraphics[width=\fitscale\tgtwidthteaser, trim={280 110 300 140}, clip]{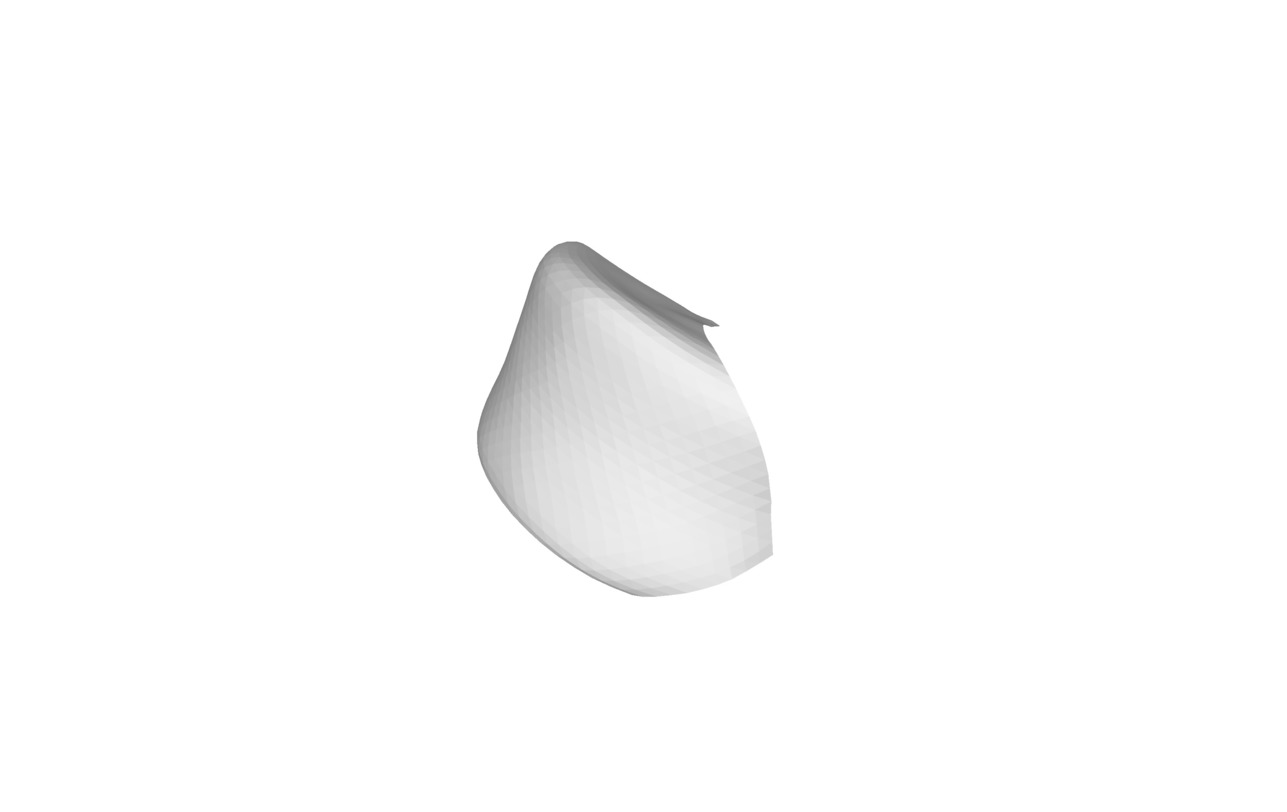}}

\end{small}

%% file: figures/ablation/ablation.tex

\begin{small}
\captionsetup[subfigure]{labelformat=empty}
\centering
\subfloat{\includegraphics[width=.25\columnwidth, trim={150 165 150 170}, clip]{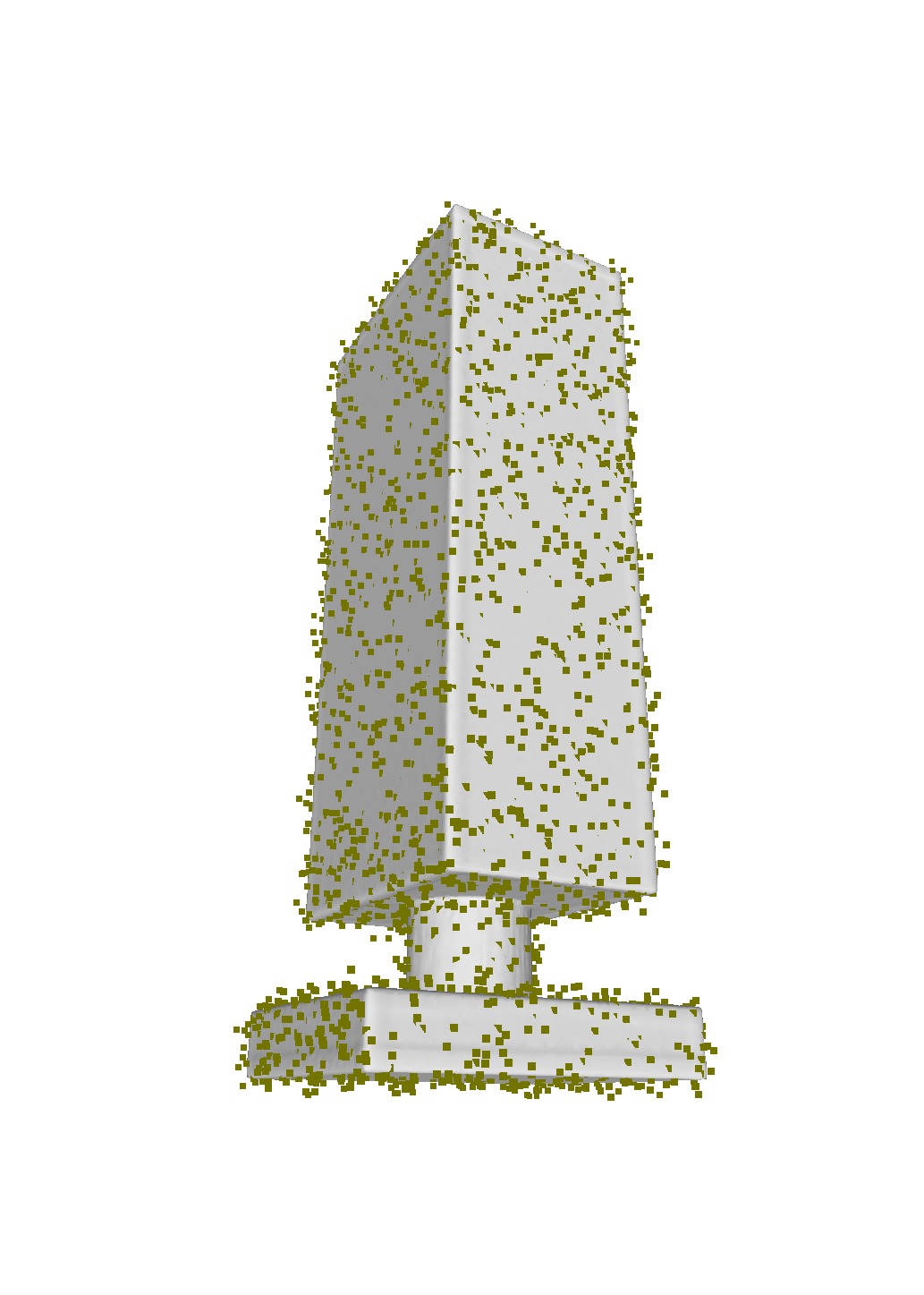}}
\subfloat{\includegraphics[width=.25\columnwidth, trim={150 165 150 170}, clip]{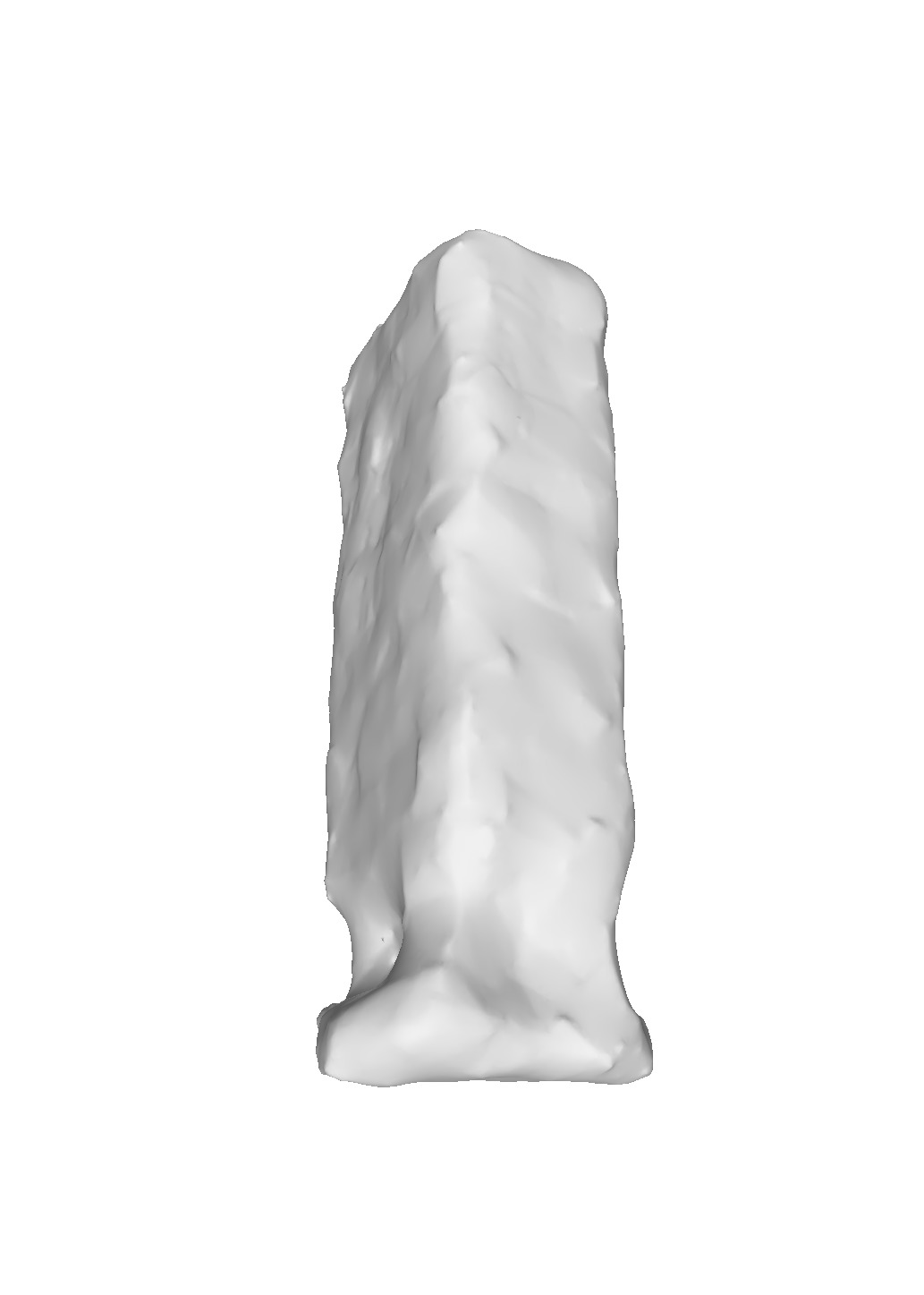}}
\subfloat{\includegraphics[width=.25\columnwidth, trim={150 165 150 170}, clip]{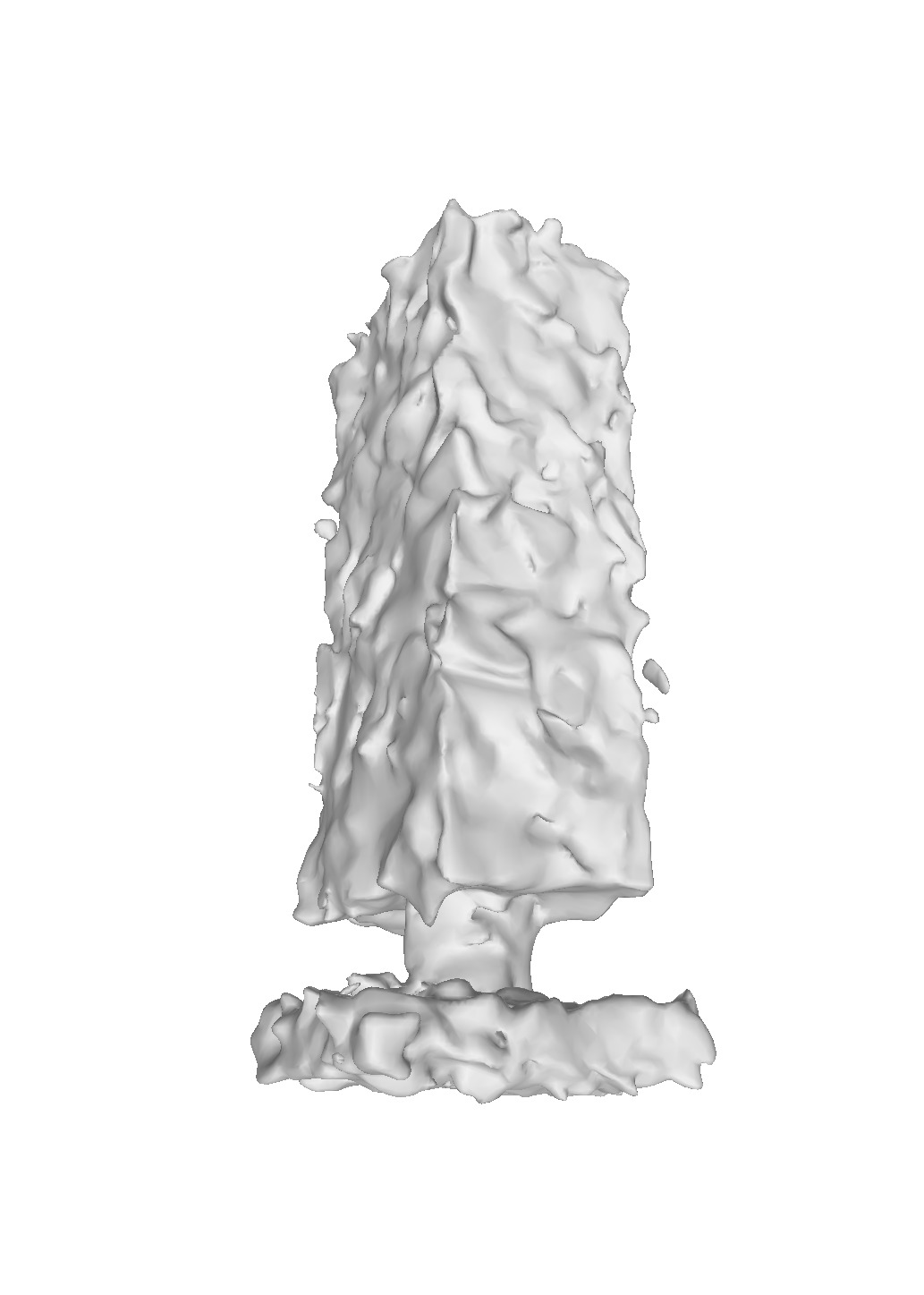}}
\subfloat{\includegraphics[width=.25\columnwidth, trim={150 165 150 170}, clip]{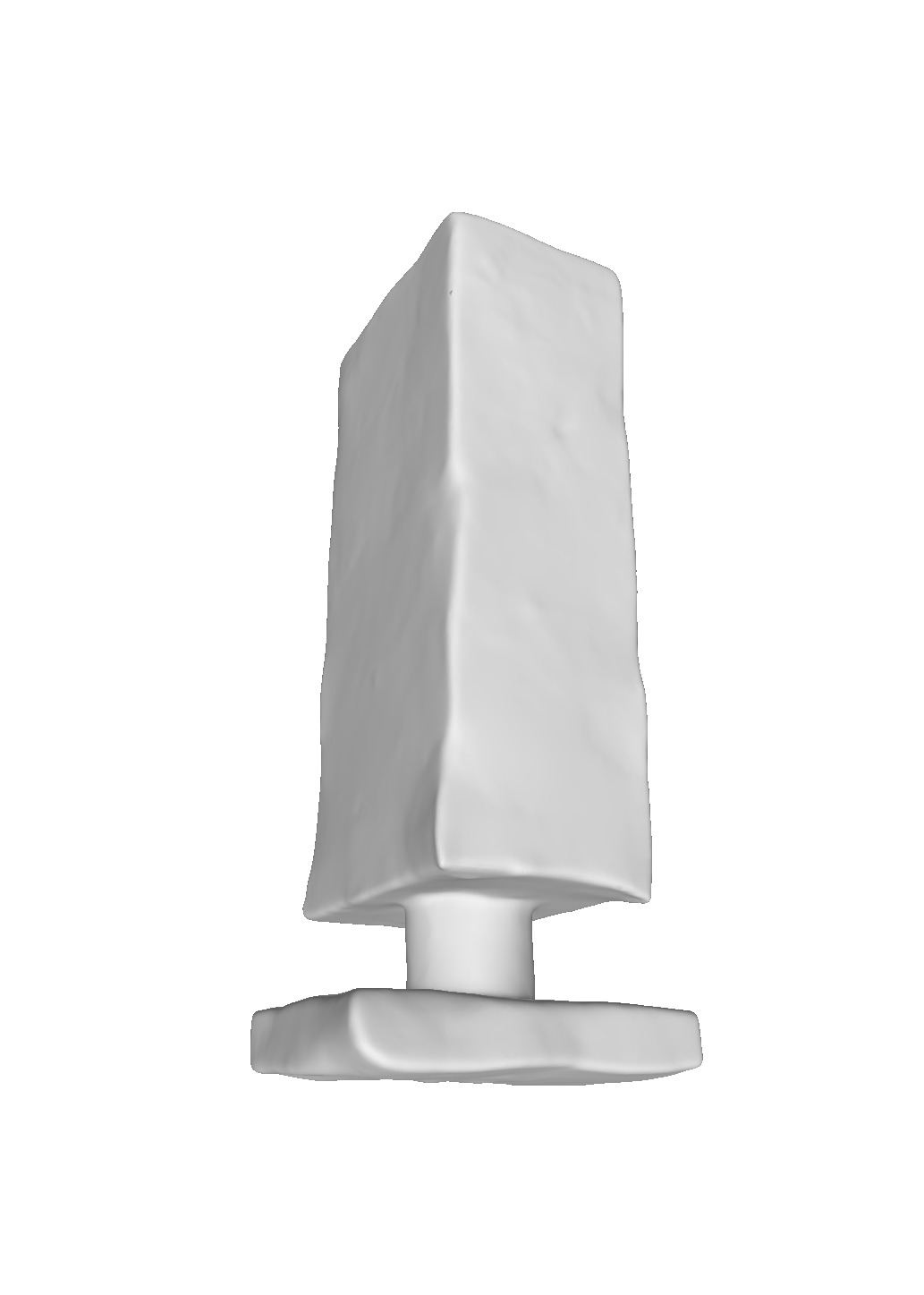}}\\
\vspace{1mm}
\subfloat[GT+PC]{\includegraphics[width=.25\columnwidth, trim={150 165 150 250}, clip]{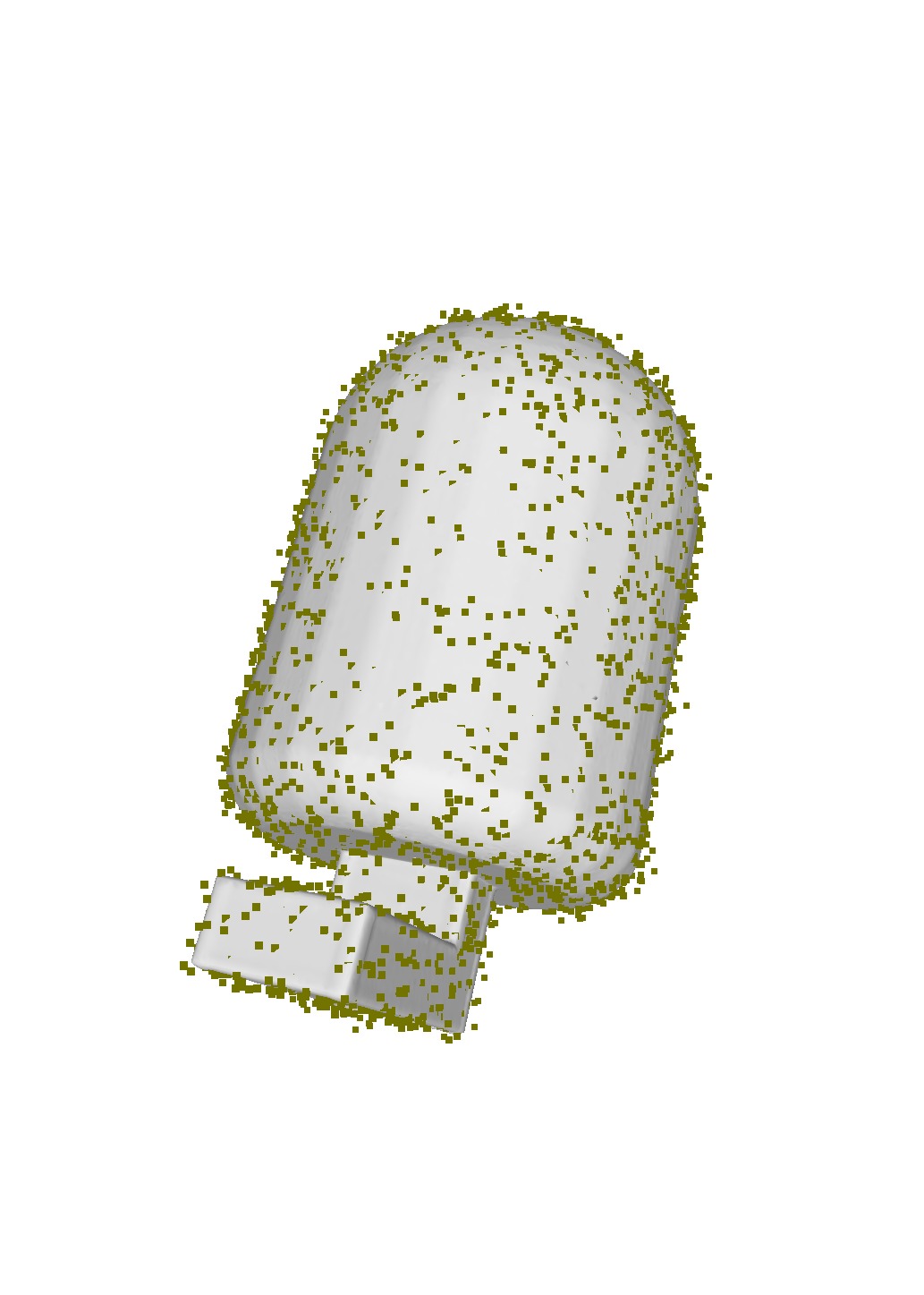}}
\subfloat[Laplacian Prior]{\includegraphics[width=.25\columnwidth, trim={150 165 150 250}, clip]{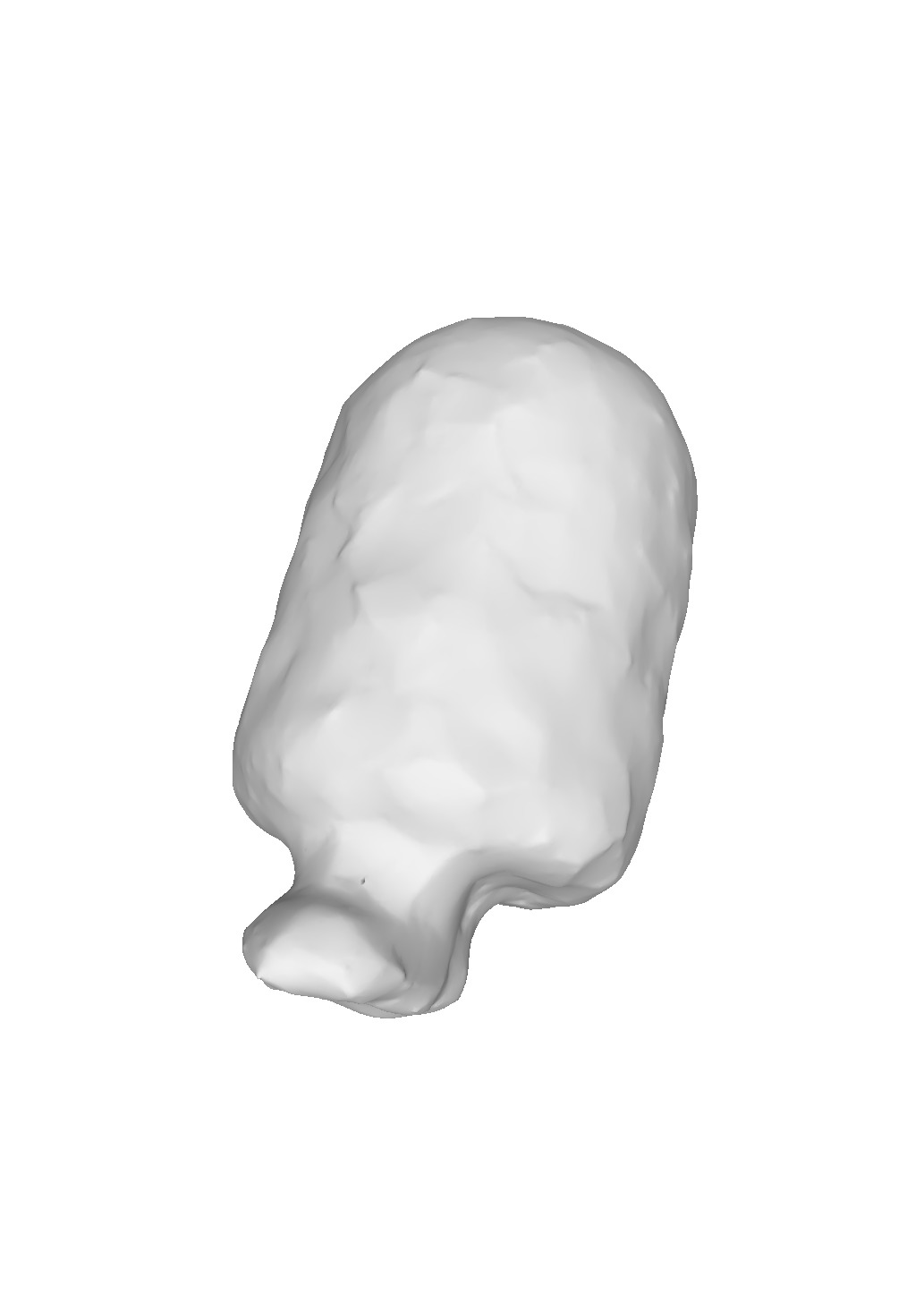}}
\subfloat[No GC]{\includegraphics[width=.25\columnwidth, trim={150 165 150 250}, clip]{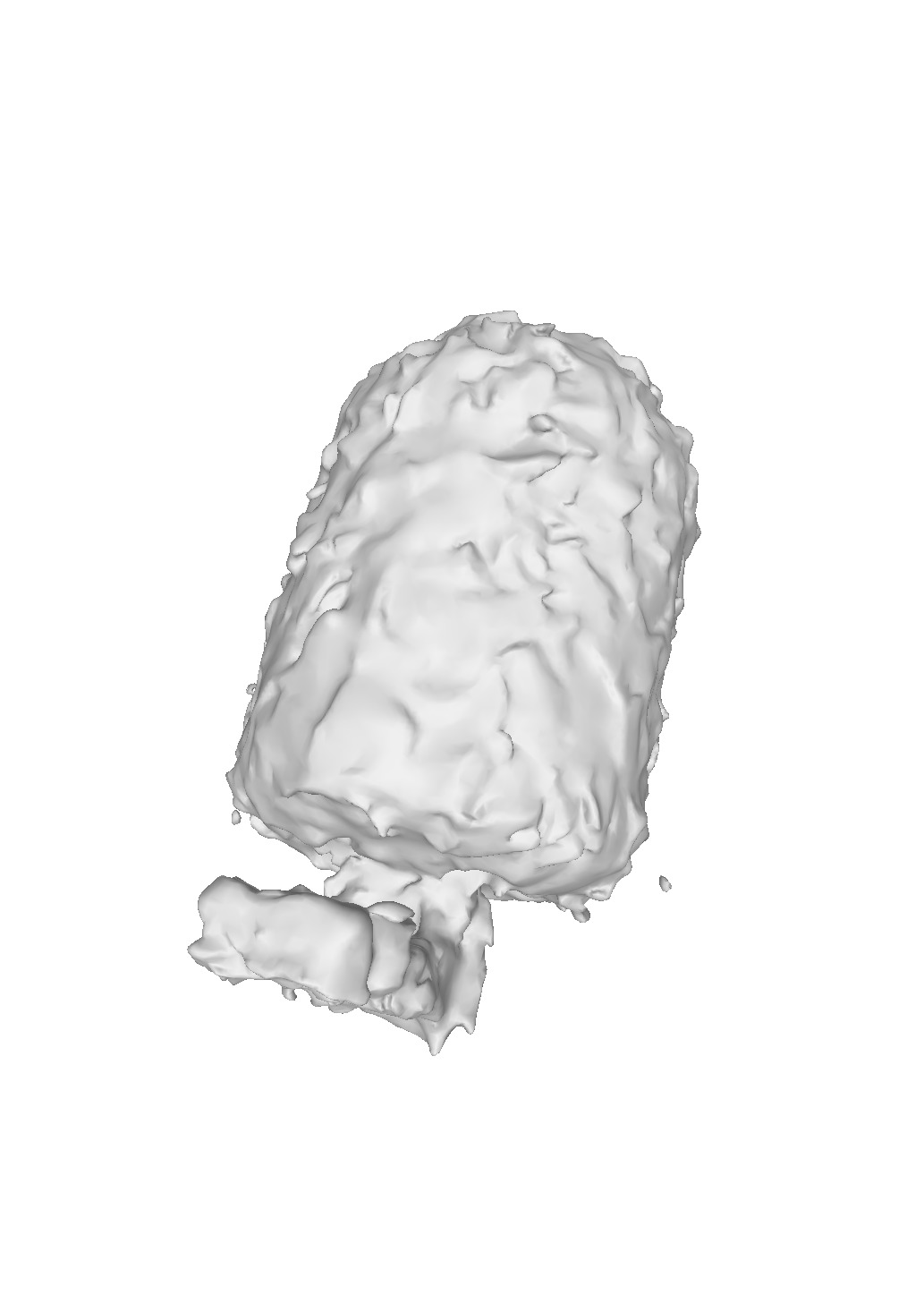}}
\subfloat[Ours]{\includegraphics[width=.25\columnwidth, trim={150 165 150 250}, clip]{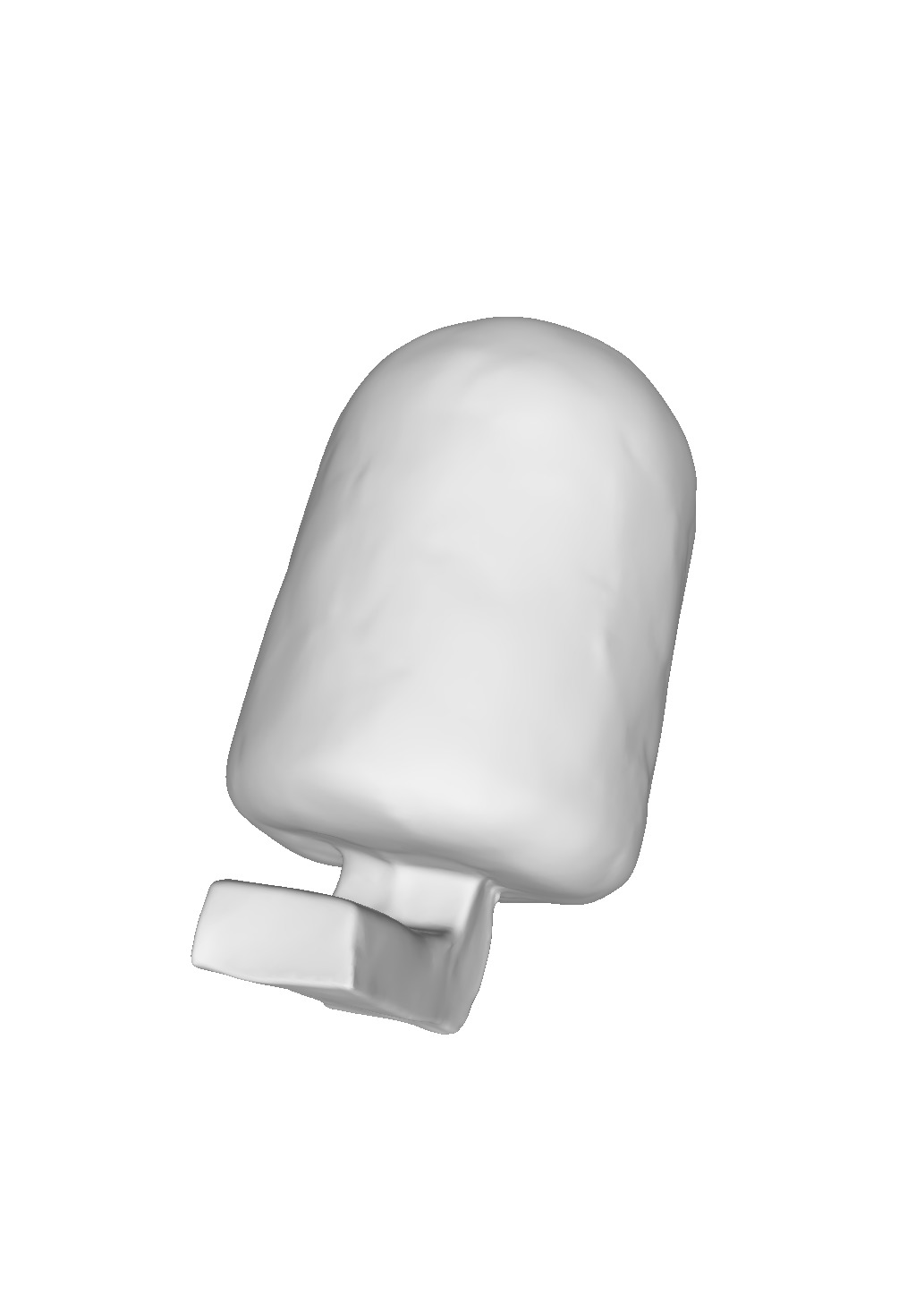}}
\end{small}

%% file: discussion.tex

Our approach optimizes a mesh and a number of meshlets based on the gradients available at mesh vertexes, while enforcing the meshlets priors. In this paper, the gradients for the mesh were obtained by computing the distance of the mesh to the point cloud.
However, our method can take gradients from any source, including a differentiable renderer~\cite{kato2018renderer}.
This adds to the flexibility of our approach.

\begin{figure}
    \vspace{-5mm}
    \centering
    \subfloat[]{\includegraphics[width=.15\columnwidth, trim={0 50 0 50}, clip]{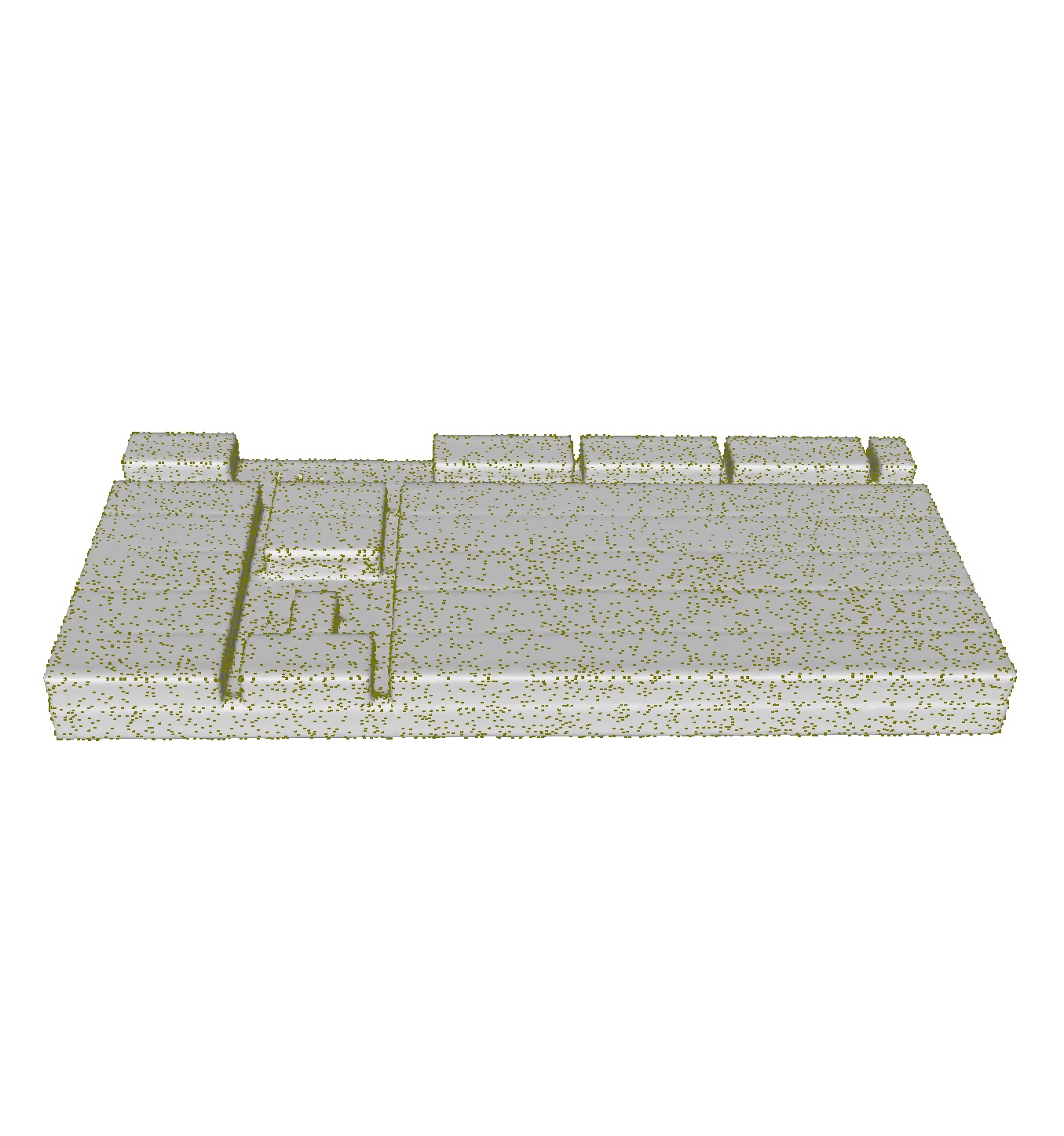}\includegraphics[width=.15\columnwidth, trim={0 50 0 50}, clip]{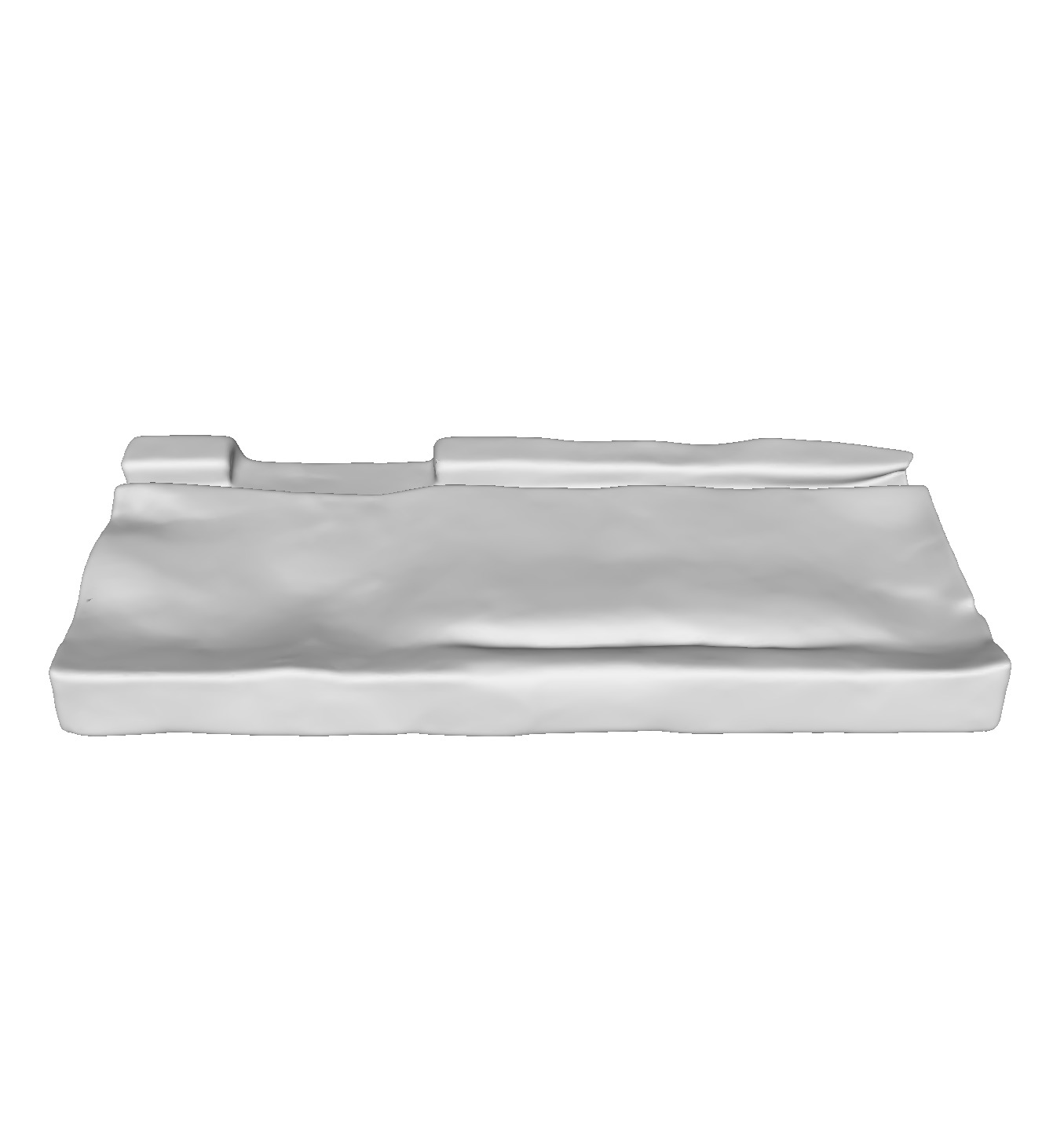}}\hspace{1mm}
    \subfloat[]{\includegraphics[width=.15\columnwidth, trim={0 0 40 0}, clip]{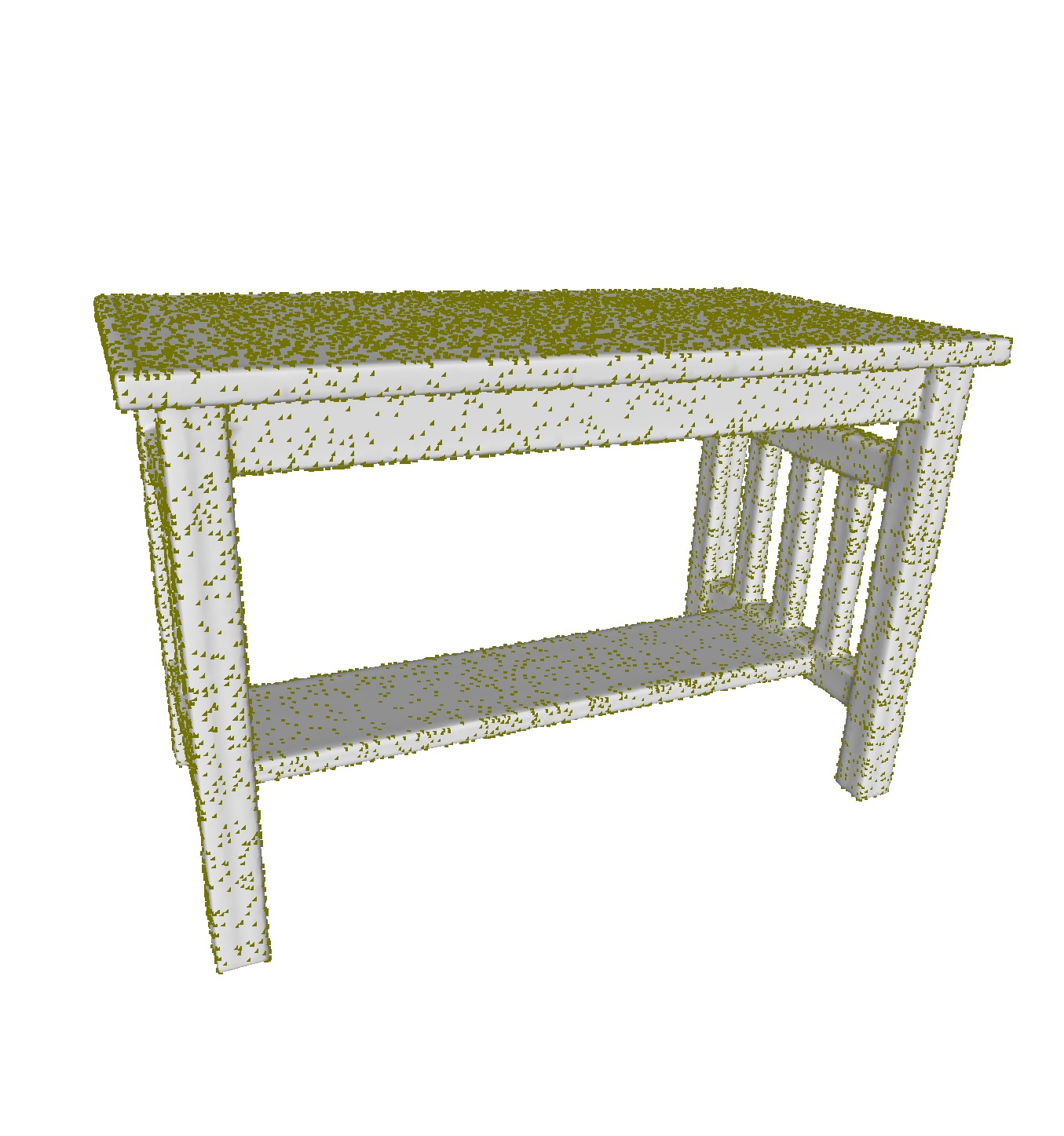}\includegraphics[width=.15\columnwidth, trim={0 125 40 0}, clip]{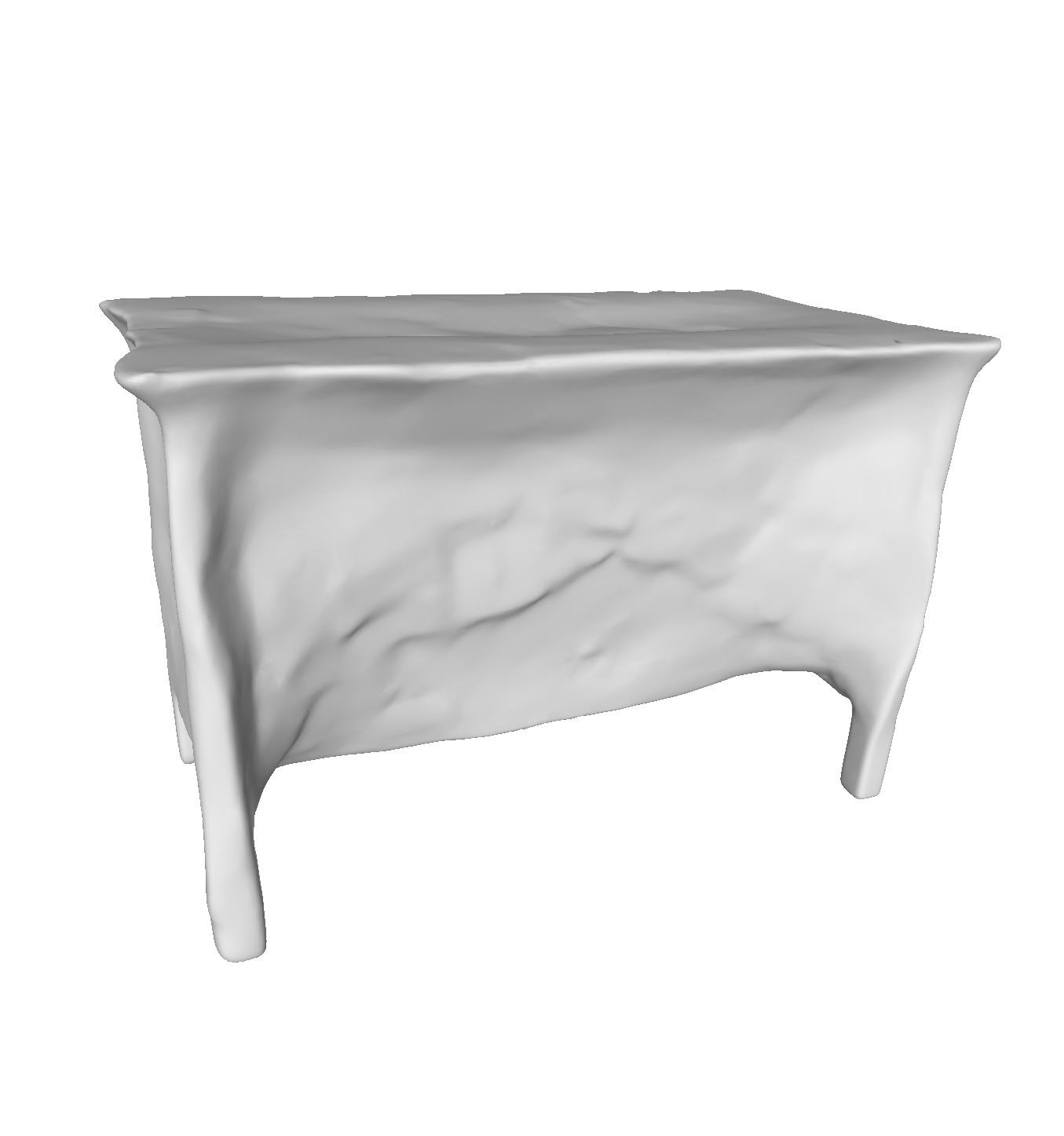}}\hspace{1mm}
    \subfloat[]{\includegraphics[width=.15\columnwidth, trim={120 50 125 175}, clip]{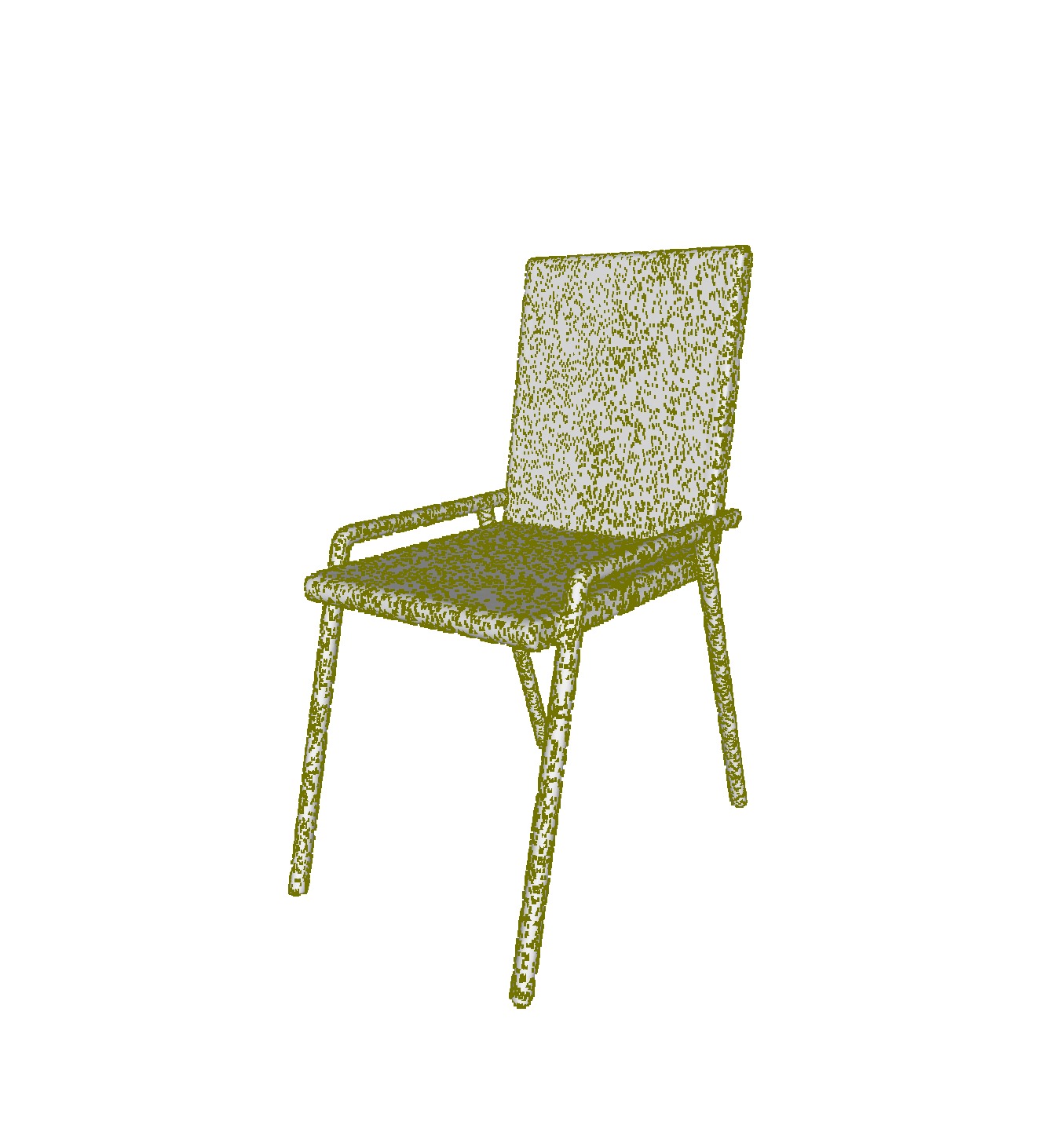}\includegraphics[width=.15\columnwidth, trim={120 175 125 175}, clip]{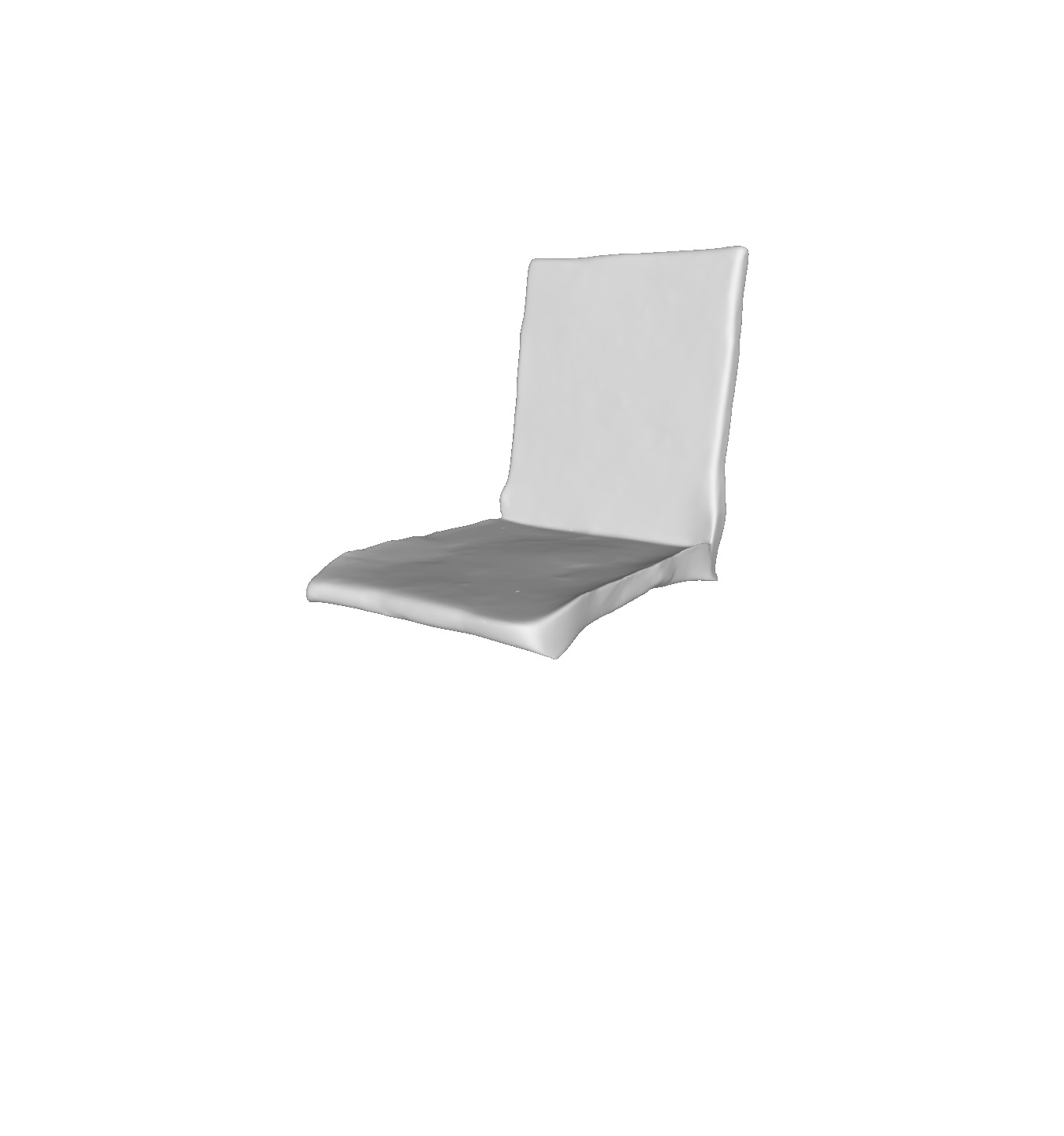}}
    \caption{Our reconstruction method may yield poor reconstructions if when the object features are smaller than the resolution of the meshlets (a), when the optimization falls in a local minimum (b), or for thin structures (c).}\label{fig:failure}
    \vspace{1mm}
\end{figure}

In our work we learn and enforce priors for mesh estimation using meshlets, which have an intrinsic scale and resolution. Our current approach uses a single fixed scale of the meshlet for all the object reconstructions, although we effectively learn meshlets at multiple scales (see Section~\ref{sec:train_details}).
This poses limitations on the level of details we can reconstruct: they cannot be smaller than the resolution of the meshlet.
When this happens, the fine details may not be reconstructed, as shown in Figure~\ref{fig:failure}(a).
Using a meshlet at a single scale throughout the mesh deformation process may also lead to local minima, a particularly pressing problem if the initial mesh is significantly far from the target.
Figure~\ref{fig:failure}(b) shows a reconstruction that fell in a local minimum.
A natural extension to our approach, then, would be to use a coarse-to-fine approach. 
Finally, our approach may fail for objects that present very thin structures, such as those shown in Figure~\ref{fig:failure}(c).

To ensure that our auxiliary mesh stays regular throughout the optimization process, we perform Poisson reconstruction (see Section~\ref{sec:optim_details}).
As a by-product, we inherit its ability to deal with different topologies as can be seen for the Stanford Teacup in Figure~\ref{fig:big_table}, which is genus-one.

Our current approach is computationally expensive and not optimized for speed. Hence, it can take from hours to dozens of hours, depending on the initialization to run the full optimization.
The Chamfer distance computation in Equations~\ref{eq:PC} and \ref{eq:Cm} has quadratic complexity in the number of points and is the bottleneck of our approach.
In the current implementation, we consider all the points in the point cloud every time we compute the distance.
Limiting the search to a local neighborhood would help.
Improving the efficiency of the meshlets extraction is another obvious venue to speed up the algorithm.

%% file: conclusions.tex

We present meshlets, a novel local shape representation that allows to reconstruct 3D meshes from sparse, noisy point clouds.
To do this, we train a variational autoencoder to learn the manifold of naturally-occurring meshlets.
Meshlets on this manifold act as priors for local, class-agnostic, natural features.
Therefore, meshlets allow us to disentangle the overall pose of the object from its shape, and can be used to reconstruct objects from classes not seen in training.
To reconstruct a full mesh we use a number of meshlets.
We therefore propose an alternating optimization procedure that first optimizes the meshlets to match the points (locally) and then enforces their consistency (globally).
Our algorithm reconstructs objects from classes not seen in training, in arbitrary poses, and under significant noise and sparsity of the input points, even when existing state-of-the-art methods fail.

%% file: meshlets_cvpr20_supplementary_additional_exp_details.tex

In this section we give additional details about the experiments shown in the main paper.

\subsection{Generating Water-tight Meshes}
We used meshlets extracted from the ShapeNet~\cite{shapenet2015} dataset for training. 
The test dataset for mesh reconstruction was formed by selecting objects from the test set of ShapeNet dataset as well as few objects from outside the ShapeNet. 
However, most ShapeNet objects are not watertight. 
To generate water-tight meshes for our objects we used the process described by Stutz and Geiger~\cite{Stutz2018ARXIV}. 
First, the object is scaled to lie in $[-1,\ 1]^3$. 
Next, depth maps are rendered from 200 views and with a resolution of $1024\times 1024$. 
These depth maps are used to perform TSDF fusion. We use $512\times 512\times 512$ volume for TSDF fusion. 
Finally a mesh simplification step is performed using meshlab to give us a final mesh with 50k vertices. 
These vertices are roughly uniform over the surface of the mesh.

\subsection{Noise, Sparsity and Outliers Parameters used in Experiments}
To test different approaches we designed three different settings for noise and sparsity. 
Given a GT object with 50k vertices, we first randomly sample a $x\%$ of the vertices to get a sparse point cloud. 
Following this we add a Gaussian noise of magnitude $y$ to each point in the sparse point cloud. 
The three different settings used for the experiments are as follows:
\begin{itemize}
	\item Setting 1 (S1): $x=10\%$ and $y=0.0150$
	\item Setting 2 (S2): $x=20\%$ and $y=0.0225$
	\item Setting 3 (S3): $x=40\%$ and $y=0.0300$
\end{itemize}

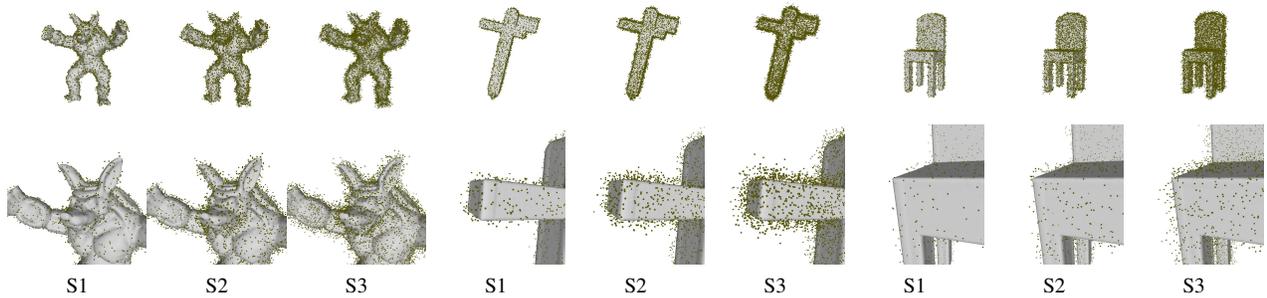
\begin{figure*}[p]
	\input{figures_supplementary/noise_visualization/noise_viz}
	\caption{In this figure we show three different noise and sparsity settings used in our experiments. 
			 Setting S1 has less noise but a more sparse point cloud.
			 Setting S3 has denser but a more noisier point cloud. 
			 These settings are designed to test the robustness of our approach to sparse and noisy observations.}
	\label{fig:noise_viz}
\end{figure*}

To appreciate the level of noise we provide a visualization of these noise settings in Figure~\ref{fig:noise_viz}. 

\subsection{Test dataset and Initialization}
\begin{figure*}[p]
	\input{figures_supplementary/gt_mesh_init_cvpr2020/gt_mesh_init_1}
	\caption{In this figure we show the initialization mesh $\mathcal{M}(t_0)$ and GT pair for all of the test objects. 
	We use $\mathcal{M}(t_0)$ to initialize our method, as well as the Laplacian mesh optimization.}
	\label{fig:gt_mesh_init_1}
\end{figure*}
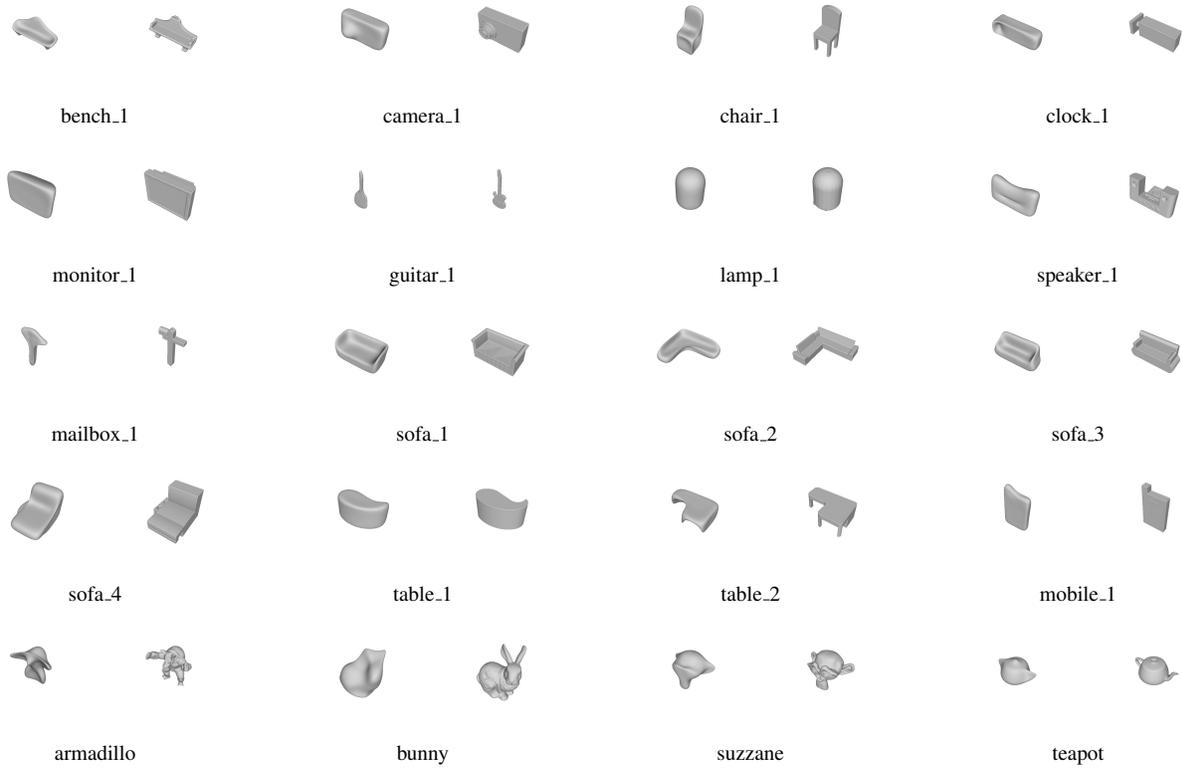

In Figure~\ref{fig:gt_mesh_init_1} we show all the GT objects used for the mesh reconstruction evaluation.
We also show the initial mesh, $\mathcal{M}(t_0)$. 
Note that we used the exact same $\mathcal{M}(t_0)$ to initialize both our method and the Laplacian mesh optimization.

\

%% file: figures_supplementary/gt_mesh_init_cvpr2020/gt_mesh_init_1.tex

\setlength{\numcrops}{8pt}
\newcommand{\fitscale}{.8}

\settowidth{\cropwidth}{\includegraphics{figures/results/teapot_new/snapshot00.jpg}}

\setlength{\one}{1pt}

\setlength\tgtwidth{\textwidth*\ratio{\one}{\numcrops}}
\newlength{\spacing}
\setlength{\spacing}{6mm}

\captionsetup[subfigure]{labelformat=empty}
\centering
\vspace{-4mm}
\subfloat[bench\_1]{
    \includegraphics[width=\fitscale\tgtwidth, trim={262 76 319  76}, clip]{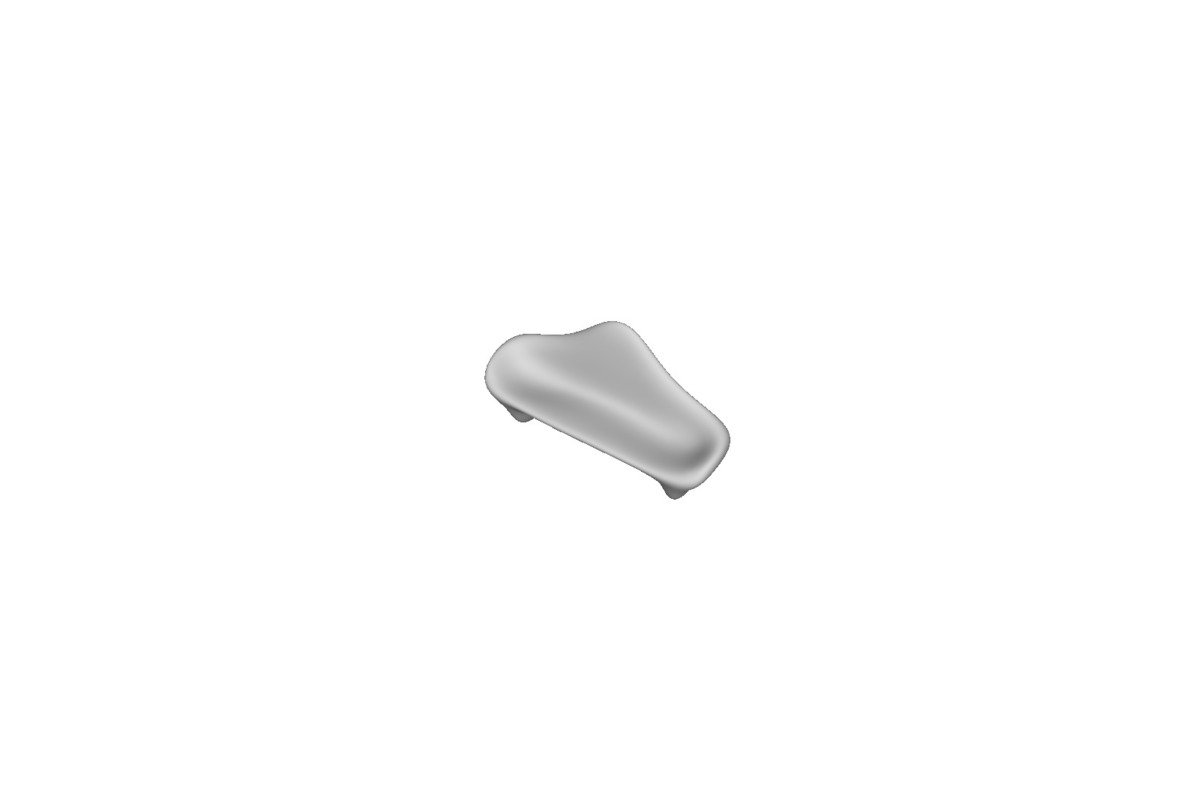}
    \includegraphics[width=\fitscale\tgtwidth, trim={262 76 319  76}, clip]{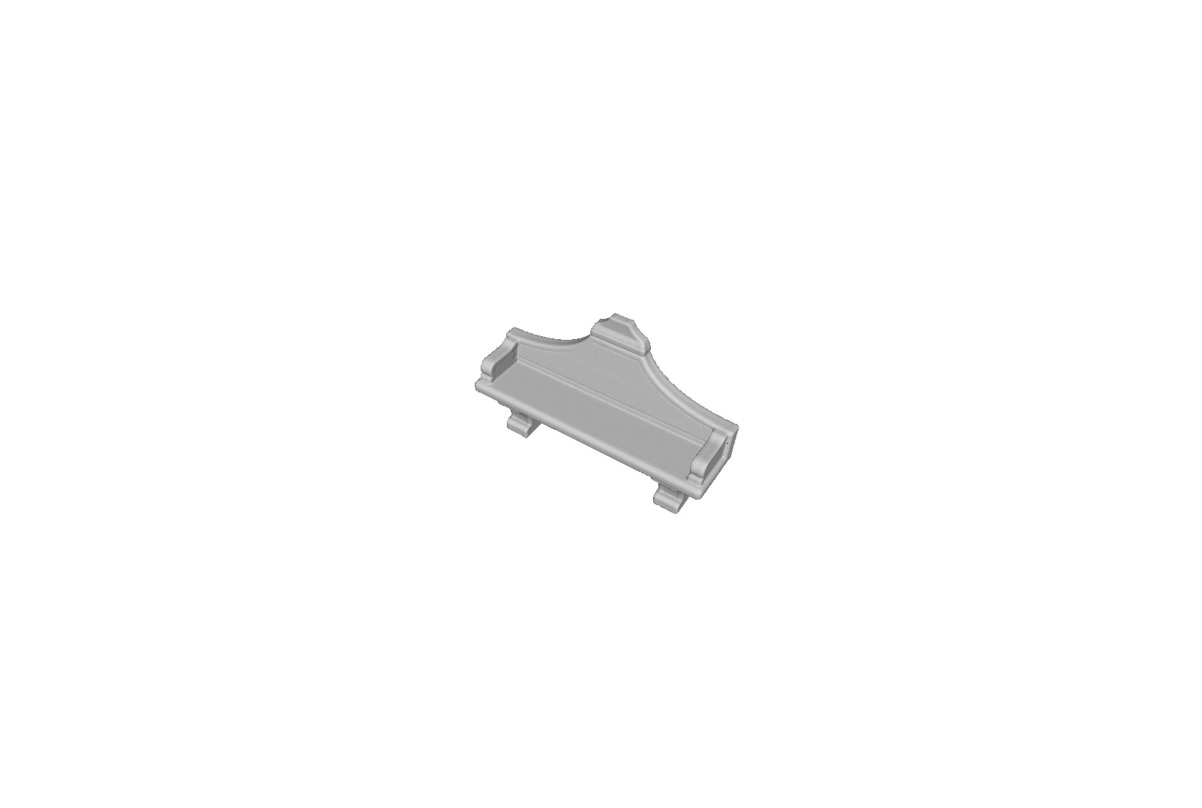}}\hspace{\spacing}
\subfloat[camera\_1]{
    \includegraphics[width=\fitscale\tgtwidth, trim={262 76 319  76}, clip]{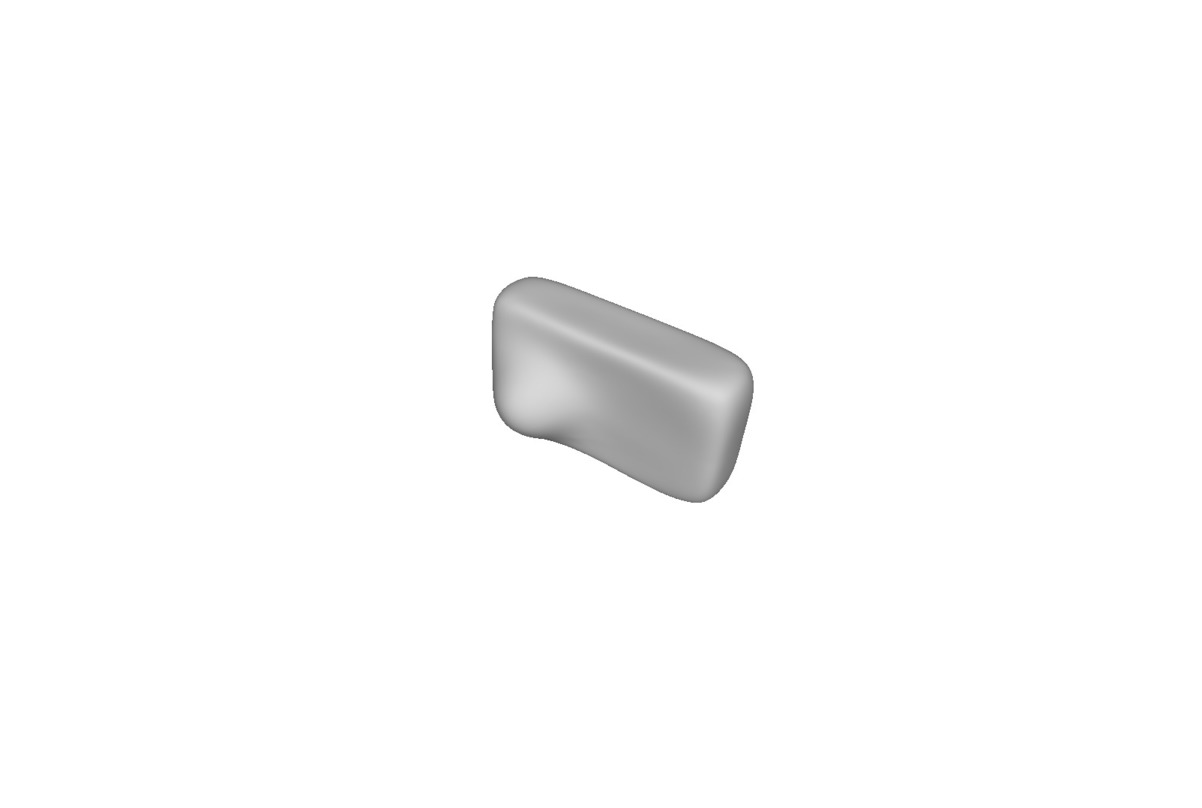}
    \includegraphics[width=\fitscale\tgtwidth, trim={262 76 319  76}, clip]{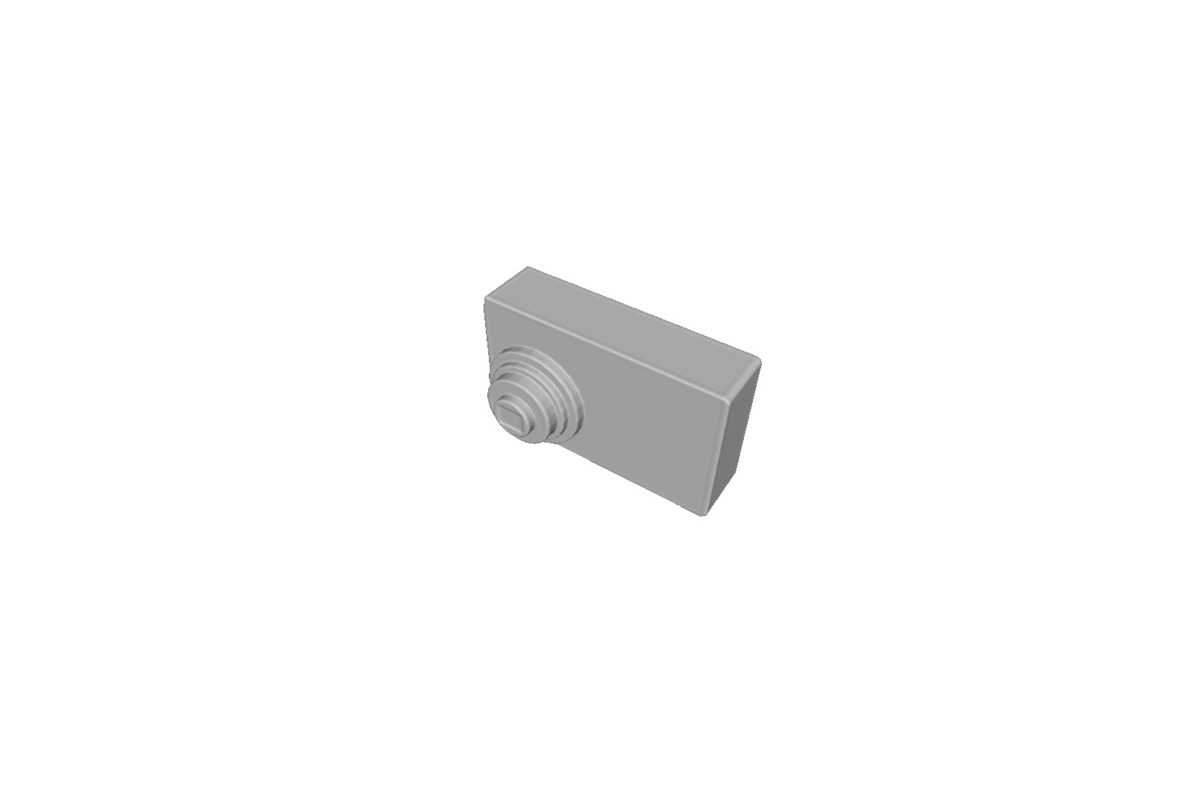}}\hspace{\spacing}
\subfloat[chair\_1]{
    \includegraphics[width=\fitscale\tgtwidth, trim={262 76 319  76}, clip]{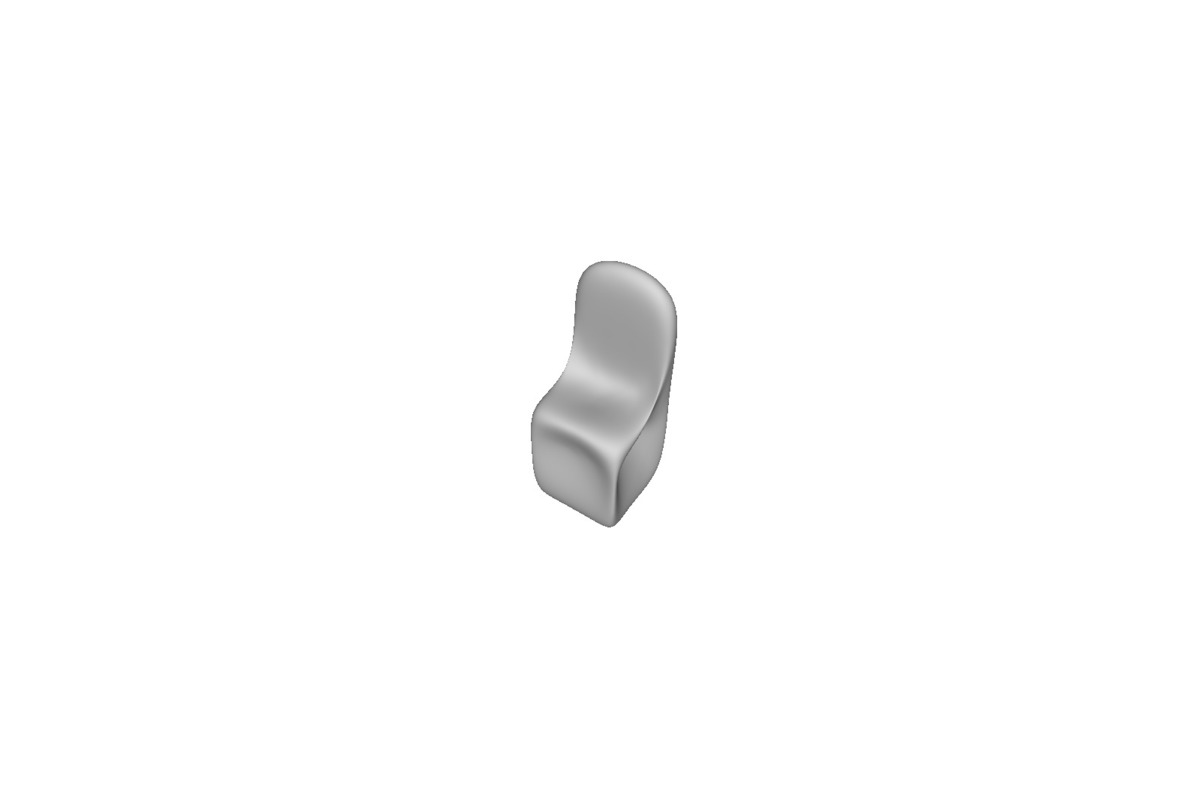}
    \includegraphics[width=\fitscale\tgtwidth, trim={262 76 319  76}, clip]{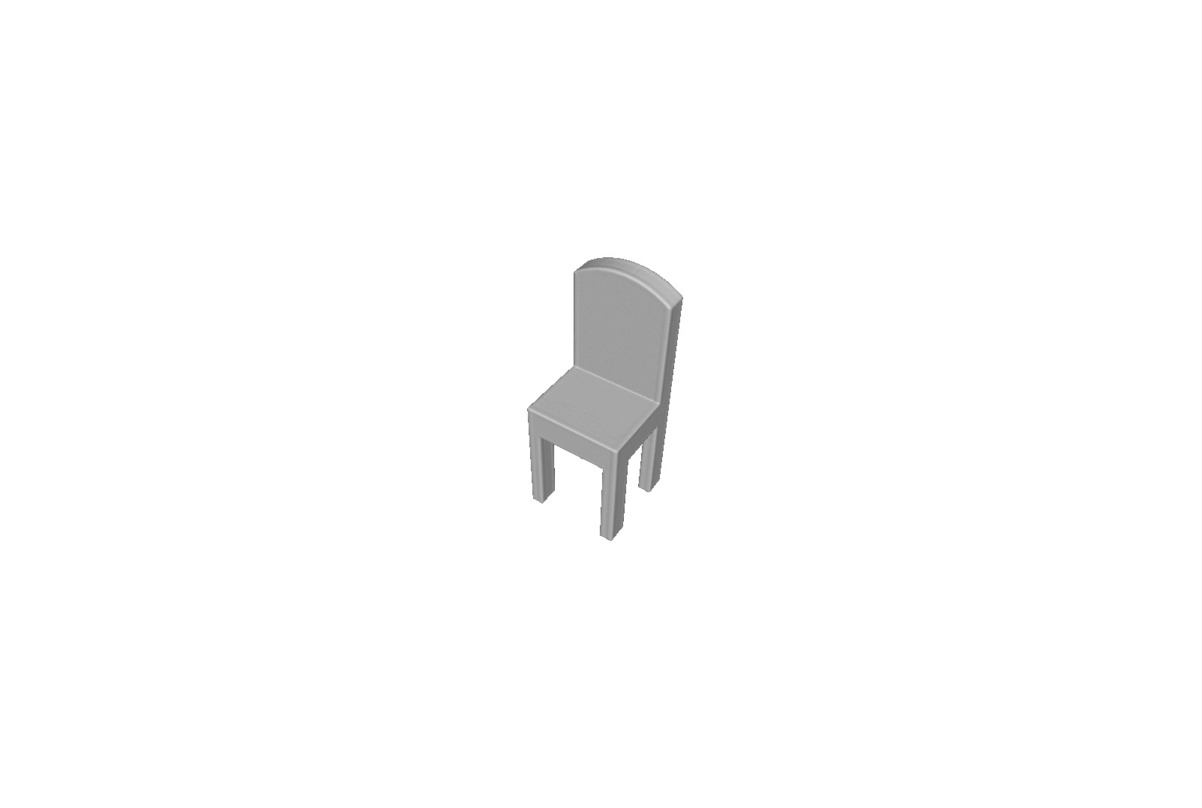}}\hspace{\spacing}
\subfloat[clock\_1]{
    \includegraphics[width=\fitscale\tgtwidth, trim={262 76 319  76}, clip]{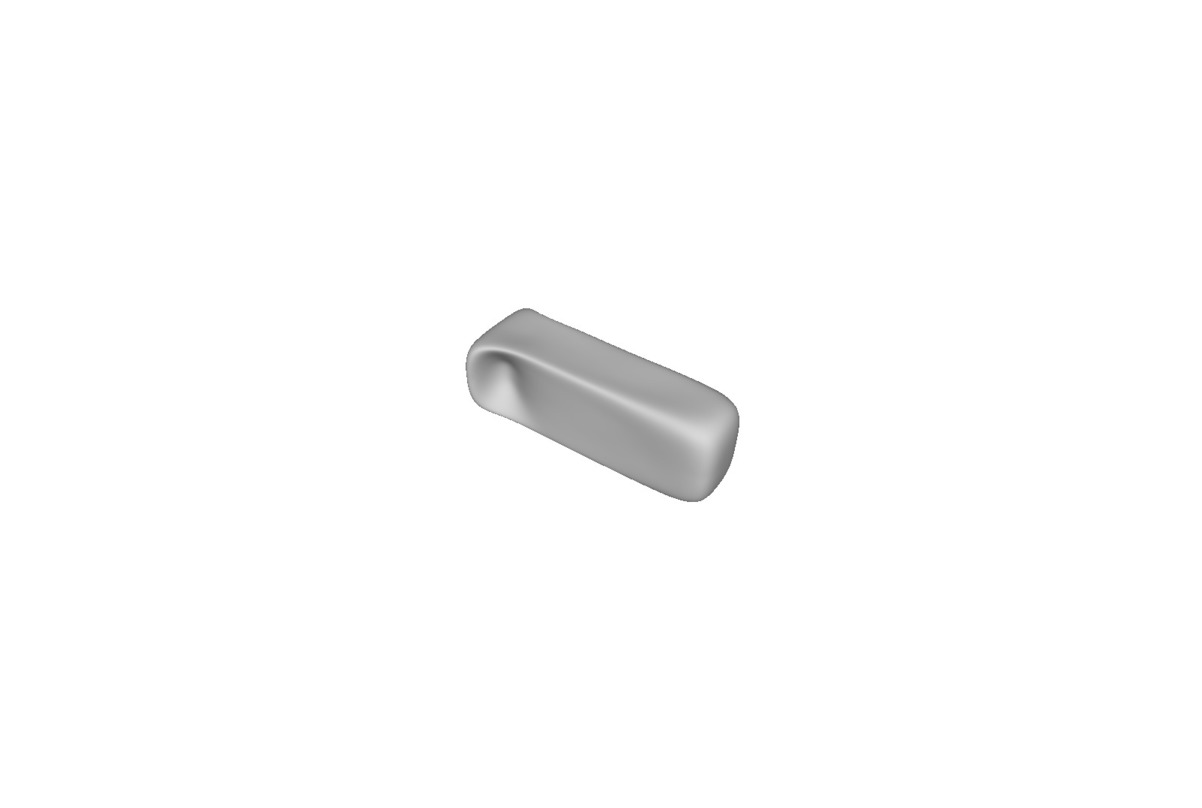}
    \includegraphics[width=\fitscale\tgtwidth, trim={262 76 319  76}, clip]{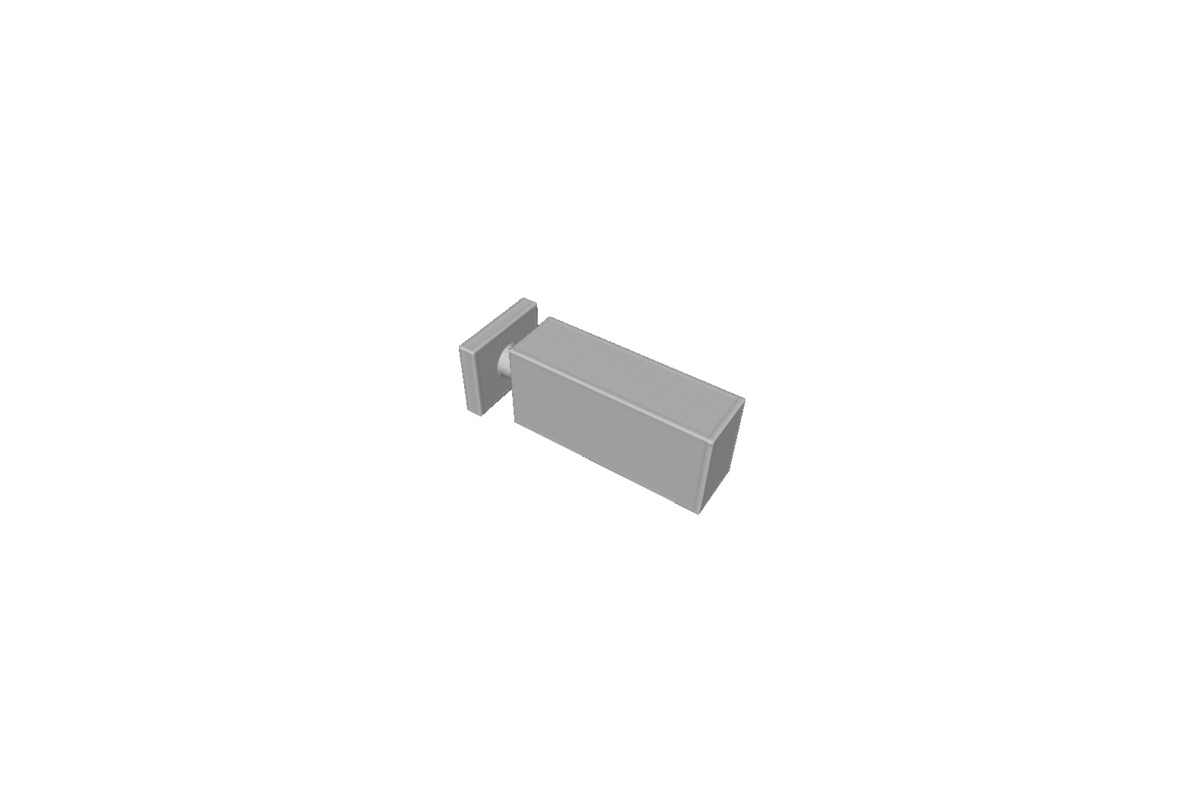}}
\\
\vspace{-4mm}
\subfloat[monitor\_1]{
    \includegraphics[width=\fitscale\tgtwidth, trim={262 76 319  76}, clip]{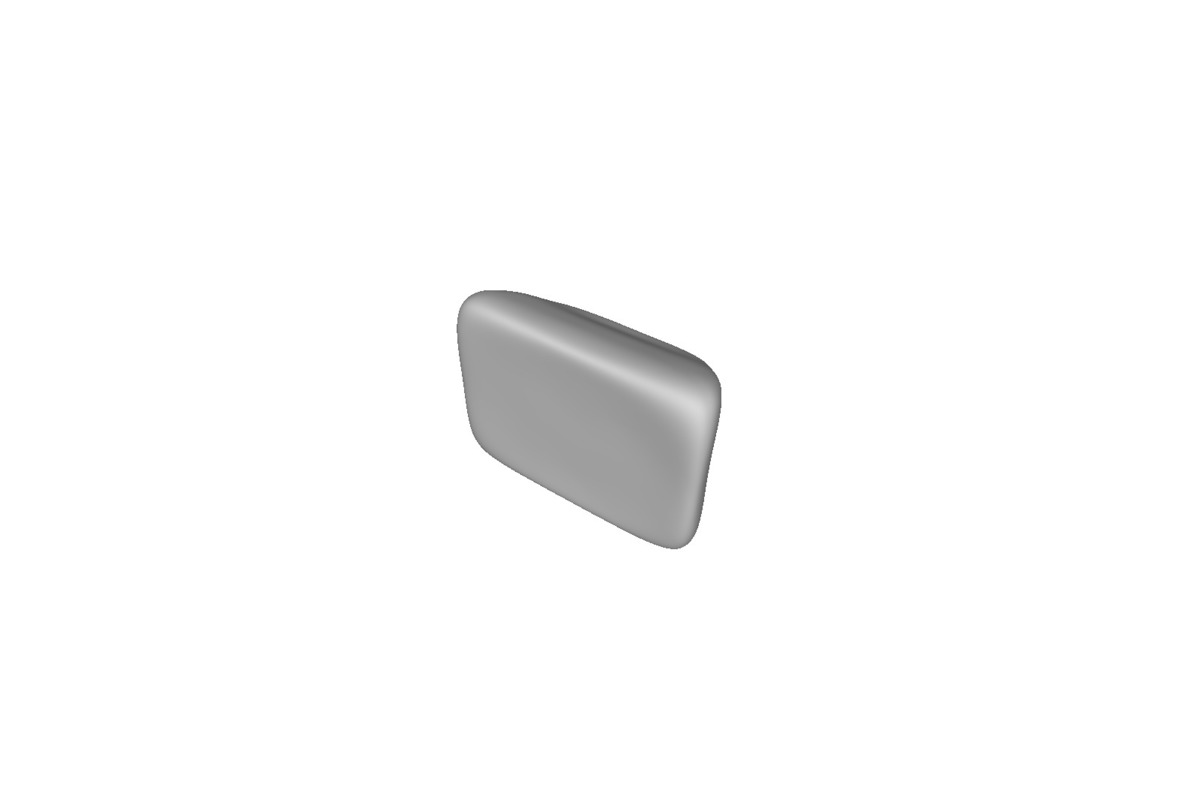}
    \includegraphics[width=\fitscale\tgtwidth, trim={262 76 319  76}, clip]{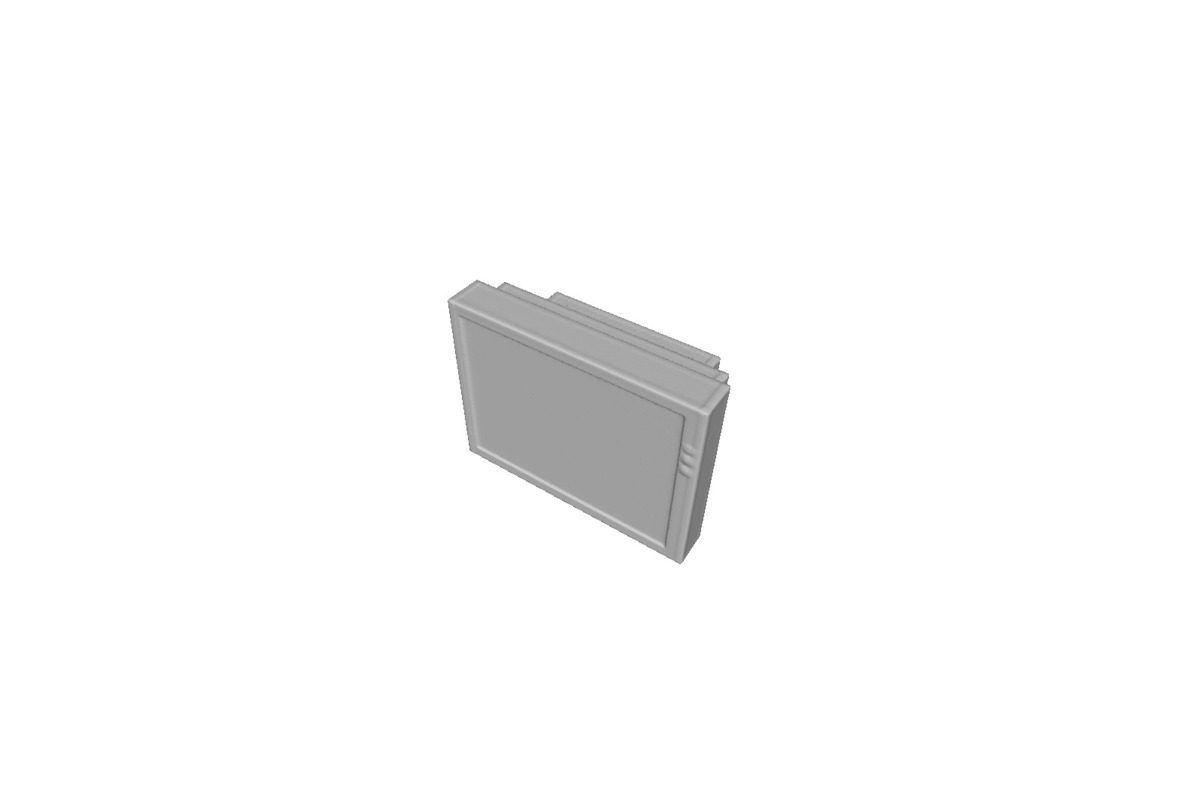}}\hspace{\spacing}
\subfloat[guitar\_1]{
    \includegraphics[width=\fitscale\tgtwidth, trim={262 76 319  76}, clip]{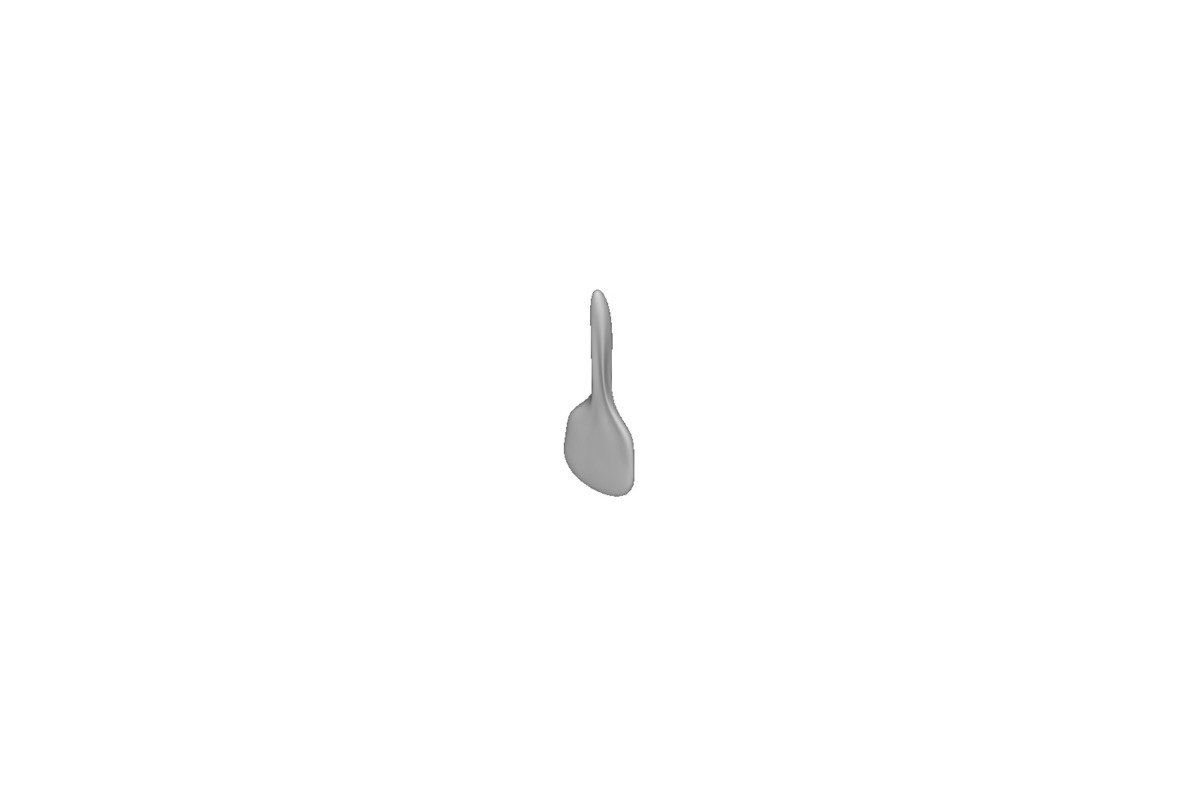}
    \includegraphics[width=\fitscale\tgtwidth, trim={262 76 319  76}, clip]{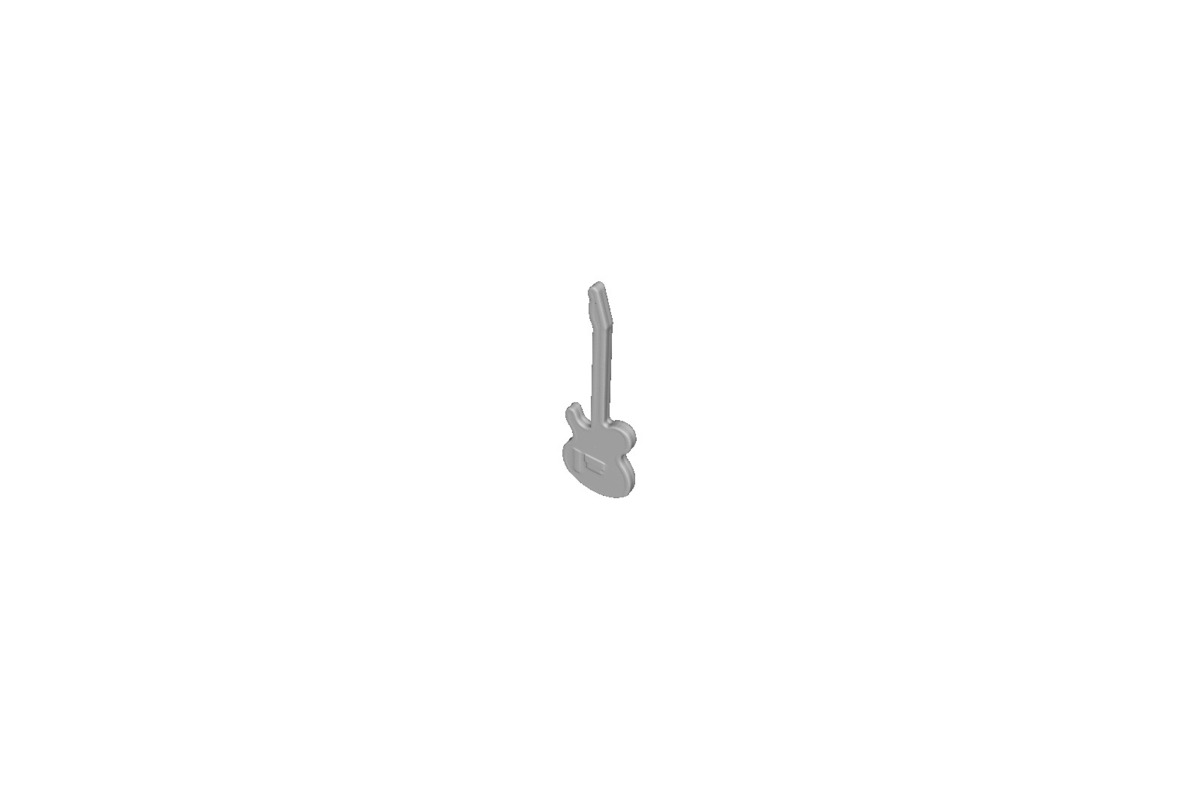}}\hspace{\spacing}
\subfloat[lamp\_1]{
    \includegraphics[width=\fitscale\tgtwidth, trim={262 76 319  76}, clip]{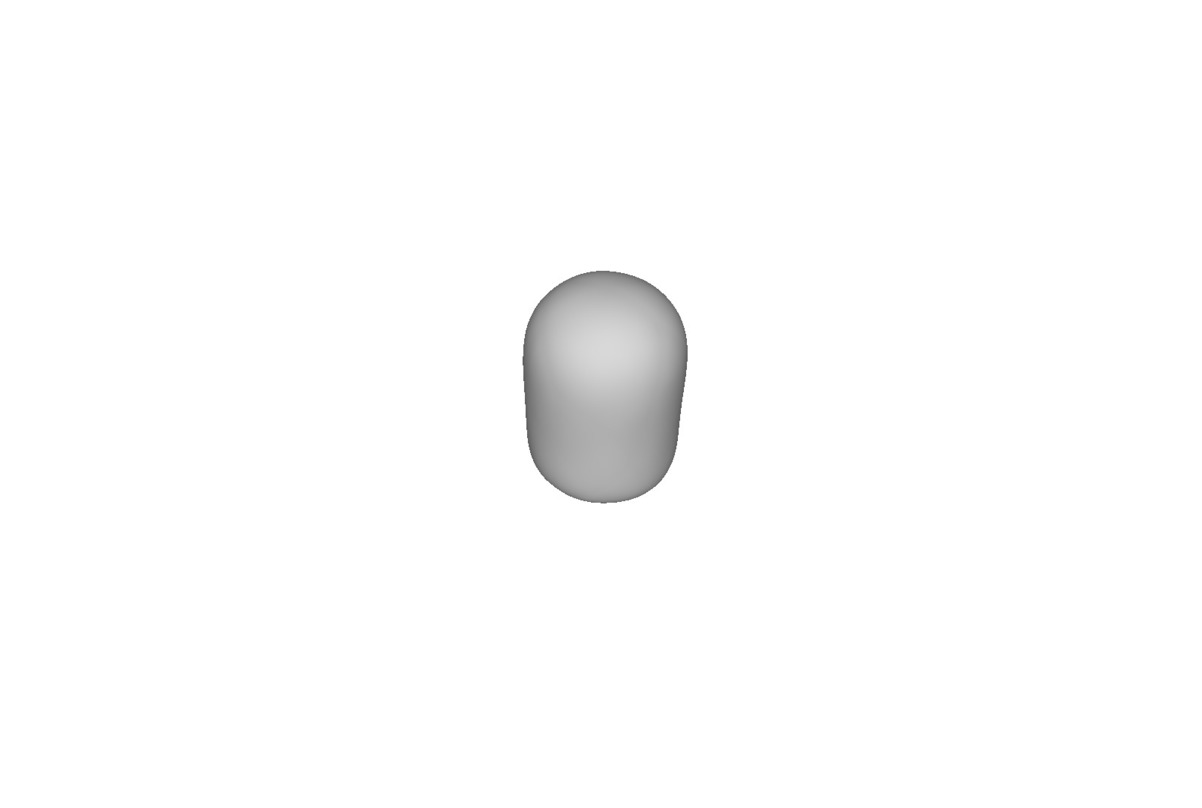}
    \includegraphics[width=\fitscale\tgtwidth, trim={262 76 319  76}, clip]{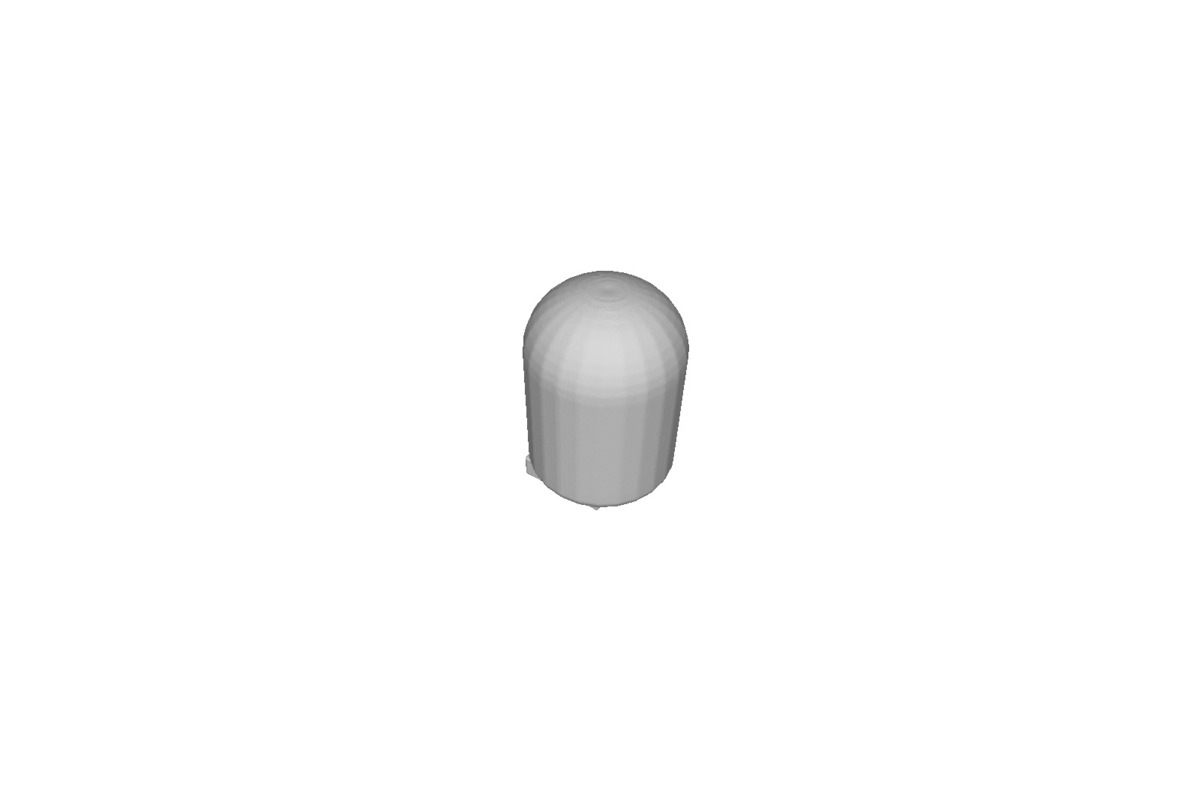}}\hspace{\spacing}
\subfloat[speaker\_1]{
    \includegraphics[width=\fitscale\tgtwidth, trim={262 76 319  76}, clip]{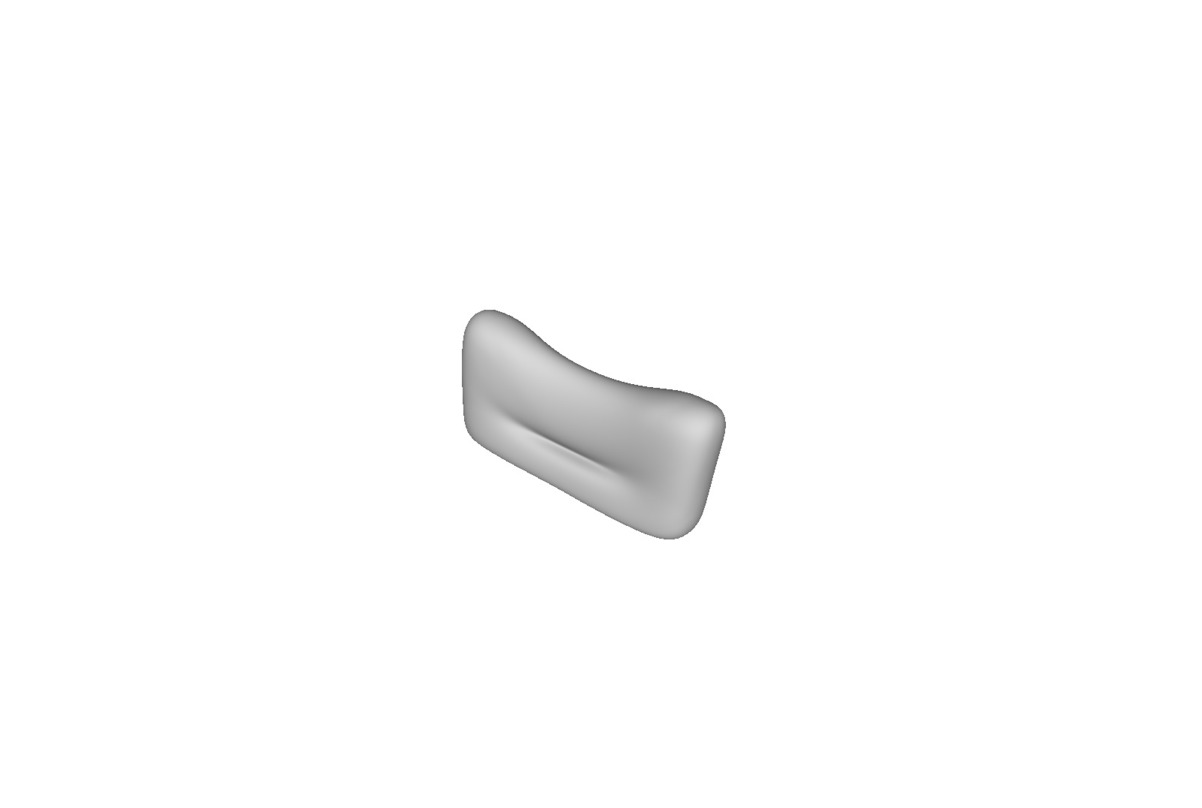}
    \includegraphics[width=\fitscale\tgtwidth, trim={262 76 319  76}, clip]{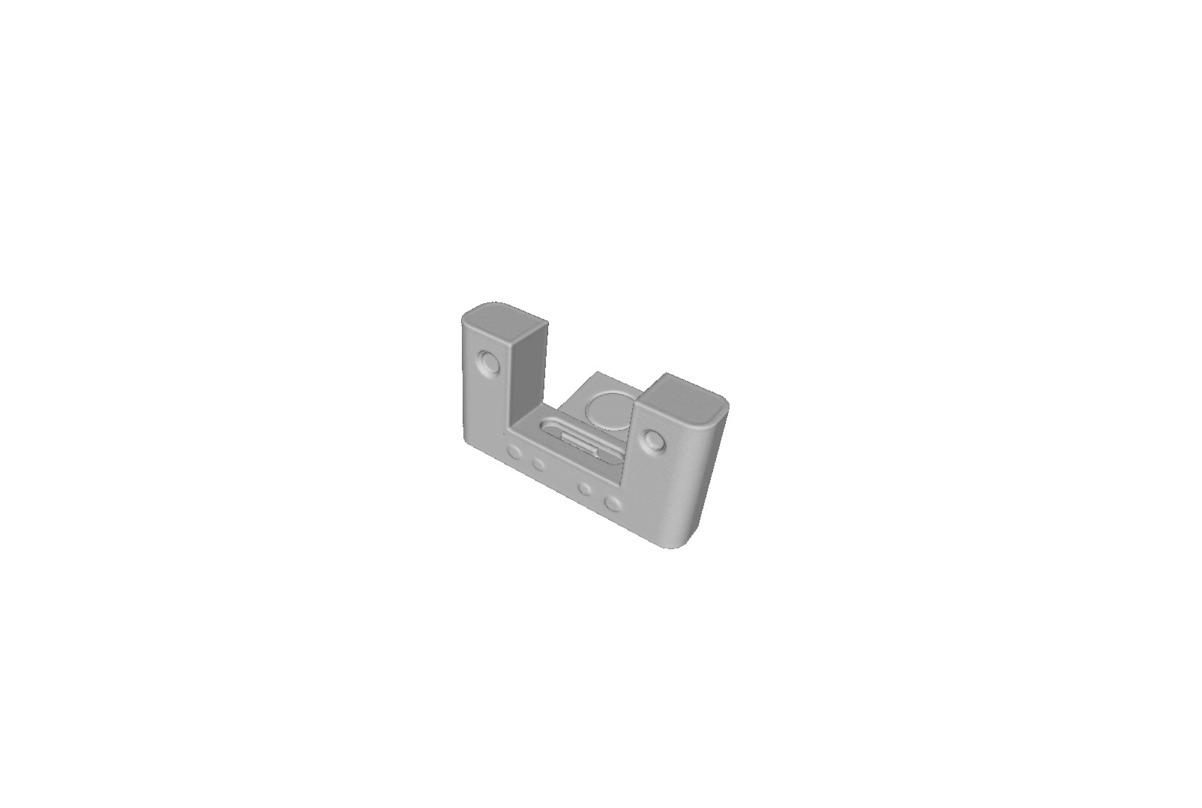}}
\\
\vspace{-4mm}
\subfloat[mailbox\_1]{
    \includegraphics[width=\fitscale\tgtwidth, trim={262 76 319  76}, clip]{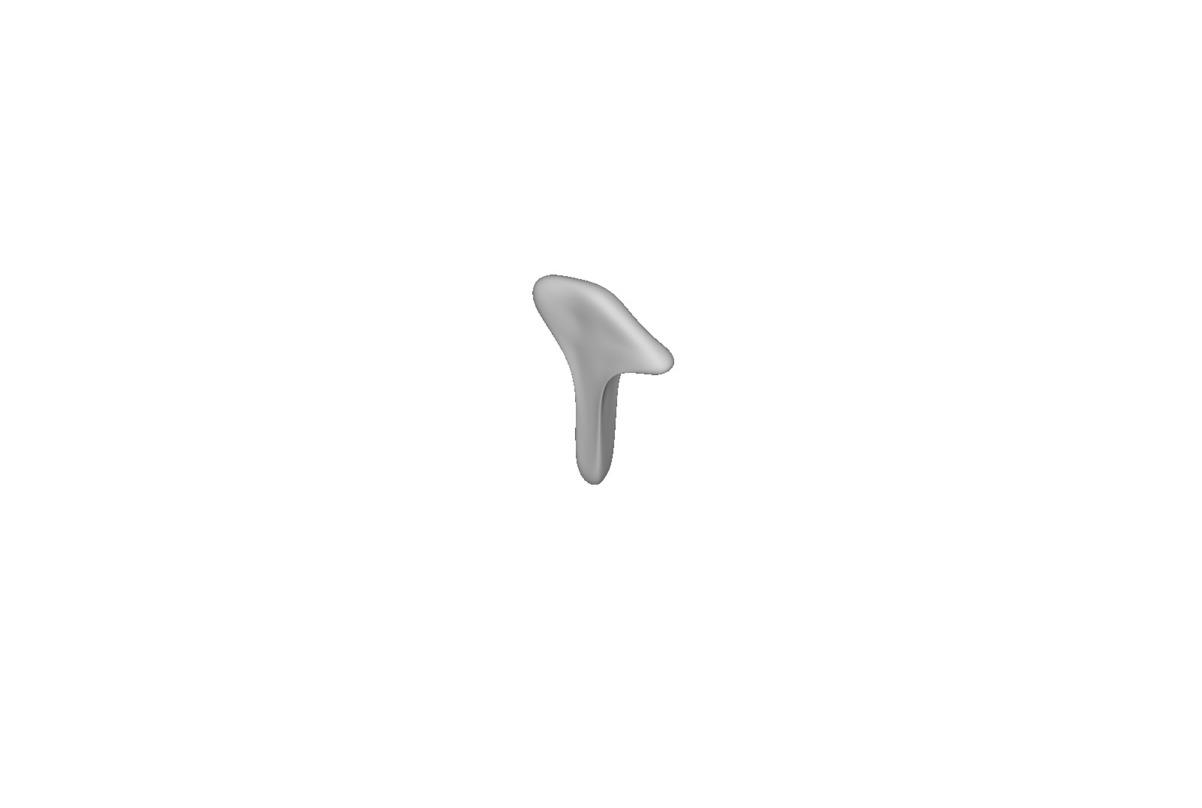}
    \includegraphics[width=\fitscale\tgtwidth, trim={262 76 319  76}, clip]{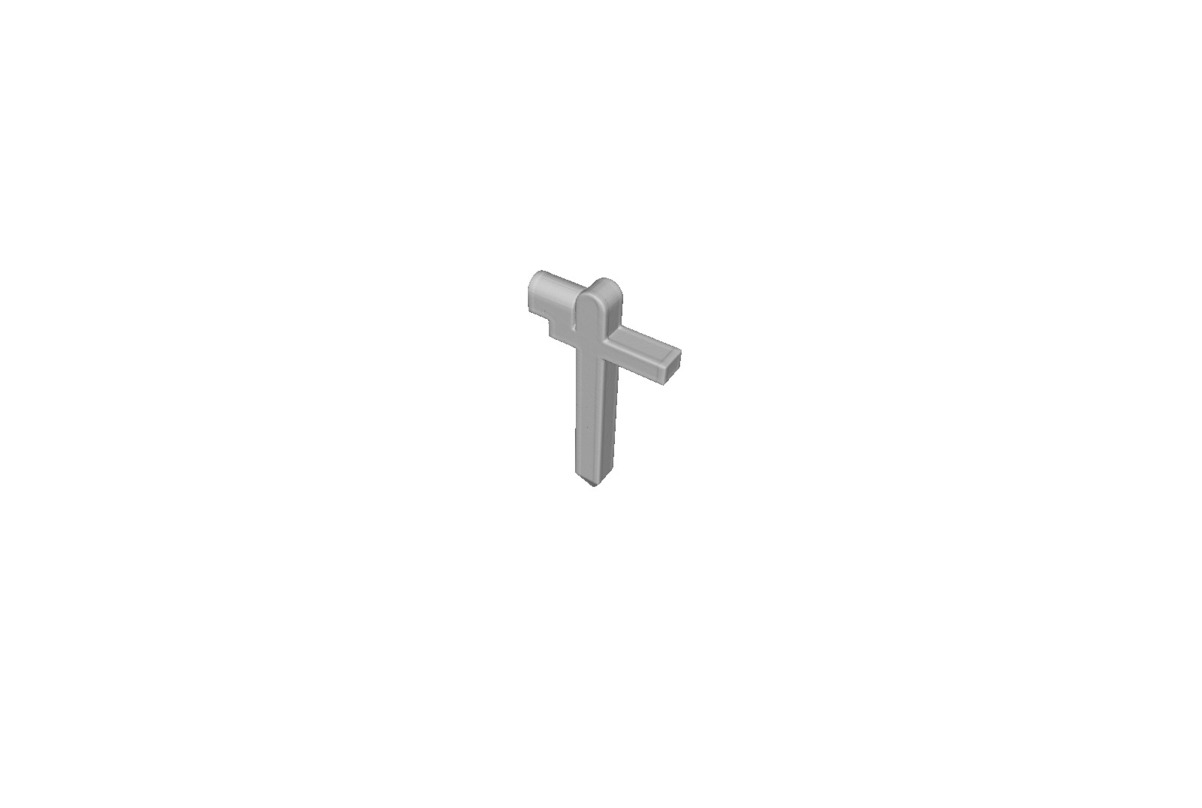}}\hspace{\spacing}
\subfloat[sofa\_1]{
    \includegraphics[width=\fitscale\tgtwidth, trim={262 76 319  76}, clip]{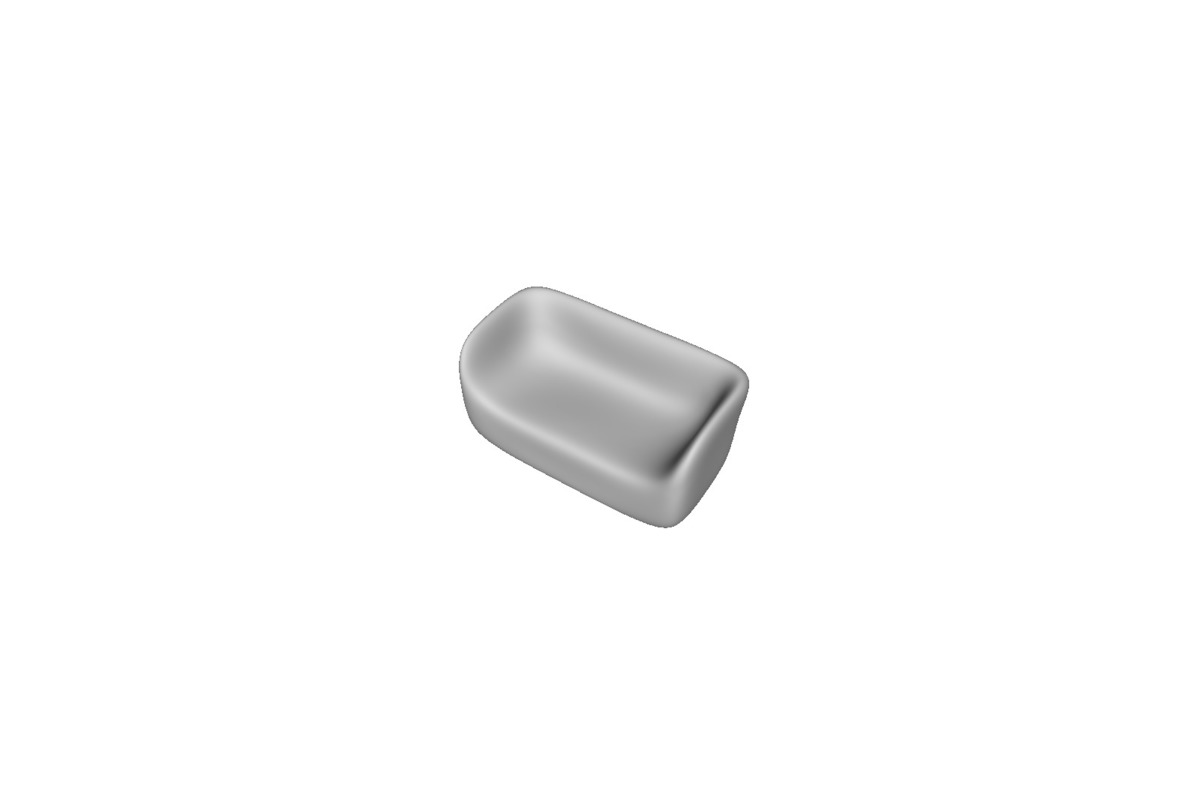}
    \includegraphics[width=\fitscale\tgtwidth, trim={262 76 319  76}, clip]{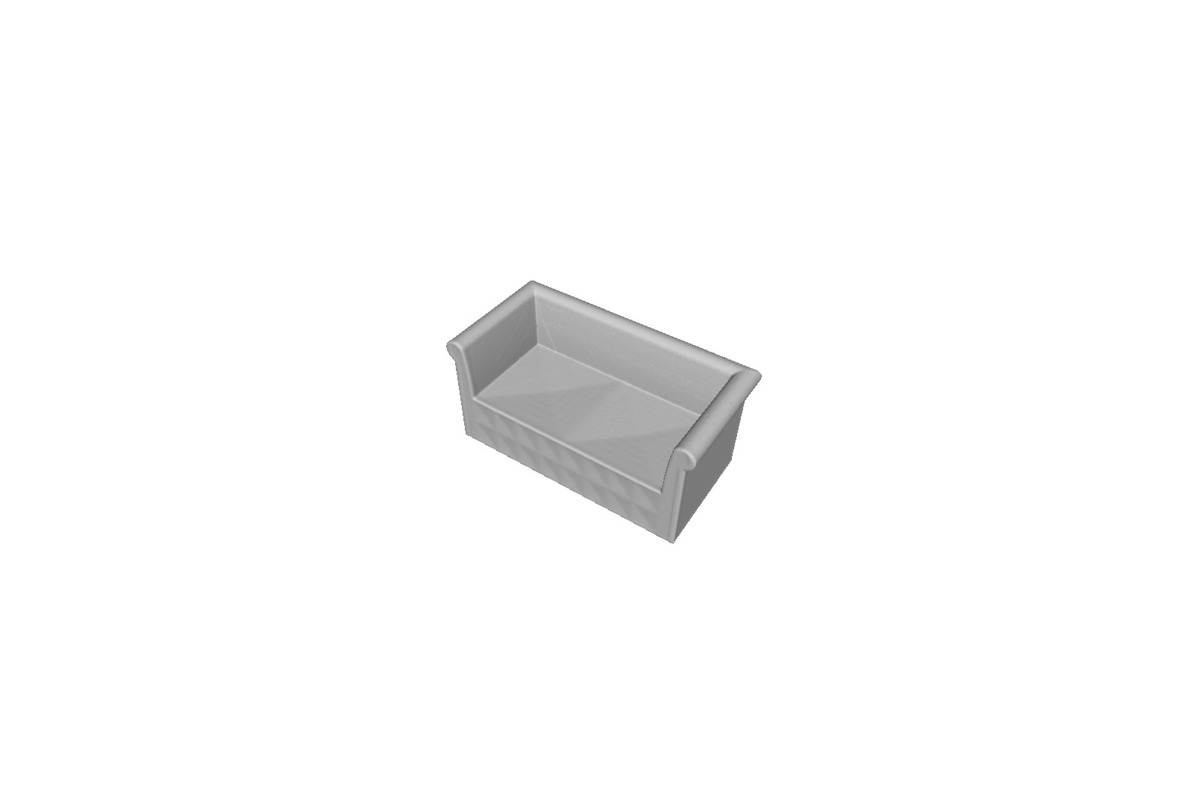}}\hspace{\spacing}
\subfloat[sofa\_2]{
    \includegraphics[width=\fitscale\tgtwidth, trim={262 76 319  76}, clip]{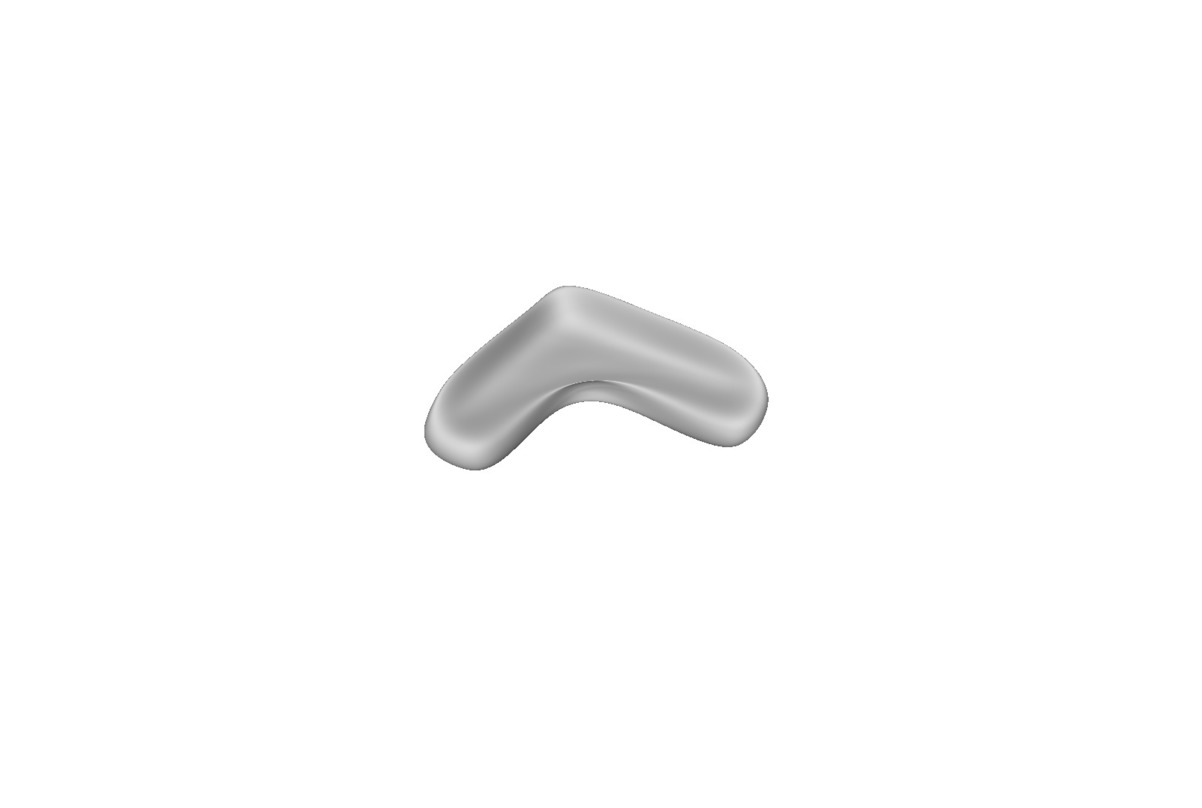}
    \includegraphics[width=\fitscale\tgtwidth, trim={262 76 319  76}, clip]{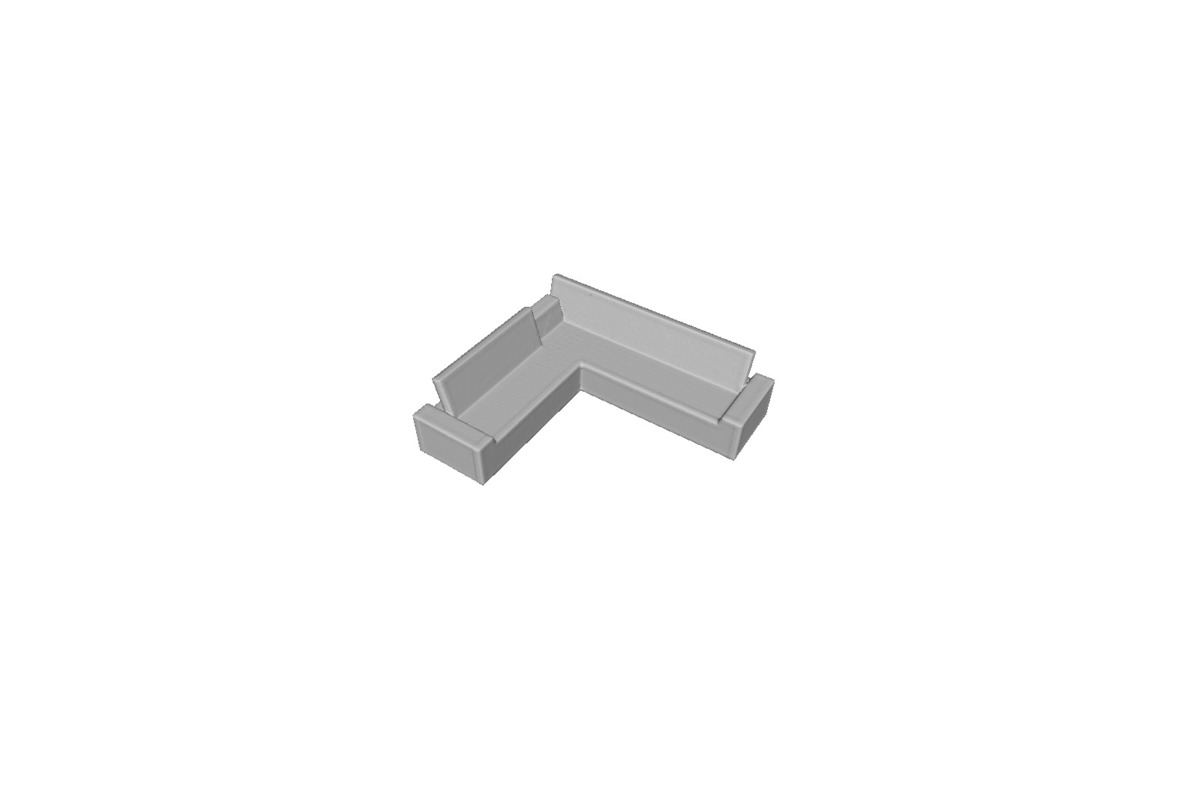}}\hspace{\spacing}
\subfloat[sofa\_3]{
    \includegraphics[width=\fitscale\tgtwidth, trim={262 76 319  76}, clip]{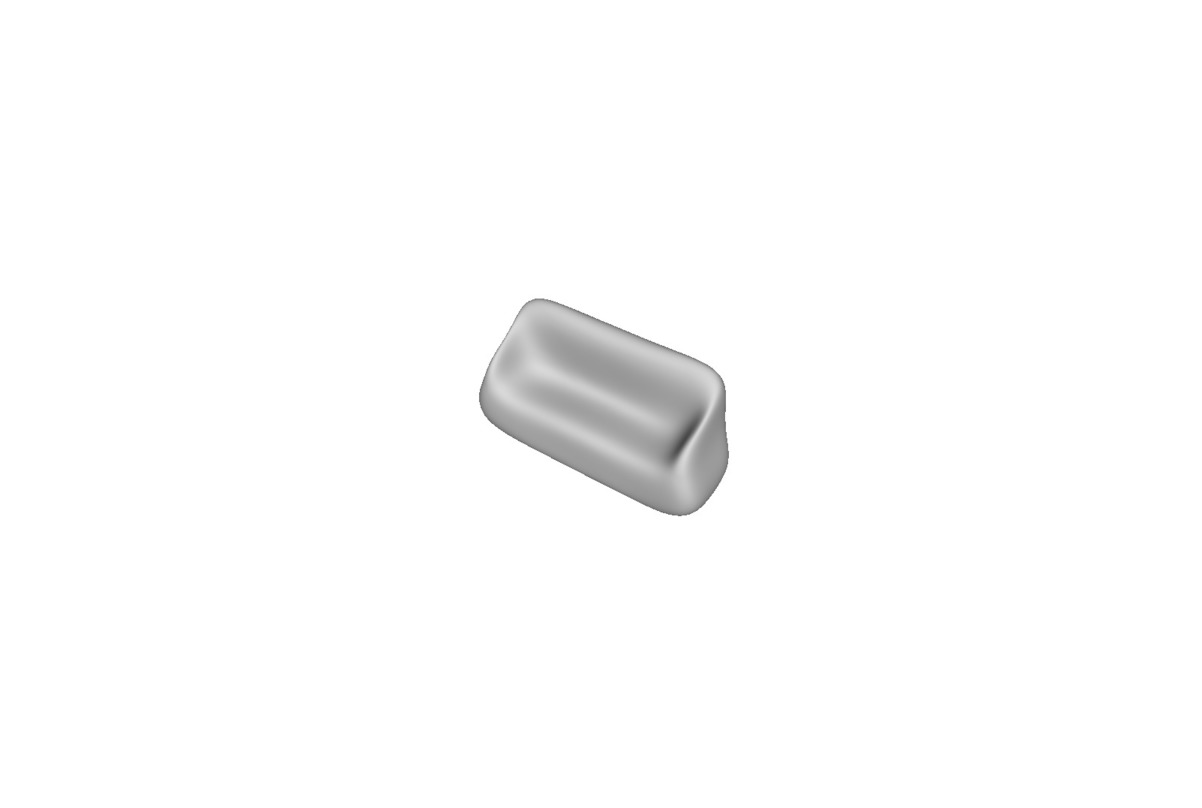}
    \includegraphics[width=\fitscale\tgtwidth, trim={262 76 319  76}, clip]{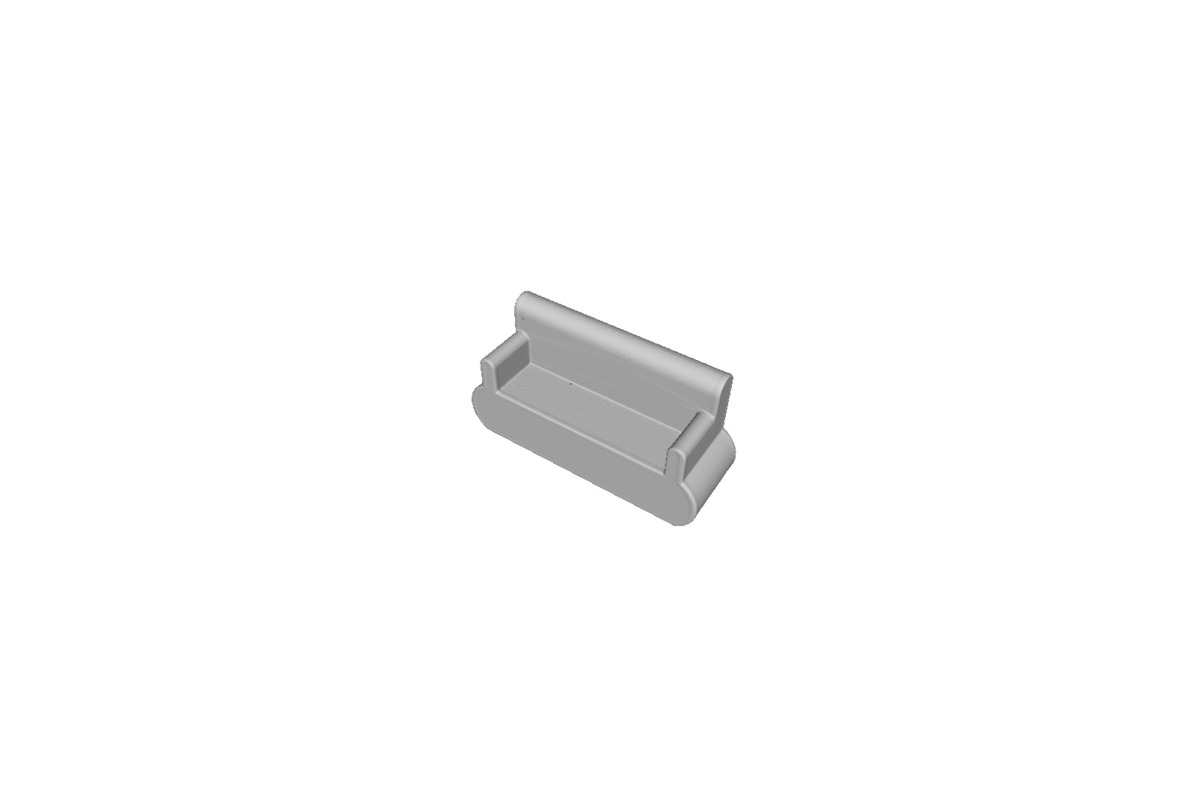}}
\\
\vspace{-4mm}
\subfloat[sofa\_4]{
    \includegraphics[width=\fitscale\tgtwidth, trim={262 76 319  76}, clip]{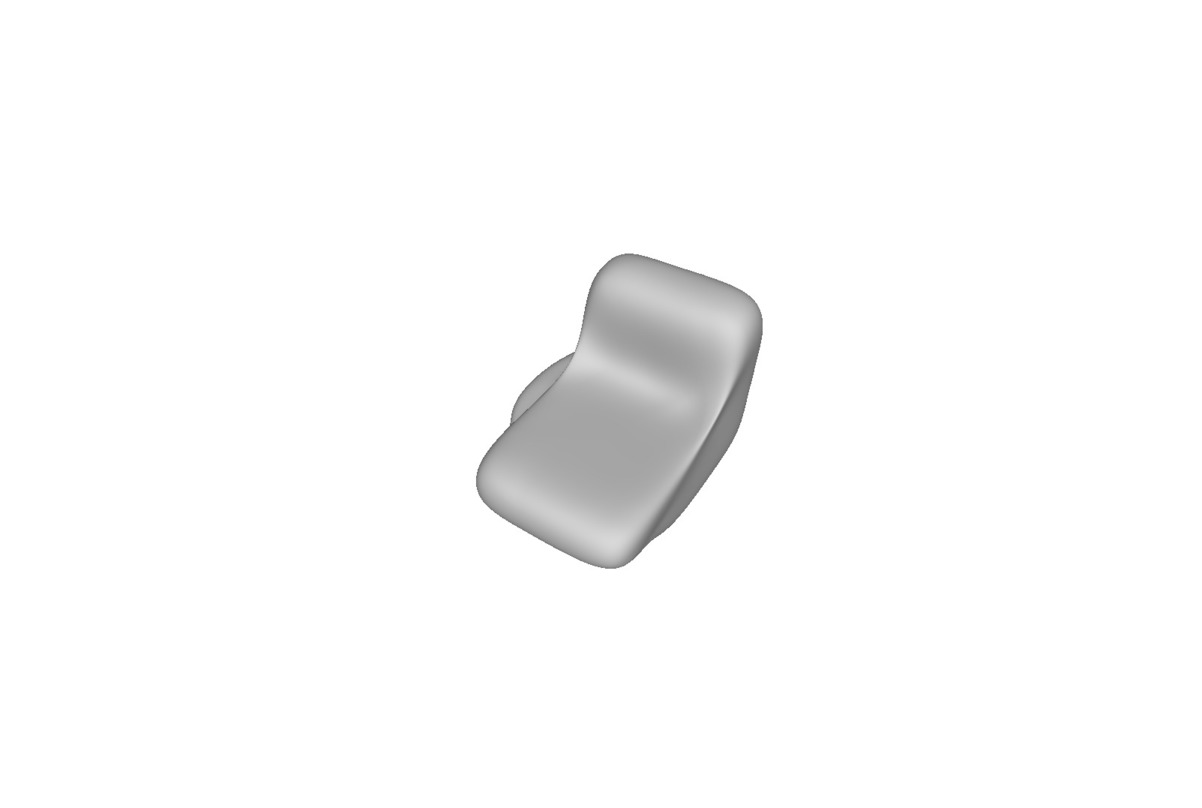}
    \includegraphics[width=\fitscale\tgtwidth, trim={262 76 319  76}, clip]{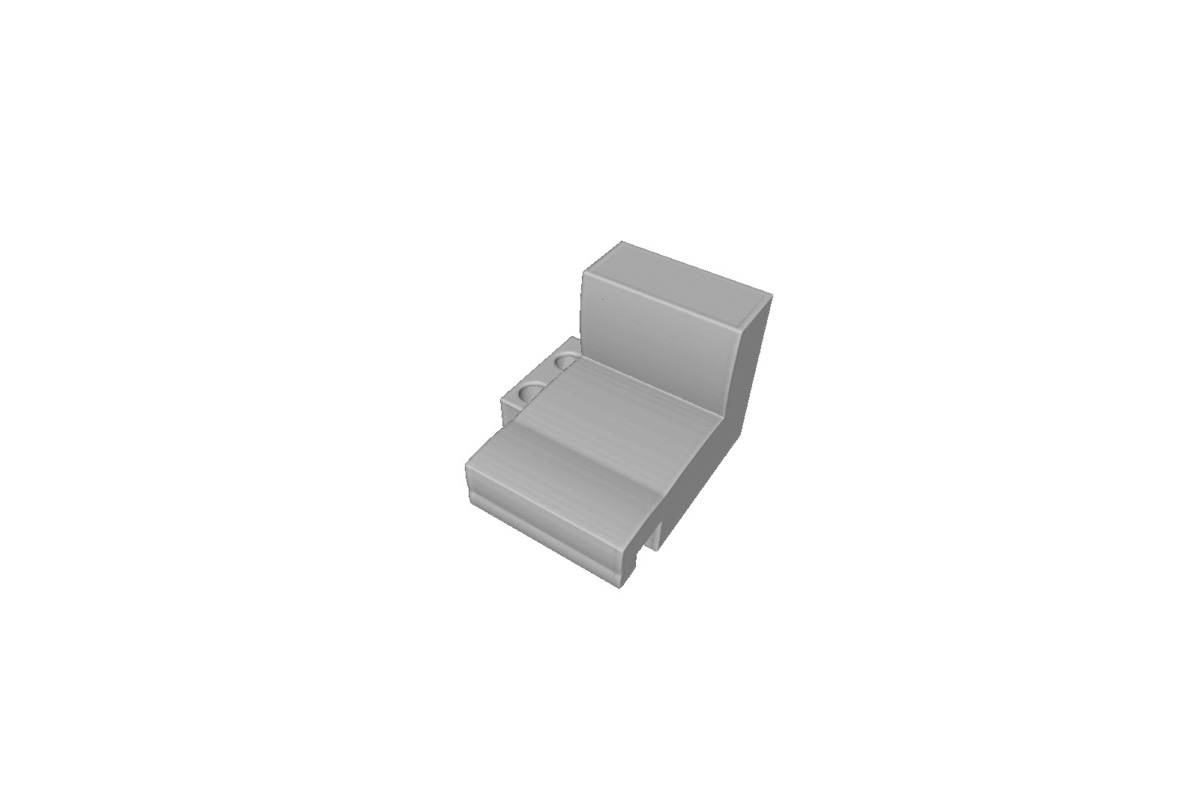}}\hspace{\spacing}
\subfloat[table\_1]{
    \includegraphics[width=\fitscale\tgtwidth, trim={262 76 319  76}, clip]{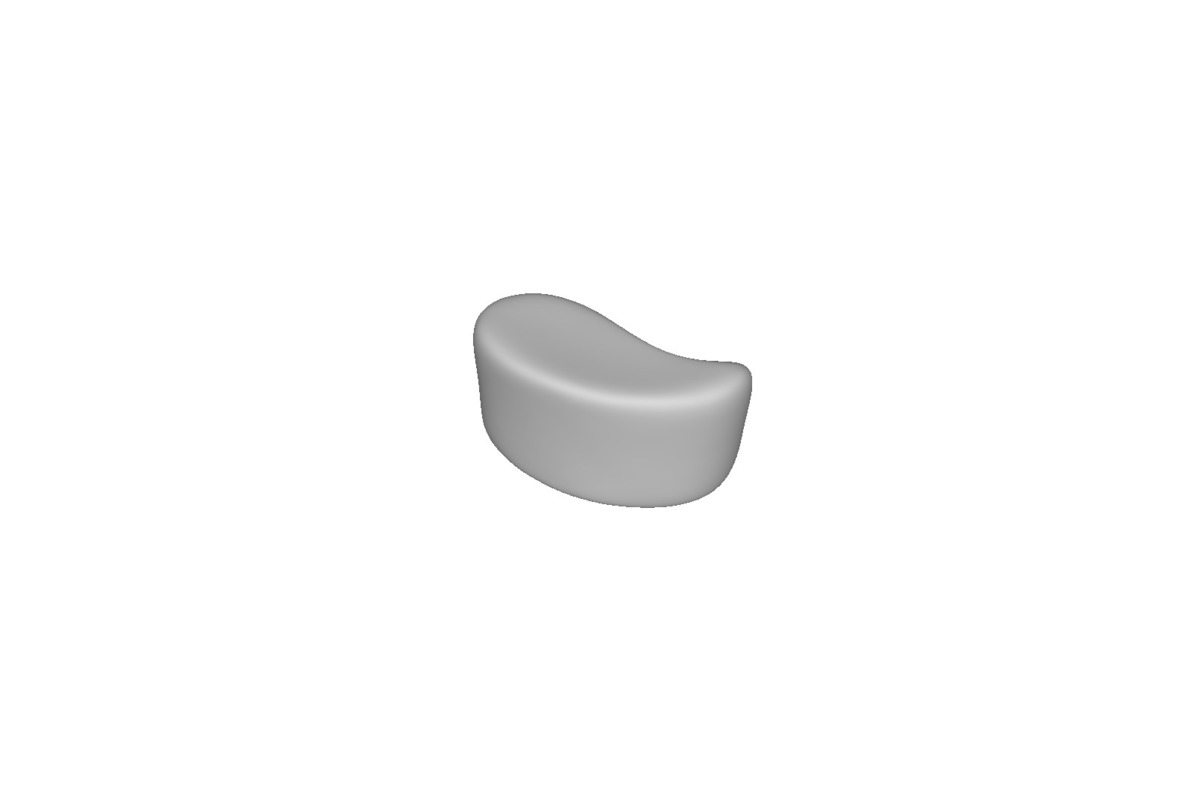}
    \includegraphics[width=\fitscale\tgtwidth, trim={262 76 319  76}, clip]{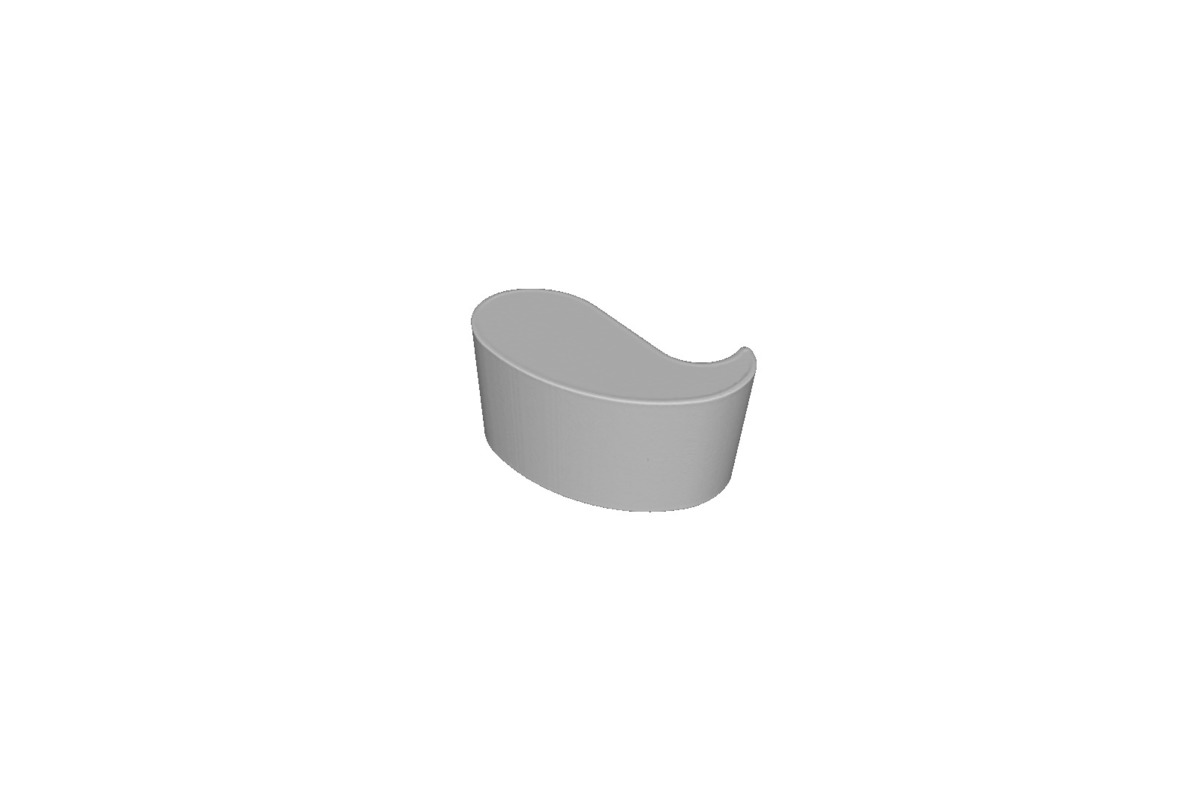}}\hspace{\spacing}
\subfloat[table\_2]{
    \includegraphics[width=\fitscale\tgtwidth, trim={262 76 319  76}, clip]{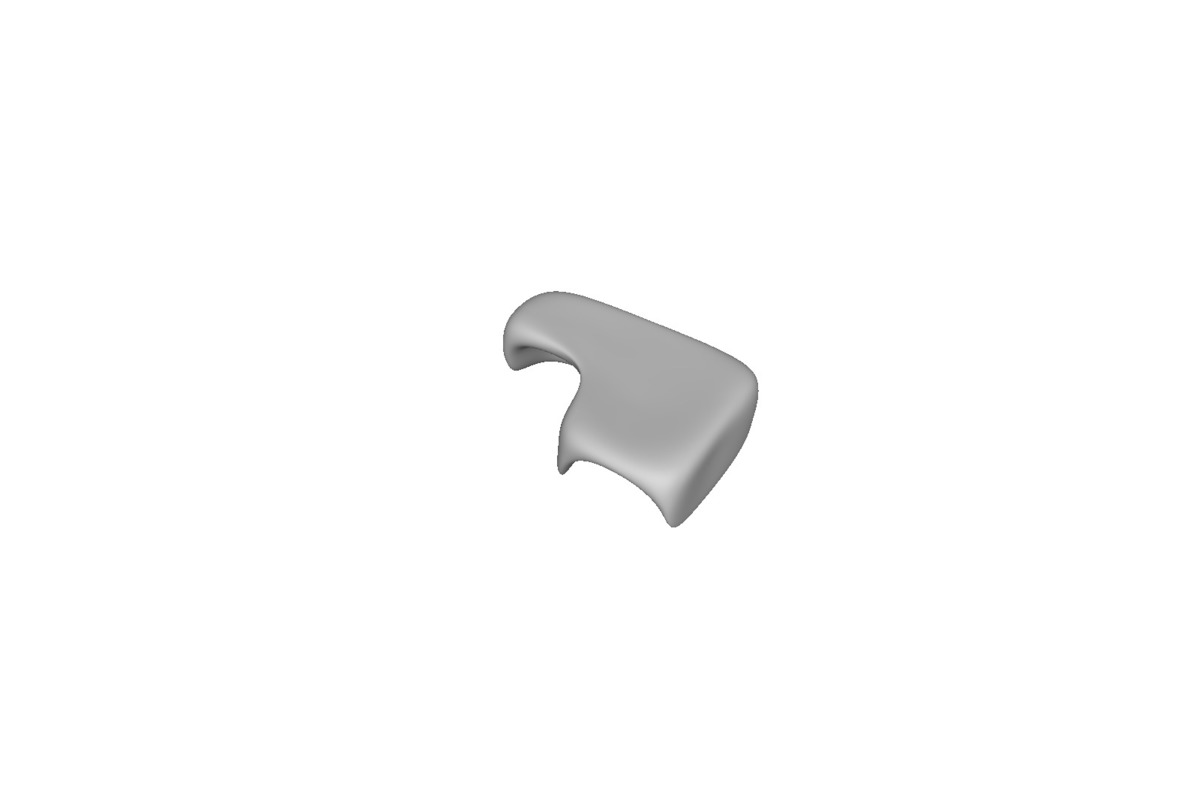}
    \includegraphics[width=\fitscale\tgtwidth, trim={262 76 319  76}, clip]{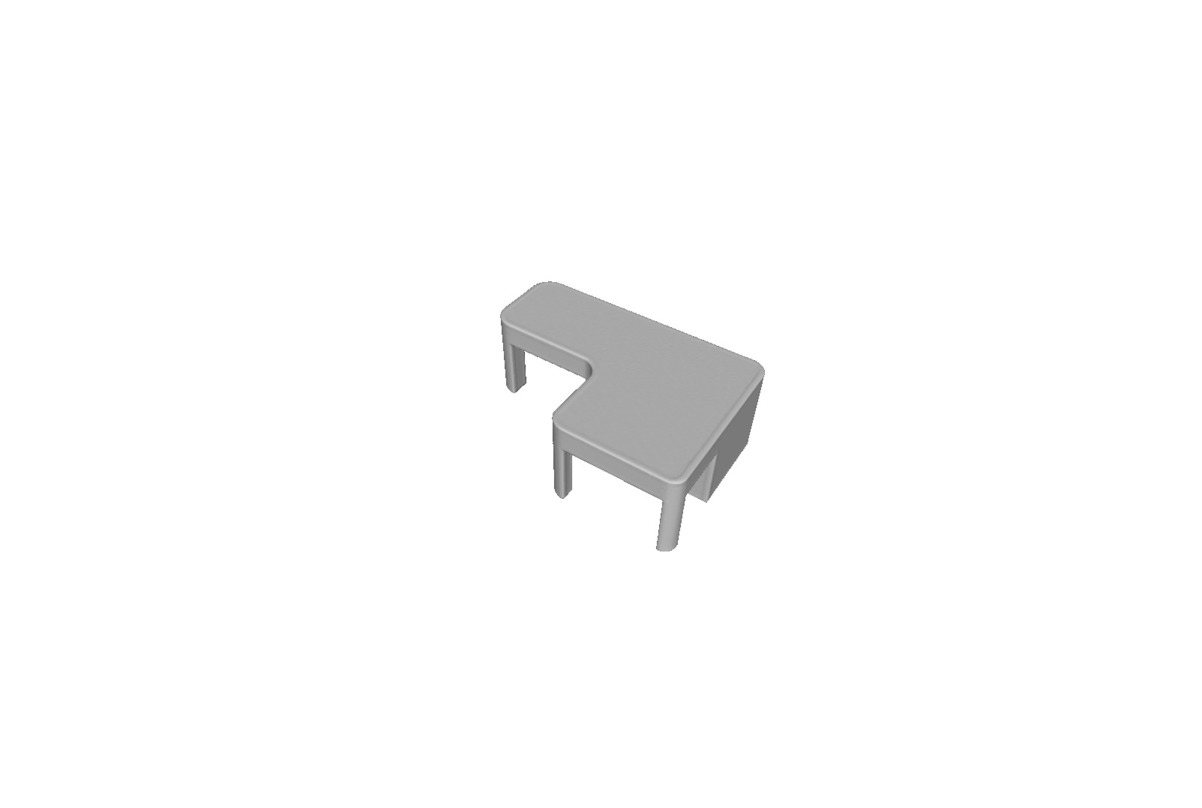}}\hspace{\spacing}
\subfloat[mobile\_1]{
    \includegraphics[width=\fitscale\tgtwidth, trim={262 76 319  76}, clip]{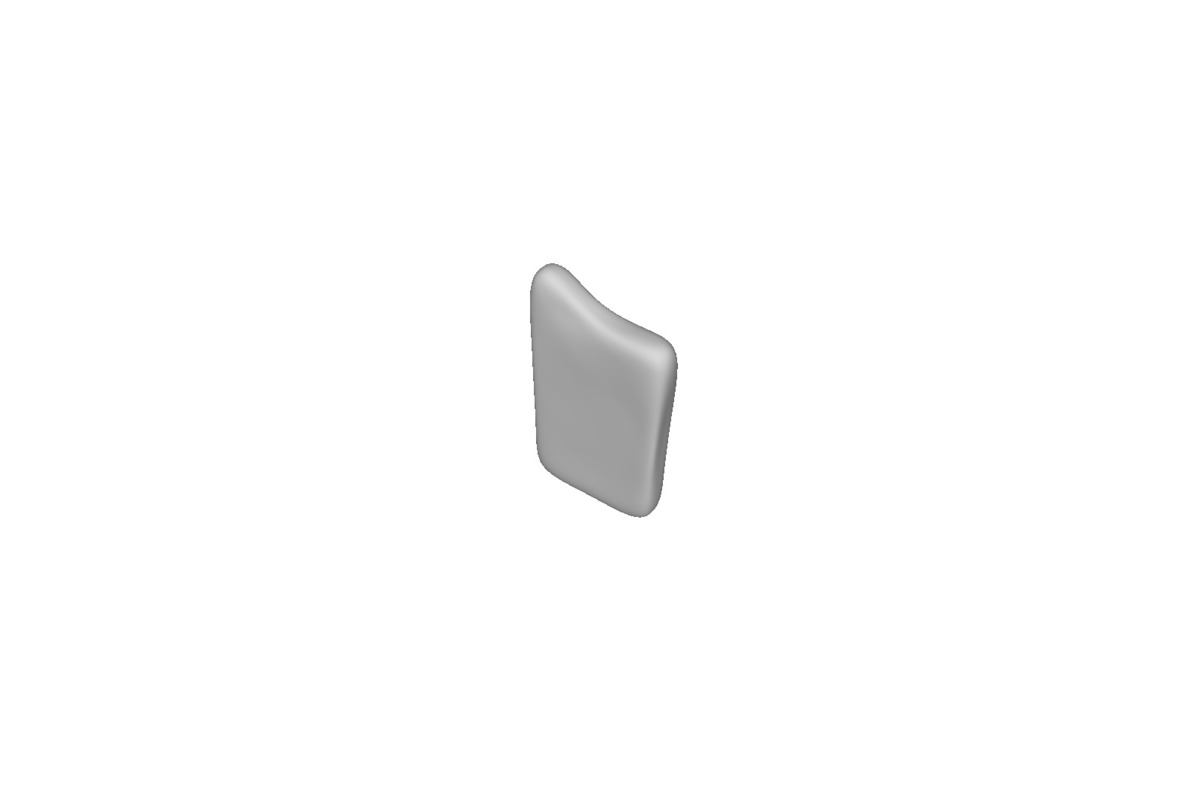}
    \includegraphics[width=\fitscale\tgtwidth, trim={262 76 319  76}, clip]{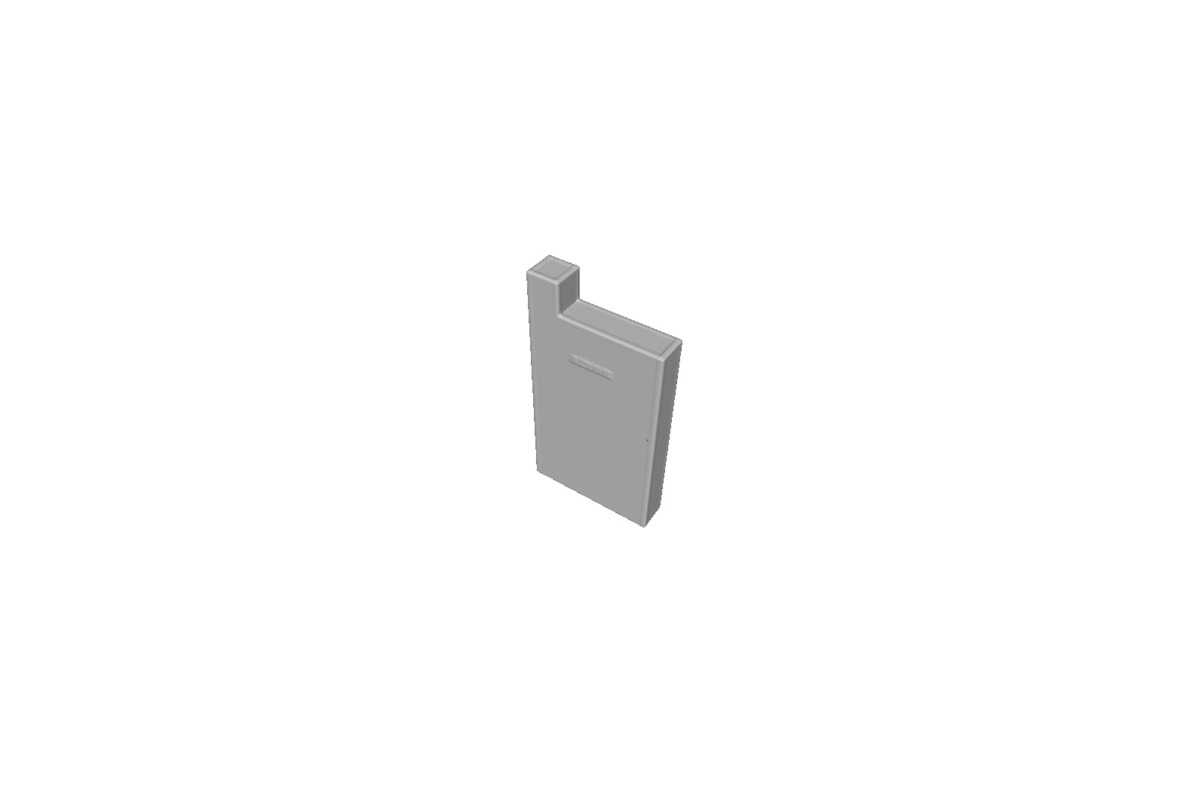}}
\\
\vspace{-4mm}
\subfloat[armadillo]{
    \includegraphics[width=\fitscale\tgtwidth, trim={262 76 319  76}, clip]{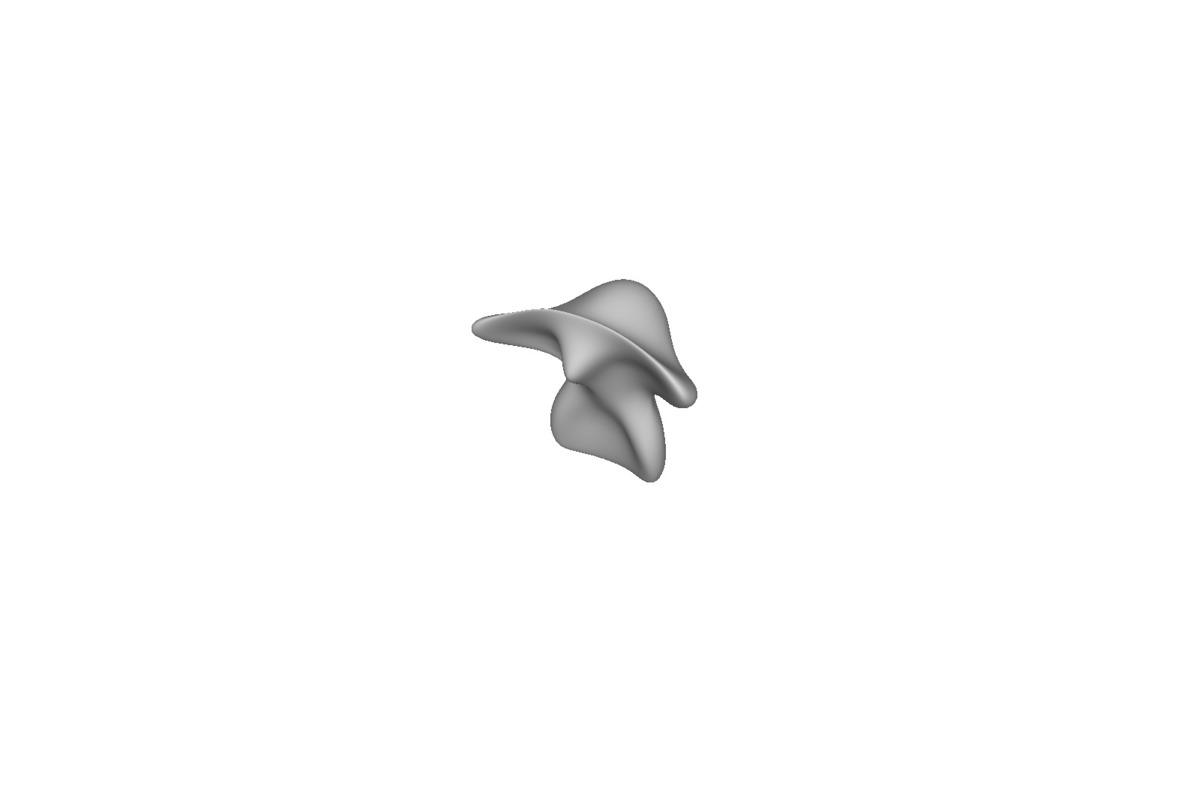}
    \includegraphics[width=\fitscale\tgtwidth, trim={262 76 319  76}, clip]{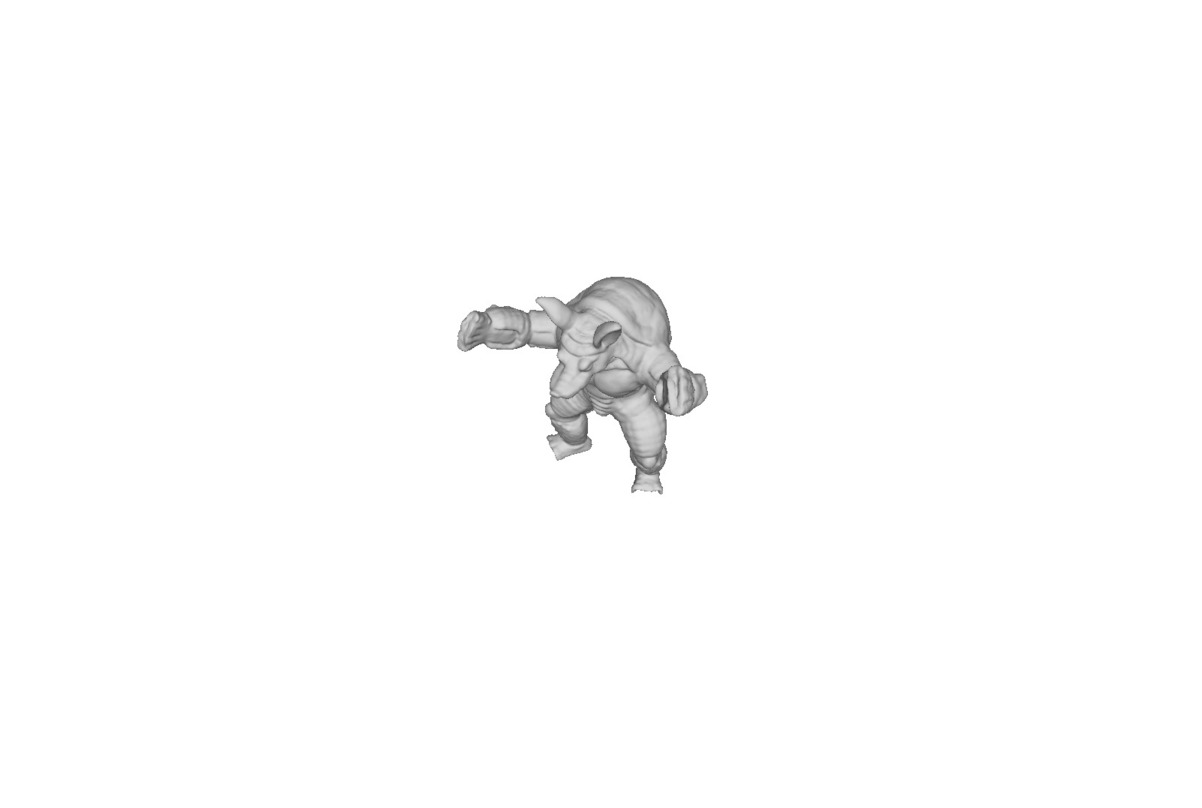}}\hspace{\spacing}
\subfloat[bunny]{
    \includegraphics[width=\fitscale\tgtwidth, trim={262 76 319  76}, clip]{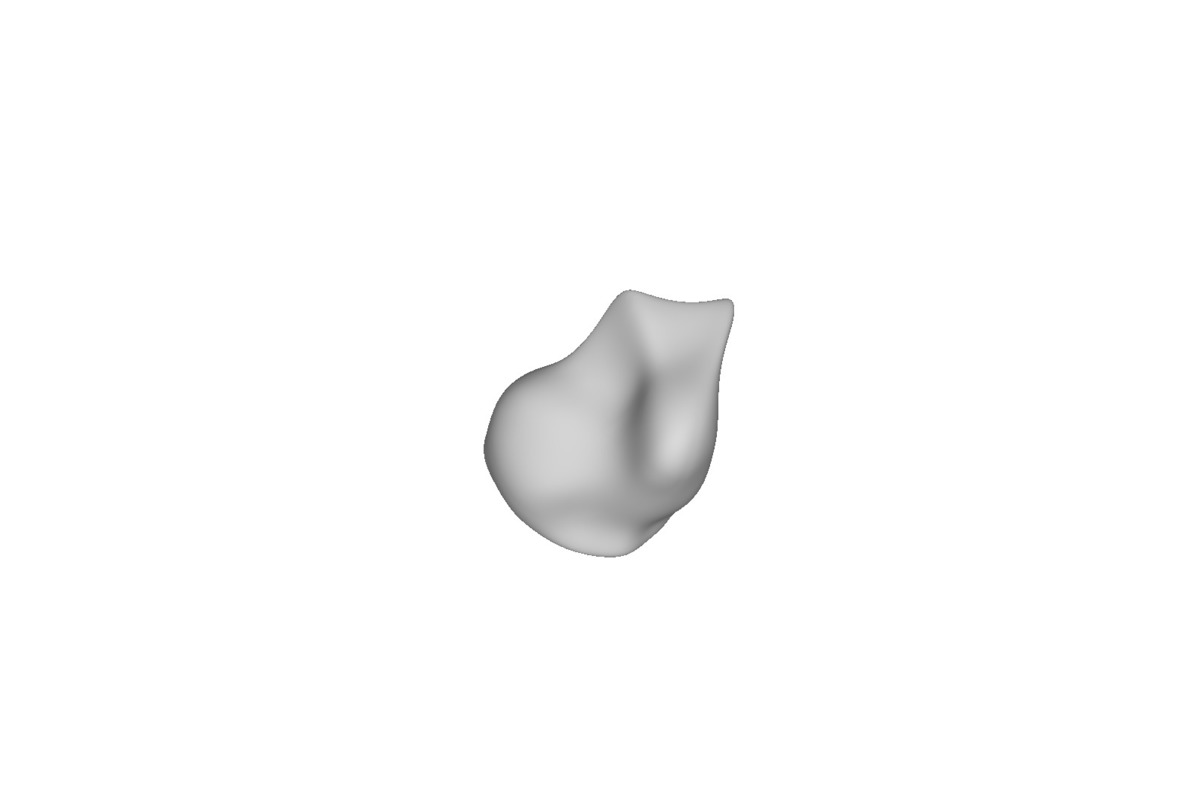}
    \includegraphics[width=\fitscale\tgtwidth, trim={262 76 319  76}, clip]{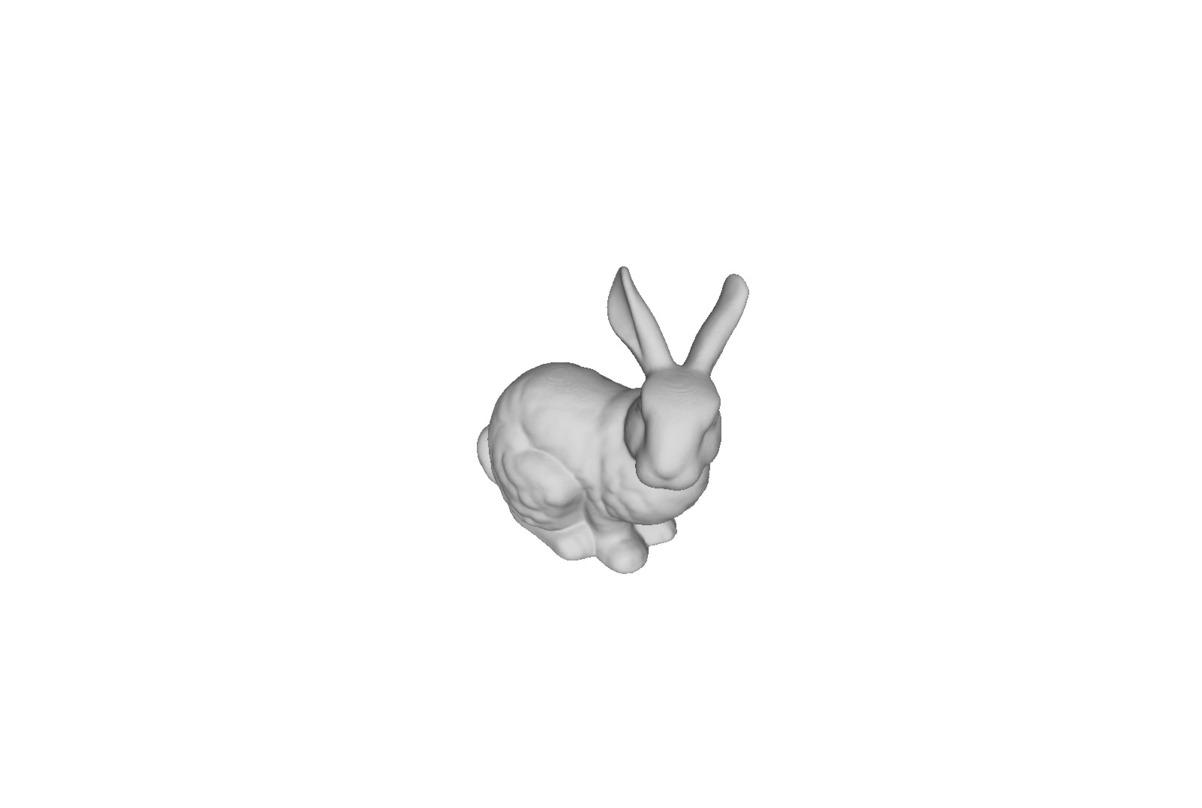}}\hspace{\spacing}
\subfloat[suzzane]{
    \includegraphics[width=\fitscale\tgtwidth, trim={262 76 319  76}, clip]{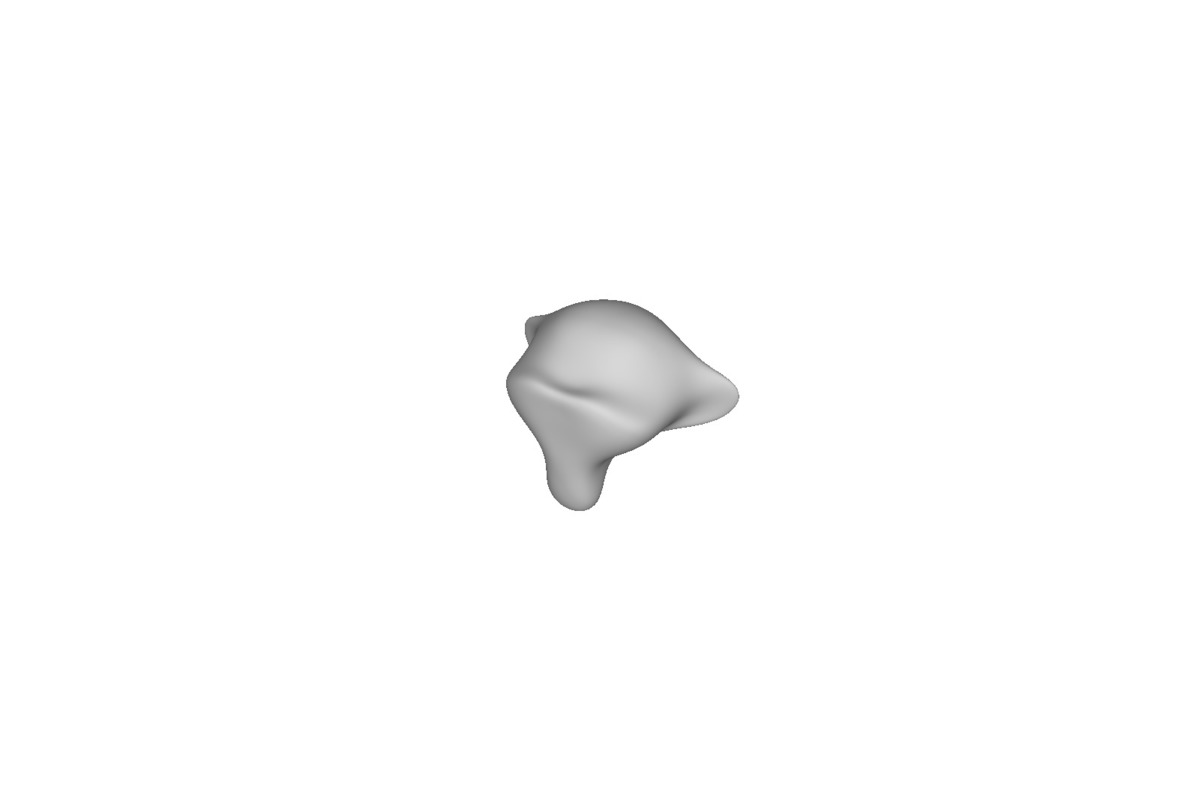}
    \includegraphics[width=\fitscale\tgtwidth, trim={262 76 319  76}, clip]{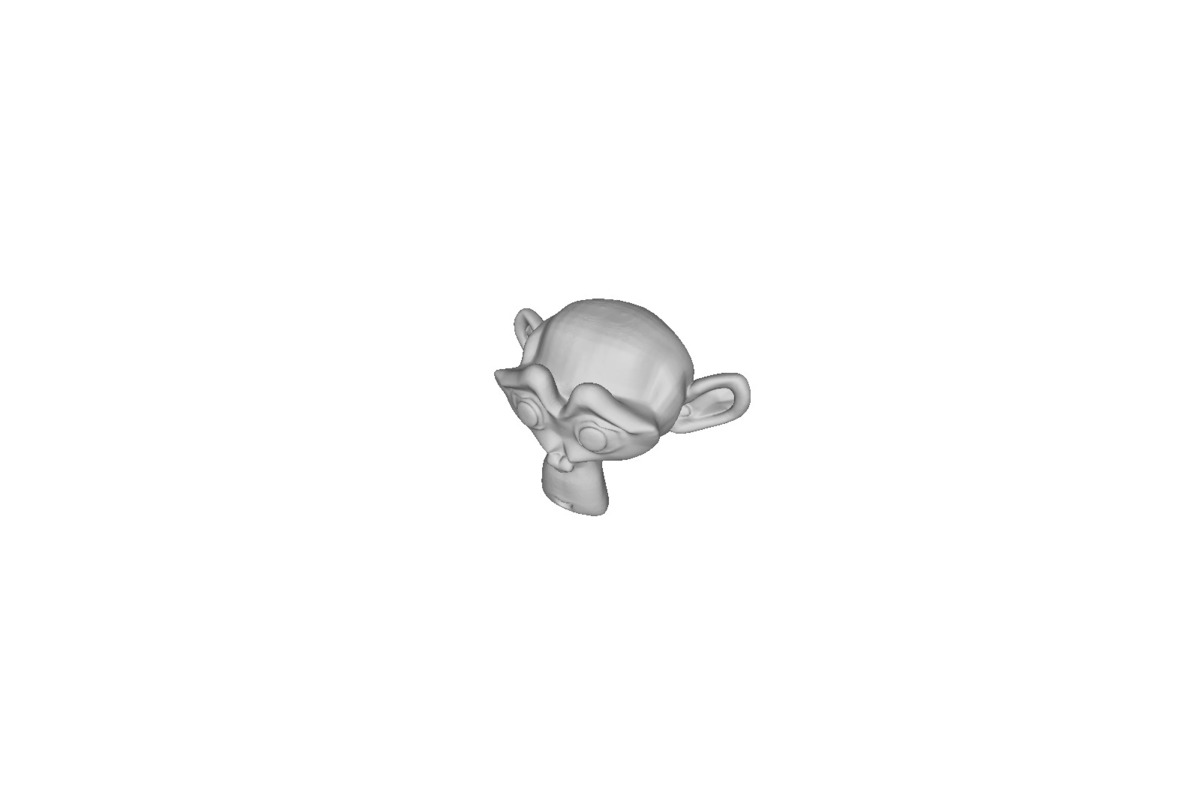}}\hspace{\spacing}
\subfloat[teapot]{
    \includegraphics[width=\fitscale\tgtwidth, trim={262 76 319  76}, clip]{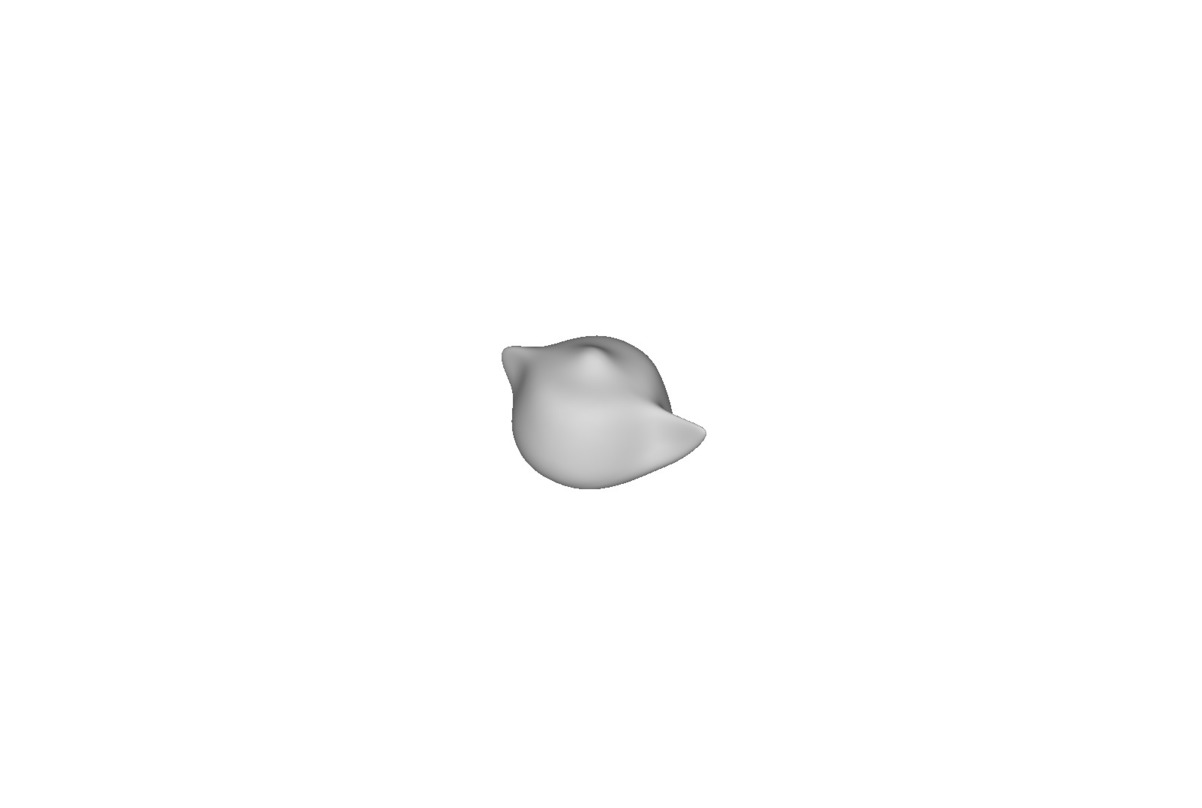}
    \includegraphics[width=\fitscale\tgtwidth, trim={262 76 319  76}, clip]{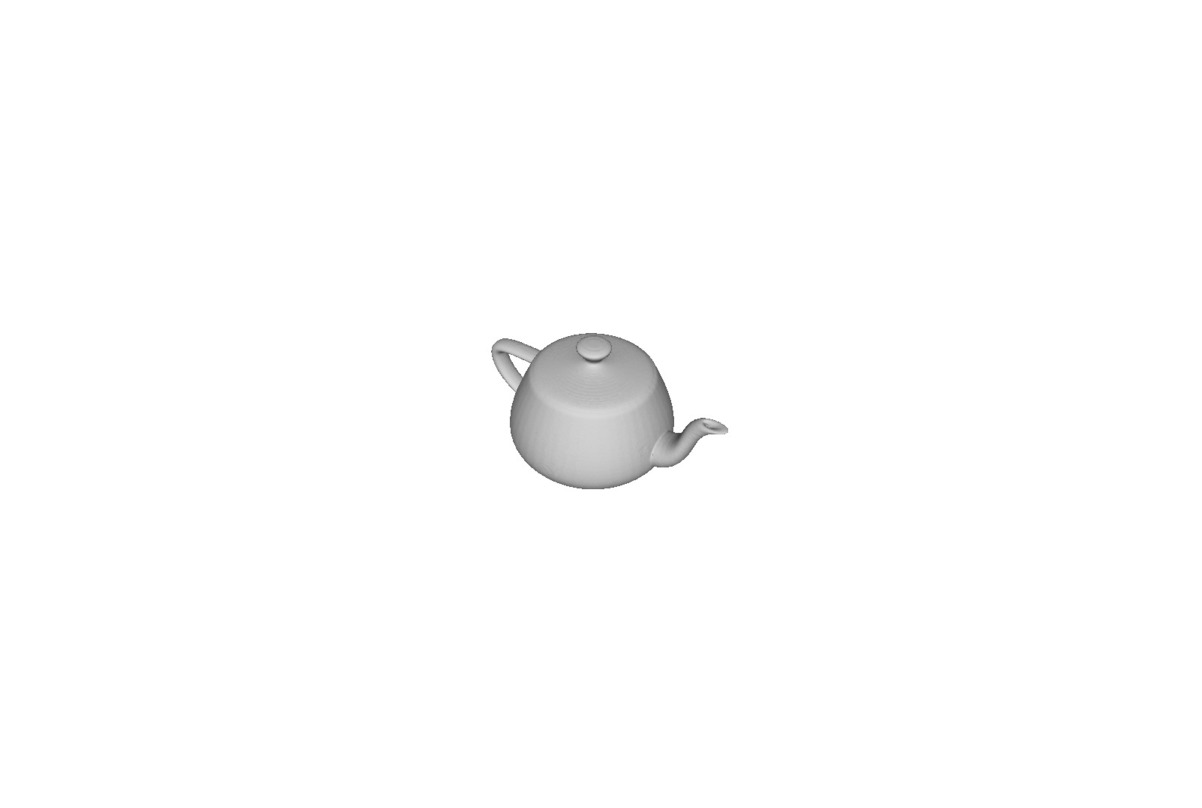}}
\\

%% file: figures_supplementary/noise_visualization/noise_viz.tex

\setlength{\numcrops}{8pt}
\newcommand{\fitscale}{.85}

\settowidth{\cropwidth}{\includegraphics{figures/results/teapot_new/snapshot00.jpg}}

\setlength{\one}{1pt}

\setlength\tgtwidth{\textwidth*\ratio{\one}{\numcrops}}

\captionsetup[subfigure]{labelformat=empty}
\centering
\vspace{-4mm}
\subfloat{\includegraphics[width=\fitscale\tgtwidth, trim={262 76 319 76}, clip]{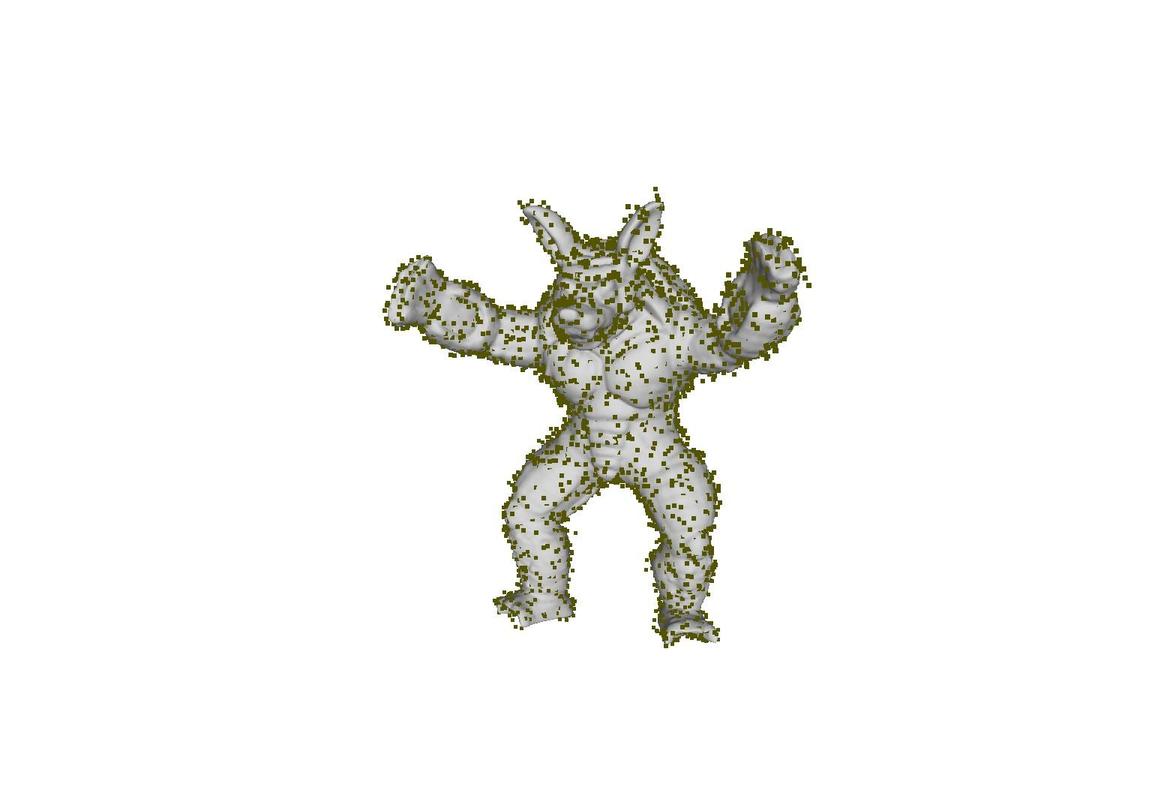}}
\subfloat{\includegraphics[width=\fitscale\tgtwidth, trim={262 76 319 76}, clip]{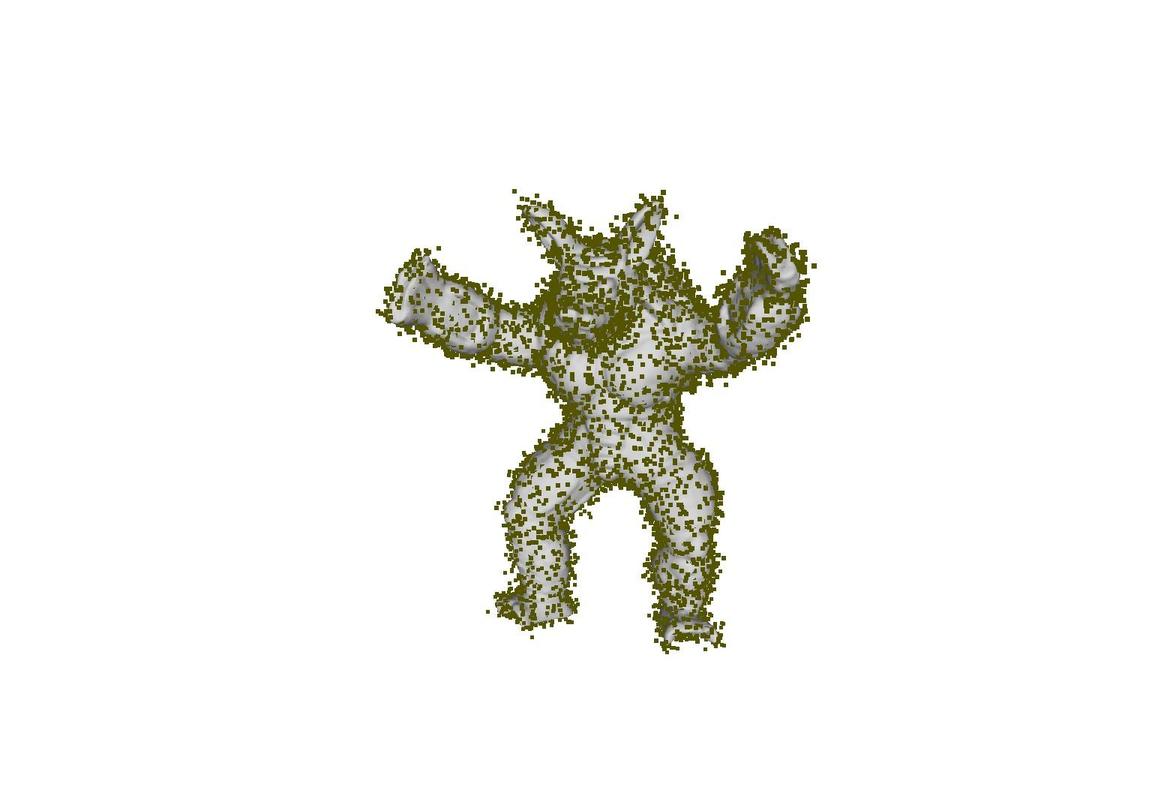}}
\subfloat{\includegraphics[width=\fitscale\tgtwidth, trim={262 76 319 76}, clip]{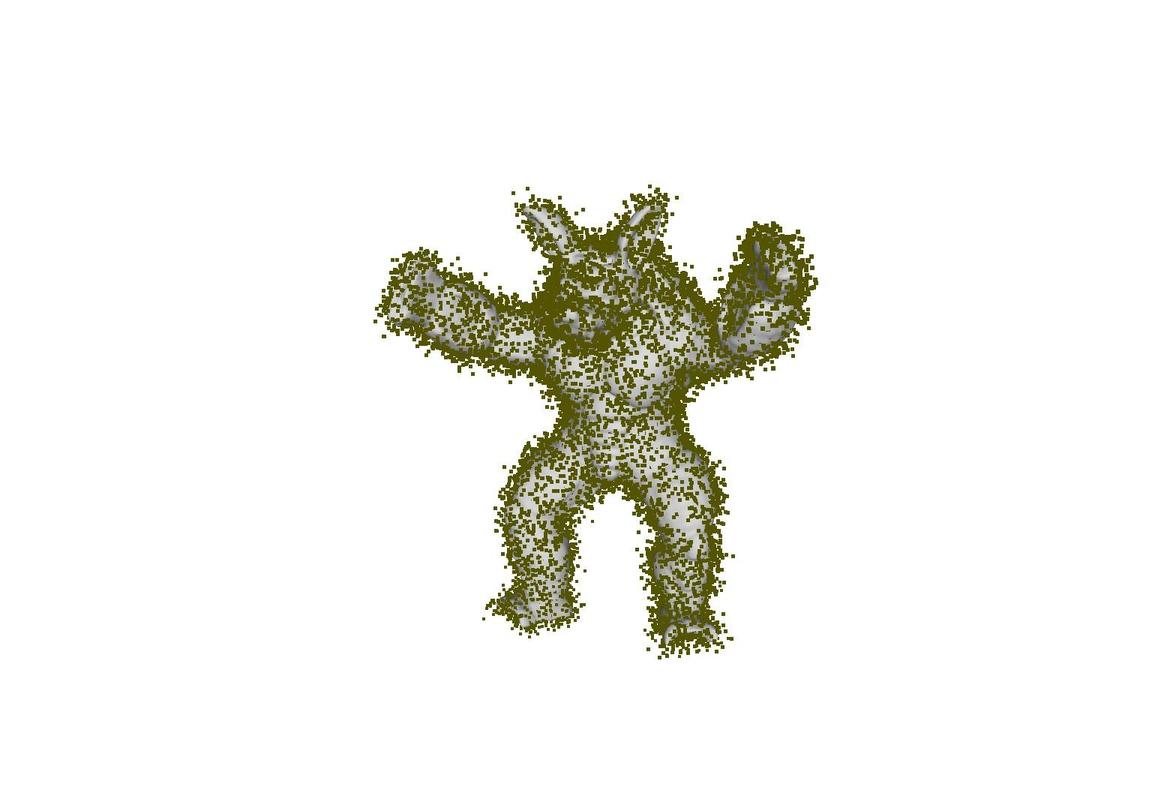}}
\subfloat{\includegraphics[width=\fitscale\tgtwidth, trim={262 76 319 76}, clip]{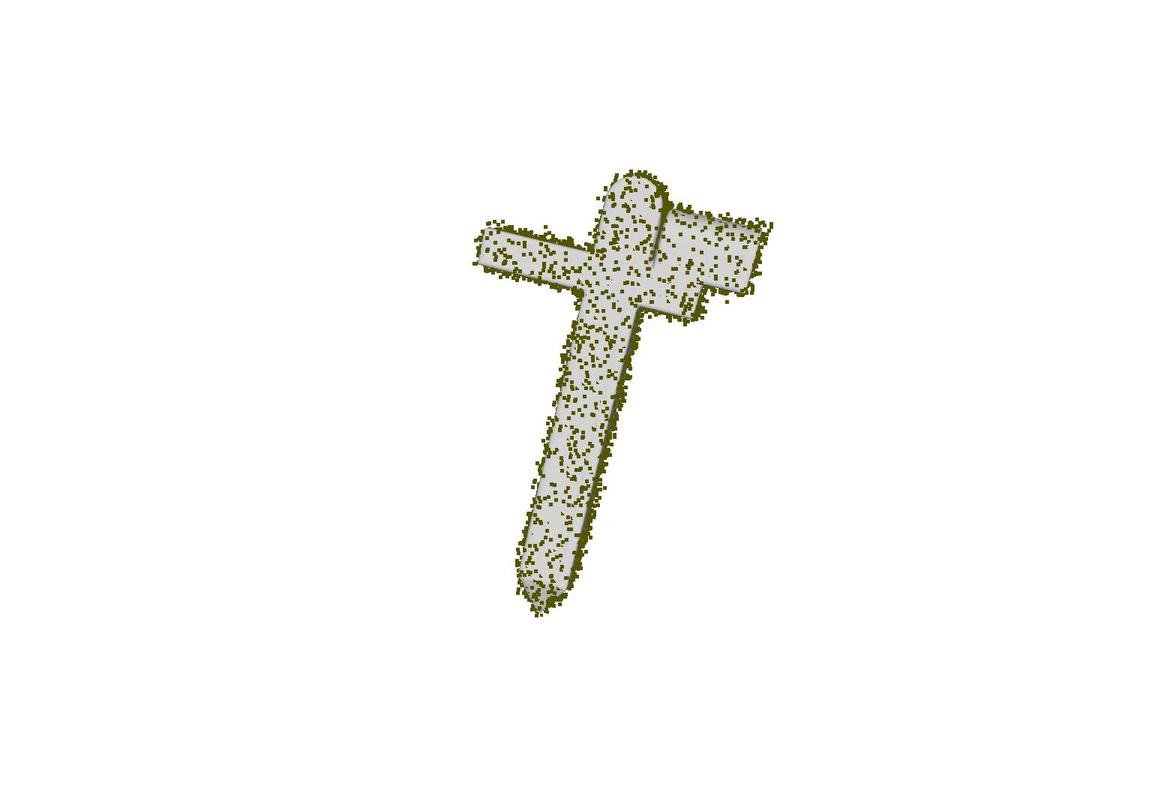}}
\subfloat{\includegraphics[width=\fitscale\tgtwidth, trim={262 76 319 76}, clip]{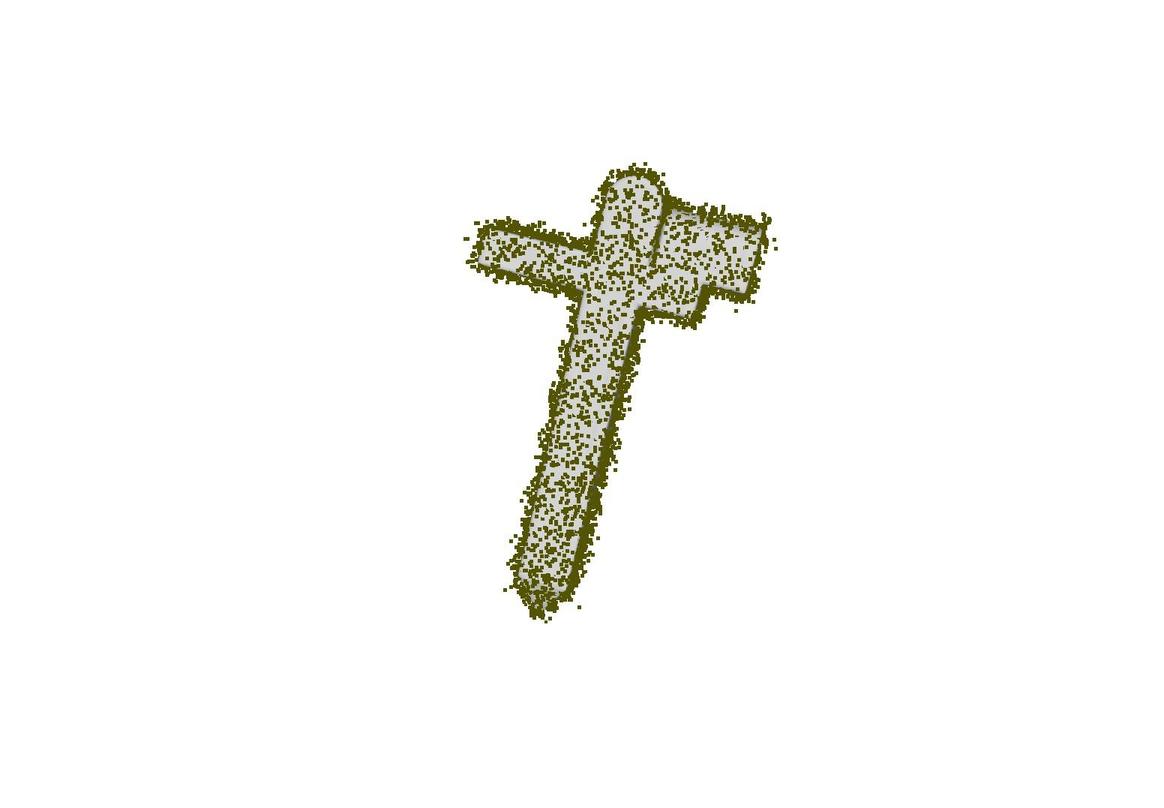}}
\subfloat{\includegraphics[width=\fitscale\tgtwidth, trim={262 76 319 76}, clip]{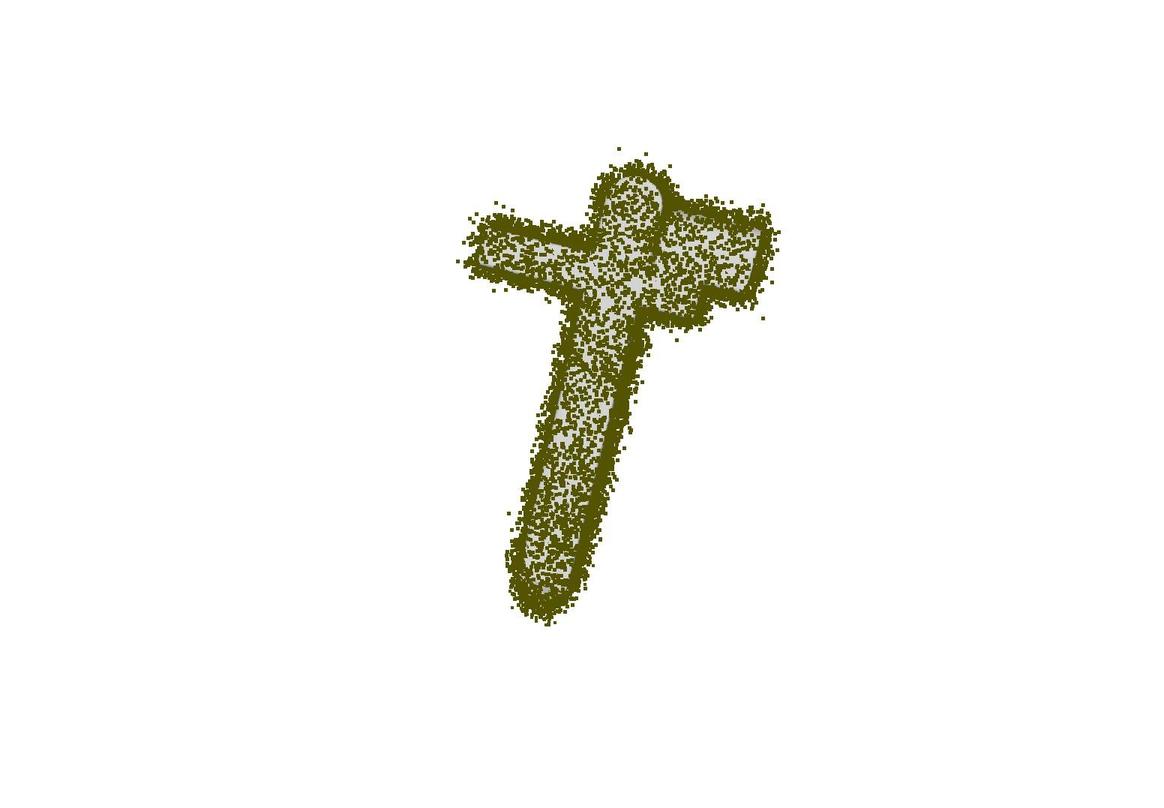}}
\subfloat{\includegraphics[width=\fitscale\tgtwidth, trim={262 76 319 76}, clip]{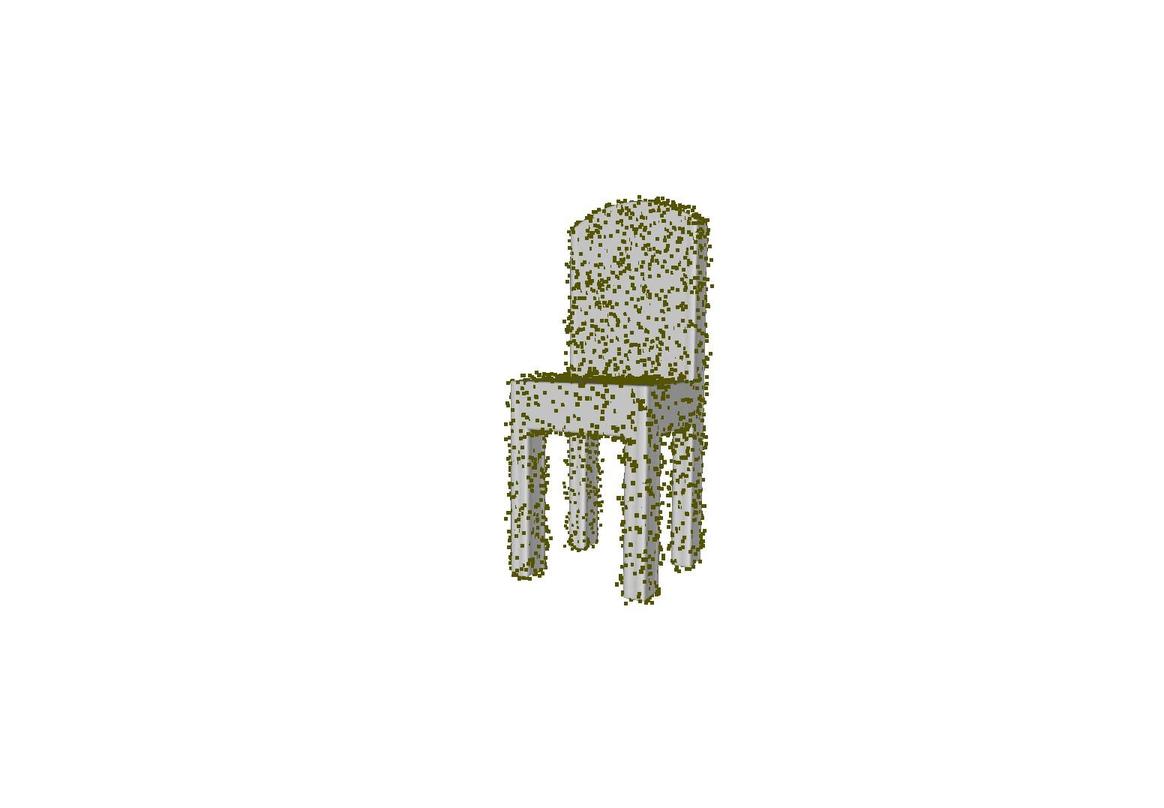}}
\subfloat{\includegraphics[width=\fitscale\tgtwidth, trim={262 76 319 76}, clip]{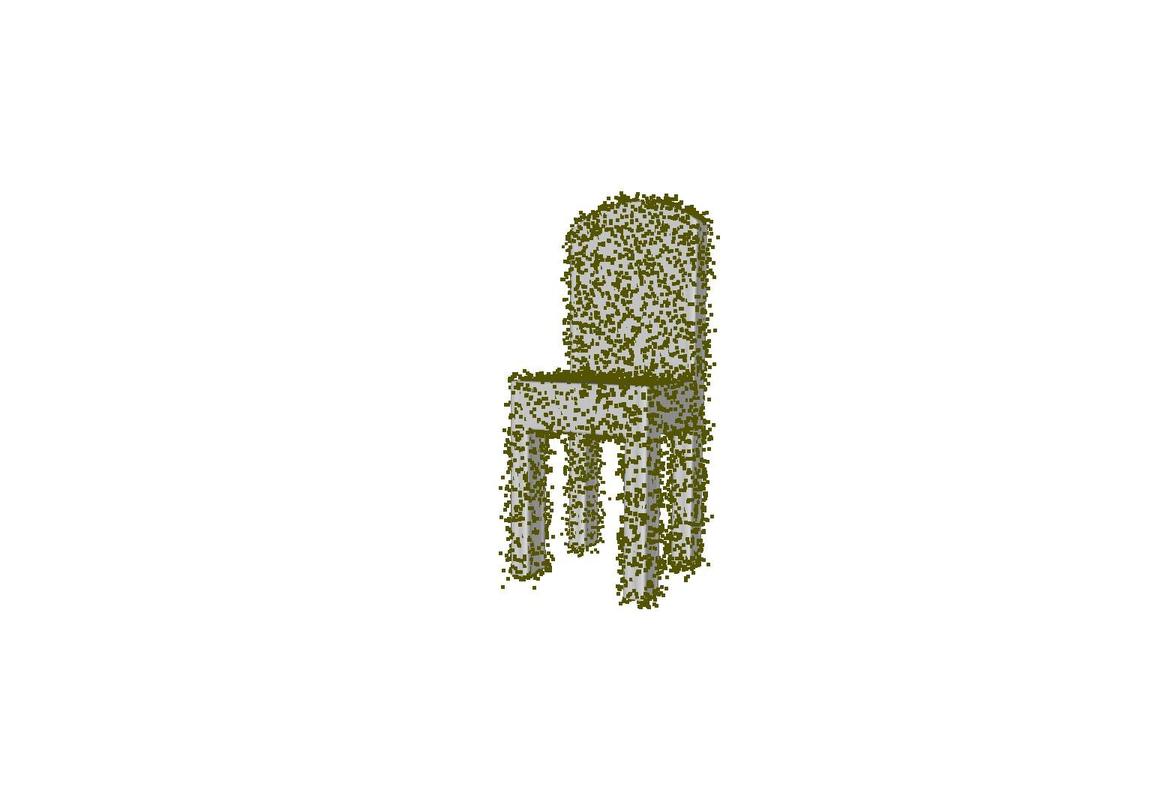}}
\subfloat{\includegraphics[width=\fitscale\tgtwidth, trim={262 76 319 76}, clip]{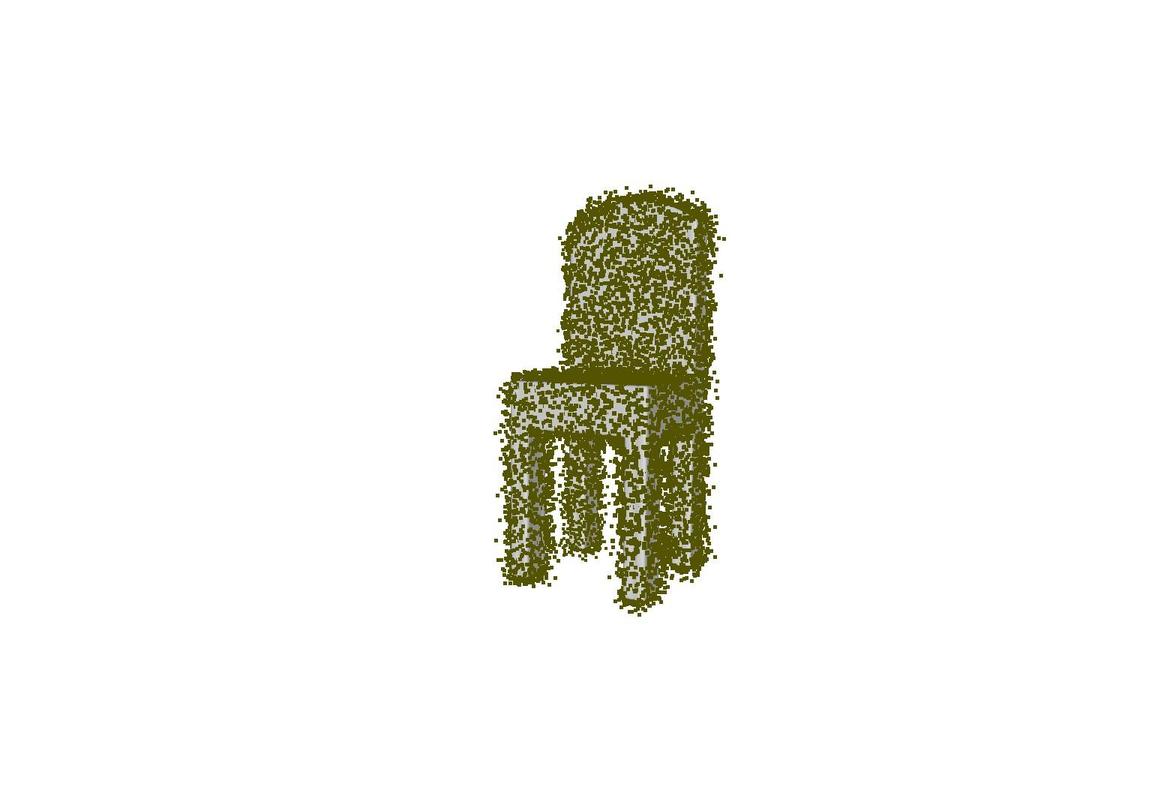}}
\\
\vspace{-4mm}
\subfloat[S1]{\includegraphics[width=\fitscale\tgtwidth, trim={262 76 319 76}, clip]{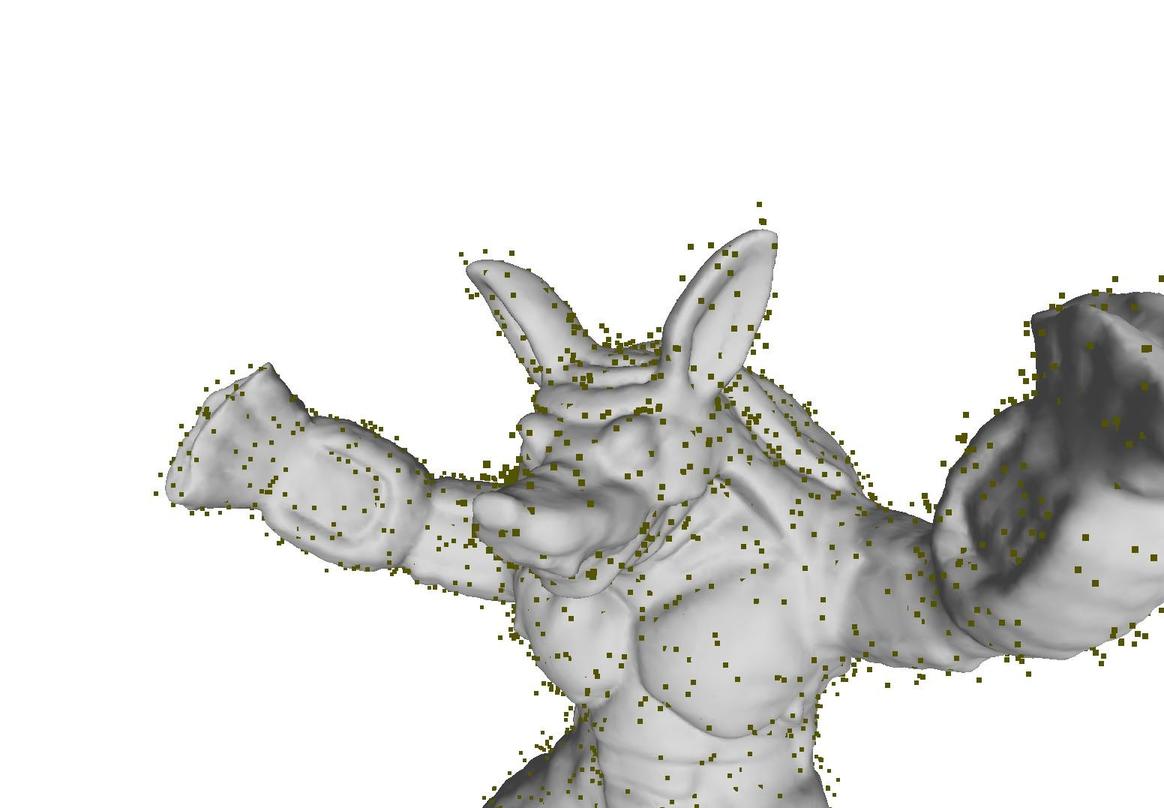}}
\subfloat[S2]{\includegraphics[width=\fitscale\tgtwidth, trim={262 76 319 76}, clip]{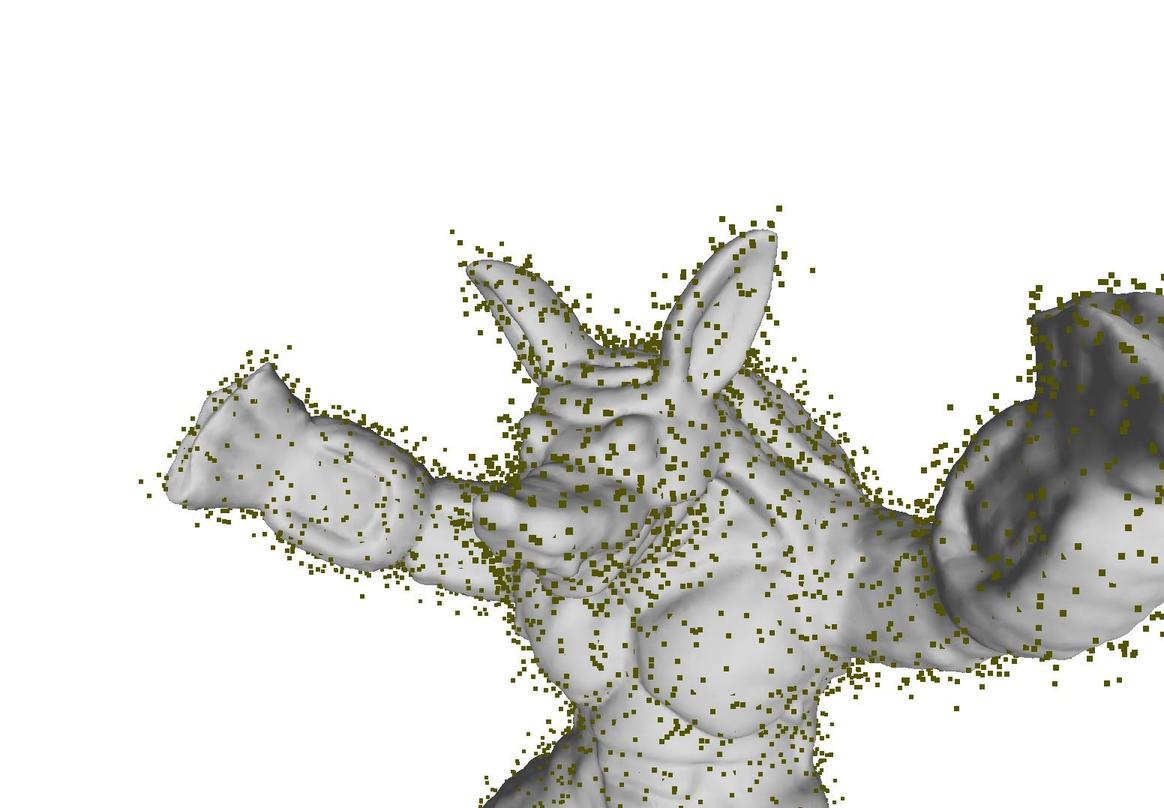}}
\subfloat[S3]{\includegraphics[width=\fitscale\tgtwidth, trim={262 76 319 76}, clip]{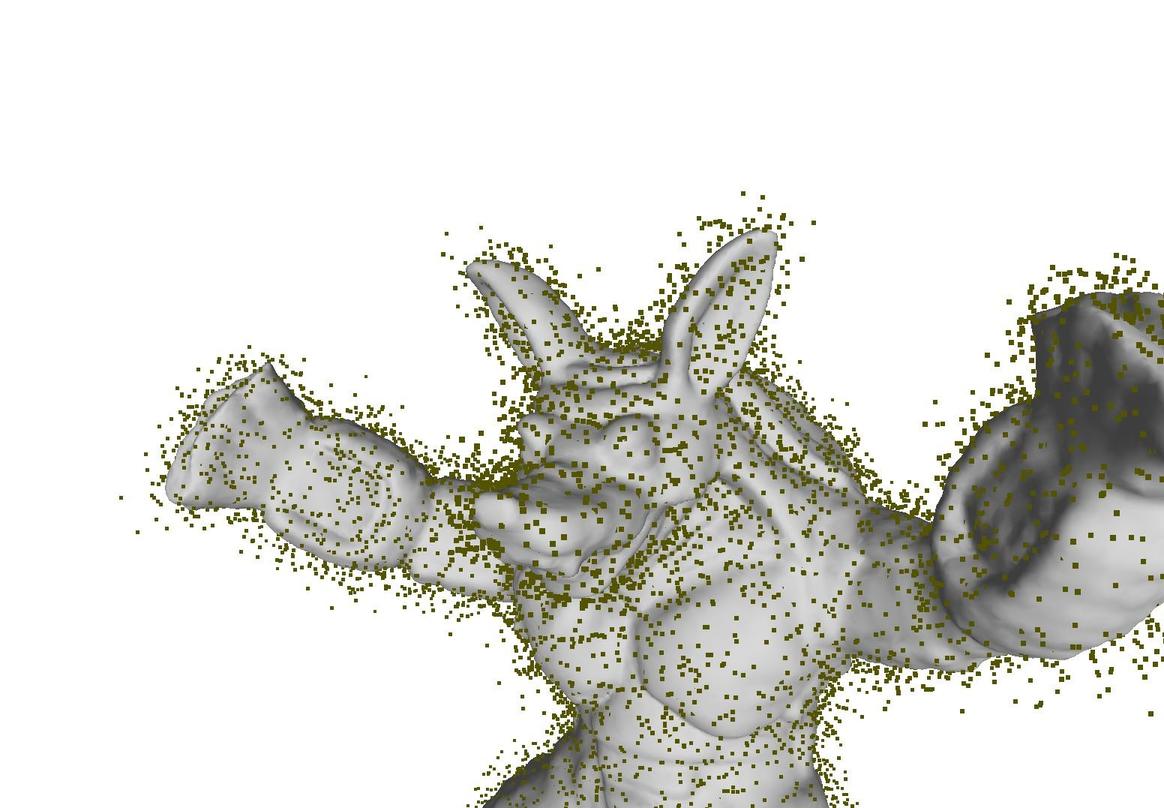}}
\subfloat[S1]{\includegraphics[width=\fitscale\tgtwidth, trim={262 76 319 76}, clip]{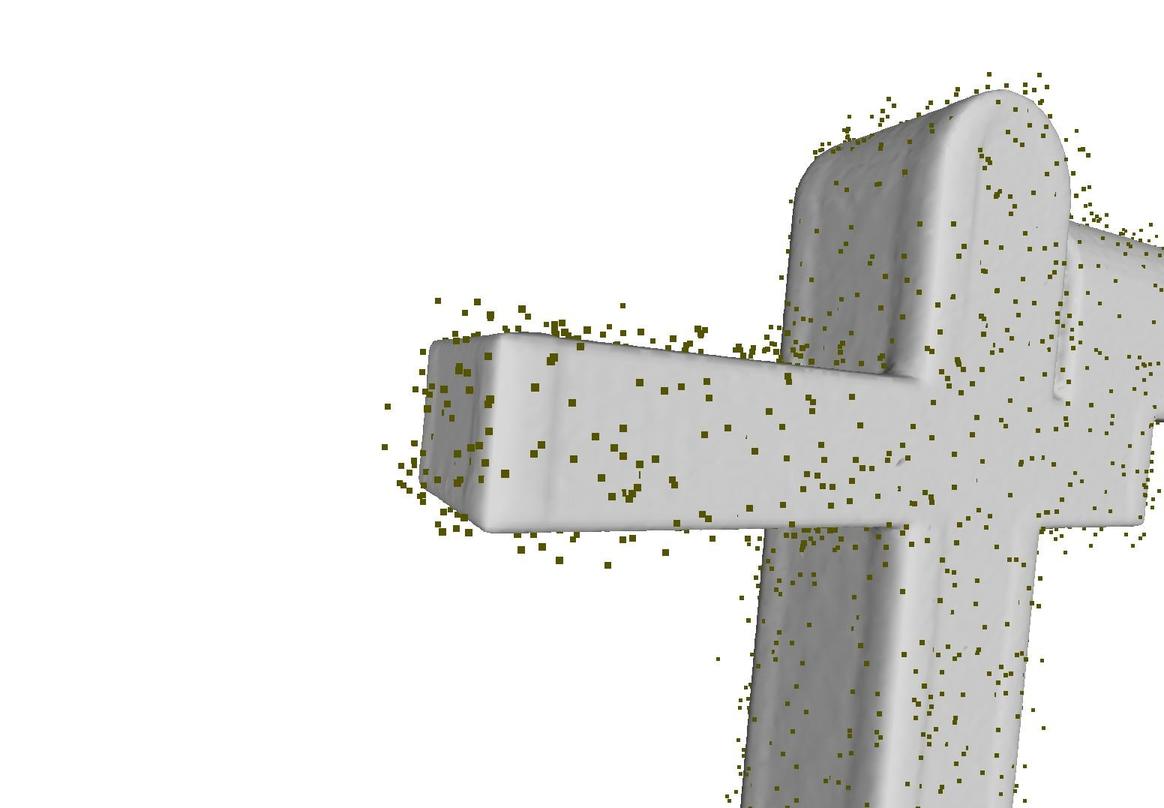}}
\subfloat[S2]{\includegraphics[width=\fitscale\tgtwidth, trim={262 76 319 76}, clip]{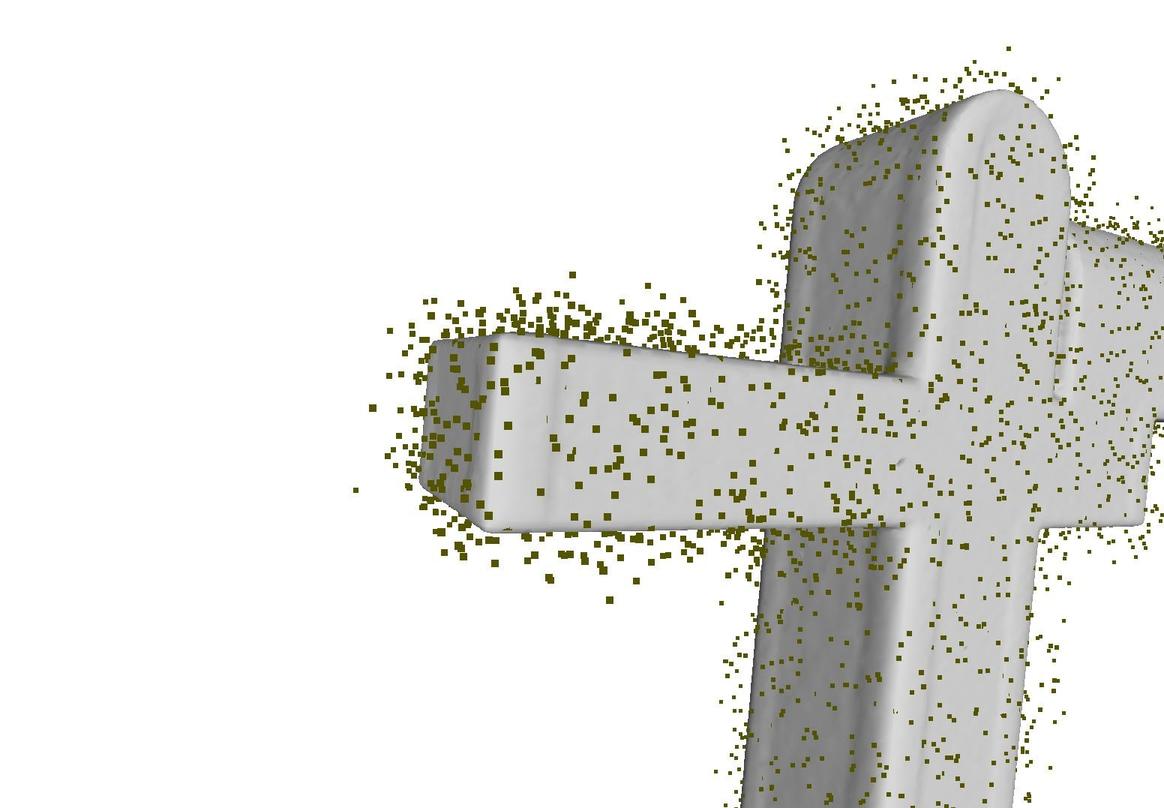}}
\subfloat[S3]{\includegraphics[width=\fitscale\tgtwidth, trim={262 76 319 76}, clip]{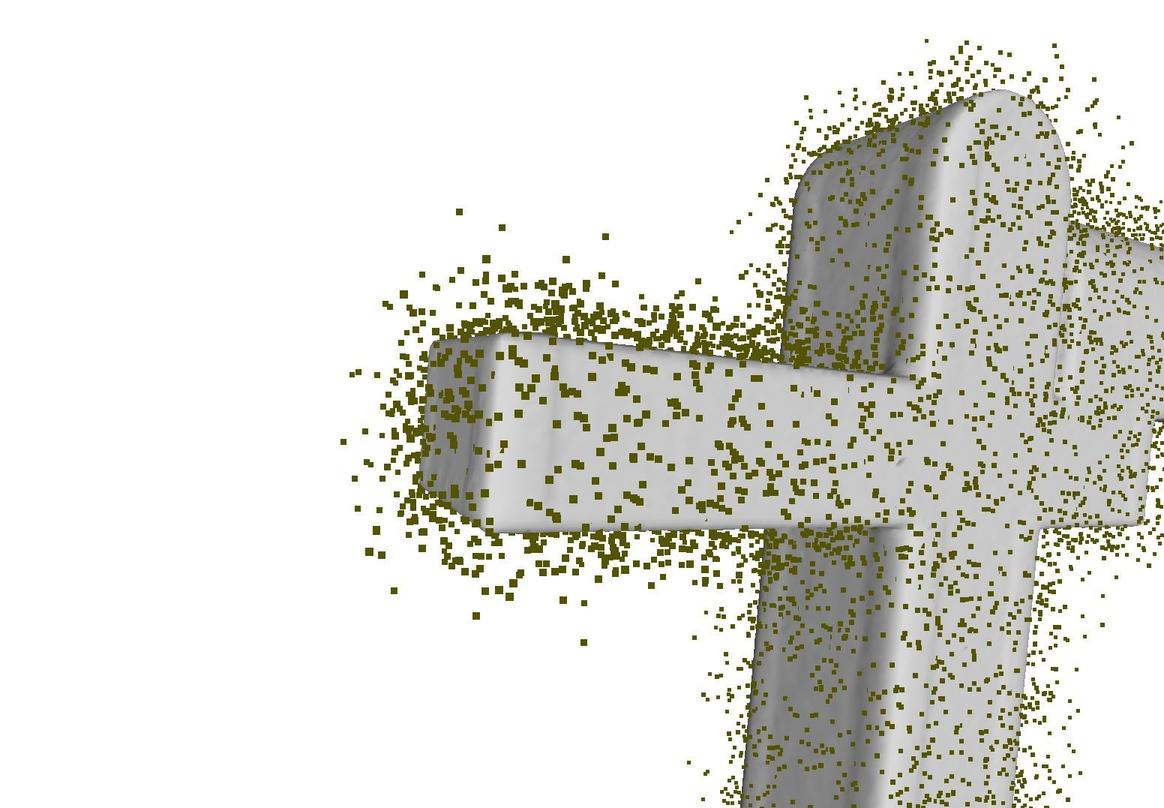}}
\subfloat[S1]{\includegraphics[width=\fitscale\tgtwidth, trim={262 76 319 76}, clip]{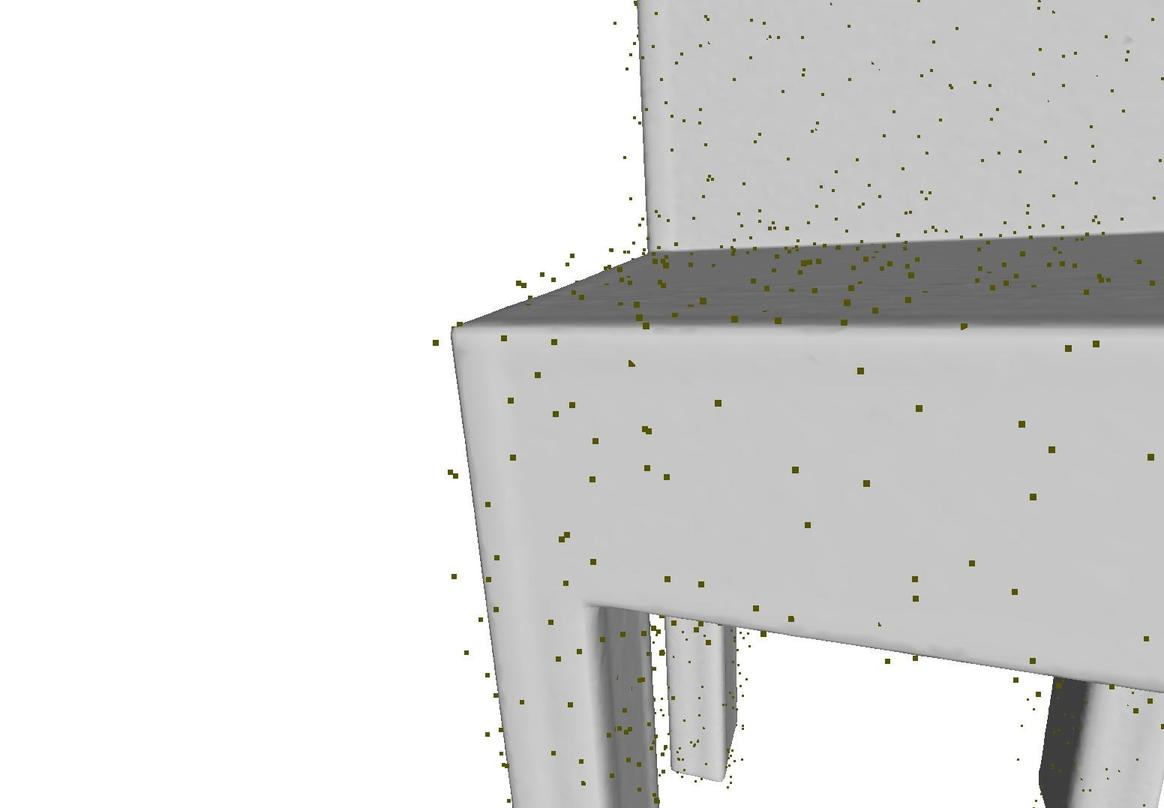}}
\subfloat[S2]{\includegraphics[width=\fitscale\tgtwidth, trim={262 76 319 76}, clip]{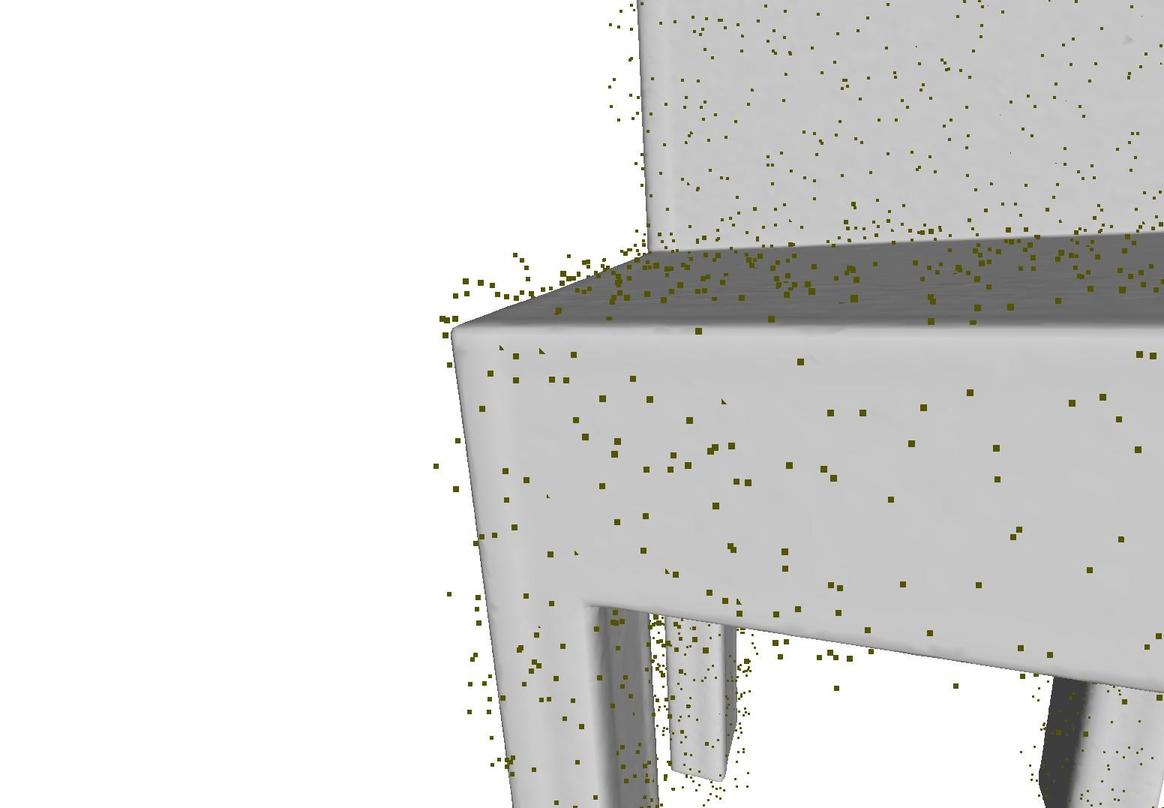}}
\subfloat[S3]{\includegraphics[width=\fitscale\tgtwidth, trim={262 76 319 76}, clip]{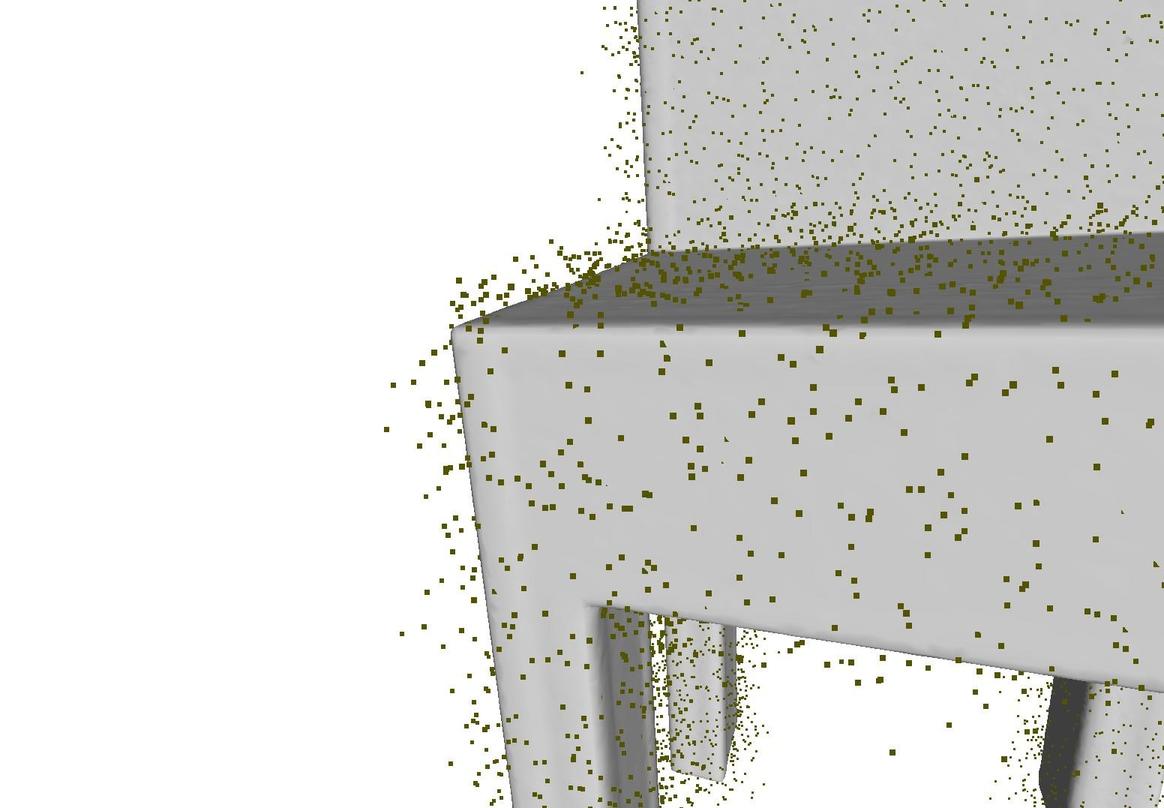}}
\\

%% file: meshlets_cvpr20_supplementary_additional_results.tex

Figure~\ref{fig:results} shows additional qualitative comparisons between different approaches. 
\begin{figure*}
    \input{figures_supplementary/results_cvpr2020/result_page}
    \caption{Qualitative comparison of several reconstruction methods and our approach. 
    On the left is the input, the ground-truth (GT) mesh overlaid with the sparse, noisy point cloud (PC). 
    We show results with normals estimated with Meshlab~\cite{meshlab} and PCPNet~\cite{pcpnet}. We reconstruct the resulting point clouds with both RILMS~\cite{oztireli2009feature} and Screened Poisson~\cite{screenedpoisson}. 
    Laplacian regularizer~\cite{laplacian} is shown for two levels of smoothing. We also show three recent deep learning approaches: Deep Geometric Prior~\cite{Williams2019}, AtlasNet~\cite{Groueix_2018_CVPR} and OccNet~\cite{occNet}.}\label{fig:results}
\end{figure*}

We provide the numbers for each object across the different noise settings in Table~\ref{tab:hausdorff} (Symmetric Hausdorff) and Table~\ref{tab:chamfer} (Chamfer-$\ell_1$).

\begin{table*}[]
    \centering
    \scriptsize
    \begin{tabular}{|l|l|l|l|l|l|l|l|l|l|l|}
    \hline
    \textbf{obj\_name} & \textbf{Ours}  & \textbf{DGP} & \textbf{Lap-low} & \textbf{Lap-high} & \textbf{\cite{meshlab}+\cite{screenedpoisson}} & \cite{meshlab}+\cite{oztireli2009feature} & \cite{pcpnet}+\cite{screenedpoisson} & \cite{pcpnet}+\cite{oztireli2009feature} & \textbf{OccNet} & \textbf{AtlasNet} \\ \hline
    clock\_1\_S3       & \textbf{3.653} & 17.457                                                                            & 11.052          & 11.005           & 122.523          & 15.532             & 11.359             & 15.511               & 29.283          & 24.631            \\ \hline
    clock\_1\_S2       & \textbf{3.536} & 13.955                                                                            & 10.953          & 10.891           & 6.235            & 13.111             & 11.333             & 11.858               & 25.27           & 21.75             \\ \hline
    clock\_1\_S1       & \textbf{2.853} & 13.349                                                                            & 11.419          & 11.356           & 3.971            & 10.19              & 11.778             & 15.908               & 25.274          & 21.229            \\ \hline
    chair\_1\_S3       & \textbf{5.434} & 16.978                                                                            & 32.646          & 33.899           & 115.056          & 17.746             & 14.692             & 16.042               & 22.796          & 22.476            \\ \hline
    chair\_1\_S2       & \textbf{3.154} & 12.676                                                                            & 30.231          & 32.58            & 112.8            & 13.675             & 14.424             & 13.751               & 12.777          & 19.855            \\ \hline
    chair\_1\_S1       & \textbf{4.204} & 10.124                                                                            & 29.066          & 32.861           & 4.012            & 12.236             & 14.455             & 12.642               & 12.651          & 17.271            \\ \hline
    sofa\_4\_S3        & \textbf{5.327} & 17.874                                                                            & 5.597           & 6.561            & 39.052           & 18.864             & 24.706             & 18.801               & 65.388          & 35.434            \\ \hline
    sofa\_4\_S2        & 6.13           & 15.129                                                                            & \textbf{4.933}  & 6.111            & 16.215           & 14.744             & 26.939             & 16.667               & 33.341          & 33.299            \\ \hline
    sofa\_4\_S1        & 12.935         & 13.947                                                                            & \textbf{9.957}  & 11.439           & 11.61            & 15.108             & 31.597             & 19.499               & 27.108          & 32.914            \\ \hline
    sofa\_2\_S3        & \textbf{5.552} & 18.691                                                                            & 8.772           & 12.991           & 166.4            & 17.276             & 20.451             & 15.399               & 26.107          & 54.051            \\ \hline
    sofa\_2\_S2        & 6.69           & 14.925                                                                            & \textbf{5.316}  & 7.299            & 31.524           & 14.784             & 20.95              & 14.418               & 22.708          & 49.475            \\ \hline
    sofa\_2\_S1        & \textbf{3.857} & 10.758                                                                            & 5.074           & 6.611            & 13.18            & 13.112             & 24.549             & 14.801               & 16.465          & 40.677            \\ \hline
    sofa\_3\_S3        & \textbf{4.272} & 17.02                                                                             & 6.688           & 6.387            & 12.701           & 18.736             & 19.288             & 15.456               & 23.524          & 21.608            \\ \hline
    sofa\_3\_S2        & \textbf{4.567} & 16.181                                                                            & 6.169           & 5.986            & 6.57             & 15.084             & 20.977             & 14.355               & 23.436          & 22.014            \\ \hline
    sofa\_3\_S1        & \textbf{3.413} & 11.27                                                                             & 4.766           & 4.185            & 4.412            & 13.232             & 23.584             & 14.913               & 16.549          & 14.023            \\ \hline
    sofa\_1\_S3        & 4.513          & 16.788                                                                            & 4.485           & \textbf{4.01}    & 10.122           & 17.564             & 27.773             & 16.158               & 22.573          & 23.399            \\ \hline
    sofa\_1\_S2        & 5.796          & 13.385                                                                            & 3.687           & \textbf{4.92}    & 4.658            & 13.339             & 27.74              & 14.893               & 22.384          & 22.026            \\ \hline
    sofa\_1\_S1        & 5.474          & 10.798                                                                            & 5.449           & \textbf{5.23}    & 5.2              & 15.536             & 30.814             & 18.724               & 7.518           & 20.873            \\ \hline
    bench\_1\_S3       & 5.142          & 16.187                                                                            & 5.887           & \textbf{4.674}   & 153.809          & 16.129             & 17.476             & 15.43                & 30.692          & 23.743            \\ \hline
    bench\_1\_S2       & 4.551          & 14.132                                                                            & 3.899           & \textbf{3.906}   & 112.483          & 13.554             & 18.607             & 12.978               & 26.192          & 17.842            \\ \hline
    bench\_1\_S1       & \textbf{3.543} & 10.362                                                                            & 4.36            & 4.379            & 4.493            & 10.929             & 20.693             & 12.559               & 20.622          & 15.442            \\ \hline
    guitar\_1\_S3      & \textbf{7.381} & 14.657                                                                            & 10.117          & 10.177           & 163.681          & 16.889             & 7.67               & 15.33                & 20.392          & 25.868            \\ \hline
    guitar\_1\_S2      & \textbf{6.18}  & 11.316                                                                            & 7.614           & 7.65             & 137.846          & 12.287             & 6.419              & 10.392               & 19.465          & 25.338            \\ \hline
    guitar\_1\_S1      & \textbf{3.011} & 11.2                                                                              & 7.332           & 7.415            & 9.681            & 9.027              & 5.831              & 8.532                & 18.356          & 26.169            \\ \hline
    suzanne\_S3        & \textbf{5.719} & 16.841                                                                            & 6.748           & 7.252            & 25.533           & 16.781             & 6.49               & 15.598               & 34.363          & 34.366            \\ \hline
    suzanne\_S2        & \textbf{4.141} & 14.976                                                                            & 6.327           & 6.516            & 5.124            & 15.299             & 7.623              & 14.374               & 32.841          & 27.383            \\ \hline
    suzanne\_S1        & \textbf{4.797} & 10.417                                                                            & 5.473           & 5.442            & \textbf{4.721}   & 9.305              & 7.051              & 14.159               & 30.643          & 23.145            \\ \hline
    teapot\_S3         & \textbf{6.152} & 17.25                                                                             & 11.014          & 11.161           & 118.128          & 18.79              & 11.768             & 15.414               & 26.124          & 25.139            \\ \hline
    teapot\_S2         & \textbf{7.314} & 15.836                                                                            & 10.832          & 11.473           & 17.74            & 15.421             & 11.692             & 13.771               & 23.316          & 22.191            \\ \hline
    teapot\_S1         & \textbf{7.58}  & 11.849                                                                            & 10.854          & 11.121           & 6.713            & 10.7               & 11.139             & 13.41                & 24.015          & 18.796            \\ \hline
    armadillo\_S3      & \textbf{5.812} & 19.349                                                                            & 23.069          & 23.48            & 117.21           & 18.891             & 10.438             & 17.449               & 41.609          & 37.975            \\ \hline
    armadillo\_S2      & \textbf{4.247} & 16.592                                                                            & 23.446          & 23.703           & 8.058            & 14.834             & 12.623             & 12.515               & 40.507          & 36.749            \\ \hline
    armadillo\_S1      & \textbf{4.342} & 12.021                                                                            & 22.962          & 24.023           & 6.943            & 16                 & 14.953             & 14.678               & 39.251          & 35.446            \\ \hline
    bunny\_S3          & \textbf{3.917} & 25.159                                                                            & 19.27           & 19.099           & 25.02            & 20.017             & 20.489             & 17.757               & 75.799          & 50.199            \\ \hline
    bunny\_S2          & \textbf{6.974} & 18.543                                                                            & 19.11           & 19.198           & 8.336            & 14.737             & 21.744             & 18.383               & 66.464          & 47.761            \\ \hline
    bunny\_S1          & \textbf{3.967} & 13.767                                                                            & 19.421          & 19.429           & 5.853            & 13.703             & 23.577             & 14.289               & 60.246          & 52.51             \\ \hline
    mobile\_1\_S3      & 4.189          & 16.22                                                                             & 4.63            & \textbf{3.216}   & 82.184           & 18.032             & 5.012              & 15.388               & 29.668          & 18.336            \\ \hline
    mobile\_1\_S2      & 3.579          & 13.233                                                                            & 3.94            & \textbf{3.212}   & 4.531            & 14.088             & 3.509              & 12.554               & 28.743          & 16.21             \\ \hline
    mobile\_1\_S1      & 2.869          & 9.153                                                                             & 2.979           & \textbf{2.484}   & 3.201            & 10.468             & 5.165              & 10.361               & 28.298          & 16.223            \\ \hline
    speaker\_1\_S3     & \textbf{5.033} & 18.527                                                                            & 8.624           & 8.327            & 17.667           & 19.866             & 14.627             & 16.993               & 34.595          & 48.009            \\ \hline
    speaker\_1\_S2     & \textbf{5.141} & 14.938                                                                            & 7.281           & 7.328            & 21.282           & 15.363             & 14.223             & 14.691               & 17.407          & 47.535            \\ \hline
    speaker\_1\_S1     & \textbf{4.737} & 14.208                                                                            & 5.145           & \textbf{4.72}    & 5.276            & 13.521             & 12.756             & 12.722               & 9.65            & 47.674            \\ \hline
    mailbox\_1\_S3     & 10.031         & 16.761                                                                            & 6.169           & \textbf{5.413}   & 38.719           & 20.275             & 5.141              & 15.046               & 35.726          & 27.246            \\ \hline
    mailbox\_1\_S2     & 10.368         & 12.95                                                                             & 4.312           & \textbf{4.874}   & 9.469            & 12.663             & 4.967              & 12.598               & 30.253          & 26.995            \\ \hline
    mailbox\_1\_S1     & 9.868          & 11.426                                                                            & 4.425           & \textbf{6.325}   & 4.245            & 8.704              & 6.586              & 8.777                & 26.279          & 27.183            \\ \hline
    camera\_1\_S3      & 25.875         & 19.303                                                                            & 5.048           & \textbf{5.05}    & 29.712           & 16.55              & 5.022              & 16.955               & 33.868          & 39.07             \\ \hline
    camera\_1\_S2      & 5.622          & 12.472                                                                            & 5.091           & \textbf{5.331}   & 6.308            & 14.748             & 9.199              & 15.082               & 35.641          & 35.475            \\ \hline
    camera\_1\_S1      & 4.787          & 7.959                                                                             & 4.943           & \textbf{4.081}   & 4.349            & 13.25              & 14.537             & 17.113               & 26.015          & 30.47             \\ \hline
    table\_1\_S3       & \textbf{2.944} & 21.648                                                                            & 4.165           & \textbf{2.996}   & 30.838           & 18.05              & 14.61              & 16.602               & 46.057          & 26.448            \\ \hline
    table\_1\_S2       & \textbf{3.182} & 16.677                                                                            & 3.886           & \textbf{3.124}   & 6.728            & 15.498             & 13.449             & 16.89                & 42.707          & 17.832            \\ \hline
    table\_1\_S1       & \textbf{3.185} & 8.448                                                                             & 3.563           & \textbf{3.369}   & 3.934            & 12.077             & 13.865             & 16.576               & 40.015          & 16.728            \\ \hline
    table\_2\_S3       & \textbf{4.791} & 18.325                                                                            & 21.119          & 21.256           & 47.586           & 20.083             & 24.451             & 20.052               & 39.425          & 40.386            \\ \hline
    table\_2\_S2       & \textbf{4.568} & 15.504                                                                            & 21.173          & 21.174           & 28.283           & 13.954             & 21.651             & 15.999               & 38.058          & 43.241            \\ \hline
    table\_2\_S1       & \textbf{5.523} & 9.842                                                                             & 20.715          & 21.051           & 12.037           & 13.089             & 24.821             & 18.586               & 37.659          & 38.992            \\ \hline
    monitor\_1\_S3     & \textbf{4.985} & 19.937                                                                            & \textbf{4.791}  & 5.151            & 27.943           & 18.517             & 8.319              & 16.546               & 32.03           & 26.32             \\ \hline
    monitor\_1\_S2     & \textbf{5.325} & 16.997                                                                            & 5.748           & 5.66             & 5.546            & 16.31              & 7.464              & 15.472               & 14.612          & 21.924            \\ \hline
    monitor\_1\_S1     & \textbf{5.036} & 8.414                                                                             & 5.448           & 5.518            & 5.156            & 13.91              & 5.522              & 15.862               & 9.04            & 21.262            \\ \hline
    lamp\_1\_S3        & \textbf{3.934} & 18.594                                                                            & 11.321          & 11.347           & 10.307           & 18.322             & 9.497              & 15.259               & 43.916          & 29.247            \\ \hline
    lamp\_1\_S2        & \textbf{3.112} & 13.952                                                                            & 10.25           & 10.115           & 4.465            & 13.1               & 11.141             & 14.075               & 42.604          & 27.942            \\ \hline
    lamp\_1\_S1        & \textbf{4.107} & 12.046                                                                            & 9.68            & 9.792            & 4.88             & 11.655             & 15.259             & 17.152               & 41.499          & 27.995            \\ \hline
    \multicolumn{11}{|l|}{}                                                                                                                                                                                                                                                                \\ \hline
    mean               & \textbf{5.482} & 14.655                                                                            & 9.974           & 10.255           & 33.871           & 14.921             & 14.741             & 15.069               & 30.497          & 29.363            \\ \hline
    median             & \textbf{4.789} & 14.791                                                                            & 6.507           & 6.931            & 10.2145          & 14.809             & 14.044             & 15.359               & 28.520          & 26.384            \\ \hline
    \end{tabular}
    \caption{Hausdorff distances between GT and estimated mesh. 
            The distances are multiplied by a factor of 100 for better readability. 
            We highlight in bold the best performing approach as well as other approaches that are within 0.2 error of the best approach.
            We show results with normals estimated with Meshlab~\cite{meshlab} and PCPNet~\cite{pcpnet}. We reconstruct the resulting point clouds with both RILMS~\cite{oztireli2009feature} and Screened Poisson~\cite{screenedpoisson}. 
            Laplacian regularizer~\cite{laplacian} is shown for two levels of smoothing. We also show three recent deep learning approaches: Deep Geometric Prior~\cite{Williams2019}, AtlasNet~\cite{Groueix_2018_CVPR} and OccNet~\cite{occNet}.}
    \label{tab:hausdorff}
    \end{table*}

\begin{table*}[]
    \centering
    \scriptsize
    \begin{tabular}{|l|l|l|l|l|l|l|l|l|l|l|}
    \hline
    \textbf{obj\_name} & \textbf{Ours}  & \textbf{DGP} & \textbf{Lap-low} & \textbf{Lap-high} & \textbf{\cite{meshlab}+\cite{screenedpoisson}} & \cite{meshlab}+\cite{oztireli2009feature} & \cite{pcpnet}+\cite{screenedpoisson} & \cite{pcpnet}+\cite{oztireli2009feature} & \textbf{OccNet} & \textbf{AtlasNet} \\ \hline
    clock\_1\_S3       & \textbf{0.792} & 1.658          & 1.199           & 1.071            & 9.294            & 2.562              & 1.265              & 2.378                & 8.748           & 4.26              \\ \hline
    clock\_1\_S2       & \textbf{0.759} & 1.266          & 1.019           & 0.962            & 1.008            & 1.781              & 1.252              & 1.6                  & 5.506           & 3.823             \\ \hline
    clock\_1\_S1       & \textbf{0.735} & 0.908          & 0.931           & 0.943            & 0.827            & 0.995              & 1.301              & 1.374                & 3.951           & 3.576             \\ \hline
    chair\_1\_S3       & \textbf{0.873} & 1.589          & 1.431           & 1.448            & 7.374            & 2.635              & 1.432              & 2.644                & 7.326           & 3.052             \\ \hline
    chair\_1\_S2       & \textbf{0.761} & 1.204          & 1.21            & 1.392            & 9.841            & 1.871              & 1.247              & 1.821                & 5.169           & 2.358             \\ \hline
    chair\_1\_S1       & \textbf{0.747} & 0.904          & 0.989           & 1.157            & 0.846            & 1.104              & 1.271              & 1.524                & 3.008           & 1.897             \\ \hline
    sofa\_4\_S3        & \textbf{1.14}  & 1.669          & 1.188           & 1.166            & 2.118            & 2.633              & 1.54               & 2.253                & 9.094           & 7.69              \\ \hline
    sofa\_4\_S2        & 1.197          & 1.32           & \textbf{1.13}   & 1.163            & 1.226            & 1.567              & 1.757              & 1.883                & 6.557           & 7.39              \\ \hline
    sofa\_4\_S1        & 1.361          & \textbf{1.02}  & 1.127           & 1.171            & 1.112            & 1.148              & 2.591              & 2.007                & 3.453           & 7.178             \\ \hline
    sofa\_2\_S3        & \textbf{0.985} & 1.631          & 1.24            & 1.26             & 6.973            & 2.524              & 1.359              & 2.398                & 7.678           & 7.688             \\ \hline
    sofa\_2\_S2        & \textbf{0.967} & 1.228          & 1.044           & 1.082            & 2.205            & 1.701              & 1.379              & 1.699                & 5.169           & 6.886             \\ \hline
    sofa\_2\_S1        & \textbf{0.952} & \textbf{0.956} & \textbf{0.997}  & 1.043            & 1.257            & 1.213              & 1.785              & 1.727                & 3.256           & 6.304             \\ \hline
    sofa\_3\_S3        & 0.881          & 1.714          & 1.079           & \textbf{0.823}   & 1.491            & 2.567              & 1.381              & 2.409                & 6.947           & 3.619             \\ \hline
    sofa\_3\_S2        & 0.832          & 1.272          & 0.922           & \textbf{0.792}   & 0.939            & 1.586              & 1.333              & 1.693                & 5.558           & 2.655             \\ \hline
    sofa\_3\_S1        & 0.786          & 0.936          & 0.864           & \textbf{0.759}   & \textbf{0.776}   & 0.957              & 1.583              & 1.706                & 2.555           & 2.029             \\ \hline
    sofa\_1\_S3        & \textbf{0.951} & 1.642          & 1.08            & \textbf{0.95}    & 1.369            & 2.551              & 1.672              & 2.356                & 7.21            & 3.583             \\ \hline
    sofa\_1\_S2        & 0.976          & 1.301          & 0.988           & \textbf{0.901}   & 0.976            & 1.513              & 1.714              & 1.812                & 5.508           & 2.775             \\ \hline
    sofa\_1\_S1        & 0.937          & 0.991          & 0.925           & \textbf{0.869}   & 0.891            & 1.031              & 2.07               & 1.938                & 2.602           & 2.384             \\ \hline
    bench\_1\_S3       & \textbf{0.938} & 1.547          & 1.199           & \textbf{0.953}   & 11.713           & 2.485              & 1.476              & 2.502                & 7.794           & 3.876             \\ \hline
    bench\_1\_S2       & \textbf{0.793} & 1.202          & 0.916           & 0.809            & 9.116            & 1.827              & 1.514              & 1.788                & 5.702           & 2.697             \\ \hline
    bench\_1\_S1       & \textbf{0.748} & 0.947          & 0.834           & 0.794            & 0.814            & 1.087              & 1.867              & 1.496                & 3.524           & 1.961             \\ \hline
    guitar\_1\_S3      & \textbf{0.973} & 1.487          & 1.77            & 1.455            & 18.503           & 2.51               & 1.229              & 3.014                & 9.138           & 4.396             \\ \hline
    guitar\_1\_S2      & \textbf{0.657} & 1.161          & 1.046           & 0.837            & 12.261           & 2.004              & 0.8                & 1.945                & 6.986           & 3.166             \\ \hline
    guitar\_1\_S1      & \textbf{0.475} & 0.859          & 0.712           & 0.597            & 0.869            & 1.26               & 0.642              & 1.011                & 4.413           & 2.38              \\ \hline
    suzanne\_S3        & 1.013          & 1.684          & 1.123           & \textbf{0.93}    & 2.467            & 2.551              & 0.955              & 2.217                & 7.967           & 4.529             \\ \hline
    suzanne\_S2        & 0.906          & 1.247          & 0.95            & \textbf{0.894}   & 0.937            & 1.597              & \textbf{0.873}     & 1.399                & 6.573           & 4.162             \\ \hline
    suzanne\_S1        & 0.954          & 0.933          & 0.899           & 0.873            & \textbf{0.756}   & 0.916              & 0.916              & 1.21                 & 5.825           & 3.991             \\ \hline
    teapot\_S3         & \textbf{0.827} & 1.72           & 1.184           & \textbf{0.841}   & 9.856            & 2.694              & 1.008              & 2.318                & 6.316           & 3.777             \\ \hline
    teapot\_S2         & \textbf{0.739} & 1.238          & 0.882           & \textbf{0.744}   & 1.118            & 1.832              & 0.869              & 1.465                & 5.265           & 3.336             \\ \hline
    teapot\_S1         & \textbf{0.736} & 0.912          & 0.818           & \textbf{0.73}    & 0.711            & 0.977              & 0.876              & 1.206                & 4.66            & 2.902             \\ \hline
    armadillo\_S3      & \textbf{1.033} & 1.653          & 1.301           & 1.247            & 7.371            & 2.778              & 1.062              & 2.456                & 9.835           & 6.23              \\ \hline
    armadillo\_S2      & \textbf{0.908} & 1.208          & 1.071           & 1.132            & 0.999            & 1.735              & 0.972              & 1.604                & 8.848           & 5.931             \\ \hline
    armadillo\_S1      & 0.962          & 0.914          & 1.007           & 1.1              & \textbf{0.809}   & 1.046              & 1.082              & 1.44                 & 7.576           & 5.639             \\ \hline
    bunny\_S3          & \textbf{1.037} & 1.716          & 1.133           & 1.087            & 1.852            & 2.569              & 1.312              & 2.261                & 13.601          & 7.731             \\ \hline
    bunny\_S2          & 1.02           & 1.277          & 1.033           & 1.05             & \textbf{0.962}   & 1.453              & 1.283              & 1.529                & 12.126          & 7.482             \\ \hline
    bunny\_S1          & 1.044          & 0.976          & 0.989           & 1.048            & \textbf{0.839}   & 0.965              & 1.434              & 1.392                & 11.111          & 7.348             \\ \hline
    mobile\_1\_S3      & \textbf{0.778} & 1.72           & 1.123           & 0.795            & 6.66             & 2.835              & 0.94               & 2.332                & 7.237           & 4.035             \\ \hline
    mobile\_1\_S2      & \textbf{0.686} & 1.264          & 0.91            & 0.71             & 0.938            & 1.731              & 0.772              & 1.313                & 5.552           & 2.621             \\ \hline
    mobile\_1\_S1      & \textbf{0.672} & 0.897          & 0.781           & \textbf{0.656}   & 0.738            & 0.925              & 0.717              & 0.903                & 4.136           & 1.832             \\ \hline
    speaker\_1\_S3     & 0.966          & 1.659          & 1.126           & \textbf{0.945}   & 1.686            & 2.706              & 1.08               & 2.256                & 8.276           & 3.621             \\ \hline
    speaker\_1\_S2     & 0.92           & 1.244          & 0.987           & \textbf{0.887}   & 1.381            & 1.692              & 1.024              & 1.515                & 4.767           & 3.328             \\ \hline
    speaker\_1\_S1     & 0.925          & 0.932          & 0.919           & \textbf{0.868}   & \textbf{0.873}   & 1.077              & 0.986              & 1.259                & 3.059           & 3.476             \\ \hline
    mailbox\_1\_S3     & \textbf{0.976} & 1.654          & 1.437           & 1.003            & 3.53             & 2.505              & 1.029              & 2.615                & 7.939           & 4.331             \\ \hline
    mailbox\_1\_S2     & \textbf{0.732} & 1.201          & 0.944           & \textbf{0.733}   & 1.122            & 1.908              & 0.78               & 1.665                & 6.871           & 3.403             \\ \hline
    mailbox\_1\_S1     & 0.68           & 0.805          & 0.732           & \textbf{0.648}   & 0.702            & 0.978              & 0.695              & 0.935                & 5.187           & 2.928             \\ \hline
    camera\_1\_S3      & 1.021          & 1.646          & 1.072           & \textbf{0.955}   & 2.528            & 2.495              & 0.984              & 2.057                & 6.927           & 4.052             \\ \hline
    camera\_1\_S2      & 1.075          & 1.252          & 1.027           & \textbf{0.953}   & 1.008            & 1.62               & 0.979              & 1.544                & 6.225           & 3.554             \\ \hline
    camera\_1\_S1      & 0.999          & 0.954          & 0.954           & \textbf{0.922}   & \textbf{0.93}    & 1.058              & 1.29               & 1.786                & 4.895           & 3.425             \\ \hline
    table\_1\_S3       & 0.825          & 1.697          & 1.032           & \textbf{0.792}   & 1.758            & 2.53               & 1.092              & 2.191                & 9.56            & 3.899             \\ \hline
    table\_1\_S2       & 0.788          & 1.266          & 0.908           & \textbf{0.755}   & 0.912            & 1.506              & 1.009              & 1.448                & 6.487           & 3.719             \\ \hline
    table\_1\_S1       & 0.803          & 0.949          & 0.865           & \textbf{0.745}   & \textbf{0.767}   & 0.893              & 1.018              & 1.286                & 3.635           & 3.757             \\ \hline
    table\_2\_S3       & \textbf{0.934} & 1.66           & 1.202           & 1.124            & 3.284            & 2.735              & 1.539              & 2.482                & 8.435           & 5.142             \\ \hline
    table\_2\_S2       & \textbf{0.925} & 1.279          & 1.1             & 1.08             & 1.665            & 1.622              & 1.474              & 1.91                 & 6.948           & 4.359             \\ \hline
    table\_2\_S1       & \textbf{0.926} & 0.952          & 1.023           & 1.036            & 1.058            & 1.05               & 1.832              & 1.959                & 4.537           & 3.88              \\ \hline
    monitor\_1\_S3     & 1.094          & 1.66           & 1.109           & 1.072            & 2.048            & 2.605              & \textbf{1.03}      & 2.033                & 6.693           & 4.368             \\ \hline
    monitor\_1\_S2     & 1.179          & 1.273          & 1.061           & 1.067            & 1.018            & 1.537              & \textbf{0.977}     & 1.386                & 4.236           & 4.261             \\ \hline
    monitor\_1\_S1     & 1.13           & \textbf{0.96}  & 1.005           & 1.046            & \textbf{0.957}   & 1.017              & 1.033              & 1.338                & 2.354           & 4.017             \\ \hline
    lamp\_1\_S3        & \textbf{0.8}   & 1.742          & 1.06            & 0.86             & 1.361            & 2.544              & 1.019              & 2.089                & 8.63            & 3.667             \\ \hline
    lamp\_1\_S2        & \textbf{0.766} & 1.283          & 0.933           & 0.823            & 0.879            & 1.376              & 0.934              & 1.415                & 5.624           & 3.359             \\ \hline
    lamp\_1\_S1        & \textbf{0.772} & 0.969          & 0.88            & 0.8              & 0.728            & 0.891              & 1.004              & 1.441                & 3.028           & 2.998             \\ \hline
    \multicolumn{11}{|l|}{}                                                                                                                                                                                             \\ \hline
    mean               & \textbf{0.896} & 1.280          & 1.040           & 0.956            & 2.850            & 1.768              & 1.222              & 1.811                & 6.297           & 4.145             \\ \hline
    median             & \textbf{0.922} & 1.258          & 1.025           & 0.944            & 1.115            & 1.657              & 1.087              & 1.757                & 6.270           & 3.767             \\ \hline
    \end{tabular}
    \caption{Chamfer-$\ell_1$ distances between GT and estimated mesh.
            The distances are multiplied by a factor of 100 for better readability. 
            We highlight in bold the best performing approach as well as other approaches that are within 0.02 error of the best approach.
            We show results with normals estimated with Meshlab~\cite{meshlab} and PCPNet~\cite{pcpnet}. We reconstruct the resulting point clouds with both RILMS~\cite{oztireli2009feature} and Screened Poisson~\cite{screenedpoisson}. 
            Laplacian regularizer~\cite{laplacian} is shown for two levels of smoothing. We also show three recent deep learning approaches: Deep Geometric Prior~\cite{Williams2019}, AtlasNet~\cite{Groueix_2018_CVPR} and OccNet~\cite{occNet}.}
    \label{tab:chamfer}
    \end{table*}

%% file: figures_supplementary/results_cvpr2020/result_page.tex

\setlength{\numcrops}{11pt}
\newcommand{\fitscale}{.98}

\settowidth{\cropwidth}{\includegraphics{figures/results/armadillo_new/snapshot00.jpg}}

\setlength{\one}{1pt}

\setlength\tgtwidth{\textwidth*\ratio{\one}{\numcrops}}

\captionsetup[subfigure]{labelformat=empty}
\centering
\vspace{-10mm}
\subfloat{\includegraphics[width=\fitscale\tgtwidth, trim={450 152 450 152}, clip]{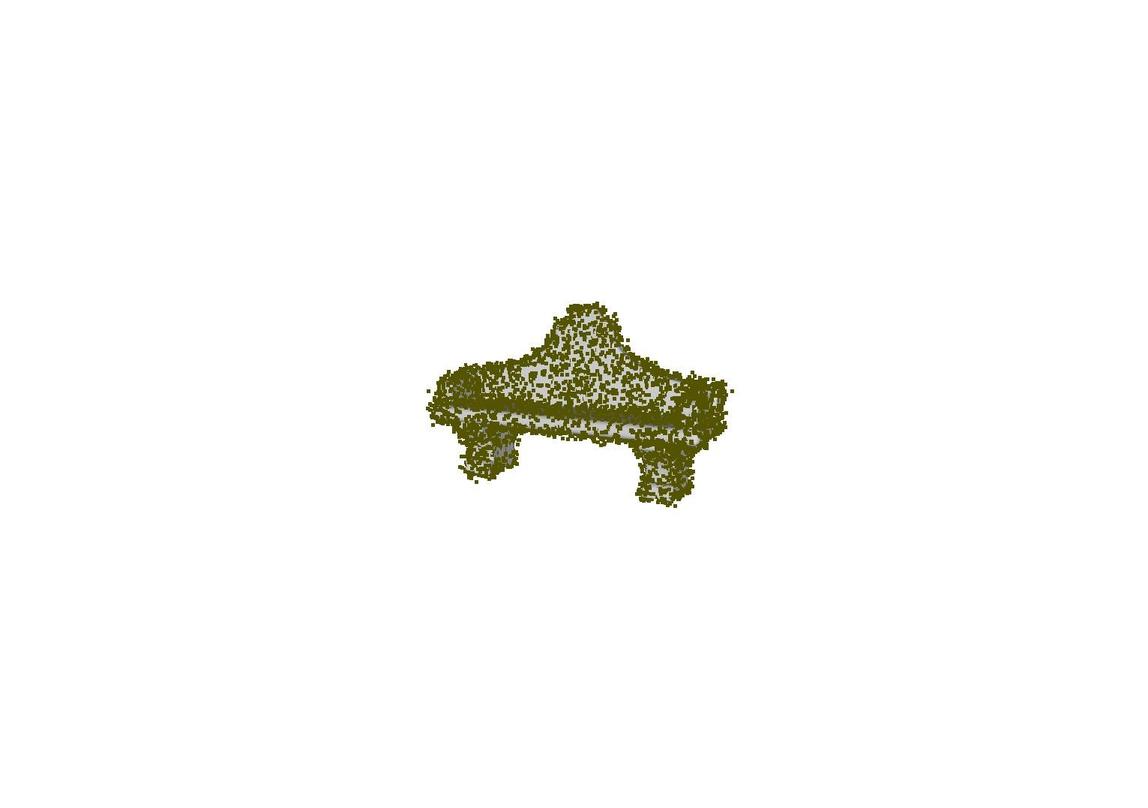}}
\subfloat{\includegraphics[width=\fitscale\tgtwidth, trim={450 152 450 152}, clip]{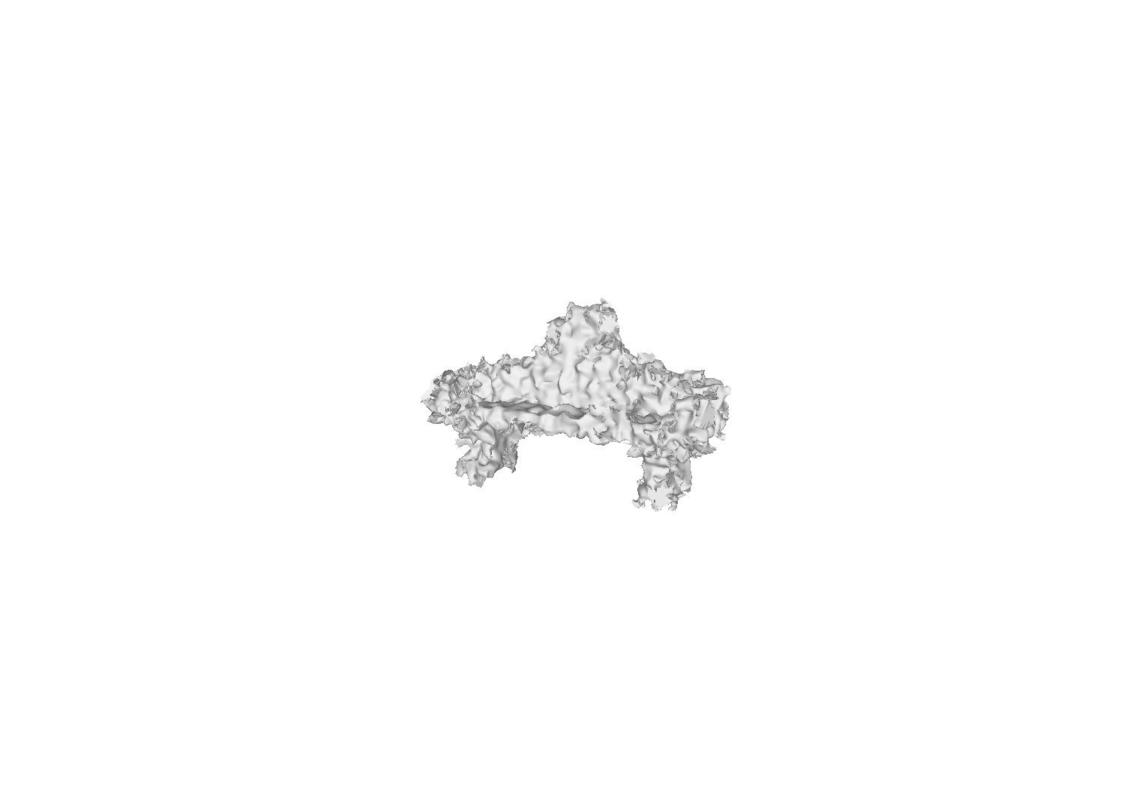}}
\subfloat{\includegraphics[width=\fitscale\tgtwidth, trim={450 152 450 152}, clip]{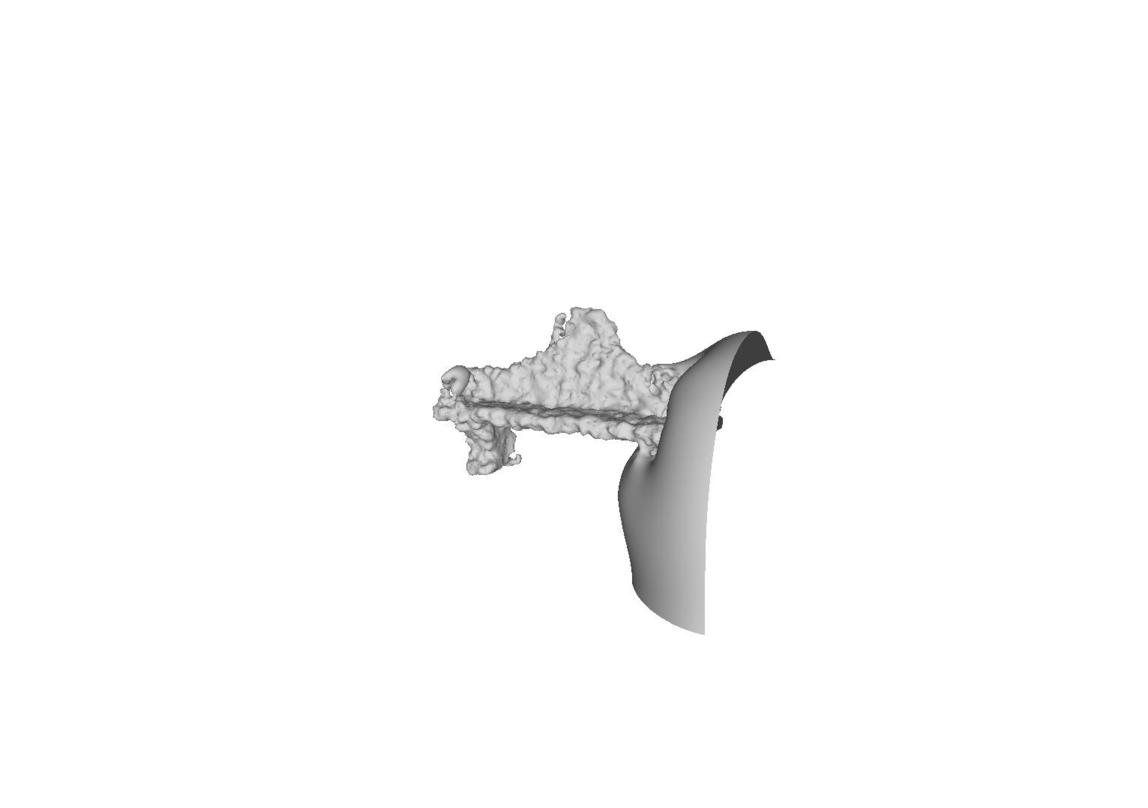}}
\subfloat{\includegraphics[width=\fitscale\tgtwidth, trim={450 152 450 152}, clip]{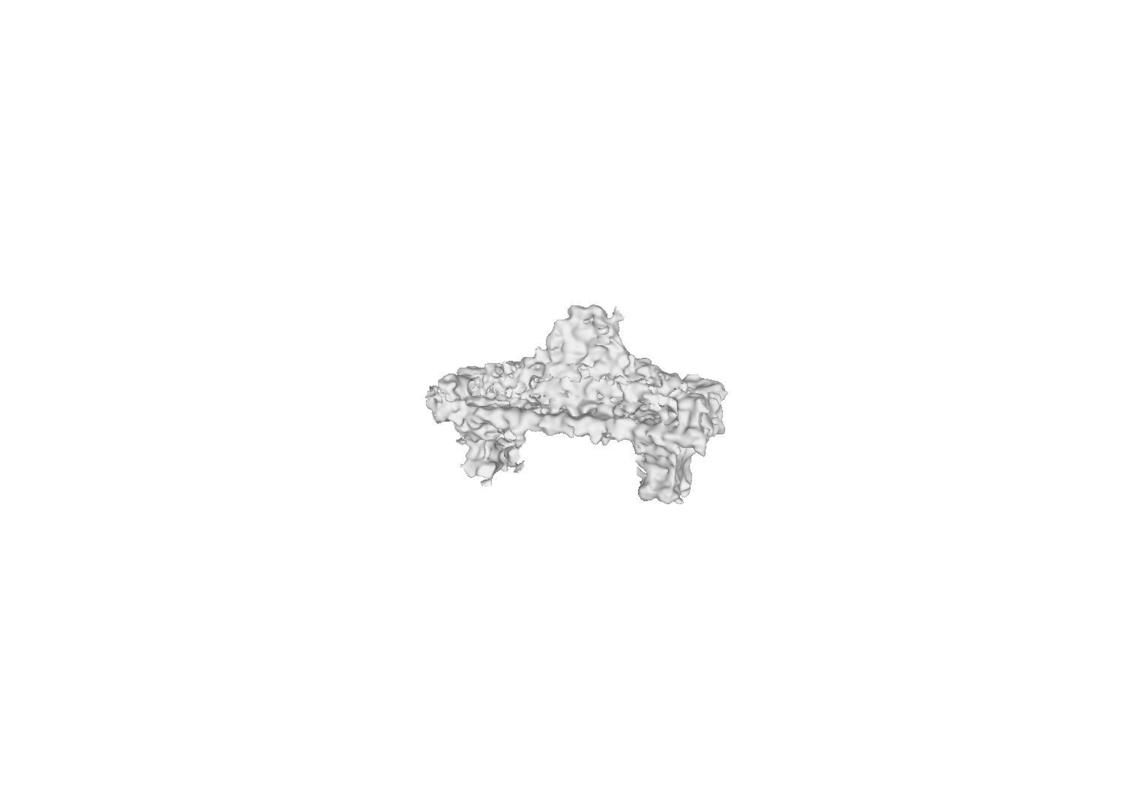}}
\subfloat{\includegraphics[width=\fitscale\tgtwidth, trim={450 152 450 152}, clip]{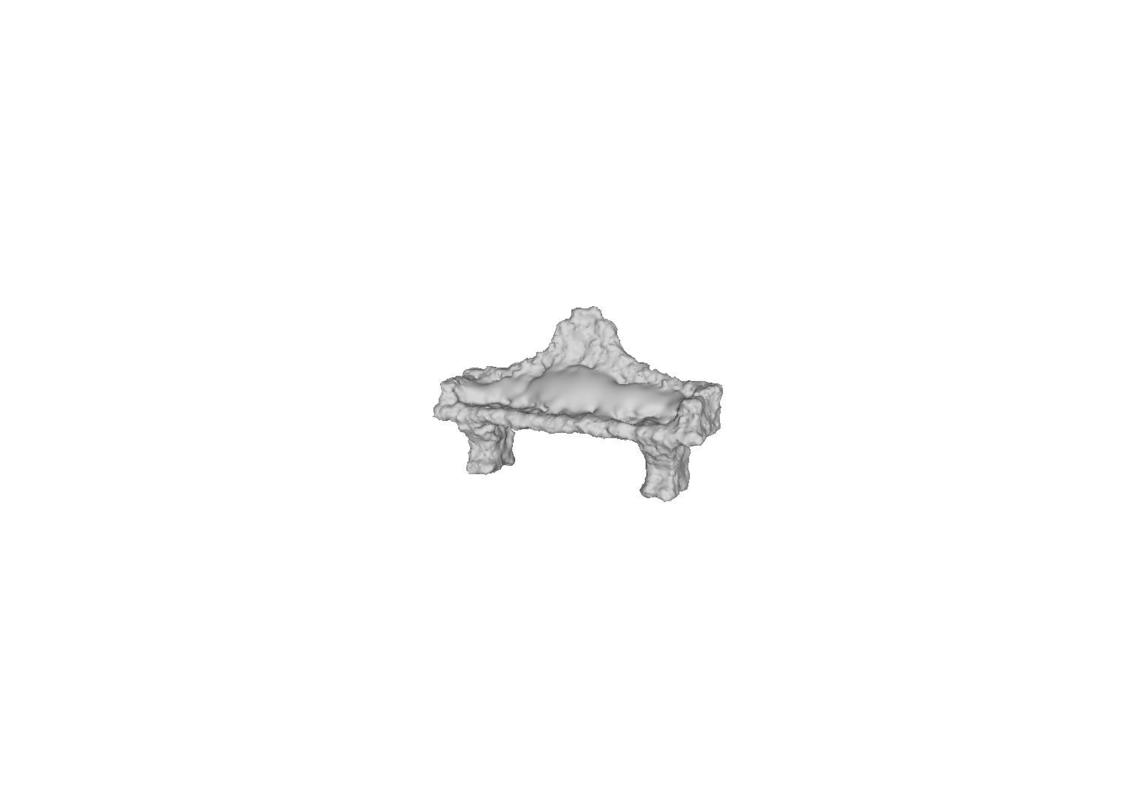}}
\subfloat{\includegraphics[width=\fitscale\tgtwidth, trim={450 152 450 152}, clip]{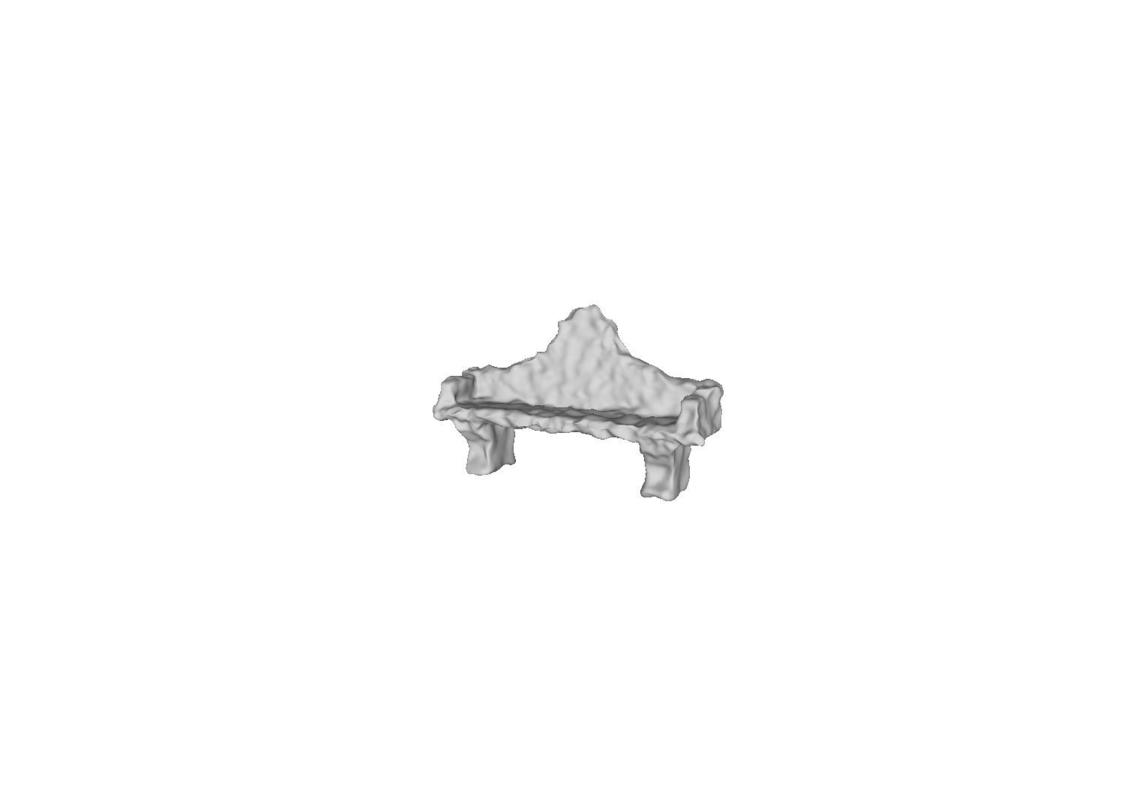}}
\subfloat{\includegraphics[width=\fitscale\tgtwidth, trim={450 152 450 152}, clip]{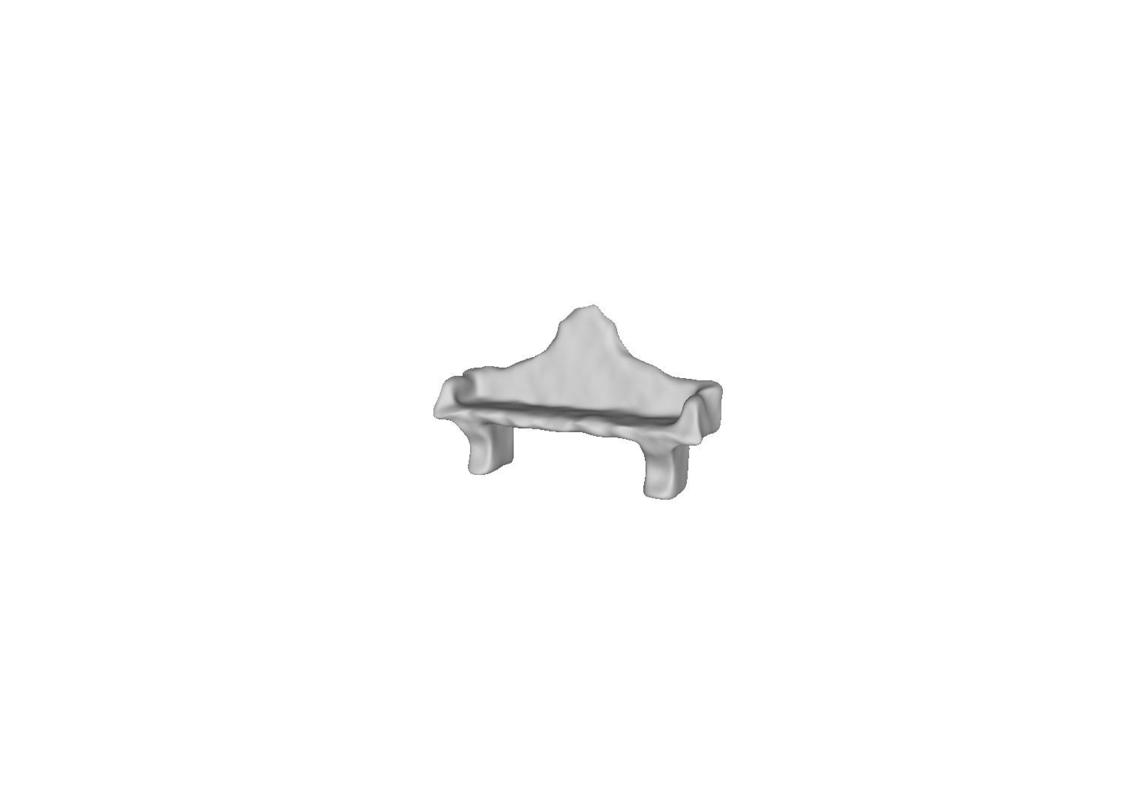}}
\subfloat{\includegraphics[width=\fitscale\tgtwidth, trim={450 152 450 152}, clip]{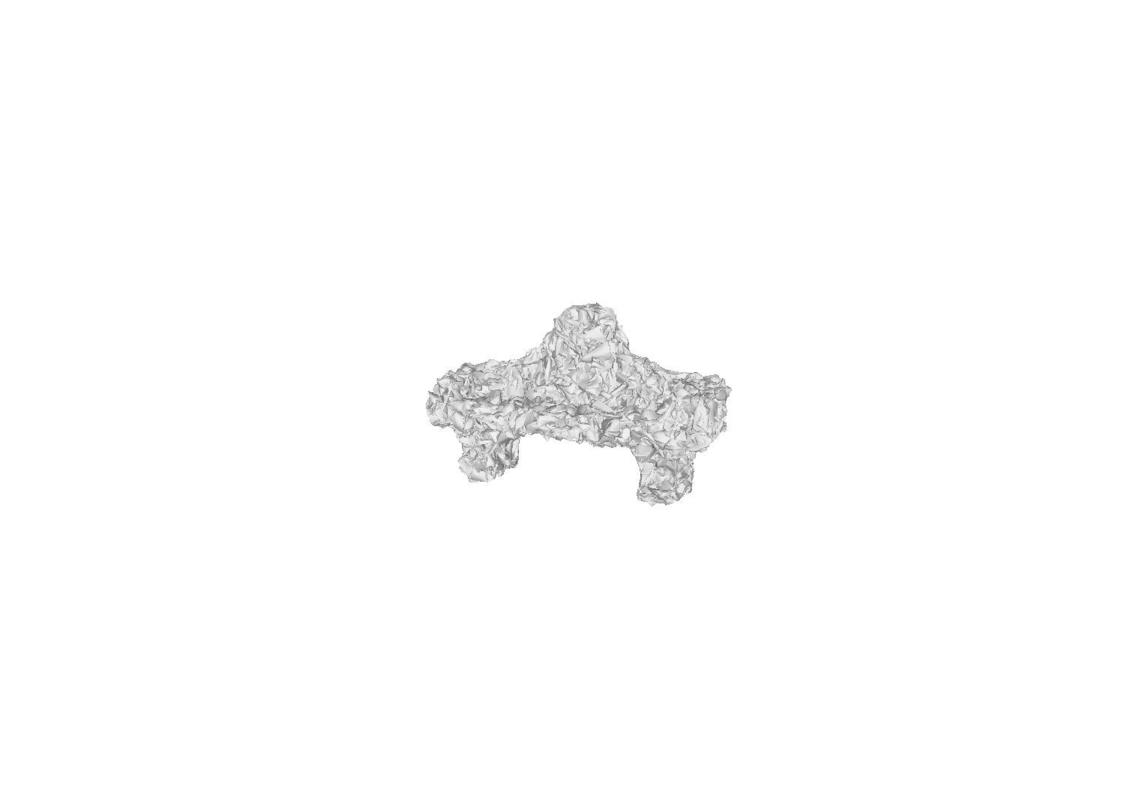}}
\subfloat{\includegraphics[width=\fitscale\tgtwidth, trim={450 152 450 152}, clip]{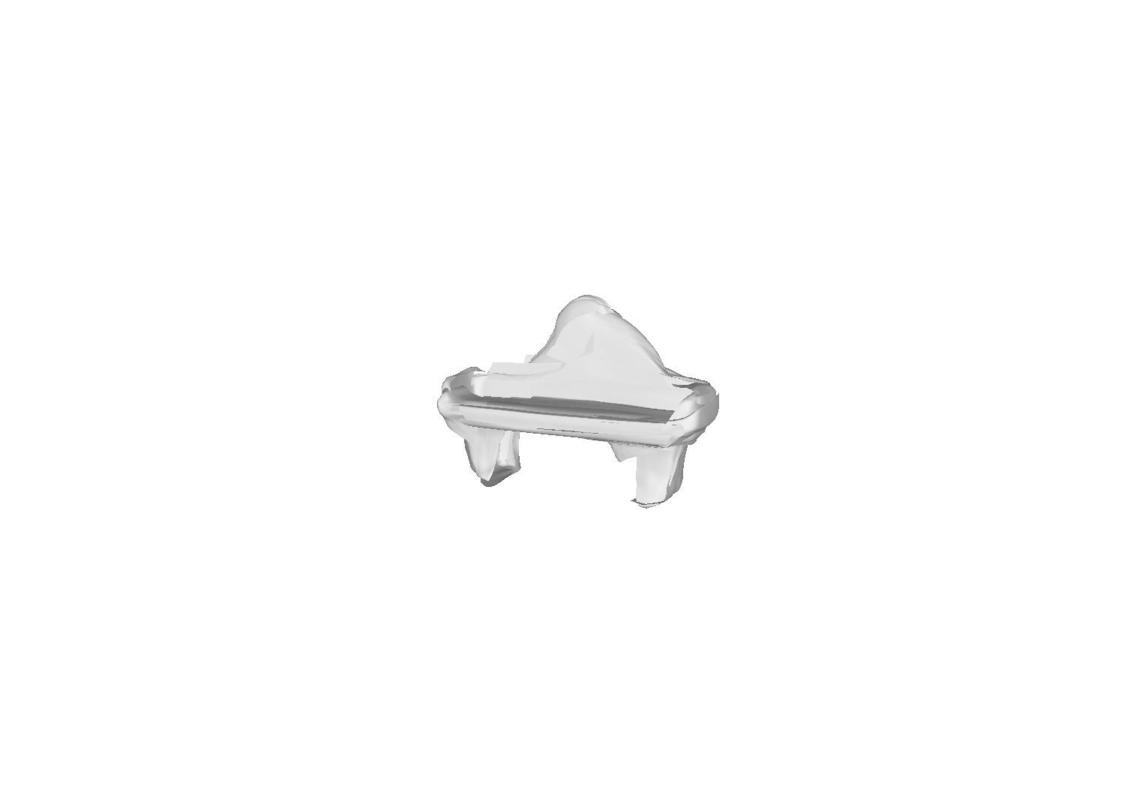}}
\subfloat{\includegraphics[width=\fitscale\tgtwidth, trim={450 152 450 152}, clip]{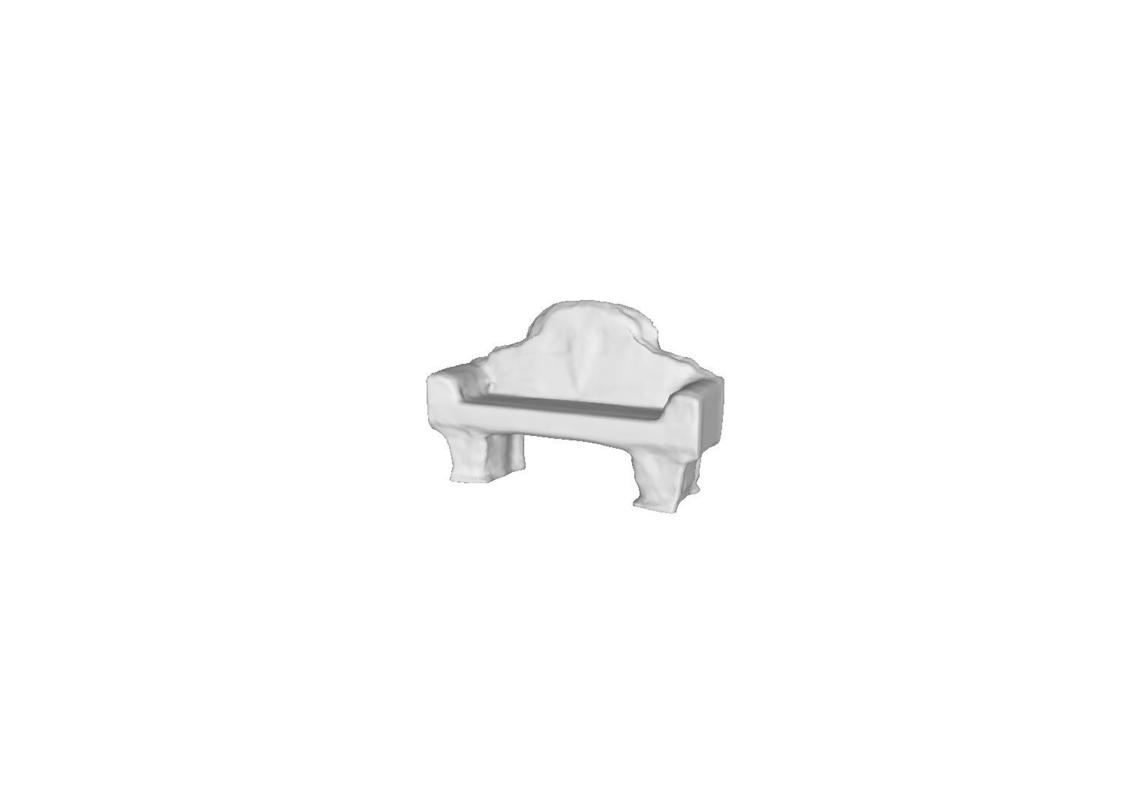}}
~
\subfloat{\includegraphics[width=\fitscale\tgtwidth, trim={450 152 450 152}, clip]{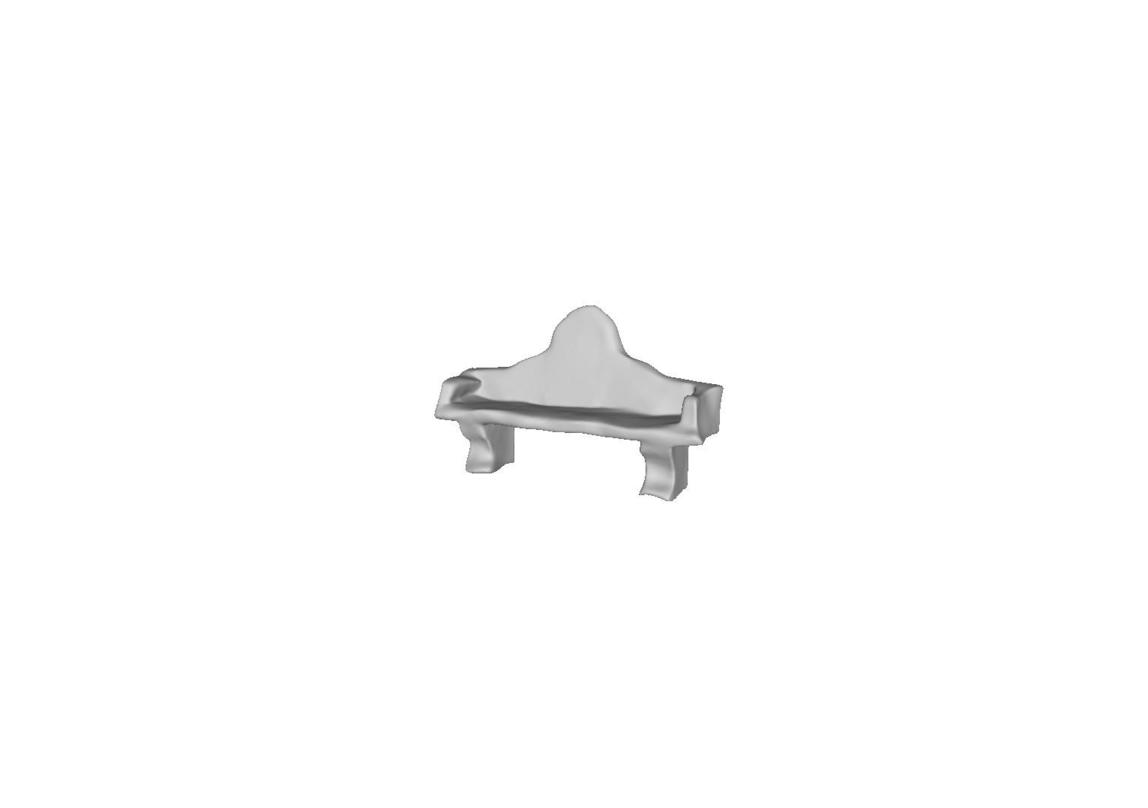}}
\\
\vspace{-5mm}
\subfloat{\includegraphics[width=\fitscale\tgtwidth, trim={450 152 450 152}, clip]{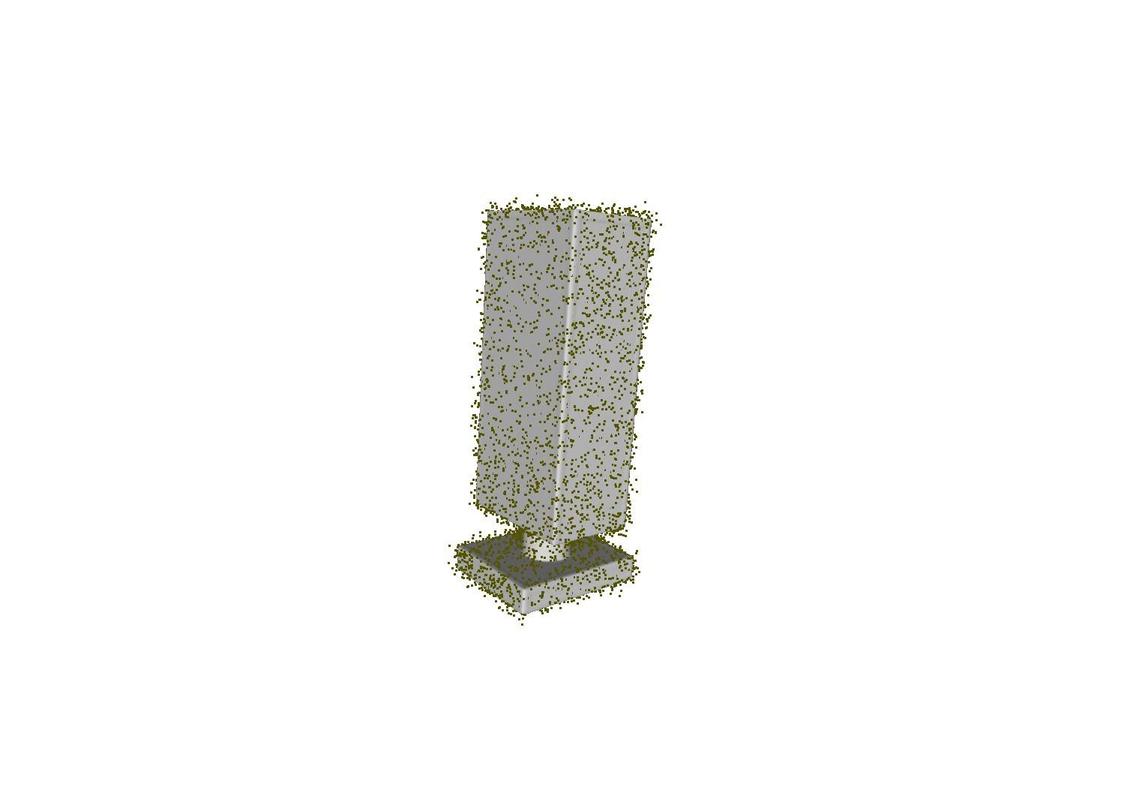}}
\subfloat{\includegraphics[width=\fitscale\tgtwidth, trim={450 152 450 152}, clip]{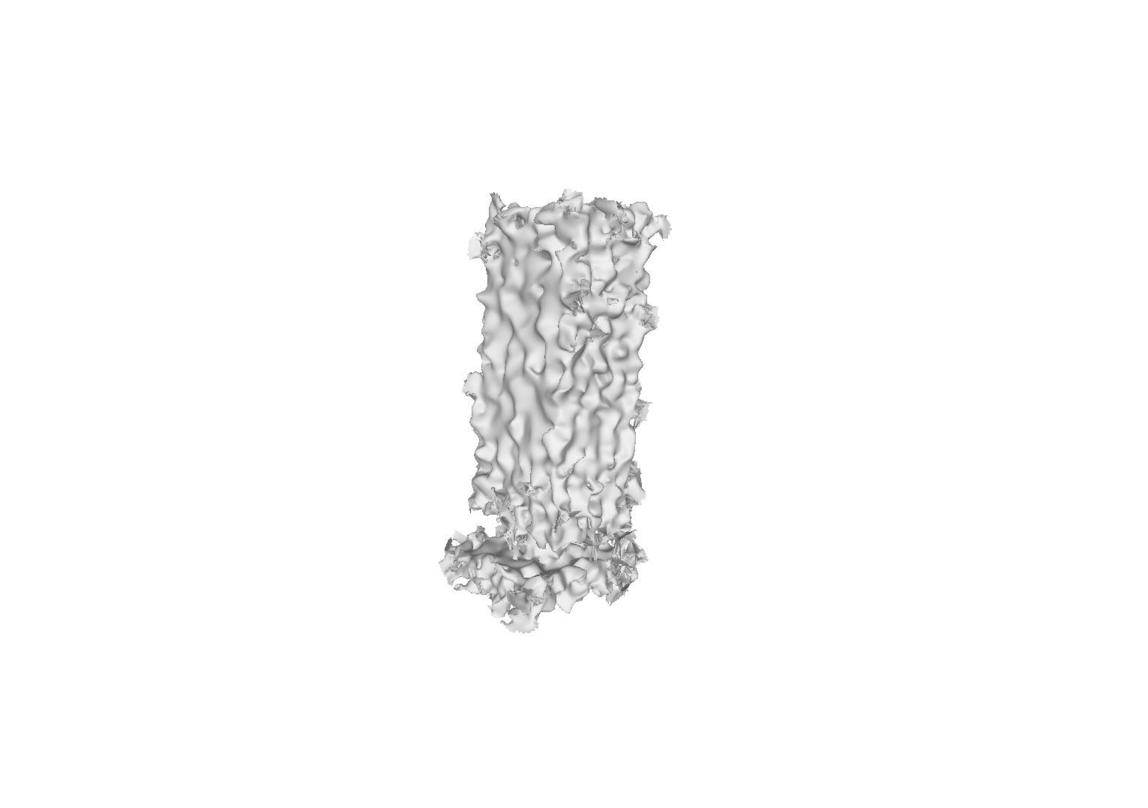}}
\subfloat{\includegraphics[width=\fitscale\tgtwidth, trim={450 152 450 152}, clip]{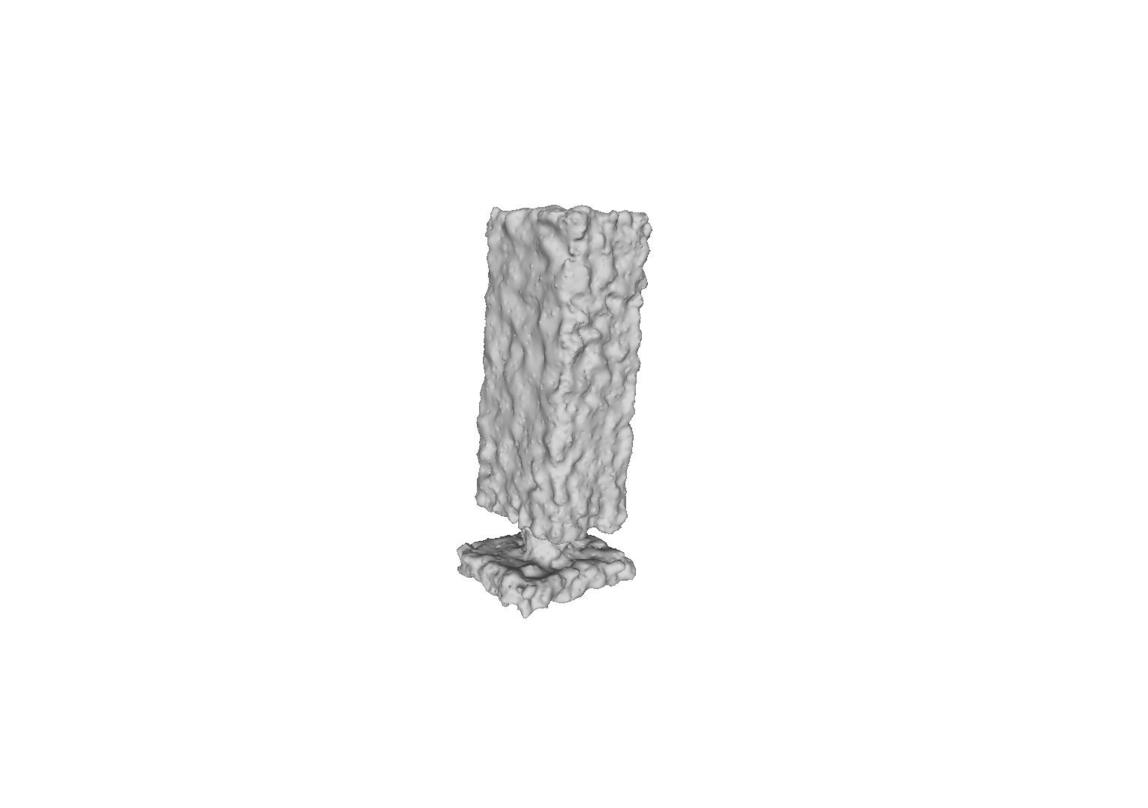}}
\subfloat{\includegraphics[width=\fitscale\tgtwidth, trim={450 152 450 152}, clip]{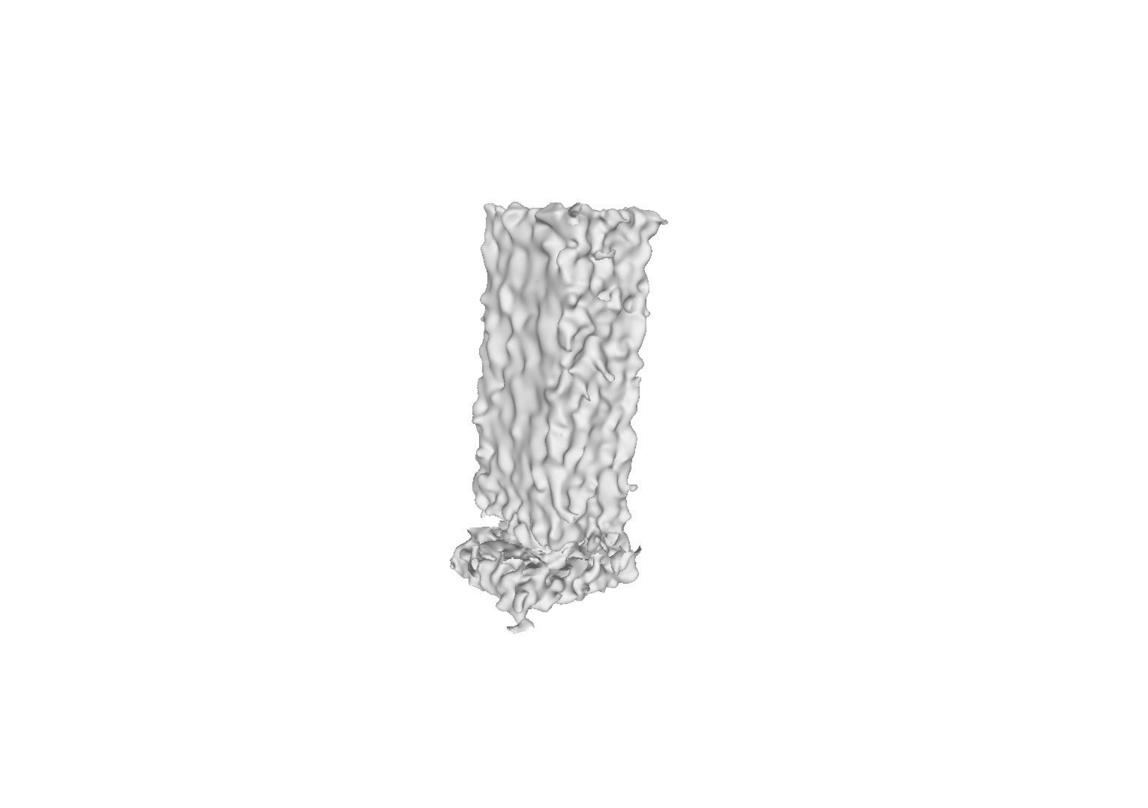}}
\subfloat{\includegraphics[width=\fitscale\tgtwidth, trim={450 152 450 152}, clip]{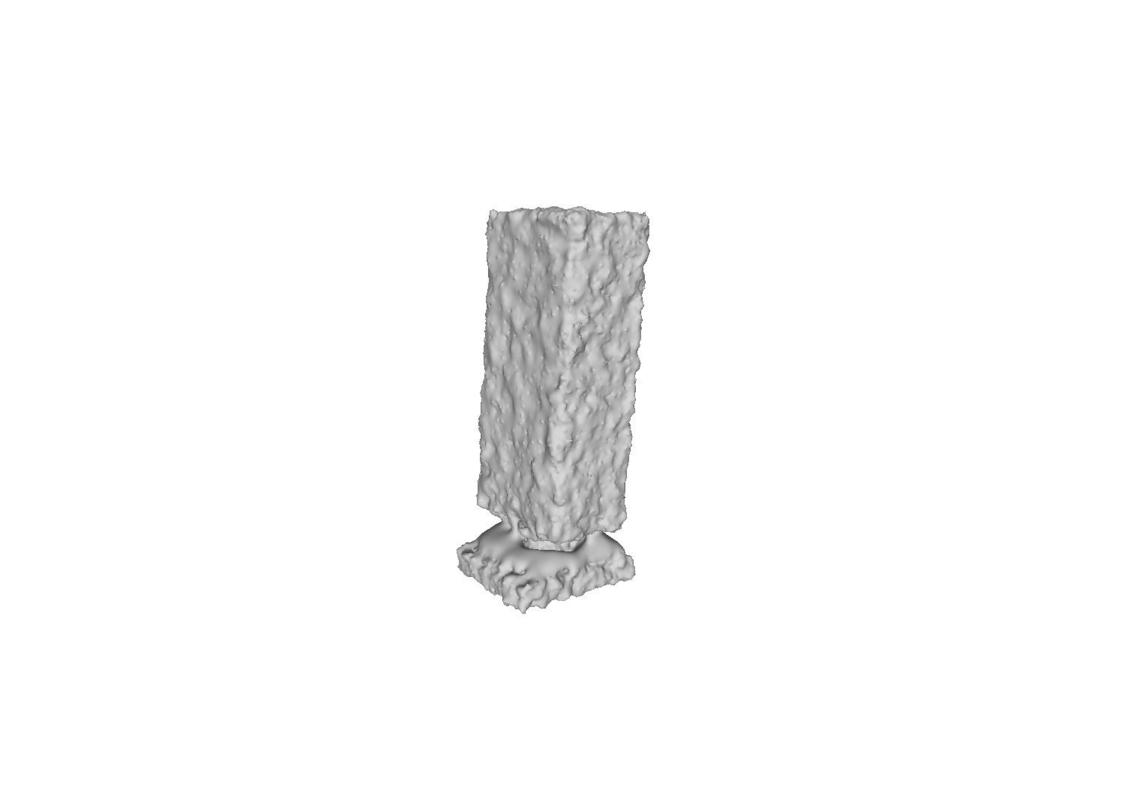}}
\subfloat{\includegraphics[width=\fitscale\tgtwidth, trim={450 152 450 152}, clip]{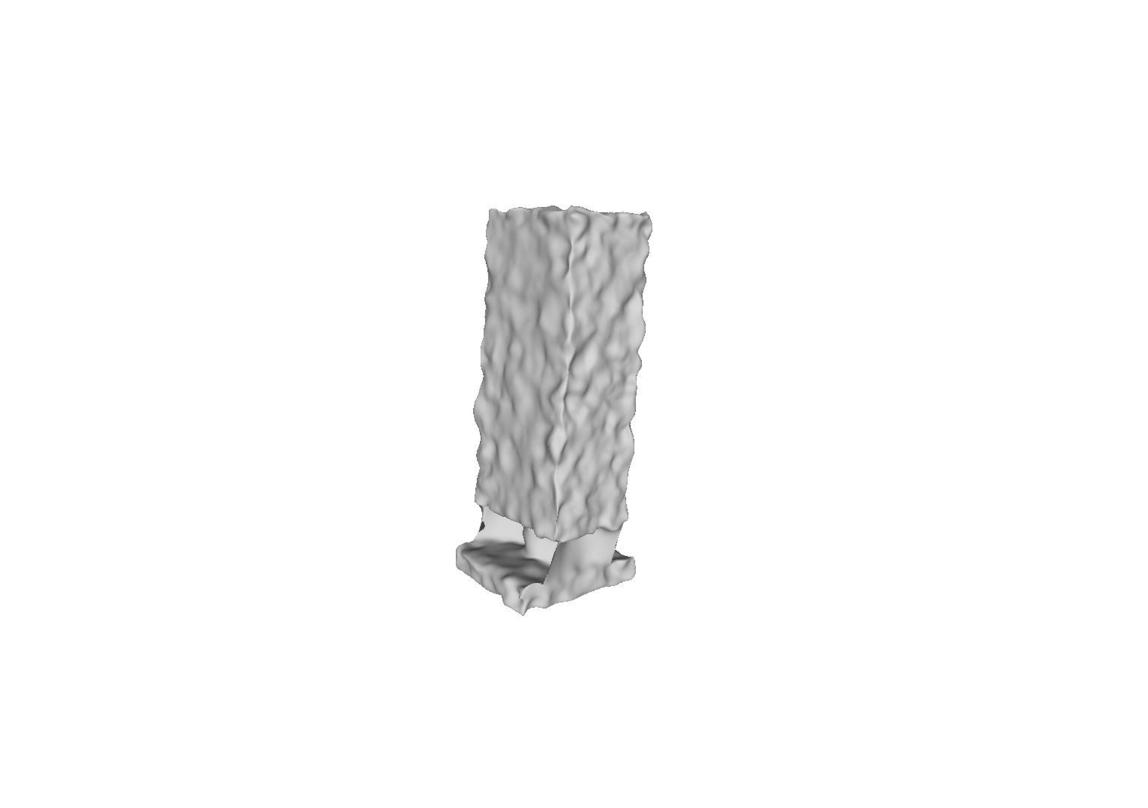}}
\subfloat{\includegraphics[width=\fitscale\tgtwidth, trim={450 152 450 152}, clip]{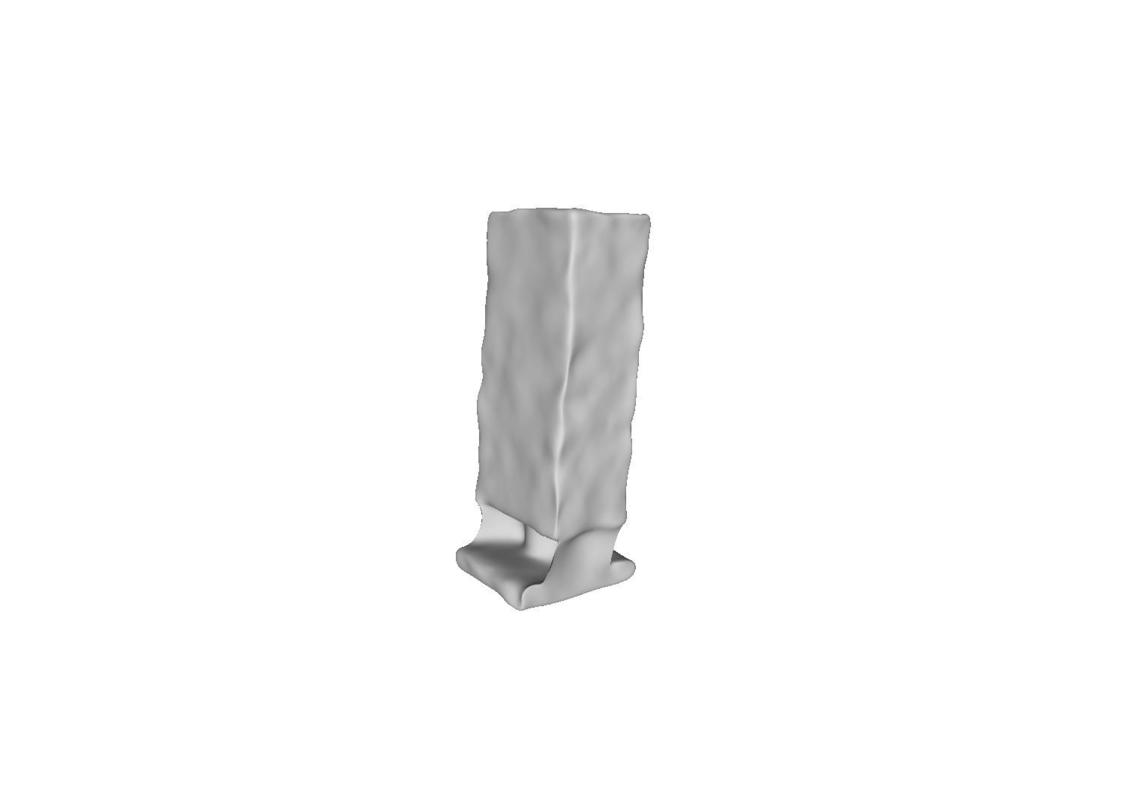}}
\subfloat{\includegraphics[width=\fitscale\tgtwidth, trim={450 152 450 152}, clip]{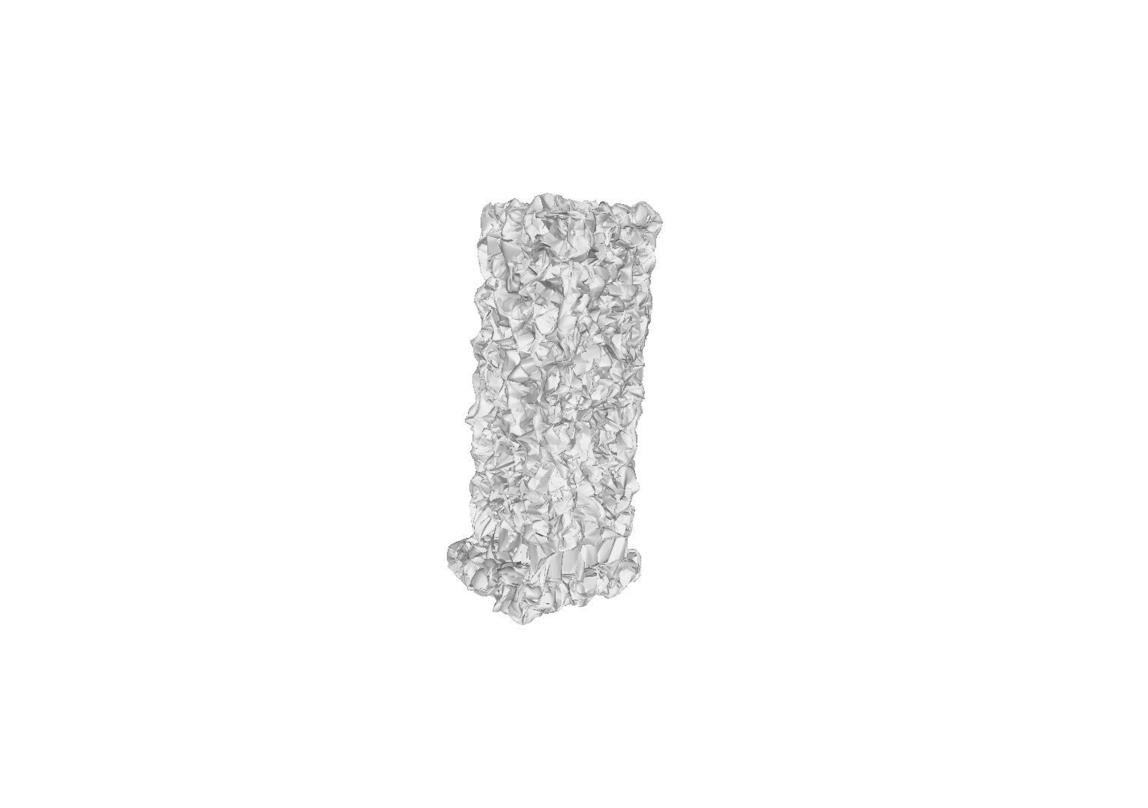}}
\subfloat{\includegraphics[width=\fitscale\tgtwidth, trim={450 152 450 152}, clip]{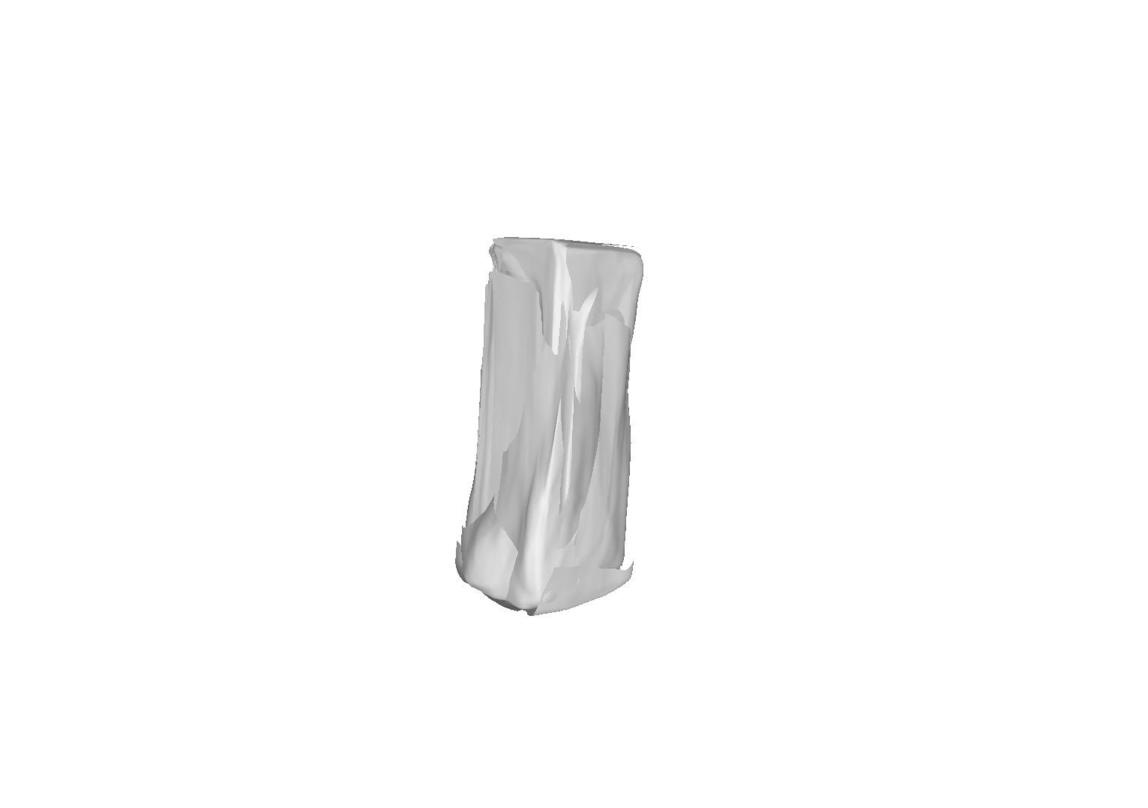}}
\subfloat{\includegraphics[width=\fitscale\tgtwidth, trim={450 152 450 152}, clip]{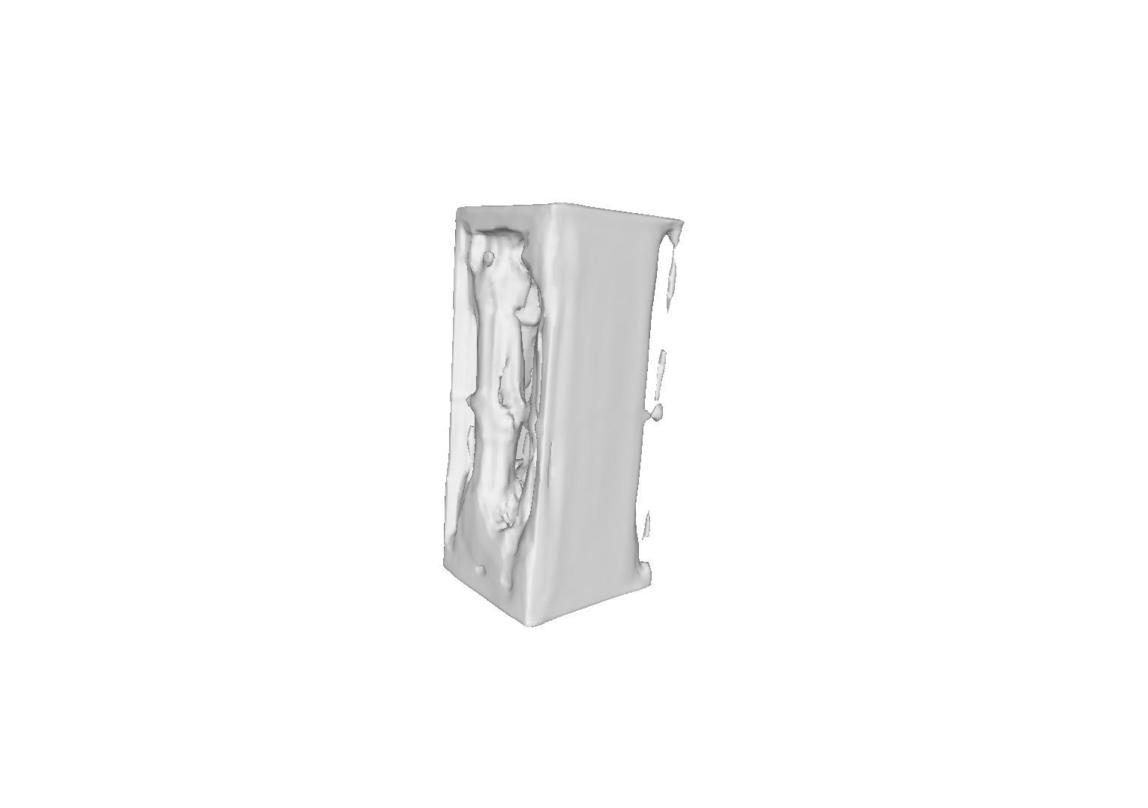}}
~
\subfloat{\includegraphics[width=\fitscale\tgtwidth, trim={450 152 450 152}, clip]{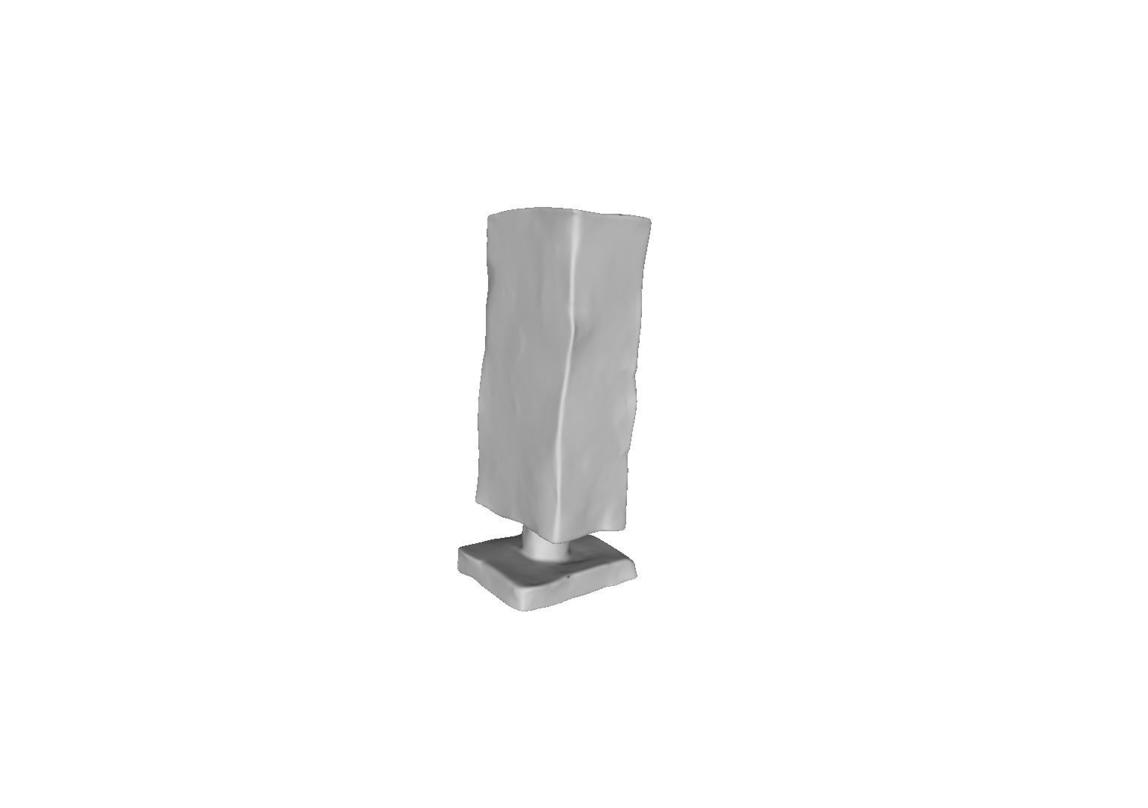}}
\\
\vspace{-5mm}
\subfloat{\includegraphics[width=\fitscale\tgtwidth, trim={450 152 450 152}, clip]{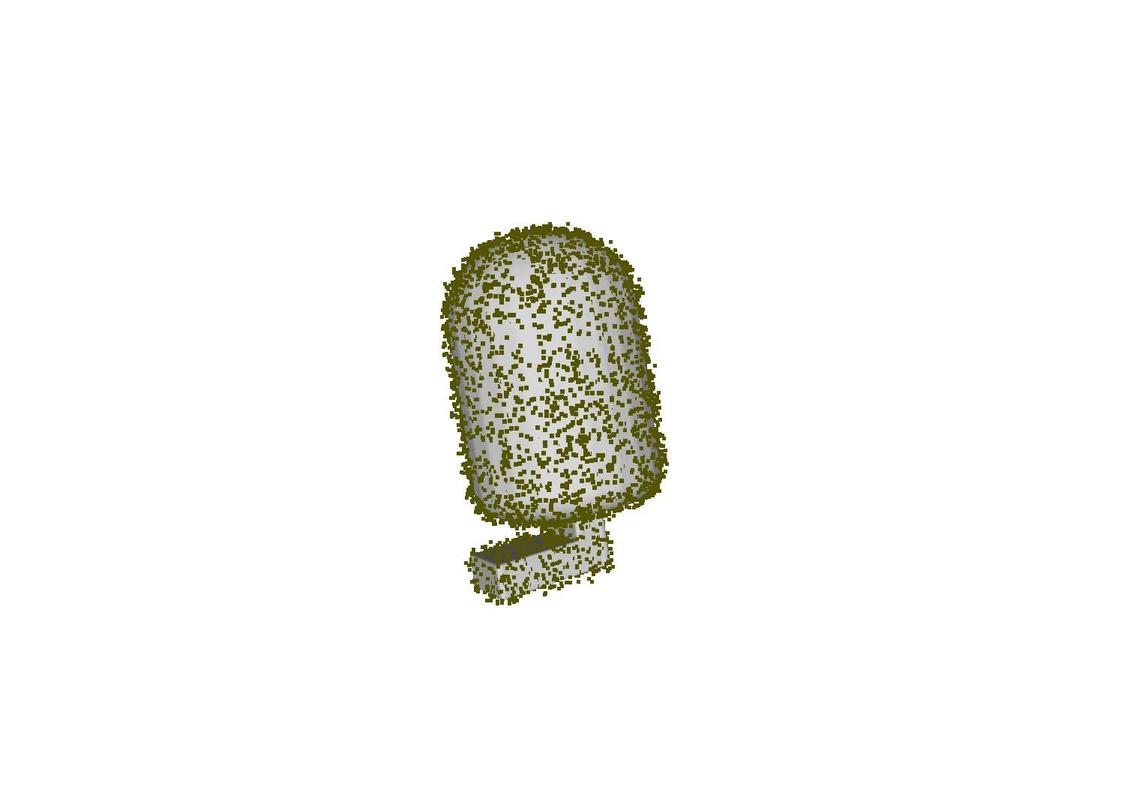}}
\subfloat{\includegraphics[width=\fitscale\tgtwidth, trim={450 152 450 152}, clip]{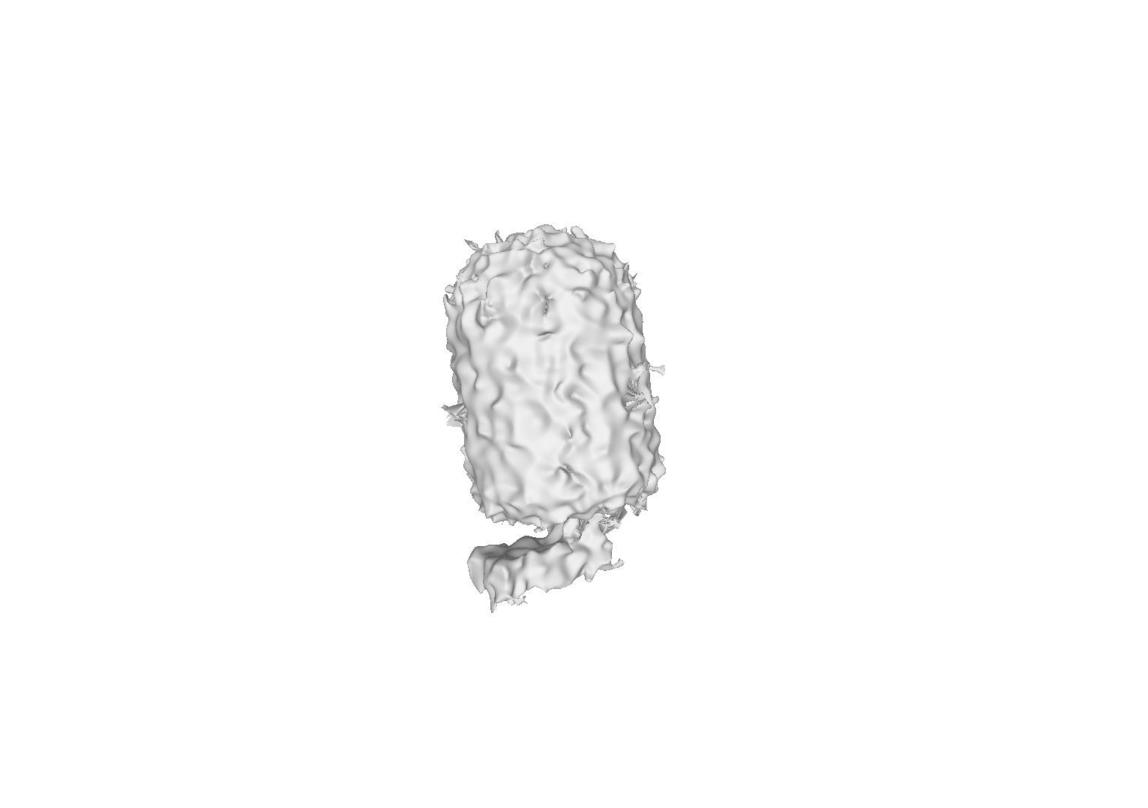}}
\subfloat{\includegraphics[width=\fitscale\tgtwidth, trim={450 152 450 152}, clip]{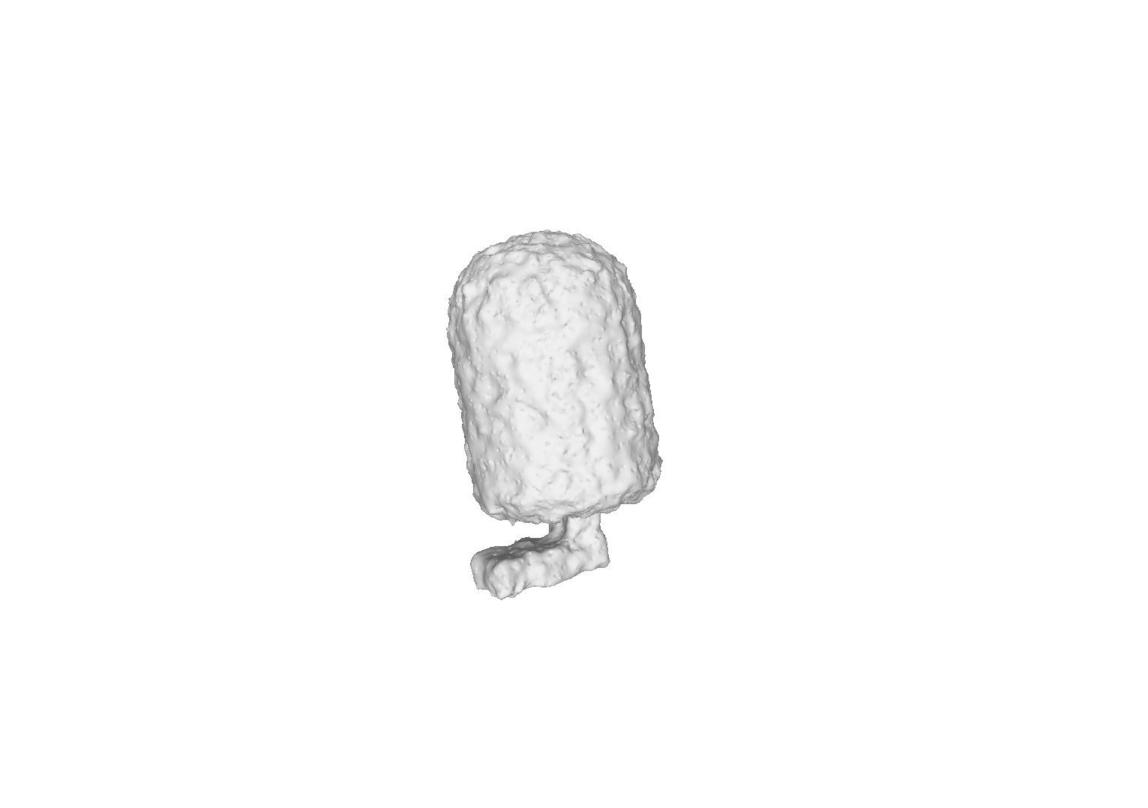}}
\subfloat{\includegraphics[width=\fitscale\tgtwidth, trim={450 152 450 152}, clip]{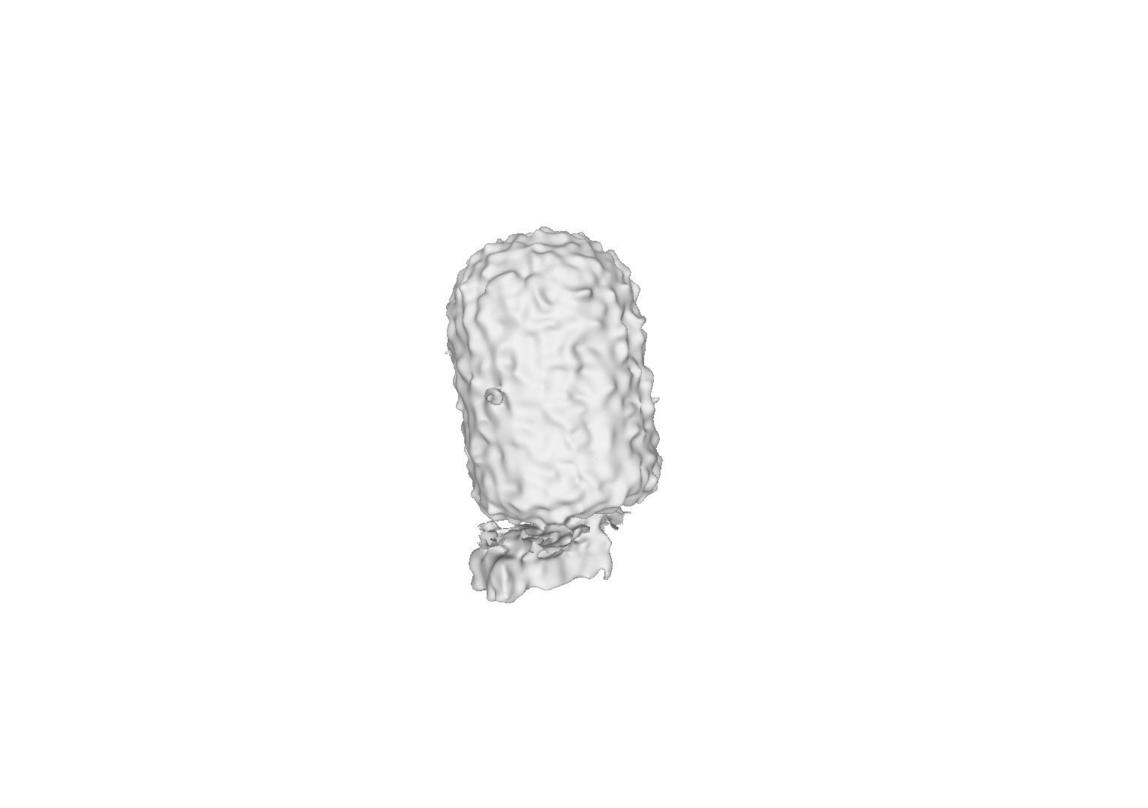}}
\subfloat{\includegraphics[width=\fitscale\tgtwidth, trim={450 152 450 152}, clip]{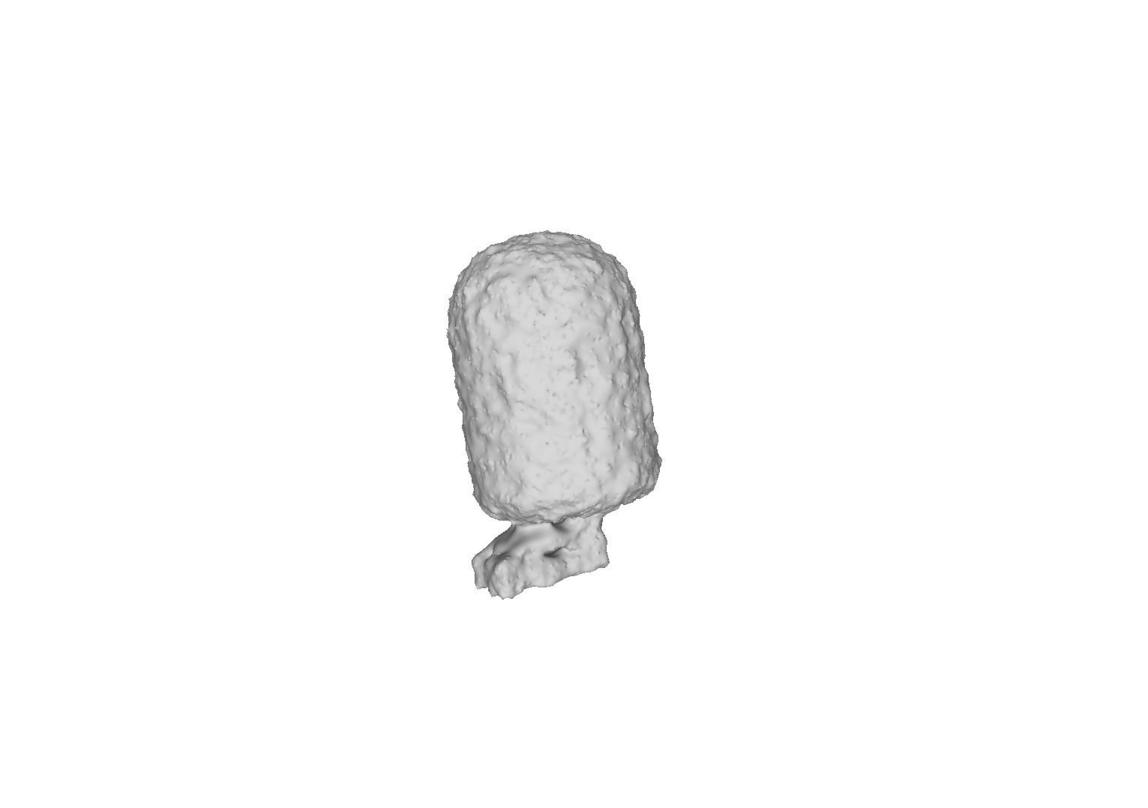}}
\subfloat{\includegraphics[width=\fitscale\tgtwidth, trim={450 152 450 152}, clip]{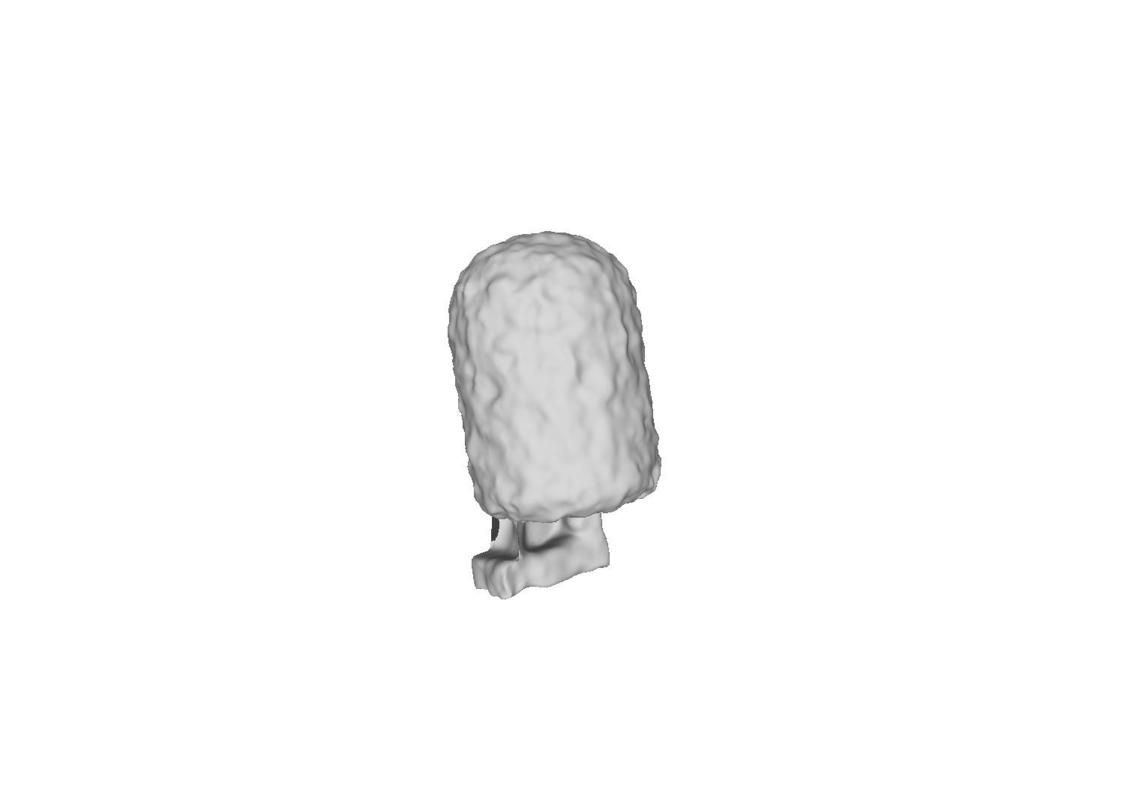}}
\subfloat{\includegraphics[width=\fitscale\tgtwidth, trim={450 152 450 152}, clip]{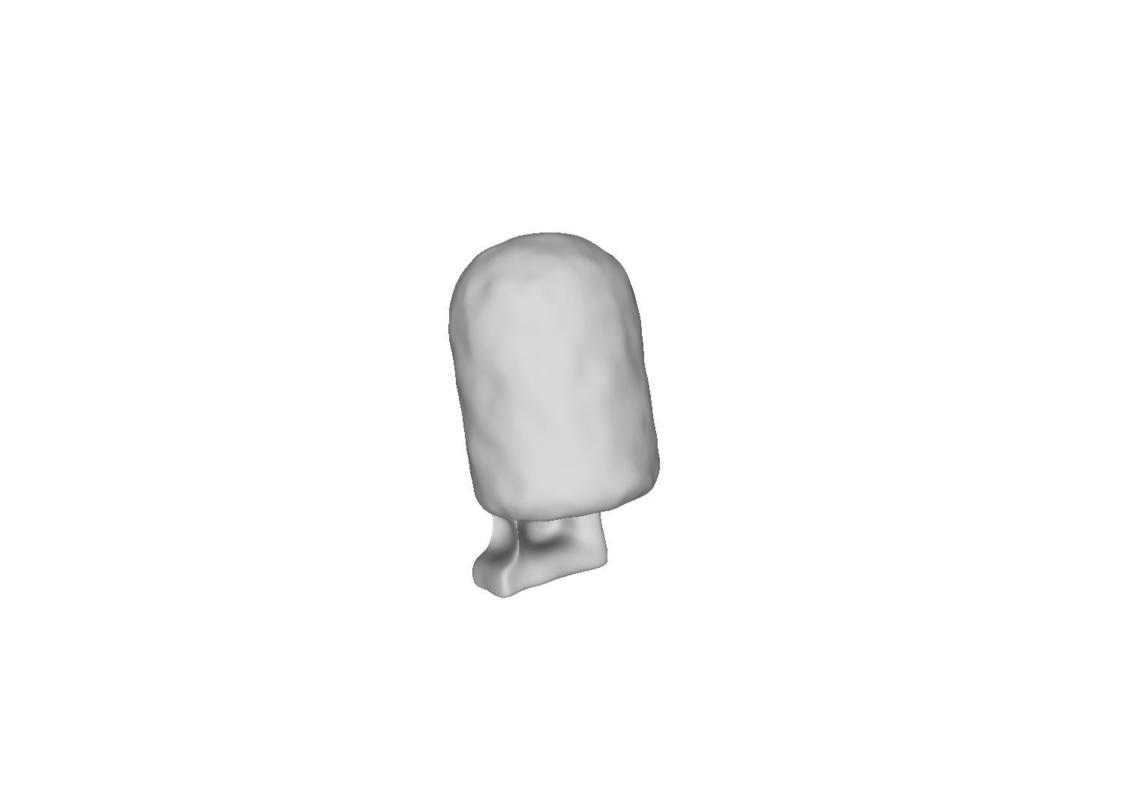}}
\subfloat{\includegraphics[width=\fitscale\tgtwidth, trim={450 152 450 152}, clip]{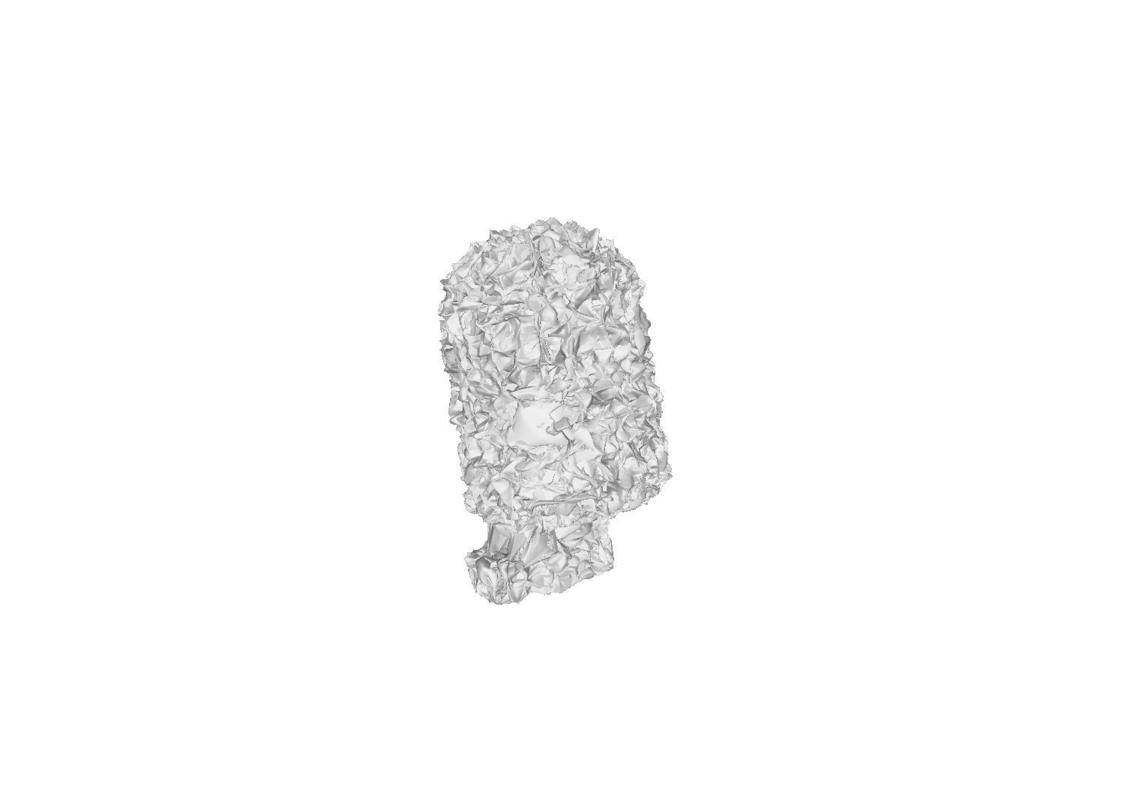}}
\subfloat{\includegraphics[width=\fitscale\tgtwidth, trim={450 152 450 152}, clip]{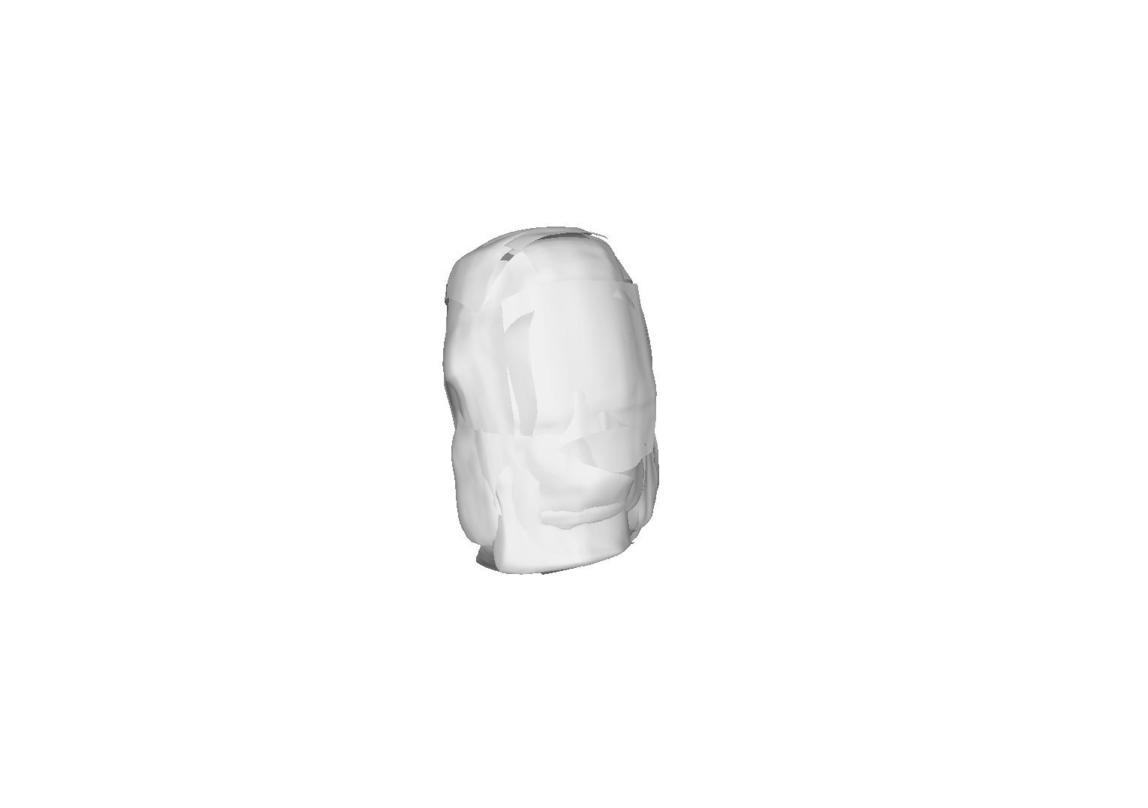}}
\subfloat{\includegraphics[width=\fitscale\tgtwidth, trim={450 152 450 152}, clip]{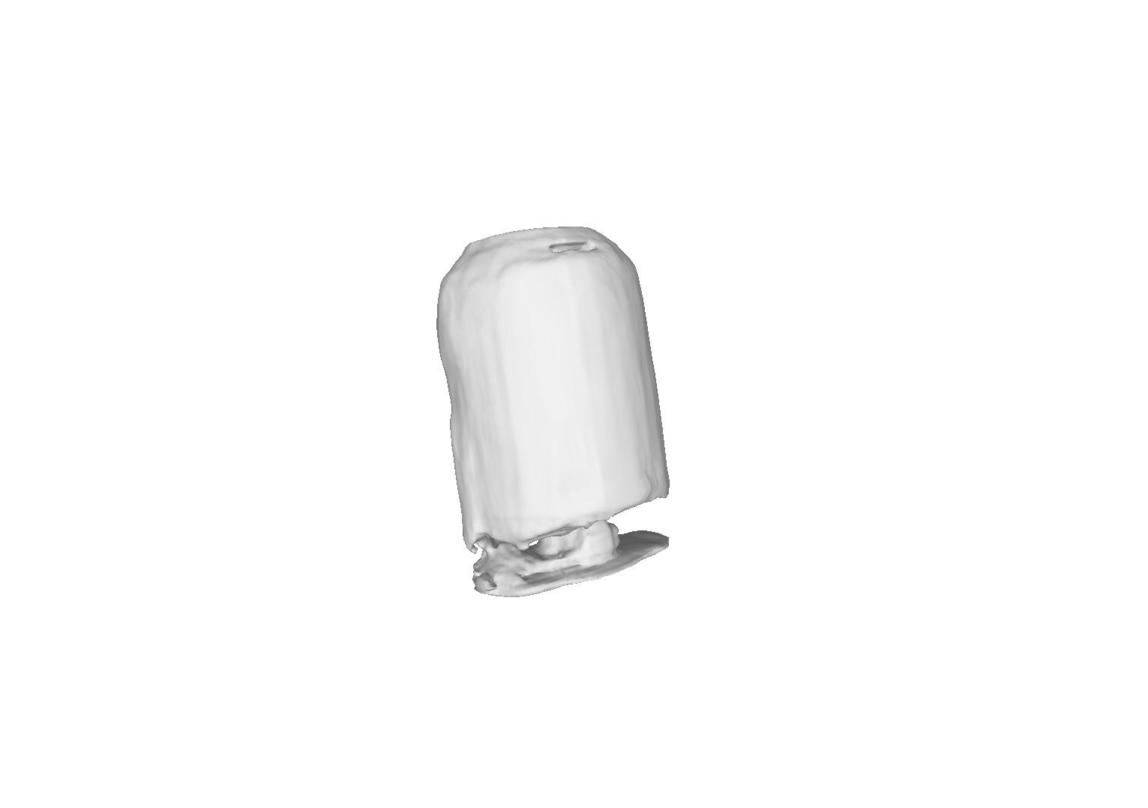}}
~
\subfloat{\includegraphics[width=\fitscale\tgtwidth, trim={450 152 450 152}, clip]{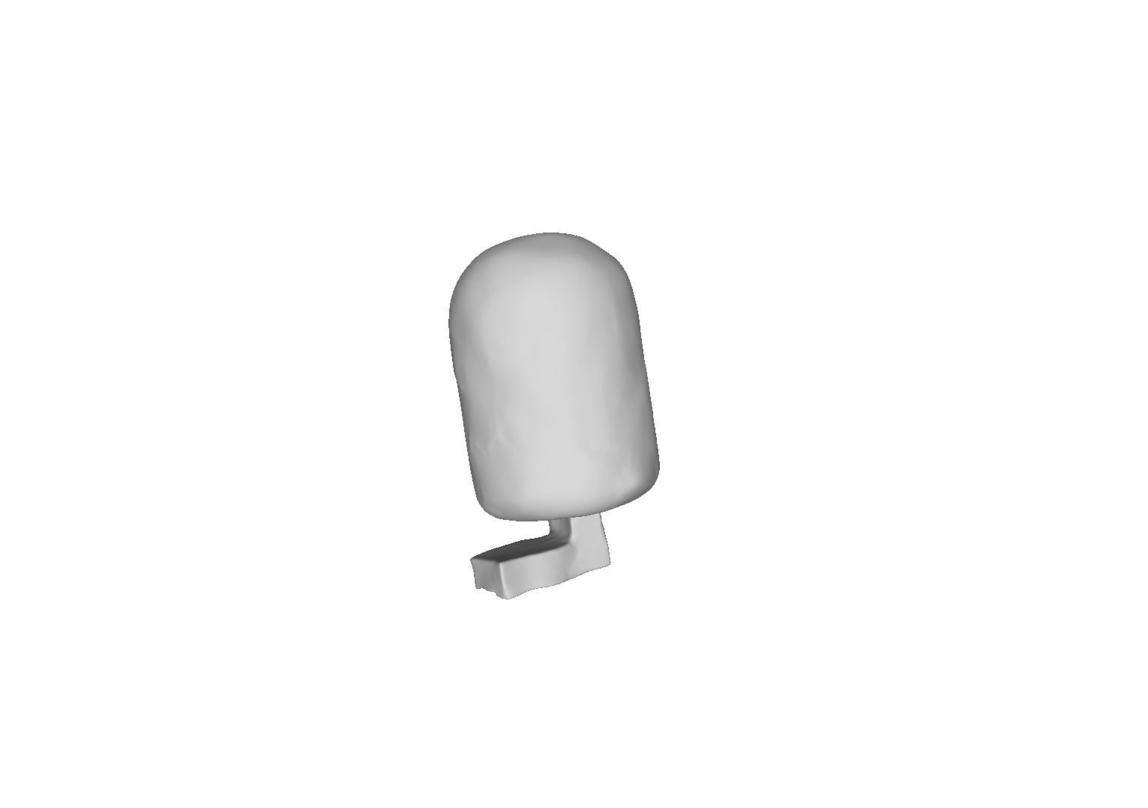}}
\\
\vspace{-5mm}
\subfloat{\includegraphics[width=\fitscale\tgtwidth, trim={450 152 450 152}, clip]{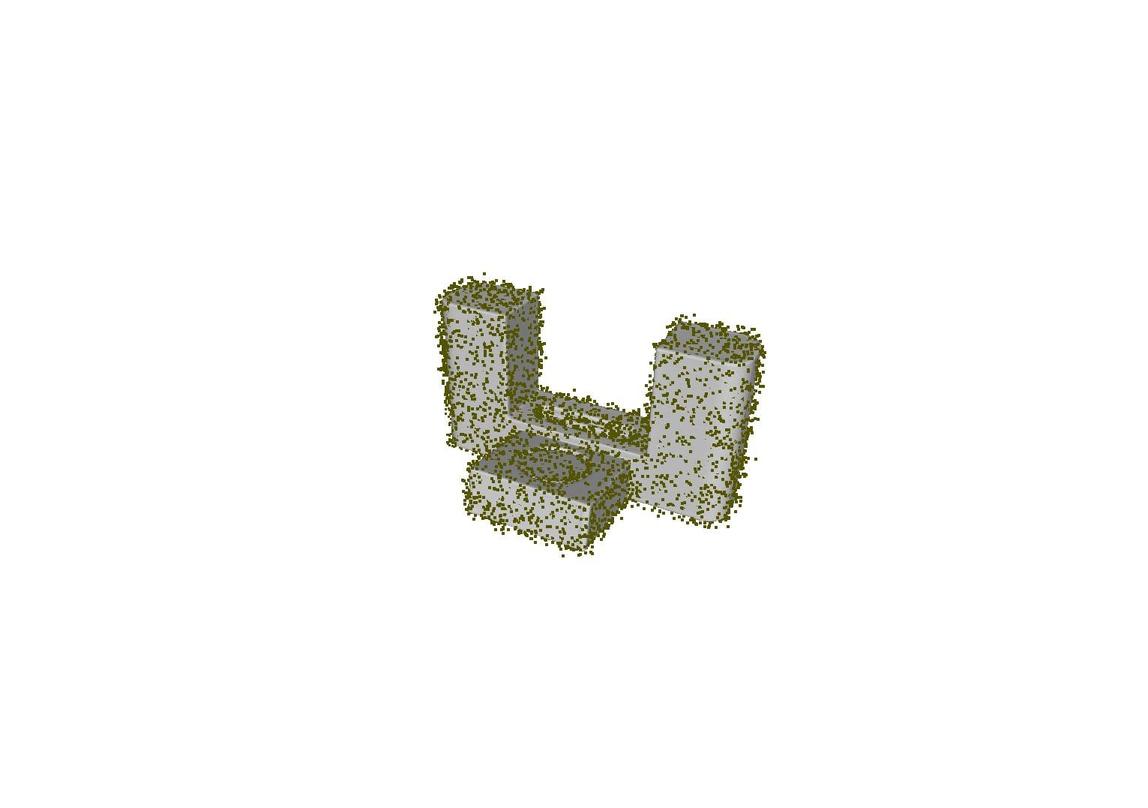}}
\subfloat{\includegraphics[width=\fitscale\tgtwidth, trim={450 152 450 152}, clip]{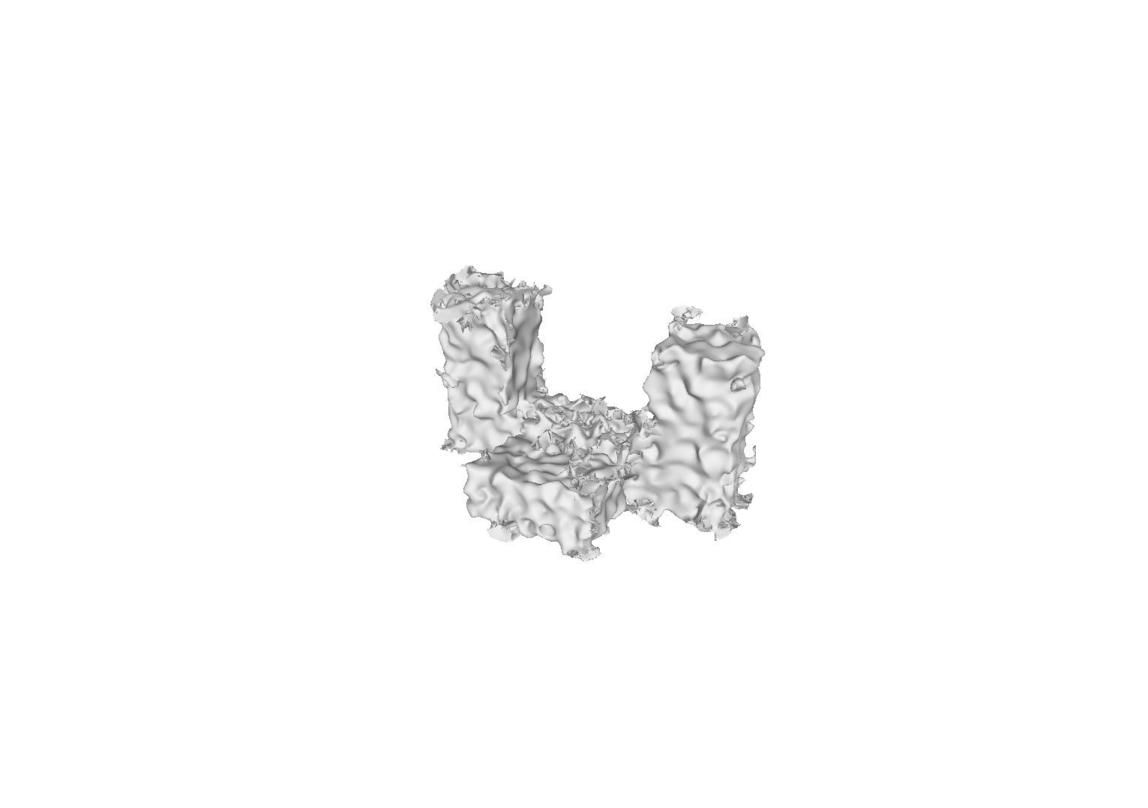}}
\subfloat{\includegraphics[width=\fitscale\tgtwidth, trim={450 152 450 152}, clip]{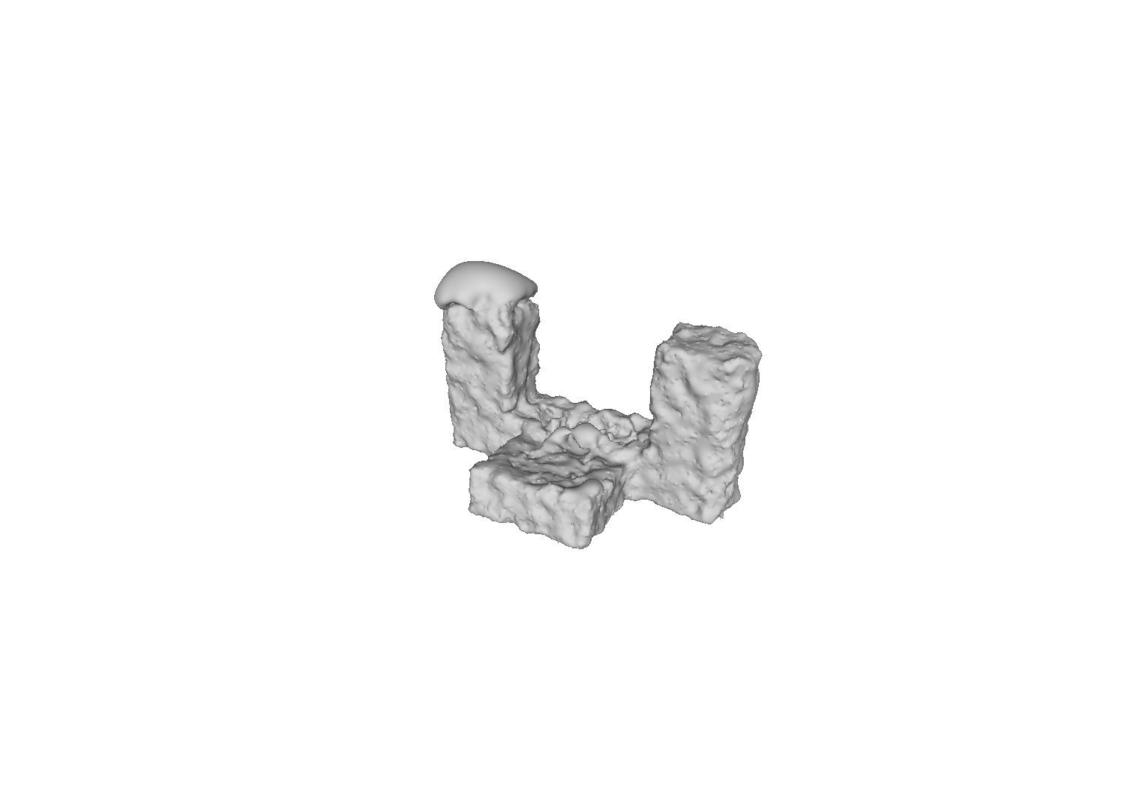}}
\subfloat{\includegraphics[width=\fitscale\tgtwidth, trim={450 152 450 152}, clip]{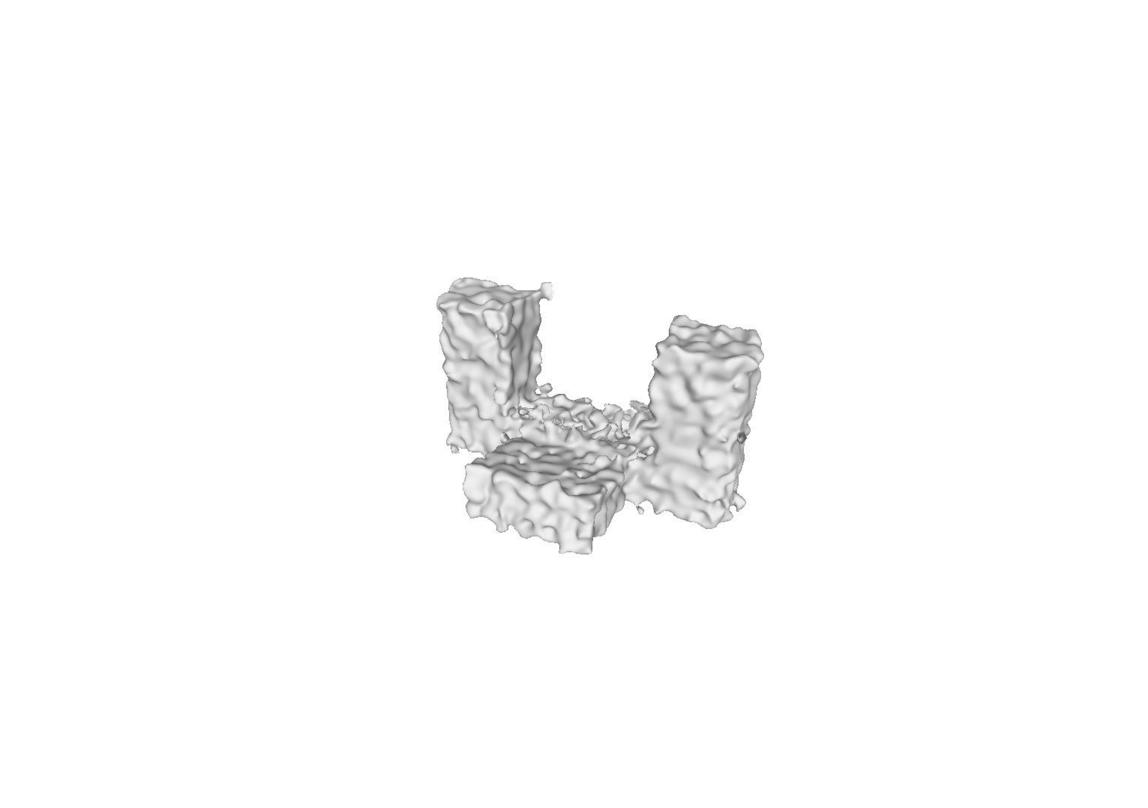}}
\subfloat{\includegraphics[width=\fitscale\tgtwidth, trim={450 152 450 152}, clip]{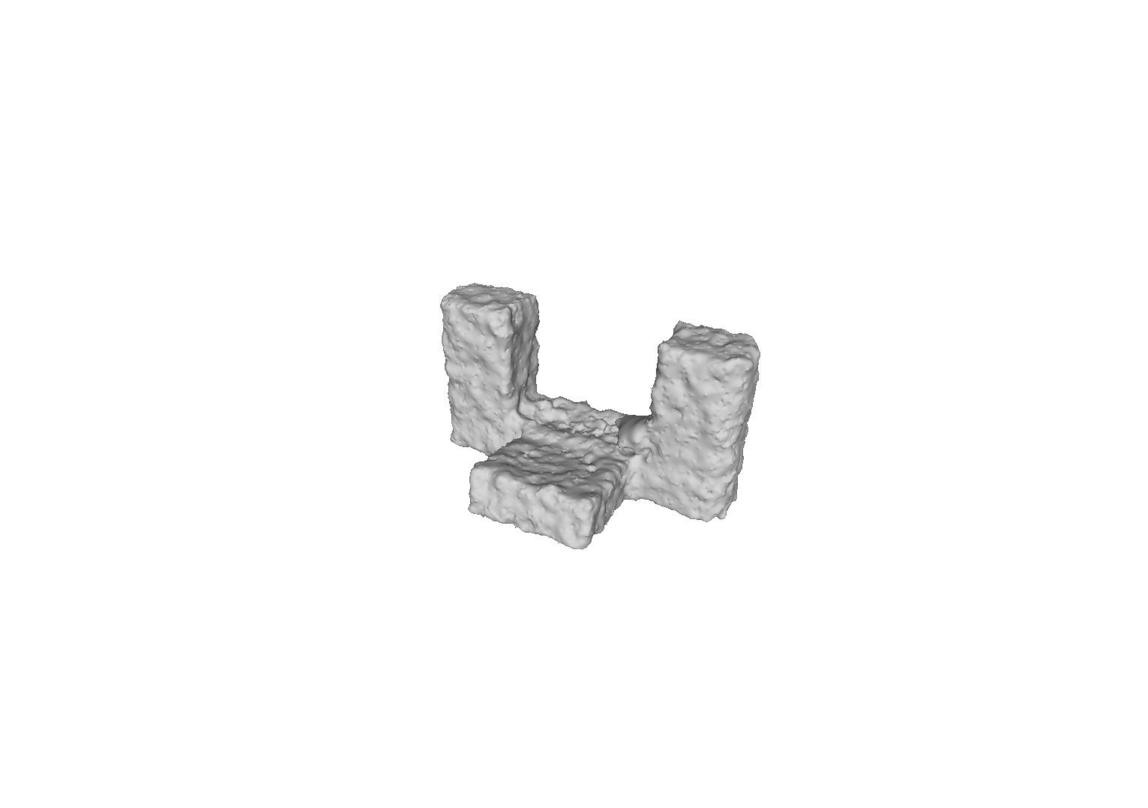}}
\subfloat{\includegraphics[width=\fitscale\tgtwidth, trim={450 152 450 152}, clip]{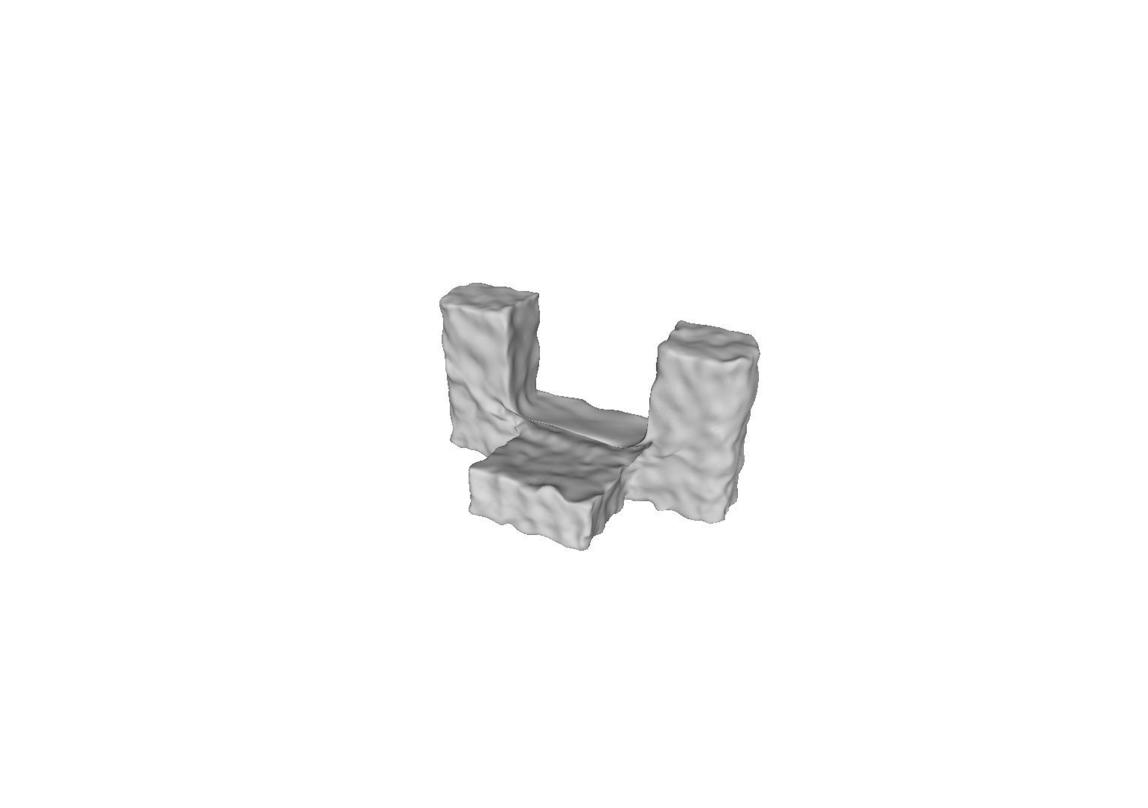}}
\subfloat{\includegraphics[width=\fitscale\tgtwidth, trim={450 152 450 152}, clip]{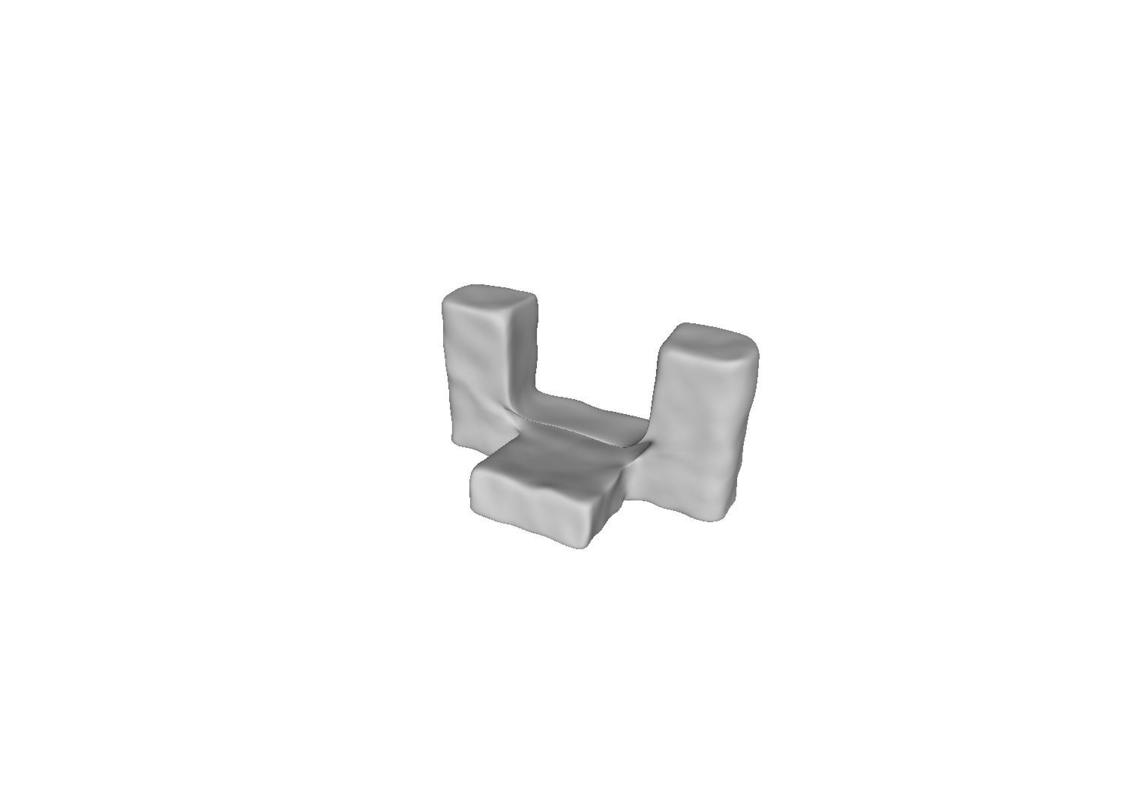}}
\subfloat{\includegraphics[width=\fitscale\tgtwidth, trim={450 152 450 152}, clip]{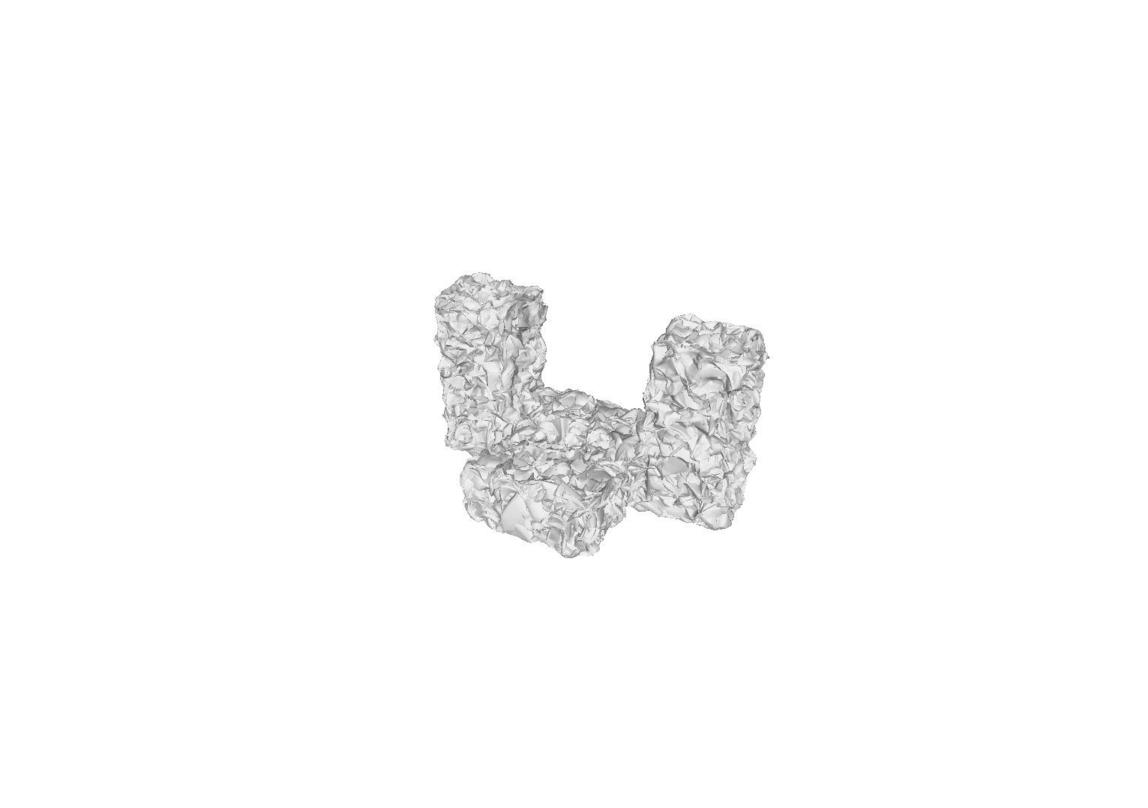}}
\subfloat{\includegraphics[width=\fitscale\tgtwidth, trim={450 152 450 152}, clip]{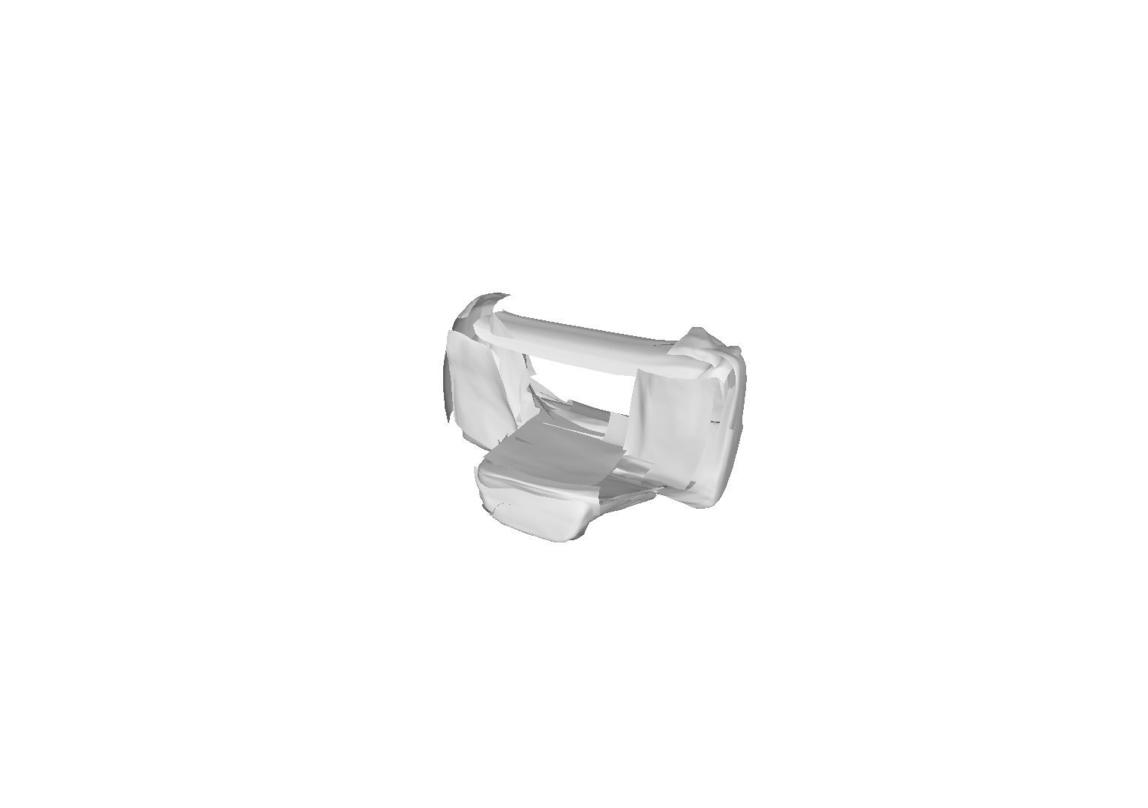}}
\subfloat{\includegraphics[width=\fitscale\tgtwidth, trim={450 152 450 152}, clip]{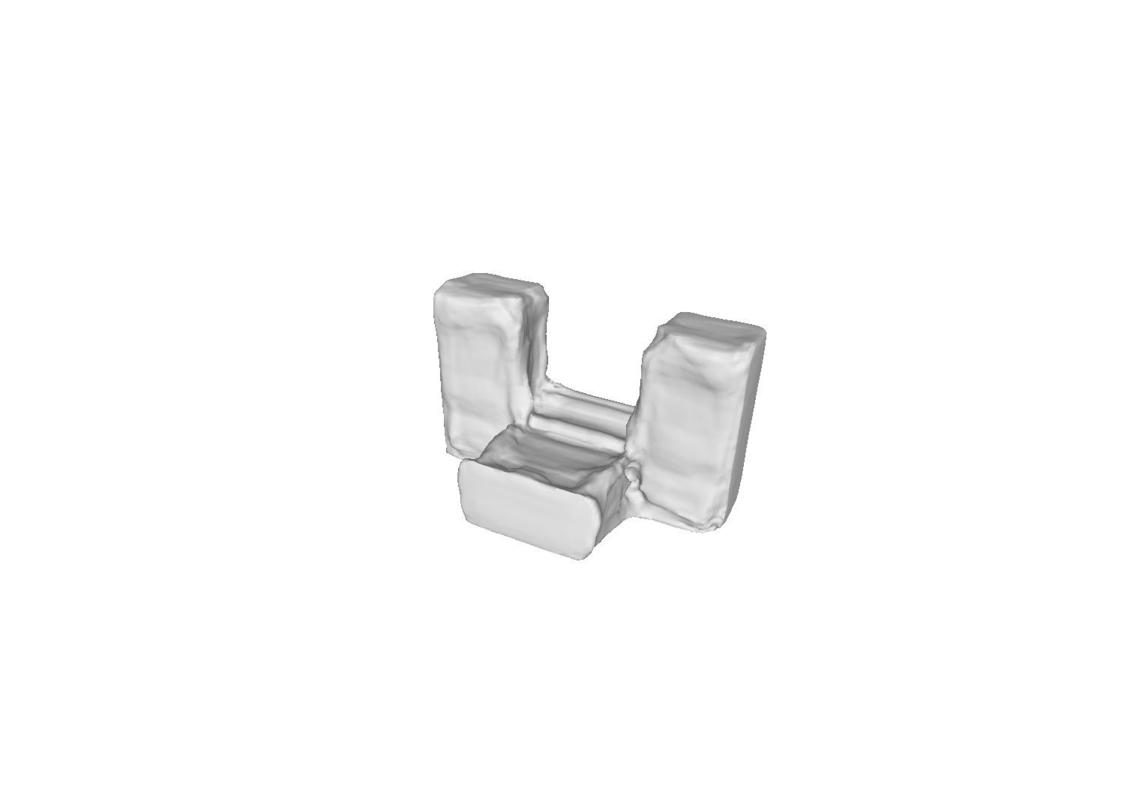}}
~
\subfloat{\includegraphics[width=\fitscale\tgtwidth, trim={450 152 450 152}, clip]{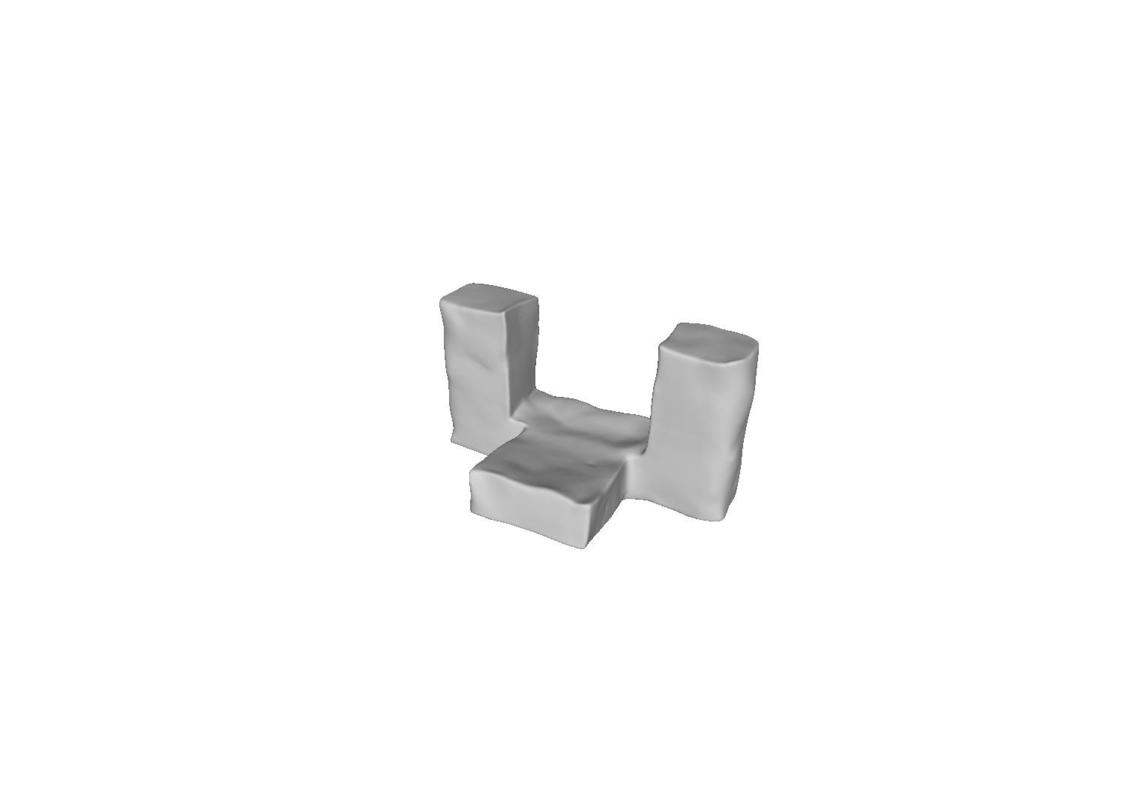}}
\\
\vspace{-5mm}
\subfloat{\includegraphics[width=\fitscale\tgtwidth, trim={450 152 450 152}, clip]{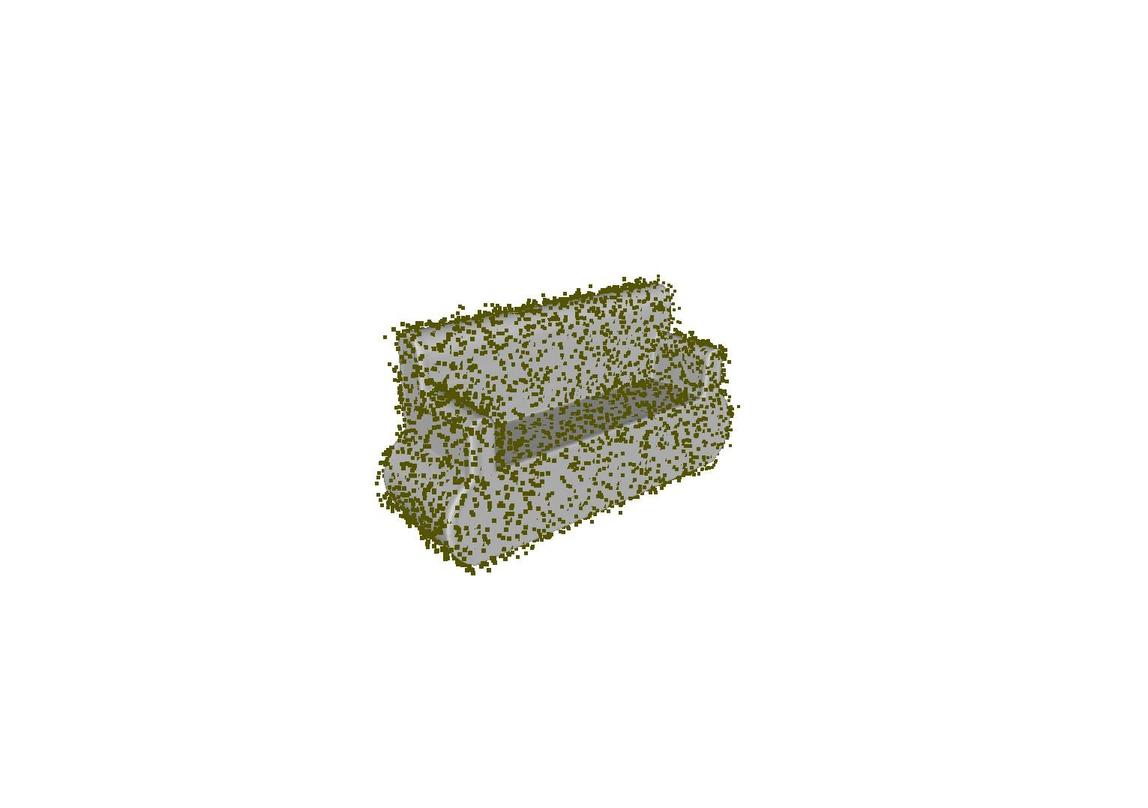}}
\subfloat{\includegraphics[width=\fitscale\tgtwidth, trim={450 152 450 152}, clip]{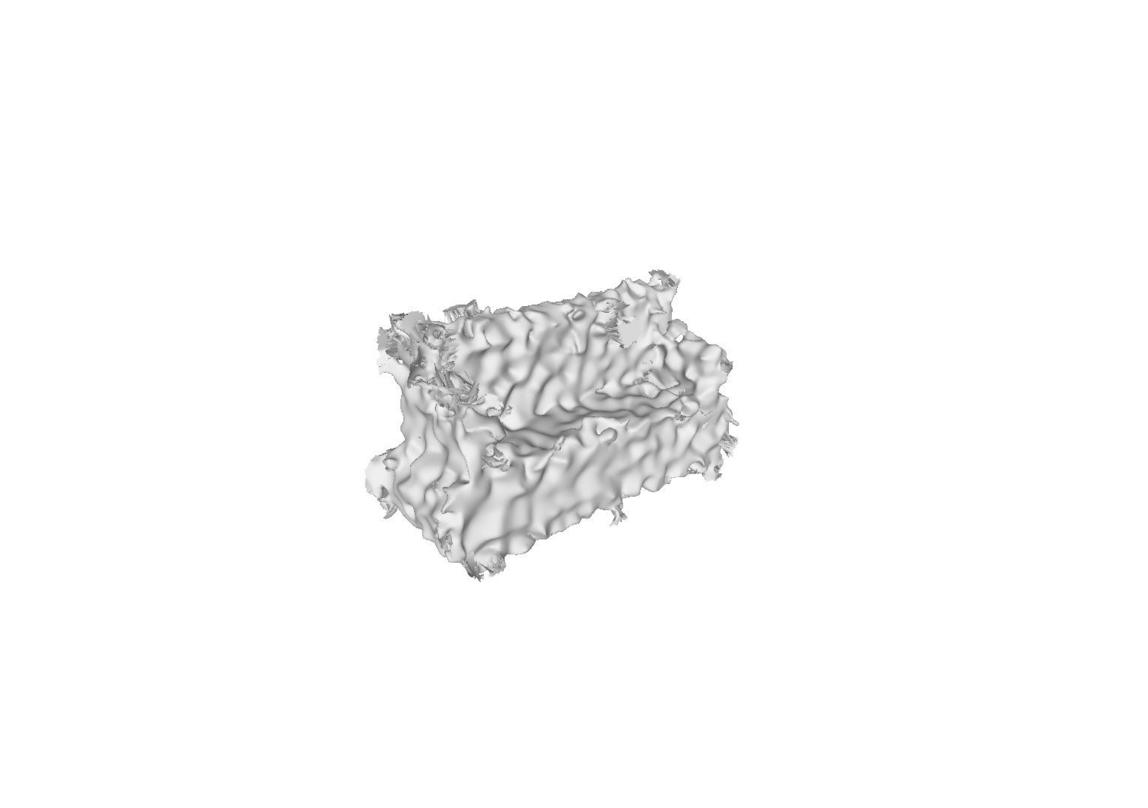}}
\subfloat{\includegraphics[width=\fitscale\tgtwidth, trim={450 152 450 152}, clip]{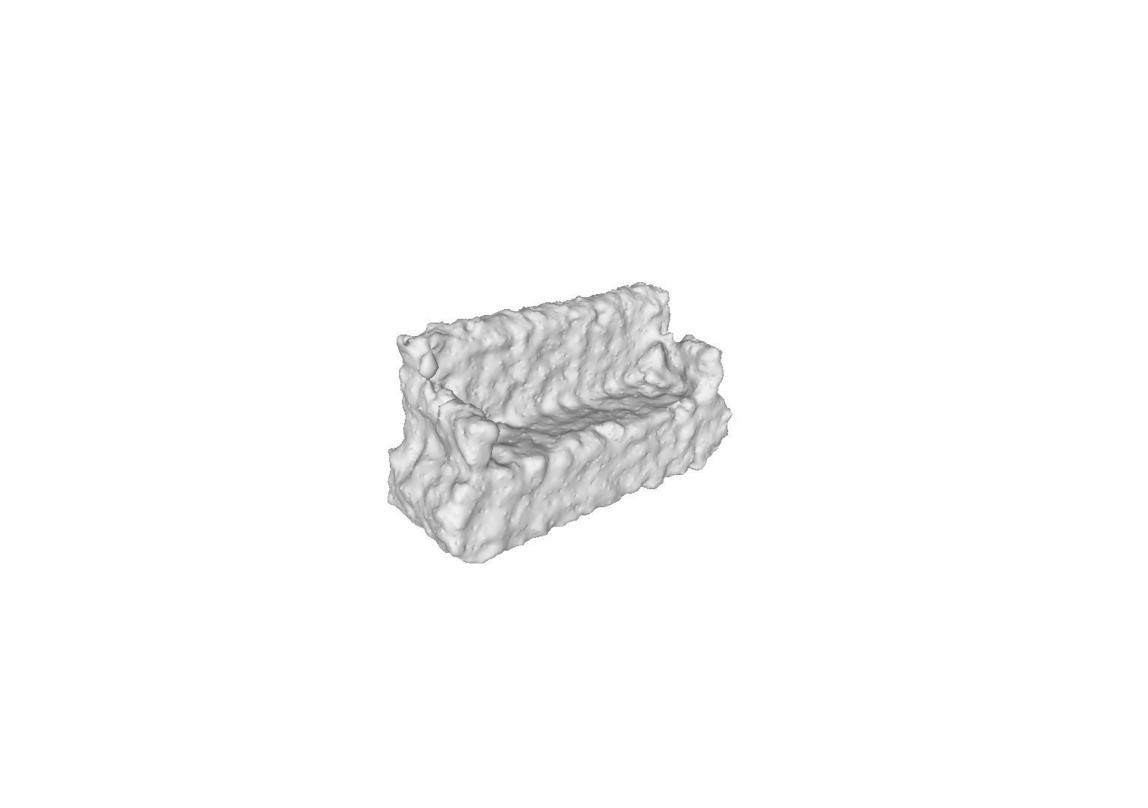}}
\subfloat{\includegraphics[width=\fitscale\tgtwidth, trim={450 152 450 152}, clip]{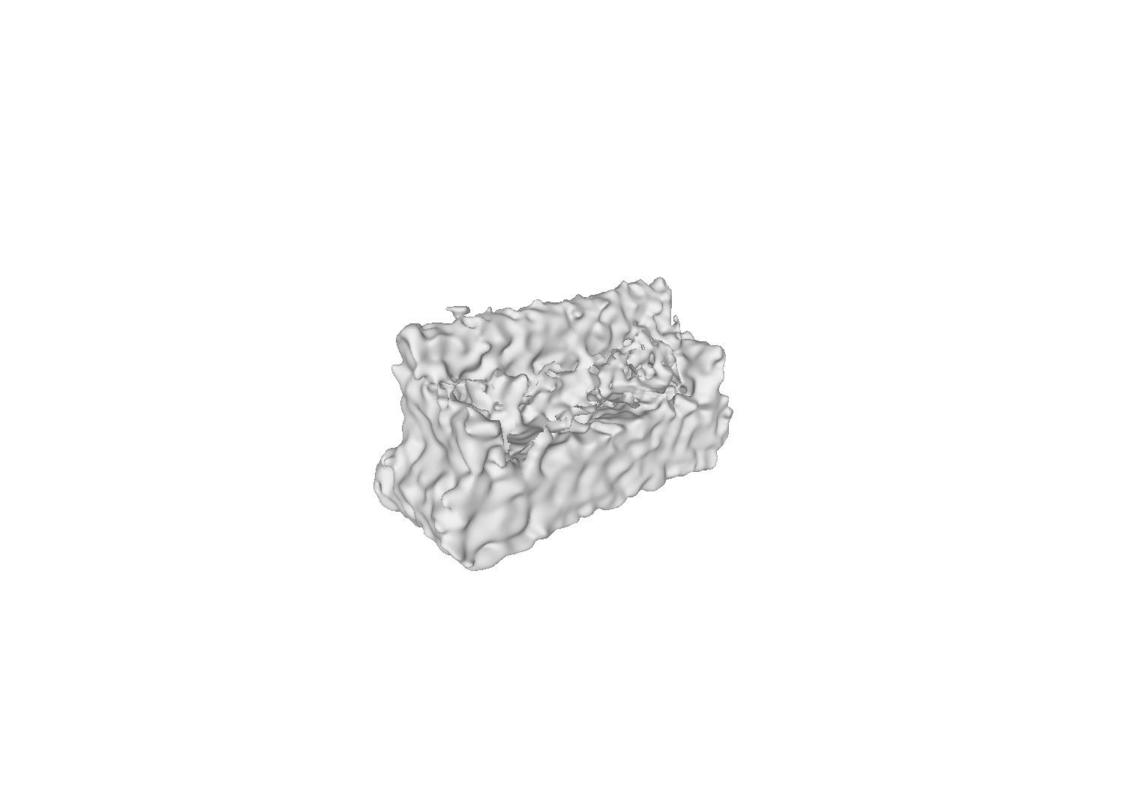}}
\subfloat{\includegraphics[width=\fitscale\tgtwidth, trim={450 152 450 152}, clip]{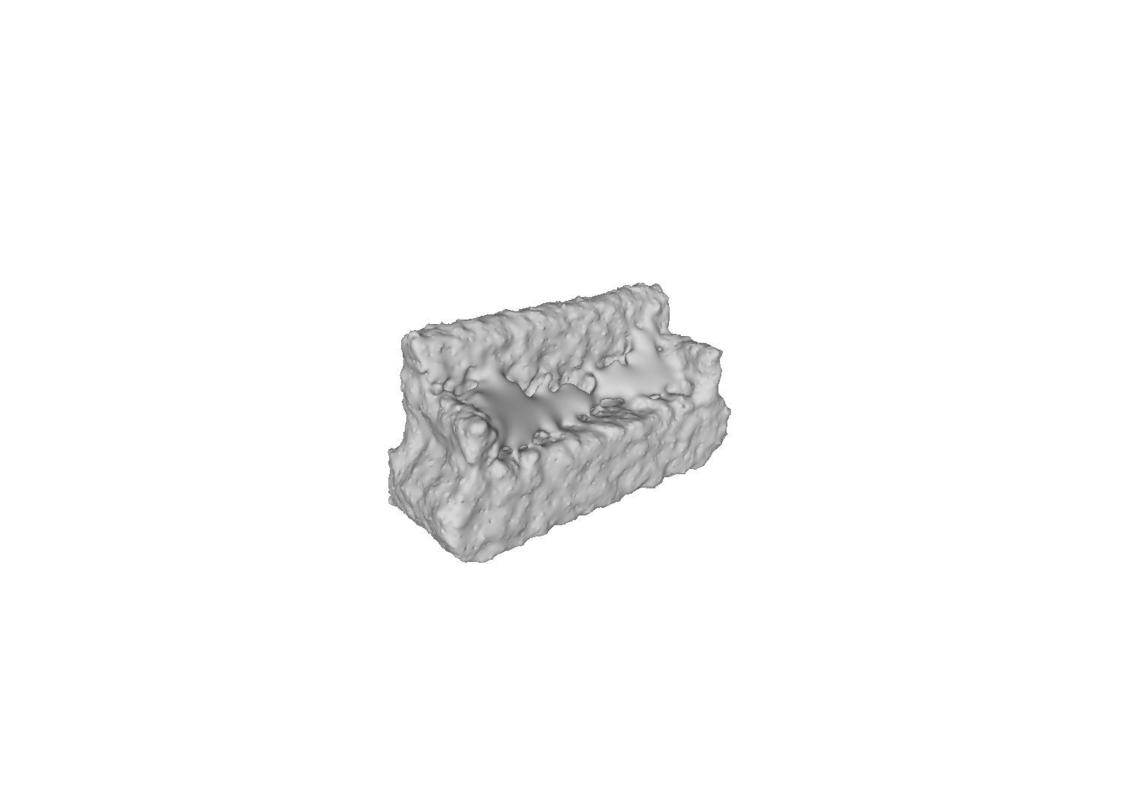}}
\subfloat{\includegraphics[width=\fitscale\tgtwidth, trim={450 152 450 152}, clip]{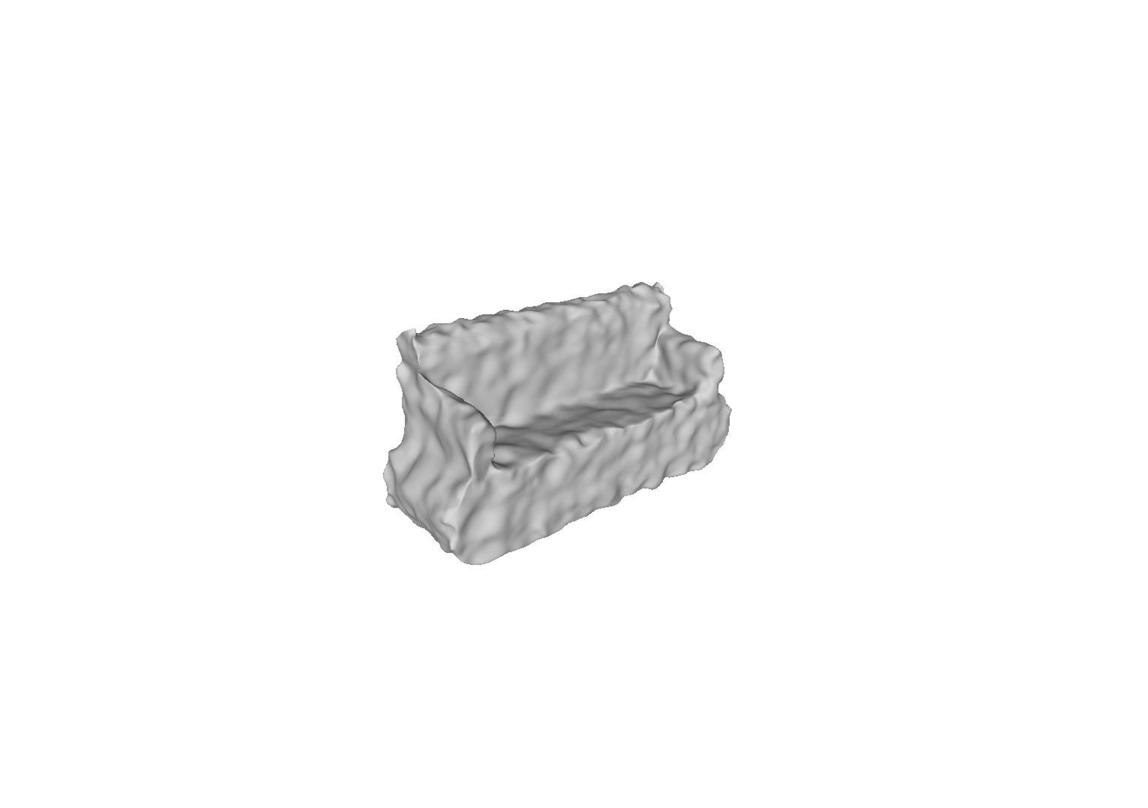}}
\subfloat{\includegraphics[width=\fitscale\tgtwidth, trim={450 152 450 152}, clip]{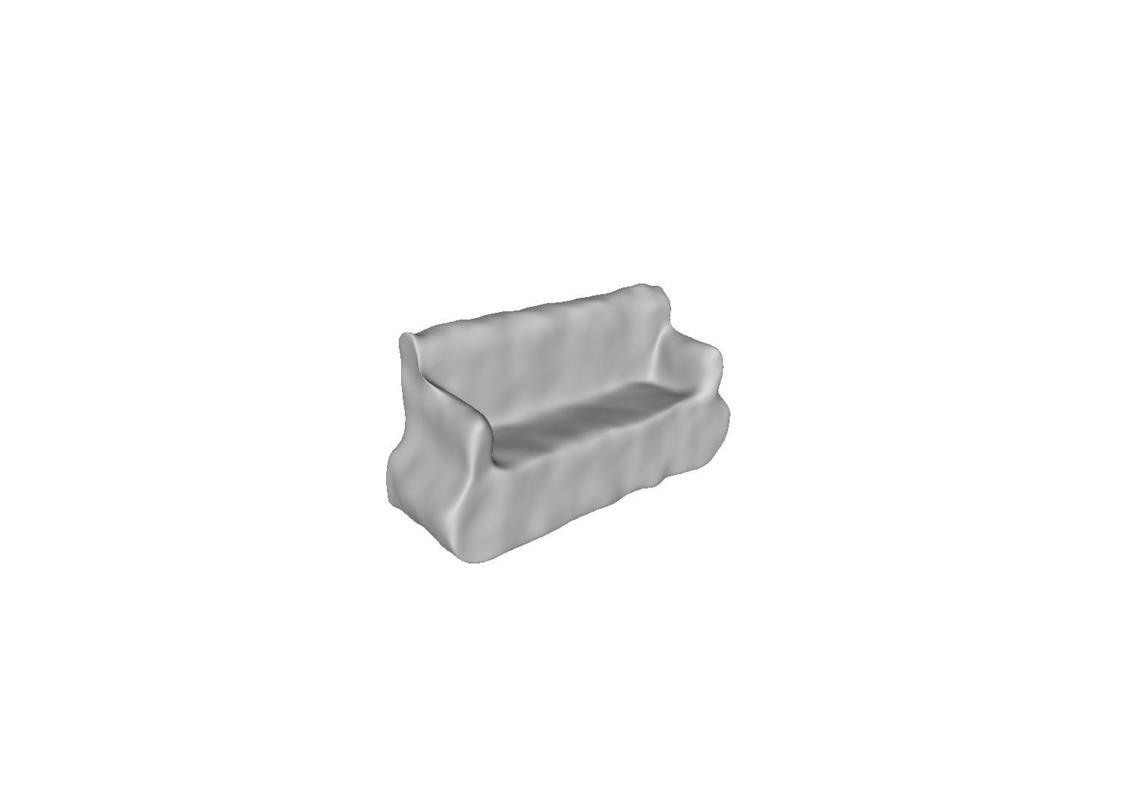}}
\subfloat{\includegraphics[width=\fitscale\tgtwidth, trim={450 152 450 152}, clip]{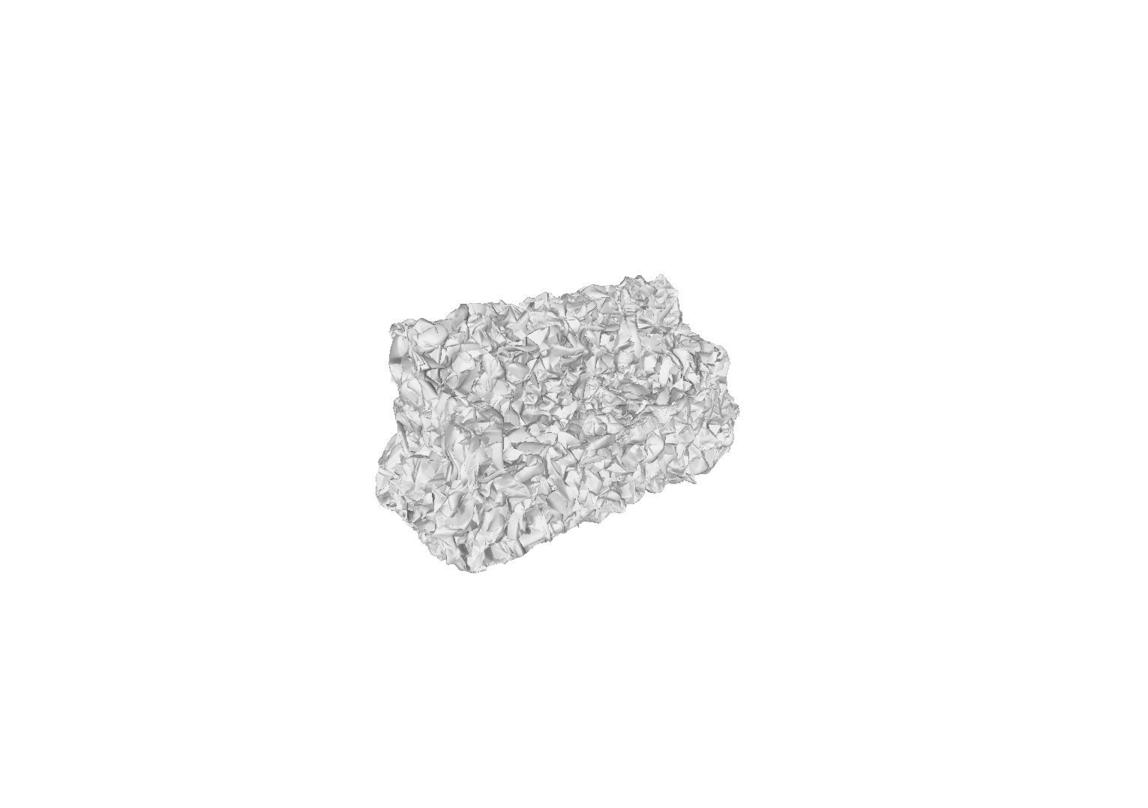}}
\subfloat{\includegraphics[width=\fitscale\tgtwidth, trim={450 152 450 152}, clip]{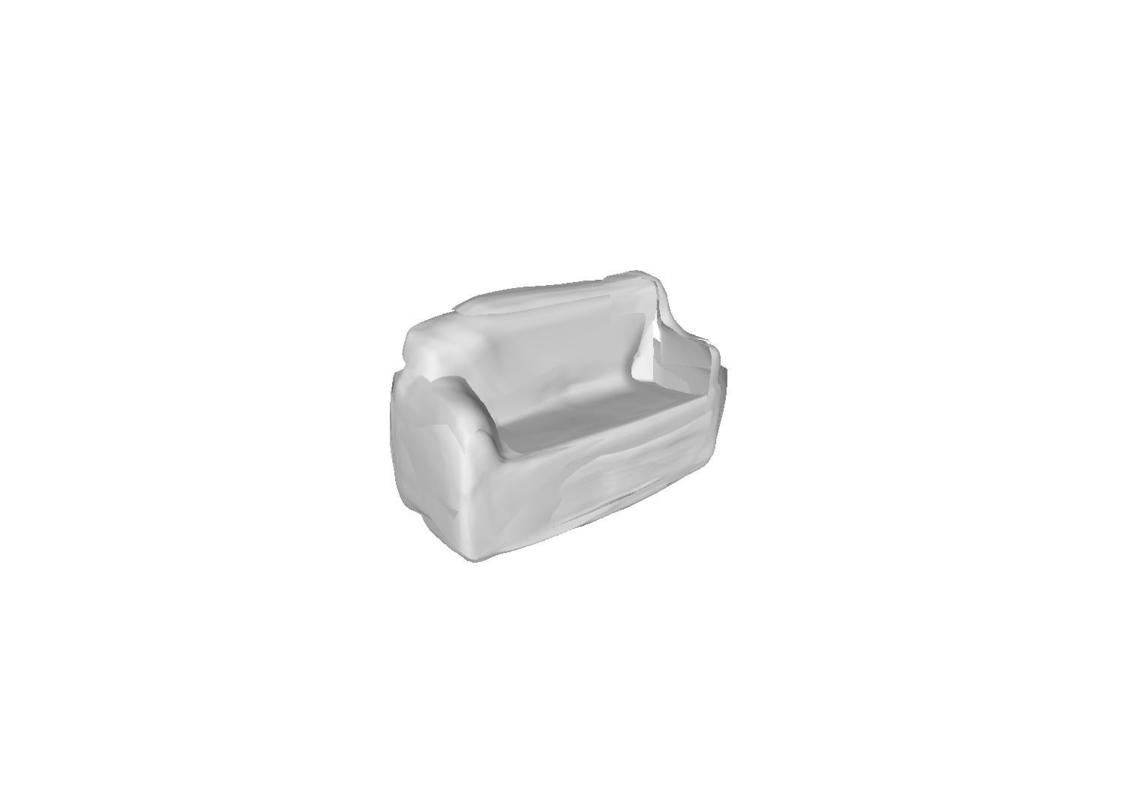}}
\subfloat{\includegraphics[width=\fitscale\tgtwidth, trim={450 152 450 152}, clip]{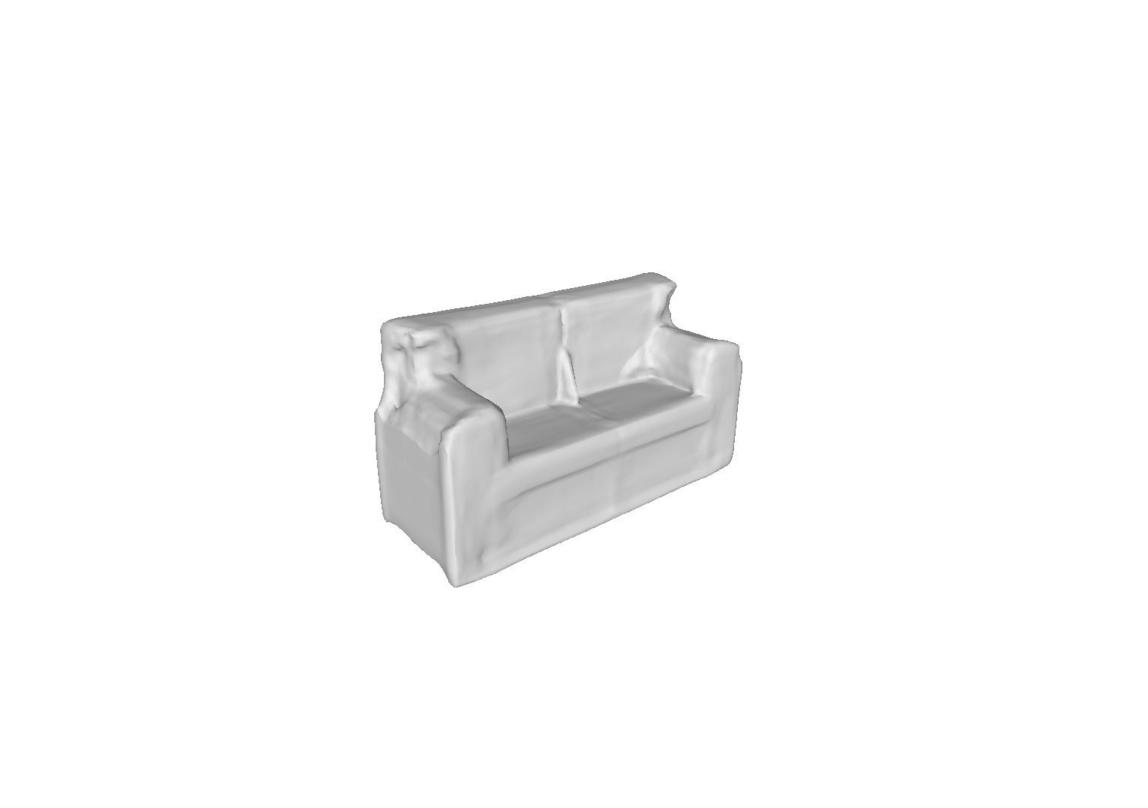}}
~
\subfloat{\includegraphics[width=\fitscale\tgtwidth, trim={450 152 450 152}, clip]{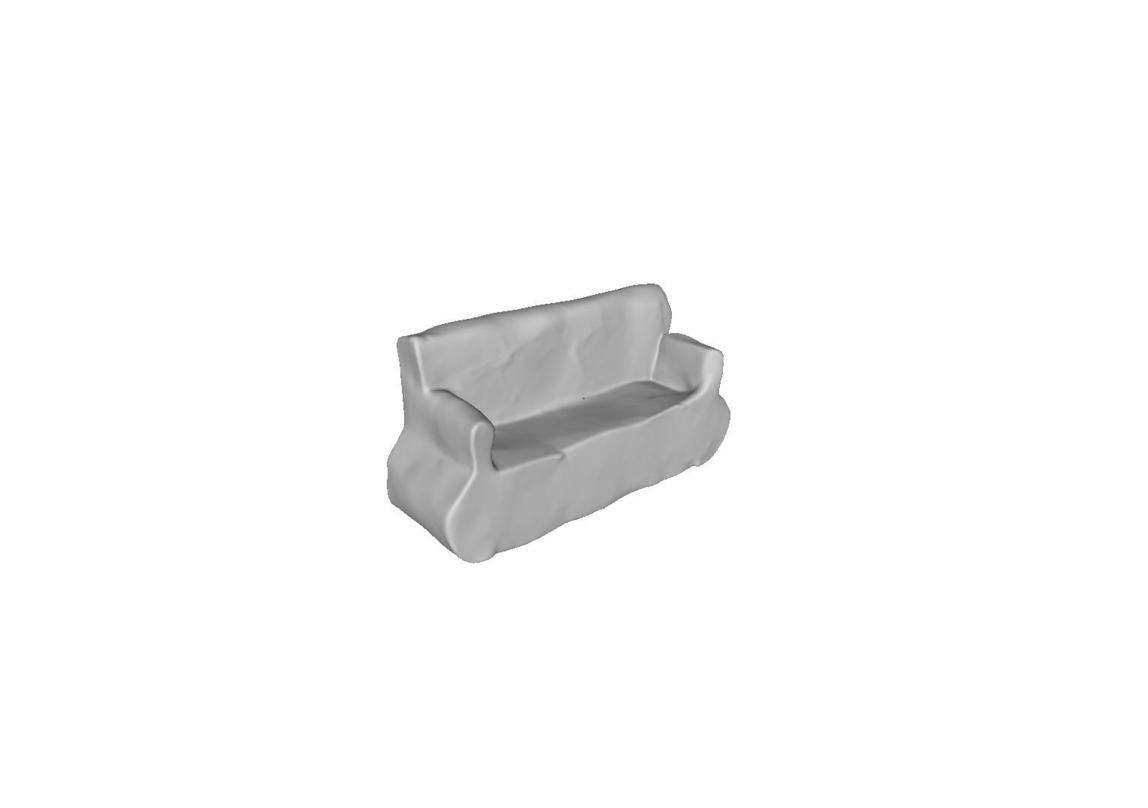}}
\\
\vspace{-5mm}
\subfloat{\includegraphics[width=\fitscale\tgtwidth, trim={450 152 450 152}, clip]{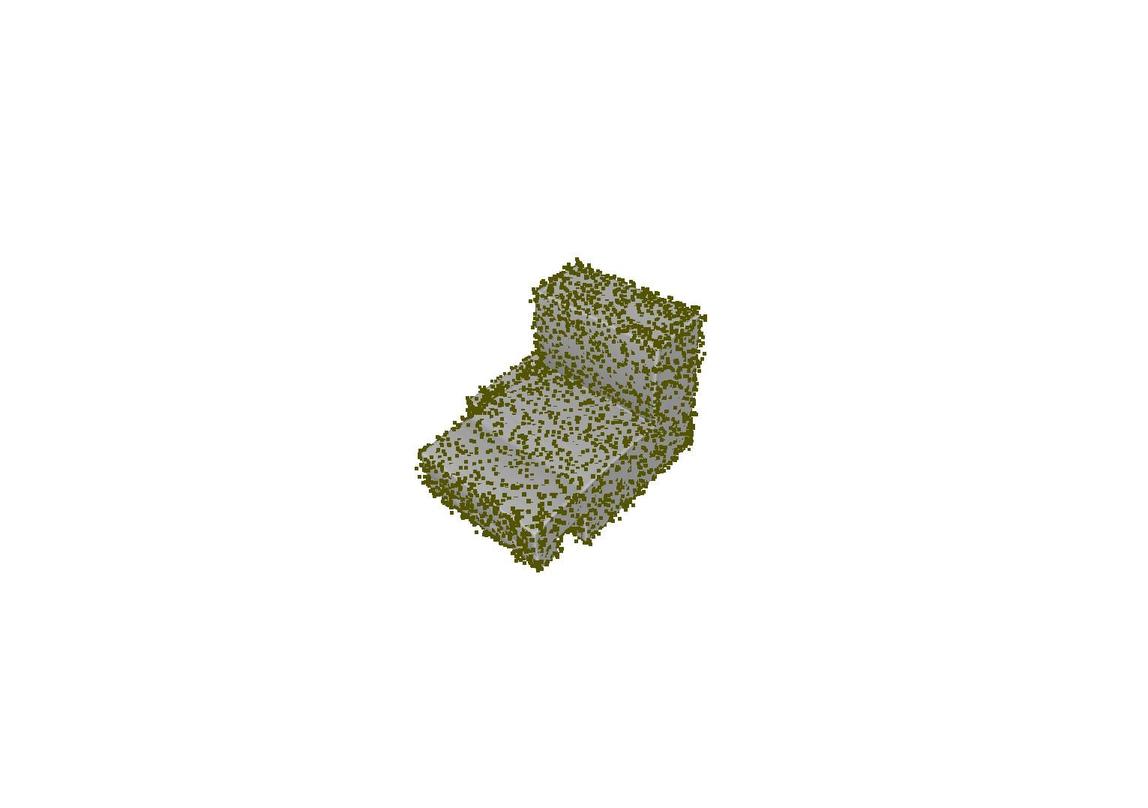}}
\subfloat{\includegraphics[width=\fitscale\tgtwidth, trim={450 152 450 152}, clip]{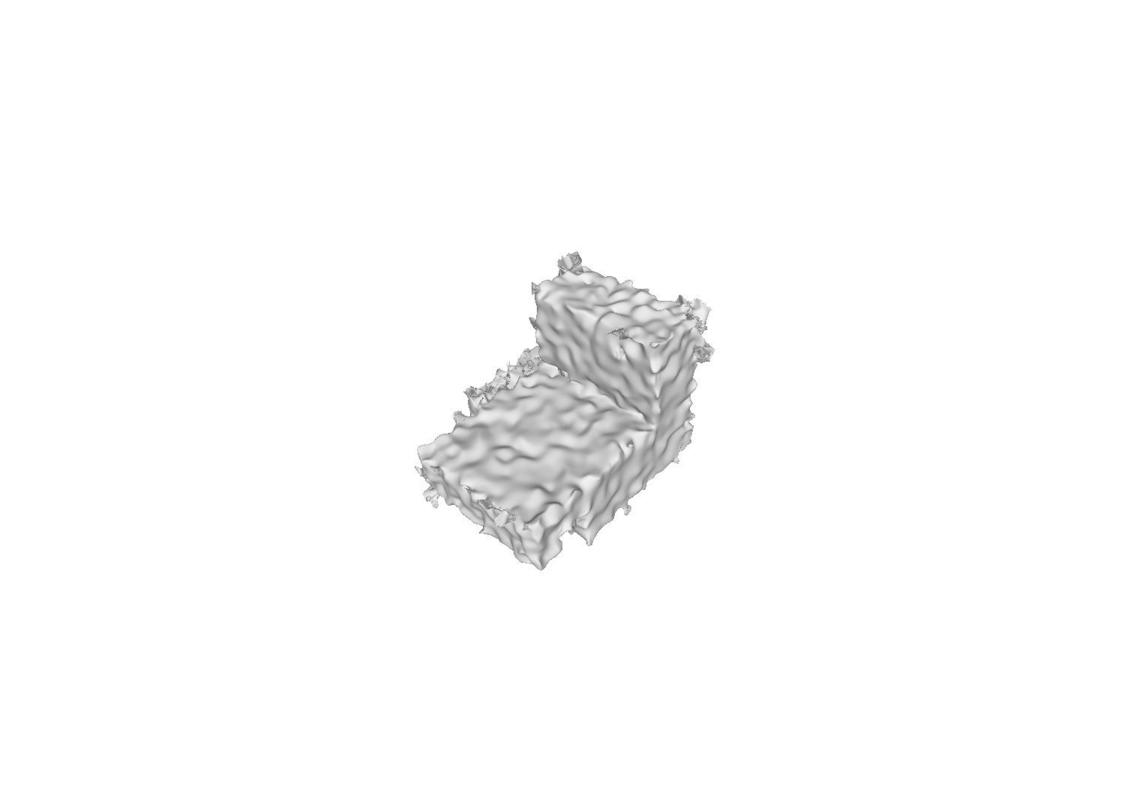}}
\subfloat{\includegraphics[width=\fitscale\tgtwidth, trim={450 152 450 152}, clip]{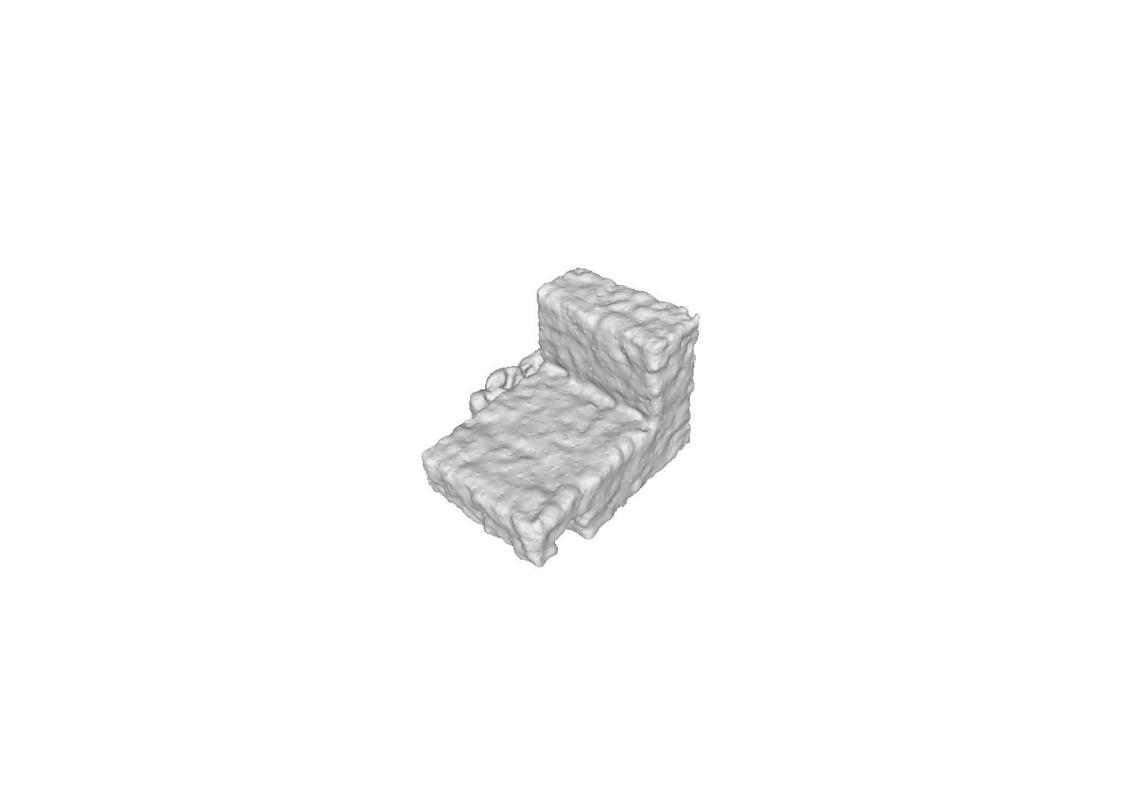}}
\subfloat{\includegraphics[width=\fitscale\tgtwidth, trim={450 152 450 152}, clip]{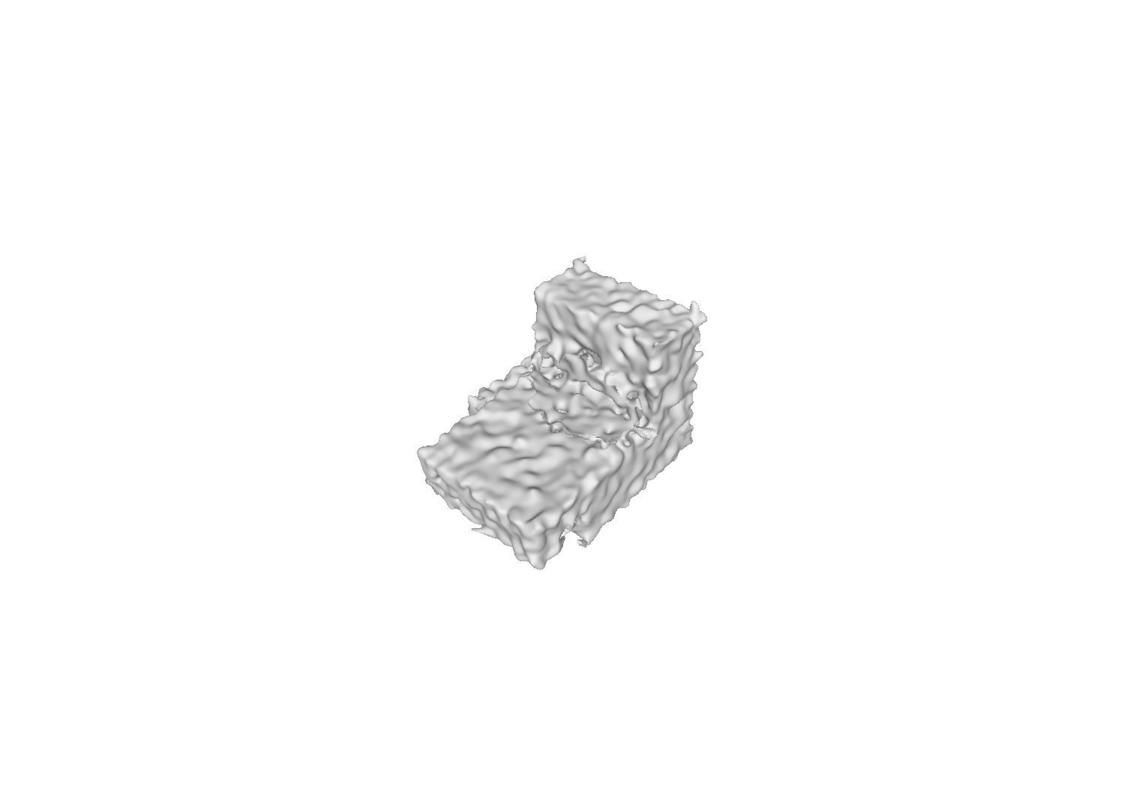}}
\subfloat{\includegraphics[width=\fitscale\tgtwidth, trim={450 152 450 152}, clip]{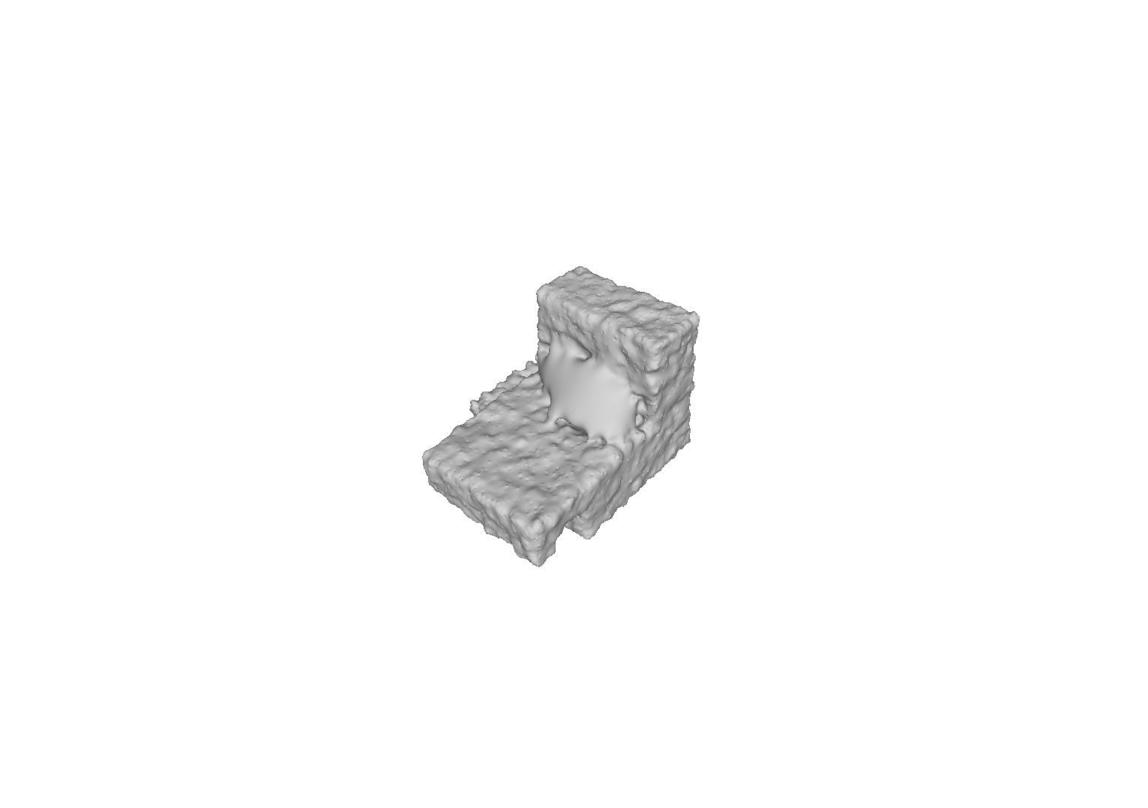}}
\subfloat{\includegraphics[width=\fitscale\tgtwidth, trim={450 152 450 152}, clip]{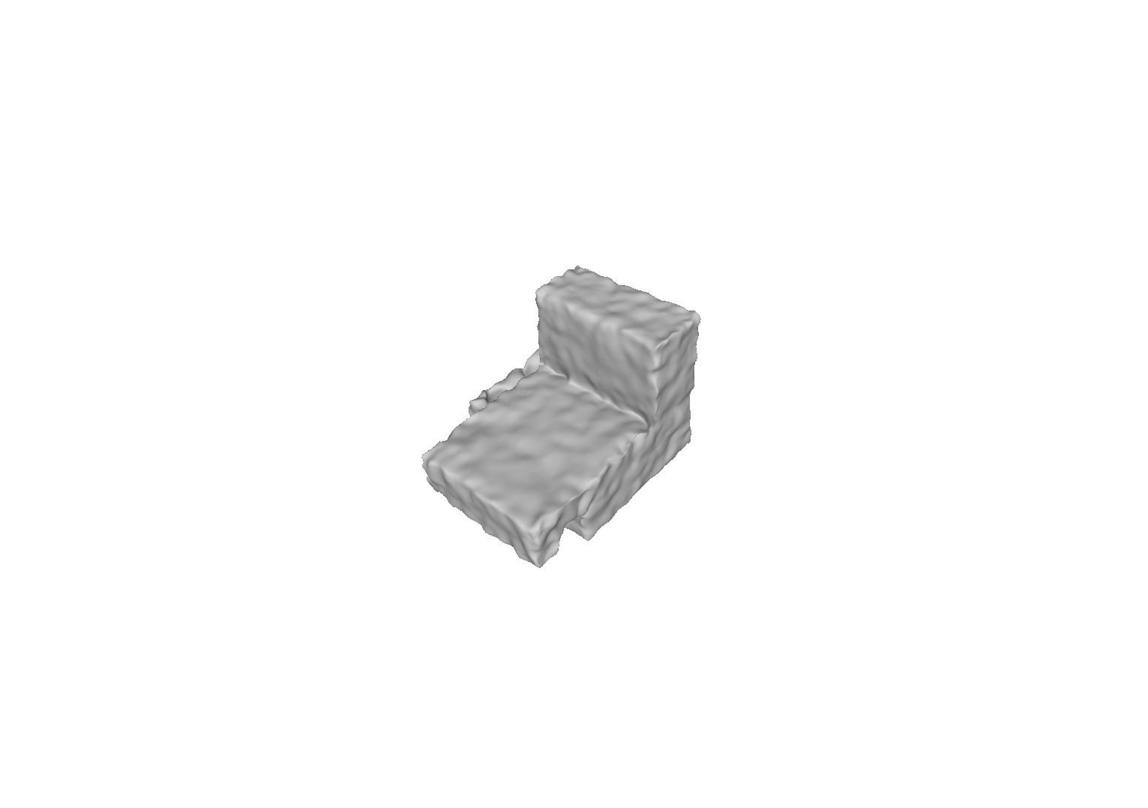}}
\subfloat{\includegraphics[width=\fitscale\tgtwidth, trim={450 152 450 152}, clip]{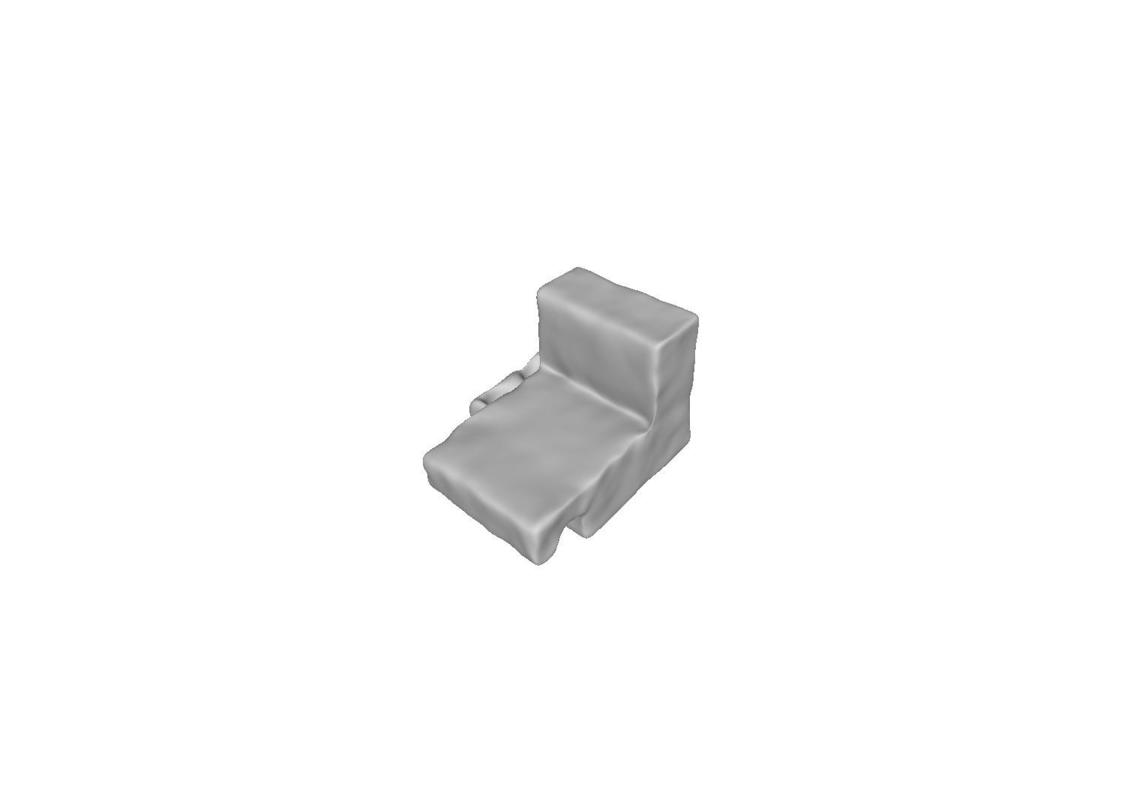}}
\subfloat{\includegraphics[width=\fitscale\tgtwidth, trim={450 152 450 152}, clip]{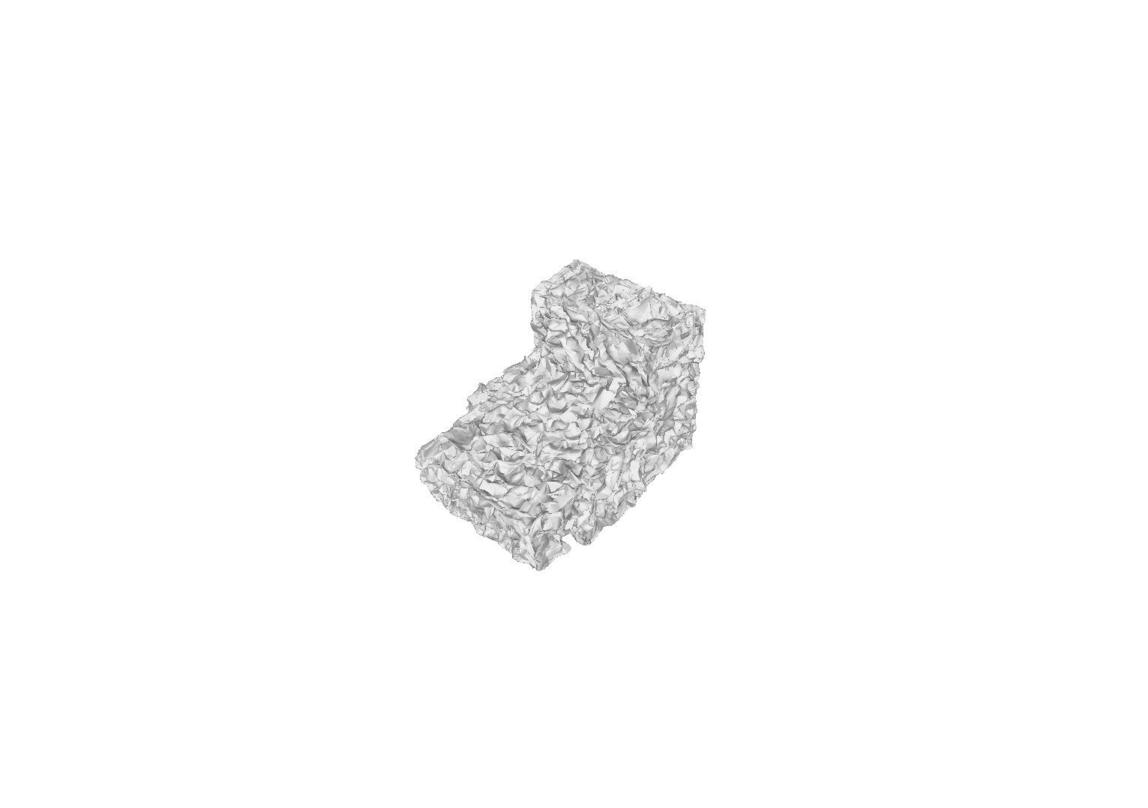}}
\subfloat{\includegraphics[width=\fitscale\tgtwidth, trim={450 152 450 152}, clip]{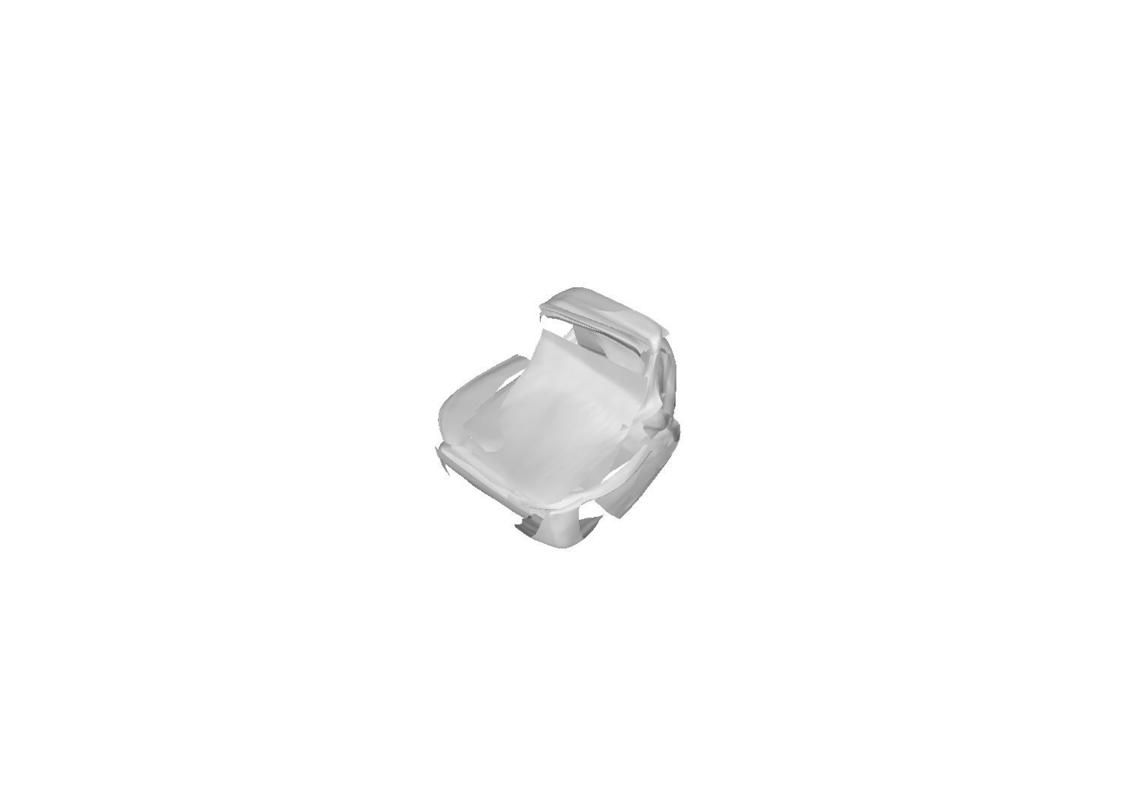}}
\subfloat{\includegraphics[width=\fitscale\tgtwidth, trim={450 152 450 152}, clip]{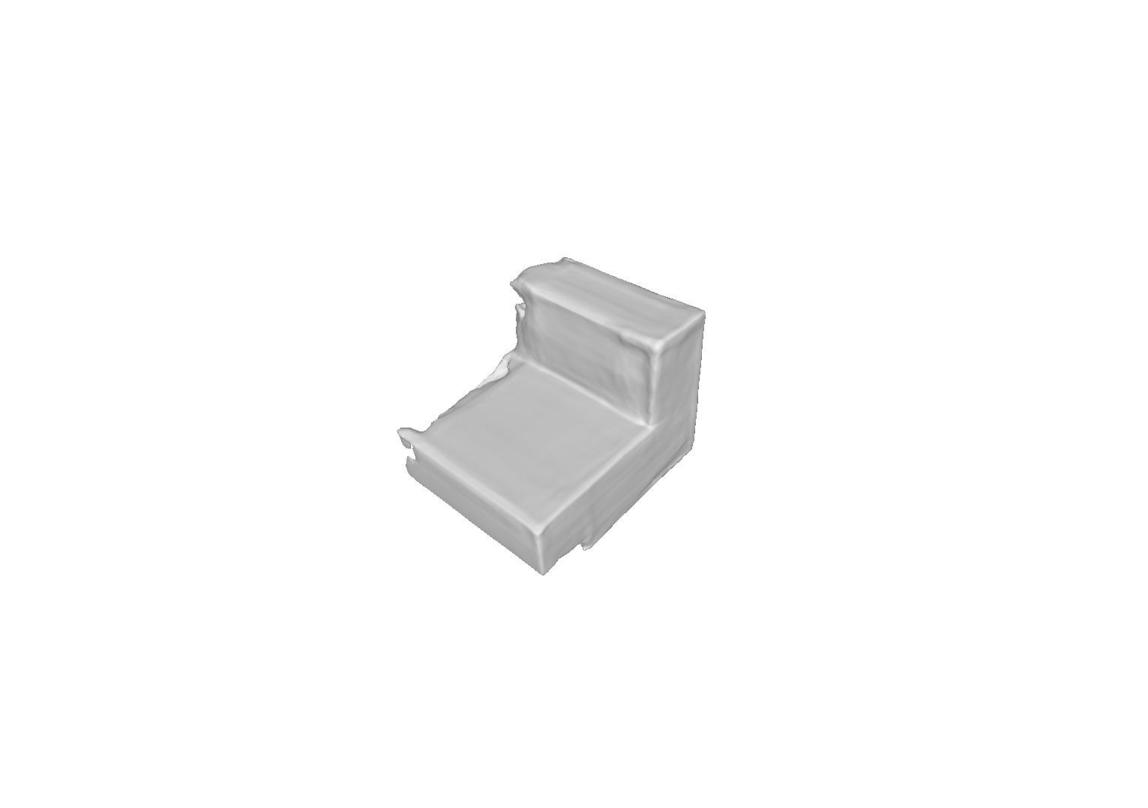}}
~
\subfloat{\includegraphics[width=\fitscale\tgtwidth, trim={450 152 450 152}, clip]{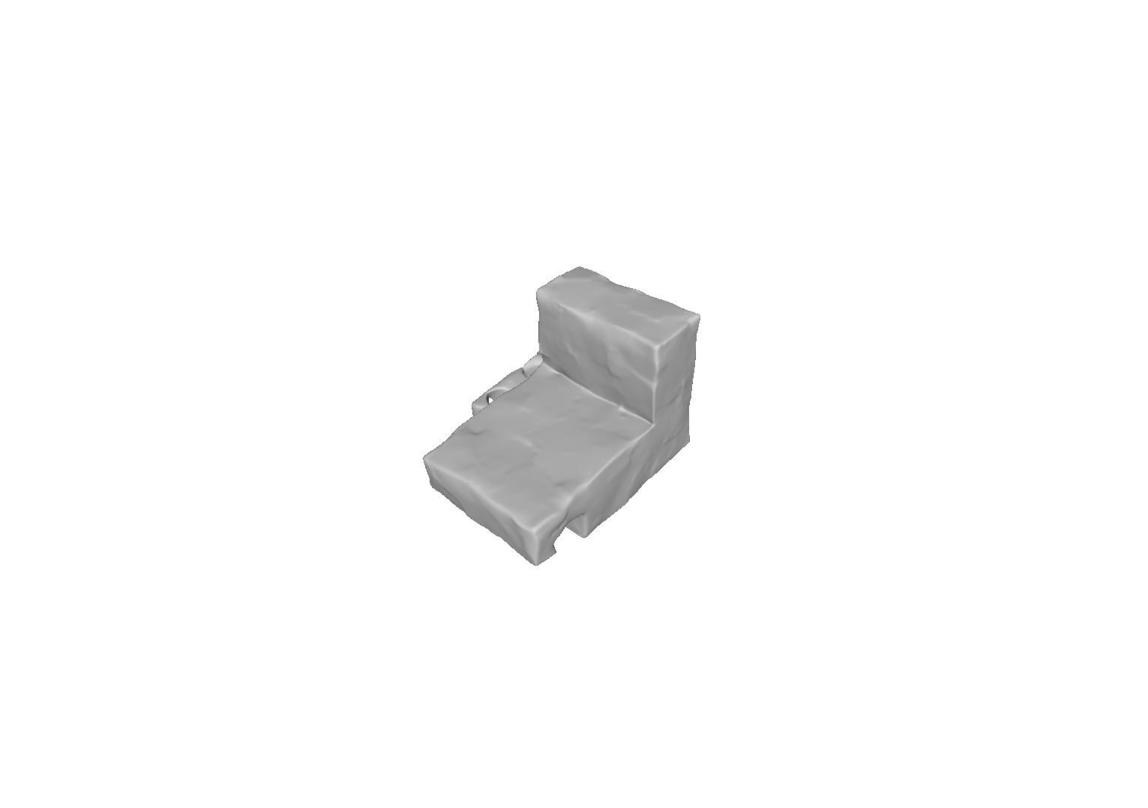}}
\\
\vspace{-5mm}
\subfloat{\includegraphics[width=\fitscale\tgtwidth, trim={450 152 450 152}, clip]{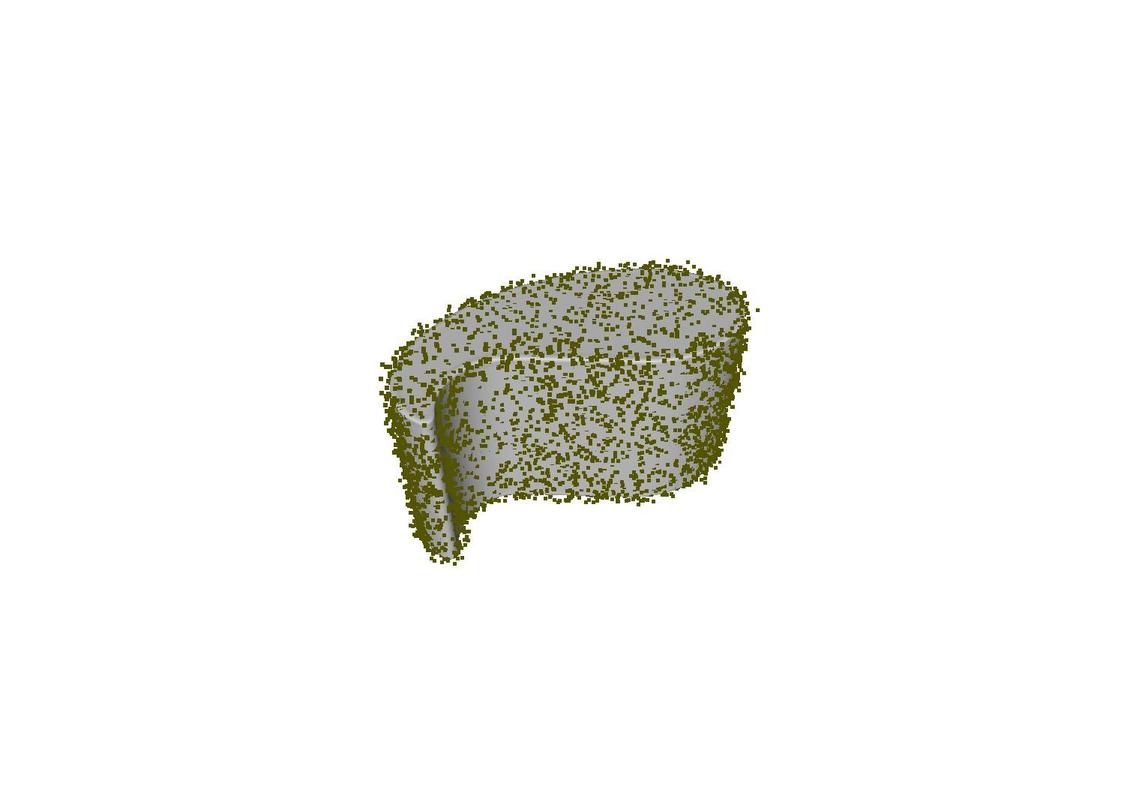}}
\subfloat{\includegraphics[width=\fitscale\tgtwidth, trim={450 152 450 152}, clip]{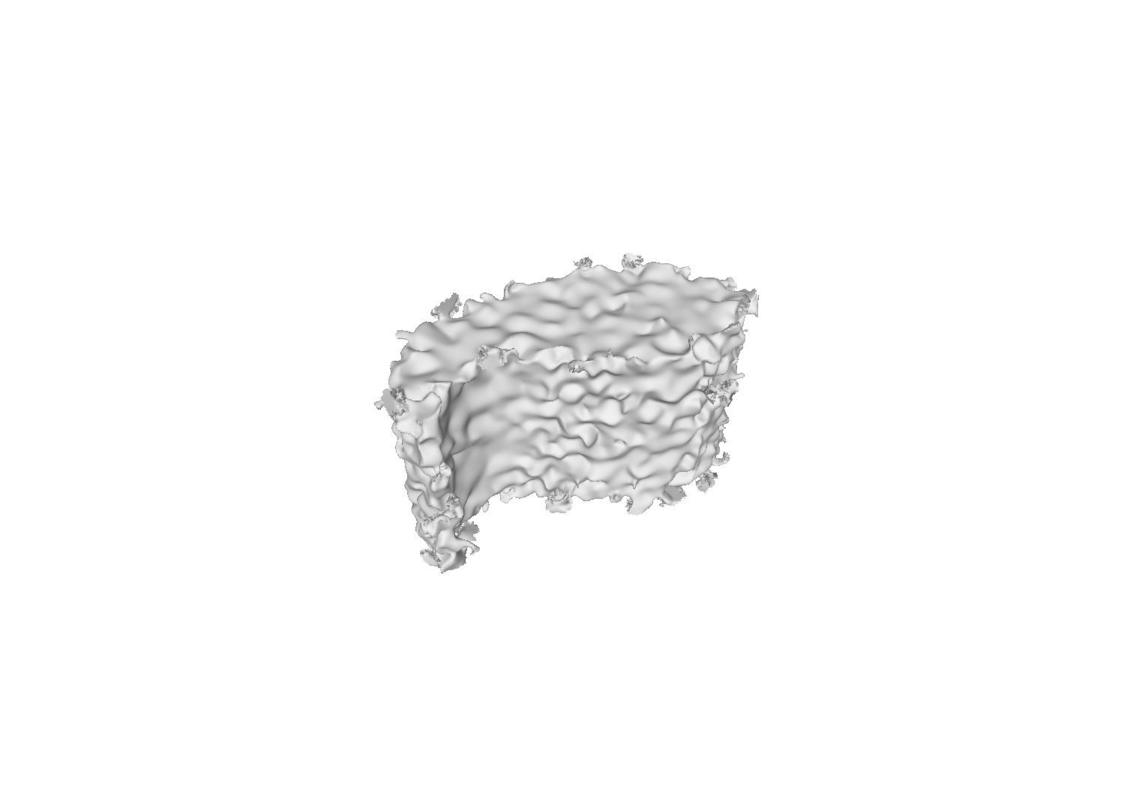}}
\subfloat{\includegraphics[width=\fitscale\tgtwidth, trim={450 152 450 152}, clip]{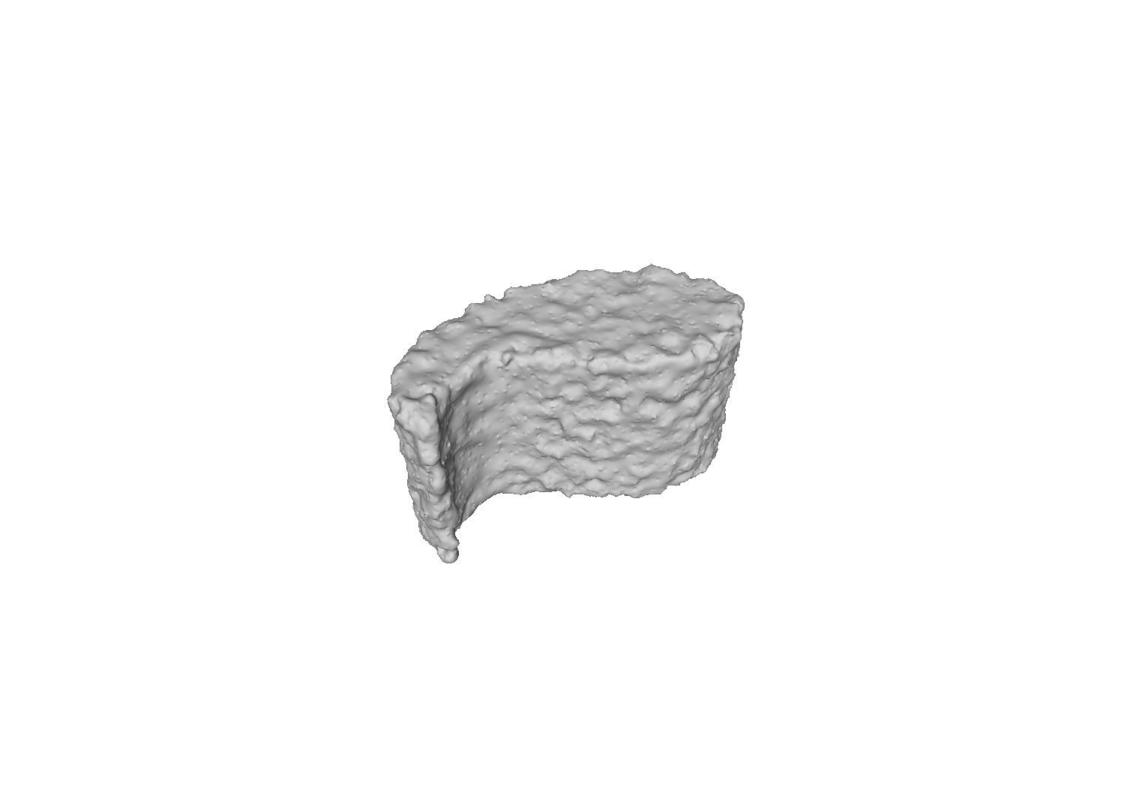}}
\subfloat{\includegraphics[width=\fitscale\tgtwidth, trim={450 152 450 152}, clip]{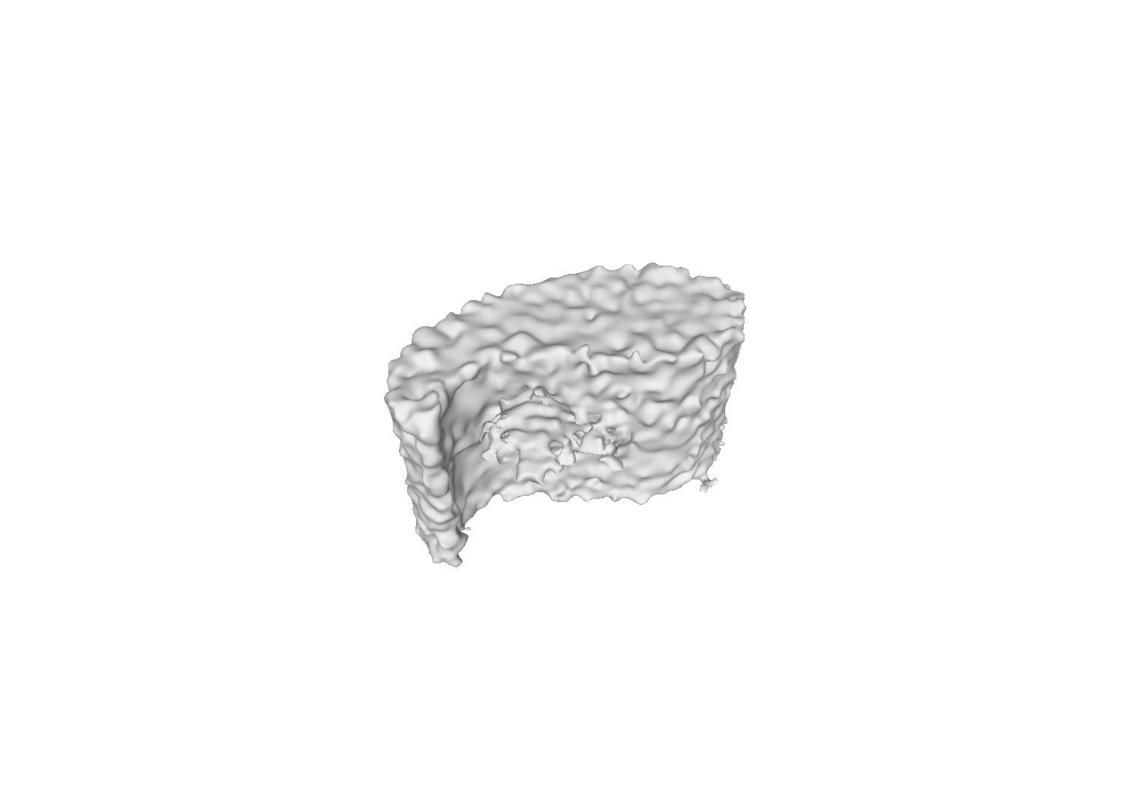}}
\subfloat{\includegraphics[width=\fitscale\tgtwidth, trim={450 152 450 152}, clip]{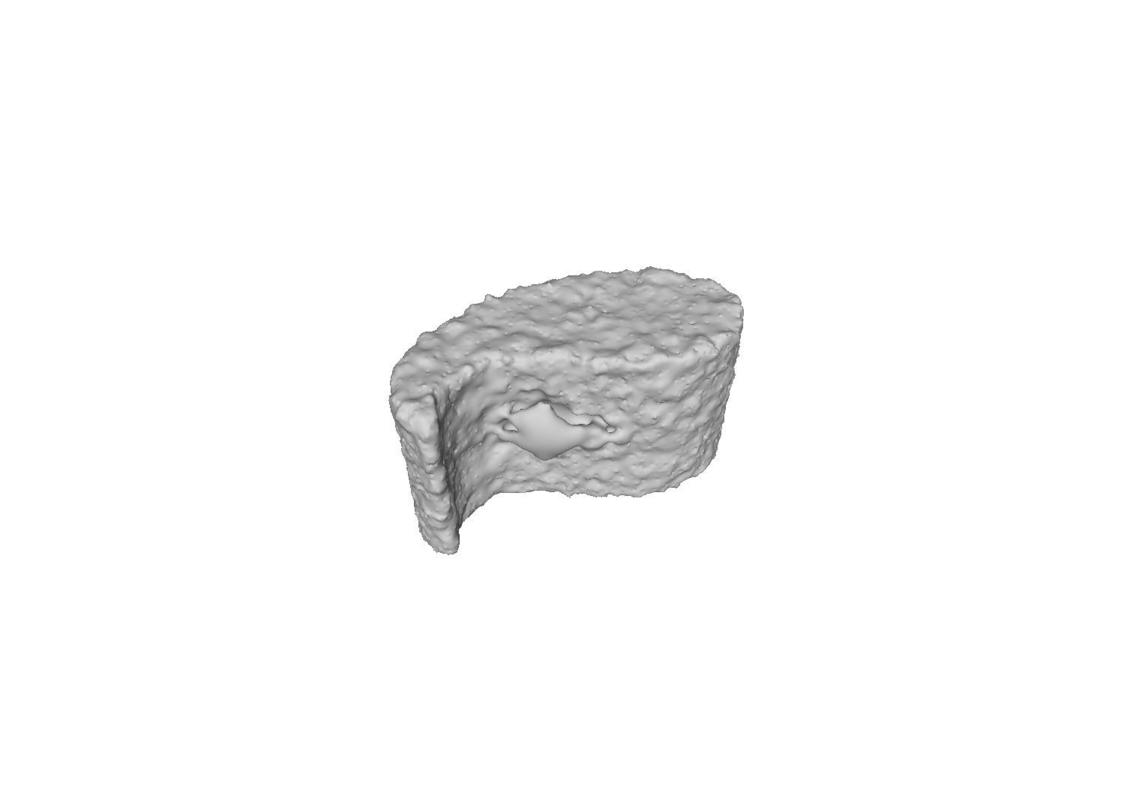}}
\subfloat{\includegraphics[width=\fitscale\tgtwidth, trim={450 152 450 152}, clip]{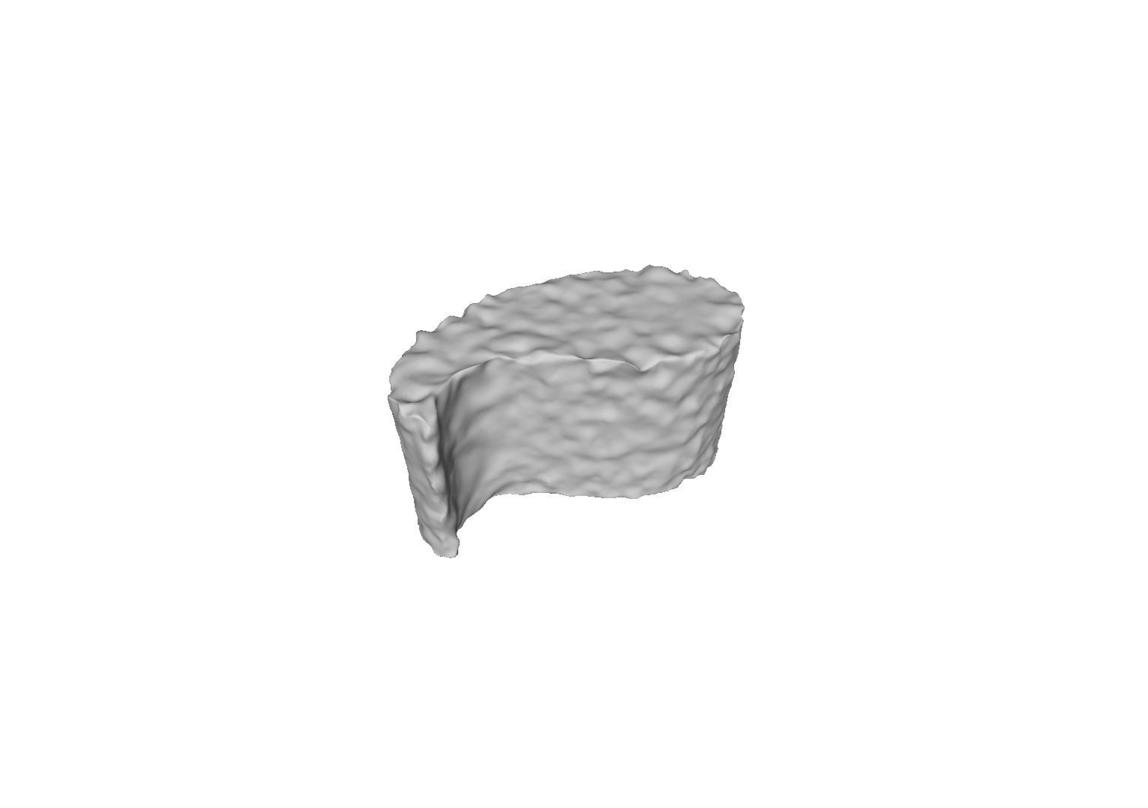}}
\subfloat{\includegraphics[width=\fitscale\tgtwidth, trim={450 152 450 152}, clip]{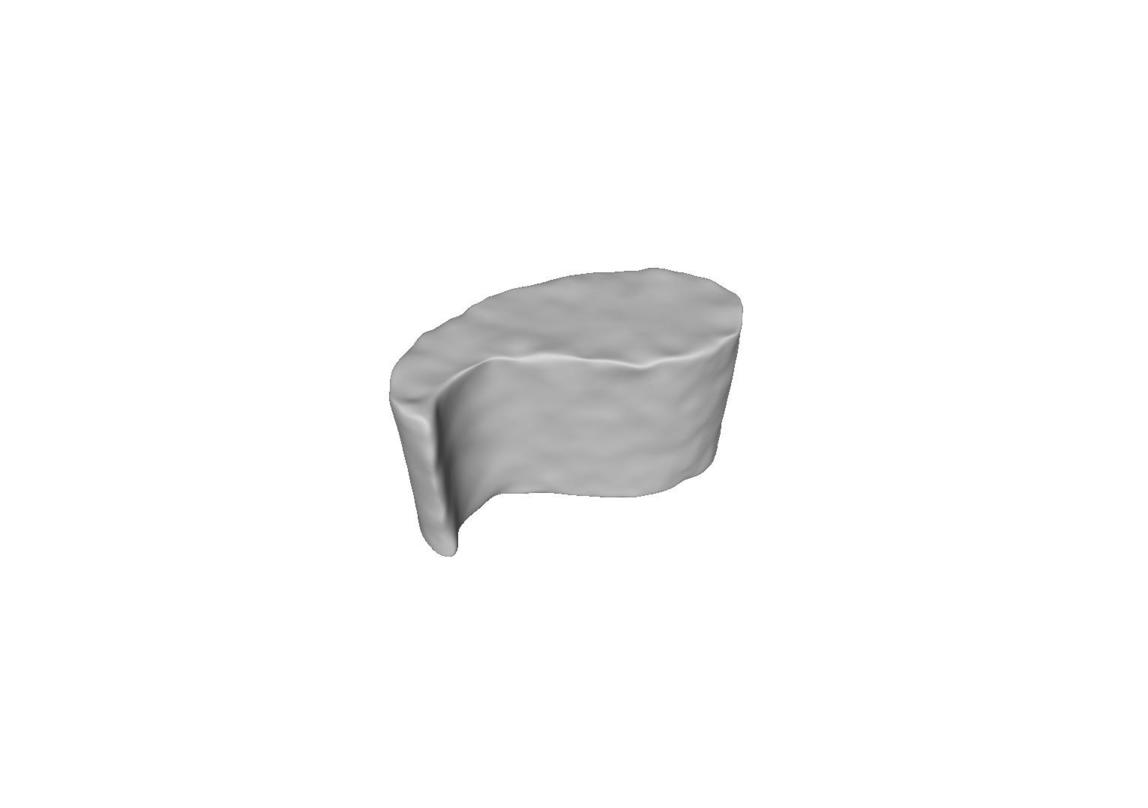}}
\subfloat{\includegraphics[width=\fitscale\tgtwidth, trim={450 152 450 152}, clip]{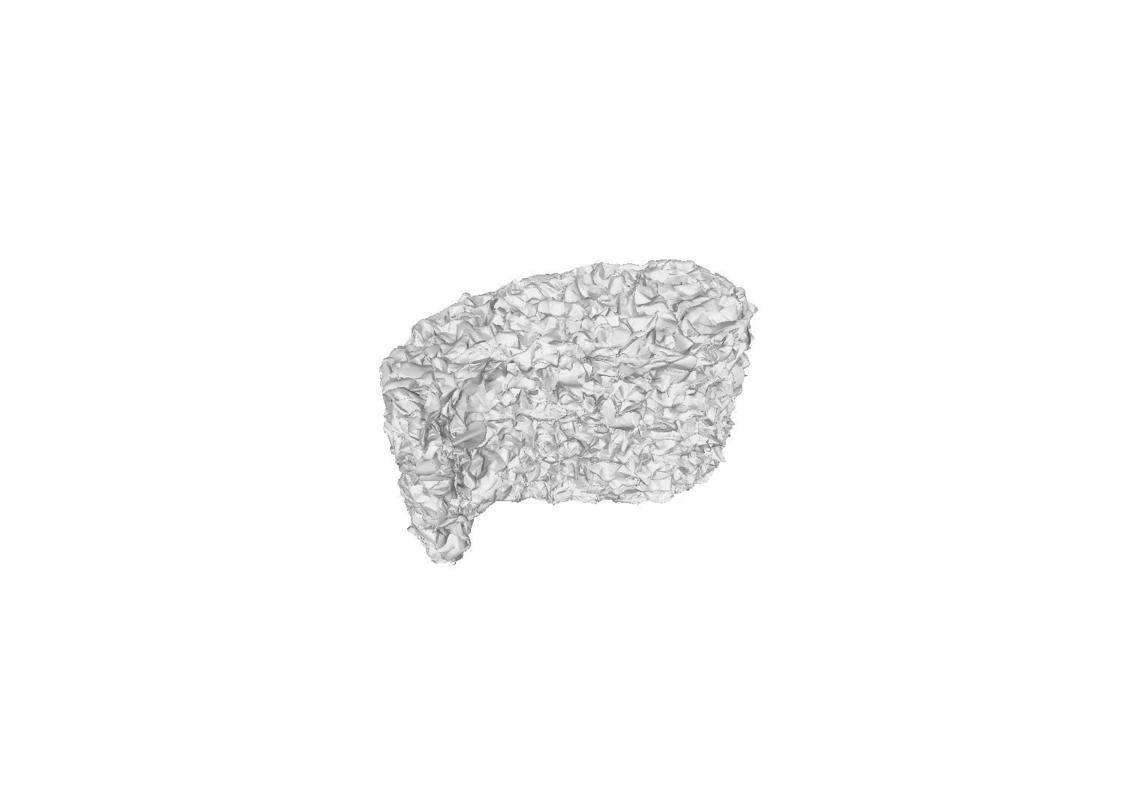}}
\subfloat{\includegraphics[width=\fitscale\tgtwidth, trim={450 152 450 152}, clip]{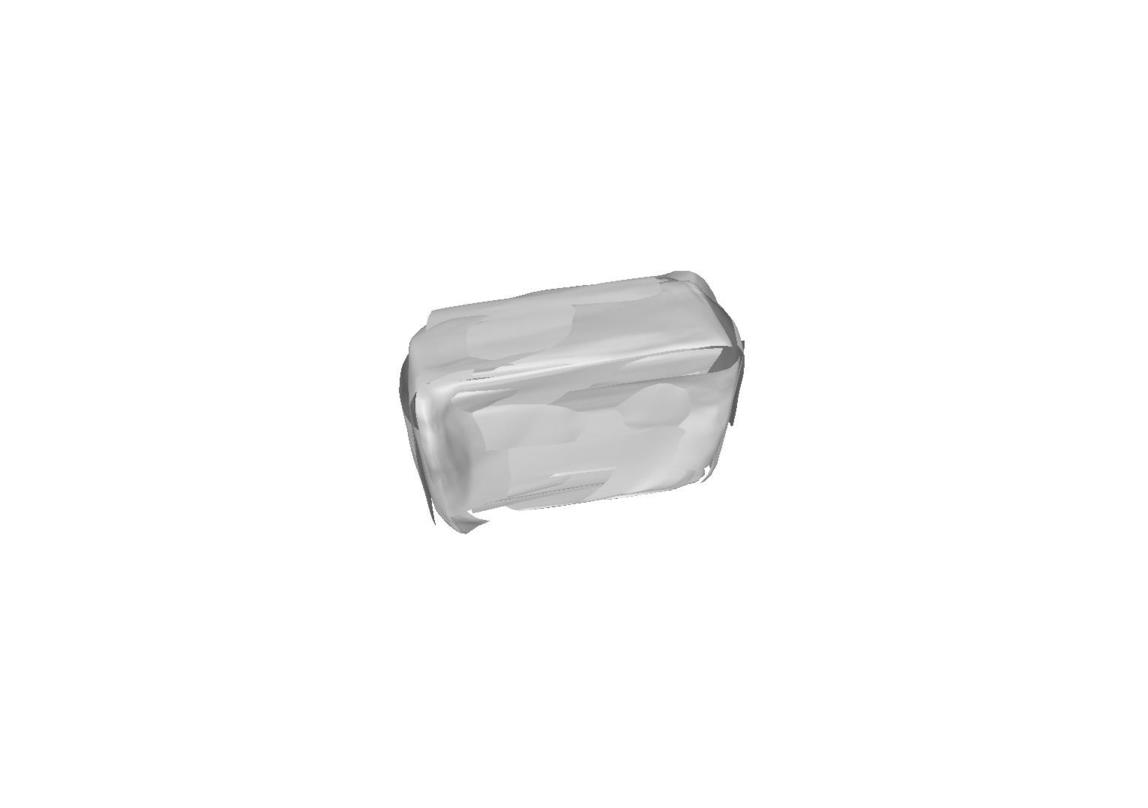}}
\subfloat{\includegraphics[width=\fitscale\tgtwidth, trim={450 152 450 152}, clip]{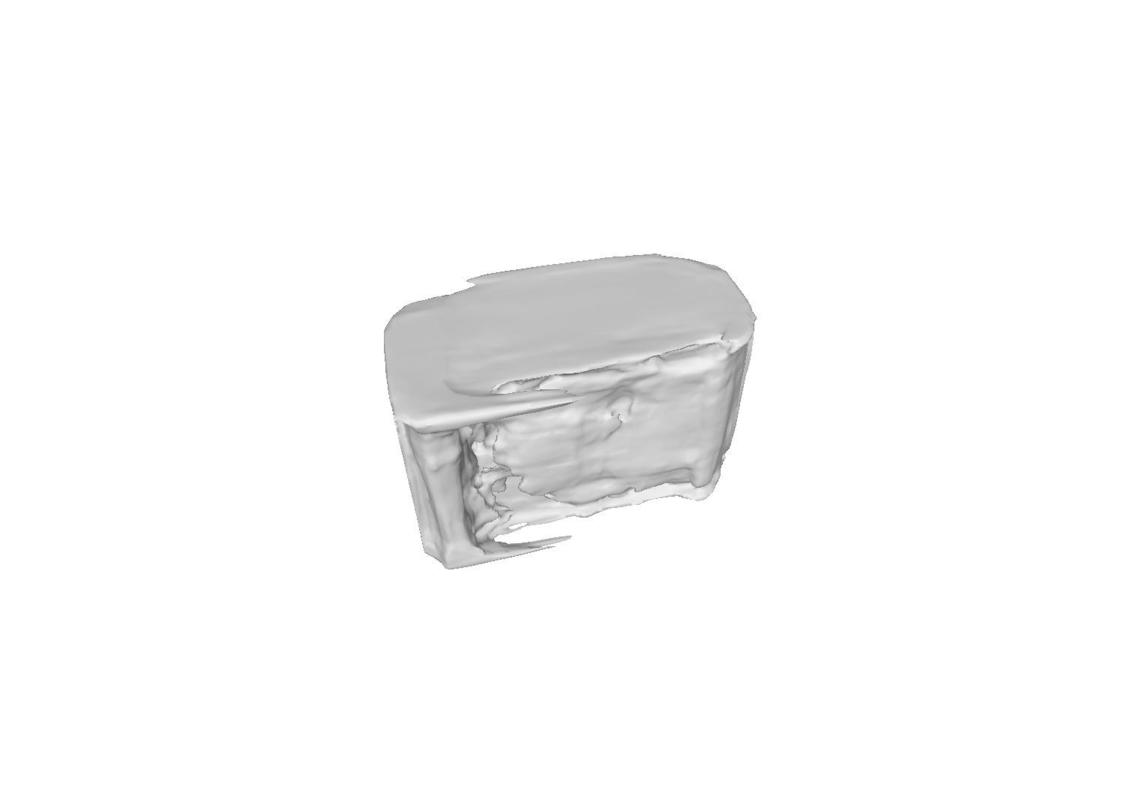}}
~
\subfloat{\includegraphics[width=\fitscale\tgtwidth, trim={450 152 450 152}, clip]{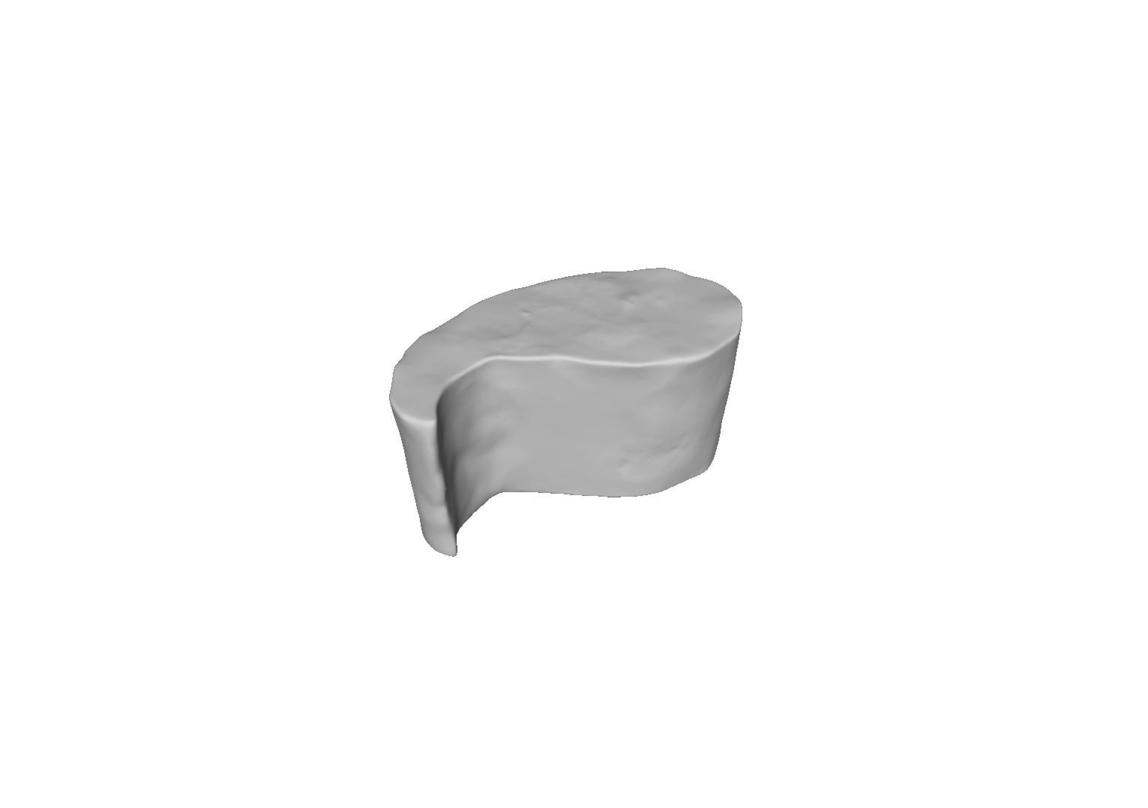}}
\\
\vspace{-5mm}
\subfloat{\includegraphics[width=\fitscale\tgtwidth, trim={450 152 450 152}, clip]{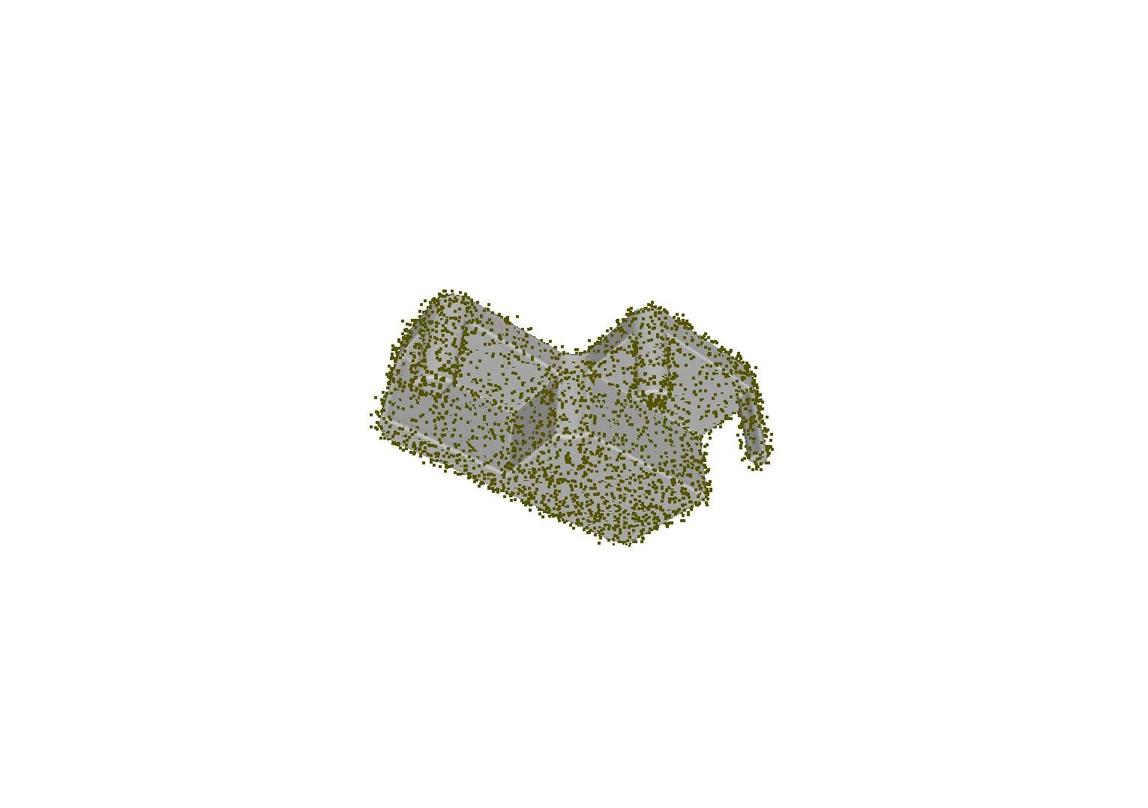}}
\subfloat{\includegraphics[width=\fitscale\tgtwidth, trim={450 152 450 152}, clip]{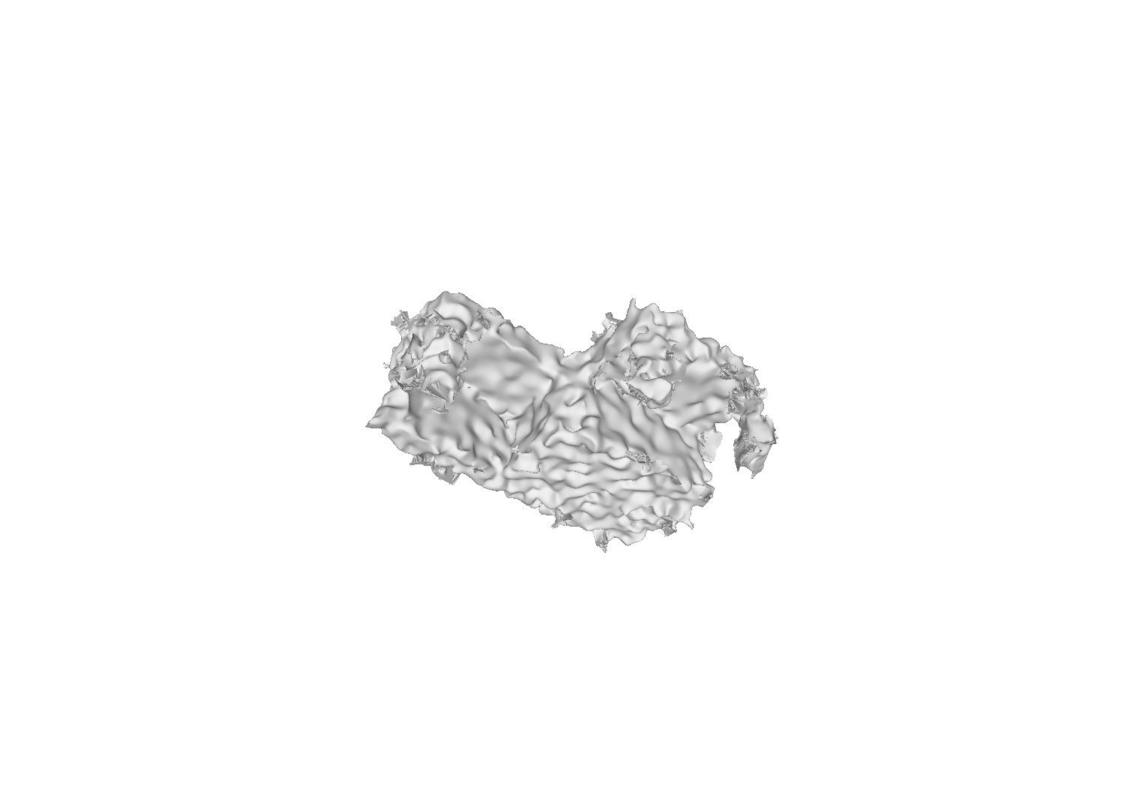}}
\subfloat{\includegraphics[width=\fitscale\tgtwidth, trim={450 152 450 152}, clip]{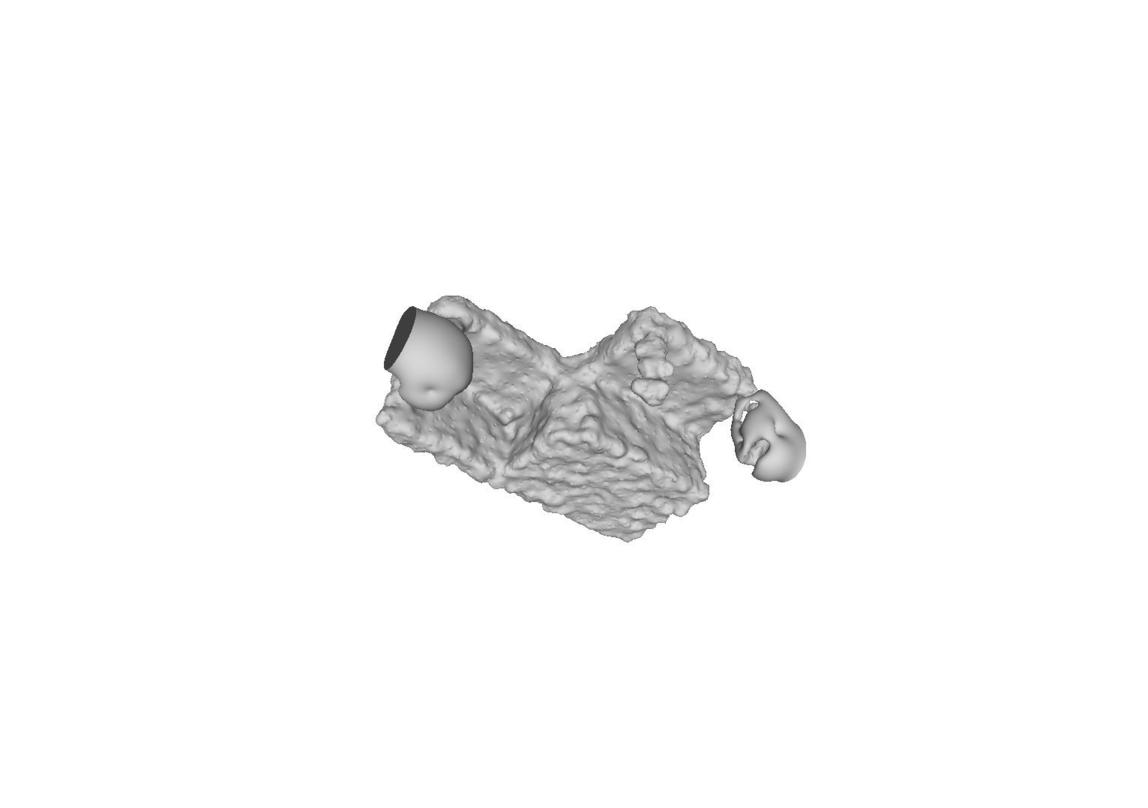}}
\subfloat{\includegraphics[width=\fitscale\tgtwidth, trim={450 152 450 152}, clip]{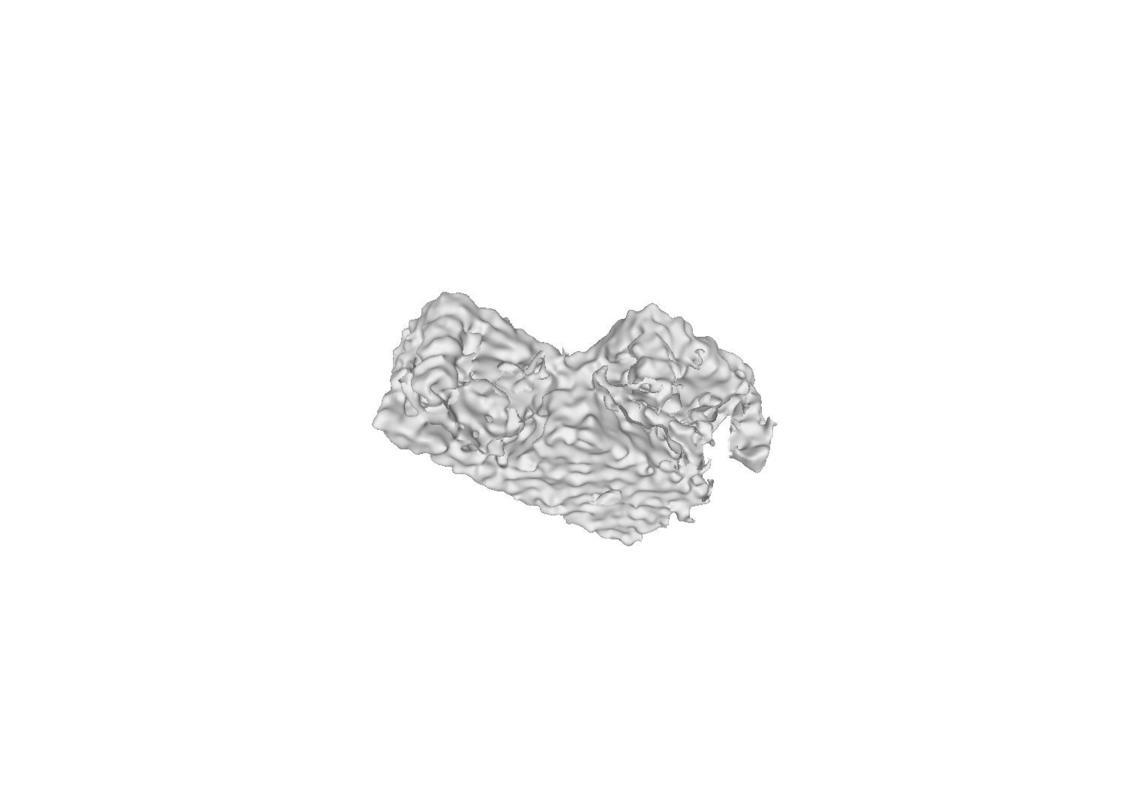}}
\subfloat{\includegraphics[width=\fitscale\tgtwidth, trim={450 152 450 152}, clip]{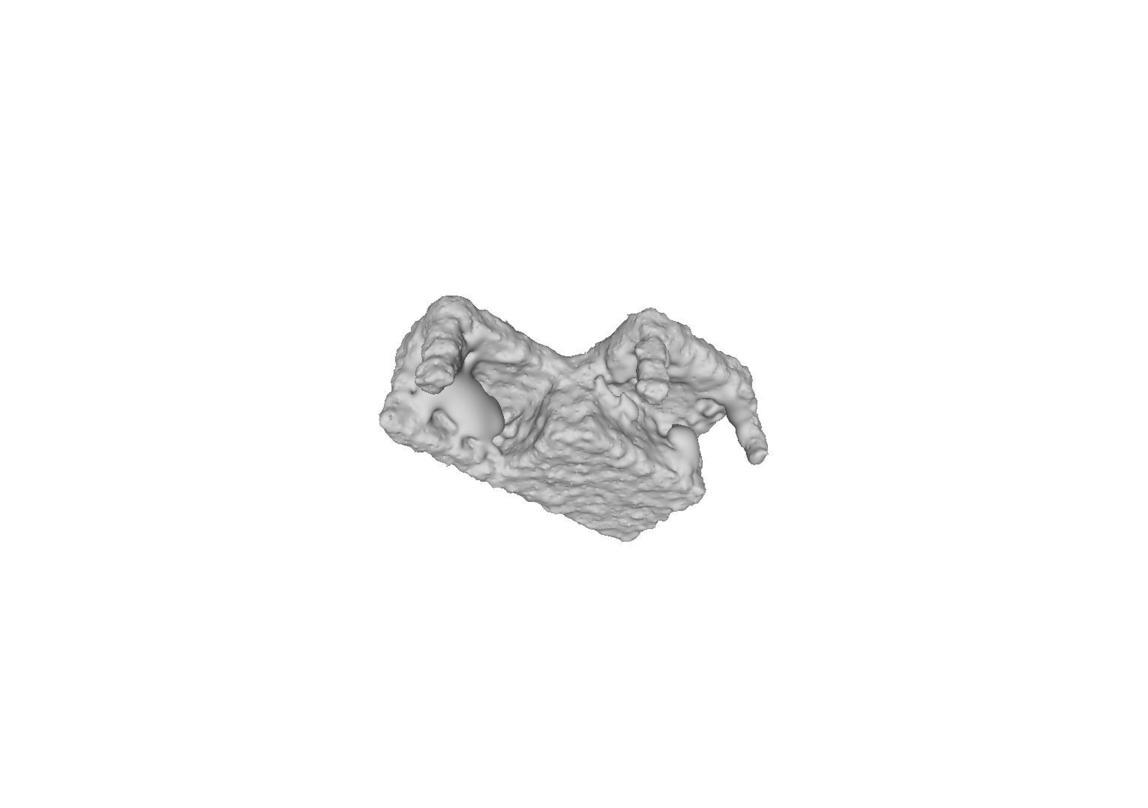}}
\subfloat{\includegraphics[width=\fitscale\tgtwidth, trim={450 152 450 152}, clip]{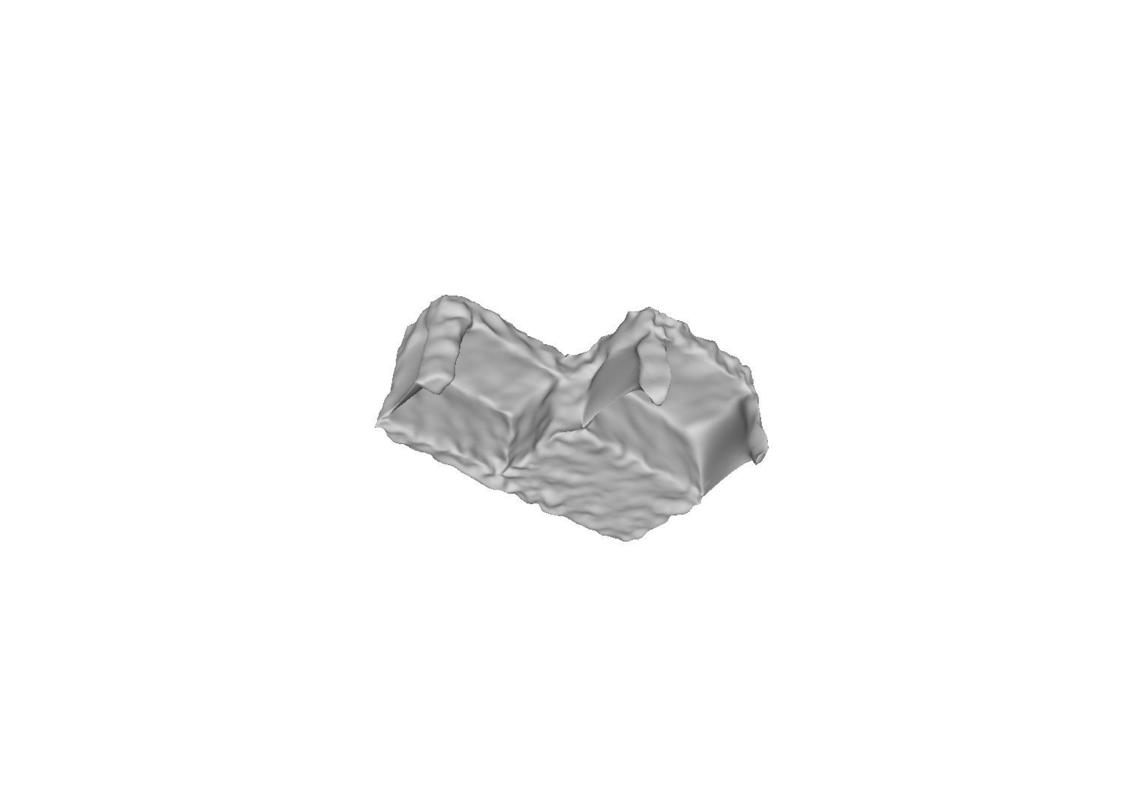}}
\subfloat{\includegraphics[width=\fitscale\tgtwidth, trim={450 152 450 152}, clip]{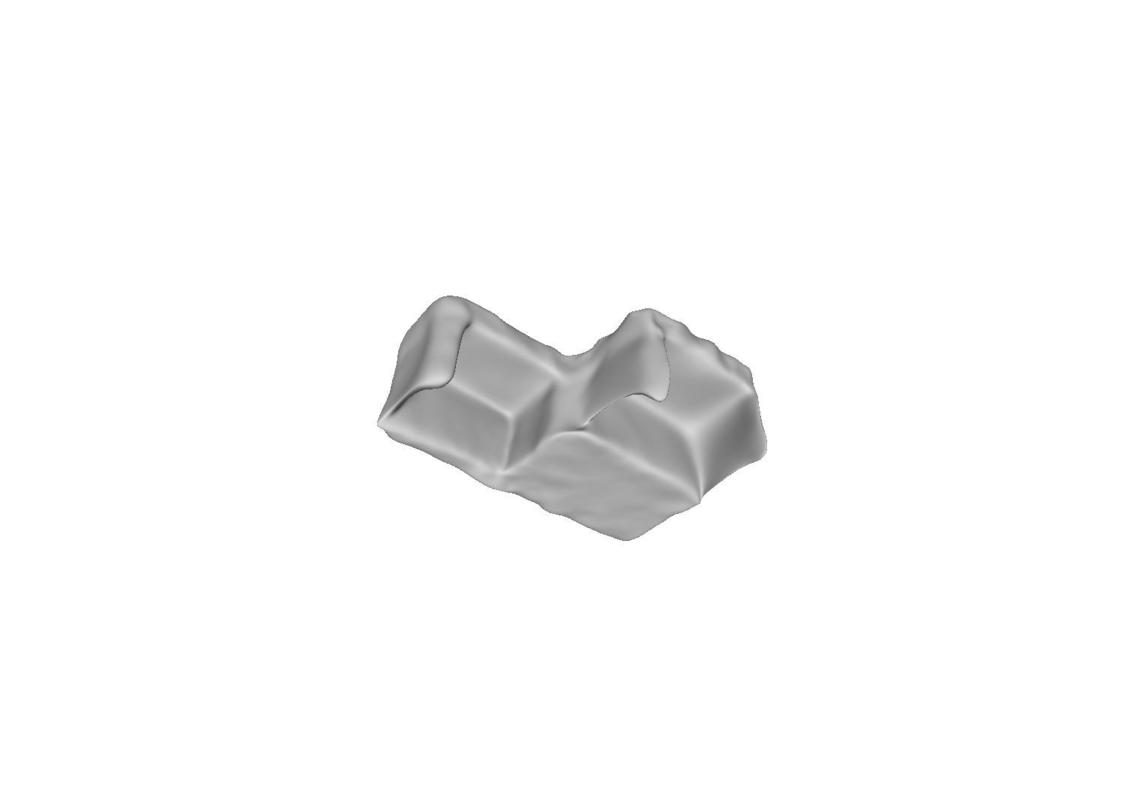}}
\subfloat{\includegraphics[width=\fitscale\tgtwidth, trim={450 152 450 152}, clip]{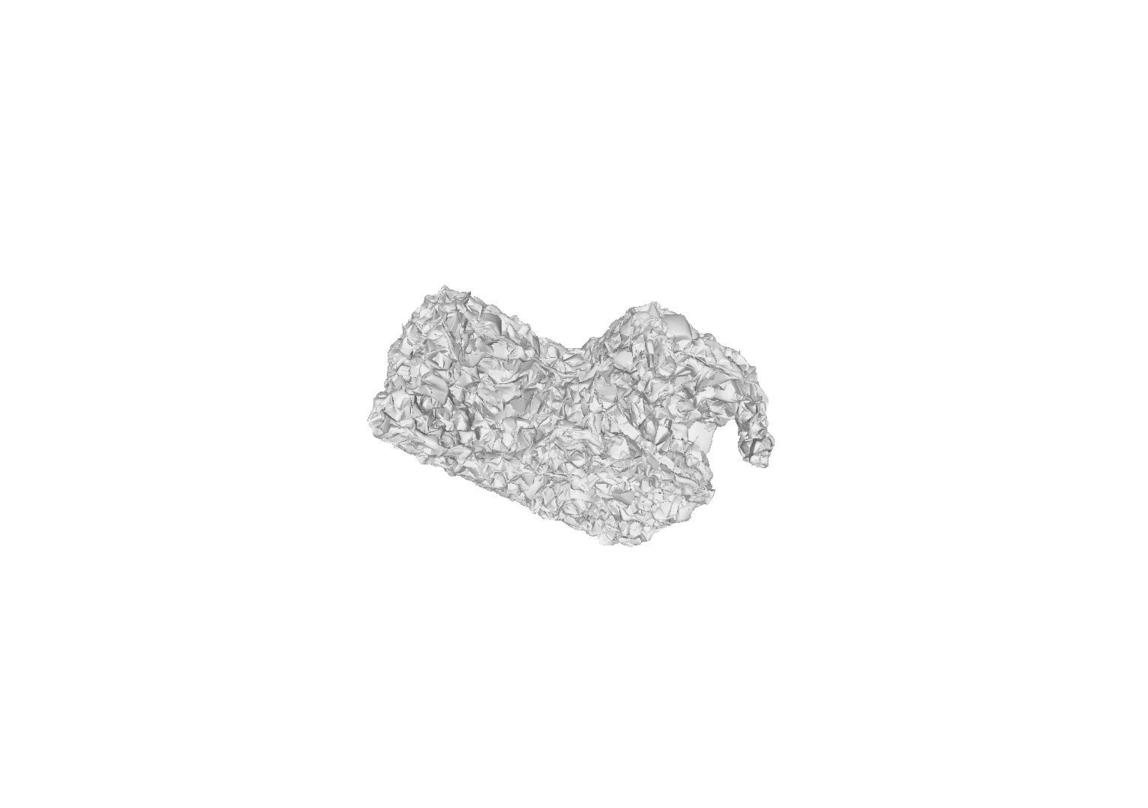}}
\subfloat{\includegraphics[width=\fitscale\tgtwidth, trim={450 152 450 152}, clip]{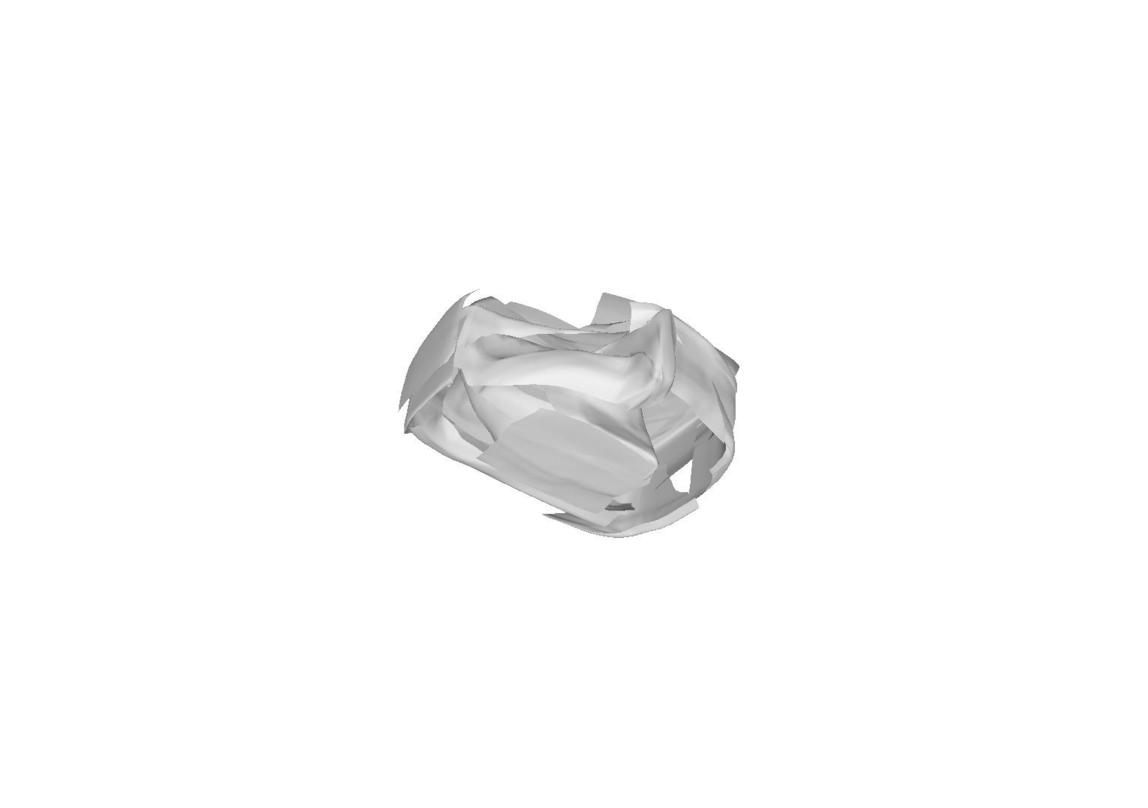}}
\subfloat{\includegraphics[width=\fitscale\tgtwidth, trim={450 152 450 152}, clip]{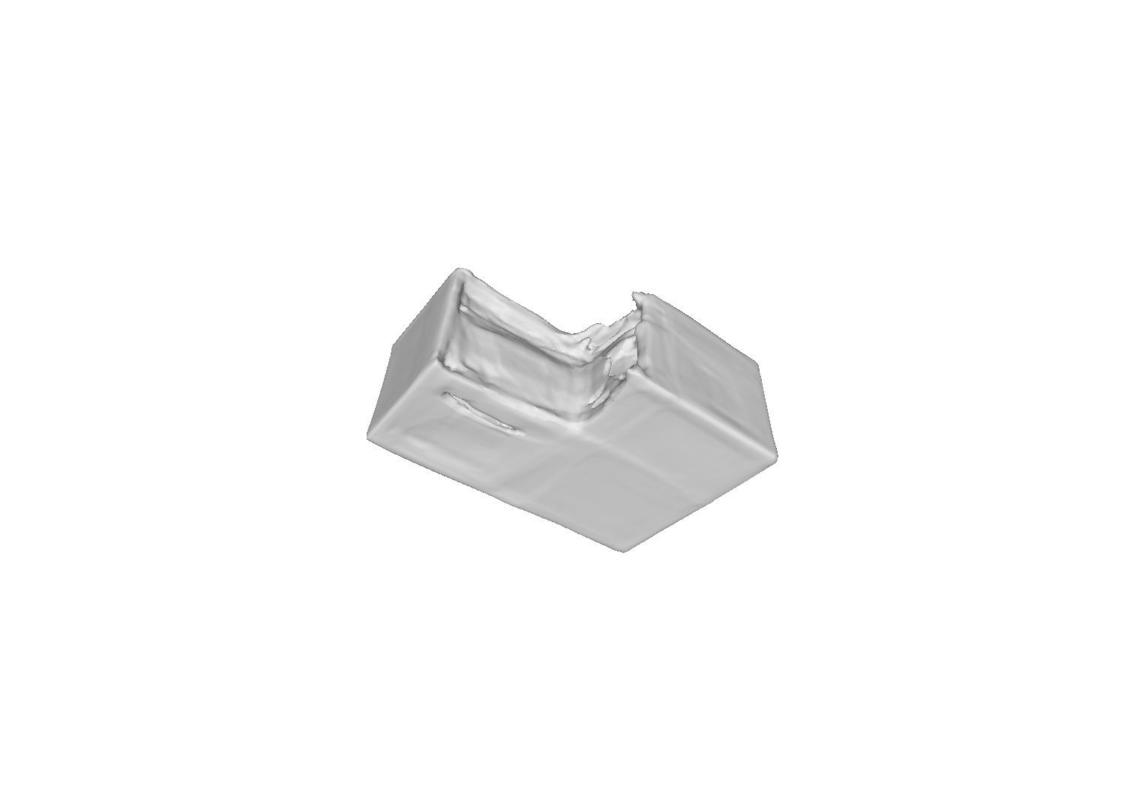}}
~
\subfloat{\includegraphics[width=\fitscale\tgtwidth, trim={450 152 450 152}, clip]{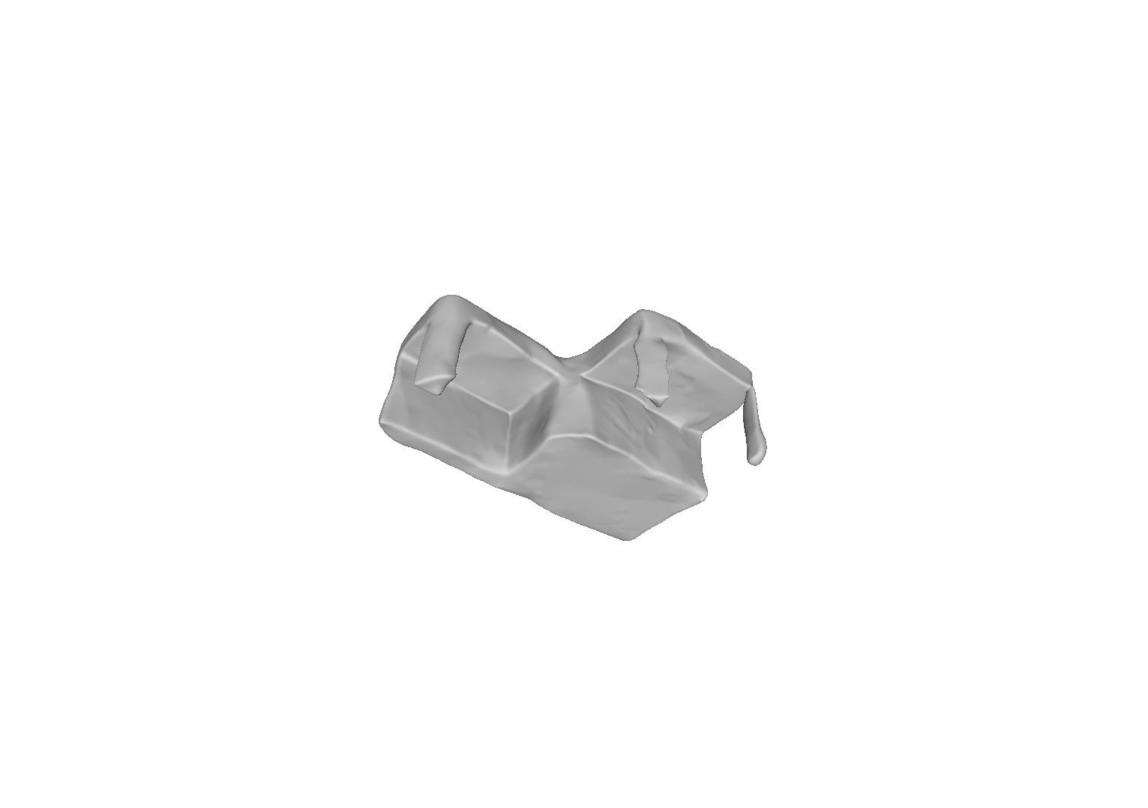}}
\\
\vspace{-5mm}
\subfloat{\includegraphics[width=\fitscale\tgtwidth, trim={450 152 450 152}, clip]{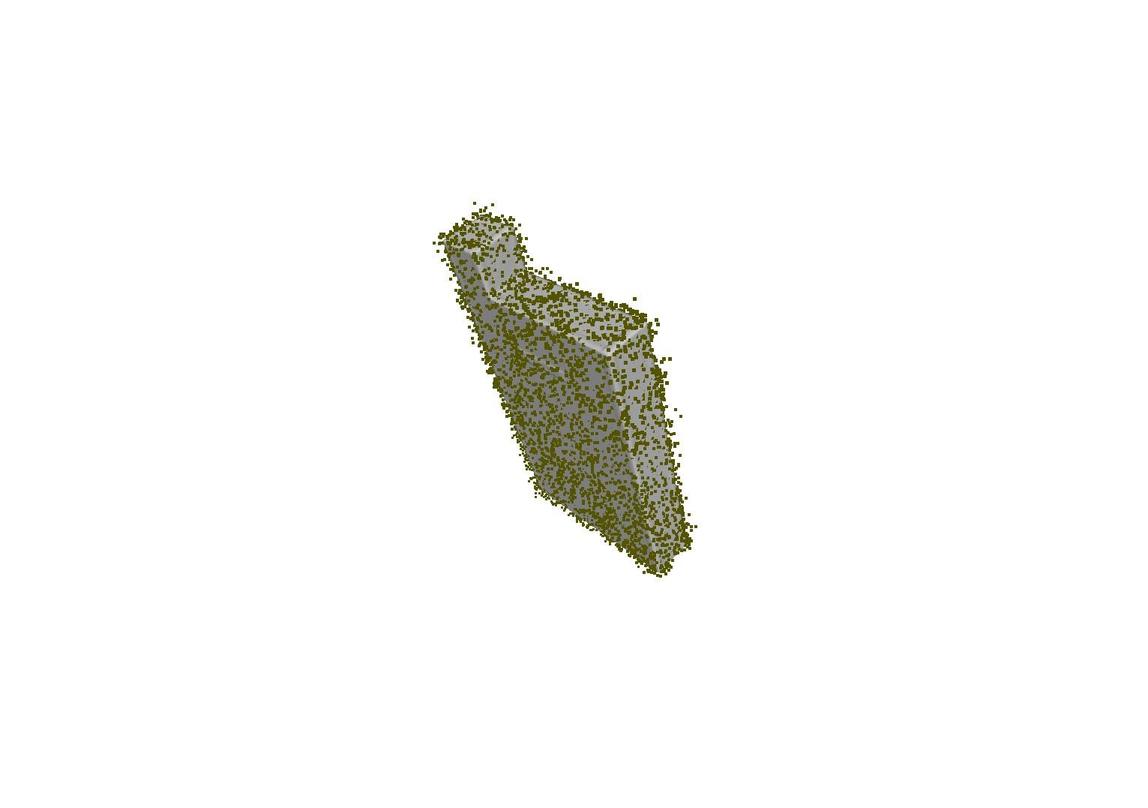}}
\subfloat{\includegraphics[width=\fitscale\tgtwidth, trim={450 152 450 152}, clip]{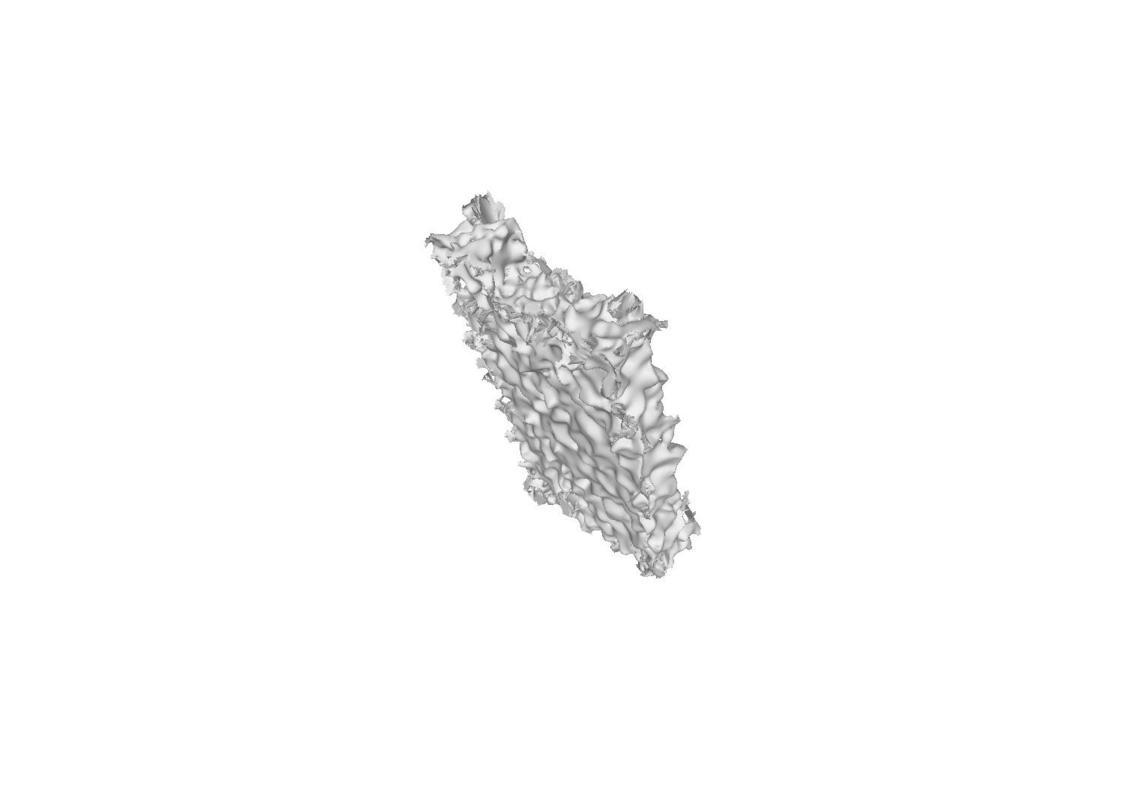}}
\subfloat{\includegraphics[width=\fitscale\tgtwidth, trim={450 152 450 152}, clip]{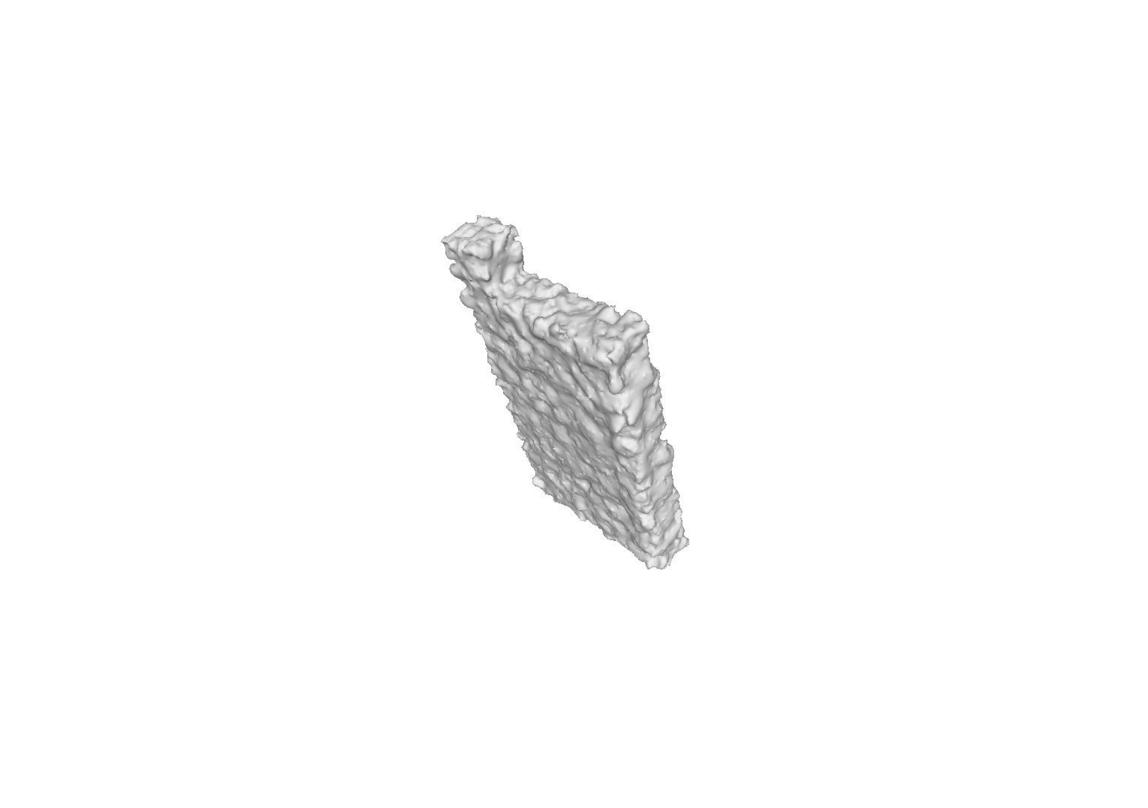}}
\subfloat{\includegraphics[width=\fitscale\tgtwidth, trim={450 152 450 152}, clip]{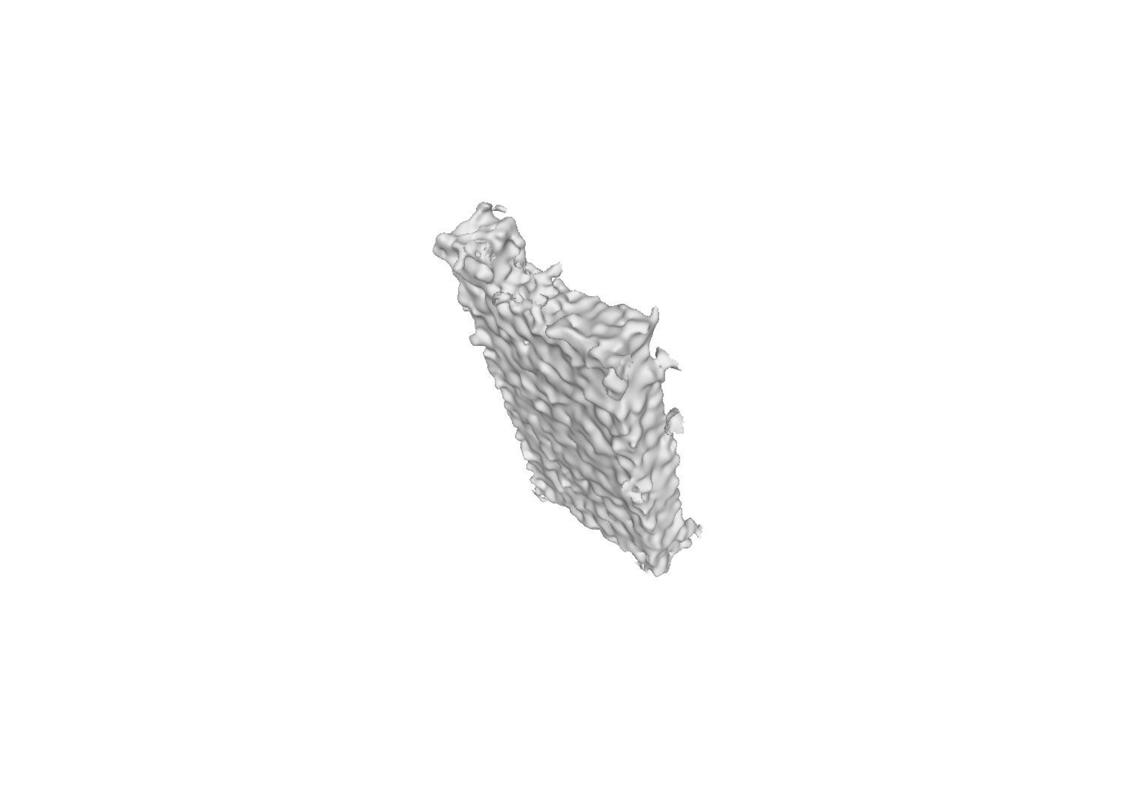}}
\subfloat{\includegraphics[width=\fitscale\tgtwidth, trim={450 152 450 152}, clip]{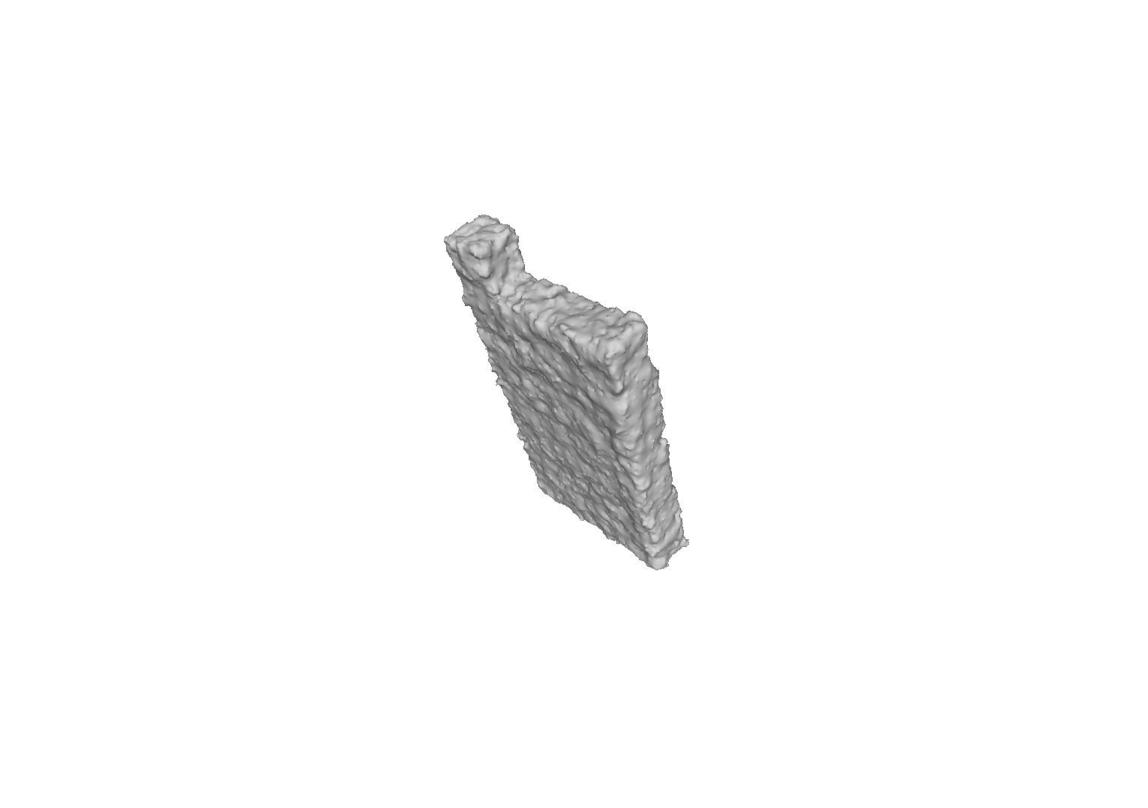}}
\subfloat{\includegraphics[width=\fitscale\tgtwidth, trim={450 152 450 152}, clip]{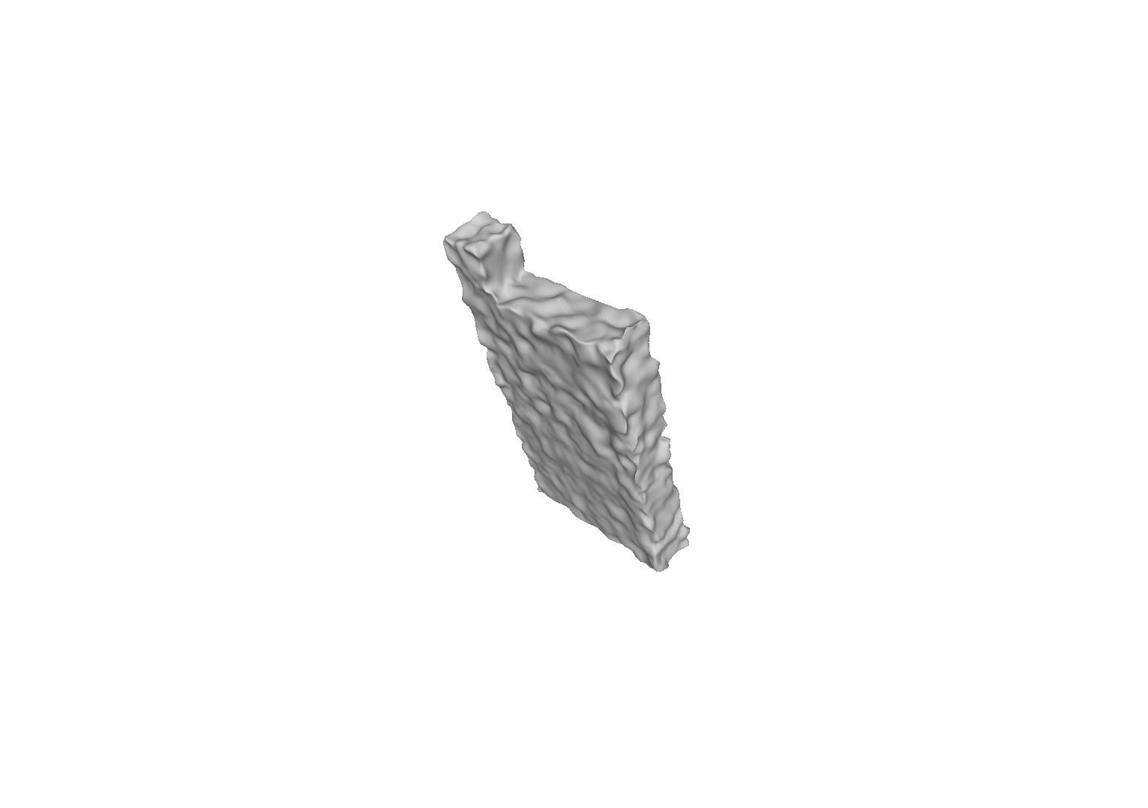}}
\subfloat{\includegraphics[width=\fitscale\tgtwidth, trim={450 152 450 152}, clip]{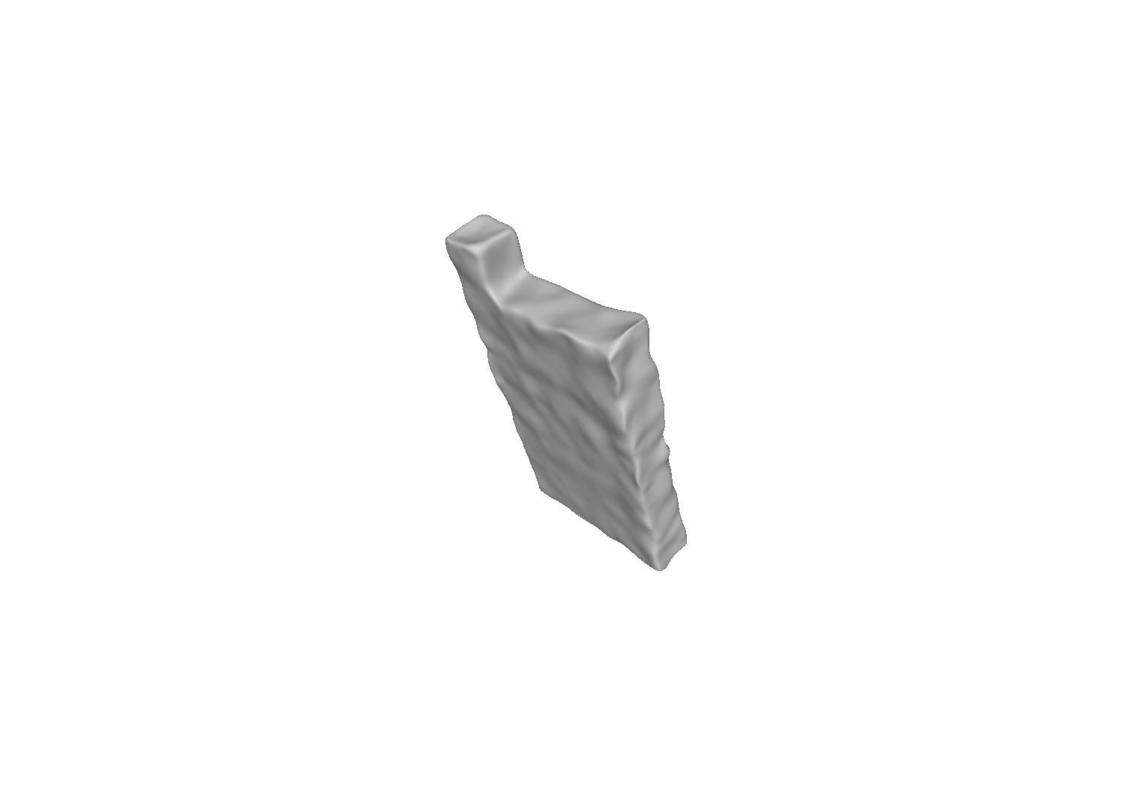}}
\subfloat{\includegraphics[width=\fitscale\tgtwidth, trim={450 152 450 152}, clip]{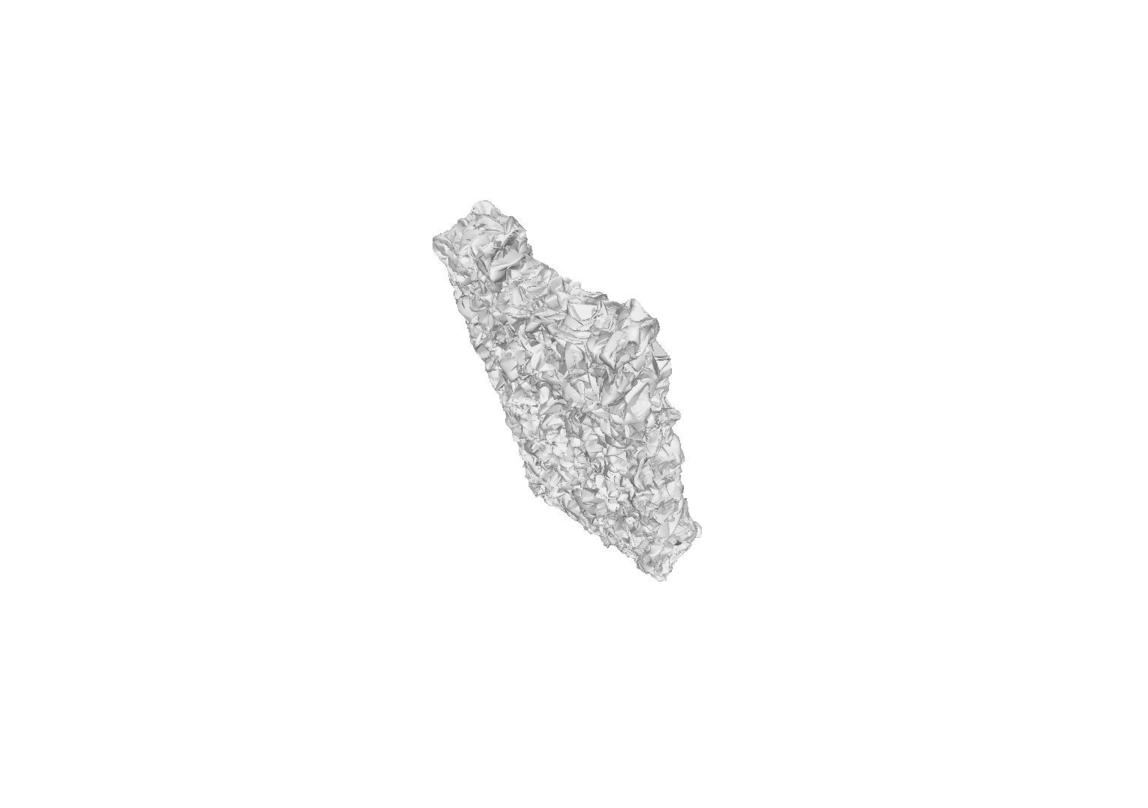}}
\subfloat{\includegraphics[width=\fitscale\tgtwidth, trim={450 152 450 152}, clip]{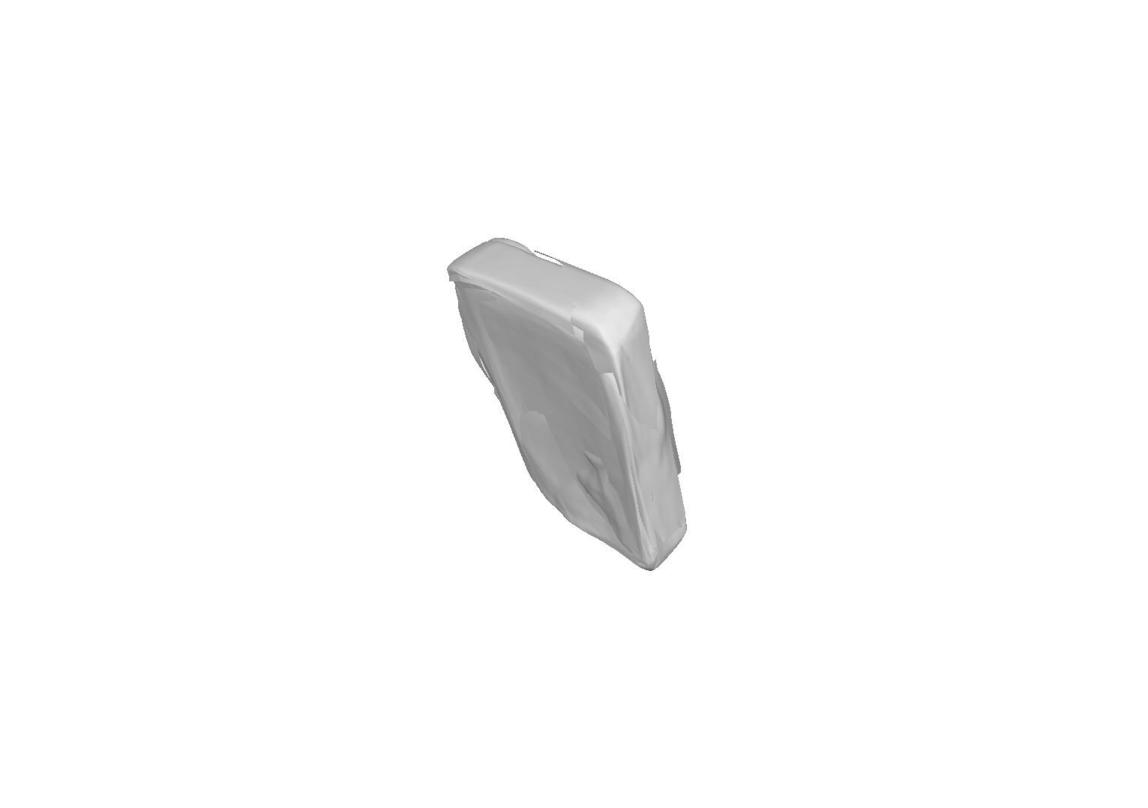}}
\subfloat{\includegraphics[width=\fitscale\tgtwidth, trim={450 152 450 152}, clip]{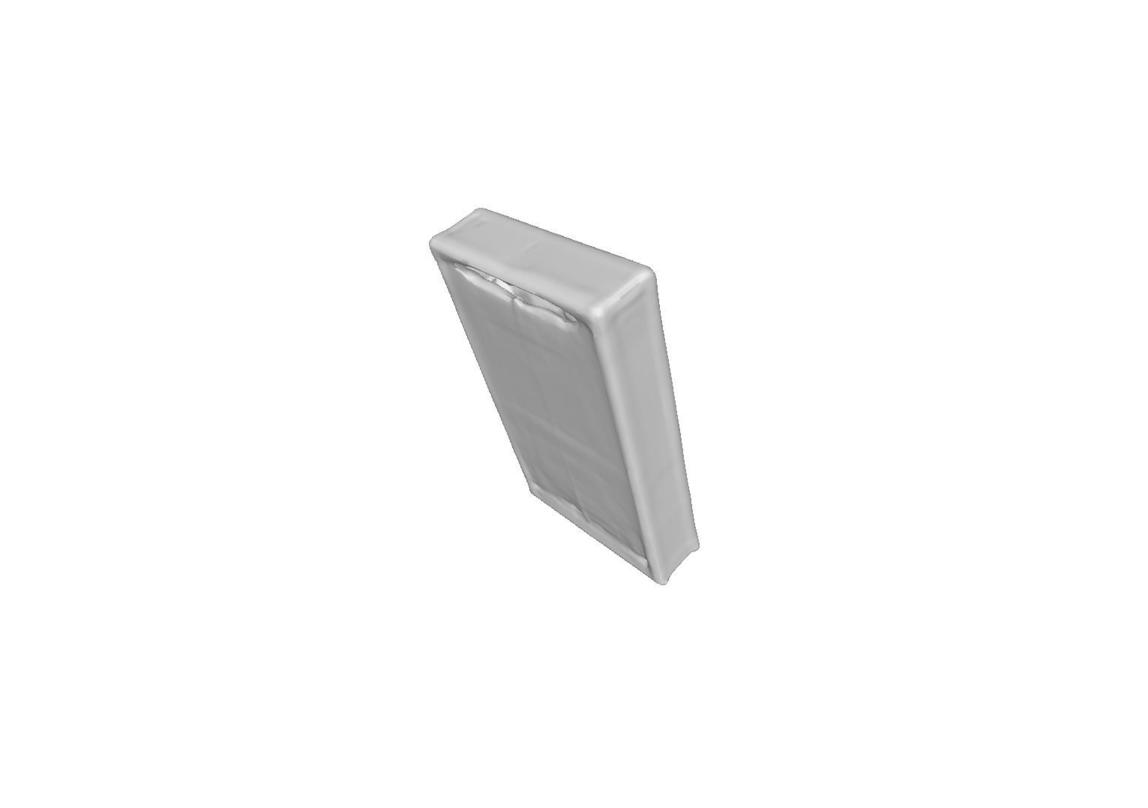}}
~
\subfloat{\includegraphics[width=\fitscale\tgtwidth, trim={450 152 450 152}, clip]{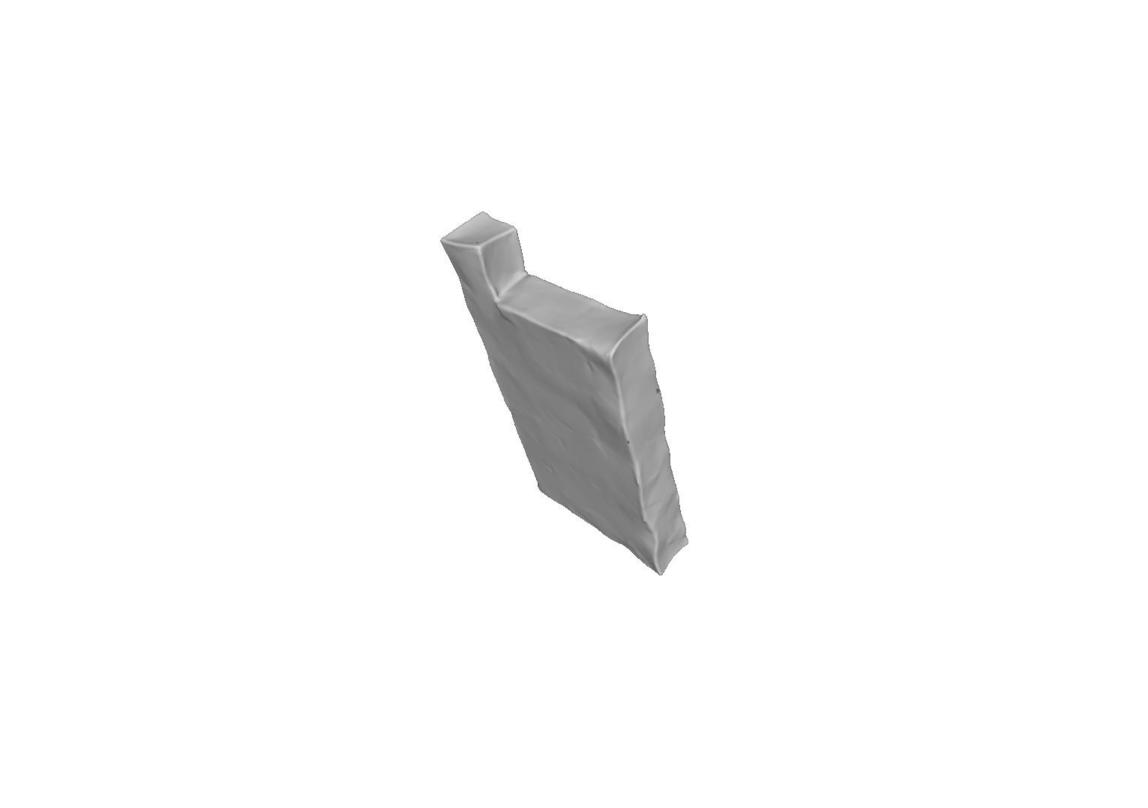}}
\\
\vspace{-5mm}
\subfloat{\includegraphics[width=\fitscale\tgtwidth, trim={450 152 450 152}, clip]{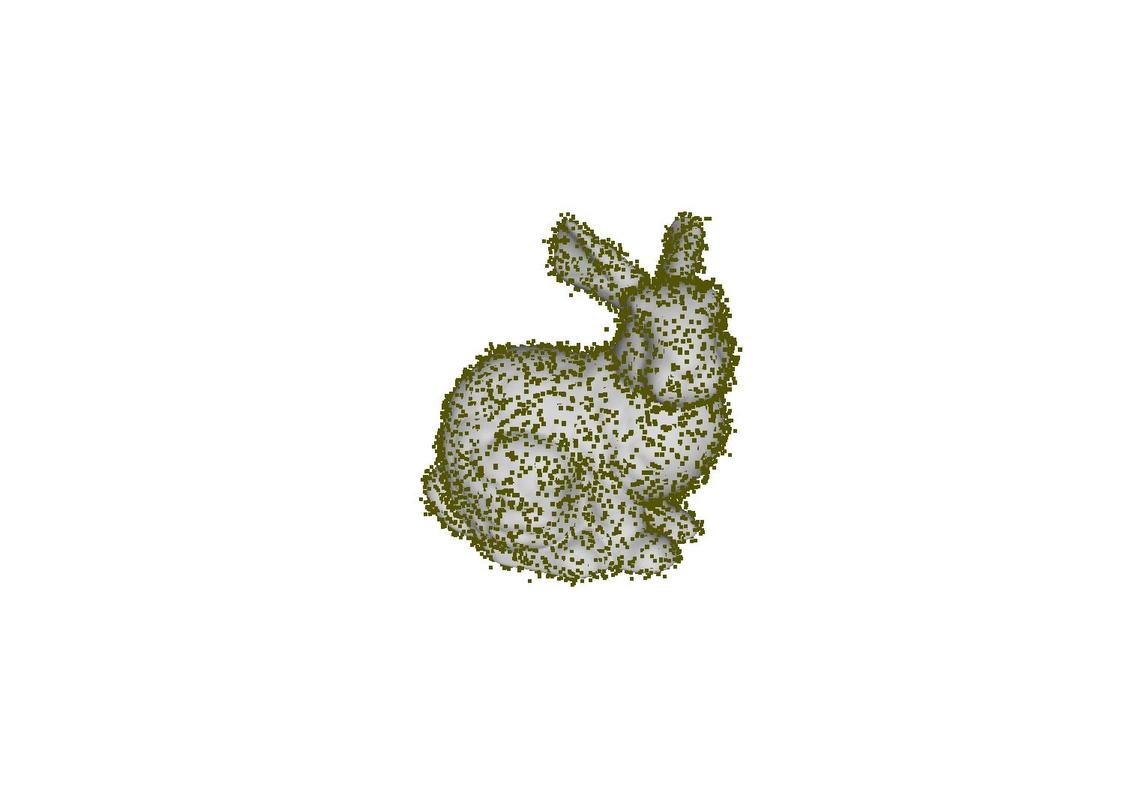}}
\subfloat{\includegraphics[width=\fitscale\tgtwidth, trim={450 152 450 152}, clip]{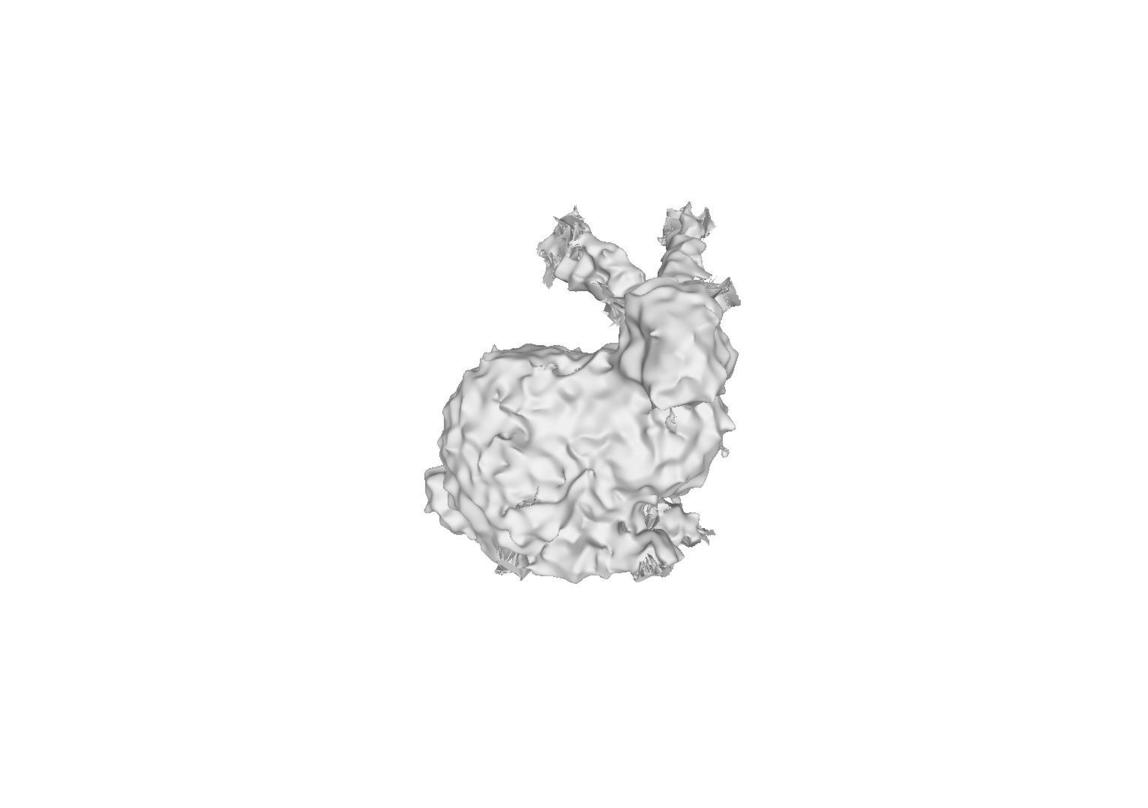}}
\subfloat{\includegraphics[width=\fitscale\tgtwidth, trim={450 152 450 152}, clip]{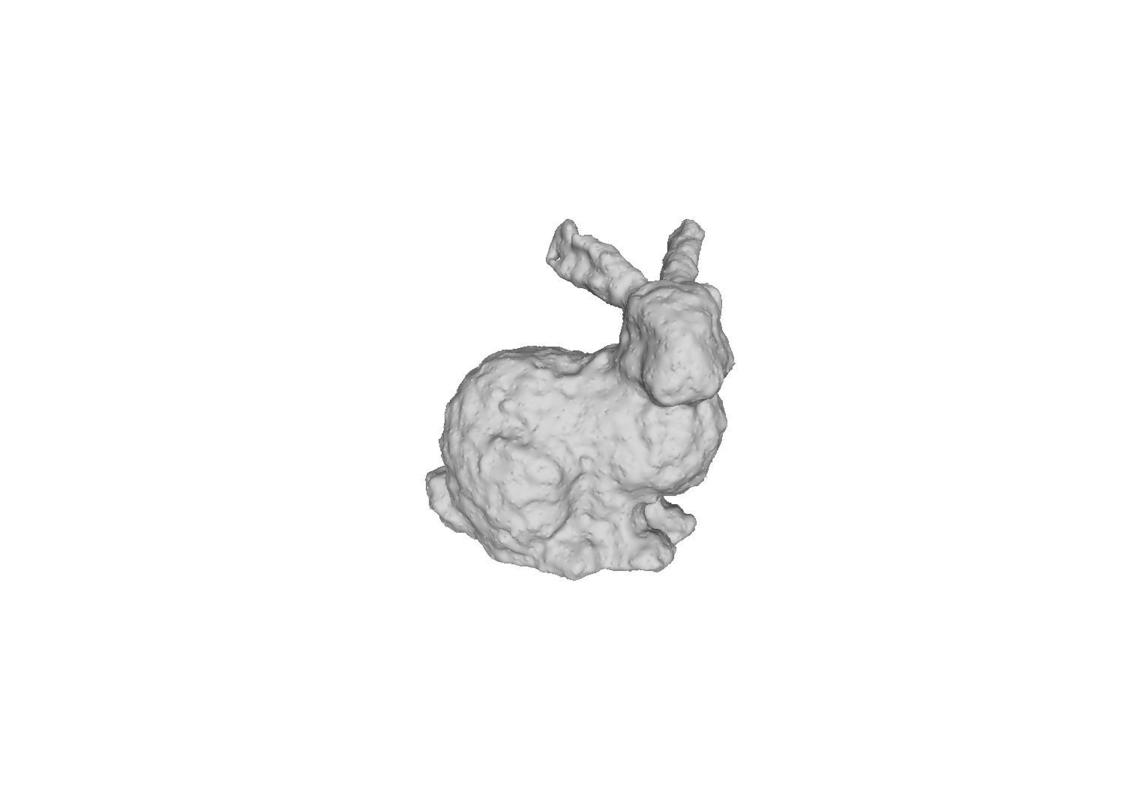}}
\subfloat{\includegraphics[width=\fitscale\tgtwidth, trim={450 152 450 152}, clip]{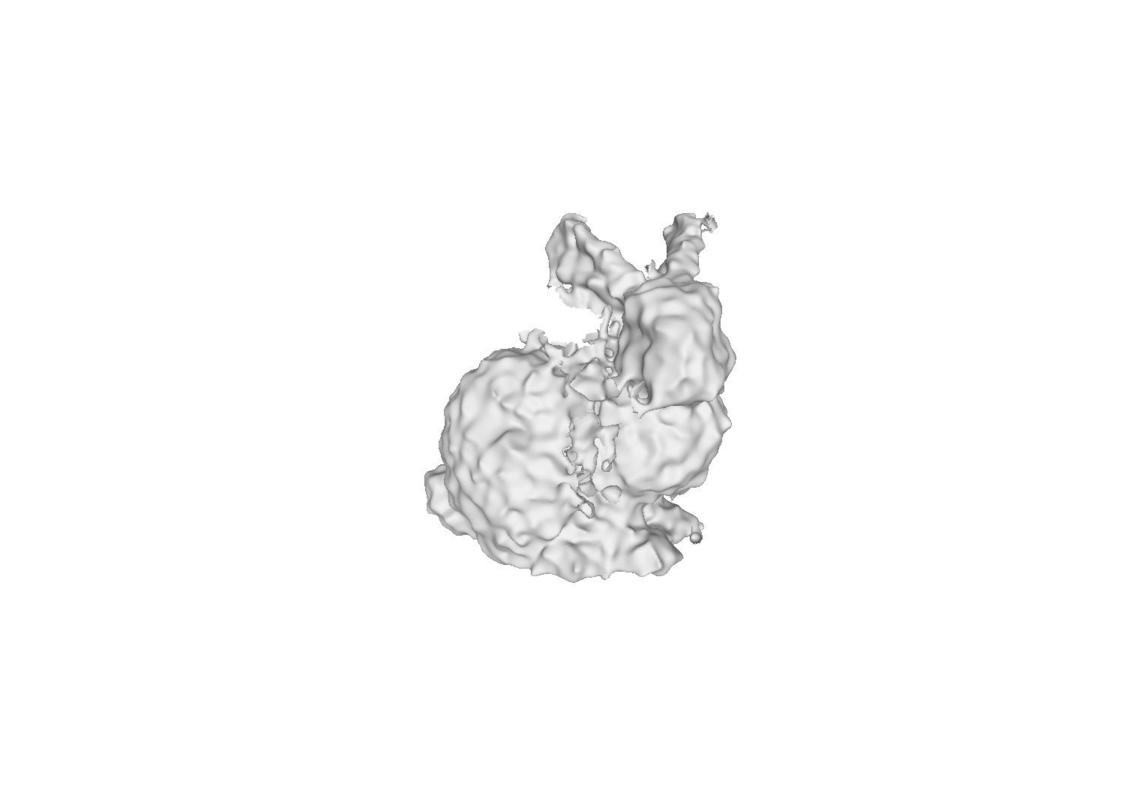}}
\subfloat{\includegraphics[width=\fitscale\tgtwidth, trim={450 152 450 152}, clip]{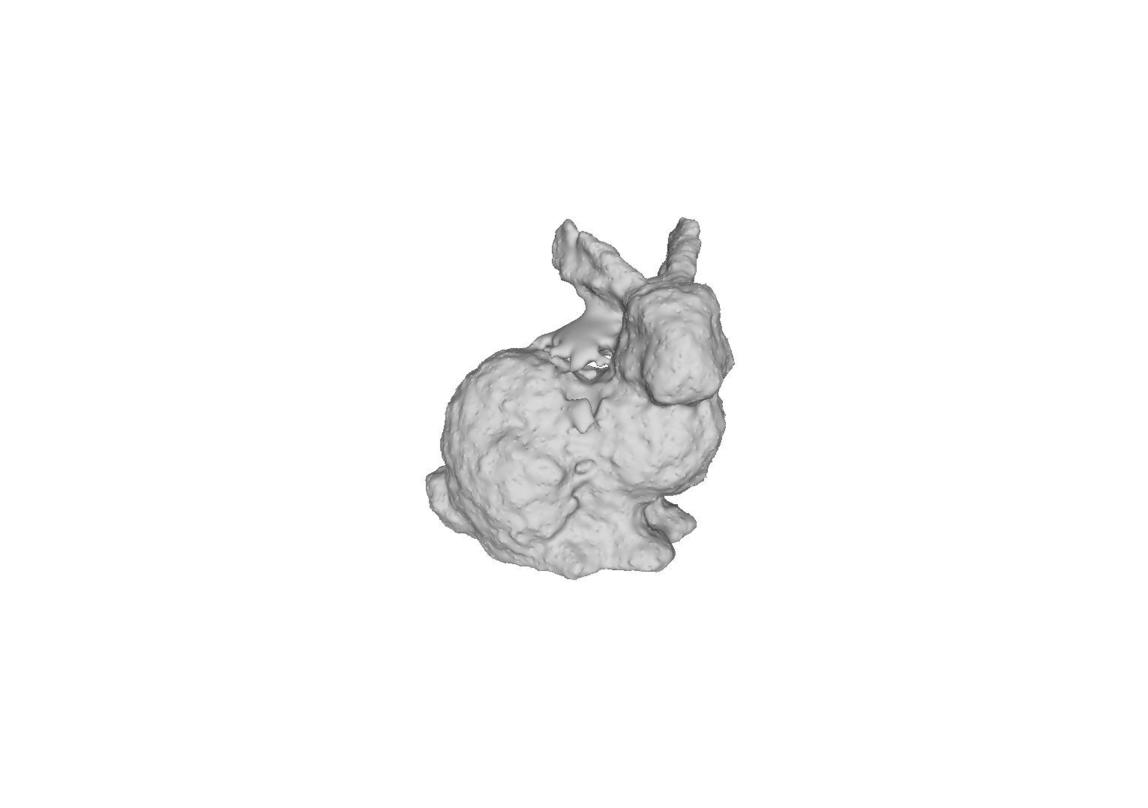}}
\subfloat{\includegraphics[width=\fitscale\tgtwidth, trim={450 152 450 152}, clip]{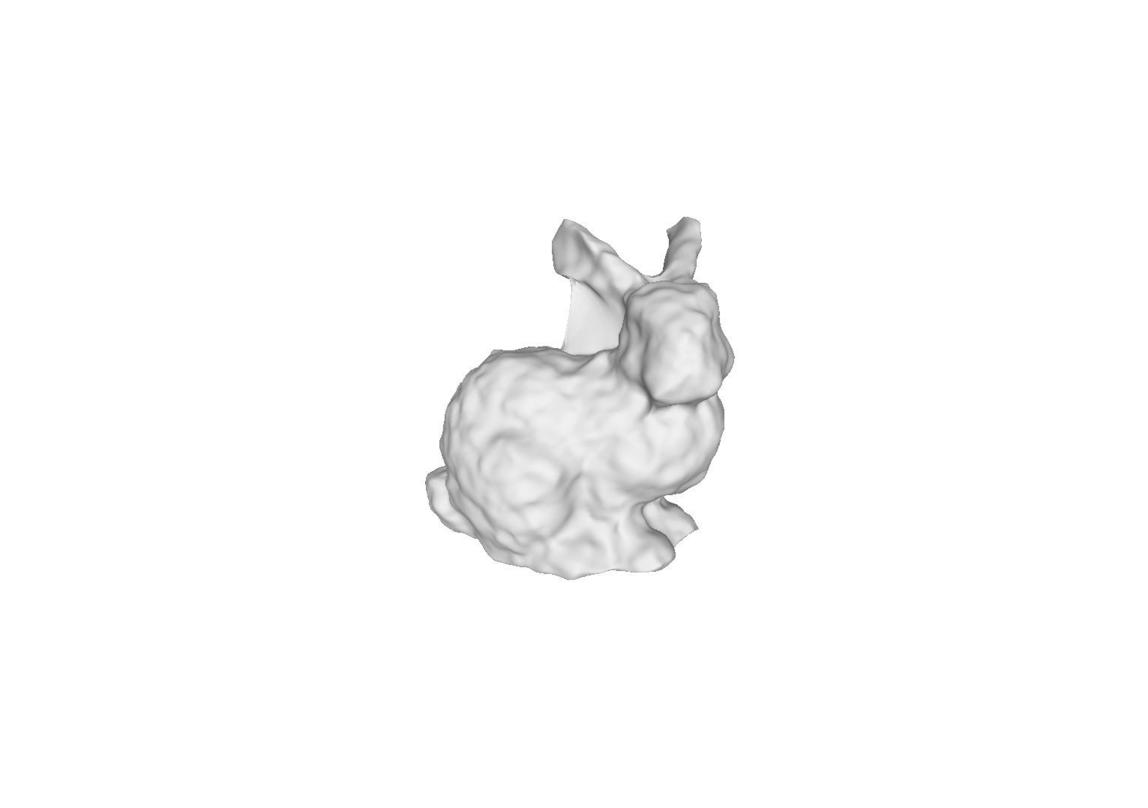}}
\subfloat{\includegraphics[width=\fitscale\tgtwidth, trim={450 152 450 152}, clip]{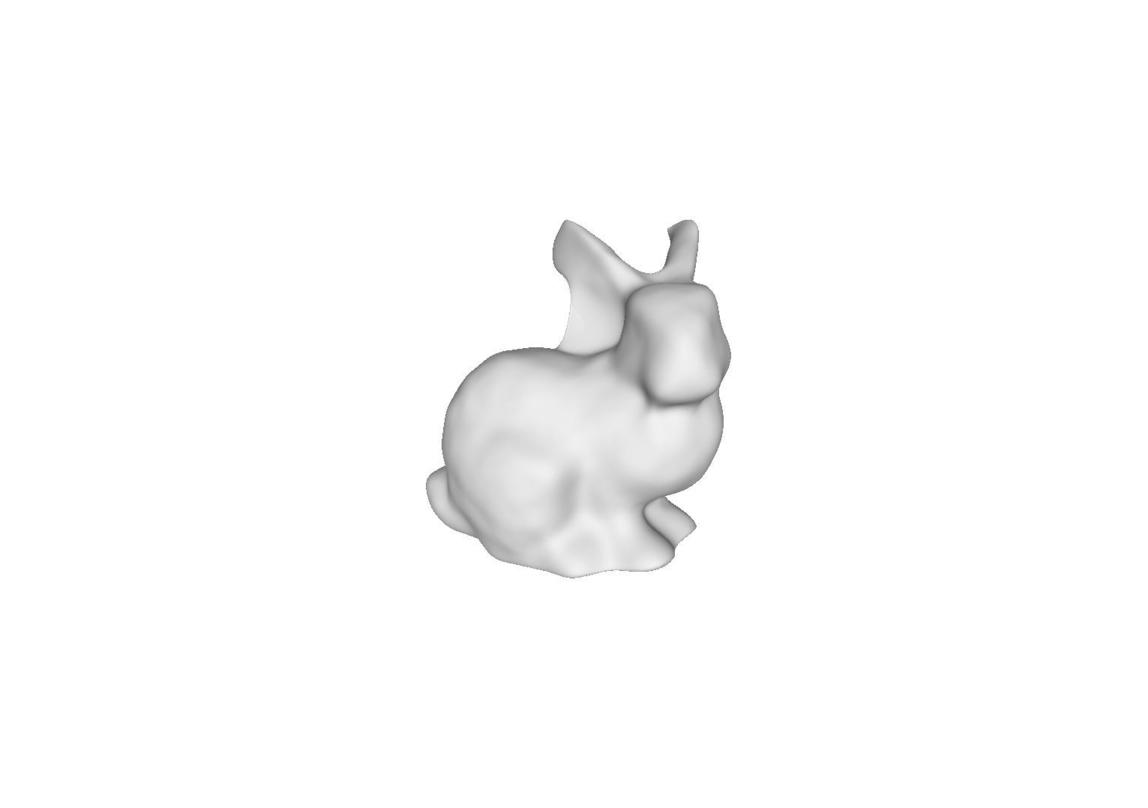}}
\subfloat{\includegraphics[width=\fitscale\tgtwidth, trim={450 152 450 152}, clip]{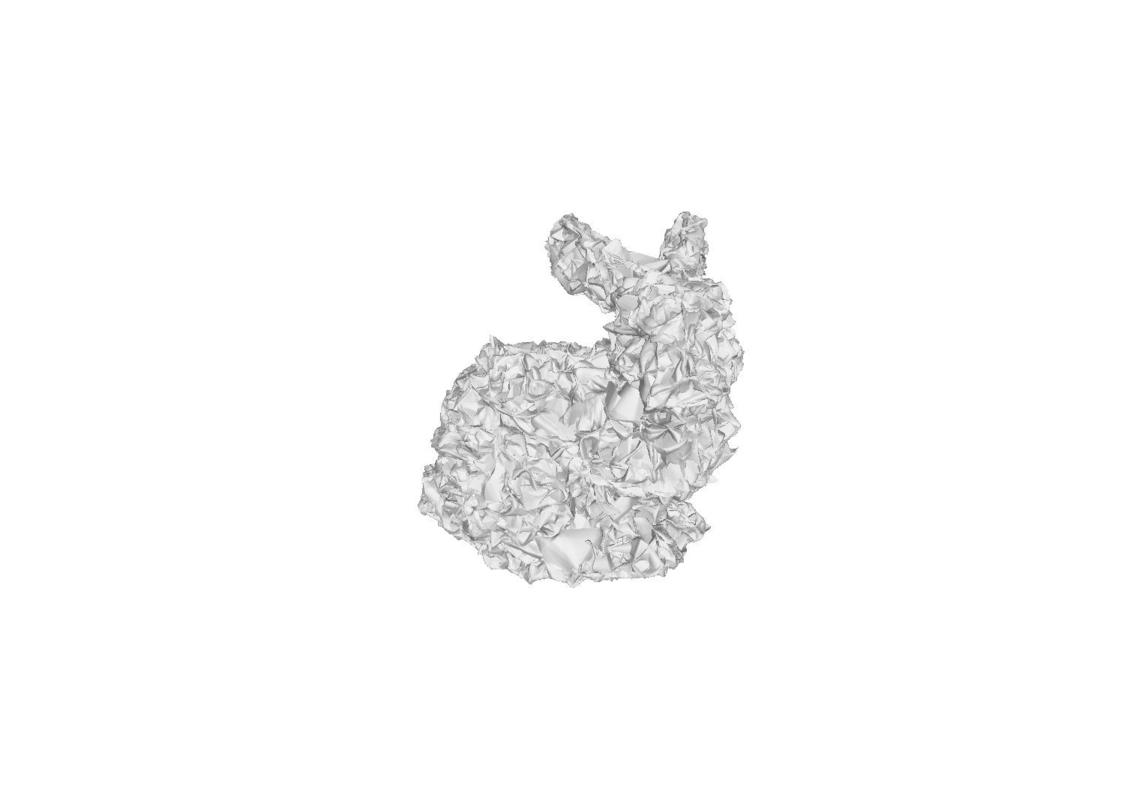}}
\subfloat{\includegraphics[width=\fitscale\tgtwidth, trim={450 152 450 152}, clip]{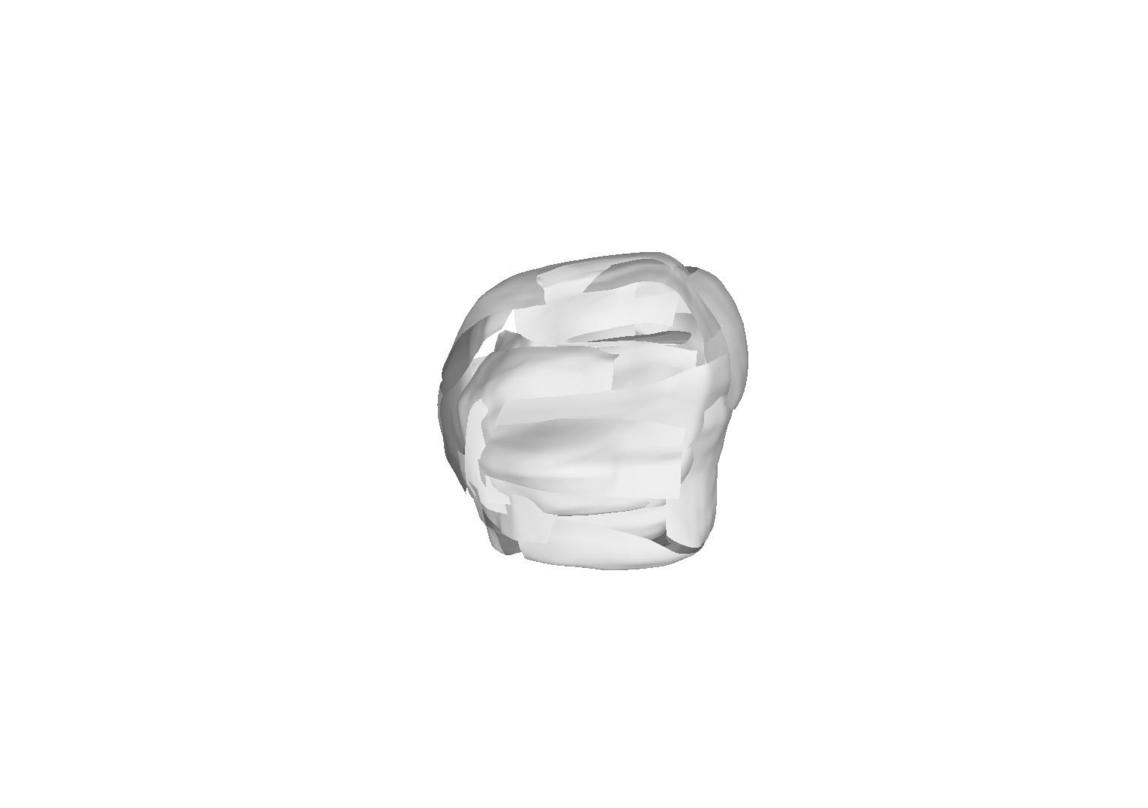}}
\subfloat{\includegraphics[width=\fitscale\tgtwidth, trim={450 152 450 152}, clip]{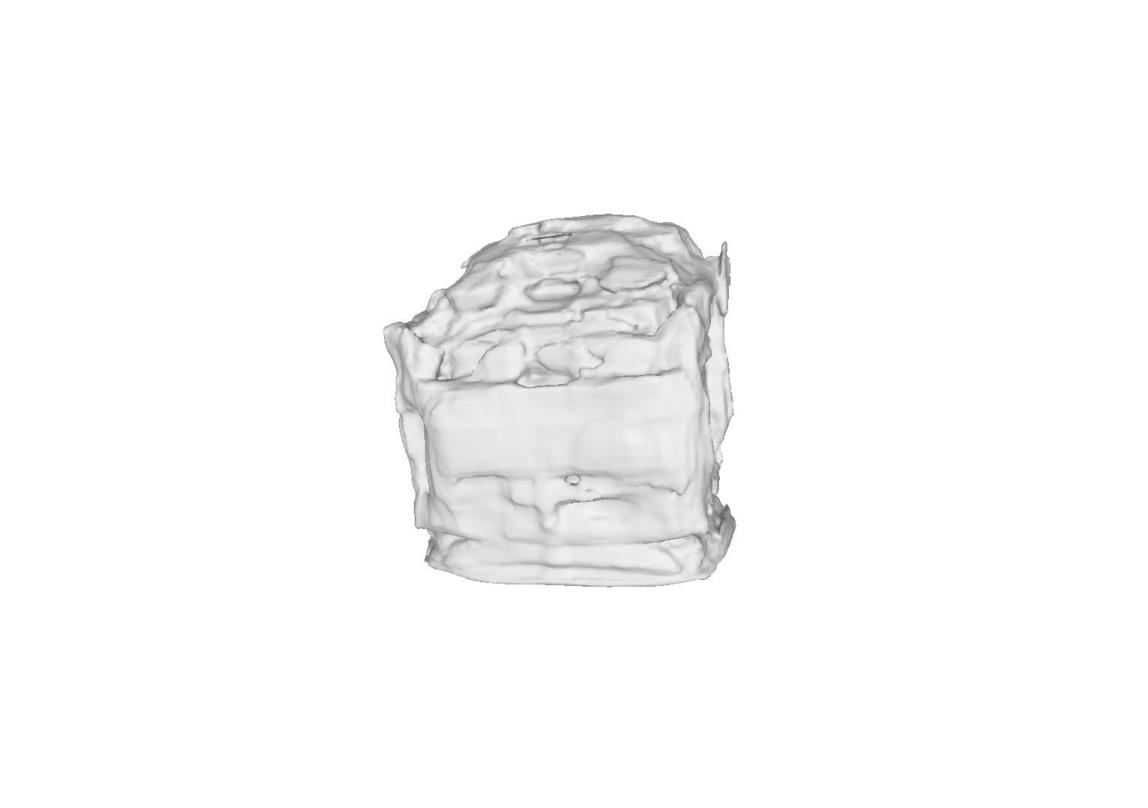}}
~
\subfloat{\includegraphics[width=\fitscale\tgtwidth, trim={450 152 450 152}, clip]{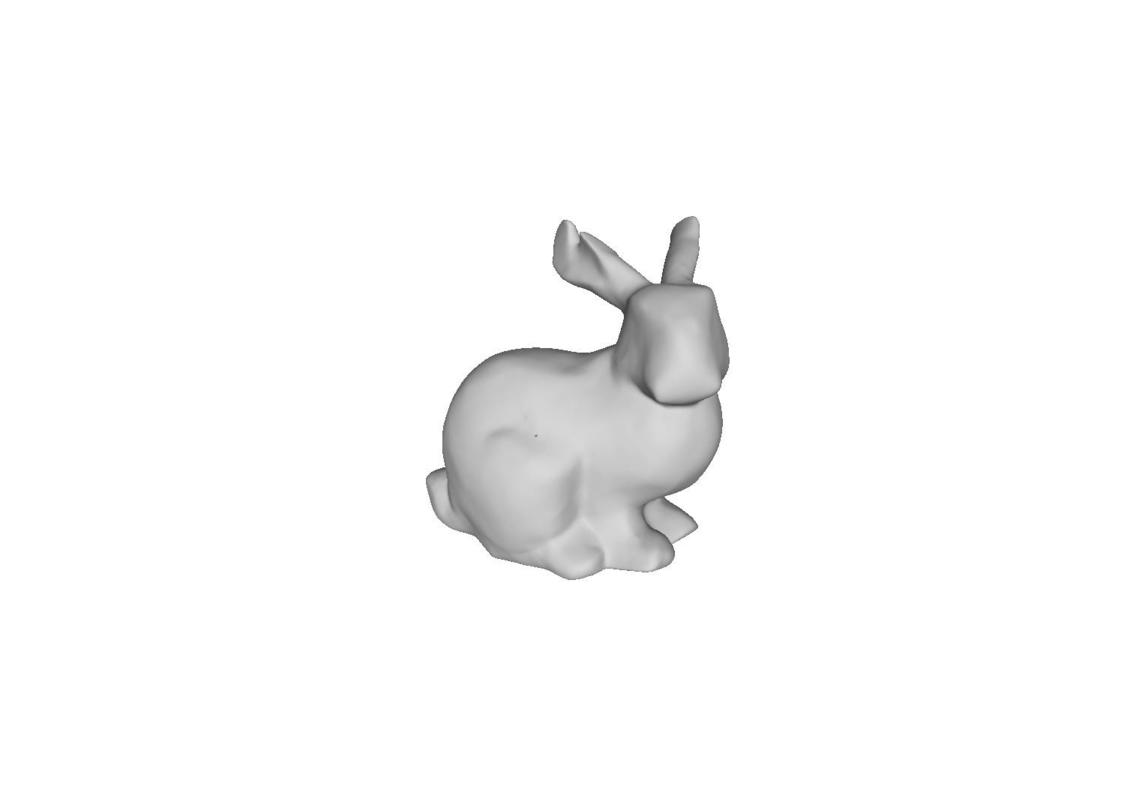}}
\\
\vspace{-5mm}
\subfloat[GT+PC]{\includegraphics[width=\fitscale\tgtwidth, trim={319 152 323 202}, clip]{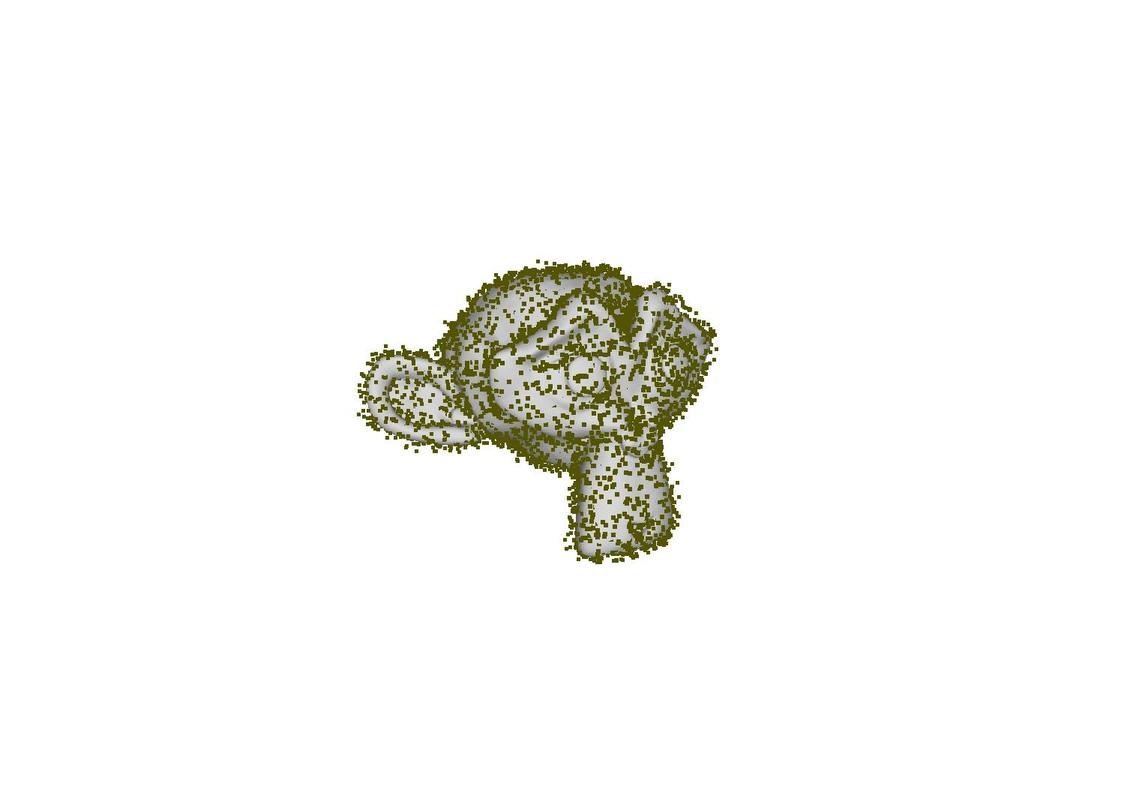}}
\subfloat[\cite{meshlab}+\cite{oztireli2009feature}]{\includegraphics[width=\fitscale\tgtwidth, trim={319 152 323 202}, clip]{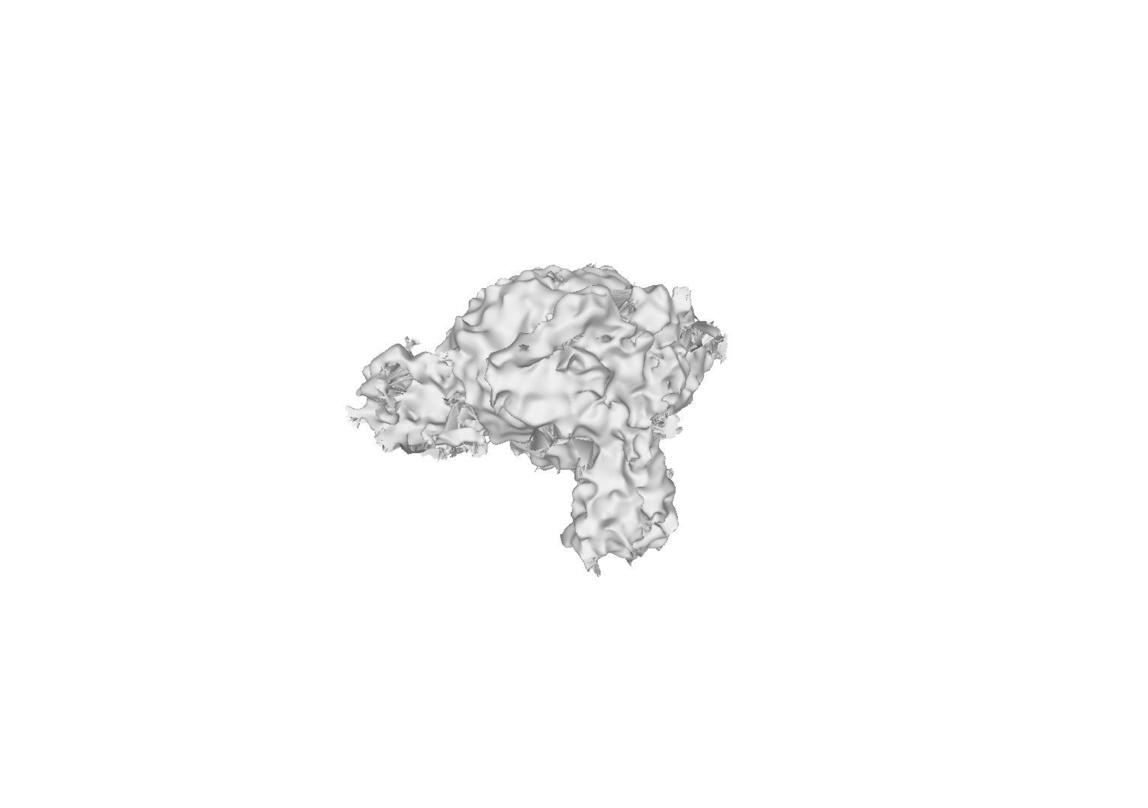}}
\subfloat[\cite{meshlab}+\cite{screenedpoisson}]{\includegraphics[width=\fitscale\tgtwidth, trim={319 152 323 202}, clip]{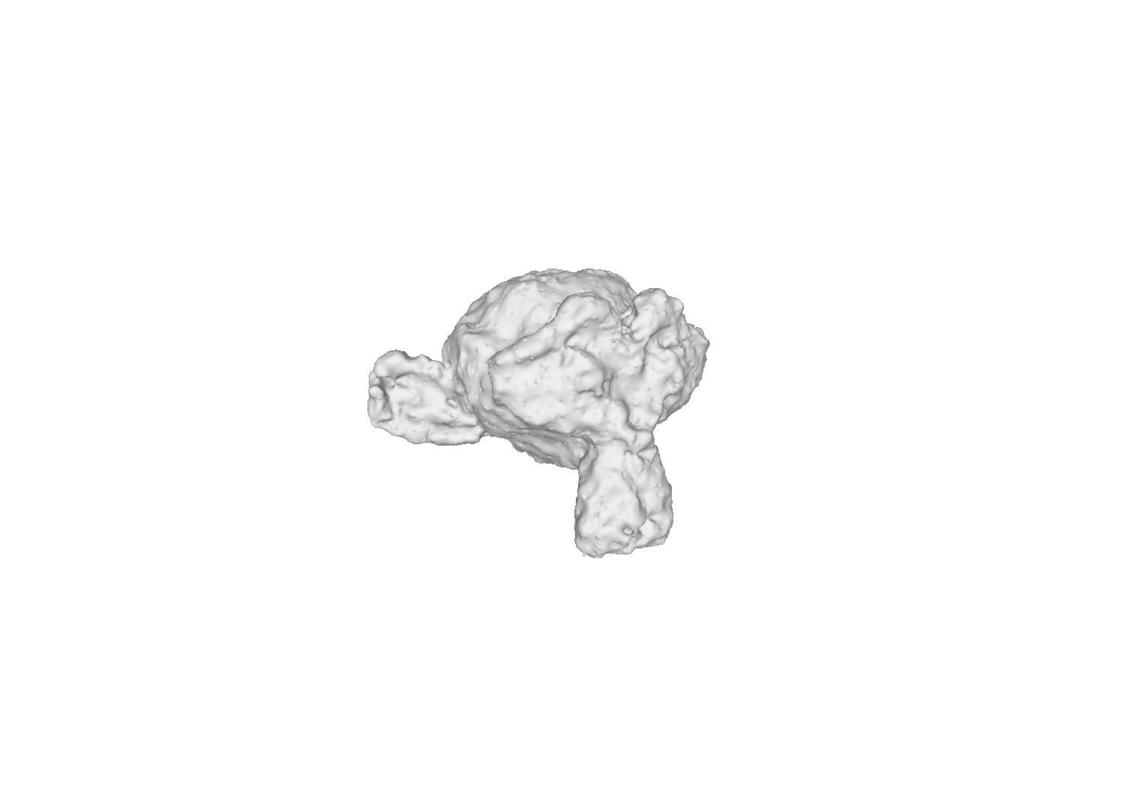}}
\subfloat[\cite{pcpnet}+\cite{oztireli2009feature}]{\includegraphics[width=\fitscale\tgtwidth, trim={319 152 323 202}, clip]{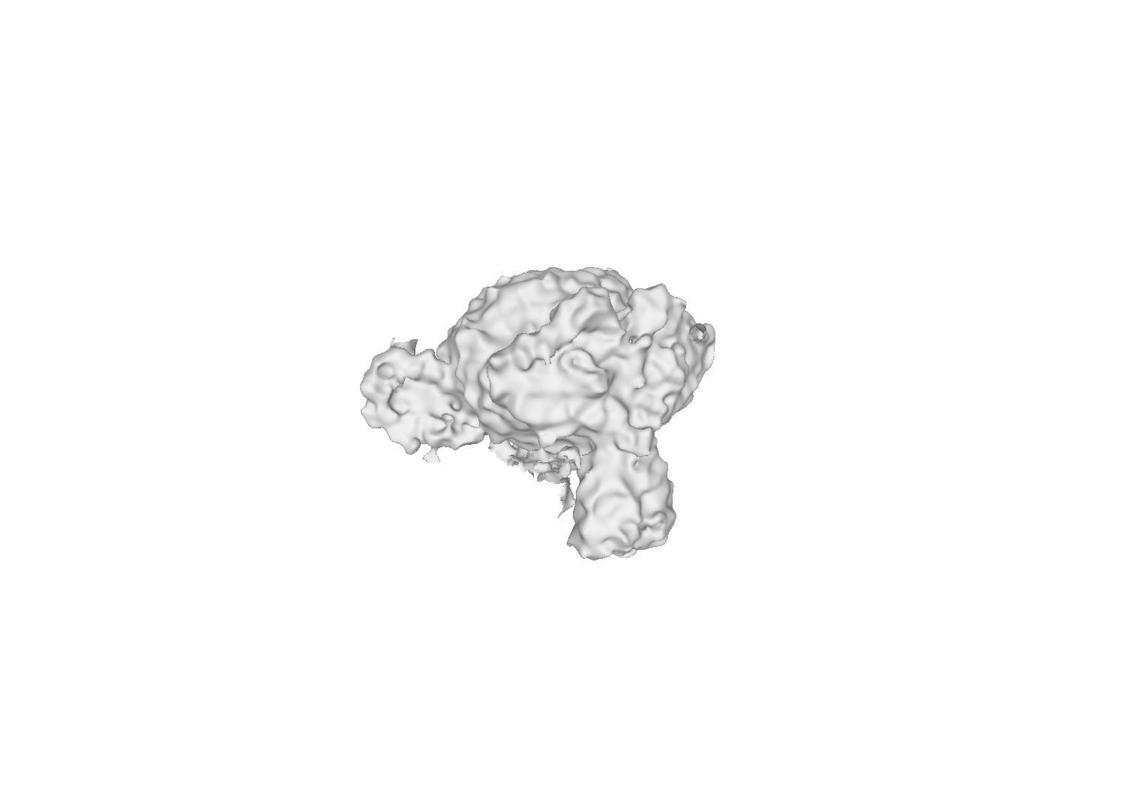}}
\subfloat[\cite{pcpnet}+\cite{screenedpoisson}]{\includegraphics[width=\fitscale\tgtwidth, trim={319 152 323 202}, clip]{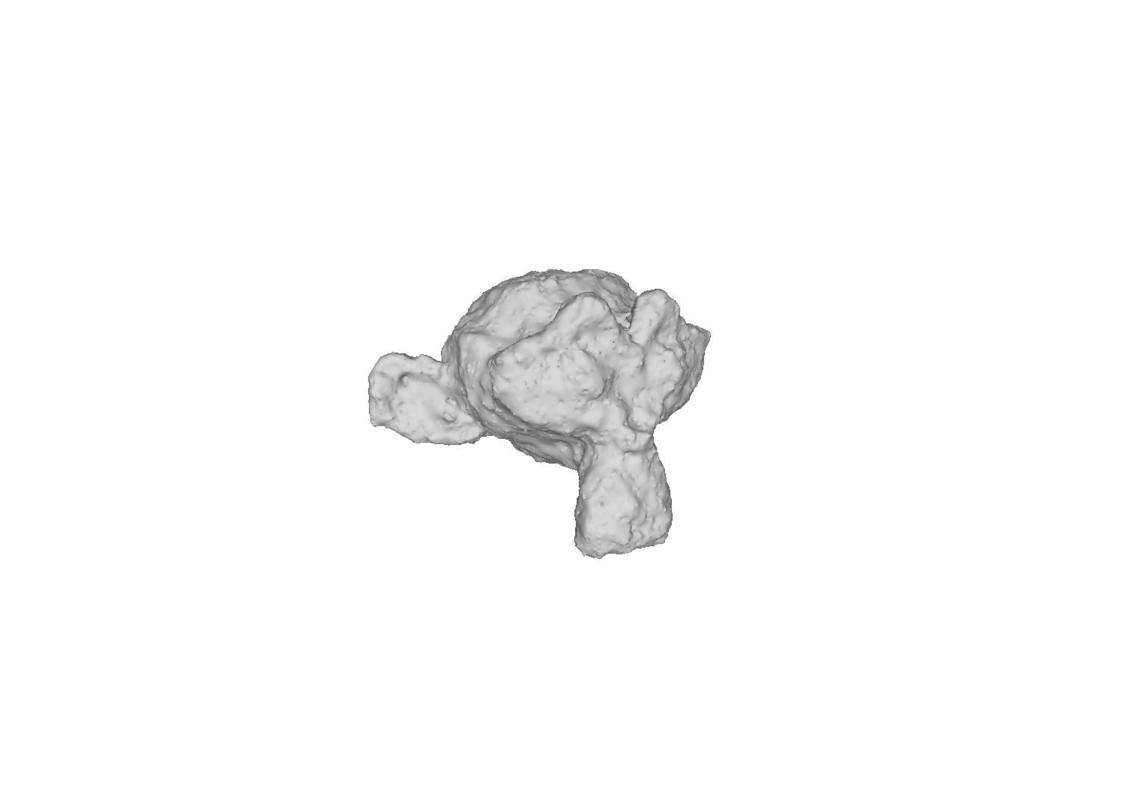}}
\subfloat[Lap-low]{\includegraphics[width=\fitscale\tgtwidth, trim={319 152 323 202}, clip]{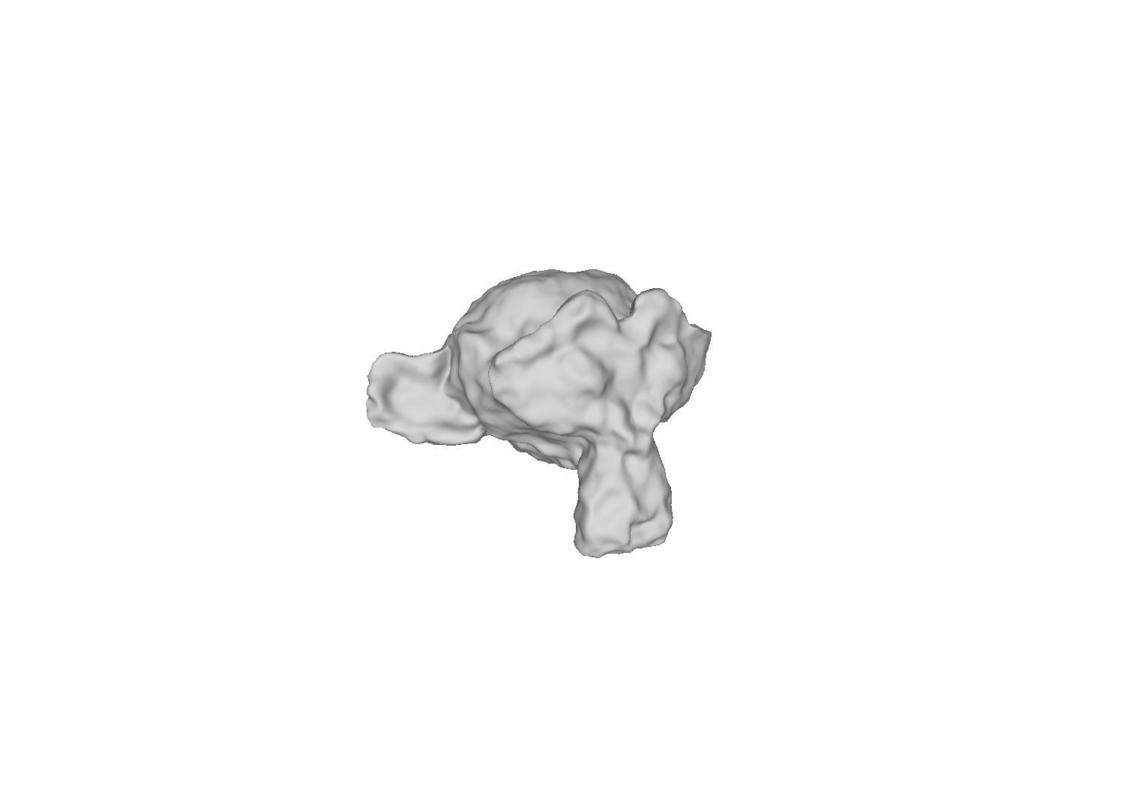}}
\subfloat[Lap-high]{\includegraphics[width=\fitscale\tgtwidth, trim={319 152 323 202}, clip]{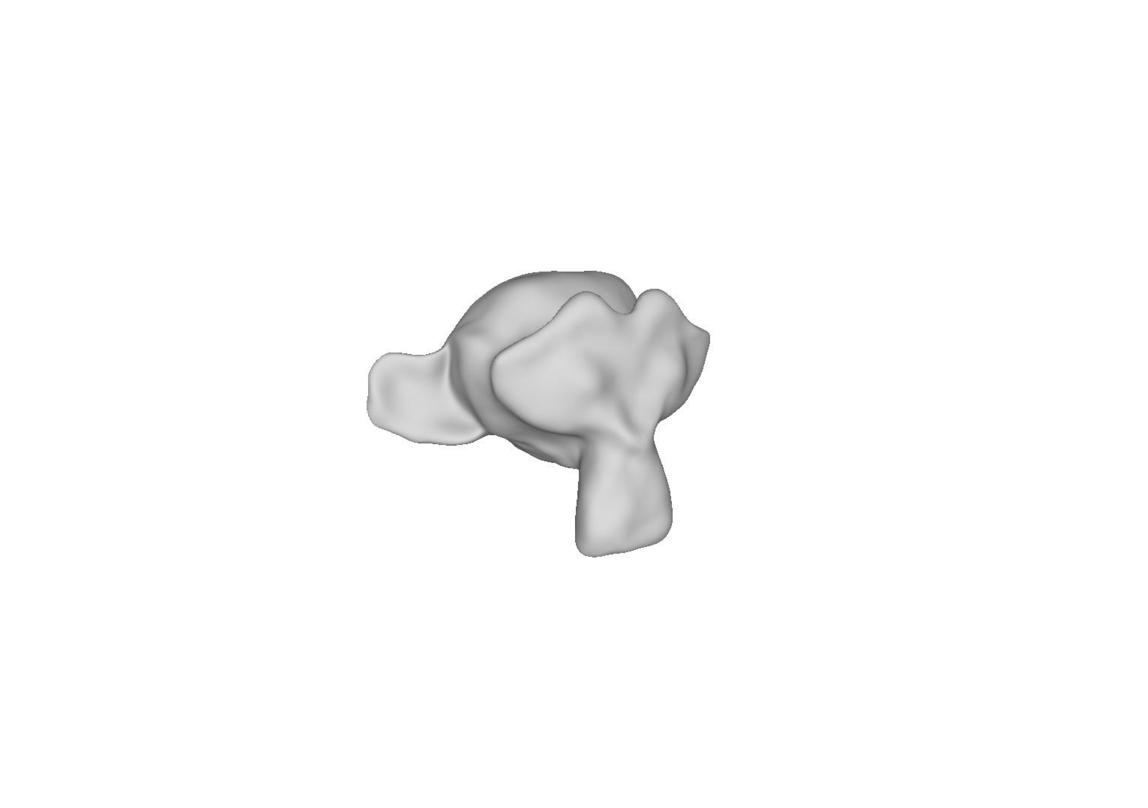}}
\subfloat[DGP]{\includegraphics[width=\fitscale\tgtwidth, trim={319 152 323 202}, clip]{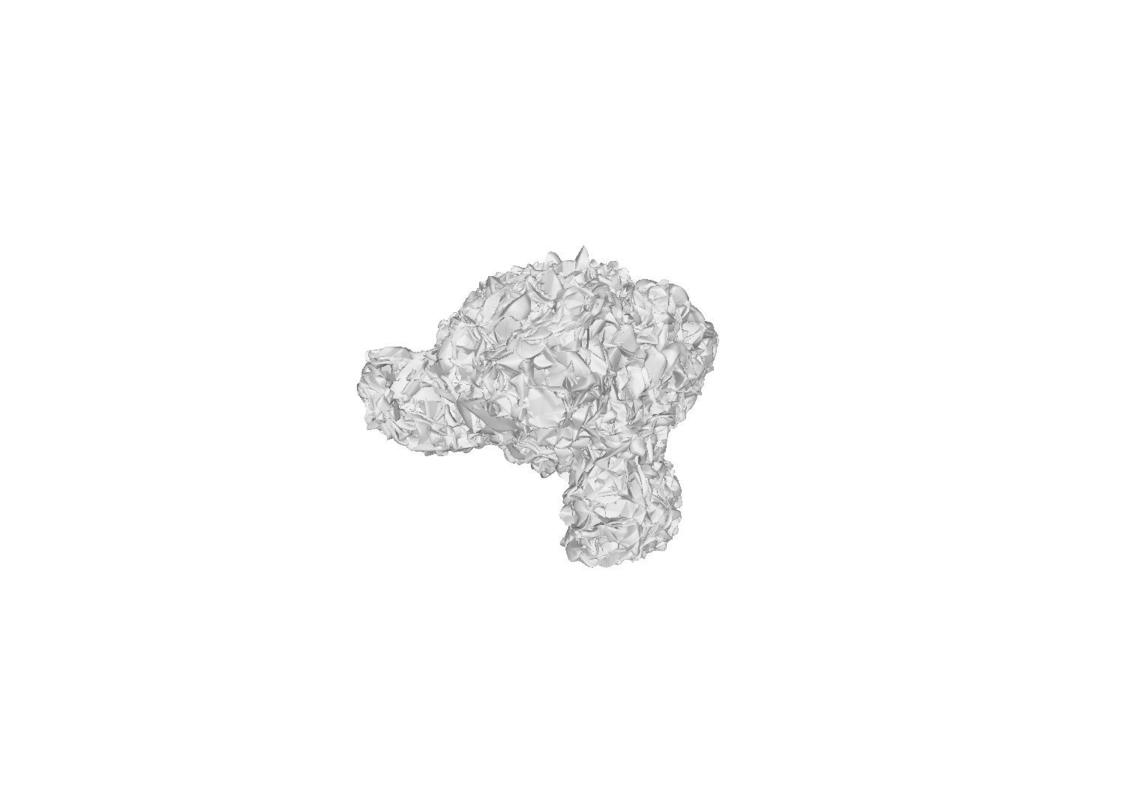}}
\subfloat[AtlasNet]{\includegraphics[width=\fitscale\tgtwidth, trim={319 152 323 202}, clip]{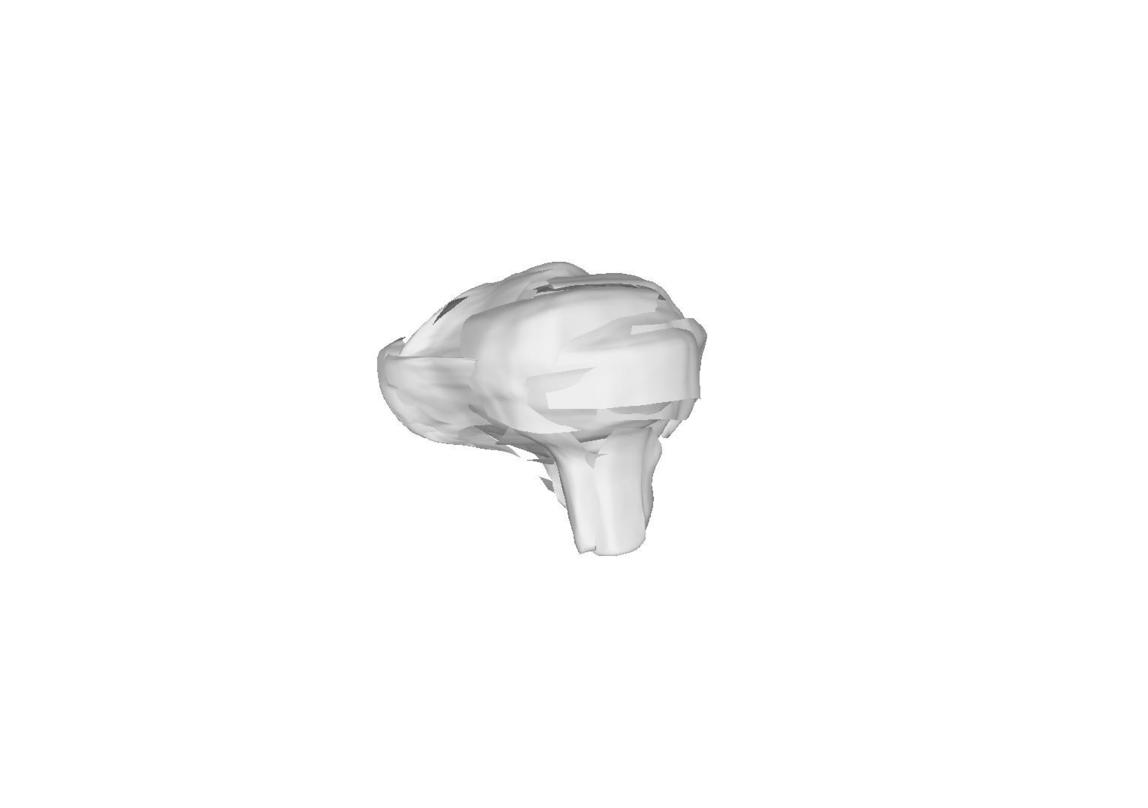}}
\subfloat[OccNet]{\includegraphics[width=\fitscale\tgtwidth, trim={319 152 323 202}, clip]{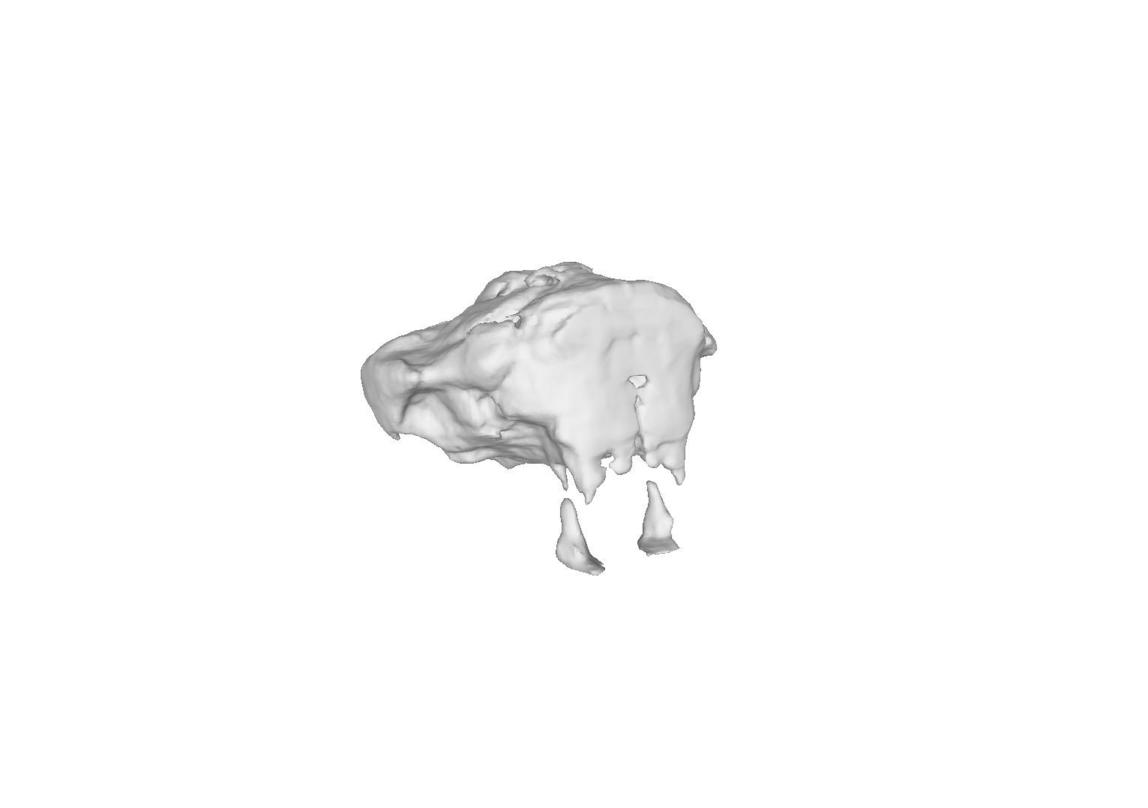}}
~
\subfloat[{\bf Ours}]{\includegraphics[width=\fitscale\tgtwidth, trim={319 152 323 202}, clip]{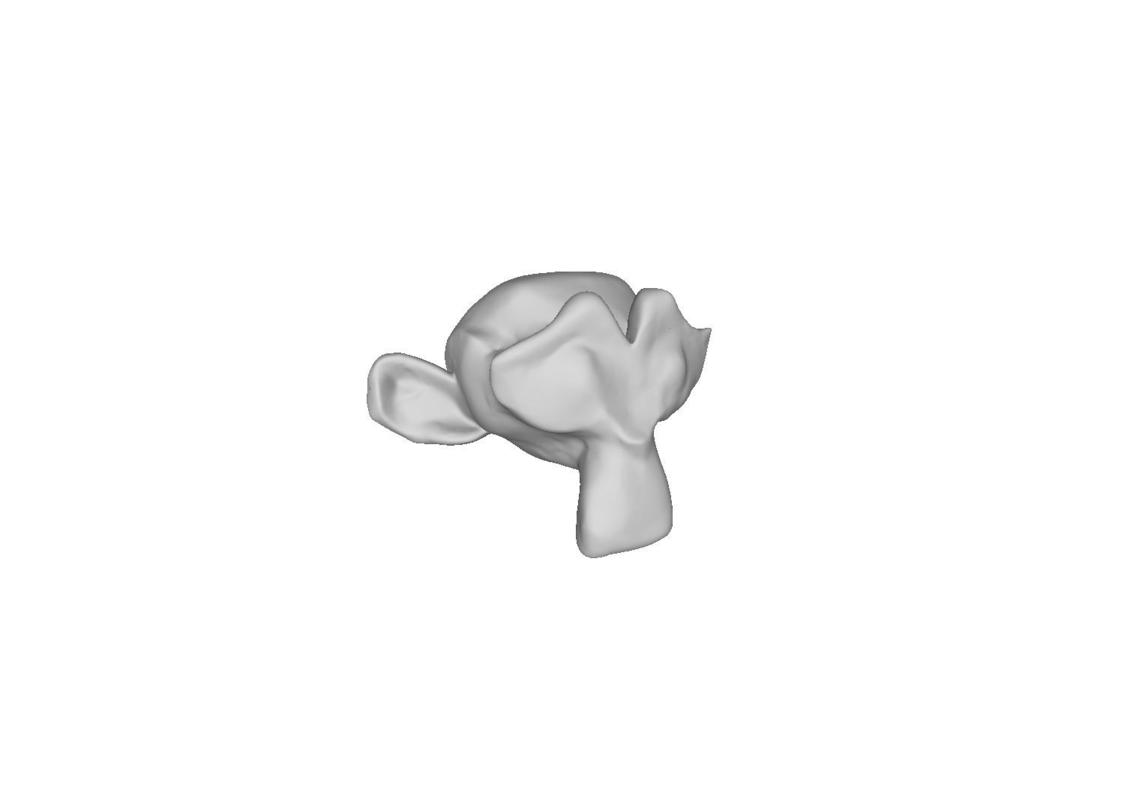}}

%% file: meshlets_cvpr20_supplementary_additional_algo_details.tex

The pseudo code for our optimization procedure to estimate a watertight mesh while enforcing meshlets priors is explained in Algorithm~\ref{algo:meshoptim}.
\\

In our optimization procedure we update both the meshlets and the auxiliary mesh. 
While meshlets priors make the updates of meshlets stable, a prior is needed while updating mesh to ensure that the mesh is watertight and vertices are uniformly distributed. 
Use of smoothness priors or other priors for mesh would hinder our ability to reconstruct sharp features. 
Hence we use Screened Poisson Reconstruction~\cite{screenedpoisson} at the end of every iteration and use the vertices and normals of globally consistent meshlets to update the mesh.

\SetKwFor{For}{for (}{) $\lbrace$}{$\rbrace$}

\begin{algorithm*}
    \SetNoFillComment
    \DontPrintSemicolon
    \KwIn{Sparse and noisy point cloud ${PC}$}
    \KwOut{Watertight mesh $\mathcal{M}$}
    \tcc{mesh and meshlets initialization}
    mesh, $\mathcal{M}(t_0)$, initialization (Section 4.1)\;
    meshlets, $\{m_i(t_0)\}_{i=1:N}$, initialization by meshlets resampling (Section 4.1)\;
    \tcc{outer loop}
    \While{$\mathcal{M} \text{converges\ to}\ {PC}$}{
        \tcc{inner loop}
        \For{$n\gets0$ \KwTo $20$}{
            \tcc{Enforce Local Priors, Section 3.2.1}
            Update $N$ meshlets $\{m_i\}_{i=1:N}$ with respect to the point cloud\;

            \tcc{Enforce Global Consistency, Section 3.2.2}
            \While{\text{global consistency not achieved}}
            {
                Update $N$ meshlets $\{m_i\}_{i=1:N}$ with respect to the $\mathcal{M}$\;
                Update $\mathcal{M}$ with respect to $N$ meshlets $\{m_i\}_{i=1:N}$\;
            }
            \tcc{Re-meshing, Section 4.1}
            Estimate $\mathcal{M}$ from meshlets vertices and normals using Screened Poisson Reconstruction~\cite{screenedpoisson}
        }
        \tcc{Ensure proper coverage of meshlets on mesh}
        Meshlets resampling (Section 4.1) on current mesh $\mathcal{M}$\;
    }
    \caption{Optimization procedure to estimate a watertight mesh while enforcing meshlets priors}
    \label{algo:meshoptim}
\end{algorithm*}